%% file: main.tex
\documentclass[PhD,plaintoc, oldfontcommands]{msu-thesis}

\usepackage[T1]{fontenc}
\usepackage{newtxtext,newtxmath} 
\usepackage{url}
\input{packages}

\title{Generalizing Monocular 3D Object Detection}
\author{Abhinav Kumar}
\fieldofstudy{Computer Science} 
\date{2025}

\dedication{Dedicated to my country India.}

\begin{document}

\frontmatter
\maketitlepage

\begin{abstract}
\input{front_materials/abstract}
\end{abstract}

\clearpage

\makecopyrightpage 

\makededicationpage
\clearpage

\chapter*{Acknowledgements}
\DoubleSpacing 
\input{front_materials/acknowledgments}
\clearpage
%
%

\SingleSpacing
\hyphenchar\font=-1 
\tableofcontents* 
\hyphenchar\font=`\- 
%
%
%
%
\mainmatter
%

\input{chapters/introduction}
\input{chapters/groomed}
\input{chapters/deviant}
\input{chapters/seabird}
\input{chapters/charmer}
\input{chapters/future}

\newpage
{
    \bibliographystyle{ieee_fullname}
    \bibliography{references}
}

\begin{appendices}
    \input{appendices/publications}
    \input{appendices/groomed_appendix}
    \input{appendices/deviant_appendix}
    \input{appendices/seabird_appendix}
    \input{appendices/charmer_appendix}
\end{appendices}

\end{document}

%% file: packages.tex
\usepackage{colortbl}
\usepackage{amsmath}

\usepackage{graphicx}

\usepackage{amsthm}           
\usepackage{booktabs}

\usepackage{epsfig}
\usepackage{tabularx}
\usepackage[table]{xcolor}       
\usepackage{booktabs}
\usepackage{relsize}
\usepackage{multirow}
\usepackage{array}               
\usepackage{makecell}            
\usepackage{subcaption}          
\usepackage{wasysym}             
\usepackage{float}
\usepackage{caption}
\usepackage{bm}
\usepackage{cite}                
\usepackage{paralist}            
\usepackage{enumitem}            
\usepackage{empheq}
\usepackage{wrapfig}             
\usepackage{xspace}              
\usepackage{arydshln}            
\usepackage{pifont}              
\usepackage{hhline}              

\setlist[itemize, 1]{label =\raisebox{-0.4\height}{\scalebox{1.7}{\textbullet}}}
\setlist[itemize]{noitemsep, topsep=0cm, leftmargin=5mm}

\allowdisplaybreaks              

\graphicspath{{images/}}

\definecolor{ultramarine}{rgb}{0, 0.522, 0.635}
\definecolor{orange}{rgb}{1.0, 0.776, 0.424}
\definecolor{lion}{rgb}{0.322, 0.392, 0.8}
\definecolor{lgreen}{rgb}{0.227, 0.709, 0.294}
\definecolor{lred}{rgb}{0.941, 0.325, 0.137}

\definecolor{darkred}{rgb}{0.65,0.1,0.1}
\definecolor{darkblue}{rgb}{0.1,0.1,0.65}
\definecolor{darkgreen}{rgb}{0.1,0.65,0.1}

\newtheorem{theorem}{Theorem}
\newtheorem{corollary}{Corollary}[theorem]
\newtheorem{lemma}{Lemma}

\makeatletter
    \def\thm@space@setup{
        \thm@preskip=0pt plus 2pt minus 2pt 
        \thm@postskip=\thm@preskip 
    }
\makeatother

\makeatletter
\@namedef{ver@everyshi.sty}{}
\makeatother
\usepackage{tikz} 
\usetikzlibrary{shapes.geometric} 
\usetikzlibrary{calc}  
\usetikzlibrary{arrows}
\usetikzlibrary{arrows.meta}
\usetikzlibrary{shapes.arrows}
\usetikzlibrary{fadings, shadows}
\usetikzlibrary{decorations.text}
\def\centerarc[#1](#2)(#3:#4:#5)
    { \draw[#1] ($(#2)+({#5*cos(#3)},{#5*sin(#3)})$) arc (#3:#4:#5); }

\usepackage[ruled,linesnumbered]{algorithm2e}  
\SetKwIF{If}{ElseIf}{Else}{if}{}{else if}{else}{}
\SetKwFor{For}{for}{do}{end}
\SetKwComment{Comment}{$\triangleright$\ }{}

\SetCommentSty{mycommfont}

\AtBeginEnvironment{thebibliography}{\linespread{1}\selectfont}

\renewenvironment{thebibliography}[1]{
  \begin{oldthebibliography}{#1}
    \setlength{\parskip}{12pt}
}
{
  \end{oldthebibliography}
}

\input{notations}

\usepackage[pagebackref,breaklinks,colorlinks,citecolor=black,urlcolor=black,linkcolor=black]{hyperref}

\usepackage[capitalize]{cleveref}
\crefname{section}{Sec.}{Secs.}
\Crefname{section}{Section}{Sections}
\Crefname{table}{Table}{Tables}
\crefname{table}{Tab.}{Tabs.}
\crefname{figure}{Fig.}{Figs.}
\Crefname{figure}{Figure}{Figures}
\crefname{theorem}{Th.}{Th.}
\Crefname{theorem}{Theorem}{Theorems}

%% file: notations.tex
\definecolor{my_blue}{rgb}{0.2, 0.6, 1}  
\definecolor{my_magenta}{rgb}{1.0, 0.2, 0.6} 
\definecolor{my_yellow}{rgb}{1.0, 0.8, 0.2} 
\definecolor{my_green}{rgb}{0.0, 0.9, 0.24}
\definecolor{my_green_2}{rgb}{0.0, 0.4, 0.0}

\definecolor{sns_blue}{rgb}{0.21, 0.06, 0.42}    
\definecolor{sns_violet}{rgb}{0.45, 0.12, 0.51}  
\definecolor{sns_orange}{rgb}{0.75, 0.22, 0.46}  
\definecolor{sns_yellow}{rgb}{0.99, 0.91, 0.66}  

\definecolor{white}{rgb}{1.0, 1.0, 1.0}
\definecolor{darkGreen}{rgb}{0.01, 0.8, 0.24}
\definecolor{darkGreen2}{rgb}{0.22,0.42, 0.33}
\definecolor{darkGreen3}{rgb}{0.20,0.66, 0.33}
\definecolor{cvprblue}{rgb}{0.21,0.49,0.74}
\definecolor{LightCyan}{rgb}{0.88,1,1}
\definecolor{lightgreen}{HTML}{90EE90}
\definecolor{new_green}{rgb}{0.75,0.97,0.44}
\definecolor{Gray}{gray}{0.95}
\definecolor{lightgray}{rgb}{0.96, 0.96, 0.96}

\definecolor{set1_cyan}{rgb}{0.23, 0.87, 1.0}
\definecolor{building}{rgb}{0.2, 0.33, 0.33}

\definecolor{my_violet}{rgb}{0.79, 0.40, 1} 
\definecolor{my_yellow_2}{rgb}{0.9, 0.8, 0.54}
\definecolor{my_red}{rgb}{1,0,0}
\definecolor{my_purple}{rgb}{0.27,0.8, 0.8}
\definecolor{my_orange}{rgb}{1.0,0.6,0.35}
\definecolor{my_golden}{rgb}{1.0, 0.75, 0.0}
\colorlet{my_gray}{gray!12}

\definecolor{projectionColor}{rgb}{0.2, 0.6, 1}
\definecolor{rayColor}{rgb}{0.0,0.0,0.0}
\definecolor{axisColor}{rgb}{0.0, 0.0, 0.0}
\definecolor{maroon}{rgb}{0.76, 0.13, 0.28}
\colorlet{projectionBorderShade}{rayColor!100}
\colorlet{projectionFillShade}{projectionColor!20}
\colorlet{rayShade}{my_yellow}
\colorlet{axisShade}{axisColor!20}
\colorlet{axisShadeDark}{axisColor!100}

\definecolor{backward_color}{rgb}{1.0, 0.6, 0.2}
\definecolor{forward_color}{rgb}{0.2, 1.0, 0.6}

\definecolor{gain}{HTML}{34a853}
\definecolor{lost}{HTML}{ea4335}

\colorlet{proposedShade}{darkGreen}
\colorlet{vanillaShade}{red!90}

\colorlet{theme_color}{sns_orange}
\colorlet{theme_color_light}{sns_yellow!25}
\colorlet{link_color}{maroon}

\colorlet{ood_color}{gray!20}
\colorlet{methodColor}{cyan!20}

\newcommand{\figureScaleFraction}{0.9}
\newcommand{\waymoFigureScaleFraction}{0.61}

\newcommand{\noIndentHeading}[1]{\noindent\textbf{#1}}

\definecolor{XLcolor}{rgb}{0.858, 0.188, 0.478}

\newcommand{\prof}{Prof.\!\xspace}
\newcommand{\dr}{Dr.\xspace}
\newcommand{\phd}{Ph.D.\xspace}
\newcommand{\myself}{\textit{Abhinav Kumar}\xspace}

\newcommand{\forExample}{\textit{e.g.}\xspace}
\newcommand{\thatIs}{\textit{i.e.}\xspace}

\newcommand{\argmax}{\operatornamewithlimits{argmax}}
\newcommand{\wrt}{\textit{w.r.t.}\xspace}

\newcommand{\cmark}{\checkmark}
\newcommand{\xmark}{\ding{53}}
\newcommand{\autoBraces}[1]{\left(#1\right)}

\newcommand{\bigO}[1]{\mathcal{O}\left(#1\right)}
\newcommand{\clip}[1]{\left\lfloor#1\right\rceil}
\newcommand{\numberElem}[1]{\left|#1\right|}
\newcommand{\elementMul}{\!\odot} 

\newcommand{\identity}{{\bf{I}}}

\DeclareMathOperator{\sign}{sgn}

\newcommand{\scaleFraction}{0.9}
\newcommand{\scaleFractionSeaBird}{0.8}

\newcommand{\myTopRule}{\Xhline{2\arrayrulewidth}}

\newcolumntype{t}{!{\vrule width 1.5\arrayrulewidth}}
\newcolumntype{m}{!{\vrule width 2.5\arrayrulewidth}}
\newcolumntype{a}{>{\columncolor{theme_color_light}}l}
\newcolumntype{b}{>{\columncolor{theme_color_light}}c}
\newcolumntype{d}{>{\columncolor{ood_color}}c}

\newcommand{\CYMyFix}{\cellcolor{theme_color_light}}

\colorlet{cyan_highlight}{my_blue!85}

\colorlet{darkGreen_highlight}{darkGreen!75}

\colorlet{my_magenta_highlight}{my_magenta!50}
\newcommand{\COFix}{\cellcolor{my_magenta_highlight}}
\colorlet{my_yellow_highlight}{my_yellow!55}

\providecommand\rightarrowRHD{\relbar\joinrel\mathrel\RHD}

\providecommand\longrightarrowRHD{\relbar\joinrel\relbar\joinrel\mathrel\RHD}

\newcommand{\uparrowRHD}  {\rotatebox[origin=c]{90}{$\rightarrowRHD$}}
\newcommand{\downarrowRHD}{\rotatebox[origin=c]{270}{$\rightarrowRHD$}}

\newcommand{\uparrowRHDSmall}  {\raisebox{0.05\normalbaselineskip}{\scalebox{0.7}{\uparrowRHD}}}   
\newcommand{\downarrowRHDSmall}{\raisebox{0.07\normalbaselineskip}{\scalebox{0.7}{\downarrowRHD}}} 
\newcommand{\rightarrowRHDSmall}{\raisebox{0.05\normalbaselineskip}{\scalebox{0.75}{$\rightarrowRHD$}}}   

\newcommand{\monoThreeD}{Mono3D\xspace}

\newcommand{\oneD}{$1$D\xspace}
\newcommand{\twoD}{$2$D\xspace}
\newcommand{\threeD}{$3$D\xspace}
\newcommand{\fourD}{$4$D\xspace}
\newcommand{\fiveD}{$5$D\xspace}

\newcommand{\twoDMath}{2\text{D}\xspace}
\newcommand{\threeDMath}{3\text{D}\xspace}
\newcommand{\iou}{IoU\xspace}
\newcommand{\iouTwoD}{IoU$_{2\text{D}}$\xspace}
\newcommand{\iouThreeD}{IoU$_{3\text{D}}$\xspace}
\newcommand{\iouThreeDMath}{\text{\iou}_{3\text{D}}\xspace}
\newcommand{\lidar}{LiDAR\xspace}

\newcommand{\radar}{radar\xspace}

\newcommand{\giouTwoD}{g\iouTwoD}
\newcommand{\giouThreeD}{g\iouThreeD}
\newcommand{\vol}{V}
\newcommand{\bev}{BEV\xspace}

\newcommand{\dla}{DLA-34\xspace}
\newcommand{\dlaThirtyFour}{DLA-34\xspace}
\newcommand{\dlaOneZeroTwo}{DLA-102\xspace}
\newcommand{\dlaOneSixNine}{DLA-169\xspace}

\newcommand{\resNetEighteen}{ResNet-18\xspace}
\newcommand{\resNetFifty}{ResNet-50\xspace}
\newcommand{\resNetOneHundredOne}{ResNet-101\xspace}

\newcommand{\vovNet}{V2-99\xspace}
\newcommand{\efficientDet}{EfficientDet\xspace}
\newcommand{\vit}{ViT\xspace}
\newcommand{\pretrain}{pretrain\xspace}

\newcommand{\kitti}{KITTI\xspace}
\newcommand{\nuscenes}{nuScenes\xspace}
\newcommand{\waymo}{Waymo\xspace}

\newcommand{\imageNet}{ImageNet\xspace}

\newcommand{\valOne}{Val 1\xspace}
\newcommand{\valTwo}{Val 2\xspace}
\newcommand{\val}{Val\xspace}

\newcommand{\frontal}{frontal\xspace}

\newcommand{\kittiThreeSixty}{KITTI-360\xspace}
\newcommand{\kittiThreeSixtyPanoptic}{\kittiThreeSixty PanopticBEV\xspace}
\newcommand{\semanticKITTI}{Semantic KITTI\xspace}
\newcommand{\carla}{CARLA\xspace}

\newcommand{\coda}{CODa\xspace}

\newcommand{\loss}{\mathcal{L}}
\newcommand{\lossBefore}{\loss_{before}}
\newcommand{\lossAfter}{\loss_{after}}
\newcommand{\lossWeigh}{\lambda}
\newcommand{\aploss}{\ap-Loss}
\newcommand{\imageWise}{Imagewise}
\newcommand{\lOne}{\loss_1}
\newcommand{\lTwo}{\loss_2}
\newcommand{\smoothLOne}{\text{Smooth}~\lOne}

\newcommand{\lDice}{\loss_{dice}}
\newcommand{\dice}{dice\xspace}
\newcommand{\Dice}{Dice\xspace}

\newcommand{\size}{\text{size}}
\newcommand{\offset}{\text{offset}}
\newcommand{\heatmap}{\text{heatmap}}
\newcommand{\class}{{class}}
\newcommand{\weightSeg}{\lambda_{seg}}

\newcommand{\MAP}{mAP\xspace}
\newcommand{\MAPLarge}{AP$_{Lrg}$\xspace}
\newcommand{\MAPCar}{AP$_{Car}$\xspace}
\newcommand{\MAPSmall}{AP$_{Sml}$\xspace}
\newcommand{\ap}{AP\xspace}
\newcommand{\apMath}{\text{\ap}}
\newcommand{\apThreeD}{$\apMath_{3\text{D}}$\xspace}
\newcommand{\apHThreeD}{APH$_{3\text{D}}$\xspace}
\newcommand{\apThreeDForty}{\ap$_{3\text{D}|R_{40}}$\xspace}
\newcommand{\apBevForty}{\ap$_{\text{BEV}|R_{40}}$\xspace}
\newcommand{\apTwoD}{\ap$_{2\text{D}}$\xspace}
\newcommand{\apTwoDForty}{\ap$_{2\text{D}|R_{40}}$\xspace}

\newcommand{\NDS}{NDS\xspace}
\newcommand{\MDE}{MDE\xspace}

\newcommand{\levelOne}{Level\_1\xspace}
\newcommand{\levelTwo}{Level\_2\xspace}
\newcommand{\bracketPercentage}{[\%]}
\newcommand{\imageOnly}{image-only\xspace}
\newcommand{\apThreeDSeventy}{\ap$_{\!3\text{D}\!}$ 70\xspace}
\newcommand{\apThreeDFifty}{\ap$_{\!3\text{D}\!}$ 50\xspace}
\newcommand{\apThreeDTwentyFive}{\ap$_{\!3\text{D}\!}$ 25\xspace}
\newcommand{\apThreeDThirty}{\ap$_{\!3\text{D}\!}$ 30\xspace}

\newcommand{\meanFor}{M$_\text{For}$}
\newcommand{\meanEleven}{M$_\text{All}$}

\newcommand{\first}[1]{$\textcolor{sns_blue}{\mathbf{#1}}$}
\newcommand{\second}[1]{$\textcolor{sns_orange}{\mathbf{#1}}$}
\newcommand{\firstKey}[1]{\textcolor{sns_blue}{\textbf{#1}}}
\newcommand{\secondKey}[1]{\textcolor{sns_orange}{\textbf{#1}}}
\newcommand{\sota}{SoTA\xspace}
\newcommand{\best}[1]{$\textcolor{sns_blue}{\mathbf{#1}}$}
\newcommand{\bestKey}[1]{\textcolor{sns_blue}{\textbf{#1}}}
\newcommand{\gain}[1]{\textcolor{gain}{#1}}
\newcommand{\lost}[1]{\textcolor{lost}{#1}}
\newcommand{\good}[2]{{$#1$} {({\gain{\textbf{+}$\mathbf{#2}$}})}}
\newcommand{\bad}[2]{{$#1$} {({\lost{--$#2$}})}}
\newcommand{\arxiv}{ArXiv\xspace}
\newcommand{\mySign}{\!+\!}
\newcommand{\mathDash}{$-$}

\def\relicon{\resizebox{.009\textwidth}{!}{\includegraphics{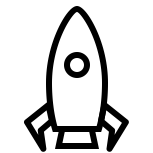}}}
\newcommand{\released}{$\!^{\relicon}$}
\newcommand{\retrained}{$\!^\dagger$}
\newcommand{\reimplemented}{$\!^*$}
\newcommand{\cbgs}{$\!^{\mathsection}$}
\newcommand{\future}{$\!^{\tiny{\fullmoon}\hspace{-0.03cm}\tiny{\newmoon}}$}

\newcommand{\mthreeDRPN}{M3D-RPN\xspace}
\newcommand{\kinematicImage}{Kinematic (Image)\xspace}
\newcommand{\kinematicVideo}{Kinematic (Video)\xspace}
\newcommand{\groomedNMS}{GrooMeD-NMS\xspace}
\newcommand{\groomedNMSFull}{Grouped Mathematically Differentiable NMS\xspace}
\newcommand{\groomedNMSShort}{GrooMeD\xspace}
\newcommand{\monoDISMulti}{MonoDIS-M\xspace}
\newcommand{\gupNet}{GUP Net\xspace}
\newcommand{\deviant}{DEVIANT\xspace}
\newcommand{\deviantFull}{Depth Equivariant Network\xspace}
\newcommand{\pseudoLidar}{Pseudo-{\lidar}\xspace}
\newcommand{\ddThreeD}{DD3D\xspace}
\newcommand{\ddmp}{DDMP-3D\xspace}
\newcommand{\caddn}{CaDDN\xspace}
\newcommand{\patchNet}{PatchNet\xspace}

\newcommand{\monodle}{MonoDLE\xspace}
\newcommand{\monodetr}{MonoDETR\xspace}

\newcommand{\detrThreeD}{DETR3D\xspace}
\newcommand{\fcosThreeD}{FCOS3D\xspace}
\newcommand{\pgd}{PGD\xspace}
\newcommand{\petr}{PETR\xspace}
\newcommand{\petrVTwo}{PETRv2\xspace}
\newcommand{\bevFormer}{BEVFormer\xspace}
\newcommand{\polarFormer}{PolarFormer\xspace}
\newcommand{\bevDepth}{BEVDepth\xspace}
\newcommand{\bevStereo}{BEVStereo\xspace}
\newcommand{\bevDetFourD}{BEVDet4D\xspace}
\newcommand{\beVerse}{BEVerse\xspace}
\newcommand{\beVerseTiny}{BEVerse-T\xspace}
\newcommand{\beVerseSmall}{BEVerse-S\xspace}
\newcommand{\orBoxNet}{Box Net\xspace}
\newcommand{\uvtr}{UVTR\xspace}
\newcommand{\veDet}{VEDet\xspace}
\newcommand{\cape}{CAPE\xspace}
\newcommand{\xThreeKDAll}{X3KD$_{\text{all}}$\xspace}
\newcommand{\frustumFormer}{FrustumFormer\xspace}
\newcommand{\threeDPPE}{3DPPE\xspace}
\newcommand{\hop}{HoP\xspace}
\newcommand{\parametricBEV}{ParametricBEV\xspace}
\newcommand{\sparseBEV}{SparseBEV\xspace}
\newcommand{\saBEV}{SA-BEV\xspace}
\newcommand{\fbBEV}{FB-BEV\xspace}

\newcommand{\beVerseWithMethod}{BEVerse+SeaBird\xspace}
\newcommand{\beVerseTinyWithMethod}{BEVerse-T+SeaBird\xspace}
\newcommand{\beVerseSmallWithMethod}{BEVerse-S+SeaBird\xspace}

\newcommand{\bevHeight}{BEVHeight\xspace}
\newcommand{\hopWithMethod}{HoP+SeaBird\xspace}
\newcommand{\stxd}{STXD\xspace}
\newcommand{\cubeRCNN}{Cube R-CNN\xspace}

\newcommand{\seabird}{SeaBird\xspace}
\newcommand{\seabirdFull}{Segmentation in Bird's View\xspace}
\newcommand{\charmer}{CHARM3R\xspace}
\newcommand{\charmerFull}{Camera Height Agnostic Monocular 3D Object Detector\xspace}

\newcommand{\imageToMaps}{I2M\xspace}
\newcommand{\imageToMapsLong}{Image2Maps\xspace}
\newcommand{\imageToMapsWithMethod}{I2M+SeaBird\xspace}
\newcommand{\panopticBEV}{PBEV\xspace}
\newcommand{\panopticBEVLong}{PanopticBEV\xspace}
\newcommand{\panopticBEVWithMethod}{PBEV+SeaBird\xspace}

\newcommand{\lidarBoxNet}{L-BoxNet\xspace}
\newcommand{\voteNet}{L-VoteNet\xspace}

\newcommand{\mirNet}{MIRNet-v2\xspace}

\newcommand{\ood}{OOD\xspace}
\newcommand{\outDomain}{OOD\xspace}
\newcommand{\inDomain}{ID\xspace}
\newcommand{\camConvs}{CAM-Convs\xspace}
\newcommand{\Variation}{Height\xspace}
\newcommand{\variation}{height\xspace}
\newcommand{\variations}{heights\xspace}
\newcommand{\zeroThreeD}{\bm{0}}
\newcommand{\egoHeight}{H}
\newcommand{\egoHeightChange}{\Delta H}
\newcommand{\uniDrive}{UniDrive\xspace}
\newcommand{\uniDrivePlus}{UniDrive++\xspace}

\newcommand{\newDepth}{\depthPred_{\egoHeightChange}}
\newcommand{\oldDepth}{\depthPred_{0}}

\newcommand{\newDepthGround}{{}^{g}\!\newDepth}
\newcommand{\oldDepthGround}{{}^{g}\!\oldDepth}
\newcommand{\newDepthRegress}{{}^{r}\!\newDepth}
\newcommand{\oldDepthRegress}{{}^{r}\!\oldDepth}
\newcommand{\depthSlope}{\beta}
\newcommand{\depthGTMax}{\depthGT_{max}}
\newcommand{\depthGTMin}{\depthGT_{min}}
\newcommand{\heightImage}{h}
\newcommand{\pixUBottom}{\pixU_b}
\newcommand{\pixVBottom}{\pixV_b}
\newcommand{\pixUCenter}{\pixU_c}
\newcommand{\pixVCenter}{\pixV_c}
\newcommand{\heightBoxTwoD}{h_{2D}}
\newcommand{\pixUCenterTwoD}{\pixU_{c,2D}}
\newcommand{\pixVCenterTwoD}{\pixV_{c,2D}}
\newcommand{\relu}{ReLU\xspace}
\newcommand{\plucker}{Plucker\xspace}
\newcommand{\ray}{\overrightarrow{r}}
\newcommand{\trend}{trend\xspace}
\newcommand{\trends}{trends\xspace}

\newcommand{\groundTruthj}{g_l}
\newcommand{\boxes}{\mathcal{B}}
\newcommand{\boxIndex}{i}
\newcommand{\boxMember}{b}
\newcommand{\boxi}{\boxMember_\boxIndex}
\newcommand{\boxj}{\boxMember_j}
\newcommand{\numboxes}{n}
\newcommand{\boxesAfterNMS}{\mathcal{D}}
\newcommand{\vindex}{\mathbf{d}}

\newcommand{\boxConfidence}{b_{conf}}

\newcommand{\normal}{\mathcal{N}}
\newcommand{\depthPred}{\hat{\posZ}}
\newcommand{\depthPredReg}{{~}^r\depthPred}
\newcommand{\depthPredDice}{{~}^d\depthPred}

\newcommand{\depthGT}{\posZ}

\newcommand{\normalSig}{\sigma}
\newcommand{\normalSigTh}{\sigma_{m}}
\newcommand{\normalSigCr}{\sigma_{c}}
\newcommand{\normalVar}{\sigma^2}
\newcommand{\normalCDF}{\Phi}
\newcommand{\normalErf}{\text{Erf}}
\newcommand{\normalErfInv}{\text{Erf}^{-1}}
\newcommand{\expect}{\mathbb{E}}
\newcommand{\var}{\text{Var}}
\newcommand{\step}{s}

\newcommand{\stepConstant}{{c_1}}
\newcommand{\stepSumTrue}{s}
\newcommand{\image}{\mathbf{h}}
\newcommand{\noise}{\eta}
\newcommand{\funcNoise}{\epsilon}
\newcommand{\layerWeight}{\mathbf{w}}
\newcommand{\layerWeightOptimal}{\mathbf{w_*}}
\newcommand{\instant}{t}
\newcommand{\instantTwo}{j}
\newcommand{\layerWeightZero}{\layerWeight_{0}}
\newcommand{\layerWeightMean}{{}^\loss\bm{\mu}}
\newcommand{\layerWeightConvVanilla}{\layerWeight_\infty}
\newcommand{\layerWeightTimeVanilla}{\layerWeight_{\instant}}
\newcommand{\layerWeightConv}{{}^\loss\layerWeightConvVanilla}
\newcommand{\layerWeightTime}{{}^\loss\layerWeightTimeVanilla}
\newcommand{\gradient}{\mathbf{g}}
\newcommand{\gradTimeVanilla}{\gradient_\instant}
\newcommand{\gradTime}{{}^\loss\gradTimeVanilla}
\newcommand{\gradTimeTwo}{{}^\loss\gradient_\instantTwo}
\newcommand{\uselessConstant}{{c_2}}
\newcommand{\length}{\ell}

\newcommand{\layerWeightConvReg}{{}^r\layerWeightConvVanilla}
\newcommand{\layerWeightConvDice}{{}^d\layerWeightConvVanilla}

\newcommand{\intrinsic}{\mathbf{K}}

\newcommand{\extrinsicTrans}{\mathbf{T}}

\newcommand{\realDomain}{\mathbb{R}}

\newcommand{\equivariant} {equivariant\xspace}
\newcommand{\Equivariant} {Equivariant\xspace}
\newcommand{\equivariance}{equivariance\xspace}
\newcommand{\Equivariance}{Equivariance\xspace}
\newcommand{\scaleEquivariant} {scale \equivariant}
\newcommand{\ScaleEquivariant} {Scale \Equivariant}
\newcommand{\scaleEquivariance}{scale \equivariance}
\newcommand{\ScaleEquivariance}{Scale \Equivariance}
\newcommand{\depthEquivariant} {depth \equivariant}
\newcommand{\DepthEquivariant} {Depth \Equivariant}
\newcommand{\depthEquivariance} {depth \equivariance}

\newcommand{\MaxScale}{Scale-Projection}
\newcommand{\ses}{SES}
\newcommand{\se}{SE}
\newcommand{\weight}{\mathbf{w}}
\newcommand{\plane}{patch plane\xspace}
\newcommand{\transformationGroup}{G}
\newcommand{\transformationGroupMember}{g}
\newcommand{\inputData}{h}
\newcommand{\outputData}{y}
\newcommand{\mapping}{\Phi}
\newcommand{\transformationMath}{\mathcal{T}}
\newcommand{\transformationInput}{\transformationMath_\transformationGroupMember^\inputData}
\newcommand{\transformationOutput}{\transformationMath_\transformationGroupMember^\outputData}
\newcommand{\manifold}{manifold\xspace}

\newcommand{\transformation}{transformation\xspace}

\newcommand{\varX}{X}
\newcommand{\varY}{Y}
\newcommand{\varZ}{Z}

\newcommand{\posX}{x}
\newcommand{\posY}{y}
\newcommand{\posZ}{z}

\newcommand{\norm}[1]{\left\lVert#1\right\rVert}

\newcommand{\pointCloudOne}{H}

\newcommand{\projectionOperator}{\mathbf{K}}

\newcommand{\rotation}{\mathbf{R}}
\newcommand{\translation}{\mathbf{t}}
\newcommand{\transX}{t_\varX}
\newcommand{\transY}{t_\varY}
\newcommand{\transZ}{t_\varZ}
\newcommand{\transXTwo}{\overline{t}_\varX}
\newcommand{\transYTwo}{\overline{t}_\varY}
\newcommand{\transZTwo}{\overline{t}_\varZ}

\newcommand{\predictionX}{\Hat{x}}
\newcommand{\predictionY}{\Hat{y}}
\newcommand{\predictionZ}{\Hat{z}}

\newcommand{\pixU}{u}
\newcommand{\pixV}{v}
\newcommand{\pixUTwo}{u'}
\newcommand{\pixVTwo}{v'}

\newcommand{\ppointU}{u_0} 
\newcommand{\ppointV}{v_0} 
\newcommand{\focal}{f}
\newcommand{\focalNormalized}{\bar{\focal}}
\newcommand{\projectionOne}{h} 
\newcommand{\projectionTwo}{h'}
\newcommand{\projectionOneIndexed}{h_i}
\newcommand{\pixUMinusPointU}{(\!\pixU\!-\!\ppointU\!)}

\newcommand{\pixVMinusPointV}{(\!\pixV\!-\!\ppointV\!)}

\newcommand{\transU}{t_u}
\newcommand{\transV}{t_v}
\newcommand{\predictionU}{\Hat{\pixU}}
\newcommand{\predictionV}{\Hat{\pixV}}
\newcommand{\scaleNotation}{s}

\newcommand{\conv}{*}
\newcommand{\filter}{\Psi}

\newcommand{\vanillaBlocks}{vanilla blocks\xspace}
\newcommand{\vanillaConv}{vanilla convolution\xspace}
\newcommand{\logPolar}{log-polar\xspace}
\newcommand{\LogPolar}{Log-polar\xspace}

\newcommand{\classicalNmsShort}{classical\xspace}
\newcommand{\classicalNmsShortCaps}{Classical\xspace}
\newcommand{\classicalNms}{classical NMS\xspace}
\newcommand{\classicalNmsCaps}{Classical NMS\xspace}
\newcommand{\softNmsCaps}{Soft-NMS\xspace}
\newcommand{\softNmsShortCaps}{Soft\xspace}
\newcommand{\distanceNmsCaps}{Distance-NMS\xspace}
\newcommand{\distanceNmsShortCaps}{Distance\xspace}

\newcommand{\nmsThresh}{N_t}
\newcommand{\validBoxThresh}{v}
\newcommand{\prune}{p}
\newcommand{\basic}{Linear\xspace}
\newcommand{\exponentialPruning}{Exponential\xspace}
\newcommand{\pruneof}[1]{\prune(#1)}
\newcommand{\temperature}{\tau}

\newcommand{\rescore}{{\bf r}}
\newcommand{\score}{{\bf s}}
\newcommand{\rescoreMember}{r_i}

\newcommand{\tempIndex}{\mathbf{t}}
\newcommand{\tempIndexTwo}{\mathbf{u}}
\newcommand{\tempIndexThree}{\mathbf{v}}
\newcommand{\tempIndexFour}{\mathbf{w}}
\newcommand{\pruneMat}{{\bf{P}}}
\newcommand{\overlapMat}{\mathbf{O}}
\newcommand{\zeroVector}{\mathbf{0}}

\def\lowertriangle{
\resizebox{.015\textwidth}{!}{
\setlength{\unitlength}{.1cm}
\begin{picture}(7,5)(0,0)
\put(6,0){\line(-1,0){6}}
\put(0,0){\line(0,1){5}}
\put(6,0){\line(-6,5){6}}
\end{picture}
}
}
\newcommand{\ourTriangle}{\triangle}
\newcommand{\overlapMatLower}{\overlapMat_{\lowertriangle}}
\newcommand{\overlap}{o}
\newcommand{\mask}{{\bf{M}}}

\newcommand{\groups}{\mathcal{G}}
\newcommand{\groupIndex}{k}
\newcommand{\groupMember}{\groups_\groupIndex}
\newcommand{\boxesGroup}{\boxes_{\groupMember}}
\newcommand{\boxesGroupTwo}{\boxes_{\groups_l}}
\newcommand{\rescoreGroup}{\rescore_{\groupMember}}
\newcommand{\scoreGroup}{\score_{\groupMember}}
\newcommand{\pruneMatGroup}{\pruneMat_{\groupMember}}
\newcommand{\identityGroup}{\identity_{\groupMember}}
\newcommand{\maskGroup}{\mask_{\groupMember}}
\newcommand{\groupSize}{\alpha}

%% file: front_materials/abstract.tex
Monocular \threeD object detection (\monoThreeD) is a fundamental computer vision task that estimates an object's class, \threeD position, dimensions, and orientation from a single image. 
Its applications, including autonomous driving, augmented reality, and robotics, critically rely on accurate \threeD environmental understanding. 
This thesis addresses the challenge of generalizing \monoThreeD models to diverse scenarios, including occlusions, datasets, object sizes, and camera parameters.
To enhance occlusion robustness, we propose a mathematically differentiable NMS (\groomedNMS). 
To improve generalization to new datasets, we explore depth equivariant (\deviant) backbones. 
We address the issue of large object detection, demonstrating that it's not solely a data imbalance or receptive field problem but also a noise sensitivity issue. 
To mitigate this, we introduce a segmentation-based approach in bird's-eye view with dice loss (\seabird). 
Finally, we mathematically analyze the extrapolation of \monoThreeD models to unseen camera heights and improve \monoThreeD generalization in such out-of-distribution settings.

%% file: front_materials/acknowledgments.tex
My long PhD journey is the result of all my advisors, mentors, collaborators, friends, and family.

First and foremost, I express my deepest gratitude to my advisor, \prof Xiaoming Liu.
\prof Liu took a bet on me at a point when I was lost in the dark - having an intense desire to do the PhD, but bereft of any support. 
Over the years, his taste, rigor, work-ethics and guidance has instilled in me an awareness of what it takes to do great research. 
The faith he had on my capabilities, even when I did not have on myself, is what I am grateful to him for.

I also thank my PhD committee – \prof Daniel Morris (MSU), \prof Georgia Gkioxari (Caltech, FAIR), \prof Vishnu Boddetti (MSU), and \prof Yu Kong (MSU) for agreeing to serve in my committee and supporting this journey.
I acknowledge \prof Daniel Morris for a three-year long collaboration on the Radar-Camera project, and sharing all his insights in developing radar-camera 3D detectors. 
I thank \prof Georgia Gkioxari, who was also my internship manager at FAIR, Meta AI. 
Her vision expanded my horizons by giving me a taste of moonshot industry grade research, and what it takes to do one.
I thank \prof Vishnu Bodetti and \prof Yu Kong for asking thought-provoking questions in this journey. 

I deeply acknowledge my mentors: \dr Tim Marks, \dr Michael Jones, \dr Anoop Cherian, \dr Ye Wang, \dr Toshi Koike-Akino and \prof Cheng Feng at MERL. 
They took a bet on me as a first year PhD student when I didn't have any significant publications. 
The work done there culminated into my first CVPR paper. 
The paper opened doors to MSU to continue my PhD. 
If not for that internship, my aspirations for a PhD would have come to a crashing end five years back. 

When I joined MSU, \dr Garrick Brazil took me under his wings for a very daunting area of 3D computer vision, and was almost a second advisor to me at MSU and FAIR. 
It was due to his strong belief in me that I applied to FAIR internship, which at that time, I believed was beyond my capacity. 

I acknowledge \dr Yuliang Guo, \dr Xinyu Huang and \dr Liu Ren from Bosch AI Research. We had a long collaboration that spanned across 1 internship and 2 CVPR submissions. 
Their continued guidance and support enabled us to tackle hard open problems.

\dr SriGanesh Madhvanath was my manager at Xerox Research, Bangalore. His mentorship gave me an early realization that as much as the calibre of a candidate matters, the environment and support system matters too. His generous endorsement opened doors for doing PhD in the US. 
As I grow in my career, I hope to pay it forward.
I also thank Vladimir Kozitsky for his outstanding leadership on the LPRv2 project at Xerox Research.

Throughout my PhD journey in the US, I have been told that my Maths and Linear Algebra skills are decent. 
My Master's advisor, \prof Animesh Kumar at IIT Bombay, is the stalwart who should get due credit. 
I stand on his shoulders and am deeply grateful to him for a rigorous foundation in mathematical thinking. 

I also thank my professors at IIT Patna: \prof Ayash Kanto Mukherjee, \prof Kailash Ray, \prof Lokman Hakim Choudhury, \prof Nutan Tomar, \prof Somnath Sarangi, \prof Sumanta Gupta, and \prof Yatendra Singh for their rigorous undergrad training. 
I thank my Physics and Maths $+2$ teacher Jitendra Bharadwaj for his inspirational and thought-provoking teaching, and my teachers at MKDAV Public School, Daltonganj: Antariksh Roy, Asha Mishra, Ashok Verma, Ganga Agarwal, Kunal Kumar, and Rita Sinha for laying a solid foundation to my higher studies.

Additionally, I thank Vincent Mattison and Brenda Hodge, the program coordinator and secretary in the CSE department at MSU for helping me with admin issues every single time. 

This research would not have been possible without the funding from Ford Motor Company and Bosch AI Research. 
I gratefully acknowledge their financial support.

\prof Liu's lab gave me an open culture, access to like-minded peers, and exposure to a setup for doing high quality research. I am thankful to all these amazing people in the lab: \prof Feng Liu, \dr Amin Jourabloo, \dr Xi Yin, \dr Garrick Brazil, \dr Yaojie Liu, Shengjie Zhu, Andrew Hou, Vishal Asnani, Masa Hu, Yunfei Long, Xiao Guo, Minchul Kim, Yiyang Su, Jei Zhu and Zhiyuan Ren for reviewing my ideas, critiquing my papers and open discussions. 
Also, the newer members of the group: Girish Ganeshan, Dinqiang Ye, Zhihao Zhong, Zhizhong Huang, Hoang Le and Ziang Gu for sharing this journey with me.
I am pretty sure each one of you have done and will be doing great in the future.

Next, I thank my friends in East Lansing - Bharat Basti Shenoy, Ankit Gupta, Rahul Dey, Ankit Kumar, Vishal Asnani, Hitesh Gakhar, Sachit Gaudi, Avrajit Ghosh and Ritam Guha, who made me feel East Lansing a second home.



I am grateful to my friends Koushik Chattopadhyay, Saurabh Kumar, Ashay Jain, Manas Pratim Haloi, Vidit Singh, and Priyanka Sinha for being my loudest supporters despite staying thousands of kilometers away. 
All of them have been friends for more than eight years with three for more than fifteen years. 
These were the people with whom I discussed all my PhD quitting plans.

I am also thankful to my parents, and my sister, Ayushi Raj, for their love, patience, support and encouragement, and keeping me sane during this demanding PhD journey.

%% file: chapters/introduction.tex
\chapter{
    Introduction
}

    Monocular 3D object detection (\monoThreeD) is a fundamental computer vision problem that estimates an object's \threeD position, dimensions, and orientation in a scene from a single image and its camera matrix. 
    Its applications, including autonomous driving \cite{park2021pseudo,kumar2022deviant,li2022bevformer}, robotics \cite{saxena2008robotic}, and augmented reality \cite{alhaija2018augmented,Xiang2018RSS,park2019pix,merrill2022symmetry}, critically rely on accurate \threeD environmental understanding. 
    To address these applications' demands, \monoThreeD networks must generalize across occlusions, diverse datasets \cite{kumar2022deviant}, object sizes \cite{kumar2024seabird}, camera intrinsics \cite{brazil2023omni3d}, extrinsics \cite{jia2023monouni,tzofi2023towards}, rotations \cite{moon2023rotation}, weather and geographical conditions \cite{dong2023benchmarking} and be robust to adversarial examples \cite{zhu2023understanding}.
    
    Although \monoThreeD popularity stems from its high accessibility from consumer vehicles compared to \lidar/Radar-based detectors \cite{shi2019pointrcnn,yin2021center,long2023radiant} and computational efficiency compared to stereo-based detectors \cite{Chen2020DSGN}, 
    \monoThreeD methods suffer from classical scale-depth ambiguity making their generalization harder. 
    This is why there are fewer works along the lines of generalizing \monoThreeD. This thesis aims to generalize \monoThreeD to these varying conditions. 

    Most \monoThreeD networks benefit from end-to-end learning idea. 
    However, they train without including NMS in the training pipeline making the final box after NMS outside the training paradigm. 
    While there were attempts to include NMS in the training pipeline for tasks such as \twoD object detection, they have been less widely adopted due to a non-mathematical expression of the NMS. 
    We present and integrate \groomedNMS -- a novel \groomedNMSFull for \monoThreeD, such that the network is trained end-to-end with a loss on the boxes after NMS. We first formulate NMS as a matrix operation and then group and mask the boxes in an unsupervised manner to obtain a simple closed-form expression of the NMS. 
    \groomedNMS addresses the mismatch between training and inference pipelines and, therefore, forces the network to select the best \threeD box in a differentiable manner. 
    As a result, \groomedNMS achieves state-of-the-art monocular \threeD object detection results on the \kitti benchmark dataset performing comparably to monocular video-based methods, and outperforming them on the hard occluded examples.

    Generalizing to datasets requires features which are dataset-independent. 
    One common way is to obtain such features is incorporating inductive bias or symmetries in the network. 
    One such symmetry is translating the ego camera along depth should result in deterministic transformations of the feature maps.
    Modern neural networks use building blocks such as convolutions that are \equivariant{} to arbitrary \twoD translations in the Euclidean manifold. 
    However, these \vanillaBlocks{} are not \equivariant{} to arbitrary \threeD translations in the projective \manifold. 
    Even then, all \monoThreeD networks use \vanillaBlocks{} to obtain the \threeD coordinates, a task for which the \vanillaBlocks{} are not designed for. 
    This paper takes the first step towards convolutions equivariant to arbitrary \threeD translations in the projective \manifold.
    Since the depth is the hardest to estimate for monocular detection, this
    paper proposes \deviantFull (\deviant) built with existing \scaleEquivariant{} steerable blocks. 
    As a result, \deviant is \equivariant{} to the depth translations in the projective \manifold{} whereas vanilla networks are not.
    The additional \depthEquivariance{} forces the \deviant to learn consistent depth estimates, and therefore,
    \deviant works better than vanilla networks in cross-dataset evaluation. 
    \deviant also achieves state-of-the-art monocular \threeD detection 
    results on \kitti and \waymo datasets in the \imageOnly{} category and performs competitively to methods using extra information.

    \monoThreeD networks achieve remarkable performance on cars and smaller objects. 
    However, their performance drops on larger objects, leading to fatal accidents. 
    Large objects like trailers, buses and trucks are harder to detect \cite{wu2023talk} in \monoThreeD, sometimes resulting in fatal accidents \cite{caldwell2022tesla,fernandez2023tesla}. 
    Some attribute these failures to training data scarcity \cite{zhu2019class} or the receptive field requirements \cite{wu2023talk} of large objects.
    We find that modern frontal detectors struggle to generalize to large objects even on nearly balanced datasets.
    We argue that the cause of failure is the sensitivity of depth regression losses to noise of larger objects.
    To bridge this gap, we comprehensively investigate regression and dice losses, examining their robustness under varying error levels and object sizes.
    We mathematically prove that the dice loss leads to superior noise-robustness and model convergence for large objects compared to regression losses for a simplified case.
    Leveraging our theoretical insights, we propose \seabird (\seabirdFull) as the first step towards generalizing to large objects.
    \seabird effectively integrates BEV segmentation on foreground objects for 3D detection, with the segmentation head trained with the dice loss.
    \seabird achieves \sota results on the \kittiThreeSixty leaderboard and improves existing detectors on the \nuscenes leaderboard, particularly for large objects. 

    With all these generalizations, the networks do not generalize well to changing extrinsics or viewpoints in testing. 
    Finally, we aim to extend \monoThreeD's capabilities to varying camera extrinsics, such as camera heights.

\section{Thesis Contributions}
The thesis focuses on generalizing \monoThreeD across occlusions, datasets, object sizes, and camera extrinsics.
The scale-depth ambiguity in \monoThreeD task requires elegant handling of the depth error.
\begin{itemize}
    \item This thesis introduces the mathematically differentiable Non-Maximal Suppression, which attempts \monoThreeD generalization to occluded and hard objects. Most detectors use a post-processing algorithm called Non-Maximal Suppression (NMS) only during inference. While there were attempts to include NMS in the training pipeline for tasks such as \twoD object detection, they have been less widely adopted due to a non-mathematical expression of the NMS. In this chapter, we present and integrate \groomedNMS– a novel Grouped Mathematically Differentiable NMS for monocular \threeD object detection, such that the network is trained end-to-end with a loss on the boxes after NMS. We first formulate NMS as a matrix operation and then group and mask the boxes in an unsupervised manner to obtain a simple closed-form expression of the NMS. \groomedNMS addresses the mismatch between training and inference pipelines and, therefore, forces the network to select the best \threeD box in a differentiable manner. As a result, \groomedNMS achieves state-of-the-art monocular \threeD object detection results on the \kitti dataset.
    
    \item We next propose the depth equivariant backbone in the projective manifold which attempts generalization to unseen datasets.
    Modern neural networks use building blocks such as convolutions that are \equivariant{} to arbitrary \twoD translations in the Euclidean manifold. 
    However, these \vanillaBlocks{} are not \equivariant{} to arbitrary \threeD translations in the projective \manifold. 
    Even then, all monocular \threeD detectors use \vanillaBlocks{} to obtain the \threeD coordinates, a task for which the \vanillaBlocks{} are not designed for. 
    This chapter takes the first step towards convolutions equivariant to arbitrary \threeD translations in the projective \manifold.
    Since the depth is the hardest to estimate for monocular detection, this
    chapter proposes \deviantFull (\deviant) built with existing \scaleEquivariant{} steerable blocks. 
    As a result, \deviant is \equivariant{} to the depth translations in the projective \manifold{} whereas vanilla networks are not.
    The additional \depthEquivariance{} forces the \deviant to learn consistent depth estimates, and therefore,
    \deviant achieves state-of-the-art monocular \threeD detection 
    results on \kitti and \waymo datasets in the \imageOnly{} category and performs competitively to methods using extra information.
    Moreover, \deviant works better than vanilla networks in cross-dataset evaluation.
    
    \item We then investigate large object detection, demonstrating that it is not solely a data imbalance or receptive field issue but also a noise sensitivity problem. To generalize \monoThreeD to large objects, it introduces a segmentation-based approach in bird’s eye view with dice loss (SeaBird). 
    Monocular \threeD detectors achieve remarkable performance on cars and smaller objects. 
    However, their performance drops on larger objects, leading to fatal accidents. 
    Some attribute the failures to training data scarcity or their receptive field requirements of large objects.
    In this chapter, we highlight this understudied problem of generalization to large objects.
    We find that modern frontal detectors struggle to generalize to large objects even on nearly balanced datasets.
    We argue that the cause of failure is the sensitivity of depth regression losses to noise of larger objects.
    To bridge this gap, we comprehensively investigate regression and dice losses, examining their robustness under varying error levels and object sizes.
    We mathematically prove that the dice loss leads to superior noise-robustness and model convergence for large objects compared to regression losses for a simplified case.
    Leveraging our theoretical insights, we propose \seabird (\seabirdFull) as the first step towards generalizing to large objects.
    \seabird effectively integrates BEV segmentation on foreground objects for 3D detection, with the segmentation head trained with the dice loss.
    \seabird achieves \sota results on the \kittiThreeSixty leaderboard and improves existing detectors on the \nuscenes leaderboard, particularly for large objects. 
    
    \item Monocular 3D object detectors, while effective on data from one ego camera height, struggle with unseen or out-of-distribution camera heights. 
    Existing methods often rely on Plucker embeddings, image transformations or data augmentation. 
    This chapter takes a step towards this understudied problem by investigating the impact of camera height variations on state-of-the-art (\sota) \monoThreeD models.
    With a systematic analysis on the extended CARLA dataset with multiple camera heights, we observe that depth estimation is a primary factor influencing performance under height variations. 
    We mathematically prove and also empirically observe consistent negative and positive \trends in mean depth error of regressed and ground-based depth models, respectively, under camera height changes. 
    To mitigate this, we propose Camera Height Robust Monocular 3D Detector (\charmer), which averages both depth estimates within the model.
    \charmer significantly improves generalization to unseen camera heights, achieving \sota performance on the CARLA dataset. 
\end{itemize}

\section{Thesis Organization}
    We organize the remaining chapters of the dissertation as follows.
    \cref{chpt:groomed} introduces the mathematically differentiable Non-Maximal Suppression, which attempts generalization to occluded and hard objects.
    \cref{chpt:deviant} describes the depth equivariant backbone which attempts generalization to unseen datasets.
    \cref{chpt:seabird} investigates large object detection, demonstrating that it is not solely a data imbalance or receptive field issue but also a noise sensitivity problem. To improve large object detection, it introduces a segmentation-based approach in bird’s eye view with dice loss (SeaBird). 
    \cref{chpt:charm3r} attempts solving the generalization of \monoThreeD trained on single camera height to multiple camera heights.
    \cref{chpt:future} introduces the future research for monocular 3D detection.

%% file: chapters/groomed.tex
\chapter{
    \groomedNMS: \groomedNMSFull for Monocular 3D Object Detection
}
\label{chpt:groomed}

Modern \threeD object detectors have immensely benefited from the end-to-end learning idea. 
However, most of them use a post-processing algorithm called Non-Maximal Suppression (NMS) only during inference. 
While there were attempts to include NMS in the training pipeline for tasks such as \twoD object detection, they have been less widely adopted due to a non-mathematical expression of the NMS. 
In this chapter, we present and integrate \groomedNMS -- a novel 
Grouped Mathematically Differentiable NMS for monocular \threeD object detection, such that the network is trained end-to-end with a loss on the boxes after NMS.
We first formulate NMS as a matrix operation and then group and mask the boxes in an unsupervised manner to obtain a simple closed-form expression of the NMS. 
\groomedNMS addresses the mismatch between training and inference pipelines and, therefore, forces the network to select the best \threeD box in a differentiable manner. 
As a result, \groomedNMS achieves state-of-the-art monocular \threeD object detection results on the \kitti benchmark dataset performing comparably to monocular video-based methods.

\section{Introduction}\label{sec:Introduction}
    \threeD object detection is one of the fundamental problems in computer vision, where the task is to infer \threeD information of the object. 
    Its applications include augmented reality~\cite{alhaija2018augmented,rematas2018soccer}, robotics~\cite{saxena2008robotic,levine2018learning}, medical surgery~\cite{rey2002automatic}, and, more recently path planning and scene understanding in autonomous driving~\cite{chen2017multi,huang2020epnet,simonelli2020disentangling,li2020rtm3d}. 
    Most of the \threeD object detectors 
    ~\cite{chen2017multi,li2019gs3d,huang2020epnet,simonelli2020disentangling,li2020rtm3d} are extensions of the \twoD~object  
    detector Faster R-CNN~\cite{ren2015faster}, which relies on the end-to-end learning idea to achieve State-of-the-Art (\sota)~object detection.
    Some of these methods have proposed changing 
    architectures~\cite{simonelli2020disentangling,shi2020distance,li2020rtm3d} or losses~\cite{brazil2019m3d,chen2020monopair}. 
    Others have tried incorporating confidence~\cite{simonelli2020disentangling,brazil2020kinematic,shi2020distance} 
    or temporal cues~\cite{brazil2020kinematic}.
    
    Almost all of them output a massive number of boxes for each object and, thus, rely on post-processing with a greedy~\cite{prokudin2017learning} clustering algorithm called Non-Maximal Suppression (NMS) during inference to reduce the number of false positives and increase performance. 
    However, these works have largely overlooked NMS's inclusion in training leading to an apparent \emph{mismatch} between training and inference pipelines as the losses are applied on all boxes before NMS but not on \emph{final} boxes after NMS (see \cref{fig:groomed_coventional_nms_pipeline}).
    We also find that \threeD object detection suffers a greater mismatch between classification and \threeD localization compared to that of \twoD~localization, as discussed further in \cref{sec:results_oracle_additional} of the supplementary and observed in~\cite{huang2020epnet, brazil2020kinematic, shi2020distance}.
    Hence, our focus is \threeD object detection. 

\begin{figure}[!tb]
    \centering
    \begin{subfigure}[t]{.3\linewidth}
        \input{images/groomed/block_diagram_others.tex}\\
        \subcaption{Conventional NMS Pipeline}\label{fig:groomed_coventional_nms_pipeline}
    \end{subfigure}
    \hspace{0.62cm}
    \begin{subfigure}[t]{.3\linewidth}
        \input{images/groomed/block_diagram_ours.tex}\\
        \subcaption{\groomedNMS Pipeline}\label{fig:groomed_nms_pipeline}
    \end{subfigure}
    \hspace{0.62cm}
    \begin{subfigure}[t]{.32\linewidth}
        \input{images/groomed/nms_inside_2.tex}\\
        \subcaption{\groomedNMS layer \label{fig:layer}}
    \end{subfigure}
    \caption[\groomedNMS Overview]{    
    \textbf{Overview of \groomedNMS.}
    (a) Conventional object detection has a mismatch between training and inference as it uses NMS only in inference. (b) To address this, we propose a novel \groomedNMS layer, such that the network is trained end-to-end with NMS applied.
    $\score$ and $\rescore$ denote the score of boxes~$\boxes$ before and after the NMS respectively. $\overlapMat$ denotes the matrix containing \iouTwoD~overlaps of $\boxes$. $\lossBefore$ denotes the losses before the NMS,  
    while $\lossAfter$ denotes the loss after the NMS. 
    (c) \groomedNMS layer calculates~$\rescore$ in a differentiable manner giving gradients from $\lossAfter$ when the best-localized box corresponding to an object is not selected after NMS.}
    \label{fig:overview}
\end{figure}
    
    Earlier attempts to include NMS in the training pipeline~\cite{hosang2016convnet, prokudin2017learning, hosang2017learning} have been made for \twoD~object detection where the improvements are less visible.
    Recent efforts to improve the correlation in \threeD object detection involve calculating~\cite{simonelli2019disentangling, simonelli2020disentangling} or predicting~\cite{brazil2020kinematic, shi2020distance} the scores via likelihood estimation~\cite{kumar2020luvli} or enforcing the correlation explicitly~\cite{huang2020epnet}. 
    Although this improves the \threeD detection performance, 
    improvements are limited as their training pipeline is not end to end in the absence of a differentiable NMS. 
    
    To address the mismatch between training and inference pipelines as well as the mismatch between classification and \threeD localization, we propose including the NMS in the training pipeline, which gives a useful gradient to the network so that it figures out which boxes are the best-localized in \threeD and, therefore, should be ranked higher (see \cref{fig:groomed_nms_pipeline}).
    
    An ideal NMS for inclusion in the training pipeline should be not only differentiable but also parallelizable. 
    Unfortunately, the inference-based \classicalNms~and \softNmsCaps~\cite{bodla2017soft} are greedy, set-based and, therefore, not parallelizable~\cite{prokudin2017learning}. 
    To make the NMS parallelizable, we first formulate the \classicalNms~as matrix operation and then obtain a closed-form mathematical expression using elementary matrix operations such as matrix multiplication, matrix inversion, and clipping. 
    We then replace the threshold pruning in the \classicalNms~with its softer version~\cite{bodla2017soft} to get useful gradients. These two changes make the NMS GPU-friendly, and the gradients are backpropagated.  
    We next group and mask the boxes in an unsupervised manner, which removes the matrix inversion and simplifies our proposed differentiable NMS expression further. We call this NMS as~\groomedNMSFull~(\groomedNMS).

    In summary, the main contributions of this work include:
    \begin{itemize}
        \item This is the first work to propose and integrate a closed-form mathematically differentiable NMS for object detection, such that the network is trained end-to-end with a loss on the boxes after NMS.
        \item We propose an unsupervised grouping and masking on the boxes to remove the matrix inversion in the closed-form NMS expression.
        \item We achieve \sota~monocular~\threeD object detection performance on the \kitti dataset performing comparably to monocular video-based methods.
    \end{itemize}

\section{Related Works}

    \noindent\textbf{3D~Object Detection.}
        Recent success in \twoD~object detection~\cite{girshick2014rich,girshick2015fast, ren2015faster,redmon2016you,lin2018focal}
        has inspired people to infer \threeD information from a single \twoD~(monocular) image. 
        However, the monocular problem is ill-posed due to the inherent scale/depth ambiguity~\cite{tang2020center3d}. 
        Hence, approaches use additional sensors such as \lidar~\cite{shi2019pointrcnn,wu2020motionnet,huang2020epnet}, stereo~\cite{li2019stereo,wang2019pseudo} or radar~\cite{vasile2005pose,moosmann2009segmentation}. 
        Although \lidar~depth estimations are accurate, \lidar~data is sparse~\cite{hu2020you} and computationally expensive to process~\cite{tang2020center3d}. 
        Moreover, \lidar s are expensive and do not work well in severe weather~\cite{tang2020center3d}. 
    
        Hence, there have been several works on monocular \threeD object detection. Earlier approaches~\cite{payet2011contours, fidler20123d, pepik2015multi, chen2016monocular} use hand-crafted features, while the recent ones are all based on deep learning.
        Some of these methods have proposed changing
        architectures~\cite{liu2019deep,li2020rtm3d,tang2020center3d} or losses~\cite{brazil2019m3d,chen2020monopair}. 
        Others have tried incorporating confidence~\cite{liu2019deep,simonelli2020disentangling,brazil2020kinematic,shi2020distance}, augmentation~\cite{simonelli2020towards},  
        depth in convolution~\cite{brazil2019m3d, ding2020learning} or temporal cues~\cite{brazil2020kinematic}. 
        Our work proposes to incorporate NMS in the training pipeline of monocular \threeD object detection.
    
    \noindent\textbf{Non-Maximal Suppression.}
        NMS has been used to reduce false positives in edge detection~\cite{rosenfeld1971edge}, feature point detection~\cite{harris1988combined, lowe2004distinctive, mikolajczyk2004scale}, face detection~\cite{viola2001rapid}, human detection~\cite{dalal2005histograms, brazil2017illuminating, brazil2019pedestrian} as well as \sota~\twoD~\cite{girshick2015fast, ren2015faster,redmon2016you,lin2018focal} and \threeD  detection~\cite{bao2019monofenet, chen2017multi, simonelli2020disentangling, brazil2020kinematic, shi2020distance, tang2020center3d}.
        Modifications to NMS in \twoD~detection~\cite{desai2011discriminative, bodla2017soft,hosang2016convnet, prokudin2017learning, hosang2017learning},~\twoD~pedestrian detection~\cite{rujikietgumjorn2013optimized, lee2016individualness, liu2019adaptive},~\twoD~salient object detection~\cite{zhang2016unconstrained} and \threeD detection~\cite{shi2020distance} can be classified into three categories -- inference NMS~\cite{bodla2017soft,shi2020distance}, optimization-based NMS~\cite{desai2011discriminative,wan2015end, rujikietgumjorn2013optimized, lee2016individualness, zhang2016unconstrained, azadi2017learning} and neural network based NMS~\cite{henderson2016end, hosang2016convnet, prokudin2017learning, hosang2017learning, liu2019adaptive}. 
        
        The inference NMS~\cite{bodla2017soft} changes the way the boxes are pruned in the final set of predictions. 
        \cite{shi2020distance} uses weighted averaging to update the $z$-coordinate after NMS.
        \cite{rujikietgumjorn2013optimized} solves quadratic unconstrained binary optimization while~\cite{lee2016individualness, azadi2017learning, some2020determinantal} and~\cite{zhang2016unconstrained} use point processes and MAP based inference respectively. 
        \cite{desai2011discriminative} and~\cite{wan2015end} formulate NMS as a structured prediction task for isolated and all object instances respectively.
        The neural network NMS use a multi-layer network and message-passing to approximate NMS~\cite{hosang2016convnet, prokudin2017learning, hosang2017learning} or to predict the NMS threshold adaptively~\cite{liu2019adaptive}.
        \cite{henderson2016end} approximates the sub-gradients of the network without modelling NMS via a transitive relationship. 
        Our work proposes a grouped closed-form mathematical approximation of the \classicalNms~and does not require multiple layers or message-passing. 
        We detail these differences in \cref{sec:difference}.

\section{Background}

    \subsection{Notations}\label{subsec:notations}
        Let $\boxes\!=\!\{b_i\}_{i=1}^\numboxes$ denote the set of boxes or proposals $\boxi$ from an image.
        Let $\score\!=\!\{s_i\}_{i=1}^\numboxes$ and $\rescore\!=\!\{r_i\}_{i=1}^\numboxes$ denote their scores (before NMS) and rescores (updated scores after NMS) respectively such that $r_i, s_i \ge 0~\forall~i$. $\boxesAfterNMS$ denotes the subset of $\boxes$ after the NMS. 
        Let $\overlapMat= [\overlap_{ij}]$  denote the $\numboxes\times \numboxes$ matrix with $\overlap_{ij}$ denoting the \twoD~Intersection over Union~(\iouTwoD)~of $\boxi$ and $\boxj$. 
        The \emph{pruning} function $\prune$ decides how to rescore a set of boxes $\boxes$ based on \iouTwoD~overlaps of its neighbors, sometimes suppressing boxes entirely.
        In other words, $\pruneof{\overlap_i} = 1$ denotes the box $\boxi$ is suppressed while $\pruneof{\overlap_i} = 0$ denotes $\boxi$ is kept in $\boxesAfterNMS$. 
        The NMS threshold $\nmsThresh$ is the threshold for which two boxes need in order for the non-maximum to be suppressed. 
        The temperature $\temperature$ controls the shape of the exponential and sigmoidal pruning functions $\prune$. 
        $\validBoxThresh$ thresholds the rescores in \groomedNMSShort~and \softNmsCaps~\cite{bodla2017code} to decide if the box remains valid after NMS.
        
        $\boxes$ is partitioned into different groups $\groups\!=\!\{\groupMember\}$. 
        $\boxesGroup$ denotes the subset of $\boxes$ belonging to group $\groupIndex$. 
        Thus, $\boxesGroup\!=\!\{\boxi\}~\forall~\boxi \in \groupMember$ and $ \boxesGroup\cap\boxesGroupTwo\!=\!\phi~\forall~\groupIndex \ne l$. 
        $\groupMember$ in the subscript of a variable denotes its subset corresponding to $\boxesGroup$. 
        Thus, $\scoreGroup$ and $\rescoreGroup$ denote the scores and the rescores of $\boxesGroup$ respectively. 
        $\alpha$ denotes the maximum group size.
        
        $\vee$ denotes the logical OR while $\clip{x}$ denotes clipping of $x$ in the range $[0,1]$. Formally,
        \vspace{-0.2cm}
        \begin{align}
            \clip{x} &= 
            \begin{cases}
                1, & x > 1\\
                x, & 0 \le x \le 1\\
                0, & x < 0
            \end{cases}
        \vspace{-0.2cm}
        \end{align}
        $\numberElem{\score}$ denotes the number of elements in $\score$. 
        $\lowertriangle$ in the subscript denotes the lower triangular version of the matrix without the principal diagonal.
        $\elementMul$ denotes the element-wise multiplication.
        $\identity$ denotes the identity matrix.

        \begin{algorithm}[!t]
            \caption{\classicalNmsShortCaps/\softNmsCaps~\cite{bodla2017soft}}
            \label{alg:classical}
            \SetAlgoLined
            \DontPrintSemicolon
            \KwIn{$\score$:~scores, $\overlapMat$:~\iouTwoD~matrix, $N_t$:~NMS threshold,
                    $\prune$: pruning function, $\temperature$:~temperature
                    }
            \KwOut{ $\vindex$: box index after NMS, $\rescore$: scores after NMS}
            \Begin
            {
                $\vindex \gets \{ \}$\;
                $\tempIndex \gets \{1, \cdots, \numberElem{\score}\}$ \Comment*[r]{All box indices}
                $\rescore \gets \score$\;
                \While{$\tempIndex \ne empty$}
                {
                    $\nu \gets \argmax~\rescore[\tempIndex]$ \Comment*[r]{Top scored box}
                    $\vindex \gets \vindex \cup \nu$ \Comment*[r]{Add to valid box index}
                    $\tempIndex \gets \tempIndex - \nu$ \Comment*[r]{Remove from $\tempIndex$}
                    \For{$\boxIndex \gets 1:\numberElem{\tempIndex}$}
                    {
                            $\rescoreMember \gets (1- p_{\temperature}(\overlapMat[\nu, i]))\rescoreMember $ \Comment*[r]{Rescore}
                    }
                }
            }
        \end{algorithm}

    \subsection{\classicalNmsShortCaps~and~\softNmsCaps}
        NMS is one of the building blocks in object detection whose
        high-level goal is to iteratively suppress boxes which have too much \iou~with a nearby high-scoring box.
        We first give an overview of the classical and \softNmsCaps~\cite{bodla2017soft}, which are greedy and used in inference. 
        \classicalNmsCaps~uses the idea that the score of a box having a high \iouTwoD~overlap with \emph{any} of the selected boxes should be suppressed to zero. 
        That is, it uses a hard pruning $p$ without any temperature $\temperature$. 
        \softNmsCaps~makes this pruning soft via temperature $\temperature$. 
        Thus, \classicalNmsShort~and \softNmsCaps~only differ in the choice of $p$. 
        We reproduce them in Alg.~\ref{alg:classical} using our notations.

\section{\groomedNMS}
    \classicalNmsCaps~(Alg.~\ref{alg:classical}) uses $\argmax$ and greedily calculates the rescore $r_i$ of boxes $\boxes$ and, is thus not parallelizable or differentiable~\cite{prokudin2017learning}.
    We wish to find its smooth approximation in closed-form for including in the training pipeline.

    \subsection{Formulation}

        \subsubsection{Sorting}
            \classicalNmsCaps~uses the non-differentiable hard $\argmax$ operation (Line $6$ of Alg.~\ref{alg:classical}). 
            We remove the $\argmax$ by hard sorting the scores $\score$ and $\overlapMat$ in decreasing order (lines $2$-$3$ of Alg.~\ref{alg:diff_nms}).
            We also try making the sorting soft. 
            Note that we require the permutation of $\score$ to sort $\overlapMat$. 
            Most soft sorting methods~\cite{ poganvcic2019differentiation,paulus2020gradient,blondel2020fast,berthet2020learning} apply the soft permutation to the same vector. 
            Only two other methods~\cite{cuturi2019differentiable,prillo2020softsort} can apply the soft permutation to another vector. Both methods use $\bigO{n^2}$ computations for soft sorting~\cite{blondel2020fast}. 
            We implement~\cite{prillo2020softsort} and find that~\cite{prillo2020softsort} is overly dependent on temperature $\temperature$ to break out the ranks, and its gradients are too unreliable to train our model. 
            Hence, we stick with the hard sorting of $\score$ and $\overlapMat$.

        \begin{algorithm}[!t]
            \caption{\groomedNMS}
            \label{alg:diff_nms}
            \SetAlgoLined
            \DontPrintSemicolon
            \KwIn{$\score$: scores, $\overlapMat$: \iouTwoD~matrix, $\nmsThresh$: NMS threshold,
                    $\prune$: pruning function,
                    $\validBoxThresh$: valid box threshold, $\groupSize$: maximum group size}
            \KwOut{ $\vindex$: box index after NMS, $\rescore$: scores after NMS}
            \Begin
            {
                $\score$, index $\gets \text{sort}(\score$, descending$=$ True$)$\Comment*[r]{Sort $\score$}
                $\overlapMat \gets \overlapMat[$index$][:$, index$]$ \Comment*[r]{Sort $\overlapMat$}
                $\overlapMatLower \gets \text{lower}(\overlapMat)$ \Comment*[r]{Lower $\ourTriangle$ular matrix}
                $\pruneMat \gets \prune(\overlapMatLower)$ \Comment*[r]{Prune matrix}
                $\identity \gets \text{Identity}(\numberElem{\score})$ \Comment*[r]{Identity matrix}
                $\groups \gets \text{group}(\overlapMat, \nmsThresh, \groupSize)$ \Comment*[r]{Group boxes $\boxes$}
                \For{$\groupIndex \gets 1:\numberElem{\groups}$}
                {
                    $\maskGroup \gets$ zeros ($\numberElem{\groupMember}, \numberElem{\groupMember}$)\Comment*[r]{Prepare mask}
                    $\maskGroup[:, \groupMember[1]] \gets 1$ \Comment*[r]{First col of $\maskGroup$}
                    $\rescore_{\groupMember} \gets \clip{\left( \identityGroup - \maskGroup \elementMul \pruneMatGroup \right) \scoreGroup}$ \Comment*[r]{Rescore}
                }
                $\vindex \gets \text{index}[ \rescore >= \validBoxThresh]$ \Comment*[r]{Valid box index}
            }
        \end{algorithm}

        \subsubsection{NMS as a Matrix Operation}\label{subsec:matrix_formulation}
            The rescoring process of the~\classicalNms~is greedy set-based~\cite{prokudin2017learning} and only considers overlaps with unsuppressed boxes. 
            We first generalize this rescoring 
            by accounting for the effect of all (suppressed and unsuppressed) boxes as
            \begin{align}
                r_i &\approx \max\left(s_i - \sum\limits_{j=1}^{i-1}\pruneof{\overlap_{ij}}r_j,~0 \right)
                \label{eq:diff_nms_rescore}
            \end{align}
            using the relaxation of logical OR $\bigvee$ operator as $\sum$~\cite{van2020analyzing, li2019augmenting}.
            See \cref{sec:NMS_explanation} of the supplementary material for an alternate explanation of \cref{eq:diff_nms_rescore}.
            The presence of $r_j$ on the RHS of \cref{eq:diff_nms_rescore} prevents suppressed boxes from influencing other boxes hugely. 
            When $\prune$ outputs discretely as $\{0, 1\}$ as in~\classicalNms, scores $s_i$ are guaranteed to be suppressed to $r_i=0$ or left unchanged $r_i=s_i$ thereby implying $r_i \leq s_i~\forall~i$. 
            We write the rescores $\rescore$ in a matrix formulation as
            \begin{align}
                \begin{bmatrix}
                r_1 \\
                r_2 \\
                r_3 \\
                \vdots \\
                r_n\\
                \end{bmatrix}
                    &\!\approx\! 
                \max\left(
                \begin{bmatrix}
                s_1\\
                s_2\\
                s_3\\
                \vdots \\
                s_n\\
                \end{bmatrix}
                -
                \begin{bmatrix}
                0 & 0 & \dots & 0\\
                \pruneof{\overlap_{21}}\!&\!0\!&\!\dots\!&\!0\\
                \pruneof{\overlap_{31}}\!&\!\pruneof{\overlap_{32}}\!&\!\dots\!&\!0\\
                \vdots\!& \vdots & \vdots\!&\!\vdots \\
                \pruneof{\overlap_{n1}}\!&\!\pruneof{\overlap_{n2}} & \dots & 0 \\
                \end{bmatrix} 
                \begin{bmatrix}
                r_1 \\
                r_2 \\
                r_3 \\
                \vdots \\
                r_n\\
                \end{bmatrix}
                ,
                \begin{bmatrix}
                0 \\
                0 \\
                0 \\
                \vdots \\
                0\\
                \end{bmatrix}
                \right).
            \end{align}
            The above equation is written compactly as 
            \begin{align}
                \rescore &\approx \max(\score - \pruneMat \rescore,\zeroVector),
                \label{eq:diff_nms_recursive}
            \end{align}
            where $\pruneMat$, called the Prune Matrix, is obtained 
            when the pruning function $\prune$ operates element-wise
            on $\overlapMatLower$. 
            Maximum operation makes \cref{eq:diff_nms_recursive} non-linear~\cite{kumar2013estimation} and, thus, difficult to solve. 
            However, to avoid recursion, we use 
            \begin{align}
                \rescore &\approx \clip{\left( \identity + \pruneMat \right)^{-1}\score}, \label{eq:diff_nms_full}
            \end{align}
            as the solution to \cref{eq:diff_nms_recursive} with $\identity$ being the identity matrix. 
            Intuitively, if the matrix inversion is considered division in \cref{eq:diff_nms_full} and the boxes have overlaps, the rescores are the scores divided by a number greater than one and are, therefore, lesser than scores. 
            If the boxes do not overlap, the division is by one and rescores equal scores.
            
            Note that the $\identity + \pruneMat$ in~\cref{eq:diff_nms_full} is a lower triangular matrix with ones on the principal diagonal. Hence, $\identity + \pruneMat$ is always full rank and, therefore, always invertible.

        \subsubsection{Grouping}
            We next observe that the object detectors output multiple boxes for an object, and a good detector outputs boxes wherever it finds objects in the monocular image. 
            Thus, we cluster the boxes in an image in an unsupervised manner based on \iouTwoD~overlaps to obtain the groups $\groups$. 
            Grouping thus mimics the grouping of the~\classicalNms, but does not rescore the boxes. 
            As clustering limits interactions to intra-group interactions among the boxes, we write~\cref{eq:diff_nms_full} as
            \begin{align}
               \rescoreGroup &\approx \clip{\left( \identityGroup + \pruneMatGroup \right)^{-1}\scoreGroup}.
               \label{eq:diff_nms_group_intermediate}
            \end{align}
            This results in taking smaller matrix inverses in \cref{eq:diff_nms_group_intermediate} than \cref{eq:diff_nms_full}.

        \begin{algorithm}[!t]
            \caption{Grouping of boxes}
            \label{alg:group}
            \SetAlgoLined
            \DontPrintSemicolon
            \KwIn{$\overlapMat$: sorted \iouTwoD~matrix, $\nmsThresh$: NMS threshold,
                    $\groupSize$: maximum group size}
            \KwOut{ $\groups$: Groups}
            \Begin{
                    $\groups \gets \{ \}$\;
                    $\tempIndex \gets \{1,\cdots,\overlapMat.\text{shape}[1]\}$\Comment*[r]{All box indices}
                    \While{$\tempIndex \ne empty$}
                    {
                        $\tempIndexTwo\!\gets\!\overlapMat[:,1]\!>\nmsThresh$ \Comment*[r]{High overlap indices}
                        $\tempIndexThree \gets \tempIndex[\tempIndexTwo]$\Comment*[r]{New group}
                        $n_{\groupMember} \gets \min(|\tempIndexThree|, \alpha)$\;
                        $\groups$.insert$(\tempIndexThree[:n_{\groupMember}])$\Comment*[r]{Insert new group}
                        $\tempIndexFour\!\gets\!\overlapMat[:,1]\!\le\nmsThresh$\Comment*[r]{Low overlap indices}
                        $\tempIndex \gets \tempIndex[\tempIndexFour]$\Comment*[r]{Keep $\tempIndexFour$ indices in $\tempIndex$}
                        $\overlapMat \gets \overlapMat[\tempIndexFour][:,\tempIndexFour]$\Comment*[r]{Keep $\tempIndexFour$ indices in $\overlapMat$}
                }
            }
        \end{algorithm}

            We use a simplistic grouping algorithm, \thatIs, we form a group $\groupMember$ with boxes  having high \iouTwoD~overlap with the top-ranked box, given that we sorted the scores. 
            As the group size is limited by $\groupSize$, we choose a minimum of $\groupSize$ and the number of boxes in $\groupMember$. 
            We next delete all the boxes of this group and iterate until we run out of boxes.
            Also, grouping uses \iouTwoD~since we can achieve meaningful clustering in \twoD. 
            We detail this unsupervised grouping in Alg.~\ref{alg:group}. 

        \subsubsection{Masking}
            \classicalNmsCaps~considers the \iouTwoD~of the top-scored box with other boxes. 
            This consideration is equivalent to only keeping the column of $\overlapMat$ corresponding to the top box while assigning the rest of the columns to be zero. 
            We implement this through masking of $\pruneMatGroup$. 
            Let $\maskGroup$ denote the binary mask corresponding to group $\groupMember$. 
            Then, entries in the binary matrix $\maskGroup$ in the column corresponding to the top-scored box are $1$ and the rest are $0$.
            Hence, only one of the columns in $\maskGroup\elementMul~\pruneMatGroup$ is non-zero. 
            Now, $\identityGroup+\maskGroup\elementMul\pruneMatGroup$ is a Frobenius matrix (Gaussian transformation) and we, therefore, invert this matrix by simply subtracting the second term~\cite{golub2013matrix}. 
            In other words, $(\identityGroup+ \maskGroup\elementMul\pruneMatGroup)^{-1} =\identityGroup - \maskGroup\elementMul\pruneMatGroup$. Hence, we simplify \cref{eq:diff_nms_group_intermediate} further to get
            \begin{align}
               \rescoreGroup &\approx \clip{ \left( \identityGroup - \maskGroup\elementMul\pruneMatGroup \right)\scoreGroup}.
               \label{eq:diff_nms_group_mask}
            \end{align}
            Thus, masking allows to bypass the computationally expensive matrix inverse operation altogether.
            
            We call the NMS based on \cref{eq:diff_nms_group_mask} as Grouped Mathematically Differentiable Non-Maximal Suppression or \groomedNMS. 
            We summarize the complete \groomedNMS in Alg.~\ref{alg:diff_nms} and show its block-diagram in \cref{fig:layer}. 
            \groomedNMS in \cref{fig:layer} provides two gradients - one through $\score$ and other through $\overlapMat$.

        \subsubsection{Pruning Function}\label{sec:pruning}

            As explained in \cref{subsec:notations}, the pruning function $\prune$ decides whether to keep the box in the final set of predictions $\boxesAfterNMS$ or not based on \iouTwoD~overlaps, \thatIs, $\pruneof{\overlap_i} = 1$ denotes the box $\boxi$ is suppressed while $\pruneof{\overlap_i} = 0$ denotes $\boxi$ is kept in $\boxesAfterNMS$. 
            
            \classicalNmsCaps~uses the threshold as the pruning function, which does not give useful gradients. 
            Therefore, we  considered three different functions for $\prune$: \basic, a temperature $(\temperature)$-controlled \exponentialPruning, and Sigmoidal function.
            \begin{compactitem}
                \item \textbf{Linear} Linear pruning function~\cite{bodla2017soft} is $\prune(\overlap)=\overlap$.
                \item \textbf{\exponentialPruning} \exponentialPruning~pruning function~\cite{bodla2017soft} is $\prune(\overlap) = 1- \mathrm{exp}\autoBraces{-\frac{\overlap^2}{\temperature}}$.
                \item \textbf{Sigmoidal} Sigmoidal pruning function is $\prune(\overlap) = \sigma\autoBraces{\frac{\overlap-\nmsThresh}{\temperature}}$ with $\sigma$ denoting the standard sigmoid. Sigmoidal function appears as the binary cross entropy relaxation of the subset selection problem~\cite{paulus2020gradient}.
            \end{compactitem}
            
\begin{figure}[!tb]
                \centering
                \includegraphics[width=0.99\linewidth]{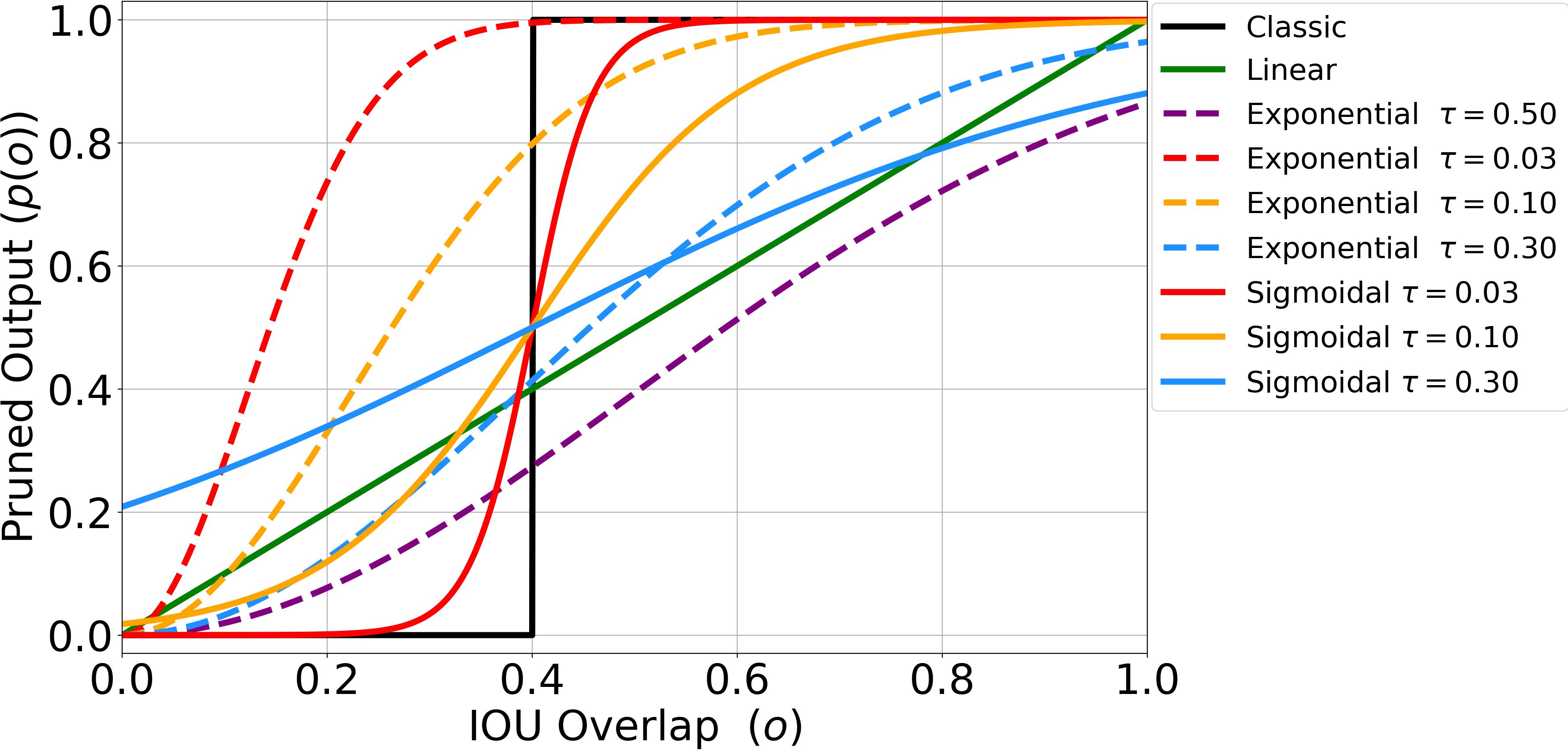}
                \vspace{1mm}
                \caption[Pruning functions of the \classicalNmsShort~and \groomedNMS.]{\textbf{Pruning functions} $\prune$ of the \classicalNmsShort~and \groomedNMS. We use the Linear and Exponential pruning of the \softNmsCaps~\cite{bodla2017soft} while training with the \groomedNMS.}
                \label{fig:pruning}
            \end{figure}

            We show these pruning functions in \cref{fig:pruning}. 
            The ablation studies (\cref{sec:results_ablation}) show that choosing $\prune$ as \basic~yields the simplest and the best \groomedNMS.

    \subsection{Differences from Existing NMS}\label{sec:difference}

        Although no differentiable NMS has been proposed for the monocular \threeD object detection, we compare our \groomedNMS with the NMS proposed for \twoD~object detection,~\twoD~pedestrian detection, \twoD~salient object detection, and~\threeD object detection in \cref{tab:nms_overview}.
        No method described in \cref{tab:nms_overview} has a matrix-based closed-form mathematical expression of the NMS.
        ~\classicalNmsShortCaps, \softNmsShortCaps~\cite{bodla2017soft} and \distanceNmsCaps~\cite{shi2020distance} are used at the inference time, while \groomedNMS is used during both training and inference.
        \distanceNmsCaps~\cite{shi2020distance} updates the $z$-coordinate of the box after NMS as the weighted average of the $z$-coordinates of top-$\kappa$ boxes.
        QUBO-NMS~\cite{rujikietgumjorn2013optimized}, Point-NMS~\cite{lee2016individualness, some2020determinantal}, and MAP-NMS~\cite{zhang2016unconstrained} are not used in end-to-end training.
        \cite{azadi2017learning} proposes a trainable Point-NMS.
        The Structured-SVM based NMS~\cite{desai2011discriminative, wan2015end} rely on structured SVM to obtain the rescores. 
        Adaptive-NMS\cite{liu2019adaptive} uses a separate neural network to predict the \classicalNms~threshold~$\nmsThresh$.
        The trainable neural network based NMS (NN-NMS)~\cite{hosang2016convnet, prokudin2017learning, hosang2017learning} use a separate neural network containing multiple layers and/or message-passing to approximate the NMS and do not use the pruning function.
        Unlike these methods, \groomedNMS uses a single layer and does not require multiple layers or message passing. Our NMS is parallel up to group (denoted by $\groups$). 
        However, $\numberElem{\groups}$ is, in general, $<< \numberElem{\boxes}$ in the NMS. 

        \begin{table}[!tb]
            \caption[Comparison of different NMS.]{\textbf{Comparison of different NMS}. [Key: Train= End-to-end Trainable, Prune= Pruning function, \#Layers= Number of layers, Par= Parallelizable]}
            \label{tab:nms_overview}
            \setlength{\tabcolsep}{0.03cm}
            \centering
            \begin{tabular}{lcccccc}
                \myTopRule
                \textbf{NMS}                                            & \textbf{Train} & \textbf{Rescore} & \textbf{Prune}  & \textbf{\#Layers} & \textbf{Par} & \\
                \myTopRule
                \classicalNmsShortCaps~                                       & \xmark         & \xmark           & Hard            & -          & $\bigO{\numberElem{\groups}}$ \\
                \softNmsCaps~\cite{bodla2017soft}                             & \xmark         & \xmark           & Soft            & -          & $\bigO{\numberElem{\groups}}$ \\
                \distanceNmsCaps~\cite{shi2020distance}                       & \xmark         & \xmark           & Hard            & -          & $\bigO{\numberElem{\groups}}$ \\
                QUBO-NMS~\cite{rujikietgumjorn2013optimized}                  & \xmark         & Optimization     & \xmark          & -          & -\\
                Point-NMS~\cite{lee2016individualness, some2020determinantal} & \xmark         & Point Process    & \xmark          & -          & -\\
                Trainable Point-NMS~\cite{azadi2017learning}                  & \cmark         & Point Process    & \xmark          & -          & -\\
                MAP-NMS~\cite{zhang2016unconstrained}                         & \xmark         & MAP              & \xmark          & -          & -\\
                Structured-NMS~\cite{desai2011discriminative, wan2015end}     & \xmark         & SSVM             & \xmark          & -          & -\\
                Adaptive-NMS~\cite{liu2019adaptive}                           & \xmark         & \xmark           & Hard            & $\!>\!1$  & $\bigO{\numberElem{\groups}}$\\
                NN-NMS~\cite{hosang2016convnet, prokudin2017learning, hosang2017learning} & \cmark & Neural Network    & \xmark      & $\!>\!1$ & $\bigO{1}$ \\
                \bottomrule
                \!\groomedNMS (Ours)                                          & \cmark         & Matrix           & Soft           & $1$         & $\bigO{\numberElem{\groups}}$ \\
                \myTopRule
            \end{tabular}
        \end{table}

    \subsection{Target Assignment and Loss Function}\label{sec:target_loss}
    
        \noindent\textbf{Target Assignment.}
            Our method consists of M3D-RPN~\cite{brazil2019m3d} and uses binning and self-balancing confidence~\cite{brazil2020kinematic}. The boxes' self-balancing confidence are used as scores $\score$, which pass through the \groomedNMS layer to obtain the rescores $\rescore$. The rescores signal the network if the \emph{best} box has not been selected for a particular object. 
            
            We extend the notion of the best \twoD~box~\cite{prokudin2017learning} to \threeD. The best box has the highest product of \iouTwoD~and \giouThreeD~\cite{rezatofighi2019generalized} with ground truth $\groundTruthj$. If the product is greater than a certain threshold $\beta$, it is assigned a positive label. Mathematically,
            \begin{align}
                \text{target}(\boxi) &= \left\{\begin{array}{@{}ll@{}}
                    \multirow{2}{*}{$1$,} & \text{if}~\exists~\groundTruthj~\text{st}~i=\argmax q(\boxj, \groundTruthj)\\ 
                                          & \phantom{\text{if}~\exists~\groundTruthj~s.~t.~i= }\text{and}~q(\boxi, \groundTruthj) \ge \beta\\
                    0, & \text{otherwise}
                    \end{array}\right.
            \end{align}
            with $q(\boxj, \groundTruthj) = \text{\iouTwoD}(\boxj,\groundTruthj)~\left(\frac{1+\text{\giouThreeD}(\boxj,\groundTruthj)}{2}\right)$. 
            \giouThreeD~is known to provide signal even for non-intersecting boxes~\cite{rezatofighi2019generalized}, where the usual \iouThreeD~is always zero. 
            Therefore, we use \giouThreeD~instead of regular \iouThreeD~for figuring out the best box in \threeD as many \threeD boxes have a zero \iouThreeD~overlap with the ground truth.
            For calculating \giouThreeD, we first calculate the volume $\vol$ and hull volume $\vol_{hull}$ of the \threeD boxes.  $\vol_{hull}$ is the product of \giouTwoD~in Birds Eye View (BEV), removing the rotations and hull of the $Y$ dimension. \giouThreeD~is then given by
            \begin{align}
                \text{\giouThreeD}(\boxi,\boxj) &= \frac{\vol(\boxi \cap \boxj)}{\vol(\boxi \cup \boxj)} + \frac{\vol(\boxi \cup \boxj)}{\vol_{hull}(\boxi, \boxj)} - 1.
            \end{align} 
            
        \noindent\textbf{Loss Function.}
            Generally the number of best boxes is less than the number of ground truths in an image, as there could be some ground truth boxes for which no box is predicted. The tiny number of best boxes introduces a far-heavier skew than the foreground-background classification. Thus, we use the modified \aploss~\cite{chen2020ap} as our loss after NMS since \aploss~does not suffer from class imbalance~\cite{chen2020ap}. 
            
            Vanilla \aploss~treats boxes of all images in a mini-batch equally, and the gradients are back-propagated through all the boxes. We remove this condition and rank boxes in an image-wise manner. In other words, if the best boxes are correctly ranked in one image and are not in the second, then the gradients only affect the boxes of the second image. We call this modification of \aploss~
            as \emph{\imageWise} \aploss. In other words,
            \vspace{-0.15cm}
            \begin{align}
                \loss_{\imageWise} = \frac{1}{N}\sum_{m=1}^N \apMath(\rescore^{(m)}, \text{target}(\boxes^{(m)})),
            \end{align}
            where $\rescore^{(m)}$ and $\boxes^{(m)}$ denote the rescores and the boxes of the $m^{\text{th}}$ image in a mini-batch respectively. This is different from previous NMS approaches~\cite{hosang2016convnet, henderson2016end, prokudin2017learning, hosang2017learning}, which use classification losses. Our ablation studies (\cref{sec:results_ablation}) show that the \imageWise~\aploss~is better suited to be used after NMS than the classification loss.
    
            Our overall loss function is thus given by $\loss = \lossBefore + \lossWeigh \lossAfter$ where $\lossBefore$ denotes the losses before the NMS including classification, \twoD~and \threeD regression as well as confidence losses, and $\lossAfter$ denotes the loss term after the NMS, which is the \imageWise~\aploss~with $\lossWeigh$ being the weight. 
            See \cref{sec:appendix_loss} of the supplementary material for more details of the loss function.

\section{Experiments}\label{sec:experiments}

    Our experiments use the most widely used \kitti autonomous driving dataset~\cite{geiger2012we}. 
    We modify the publicly-available PyTorch~\cite{paszke2019pytorch} code of Kinematic-3D~\cite{brazil2020kinematic}.
    \cite{brazil2020kinematic} uses DenseNet-121~\cite{huang2017densely} trained on ImageNet as the backbone and $n_h\!=\!1{,}024$ using \threeD-RPN settings of~\cite{brazil2019m3d}. 
    As \cite{brazil2020kinematic} is a video-based method while \groomedNMS is an image-based method, we use the best image model of~\cite{brazil2020kinematic} henceforth called~\kinematicImage~as our baseline for a fair comparison.
    \kinematicImage~is built on M3D-RPN~\cite{brazil2019m3d} and uses binning and self-balancing confidence. 

    \noindent\textbf{Data Splits.}
        There are three commonly used data splits of the \kitti dataset; we evaluate our method on all three.
    
        \textit{\kitti Test (Full) split}: Official \kitti \threeD benchmark~\cite{kitti2012benchmark}
        consists of $7{,}481$ training and $7{,}518$ testing images~\cite{geiger2012we}.
        
        \textit{\kitti \valOne split}: It partitions the $7{,}481$ training images into $3{,}712$ training and $3{,}769$ validation images~\cite{chen20153d,simonelli2020disentangling, brazil2020kinematic}.
        
        \textit{\kitti \valTwo split}: It partitions the $7{,}481$ training images into $3{,}682$ training and $3{,}799$ validation images~\cite{xiang2017subcategory}.
    
    \noindent\textbf{Training.}
        Training is done in two phases - warmup and full~\cite{brazil2020kinematic}. 
        We initialize the model with the confidence prediction branch from warmup weights and finetune using the self-balancing loss~\cite{brazil2020kinematic} and \imageWise~\aploss~\cite{chen2020ap} after our \groomedNMS. 
        See \cref{sec:training_additional} of the supplementary material for more training details. 
        We keep the weight $\lossWeigh$ at $0.05$. 
        Unless otherwise stated, we use $\prune$ as the \basic~function (this does not require $\temperature$) with $\groupSize=100$. 
        $\nmsThresh, \validBoxThresh$ and $\beta$ are set to $0.4$~\cite{brazil2019m3d, brazil2020kinematic}, $0.3$ and $0.3$ respectively.

    \noindent\textbf{Inference.}
        We multiply the class and predicted confidence to get the box's overall score in inference as in~\cite{tychsen2018improving, shi2020distance, kim2020probabilistic}. 
        See \cref{sec:results_kitti_val1} for training and inference times.

        \begin{table}[!tb]
            \caption[\kitti Test cars detection results.]{\textbf{\kitti Test cars} \apThreeDForty~and \apBevForty~comparisons (\iouThreeD~$\geq 0.7$). Previous results are quoted from the official leader-board or from papers.[Key: \firstKey{Best}, \secondKey{Second Best}].
            }
            \label{tab:results_kitti_test}
            \centering
            \setlength\tabcolsep{2.00pt}
            \begin{tabular}{tl m ccc  m ccct}
                \myTopRule
                \addlinespace[0.01cm]
                \multirow{2}{*}{Method} & \multicolumn{3}{cm}{\apThreeDForty ($\uparrowRHDSmall$)} & \multicolumn{3}{ct}{\apBevForty ($\uparrowRHDSmall$)}\\ 
                & Easy & Mod & Hard & Easy & Mod & Hard\\ 
                \myTopRule
                FQNet~\cite{liu2019deep}                   &   $2.77$       &   $1.51$       &   $1.01$      &   $5.40$       &   $3.23$       &   $2.46$      \\
                ROI-10D~\cite{manhardt2019roi}             &   $4.32$       &   $2.02$       &   $1.46$      &   $9.78$       &   $4.91$       &   $3.74$      \\
                GS3D~\cite{li2019gs3d}                     &   $4.47$       &   $2.90$       &   $2.47$      &   $8.41$       &   $6.08$       &   $4.94$      \\
                MonoGRNet~\cite{qin2019monogrnet}          &   $9.61$       &   $5.74$       &   $4.25$      & $18.19$        &   $11.17$      &   $8.73$      \\
                MonoPSR~\cite{ku2019monocular}             &  $10.76$       &   $7.25$       &   $5.85$      & $18.33$        & $12.58$        &   $9.91$      \\
                MonoDIS~\cite{simonelli2019disentangling}  &  $10.37$       &   $7.94$       &   $6.40$      & $17.23$        & $13.19$        & $11.12$       \\
                UR3D~\cite{shi2020distance}                &  $15.58$       &   $8.61$       &   $6.00$      & $21.85$        & $12.51$        & $9.20$        \\
                M3D-RPN~\cite{brazil2019m3d}               &  $14.76$       &   $9.71$       &   $7.42$      & $21.02$        & $13.67$        & $10.23$       \\
                SMOKE~\cite{liu2020smoke}                  & $14.03$        &   $9.76$       & $7.84$        & $20.83$        & $14.49$        & $12.75$       \\
                MonoPair~\cite{chen2020monopair}           & $13.04$        &   $9.99$       & $8.65$        & $19.28$        & $14.83$        & $12.89$       \\
                RTM3D~\cite{li2020rtm3d}                   & $14.41$        &   $10.34$      & $8.77$        & $19.17$        & $14.20$        & $11.99$       \\
                AM3D~\cite{ma2019accurate}                 & $16.50$        &   $10.74$      & \second{9.52}  & $25.03$        & 17.32        & \first{14.91} \\
                MoVi-3D~\cite{simonelli2020towards}        & $15.19$        &   $10.90$      & $9.26$        & $22.76$        & $17.03$        & $10.86$       \\
                RAR-Net~\cite{liu2020reinforced}           & $16.37$        &   $11.01$      & \second{9.52} & $22.45$        & $15.02$        & $12.93$       \\
                M3D-SSD~\cite{luo2021m3dssd}               & $17.51$        &   $11.46$      & 8.98          & $24.15$        & $15.93$        & $12.11$       \\
                DA-3Ddet~\cite{ye2020monocular}            & $16.77$        &   $11.50$      & 8.93          & -              & -              & -             \\
                D4LCN~\cite{ding2020learning}              & $16.65$        & $11.72$        & 9.51          & $22.51$        & $16.02$        &  $12.55$      \\
                \kinematicVideo~\cite{brazil2020kinematic} & \first{19.07}  & \first{12.72}  & $9.17$        & \first{26.69}  & \second{17.52}  & 13.10\\
                \hline
                \groomedNMS (Ours) & \second{18.10} & \second{12.32} & \first{9.65} & \second{26.19} & \first{18.27} & \second{14.05}\\
                \myTopRule
            \end{tabular}
        \end{table}    
    
    \noindent\textbf{Evaluation Metrics.} 
        \kitti uses \apThreeDForty~metric to evaluate object detection following~\cite{simonelli2019disentangling, simonelli2020disentangling}.
        \kitti benchmark evaluates on three object categories: Easy, Moderate and Hard. 
        It assigns each object to a category based on its occlusion, truncation, and height in the image space. 
        The \apThreeDForty~performance on the Moderate category compares different models in the benchmark~\cite{geiger2012we}. 
        We focus primarily on the Car class following~\cite{brazil2020kinematic}.

    \subsection{\kitti Test \monoThreeD}\label{sec:results_kitti_test}
        \cref{tab:results_kitti_test} summarizes the results of \threeD object detection and BEV evaluation on \kitti Test Split. The results in \cref{tab:results_kitti_test} show that \groomedNMS outperforms the baseline M3D-RPN~\cite{brazil2019m3d} by a significant margin and several other \sota~methods on both the tasks.~\groomedNMS also outperforms augmentation based approach MoVi-3D~\cite{simonelli2020towards} and depth-convolution based D4LCN~\cite{ding2020learning}. 
        Despite being an image-based method, \groomedNMS performs competitively to the video-based method \kinematicVideo~\cite{brazil2020kinematic}, outperforming it on the most-challenging Hard set.
        \begin{table}[t]
            \caption[\kitti \valOne~cars detection results.]{\textbf{\kitti \valOne~cars} \apThreeDForty and \apBevForty~results. [Key: \firstKey{Best}, \secondKey{Second Best}].
            }
            \label{tab:results_kitti_val1}
            \centering
            \scalebox{0.85}{
            \setlength\tabcolsep{0.1cm}
            \begin{tabular}{tl m ccc t ccc m ccc t ccct}
                \myTopRule
                \addlinespace[0.01cm]
                \multirow{3}{*}{Method} & \multicolumn{6}{cm}{\iouThreeD~$\geq 0.7$} & \multicolumn{6}{ct}{\iouThreeD~$\geq 0.5$}\\\cline{2-13}
                & \multicolumn{3}{ct}{\apThreeDForty ($\uparrowRHDSmall$)} & \multicolumn{3}{cm}{\apBevForty ($\uparrowRHDSmall$)} & \multicolumn{3}{ct}{\apThreeDForty ($\uparrowRHDSmall$)} & \multicolumn{3}{ct}{\apBevForty ($\uparrowRHDSmall$)}\\
                & Easy & Mod & Hard & Easy & Mod & Hard & Easy & Mod & Hard & Easy & Mod & Hard\\ 
                \myTopRule
                MonoDR~\cite{beker2020monocular}                                 & $12.50$        & $7.34$         & $4.98$        & $19.49$        & $11.51$        & $8.72$       & -              & -              & -              & -              & -              & -             \\
                MonoGRNet~\cite{qin2019monogrnet} in~\cite{chen2020monopair}             & $11.90$        & $7.56$         & $5.76$        & $19.72$        & $12.81$        & $10.15$       & $47.59$        & $32.28$        & $25.50$        & $52.13$        & $35.99$        & $28.72$\\
                MonoDIS~\cite{simonelli2019disentangling} in~\cite{simonelli2020disentangling} & $11.06$        & $7.60$         & $6.37$        & $18.45$        & $12.58$        & $10.66$       & -              & -              & -              & -              & -              & -             \\
                M3D-RPN~\cite{brazil2019m3d} in~\cite{brazil2020kinematic}               & $14.53$        & $11.07$        & $8.65$        & $20.85$        & $15.62$        & $11.88$       & $48.56$        & $35.94$        & $28.59$        & $53.35$        & $39.60$        & $31.77$       \\
                MoVi-3D~\cite{simonelli2020towards}                                          & $14.28$        & $11.13$        & $9.68$        & $22.36$        & $17.87$        & $15.73$       & -              & -              & -              & -              & -              & -             \\
                MonoPair~\cite{chen2020monopair}                                         & $16.28$        & $12.30$        & $10.42$       & $24.12$        & $18.17$        & \second{15.76}& $55.38$        & \first{42.39}  & \first{37.99}  & $61.06$        & \first{47.63}        & \first{41.92}       \\
                \kinematicImage~\cite{brazil2020kinematic}                              & $18.28$        & $13.55$        & $10.13$       & $25.72$        & $18.82$        & $14.48$       & $54.70$        & $39.33$        & $31.25$        & $60.87$        & $44.36$        & $34.48$       \\
                \kinematicVideo~\cite{brazil2020kinematic}                             & \first{19.76}  & \second{14.10} & \second{10.47} &  \first{27.83} & \second{19.72} & $15.10$       & \second{55.44} & $39.47$        & $31.26$        & \second{61.79} & $44.68$        & $34.56$\\
                \hline
                \groomedNMS (Ours)                                                       & \second{19.67} & \first{14.32}  & \first{11.27} & \second{27.38} & \first{19.75}  & \first{15.92} & \first {55.62} & \second{41.07} & \second {32.89}& \first {61.83} & \second{44.98} & \second{36.29}\\
                \myTopRule
            \end{tabular}
            }
        \end{table}            

        \begin{figure}[!tb]
            \centering
            \begin{subfigure}{.5\linewidth}
              \centering
              \includegraphics[width=\linewidth]{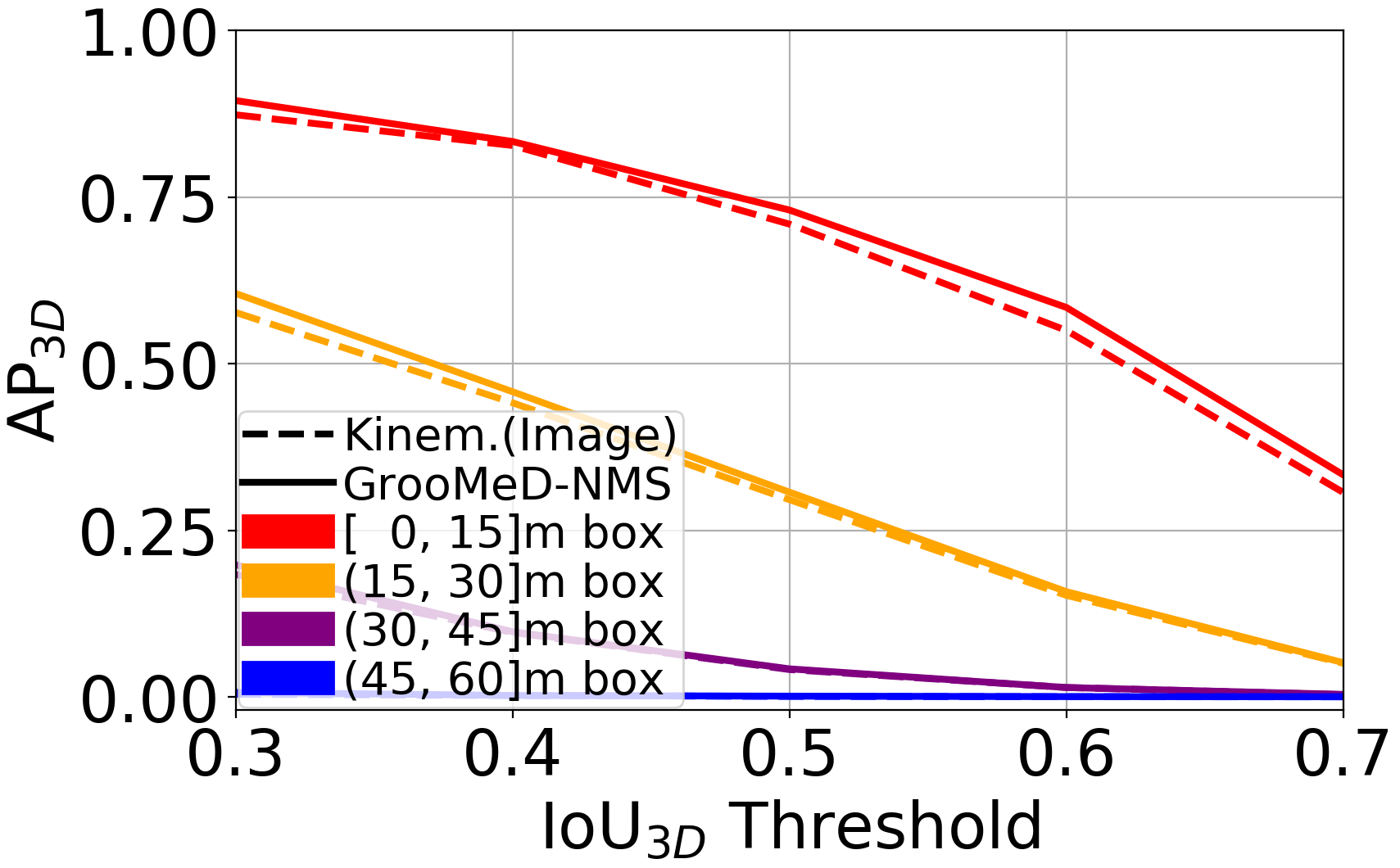}
              \caption{Linear Scale}
            \end{subfigure}%
            \begin{subfigure}{.5\linewidth}
              \centering
              \includegraphics[width=\linewidth]{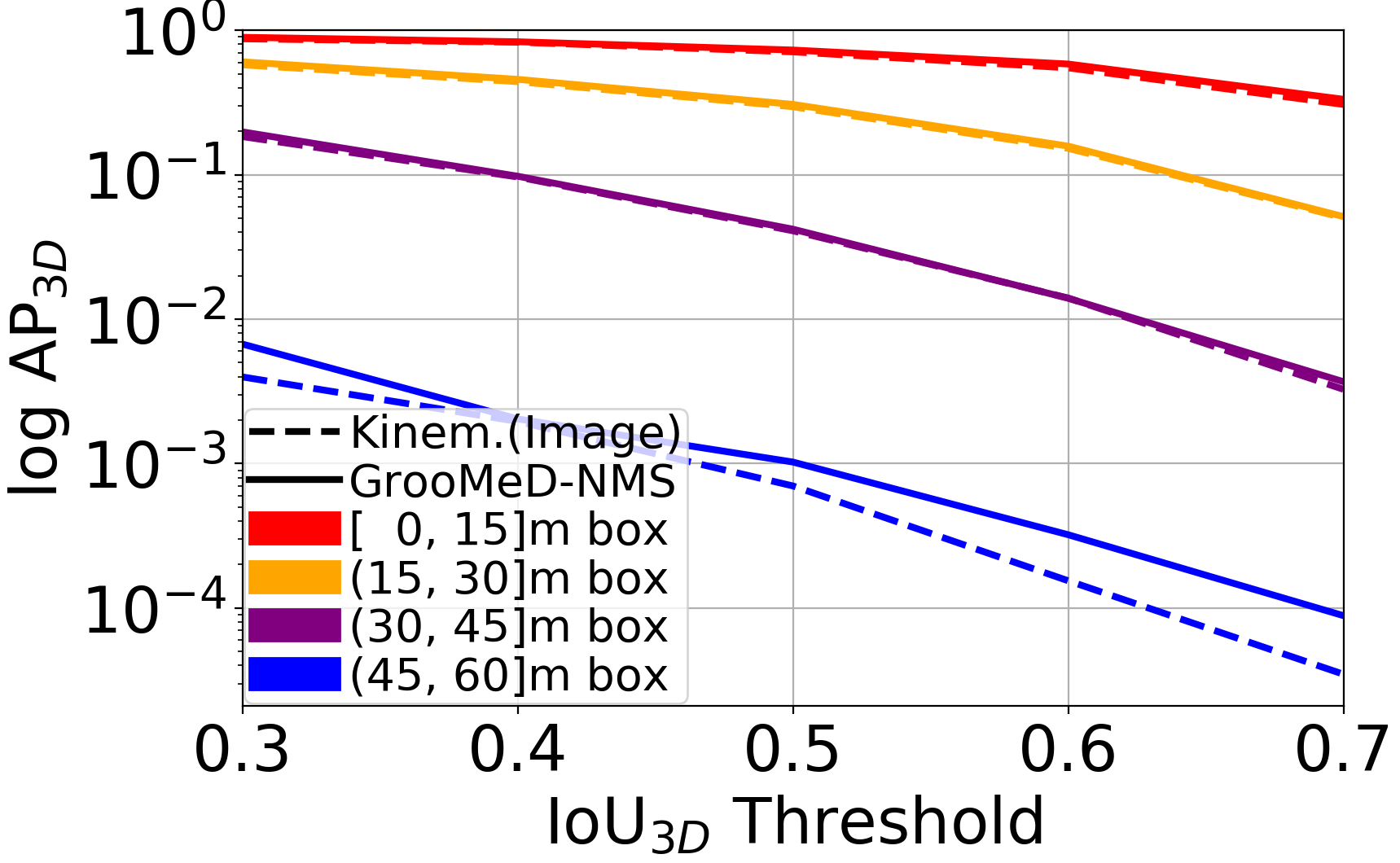}
              \caption{Log Scale}
            \end{subfigure}
            \caption[\apThreeD Comparison at different depths and \iouThreeD~matching thresholds.]{\textbf{\apThreeD Comparison at different depths and \iouThreeD~matching thresholds} on \kitti \valOne~Split.}
            \label{fig:ap_ground_truth_threshold}
        \end{figure}
    
        \begin{table}[t]
            \caption[Comparisons with other NMS on \kitti \valOne~cars.]{\textbf{Comparisons with other NMS} on \kitti \valOne~cars (\iouThreeD~$\geq 0.7$). [Key: C= \classicalNmsShortCaps, S= \softNmsCaps\cite{bodla2017soft}, D= \distanceNmsCaps\cite{shi2020distance}, G= \groomedNMS]}
            \label{tab:results_kitti_val1_other_nms}
            \centering
            \setlength\tabcolsep{0.075cm}
            \begin{tabular}{tl t c m ccc t ccct}
                \myTopRule
                \addlinespace[0.01cm]
                \multirow{2}{*}{Method}& \multirow{2}{*}{\shortstack{Infer\\NMS}} & \multicolumn{3}{ct}{\apThreeDForty ($\uparrowRHDSmall$)} & \multicolumn{3}{ct}{\apBevForty ($\uparrowRHDSmall$)}\\
                & & Easy & Mod & Hard & Easy & Mod & Hard\\ 
                \myTopRule
                \kinematicImage      &C& $18.28$        & $13.55$        & $10.13$       & $25.72$        & $18.82$        & $14.48$\\
                \kinematicImage      &S& $18.29$        & $13.55$        & $10.13$       & $25.71$        & $18.81$        & $14.48$\\
                \kinematicImage      &D& $18.25$        & $13.53$        & $10.11$       & $25.71$        & $18.82$        & $14.48$\\
                \kinematicImage      &G& $18.26$        & $13.51$        & $10.10$       & $25.67$        & $18.77$        & $14.44$\\
                \hline
                \groomedNMS          &C& $19.67$ & $14.31$  & $11.27$ & $27.38$ & $19.75$  & $15.93$\\
                \groomedNMS          &S& $19.67$ & $14.31$  & $11.27$ & $27.38$ & $19.75$  & $15.93$\\
                \groomedNMS          &D& $19.67$ & $14.31$  & $11.27$ & $27.38$ & $19.75$  & $15.93$\\
                \groomedNMS          &G& $19.67$ & $14.32$  & $11.27$ & $27.38$ & $19.75$  & $15.92$\\
                \myTopRule
            \end{tabular}
        \end{table}

    \subsection{\kitti \valOne \monoThreeD}\label{sec:results_kitti_val1}
        
        \noindent\textbf{Results.}
            \cref{tab:results_kitti_val1} summarizes the results of \threeD object detection and BEV evaluation on \kitti \valOne~Split at two \iouThreeD~thresholds of $0.7$ and $0.5$~\cite{chen2020monopair,brazil2020kinematic}. 
            \cref{tab:results_kitti_val1} results show that \groomedNMS outperforms the baseline of M3D-RPN~\cite{brazil2019m3d} and \kinematicImage~\cite{brazil2020kinematic} by a significant margin. 
            Interestingly, \groomedNMS (an image-based method) also outperforms the video-based method \kinematicVideo~\cite{brazil2020kinematic} on most of the metrics. 
            Thus, \groomedNMS performs best on $6$ out of the $12$ cases ($3$ categories $\times~2$~tasks $\times~2$~thresholds) while second-best on all other cases. 
            The performance is especially impressive since the biggest improvements are shown on the Moderate and Hard set, where objects are more distant and occluded.

        \noindent\textbf{\apThreeD at different depths and \iouThreeD~thresholds.}
            We next compare the \apThreeD performance of \groomedNMS and \kinematicImage~on linear and log scale for objects at different depths of $[15,30, 45, 60]$ meters and \iouThreeD~matching criteria of $0.3\!\rightarrowRHD\!0.7$ in \cref{fig:ap_ground_truth_threshold} as in~\cite{brazil2020kinematic}. 
            \cref{fig:ap_ground_truth_threshold} shows that~\groomedNMS outperforms the \kinematicImage~\cite{brazil2020kinematic} at all depths and all \iouThreeD~thresholds.
    
        \noindent\textbf{Comparisons with other NMS.}
            We compare with the 
            \classicalNms, \softNmsCaps~\cite{bodla2017soft} and \distanceNmsCaps~\cite{shi2020distance} in \cref{tab:results_kitti_val1_other_nms}.
            More detailed results are in \cref{tab:results_kitti_val1_other_nms_detailed} of the supplementary material.
            The results show that NMS inclusion in the training pipeline benefits the performance, unlike~\cite{bodla2017soft}, which suggests otherwise. 
            Training with \groomedNMS helps because the network gets an additional signal through the \groomedNMS layer whenever the best-localized box corresponding to an object is not selected. 
            Interestingly, \cref{tab:results_kitti_val1_other_nms} also suggests that replacing \groomedNMS with the~\classicalNms~in inference does not affect the performance.

        \begin{figure}[t]
            \centering
            \includegraphics[width=0.67\linewidth]{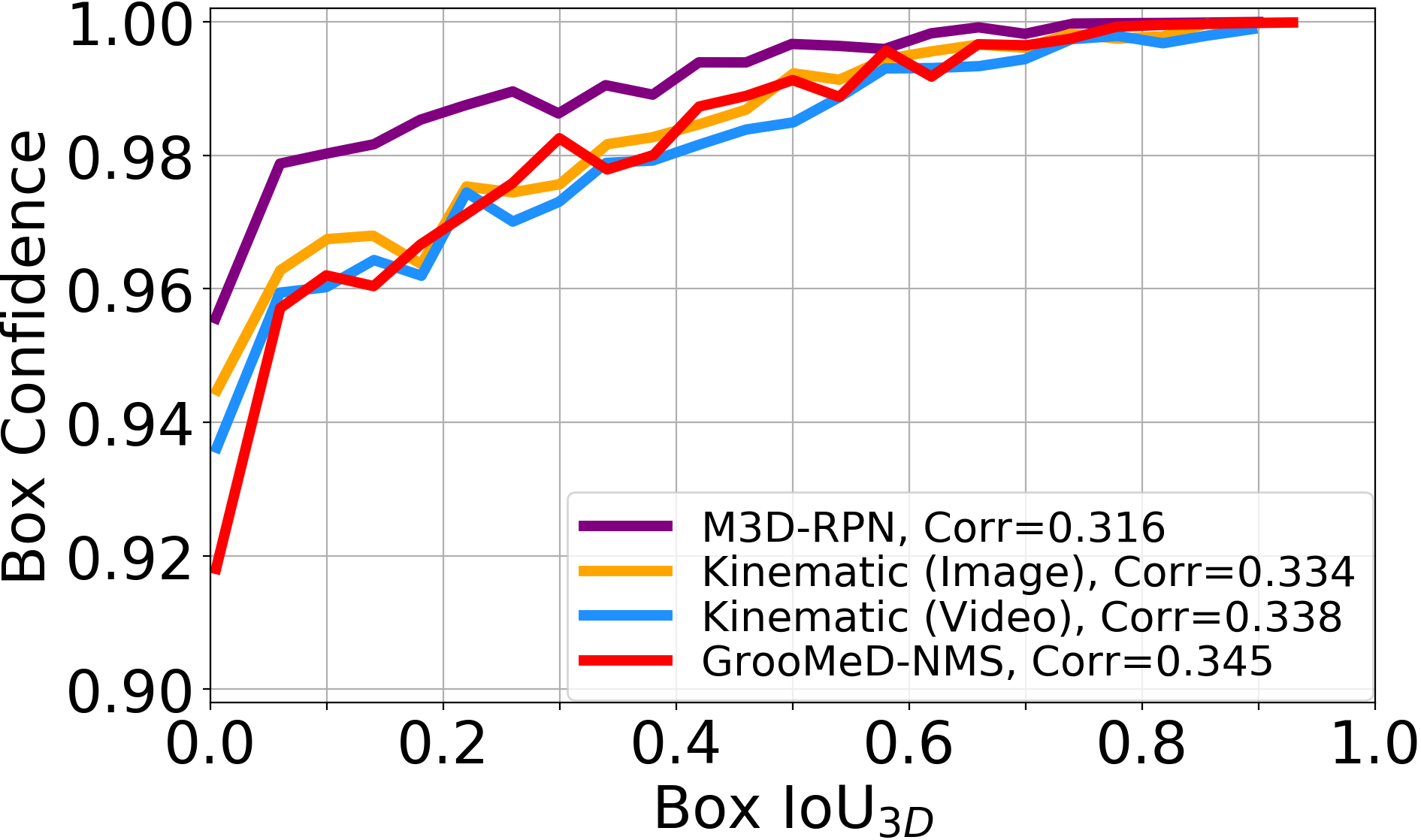}
            \caption{\textbf{Score-\iouThreeD~plot} after the NMS. \groomedNMS achieves the best correlation.}
            \label{fig:score_iou}
        \end{figure}
        
        \noindent\textbf{Score-\iouThreeD~Plot.}
            We further correlate the scores with \iouThreeD~after NMS of our~model with two baselines - M3D-RPN~\cite{brazil2019m3d} and \kinematicImage~\cite{brazil2020kinematic} and also the \kinematicVideo\cite{brazil2020kinematic}~in \cref{fig:score_iou}. 
            We obtain the best correlation of $0.345$ exceeding the correlations of M3D-RPN, \kinematicImage~and, also \kinematicVideo. This proves that including NMS in the training pipeline is beneficial.

        \noindent\textbf{Training and Inference Times.}
            We now compare the training and inference times of including~\groomedNMS in the pipeline. 
            Warmup training phase takes about $13$ hours to train on a single $12$ GB GeForce GTX Titan-X GPU. 
            Full training phase of \kinematicImage~and~\groomedNMS takes about $8$ and $8.5$ hours respectively. 
            The inference time per image using \classicalNmsShort~and \groomedNMS is $0.12$ and $0.15$ ms respectively. 
            \cref{tab:results_kitti_val1_other_nms} suggests that changing the NMS from \groomedNMSShort~to \classicalNmsShort~during inference does not alter the performance. 
            Then, the inference time of our method is the same as $0.12$ ms.

        \begin{table}[!tb]
            \caption[\kitti \valTwo cars detection results.]{\textbf{\kitti \valTwo~cars} \apThreeDForty~and \apBevForty~comparisons. [Key: \firstKey{Best}, *= Released, $^\dagger$= Retrained].
            }
            \label{tab:results_kitti_val2}
            \centering
            \scalebox{\scaleFraction}{
            \setlength\tabcolsep{0.1cm}
            \begin{tabular}{tl m ccc t ccc m ccc t ccct}
                \myTopRule
                \addlinespace[0.01cm]
                \multirow{3}{*}{Method} & \multicolumn{6}{cm}{\iouThreeD~$\geq 0.7$} & \multicolumn{6}{ct}{\iouThreeD~$\geq 0.5$}\\\cline{2-13}
                & \multicolumn{3}{ct}{\apThreeDForty ($\uparrowRHDSmall$)} & \multicolumn{3}{cm}{\apBevForty ($\uparrowRHDSmall$)} & \multicolumn{3}{ct}{\apThreeDForty ($\uparrowRHDSmall$)} & \multicolumn{3}{ct}{\apBevForty ($\uparrowRHDSmall$)}\\
                & Easy & Mod & Hard & Easy & Mod & Hard & Easy & Mod & Hard & Easy & Mod & Hard\\ 
                \myTopRule
                M3D-RPN~\cite{brazil2019m3d}*               &$14.57$ &$10.07$ &$7.51$ &$21.36$ &$15.22$ &$11.28$ &$49.14$ &$34.43$ &$26.39$ &$53.44$ &$37.79$ &$29.36$\\
                \kinematicImage~\cite{brazil2020kinematic}$^\dagger$      &$13.54$        & $10.21$        & $7.24$        & $20.60$        & $15.14$        & $11.30$        & $51.53$        & $36.55$        & $28.26$        & $56.20$        & $40.02$        & $31.25$\\
                \hline
                \groomedNMS (Ours)                     & \first{14.72} & \first{10.87} & \first{7.67} & \first{22.03} & \first{16.05} & \first{11.93} & \first{51.91} & \first{36.78} & \first{28.40} & \first{56.29} & \first{40.31} & \first{31.39}
                \\
                \myTopRule
            \end{tabular}
            }
        \end{table}

    \subsection{\kitti \valTwo \monoThreeD}\label{sec:results_kitti_val2}
        \cref{tab:results_kitti_val2} summarizes the results of \threeD object detection and BEV evaluation on \kitti \valTwo~Split at two \iouThreeD~thresholds of $0.7$ and $0.5$~\cite{chen2020monopair,brazil2020kinematic}. 
        Again, we use M3D-RPN~\cite{brazil2019m3d} and Kinematic (Image)~\cite{brazil2020kinematic} as our baselines. 
        We evaluate the released model of M3D-RPN~\cite{brazil2019m3d} using the \kitti metric.~\cite{brazil2020kinematic} does not report \valTwo~results, so we retrain on \valTwo~using their public code. 
        The results in \cref{tab:results_kitti_val2} show that \groomedNMS performs best in all cases. 
        This is again impressive because the improvements are shown on Moderate and Hard set, consistent with Tabs.~\ref{tab:results_kitti_test} and~\ref{tab:results_kitti_val1}.    

        \begin{table}[!tb]
            \caption[Ablation studies on~\kitti \valOne~cars.]{\textbf{Ablation studies} of \groomedNMS on~\kitti \valOne~cars.}
            \label{tab:ablation}
            \centering
            \scalebox{0.7}{
            \setlength{\tabcolsep}{0.08cm}
            \begin{tabular}{tc|lm ccc t ccc m ccc t ccct}
                \myTopRule
                \addlinespace[0.01cm]
                \multicolumn{2}{tcm}{\textbf{Change from \groomedNMS model:}} & \multicolumn{6}{cm}{\iouThreeD~$\geq 0.7$} & \multicolumn{6}{ct}{\iouThreeD~$\geq 0.5$}\\\cline{1-14}
                \multirow{2}{*}{Changed} & \multirow{2}{*}{From $\longrightarrowRHD$To} & \multicolumn{3}{ct}{\apThreeDForty ($\uparrowRHDSmall$)} & \multicolumn{3}{cm}{\apBevForty ($\uparrowRHDSmall$)} & \multicolumn{3}{ct}{\apThreeDForty ($\uparrowRHDSmall$)} & \multicolumn{3}{ct}{\apBevForty ($\uparrowRHDSmall$)}\\
                && Easy & Mod & Hard & Easy & Mod & Hard & Easy & Mod & Hard & Easy & Mod & Hard\\
                \myTopRule
                \multirow{3}{*}{Training}
                & Conf+NMS$\rightarrowRHD$No Conf+No NMS              &$16.66$ &$12.10$ &$9.40$  & $23.15$ &$17.43$ &$13.48$ &$51.47$ &$38.58$ &$30.98$ &$56.48$ &$42.53$ &$34.37$\\
                & Conf+NMS$\rightarrowRHD$Conf+No NMS                 &$19.16$ &$13.89$ &$10.96$ &	$27.01$ &$19.33$ &$14.84$ &$57.12$ &$41.07	$ &$32.79$ &$61.60$ &$44.58$ &$35.97$\\
                & Conf+NMS$\rightarrowRHD$No Conf+NMS                 &$15.02$ &$11.21$ &$8.83$ &$21.07$ &$16.27$ &$12.77$ &$48.01$ &$36.18$ &$29.96$ &$53.82$ &$40.94$ &$33.35$\\%
                \hline
                Initialization
                & No Warmup                                         &$15.33$ &$11.68$ &$8.78$ &$21.32$ &$16.59$ &$12.93$ &$49.15$ &$37.42$ &$30.11$ &$54.32$ &$41.44$ &$33.48$\\%
                \hline
                \multirow{4}{*}{\shortstack{Pruning\\Function}}
                & \basic$\rightarrowRHD$\exponentialPruning, $\temperature=1$     &$12.81$ &$9.26$ &$7.10$ &$17.07$ &$12.17$ &$9.25$ &$29.58$ &$20.42$ &$15.88$ &$32.06$ &$22.16$ &$17.20$\\
                & \basic$\rightarrowRHD$\exponentialPruning, $\temperature=0.5$\cite{bodla2017soft}   &$18.63$ &$13.85$ &$10.98$ &$27.52	$ &$20.14$ &$15.76$ &$56.64$ &$41.01$ &$32.79$ &$61.43$ &$44.73$ &$36.02$\\
                & \basic$\rightarrowRHD$\exponentialPruning, $\temperature=0.1$   &$18.34$ &$13.79$ &$10.88$ &$27.26$ &$19.71$ &$15.90$ &$56.98$ &$41.16$ &$32.96$ &$62.77$ &$45.23$ &$36.56$\\
                &\basic$\rightarrowRHD$Sigmoidal, $\temperature=0.1$ &$17.40$ &$13.21$ &$9.80$ &$26.77$ &$19.26$ &$14.76$ &$55.15$ &$40.77$ &$32.63$ &$60.56$ &$44.23$ &$35.74$\\
                \hline 
                \multirow{2}{*}{Group+Mask}
                & Group+Mask$\rightarrowRHD$No Group              &$18.43$ &$13.91$ &$11.08$ &$26.53$ &$19.46$ &$15.83$ &$55.93$ &$40.98$ &$32.78$ &$61.02$ &$44.77$ &$36.09$\\%
                & Group+Mask$\rightarrowRHD$Group+No Mask        &$18.99$ &$13.74$ &$10.24$ &$26.71$ &$19.21$ &$14.77$ &$55.21$ &$40.69$ &$32.55$ &$61.74$ &$44.67$ &$36.00$    \\%
                \hline
                \multirow{2}{*}{Loss}
                & \imageWise~\ap$\rightarrowRHD$Vanilla~\ap          &$18.23$ &$13.73$ &$10.28$ &$26.42$ &$19.31$ &$14.76$ &$54.47$ &$40.35$ &$32.20$ &$60.90$ &$44.08$ &$35.47$    \\%
                & \imageWise~\ap$\rightarrowRHD$BCE                    &$16.34$ &$12.74$ &$ 9.73$ &$22.40$ &$17.46$ &$13.70$ &$52.46$ &$39.40$ &$31.68$ &$58.22$ &$43.60$ &$35.27$            \\%
                \hline
                Inference
                & Class*Pred$\rightarrowRHD$Class                   &$18.26$& $13.36$& $10.49$& $25.39$& $18.64$& $15.12$& $52.44$& $38.99$& $31.3$& $57.37$& $42.89$& $34.68$\\%
                NMS Scores
                & Class*Pred$\rightarrowRHD$Pred                    & $17.51$& $12.84$& $9.55$& $24.55$& $17.85$& $13.63$& $52.78$& $37.48$& $29.37$& $58.30$& $41.26$& $32.66$\\%
                \hline
                {---} & \textbf{\groomedNMS (best model)}     & $\mathbf{19.67}$ & $\mathbf{14.32}$  & $\mathbf{11.27}$ & $\mathbf{27.38}$ & $\mathbf{19.75}$  & $\mathbf{15.92}$ & $\mathbf {55.62}$ & $\mathbf{41.07}$ & $\mathbf {32.89}$& $\mathbf {61.83}$ & $\mathbf{44.98}$ & $\mathbf{36.29}$\\
                \myTopRule
            \end{tabular}
            }
        \end{table}
    \subsection{Ablation Studies on \kitti \valOne}\label{sec:results_ablation}
        \cref{tab:ablation} compares the modifications of our approach on \kitti \valOne~cars. 
        Unless stated otherwise, we stick with the experimental settings described in \cref{sec:experiments}. 
        Using a confidence head (Conf+No NMS) proves beneficial compared to the warmup model (No Conf+No NMS), which is consistent with the observations of \cite{shi2020distance, brazil2020kinematic}. 
        Further, \groomedNMS on classification scores (denoted by No Conf + NMS) is detrimental as the classification scores are not suited for localization \cite{huang2020epnet, brazil2020kinematic}. 
        Training the warmup model and then finetuning also works better than training without warmup as in~\cite{brazil2020kinematic} since the warmup phase allows \groomedNMS to carry meaningful grouping of the boxes. 
        
        As described in \cref{sec:pruning}, in addition to Linear, we compare two other functions for pruning function $\prune$: Exponential and Sigmoidal. 
        Both of them do not perform as well as the Linear $\prune$ possibly because they have vanishing gradients close to overlap of zero or one. Grouping and masking both help our model to reach a better minimum. 
        As described in \cref{sec:target_loss}, \imageWise~\ap~loss is better than the Vanilla \ap~loss since it treats boxes of two images differently. 
        \imageWise~\ap~also performs better than the binary cross-entropy (BCE) loss proposed in~\cite{hosang2016convnet, henderson2016end, prokudin2017learning, hosang2017learning}. 
        Using the product of self-balancing confidence and classification scores instead of using them individually as the scores to the NMS in inference is better, consistent with~\cite{tychsen2018improving, shi2020distance, kim2020probabilistic}. 
        Class confidence performs worse since it does not have the localization information while the self-balancing confidence (Pred) gives the localization without considering whether the box belongs to foreground or background.

\section{Conclusions}
    In this chapter, we present and integrate \groomedNMS -- a novel 
    Grouped Mathematically Differentiable NMS for monocular \threeD object detection, such that the network is trained end-to-end with a loss on the boxes after NMS. 
    We first formulate NMS as a matrix operation and then do unsupervised grouping and masking of the boxes to obtain a simple closed-form expression of the NMS. 
    \groomedNMS addresses the mismatch between training and inference pipelines and, therefore, forces the network to  select the best \threeD box in a differentiable manner. 
    As a result, \groomedNMS achieves state-of-the-art monocular \threeD object detection results on the \kitti benchmark dataset. 
    Although our implementation demonstrates monocular \threeD object detection, \groomedNMS is fairly generic for other object detection tasks. 
    Future work includes applying this method to tasks such as \lidar-based \threeD object detection and pedestrian detection.

    \noIndentHeading{Limitation.} \groomedNMS does not fully solve the generalization issue.

%% file: images/groomed/block_diagram_others.tex
\begin{tikzpicture}[scale=0.28, every node/.style={scale=0.50}, every edge/.style={scale=0.50}]
\tikzset{vertex/.style = {shape=circle, draw=black!70, line width=0.06em, minimum size=1.4em}}
\tikzset{edge/.style = {-{Triangle[angle=60:.06cm 1]},> = latex'}}

    \input{images/groomed/block_diagram_common.tex}
    \draw [draw=black, fill= black!70, line width=0.06em] (12.25,1.8) rectangle (15.75,-0.5) node[pos=0.5, scale= 1.25,align=center]{NMS};

    \draw [draw=red, line width=0.06em, very thick]         (15.5  , 3.5) rectangle (18 , 5.5   ) node[]{};
    \draw [draw=my_yellow_2, line width=0.06em, very thick] (19.0, 3.5) rectangle (19.5   , 5.25) node[]{};

    \draw [-{Triangle[angle=60:.1cm 1]}, draw=black!40, line width=0.15em, shorten <=0.5pt, shorten >=0.5pt, >=stealth]
       (15.75,1.2) node[]{}
    -- (20,1.2) node[pos=0.48, scale= 1.25, align=center]{Predictions\\};
    
    \draw [-{Triangle[angle=60:.1cm 1]}, draw=black!40, line width=0.15em, shorten <=0.5pt, shorten >=0.5pt, >=stealth]
       (15.75,0) node[]{}
    -- (20.1,0) node[pos=0.5, scale= 1.25, align=center]{$\rescore$\\};

    \draw [-{Triangle[angle=60:.1cm 1]}, draw=black!80, line width=0.1em, shorten <=0.5pt, shorten >=0.5pt, >=stealth]
       (0.5,6) node[]{}
    -- (10,6) node[pos= 0.8, align=center, scale= 1.25]{Training\\};
    \draw [draw=black!80, line width=0.15em, shorten <=0.5pt, shorten >=0.5pt, >=stealth]
       (10,6.5) node[]{}
    -- (10,5.5) node[pos= 0.8, align=center, scale= 1.25]{};

    \draw[draw=black, fill=my_red!30, thick]( 8.5,-4.2) circle (1.5) node[scale= 1.25]{$\lossBefore$};

\end{tikzpicture}

%% file: images/groomed/block_diagram_ours.tex
\begin{tikzpicture}[scale=0.28, every node/.style={scale=0.50}, every edge/.style={scale=0.50}]
\tikzset{vertex/.style = {shape=circle, draw=black!70, line width=0.06em, minimum size=1.4em}}
\tikzset{edge/.style = {-{Triangle[angle=60:.06cm 1]},> = latex'}}
    
    \input{images/groomed/block_diagram_common.tex}
    
    \draw [draw=black, fill= forward_color, line width=0.06em] (12.25,1.8) rectangle (15.75,-0.5) node[pos=0.5, scale= 1.25,align=center]{\groomedNMSShort\\NMS};
    
    \draw [draw=green!80, line width=0.06em,  densely dashed]         (16.1, 3.2) rectangle (18.65, 4.85) node[]{};
    \draw [draw=cyan, line width=0.06em,  densely dashed]             (15.9, 2.8) rectangle (18.4 , 5   ) node[]{};
    \draw [draw=orange!80, line width=0.06em, thick,  densely dashed] (15.7, 3.0) rectangle (18.2 , 5.25) node[]{};
    \draw [draw=red, line width=0.06em, very thick]                   (15.5, 3.5) rectangle (18.0 , 5.5 ) node[]{};
    
    \draw [draw=purple!80, line width=0.06em, thick,  densely dashed] (19.2, 3.0) rectangle (19.70, 5   ) node[]{};
    \draw [draw=my_yellow_2, line width=0.06em, very thick]           (19.0, 3.5) rectangle (19.5 , 5.25) node[]{};
    
    \draw [-{Triangle[angle=60:.1cm 1]}, draw=forward_color, line width=0.15em, shorten <=0.5pt, shorten >=0.5pt, >=stealth]
       (15.75,1.2) node[]{}
    -- (20,1.2) node[pos=0.48, scale= 1.25, align=center]{Predictions\\};
    \draw [-{Triangle[angle=60:.1cm 1]}, draw=forward_color, line width=0.15em, shorten <=0.5pt, shorten >=0.5pt, >=stealth]
       (15.75,0) node[]{}
    -- (20,0) node[pos=0.5, scale= 1.25, align=center]{$\rescore$\\};
    \draw[draw=forward_color, fill=forward_color, thick](17,0) circle (0.15) node[scale= 1.25]{};
    
    \draw [-{Triangle[angle=60:.1cm 1]}, draw=forward_color, line width=0.15em, shorten <=0.5pt, shorten >=0.5pt, >=stealth]
       (17, 0.0) node[]{}
    -- (17,-2.7) node[]{};

    \draw [-{Triangle[angle=60:.1cm 1]}, draw=black!80, line width=0.15em, shorten <=0.5pt, shorten >=0.5pt, >=stealth]
       (0.5,6) node[]{}
    -- (20,6) node[pos= 0.9, align=center, scale= 1.25]{Training\\};

    \draw[draw=black, fill=my_red!60, thick](17.,-4.2) circle (1.5) node[scale= 1.25]{$\lossAfter$};

\end{tikzpicture}

%% file: images/groomed/nms_inside_2.tex
\begin{tikzpicture}[scale=0.21, every node/.style={scale=0.39}, every edge/.style={scale=0.39}]
\tikzset{vertex/.style = {shape=circle, draw=black!70, line width=0.12em, minimum size=1.4em}}
\tikzset{edge/.style = {-{Triangle[angle=60:.06cm 1]},> = latex'}}
    
    \draw [-{Triangle[angle=60:.1cm 1]},draw=forward_color, line width=0.1em, shorten <=0.5pt, shorten >=0.5pt, >=stealth]
       (-1.5,5.0) node[]{}
    -- (0,5.0) node[pos= 0, scale= 1.5,align= center]{$\score$~~~~};
    \draw [,draw=forward_color, line width=0.1em, shorten <=0.5pt, shorten >=0.5pt, >=stealth]
       (-0.15,5.0) node[]{}
    -- (17.3,5.0) node[]{};
    \draw [-{Triangle[angle=60:.1cm 1]},draw=forward_color, line width=0.1em, shorten <=0.5pt, shorten >=0.5pt, >=stealth]
       (17.15,5.0) node[]{}
    -- (17.15,1.5) node[pos= 0.7, scale= 1.5]{$\score$~~~~~};

    \draw [-{Triangle[angle=60:.1cm 1]},draw=backward_color, line width=0.1em, shorten <=0.5pt, shorten >=0.5pt, >=stealth,  densely dashed]
       (17.7,5.4) node[]{}
    -- (-1.5,5.4) node[]{};
    \draw [,draw=backward_color, line width=0.1em, shorten <=0.5pt, shorten >=0.5pt, >=stealth, densely dashed]
       (17.55,5.4) node[]{}
    -- (17.55,1.5) node[]{};

    \draw [-{Triangle[angle=60:.1cm 1]},draw=forward_color, line width=0.1em, shorten <=0.5pt, shorten >=0.5pt, >=stealth]
       (-1.5,-4.80) node[]{}
    -- ( 0,-4.80) node[pos= 0, scale= 1.5]{$\overlapMat$~~~~~};
    \draw [-{Triangle[angle=60:.1cm 1]},draw=forward_color, line width=0.1em, shorten <=0.5pt, shorten >=0.5pt, >=stealth]
       (-0.15,-4.8) node[]{}
    -- (9,-4.8) node[]{};
    \draw[draw=forward_color, fill=forward_color, thick](3.7,-4.8) circle (0.2) node[]{};
    \draw [draw=forward_color, line width=0.1em, shorten <=0.5pt, shorten >=0.5pt, >=stealth]
       (3.7, 3.75) node[]{}
    -- (3.7,-4.8) node[]{};
    \draw [draw=forward_color, line width=0.1em, shorten <=0.5pt, shorten >=0.5pt, >=stealth]
       (3.70, 3.6) node[]{}
    -- (14.8, 3.6) node[]{};
    \draw [-{Triangle[angle=60:.1cm 1]},draw=forward_color, line width=0.1em, shorten <=0.5pt, shorten >=0.5pt, >=stealth]
       (14.65, 3.6) node[]{}
    -- (14.65, 1.5) node[]{};
    
    \draw [-{Triangle[angle=60:.1cm 1]}, draw=backward_color, line width=0.1em, shorten <=0.5pt, shorten >=0.5pt, >=stealth, densely dashed ]
       (3.45,-4.4) node[]{}
    -- (-1.5,-4.4) node[]{};
    \draw [, draw=backward_color, line width=0.1em, shorten <=0.5pt, shorten >=0.5pt, >=stealth, densely dashed]
       (3.3, 4.15) node[]{}
    -- (3.3,-4.4) node[]{};
    \draw [, draw=backward_color, line width=0.1em, shorten <=0.5pt, shorten >=0.5pt, >=stealth, densely dashed]
       (3.3, 4.0) node[]{}
    -- (15.2, 4.0) node[]{};
    \draw [draw=backward_color, line width=0.1em, shorten <=0.5pt, shorten >=0.5pt, >=stealth, densely dashed]
       (15.05, 4.1) node[]{}
    -- (15.05, 1.5) node[]{};

    \draw [draw=black,  fill= yellow!30, line width=0.06em] (0.5, 5.8) rectangle (2.5, -5.8   ) node[pos= 0.5, scale= 1.5, rotate= 00, align=center]{Sort};

    \draw [draw=black, fill=black!40, line width=0.06em, thick] (9, -3.8) rectangle (13, -5.8   ) node[pos=0.5, scale= 1.5]{Group};   
    
    \draw [, draw=black!40, line width=0.1em, shorten <=0.5pt, shorten >=0.5pt, >=stealth,]
       (6.8, -2.8) node[]{}
    -- (17.5, -2.8   ) node[]{};
    \draw [-{Triangle[angle=60:.1cm 1]}, draw=black!40, line width=0.1em, shorten <=0.5pt, shorten >=0.5pt, >=stealth,]
       (6.95, -2.8) node[]{}
    -- (6.95, -1.5) node[]{};
    \draw[draw=black!40, fill=black!40, thick](11.05,-2.8) circle (0.2) node[]{};
    \draw[draw=black!40, fill=black!40, thick](14.3,-2.8) circle (0.2) node[]{};
    \draw [-{Triangle[angle=60:.1cm 1]}, draw=black!40, line width=0.1em, shorten <=0.5pt, shorten >=0.5pt, >=stealth,]
       (11.05, -3.8) node[]{}
    -- (11.05, -1.5) node[pos= 0.23, scale= 1.5]{~~~~~~$\groups$};
    \draw [-{Triangle[angle=60:.1cm 1]}, draw=black!40, line width=0.1em, shorten <=0.5pt, shorten >=0.5pt, >=stealth,]
       (14.3, -2.8) node[]{}
    -- (14.3, -1.5) node[]{};
    \draw [-{Triangle[angle=60:.1cm 1]}, draw=black!40, line width=0.1em, shorten <=0.5pt, shorten >=0.5pt, >=stealth,]
       (17.35, -2.8) node[]{}
    -- (17.35, -1.5) node[]{};
    
    \draw [draw=black, fill=white, line width=0.06em] (2.7, -0.5) rectangle (4.7, -1.7   ) node[pos= 0.5, scale= 1.2, rotate= 00, align=center]{lower};
    \draw [draw=black, fill=white, line width=0.06em] (2.7, 0.5) rectangle (4.7, 1.7   ) node[pos= 0.5, scale= 1.2, rotate= 00, align=center]{$\prune$};
    
    \node[inner sep=0pt] (input) at (5.25,0) {$\left(\vphantom{\includegraphics[height=1.2cm]{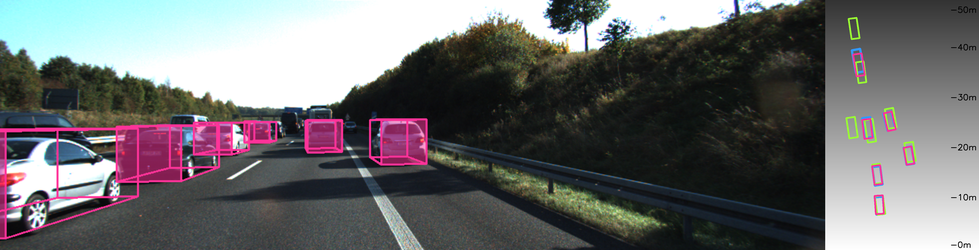}}\right.$};
    
    \draw [draw=black, line width=0.06em] (5.45, 1.5) rectangle (8.45, -1.5   ) node[pos= 0.5, scale= 1.25, rotate= 00, align=center]{};
    \node [inner sep=1pt, scale= 1.5] at (6.8, 2.55)  {$\identity$};
    
    \draw [draw=black, fill=my_blue!20, line width=0.0em] (5.48, 1.47) rectangle (7.45, -0.5   ) node[]{};
    \draw [draw=black, fill=my_magenta!20, line width=0.0em] (7.45,-0.5) rectangle (8.42, -1.47   ) node[]{};
    
    \draw [draw=black, fill=my_blue, line width=0.0em] (5.48, 1.47) rectangle (5.95, 1.   ) node[]{};    
    \draw [draw=black, fill=my_blue, line width=0.0em] (5.95, 1.) rectangle (6.45, 0.5   ) node[]{}; 
    \draw [draw=black, fill=my_blue, line width=0.0em] (6.45, 0.5) rectangle (6.95, 0.   ) node[]{}; 
    \draw [draw=black, fill=my_blue, line width=0.0em] (6.95, 0.) rectangle (7.45, -0.5   ) node[]{}; 
    \draw [draw=black, fill=my_magenta, line width=0.0em] (7.45, -0.5) rectangle (7.95, -1.0   ) node[]{}; 
    \draw [draw=black, fill=my_magenta, line width=0.0em] (7.95, -1.0) rectangle (8.42, -1.47) node[]{}; 
    
    \draw [,draw=black, line width=0.075em, shorten <=0.5pt, shorten >=0.5pt, >=stealth]
       (8.6,0.0) node[]{}
    -- (9.3,0.0) node[]{};

    \draw [draw=black, line width=0.06em] (9.55, 1.5) rectangle (12.55, -1.5   ) node[pos= 0.5, scale= 1.25, rotate= 00, align=center]{};
    \node [inner sep=1pt, scale= 1.5] at (10.9, 2.55)  {$\mask$};
    \draw [draw=black, fill=my_blue!20, line width=0.0em] (9.58, 1.47) rectangle (11.55, -0.5   ) node[]{};
    \draw [draw=black, fill=my_magenta!20, line width=0.0em]  (11.55,-0.5) rectangle (12.52, -1.47  ) node[]{}; \draw [draw=black, fill=my_blue, line width=0.0em] (9.58, 1.47) rectangle (10.25, -0.5   ) node[]{};
    \draw [draw=black, fill=my_magenta, line width=0.0em]  (11.55,-0.5) rectangle (12.05, -1.47  ) node[]{}; 
    
    \draw[draw=black](12.95,0) circle (0.25) node[scale= 1.25]{};
    \draw[fill=black](12.95,0) circle (0.07) node[scale= 1.25]{};
    
    \draw [draw=black, fill= white, line width=0.06em] (13.35, 1.5) rectangle (16.35, -1.5   ) node[pos= 0.5, scale= 1.25, rotate= 00, align=center]{};
    \node [inner sep=1pt, scale= 1.5] at (14.2, 2.55)  {$\pruneMat$};
    \draw [draw=black, fill=my_blue!20, line width=0.0em] (13.38, 1.47) rectangle (15.35, -0.5   ) node[]{};
    \draw [draw=black, fill=my_magenta!20, line width=0.0em]  (15.35,-0.5) rectangle (16.32, -1.47  ) node[]{};
    
    \draw [draw=black, fill=my_blue, line width=0.0em] (13.38, 1) rectangle (13.85, 0.5   ) node[]{};   
    \draw [draw=black, fill=my_blue, line width=0.0em] (13.38, 0.5) rectangle (13.85, 0   ) node[]{};
    \draw [draw=black, fill=my_blue, line width=0.0em] (13.85, 0.5) rectangle (14.35, 0.   ) node[]{};   
    \draw [draw=black, fill=my_blue, line width=0.0em] (13.38, 0) rectangle (13.85, -0.5   ) node[]{};
    \draw [draw=black, fill=my_blue, line width=0.0em] (13.85, 0) rectangle (14.35, -0.5   ) node[]{};
    \draw [draw=black, fill=my_blue, line width=0.0em] (14.35, 0) rectangle (14.85, -0.5   ) node[]{};
    \draw [draw=black, fill=my_magenta, line width=0.0em] (15.38, -1.0) rectangle (15.85, -1.47   ) node[]{};

    \node[inner sep=0pt] (input) at (16.6,0) {$\left.\vphantom{\includegraphics[height=1.2cm]{images/groomed/qualitative/000514.png}}\right)$};
    
    \draw [draw=black, fill= white, line width=0.06em] (17.0, 1.5) rectangle (17.7, -1.5   ) node[pos= 0.5, scale= 1.25, rotate= 00, align=center]{};
    \draw [draw=black, fill=my_blue, line width=0.0em] (17.03, 1.47) rectangle (17.67, -0.5   ) node[]{};
    \draw [draw=black, fill=my_magenta, line width=0.0em]  (17.03, -0.5) rectangle (17.67, -1.47  ) node[]{};
    
    \draw [, draw=forward_color, line width=0.1em, shorten <=0.5pt, shorten >=0.5pt, >=stealth,]
       (17.7,0.2) node[]{}
    -- (20.4,0.2) node[]{};
    \draw [, draw=backward_color, line width=0.1em, shorten <=0.5pt, shorten >=0.5pt, >=stealth, densely dashed]
       (17.7,-0.2) node[]{}
    -- (20.4,-0.2) node[]{};
    \draw [, draw=backward_color, line width=0.1em, shorten <=0.5pt, shorten >=0.5pt, >=stealth, densely dashed]
       (18.3,-0.) node[]{}
    -- (18.3,-4.6) node[]{};
    \draw [-{Triangle[angle=60:.1cm 1]}, draw=backward_color, line width=0.1em, shorten <=0.5pt, shorten >=0.5pt, >=stealth, densely dashed]
       (15.05, -4.55) node[]{}
    -- (15.05, -1.5) node[]{};
    \draw [draw=black, fill=white, line width=0.06em] (18.9, .6) rectangle (20.1, -.6   ) node[pos= 0.5, scale= 1.2, rotate= 00, align=center]{$\clip{.}$};
    
    \draw [draw=black, fill= white, line width=0.06em] (20.4, 1.5) rectangle (21.1, -1.5   ) node[]{};
    \draw [draw=black, fill=my_blue, line width=0.0em] (20.43, 1.47) rectangle (21.07, -0.5   ) node[]{};
    \draw [draw=black, fill=my_magenta, line width=0.0em]  (20.43, -0.5) rectangle (21.07, -1.47  ) node[]{};
    
    \draw [, draw=forward_color, line width=0.1em, shorten <=0.5pt, shorten >=0.5pt, >=stealth,]
       (20.55,-1.5) node[]{}
    -- (20.55,-4.95) node[]{};
    \draw [-{Triangle[angle=60:.1cm 1]}, draw=forward_color, line width=0.1em, shorten <=0.5pt, shorten >=0.5pt, >=stealth,]
           (20.45,-4.8) node[]{}
        -- (22.75,-4.8) node[pos= 0.9, scale= 1.5]{~~~~~$\rescore$};
    
    \draw[draw=backward_color, fill=backward_color, thick](18.3,-.2) circle (0.18) node[scale= 1.25]{};
    \draw [, draw=backward_color, line width=0.1em, shorten <=0.5pt, shorten >=0.5pt, >=stealth, densely dashed]
           (18.15,-4.4) node[]{}
        -- (14.90,-4.4) node[]{};
    \draw [, draw=backward_color, line width=0.1em, shorten <=0.5pt, shorten >=0.5pt, >=stealth, densely dashed]
           (22.35,-4.4) node[]{}
        -- (20.85,-4.4) node[]{};
    \draw [, draw=backward_color, line width=0.1em, shorten <=0.5pt, shorten >=0.5pt, >=stealth, densely dashed]
           (20.85,-4.55) node[]{}
        -- (20.85,-1.5) node[]{};
    \draw [-{Triangle[angle=60:.1cm 1]}, draw=backward_color, line width=0.1em, shorten <=0.5pt, shorten >=0.5pt, >=stealth,  densely dashed]
           (22.75,-4.4) node[]{}
        -- (21.50,-4.4) node[]{};
    
    \draw [, draw=forward_color, line width=0.1em, shorten <=0.5pt, shorten >=0.5pt, >=stealth,]
       (20.75,1.5) node[]{}
    -- (20.75,5.15) node[]{};
    \draw [draw=black, fill= white, line width=0.06em] (19.8, 3.5) rectangle (21.3, 4.5   ) node[pos= 0.5, scale= 1.5, rotate= 00, align=center]{$\!>\!v$};
    \draw [-{Triangle[angle=60:.1cm 1]}, draw=forward_color, line width=0.1em, shorten <=0.5pt, shorten >=0.5pt, >=stealth,]
           (20.75,5.0) node[]{}
        -- (22.75,5.0) node[pos= 0.9, scale= 1.5]{~~~~~$\vindex$};    
        
    \draw [,draw=forward_color, line width=0.1em, shorten <=0.5pt, shorten >=0.5pt, >=stealth]
       ( 2.5,-7) node[]{}
    -- ( 4.5,-7) node[pos= 1, scale= 1.5]{~~~~~~~~~~~~~~~~~~~Forward};
    \draw [,draw=backward_color, line width=0.1em, shorten <=0.5pt, shorten >=0.5pt, >=stealth, densely dashed]
       ( 12.5,-7) node[]{}
    -- ( 14.5,-7) node[pos= 1, scale= 1.5]{~~~~~~~~~~~~~~~~~~~Backward};
    
    \draw [draw=black, line width=0.06em, thick] (0, 6.4) rectangle (21.5, -6.4   ) node[]{};

\end{tikzpicture}

%% file: chapters/deviant.tex
\chapter{
    \deviant: \deviantFull for Monocular 3D Object Detection
}
\label{chpt:deviant}

    Modern neural networks use building blocks such as convolutions that are \equivariant{} to arbitrary \twoD translations in the Euclidean manifold. 
    However, these \vanillaBlocks{} are not \equivariant{} to arbitrary \threeD translations in the projective \manifold. 
    Even then, all monocular \threeD detectors use \vanillaBlocks{} to obtain the \threeD coordinates, a task for which the \vanillaBlocks{} are not designed for. 
    This chapter takes the first step towards convolutions equivariant to arbitrary \threeD translations in the projective \manifold.
    Since the depth is the hardest to estimate for monocular detection, this
    chapter proposes \deviantFull (\deviant) built with existing \scaleEquivariant{} steerable blocks. 
    As a result, \deviant is \equivariant{} to the depth translations in the projective \manifold{} whereas vanilla networks are not.
    The additional \depthEquivariance{} forces the \deviant to learn consistent depth estimates, and therefore,
    \deviant achieves state-of-the-art monocular \threeD detection 
    results on \kitti and \waymo datasets in the \imageOnly{} category and performs competitively to methods using extra information.
    Moreover, \deviant works better than vanilla networks in cross-dataset evaluation.

\section{Introduction}\label{sec:deviant_intro}

    Monocular \threeD object detection is a fundamental task in computer vision, where the task is to infer \threeD information including depth from a single monocular image. 
    It has applications in augmented reality \cite{alhaija2018augmented},  gaming \cite{rematas2018soccer}, robotics \cite{saxena2008robotic}, and more recently in autonomous driving \cite{brazil2019m3d, simonelli2020disentangling} as a fallback solution for \lidar{}. 

    Most of the monocular \threeD methods attach extra heads to the \twoD Faster-RCNN \cite{ren2015faster} or CenterNet \cite{zhou2019objects} for \threeD detections.
    Some change architectures \cite{liu2019deep,li2020rtm3d,tang2020center3d} or losses \cite{brazil2019m3d,chen2020monopair}. 
    Others incorporate augmentation \cite{simonelli2020towards}, or confidence \cite{ liu2019deep,brazil2020kinematic}. 
    Recent ones use in-network ensembles \cite{zhang2021objects, lu2021geometry} for better depth estimation.
    
    Most of these methods use \vanillaBlocks{} such as convolutions that are \textit{equivariant} to arbitrary \twoD translations \cite{rath2020boosting, bronstein2021convolution}. 
    In other words, whenever we shift the ego camera in \twoD (See {\color{vanillaShade}{$\transU$}} of \cref{fig:deviant_teaser}), the new image (projection) is a translation of the original image, and therefore, 
    these methods output a translated feature map. 
    However, in general, the camera moves in depth in driving scenes instead of \twoD (See {\color{proposedShade}{$\transZ$}} of \cref{fig:deviant_teaser}). 
    So, the new image is not a translation of the original input image due to the projective transform.
    Thus, using \vanillaBlocks in monocular methods is a mismatch between the assumptions and the regime where these blocks operate.
    Additionally, there is a huge generalization gap between training and validation for monocular \threeD detection (See 
    Modeling translation \equivariance in the correct \manifold improves generalization for tasks in spherical \cite{cohen2018spherical} and hyperbolic \cite{ganea2018hyperbolic} manifolds. 
    Monocular detection involves processing pixels (\threeD point projections) to obtain the \threeD information, and is thus a task in the projective \manifold.
    Moreover, the depth in monocular detection is ill-defined \cite{tang2020center3d}, and thus, the hardest to estimate \cite{ma2021delving}.
    Hence, using building blocks \textit{\equivariant{} to depth translations in the projective \manifold} is a natural choice for improving generalization and is also at the core of this work (See \cref{sec:deviant_why_better_generalize}).

    \begin{figure}[!tb]
        \centering
        \begin{subfigure}[align=bottom]{.35\linewidth}
            \centering
            \input{images/deviant/teaser.tex}
            \caption{Idea.}
        \end{subfigure}
        \hfill
        \begin{subfigure}[align=bottom]{.63\linewidth}
              \centering
              \input{images/deviant/depth_equivariance}
              \caption{Depth \Equivariance{}.}
        \end{subfigure}
        \caption[\deviant Idea and Depth \Equivariance]{
        \textbf{(a) Idea.} Vanilla CNN is \equivariant{} to {{projected \twoD translations $\transU,\transV$}} (in red) of the ego camera. 
        The ego camera moves in \threeD in driving scenes which breaks this assumption. 
        We propose \deviant which is additionally \equivariant{} to {{depth translations $\transZ$}} (in green) in the projective \manifold.
        \textbf{(b) Depth \Equivariance{}}. 
        \deviant enforces additional consistency among the feature maps of an image and its transformation caused by the ego depth translation.
        $\transformationMath_\scaleNotation\!=\!$ scale transformation, $*\!=\!$ vanilla convolution.
        }
        \label{fig:deviant_teaser}
    \end{figure}

    Recent monocular methods use flips \cite{brazil2019m3d}, scale \cite{simonelli2020towards, lu2021geometry}, mosaic \cite{bochkovskiy2020yolov4, sugirtha2021exploring} or copy-paste \cite{lian2021geometry} augmentation, depth-aware convolution \cite{brazil2019m3d}, or geometry \cite{liu2021ground, lu2021geometry, shi2021geometry, zhang2021learning} to improve generalization. 
    Although all these methods improve performance, a major issue is that their backbones are not designed for the projective world. 
    This results in the depth estimation going haywire with a slight ego movement \cite{zhou2021monoef}. 
    Moreover, data augmentation, \forExample, flips, scales, mosaic, copy-paste, is not only limited for the projective tasks, but also does not guarantee desired behavior \cite{gandikota2021training}.

    \begin{table}[!t]
        \caption[\Equivariance comparisons]{\textbf{\Equivariance comparisons}. [Key: Proj.= Projected, ax= axis]}
        \label{tab:deviant_equivariance_comparison}
        \centering
        \scalebox{\scaleFraction}{
            \setlength\tabcolsep{0.1cm}
            \begin{tabular}{m l t c c c t c cm}
                \myTopRule
                 & \multicolumn{3}{ct}{\threeD} & \multicolumn{2}{cm}{Proj. \twoD}\\
                \cline{2-6}
                \textbf{Translation} \scalebox{0.85}{$\rightarrowRHD$}& $x-$ax & $y-$ax & $z-$ax & $u$-ax & $v$-ax\\
                & $(\transX)$ & $(\transY)$ & $(\transZ)$ & $(\transU)$ & $(\transV)$\\
                \myTopRule
                Vanilla CNN & \mathDash{}& \mathDash{}& \mathDash{}& \checkmark & \checkmark \\
                \LogPolar{} \cite{zwicke1983new} & \mathDash{}& \mathDash{}& \checkmark & \mathDash{}& \mathDash{}\\
                \textbf{\deviant} & \mathDash{}& \mathDash{}& \checkmark & \checkmark & \checkmark \\
                Ideal & \checkmark & \checkmark & \checkmark & \mathDash{}& \mathDash{}\\ 
                \myTopRule
            \end{tabular}
        }
    \end{table}

    To address the mismatch between assumptions and the operating regime of the vanilla blocks and improve generalization, we take the first step towards convolutions \equivariant{} to arbitrary \threeD translations in the projective manifold. 
    We propose \deviantFull (\deviant)
    which is additionally equivariant to depth translations in the projective \manifold{} as shown in \cref{tab:deviant_equivariance_comparison}.
    Building upon the classic result from \cite{hartley2003multiple}, we simplify it under reasonable assumptions about the camera movement in autonomous driving to get scale transformations.
    The \scaleEquivariant{} blocks are well-known in the literature \cite{ghosh2019scale, zhu2019scale, sosnovik2020sesn, jansson2021scale}, and consequently, we replace the vanilla blocks in the backbone with their \scaleEquivariant{} steerable counterparts \cite{sosnovik2020sesn} to additionally embed \equivariance{} to depth translations in the projective \manifold. 
    Hence, \deviant learns consistent depth estimates and improves monocular detection.

    In summary, the main contributions of this work include:
    \begin{itemize}
        \item We study the modeling error in monocular \threeD detection  and propose \depthEquivariant{} networks built with \scaleEquivariant{} steerable blocks as a solution.
        \item We achieve state-of-the-art (\sota) monocular \threeD object detection results on the \kitti and \waymo datasets in the \imageOnly{} category and perform competitively to methods which use extra information.
        \item  We experimentally show that \deviant works better in cross-dataset evaluation suggesting better generalization than vanilla CNN backbones.
    \end{itemize}

\section{Related Works}

    \noIndentHeading{\Equivariant{} Neural Networks.}
        The success of convolutions in CNN has led people to look for their generalizations \cite{cohen2016group,weiler2021coordinate}.
        Convolution is the unique solution to \twoD translation \equivariance{} in the Euclidean \manifold{} \cite{bronstein2021convolution, bronstein2021geometric, rath2020boosting}.
        Thus, convolution in CNN is a prior in the Euclidean \manifold.
        Several works explore other group actions in the Euclidean \manifold{} such as \twoD rotations \cite{cohen2014learning, dieleman2016exploiting, marcos2017rotation, weiler2018learning}, scale \cite{kanazawa2014locally, marcos2018scale}, flips \cite{yeh2019chirality}, or their combinations \cite{worrall2017harmonic, wang2021incorporating}. 
        Some consider \threeD translations \cite{worrall2018cubenet} and rotations \cite{thomas2018tensor}. 
        Few \cite{dosovitskiy2021image, wilk2018learning, zhou2020meta} attempt learning the equivariance from the data, but such methods have significantly higher data requirements \cite{worrall2018cubenet}.
        Others change the \manifold{} to spherical \cite{cohen2018spherical}, hyperbolic \cite{ganea2018hyperbolic}, graphs \cite{micheli2009neural}, or arbitrary manifolds \cite{jing2020physical}. 
        Monocular \threeD detection involves operations on pixels which are projections of \threeD point and thus, works in a different \manifold{} namely projective \manifold.
        \cref{tab:deviant_equivariance} summarizes all these \equivariance{}s known thus far.
        \begin{table}[!tb]
            \caption{\Equivariance{}s known in the literature.}
            \label{tab:deviant_equivariance}
            \centering
            \scalebox{0.8}{
            \setlength\tabcolsep{0.1cm}
            \begin{tabular}{m l t c|c|c|c|cm}
                \myTopRule
                \textbf{Transformation} \scalebox{0.85}{$\rightarrowRHD$} & \multirow{2}{*}{Translation} & \multirow{2}{*}{Rotation} & \multirow{2}{*}{Scale} & 
                \multirow{2}{*}{Flips} & \multirow{2}{*}{Learned}\\
                \cline{1-1}
                \textbf{Manifold} {\raisebox{0.1\normalbaselineskip}{\scalebox{0.85}{\downarrowRHD}}}
                &  &  & & &\\
                \myTopRule
                \multirow{2}{*}{Euclidean} & \multirow{2}{*}{Vanilla CNN\cite{lecun1998gradient}} & Polar, & \LogPolar\cite{henriques2017warped}, & \multirow{2}{*}{ChiralNets\cite{yeh2019chirality}} & \multirow{2}{*}{Transformers\cite{dosovitskiy2021image}}\\
                & & Steerable\cite{worrall2017harmonic} &  Steerable\cite{ghosh2019scale} & & \\
                \hline
                Spherical & Spherical CNN\cite{cohen2018spherical} & \mathDash{}& \mathDash{}& \mathDash{}& \mathDash\\
                \hline
                Hyperbolic & Hyperbolic CNN\cite{ganea2018hyperbolic} & \mathDash{}& \mathDash{}& \mathDash{}& \mathDash\\
                \hline
                Projective & Monocular Detector & \mathDash{}& \mathDash{}& \mathDash{}& \mathDash\\
                \myTopRule
            \end{tabular}
            }
        \end{table}

    \noIndentHeading{\ScaleEquivariant{} Networks.}
        Scale \equivariance{} in the Euclidean \manifold{} is more challenging than the rotations because of its acyclic and unbounded nature \cite{rath2020boosting}.
        There are two major lines of work for \scaleEquivariant{} networks. 
        The first \cite{henriques2017warped, esteves2018polar} infers the global scale using \logPolar{} transform \cite{zwicke1983new},
        while the other infers the scale locally by convolving with multiple scales of images \cite{kanazawa2014locally} or filters \cite{xu2014scale}. 
        Several works \cite{ghosh2019scale, zhu2019scale, sosnovik2020sesn, jansson2021scale} extend the local idea, using steerable filters \cite{freeman1991design}.     
        Another work \cite{worrall2019deep} constructs filters for integer scaling.
        We compare the two kinds of \scaleEquivariant{} convolutions on the monocular \threeD detection task
        and show that steerable convolutions are better suited to embed depth (scale) \equivariance.
        Scale equivariant networks have been used for classification \cite{esteves2018polar, ghosh2019scale, sosnovik2020sesn}, \twoD tracking \cite{sosnovik2021siamese} and \threeD object classification \cite{esteves2018polar}.
        We are the first to use \scaleEquivariant{} networks for monocular \threeD detection.

    \noIndentHeading{3D Object Detection.}
        Accurate \threeD object detection uses sparse data from \lidar{}s \cite{shi2019pointrcnn}, which are expensive and do not work well in severe weather \cite{tang2020center3d} and glassy environments. 
        Hence, several works have been on monocular camera-based \threeD object detection, which is simplistic  but has scale/depth ambiguity \cite{tang2020center3d}.
        Earlier approaches \cite{payet2011contours, fidler20123d, pepik2015multi, chen2016monocular} use hand-crafted features, while the recent ones use deep learning.
        Some change architectures \cite{liu2019deep,li2020rtm3d,tang2020center3d,liu2022learning} or losses \cite{brazil2019m3d,chen2020monopair}. 
        Some use scale \cite{simonelli2020towards, lu2021geometry}, mosaic \cite{sugirtha2021exploring} or copy-paste \cite{lian2021geometry} augmentation.
        Others incorporate depth in convolution \cite{brazil2019m3d, ding2020learning}, or confidence \cite{kumar2020luvli, liu2019deep,brazil2020kinematic}. 
        More recent ones use in-network ensembles to predict the depth deterministically \cite{zhang2021objects} or probabilistically \cite{lu2021geometry}.
        A few use temporal cues \cite{brazil2020kinematic}, NMS \cite{kumar2021groomed}, or corrected camera extrinsics \cite{zhou2021monoef} in the training pipeline.
        Some also use CAD models \cite{chabot2017deep, liu2021autoshape} or \lidar{} \cite{reading2021categorical} in training.
        Another line of work called \pseudoLidar{} \cite{wang2019pseudo, ma2019accurate,ma2020rethinking,simonelli2021we,park2021pseudo} estimates the depth first, and then uses a point cloud-based \threeD object detector.
        We refer to \cite{ma20233d} for a detailed survey.
        Our work is the first to use \scaleEquivariant{} 
        blocks in the backbone for monocular \threeD detection.

\section{Background}
    We first provide the necessary definitions which are used throughout this chapter. These are not our contributions and can be found in the literature \cite{worrall2018cubenet, burns1992non, hartley2003multiple}.
    
    \noIndentHeading{\Equivariance.} 
        Consider a group of transformations $\transformationGroup$, whose individual members are $\transformationGroupMember$. 
        Assume $\mapping$ denote the mapping of the inputs $\inputData$ to the outputs $\outputData$. 
        Let the inputs and outputs undergo the transformation $\transformationInput$ and $\transformationOutput$ respectively. 
        Then, the mapping $\mapping$ is equivariant to the group $\transformationGroup$\cite{worrall2018cubenet} if
            $\mapping (\transformationInput \inputData) = \transformationOutput (\mapping \inputData), \forall~\transformationGroupMember\in\transformationGroup.$
        Thus, \equivariance{} provides an explicit relationship~between input transformations and feature-space transformations at each layer of the neural network \cite{worrall2018cubenet}, and intuitively makes the learning easier. 
        The mapping $\mapping$ is the \vanillaConv~when the $\transformationInput= \transformationOutput= \transformationMath_\translation$ where $\transformationMath_\translation$ denotes the translation $\translation$ on the discrete grid \cite{bronstein2021convolution, bronstein2021geometric, rath2020boosting}.
        These \vanillaConv~introduce weight-tying \cite{lecun1998gradient} in fully connected neural networks resulting in a greater generalization.
        A special case of \equivariance{} is the invariance \cite{worrall2018cubenet} which is given by $\mapping (\transformationInput \inputData) = \mapping \inputData, \forall~\transformationGroupMember\in\transformationGroup.$

    \noIndentHeading{Projective Transformations.} 
        Our idea is to use \equivariance{} to depth translations in the projective \manifold{} since the monocular detection task belongs to this manifold.
        A natural question to ask is whether such \equivariant{}s exist in the projective manifold.
        \cite{burns1992non} answers this question in negative, and says that such \equivariant{}s do not exist in general. 
        However, such equivariants exist for special classes, such as planes.
        An intuitive way to understand this is  to infer the rotations and translations by looking at the two projections (images). 
        For example, the result of \cite{burns1992non} makes sense if we consider a car with very different front and back sides as in
        \cref{fig:deviant_non_existence}. 
        A $180^\circ$ ego rotation around the car means the projections (images) are its front and the back sides, which are different. 
        Thus, we can not infer the translations and rotations 
        from these two projections.
        Based on this result, we stick with \textbf{locally} planar objects \thatIs{} we assume that a \threeD object is made of several \textit{patch planes}. 
        (See last row of \cref{fig:deviant_receptive} as an example). 
        It is important to stress that we do \textbf{NOT} assume that the \threeD object such as car is planar. 
        The local planarity also agrees with the property of manifolds that manifolds locally resemble $n$-dimensional Euclidean space and because the projective transform maps planes to planes, the patch planes in \threeD are also locally planar. 
        We show a sample planar patch and the \threeD object in \cref{fig:deviant_eqv_exists} in the appendix.

    \noIndentHeading{Planarity and Projective Transformation.}                              
        Example 13.2 from \cite{hartley2003multiple} links the planarity and projective transformations.
        Although their result is for stereo with two different cameras $(\projectionOperator, \projectionOperator')$, we substitute $\projectionOperator\!=\!\projectionOperator'$ to get \cref{th:projective_bigboss}. 
        \begin{theorem}\label{th:projective_bigboss}\cite{hartley2003multiple}
            Consider a \threeD point lying on a \plane~$mx\!+\!ny\!+\!oz\!+\!p\!=\!0$, and observed by an ego camera in a pinhole setup to give an image $\projectionOne$. 
            Let $\translation\!=\!(\transX,\transY,\transZ)$ and $\rotation\!=\![r_{ij}]_{3\times3}$ denote a translation and rotation of the ego camera respectively. 
            Observing the same \threeD point from a new camera position leads to an image $\projectionTwo$. 
            Then, the image $\projectionOne$ is related to the image $\projectionTwo$ by the projective \transformation{} 
            \begin{align}
                &\transformationMath: \projectionOne(\pixU-\ppointU, \pixV-\ppointV) =  
                \label{eq:bigboss}\\
                &\!\projectionTwo\!\left(\!\focal\!\dfrac
                {\left(\!r_{11}\mySign\transXTwo\!\frac{m}{p}\right)\!\pixUMinusPointU
                \!+\!\left(\!r_{21}\mySign\transXTwo\!\frac{n}{p}\right)\!\pixVMinusPointV
                \!+\!\left(\!r_{31}\mySign\transXTwo\!\frac{o}{p}\right)\!\focal
                }
                {\left(\!r_{13}\mySign\transZTwo\frac{m}{p}\right)\!\pixUMinusPointU
                \!+\!\left(\!r_{23}\mySign\transZTwo\!\frac{n}{p}\right)\!\pixVMinusPointV
                \!+\!\left(\!r_{33}\mySign\transZTwo\!\frac{o}{p}\right)\!\focal
                },
                \right.\nonumber \\
                &\!\left.\!\focal\!\dfrac
                {\left(\!r_{12}\mySign\transYTwo\!\frac{m}{p}\right)\!\pixUMinusPointU 
                \!+\!\left(\!r_{22}\mySign\transYTwo\!\frac{n}{p}\right)\!\pixVMinusPointV  
                \!+\!\left(\!r_{32}\mySign\transYTwo\!\frac{o}{p}\right)\!\focal
                }
                {\left(\!r_{13}\mySign\transZTwo\!\frac{m}{p}\right)\!\pixUMinusPointU
                \!+\!\left(\!r_{23}\mySign\transZTwo\!\frac{n}{p}\right)\!\pixVMinusPointV
                \!+\!\left(\!r_{33}\mySign\transZTwo\!\frac{o}{p}\right)\!\focal
                } 
                \!\right), \nonumber
            \end{align}
             where $\focal$ and $(\ppointU, \ppointV)$ denote the 
            focal length and principal point of the ego camera, and  $(\transXTwo, \transYTwo, \transZTwo) = \rotation^T\translation$. 
        \end{theorem}

\section{\DepthEquivariant{} Backbone}\label{sec:deviant_proposed}
    The projective transformation in \cref{eq:bigboss} from \cite{hartley2003multiple} is complicated and also involves rotations, and we do not know which convolution obeys this projective \transformation. 
    Hence, we simplify \cref{eq:bigboss} under reasonable assumptions to obtain a familiar \transformation{} for which the \textit{convolution} is known.
        \begin{corollary}\label{th:projective_scaled}
            When the ego camera translates in depth without rotations 
            $(\rotation=\identity)$,
            and the \plane~is ``approximately'' parallel to the image plane, the image $\projectionOne$ locally is a scaled version of the second image $\projectionTwo$ independent of focal length, \thatIs
            \begin{align}
                &\transformationMath_\scaleNotation: \projectionOne(\pixU-\ppointU, v-\ppointV)
                \approx \projectionTwo\left(\dfrac{\pixU-\ppointU}{1\mySign\transZ\frac{o}{p}}, \dfrac{v-\ppointV}{1\mySign\transZ\frac{o}{p}}\right).
                \label{eq:projective_scaled}
            \end{align}
            where $\focal$ and $(\ppointU, \ppointV)$ denote the 
            focal length and principal point of the ego camera, and $\transZ$ denotes the ego translation.
        \end{corollary}
        See \cref{sec:deviant_approximation_proof} for the detailed explanation of \cref{th:projective_scaled}.
        \cref{th:projective_scaled} says
        \begin{align}
            &\transformationMath_\scaleNotation : \projectionOne(\pixU-\ppointU, \pixV-\ppointV)
            \approx \projectionTwo\left(\frac{\pixU-\ppointU}{s}, \frac{\pixV-\ppointV}{s}\right),
        \end{align}
        where, $s\!=\!1+\transZ\frac{o}{p}$ denotes the scale and $\transformationMath_\scaleNotation$ denotes the scale \transformation. 
        The scale $s\! <\! 1$ suggests downscaling, while $s \!>\! 1$ suggests upscaling.
        \cref{th:projective_scaled} shows that the \transformation{} $\transformationMath_\scaleNotation$  is independent of the focal length and that scale is a linear function of the depth translation.
        Hence, the depth translation in the projective \manifold{} induces scale \transformation{} and thus, the depth \equivariance{} in the projective manifold is the \scaleEquivariance{} in the Euclidean manifold. 
        Mathematically, the desired \equivariance{} is $\left[\transformationMath_{\scaleNotation}(\projectionOne)\conv\filter\right] = \transformationMath_{\scaleNotation}\left[\projectionOne \conv \filter_{\scaleNotation^{-1}}\right]$, where $\filter$ denotes the filter (See \cref{sec:deviant_scale_eqv_proof}).
        As CNN is not a \scaleEquivariant{} (\se) architecture \cite{sosnovik2020sesn}, 
        we aim to get \se{} backbone which makes the architecture \equivariant{} to depth translations in the projective manifold.
        The scale transformation is a familiar \transformation{} and \se{} convolutions are well known \cite{ghosh2019scale, zhu2019scale, sosnovik2020sesn, jansson2021scale}.

        \begin{figure}[!tb]
            \centering
            \begin{subfigure}[align=bottom]{.31\linewidth}
                \centering
                \includegraphics[width=\linewidth]{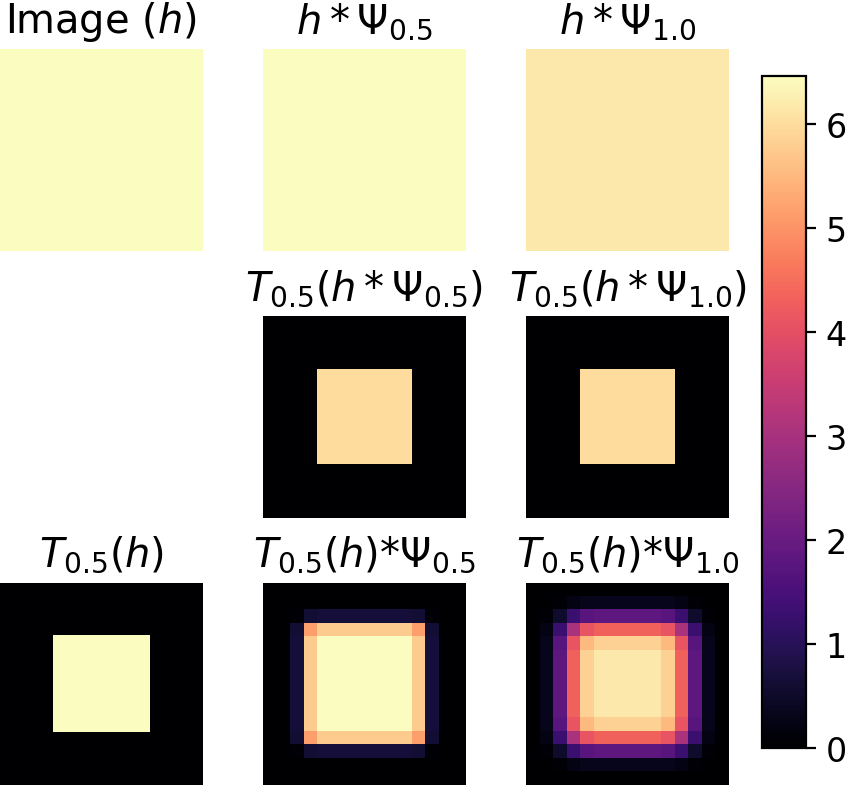}
                \caption{\ses{} Convolution Output.}
                \label{fig:deviant_scale_eq_toy}
            \end{subfigure}\hfill%
            \begin{subfigure}[align=bottom]{.27\linewidth}
                \centering
                \includegraphics[width=\linewidth]{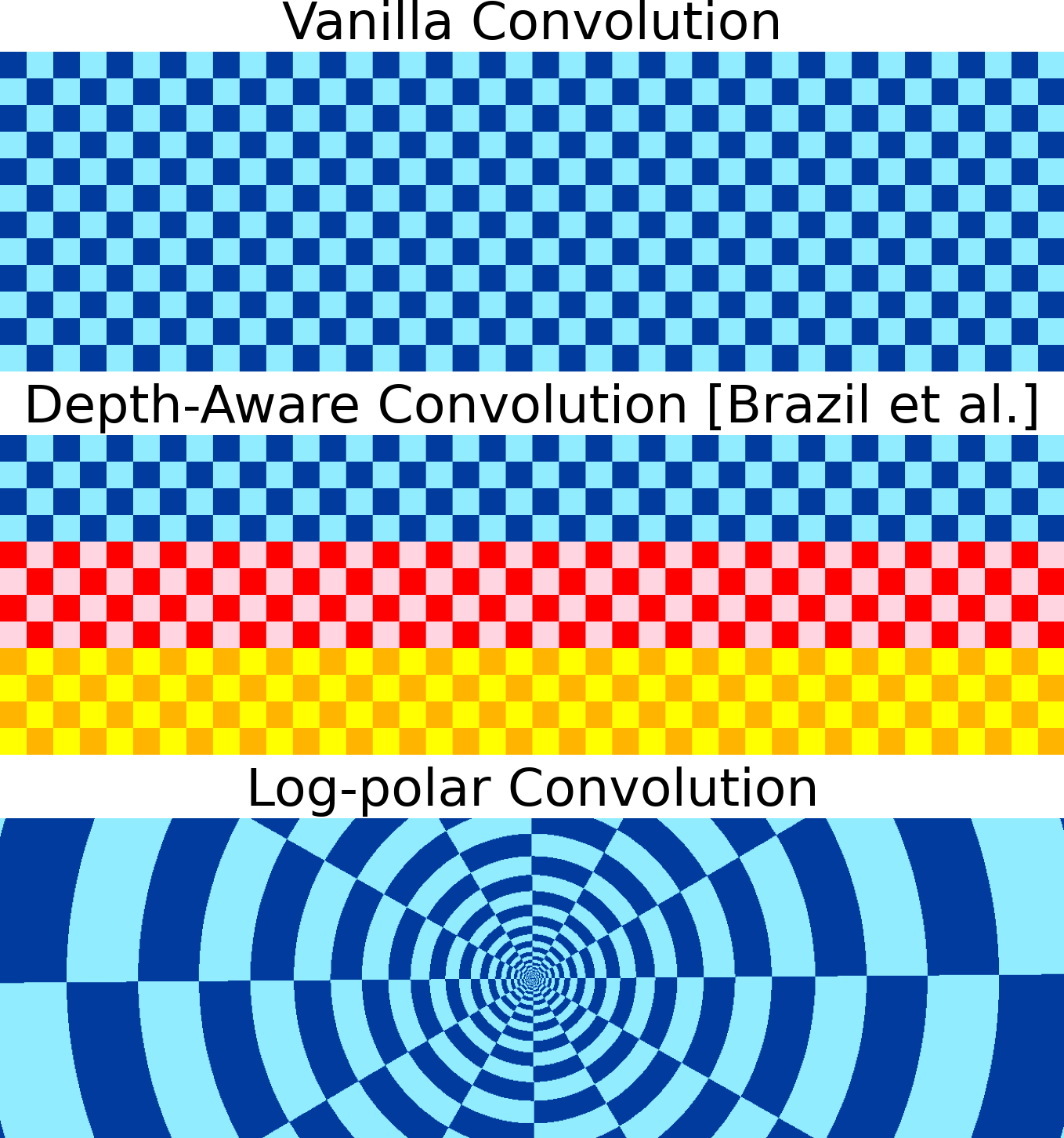}
                \caption{Receptive fields.}
                \label{fig:deviant_receptive}
            \end{subfigure}\hfill%
            \begin{subfigure}[align=bottom]{.38\linewidth}
                \centering
                \includegraphics[width=\linewidth]{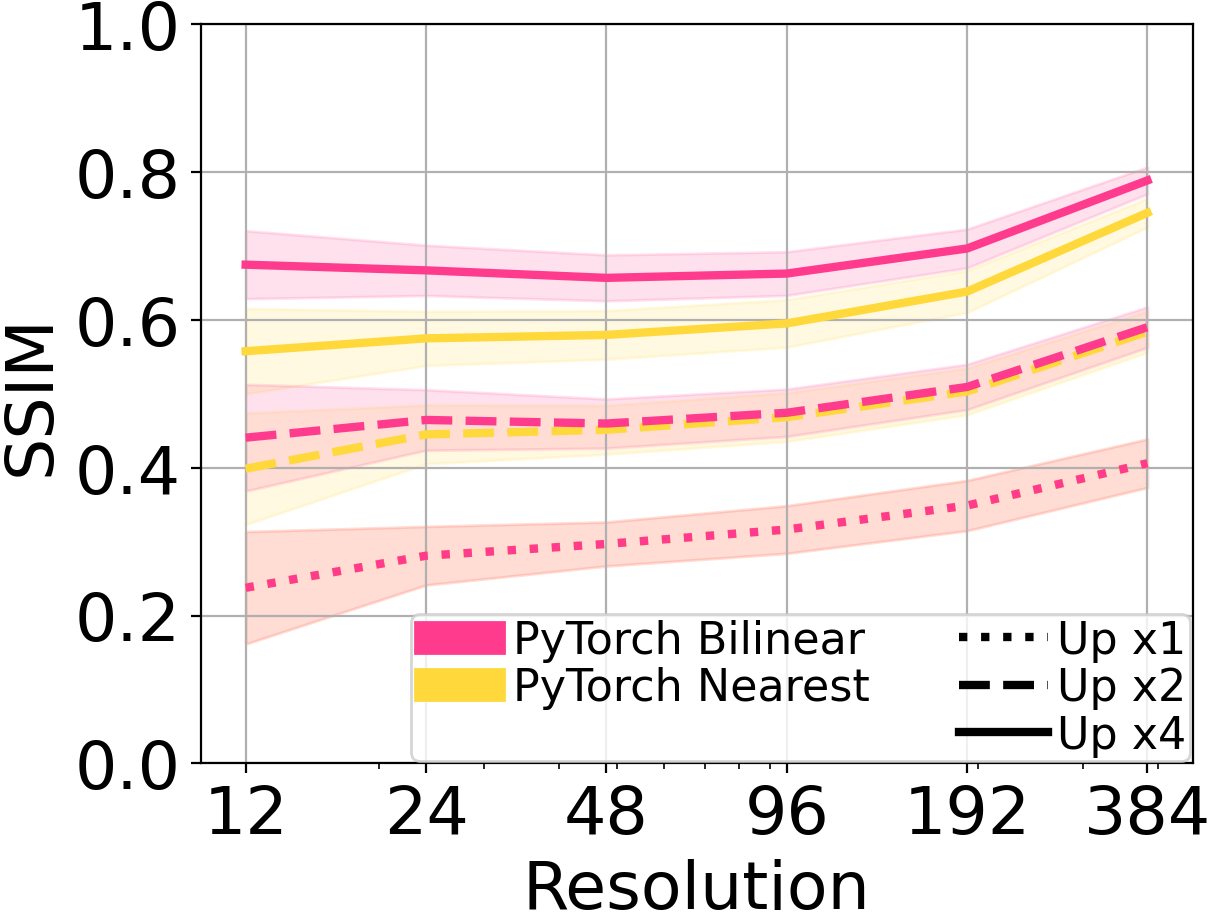}
                \caption{\LogPolar{} SSIM.}
                \label{fig:deviant_log_polar_ssim}
            \end{subfigure}
            \caption[\ScaleEquivariance, Receptive fields of convolutions in the Euclidean manifold and Impact of discretization on \logPolar convolution.]{ 
            \textbf{(a) \ScaleEquivariance}. We apply \ses{} convolution \cite{sosnovik2020sesn} with two scales on a single channel toy image $\projectionOne$. 
            \textbf{(b) Receptive fields} of convolutions in the Euclidean manifold. Colors represent different weights, while shades represent the same weight. 
            \textbf{(c) Impact of discretization on \logPolar{} convolution.} SSIM is very low at small resolutions and is not $1$ even after upscaling by $4$. 
            [Key: Up= Upscaling]
            }
        \end{figure}

        \noIndentHeading{\ScaleEquivariant{} Steerable (\ses) Blocks.}\label{sec:deviant_steerable}    
            We use the existing \ses{} blocks \cite{sosnovik2020sesn,sosnovik2021siamese} to construct our \deviantFull (\deviant) backbone.
            As \cite{sosnovik2021siamese} does not construct \se-\dla{} backbones, we construct our \deviant backbone as follows.
            We replace the vanilla convolutions by the \ses{} convolutions \cite{sosnovik2021siamese} with the basis as Hermite polynomials.
            \ses{} convolutions result in multi-scale representation of an input tensor. 
            As a result, their output is five-dimensional instead of four-dimensional. 
            Thus, we replace the \twoD pools and batch norm (BN) by \threeD pools and \threeD BN respectively.
            The \MaxScale{} layer \cite{sosnovik2020sesn} carries a $\max$ over the extra (scale) dimension to project five-dimensional tensors to four dimensions (See \cref{fig:deviant_steerable_idea}) in the supplementary).
            Ablation in \cref{sec:deviant_results_ablation} confirms that BN and Pool (BNP) should also be \se{} for the best performance.
            
            The \ses{} convolutions \cite{ghosh2019scale, zhu2019scale, sosnovik2020sesn} are based on steerable-filters \cite{freeman1991design}.
            Steerable approaches \cite{ghosh2019scale} first pre-calculate the non-trainable multi-scale basis in the Euclidean \manifold{} and then build filters by the linear combinations of the trainable weights $\weight$.
            The number of trainable weights $\weight$ equals the number of filters at one particular scale. 
            The linear combination of multi-scale basis ensures that the filters are also multi-scale.
            Thus, \ses{} blocks bypass grid conversion and do not suffer from sampling effects.
            
            We show the convolution of toy image $\projectionOne$ with a \ses{} convolution in \cref{fig:deviant_scale_eq_toy}. Let $\filter_\scaleNotation$ denote the filter at scale $\scaleNotation$.
            The convolution between downscaled image and filter $\transformationMath_{0.5}(\projectionOne)\conv\filter_{0.5}$ matches the downscaled version of original image convolved with upscaled filter $\transformationMath_{0.5}(\projectionOne\conv\filter_{1.0})$. 
            \cref{fig:deviant_scale_eq_toy} (right column) shows that the output of a CNN  exhibits aliasing in general and is therefore, not \scaleEquivariant.

        \noIndentHeading{\LogPolar{} Convolution: Impact of Discretization.}\label{sec:deviant_log_polar}
            An alternate way to convert the depth translation $\transZ$ of \cref{eq:projective_scaled} to shift is by converting the images to \logPolar{} space \cite{zwicke1983new} around the principal point $(\ppointU,\ppointV)$, as
            \begin{align}
                \projectionOne(\ln r, \theta)  &\approx \projectionTwo
                \left(
                \ln r - \ln\left(1\mySign\transZ\frac{o}{p}\right),~~\theta \right), 
                \label{eq:log_polar}
            \end{align}
            with $r\!=\!\sqrt{(\pixU\!-\!\ppointU)^2\!+\!(\pixV\!-\ppointV)^2}$, and $\theta\!=\! \tan^{-1}\left(\frac{\pixV-\ppointV}{\pixU-\ppointU}\right)$.
            The \logPolar{} transformation converts the scale to translation, so using convolution in the \logPolar{} space is \equivariant{} to the logarithm of the depth translation $\transZ$.
            We show the receptive field of \logPolar{} convolution in \cref{fig:deviant_receptive}.
            The \logPolar{} convolution uses a smaller receptive field for objects closer to the principal point, while a larger field away from the principal point.
            We implemented \logPolar{} convolution and found that its performance (See \cref{tab:deviant_ablation}) is not acceptable, consistent with \cite{sosnovik2020sesn}.
            We attribute this behavior to the discretization of pixels and loss of \twoD translation \equivariance.
            \cref{eq:log_polar} is perfectly valid in the continuous world (Note the use of parentheses instead of square brackets in \cref{eq:log_polar}). 
            However, pixels reside on discrete grids, which gives rise to sampling errors \cite{kumar2013estimation}.
            We discuss the impact of discretization on \logPolar{} convolution in \cref{sec:deviant_detection_results_kitti_val1} and show it in \cref{fig:deviant_log_polar_ssim}. 
            Hence, we do not use \logPolar{} convolution for the \deviant backbone.

        \noIndentHeading{Comparison of \Equivariance s for Monocular 3D Detection.}
            We now compare \equivariance{}s for monocular \threeD detection task.
            An ideal monocular detector should be \equivariant{} to arbitrary \threeD translations $(\transX, \transY, \transZ)$.
            However, most monocular detectors \cite{kumar2021groomed, lu2021geometry}  estimate \twoD projections of \threeD centers and the depth, which they back-project in \threeD world via known camera intrinsics. 
            Thus, a good enough detector shall be \equivariant{} to \twoD translations $(\transU,\transV)$ for projected centers as well as \equivariant{} to depth translations $(\transZ)$. 
            
            Existing detector backbones \cite{kumar2021groomed, lu2021geometry} are only \equivariant{} to \twoD translations as they use vanilla convolutions that produce \fourD{} feature maps. 
            Log-polar backbones is \equivariant{} to logarithm of depth translations but not 
            to \twoD translations.
            \deviant uses \ses{} convolutions to produce \fiveD feature maps. 
            The extra dimension in \fiveD feature map~captures the changes in scale (for depth), while these feature maps individually are \equivariant{} to \twoD translations (for projected centers).
            Hence, \deviant augments the \twoD translation \equivariance{} $(\transU,\transV)$ of the projected  centers with the depth translation \equivariance.
            We emphasize that although \deviant is \textbf{not} \equivariant{} to arbitrary \threeD translations in the projective manifold, \deviant \textbf{does} provide the \equivariance{} to depth translations $(\transZ)$ and is thus a first step towards the ideal \equivariance.
            Our experiments (\cref{sec:deviant_experiments}) show that even this additional \equivariance{} benefits monocular \threeD detection task.
            This is expected because depth is the hardest parameter to estimate \cite{ma2021delving}.
            \cref{tab:deviant_equivariance_comparison} summarizes these \equivariance{}s.
            Moreover, \cref{tab:deviant_results_kitti_compare_2d_3d} empirically shows that \twoD detection does not suffer and therefore, confirms that \deviant indeed augments the \twoD \equivariance{} with the \depthEquivariance.
            An idea similar to \deviant is the optical expansion \cite{yang2020upgrading} which augments optical flow with the scale information and benefits depth estimation.

\section{Experiments}\label{sec:deviant_experiments}

        Our experiments use the \kitti \cite{geiger2012we}, \waymo \cite{sun2020scalability} and \nuscenes datasets \cite{caesar2020nuscenes}.
        We modify the publicly-available PyTorch \cite{paszke2019pytorch} code of \gupNet{} \cite{lu2021geometry} and use the \gupNet{} model as our baseline.
        For \deviant, we keep the number of scales as three \cite{sosnovik2021siamese}. 
        \deviant takes $8.5$ hours for training and $0.04$s per image for inference on a single A100 GPU.

        \noIndentHeading{Evaluation Metrics.} 
            \kitti evaluates on three object categories: Easy, Moderate and Hard. 
            It assigns each object to a category based on its occlusion, truncation, and height in the image space. 
            \kitti uses \apThreeDForty{} percentage metric on the Moderate category to benchmark models \cite{geiger2012we} following \cite{simonelli2019disentangling, simonelli2020disentangling}.
            
            \waymo evaluates on two object levels: \levelOne~and \levelTwo. 
            It assigns each object to a level based on the number of \lidar{} points included in its \threeD box. 
            \waymo uses \apHThreeD~percentage metric which is the incorporation of heading information in \apThreeD{} to benchmark models.
            It also provides evaluation at three distances $[0,30)$, $[30,50)$ and $[50, \infty)$ meters.

        \noIndentHeading{Data Splits.}
            We use the following splits of the \kitti,\waymo and \nuscenes: 
            \begin{itemize}
                \item \textit{\kitti Test (Full) split}: Official \kitti \threeD benchmark \cite{kitti2012benchmark}
            consists of $7{,}481$ training and $7{,}518$ testing images \cite{geiger2012we}.
        
                \item \textit{\kitti \val split}: It partitions the $7{,}481$ training images into $3{,}712$ training and $3{,}769$ validation images \cite{chen20153d}.
                
                \item \textit{\waymo \val split}: This~split \cite{reading2021categorical, wang2021progressive} contains $52{,}386$ training and $39{,}848$ validation images from the front camera. 
                We construct its training set by sampling every third frame from the training sequences as in \cite{reading2021categorical, wang2021progressive}.

                \item \textit{\nuscenes \val split:} It consists of $28{,}130$ training and $6{,}019$ validation images from the front camera \cite{caesar2020nuscenes}. We use this split for evaluation \cite{shi2021geometry}.
            \end{itemize}

        \begin{table}[!tb]
            \caption[\kitti Test cars detection results.]{\textbf{Results on \kitti Test cars} 
            at \iouThreeD{} $\geq\!0.7$. Previous results are from the leader-board or papers.
            We show $3$ methods in each Extra category and $6$ methods in the \imageOnly{} category. [Key: \firstKey{Best}, \secondKey{Second Best}]
            }
            \label{tab:deviant_detection_results_kitti_test_cars}
            \centering
            \scalebox{\scaleFraction}{
                \setlength\tabcolsep{0.1cm}
                \begin{tabular}{ml m c m ccc  m cccm}
                    \myTopRule
                    \addlinespace[0.01cm]
                    \multirow{2}{*}{Method} & \multirow{2}{*}{Extra} &\multicolumn{3}{cm}{\apThreeDForty \bracketPercentage (\uparrowRHDSmall)} & \multicolumn{3}{cm}{\apBevForty \bracketPercentage (\uparrowRHDSmall)}\\ 
                    & & Easy & Mod & Hard & Easy & Mod & Hard\\ 
                    \myTopRule
                    AutoShape \cite{liu2021autoshape}         & CAD   &  $22.47$ & $14.17$ & $11.36$       &  $30.66$ & $20.08$    & $15.59$       \\
                    
                    PCT \cite{wang2021progressive}                    & Depth & $21.00$        & $13.37$        & $11.31$       & $29.65$        & $19.03$        & $15.92$ \\
                    DFR-Net \cite{zou2021devil}               & Depth & $19.40$        & $13.63$        & $10.35$       & $28.17$        & $19.17$        & $14.84$       \\
                    MonoDistill \cite{chong2022monodistill} & Depth & $22.97$ & $16.03$ & $13.60$ & $31.87$ & $22.59$ & $19.72$ \\
                    PatchNet-C \cite{simonelli2021we}         & \lidar& $22.40$        & $12.53$        & $10.60$        & \mathDash{}           & \mathDash{}           & \mathDash{}           \\
                    CaDDN \cite{reading2021categorical}       & \lidar& $19.17$        & $13.41$        & $11.46$       & $27.94$        & $18.91$        & $17.19$       \\
                    DD3D \cite{park2021pseudo}               & \lidar &  $23.22$ &  $16.34$ &  $14.20$ &  $30.98$ & $22.56$ & $20.03$\\
                    MonoEF \cite{zhou2021monoef}              & Odometry   & $21.29$        & $13.87$        & $11.71$       & $29.03$        & $19.70$        & $17.26$       \\
                    Kinematic \cite{brazil2020kinematic}      & Video & $19.07$        & $12.72$        & $9.17$        & $26.69$        & $17.52$        & $13.10$       \\
                    \myTopRule
                    \groomedNMS \cite{kumar2021groomed}           & \mathDash{}  & $18.10$        & $12.32$        & $9.65$        & $26.19$        & $18.27$        & $14.05$       \\
                    MonoRCNN \cite{shi2021geometry}           & \mathDash{}  & $18.36$        & $12.65$        & $10.03$       & $25.48$        & $18.11$        & $14.10$       \\
                    \monoDISMulti \cite{simonelli2020disentangling}& \mathDash{} & $16.54$        & $12.97$        & $11.04$       & $24.45$        & $19.25$        & $16.87$       \\
                    Ground-Aware \cite{liu2021ground}               & \mathDash{}  & \second{21.65}        & $13.25$        & $9.91$        & \first{29.81}        & $17.98$        & $13.08$       \\
                    MonoFlex \cite{zhang2021objects}          & \mathDash{}  & $19.94$        & $13.89$        & \first{12.07}& $28.23$        & \second{19.75}        & \second{16.89}       \\
                    \gupNet{} \cite{lu2021geometry} & \mathDash{}  & $20.11$ & \second{14.20} & $11.77$ & \mathDash{}& \mathDash{}& \mathDash\\
                    \hline
                    \bestKey{\deviant (Ours)}                        & \mathDash{}  & \first{21.88} & \first{14.46} & \second{11.89} & \second{29.65} & \first{20.44} & \first{17.43}   \\
                    \myTopRule
                \end{tabular}
            }
        \end{table}          
        \begin{table}[!tb]
            \caption[\kitti Test cyclists and pedestrians results.]{\textbf{Results on \kitti Test cyclists and pedestrians} (Cyc/Ped) at \iouThreeD$\geq\!0.5$. Previous results are from the leader-board or papers. [Key: \firstKey{Best}, \secondKey{Second Best}]
            }
            \label{tab:deviant_detection_results_kitti_test_ped_cyclist}
            \centering
            \scalebox{\scaleFraction}{
                \setlength\tabcolsep{0.1cm}
                \begin{tabular}{ml m c m ccc  m cccm}
                    \myTopRule
                    \addlinespace[0.01cm]
                    \multirow{2}{*}{Method} & \multirow{2}{*}{Extra} &\multicolumn{3}{cm}{Cyc \apThreeDForty \bracketPercentage (\uparrowRHDSmall)} & \multicolumn{3}{cm}{Ped \apThreeDForty \bracketPercentage (\uparrowRHDSmall)}\\ 
                    & & Easy & Mod & Hard & Easy & Mod & Hard\\ 
                    \myTopRule
                    \ddmp \cite{wang2021depth}             & Depth    & $4.18$        & $2.50$        & $2.32$  & $4.93$        & $3.55$        & $3.01$     \\
                    DFR-Net \cite{zou2021devil}               & Depth    & $5.69$        & $3.58$        & $3.10$  & $6.09$        & $3.62$        & $3.39$     \\
                    MonoDistill \cite{chong2022monodistill}       & Depth  & $5.53$        & $2.81$        & $2.40$  & $12.79$        & $8.17$        & $7.45$\\            
                    CaDDN \cite{reading2021categorical}       & \lidar  & $7.00$        & $3.41$        & $3.30$  & $12.87$        & $8.14$        & $6.76$            \\
                    DD3D \cite{park2021pseudo}               & \lidar &  $2.39$ &  $1.52$ & $1.31$ &  $13.91$ & $9.30$ &  $8.05$ \\
                    MonoEF \cite{zhou2021monoef}              & Odometry   & $1.80$ & $0.92$        & $0.71$ & $4.27$        & $2.79$        & $2.21$       \\
                    \myTopRule
                    \monoDISMulti \cite{simonelli2020disentangling}& \mathDash{}   & $1.17$        & $0.54$        & $0.48$    & $7.79$        & $5.14$        & $4.42$          \\
                    MonoFlex \cite{zhang2021objects}          & \mathDash{}  & $3.39$        & $2.10$        & $1.67$      & $11.89$        & $8.16$       & $6.81$  \\
                    \gupNet{} \cite{lu2021geometry} & \mathDash{}  & \second{4.18} & \second{2.65} & \second{2.09} & \first{14.72} & \first{9.53} & \first{7.87} \\
                    \hline	 
                    \rowcolor{white}
                    \bestKey{\deviant (Ours)}                        & \mathDash{}  & \first{5.05} & \first{3.13} & \first{2.59}  & \second{13.43} & \second{8.65} & \second{7.69} \\
                    \myTopRule
                \end{tabular}
            }
        \end{table}         
        \begin{table}[!tb]
            \caption[\kitti \val cars results.]{\textbf{Results on \kitti \val cars}. 
            Comparison with bigger CNN backbones in \cref{tab:deviant_detection_with_bigger_cnn}. 
            [Key: \firstKey{Best}, \secondKey{Second Best}, $^{-}$= No \pretrain] 
            }
            \label{tab:deviant_detection_results_kitti_val1}
            \centering
            \scalebox{0.8}{
            \setlength\tabcolsep{0.05cm}
            \begin{tabular}{ml m c m ccc t ccc m ccc t cccm}
                \myTopRule
                \addlinespace[0.01cm]
                \multirow{3}{*}{Method} & \multirow{3}{*}{Extra} & \multicolumn{6}{cm}{\iouThreeD{} $\geq 0.7$} & \multicolumn{6}{cm}{\iouThreeD{} $\geq 0.5$}\\\cline{3-14}
                & & \multicolumn{3}{ct}{\apThreeDForty \bracketPercentage(\uparrowRHDSmall)} & \multicolumn{3}{cm}{\apBevForty \bracketPercentage(\uparrowRHDSmall)} & \multicolumn{3}{ct}{\apThreeDForty \bracketPercentage(\uparrowRHDSmall)} & \multicolumn{3}{cm}{\apBevForty \bracketPercentage(\uparrowRHDSmall)}\\
                & & Easy & Mod & Hard & Easy & Mod & Hard & Easy & Mod & Hard & Easy & Mod & Hard\\ 
                \myTopRule
                \ddmp \cite{wang2021depth} & Depth & $28.12$ & $20.39$ & $16.34$ & \mathDash{}           & \mathDash{}           & \mathDash{}& \mathDash{}           & \mathDash{}           & \mathDash{}           & \mathDash{}           & \mathDash{}           & \mathDash\\
                PCT \cite{wang2021progressive} & Depth & $38.39$ & $27.53$ & $24.44$ & $47.16$            & $34.65$            & $28.47$& \mathDash{}           & \mathDash{}           & \mathDash{}           & \mathDash{}           & \mathDash{}           & \mathDash\\
                MonoDistill \cite{chong2022monodistill} & Depth & $24.31$ & $18.47$ & $15.76$ &  $33.09$ & $25.40$ & $22.16$       & $65.69$ & $49.35$        & $43.49$        & $71.45$ & $53.11$        & $46.94$ \\
                CaDDN \cite{reading2021categorical} & \lidar & $23.57$ & $16.31$ & $13.84$& \mathDash{}           & \mathDash{}           & \mathDash{}& \mathDash{}           & \mathDash{}           & \mathDash{}           & \mathDash{}           & \mathDash{}           & \mathDash\\
                PatchNet-C \cite{simonelli2021we} & \lidar & $24.51$ & $17.03$ & $13.25$ & \mathDash{}           & \mathDash{}           & \mathDash{}& \mathDash{}           & \mathDash{}           & \mathDash{}           & \mathDash{}           & \mathDash{}           & \mathDash\\
                \ddThreeD~(DLA34) \cite{park2021pseudo} & \lidar & \mathDash{}& \mathDash{}& \mathDash{}& $33.5$ & $26.0$ & $22.6$ & \mathDash{}           & \mathDash{}           & \mathDash{}           & \mathDash{}           & \mathDash{}           & \mathDash\\
                \ddThreeD$^{-}$\!(DLA34)\cite{park2021pseudo} & \lidar & \mathDash{}& \mathDash{}& \mathDash{}& $26.8$ & $20.2$ & $16.7$ & \mathDash{}          & \mathDash{}           & \mathDash{}           & \mathDash{}           & \mathDash{}           & \mathDash\\
                MonoEF \cite{zhou2021monoef} & Odometry & $18.26$& $16.30$ & $15.24$& $26.07$ & $25.21$ & $21.61$ & $57.98$     & $51.80$            & $49.34$            & $63.40$            & $61.13$            & $53.22$\\        
                Kinematic \cite{brazil2020kinematic}  & Video                           & $19.76$ & $14.10$ & $10.47$ &  $27.83$ & $19.72$ & $15.10$       & $55.44$ & $39.47$        & $31.26$        & $61.79$ & $44.68$        & $34.56$\\
                \myTopRule
                MonoRCNN \cite{shi2021geometry}  & \mathDash{} & $16.61$        & $13.19$ & $10.65$ & $25.29$ & $19.22$ & $15.30$ & \mathDash{}           & \mathDash{}           & \mathDash{}           & \mathDash{}           & \mathDash{}           & \mathDash\\
                MonoDLE \cite{ma2021delving}   & \mathDash{}& $17.45$ & $13.66$ & $11.68$ & $24.97$ & $19.33$ & $17.01$ & $55.41$ & $43.42$ & $37.81$ & $60.73$ & $46.87$ & $41.89$\\
                \groomedNMS \cite{kumar2021groomed}  & \mathDash{}                                                 & $19.67$ & $14.32$ & $11.27$ & $27.38$ & $19.75$ & $15.92$ & $55.62$ & $41.07$ & $32.89$ & $61.83$ & $44.98$ & $36.29$\\
                Ground-Aware \cite{liu2021ground} & \mathDash{}& $23.63$ & $16.16$ & $12.06$ & \mathDash{}& \mathDash{}& \mathDash{}& \second{60.92} & $42.18$ & $32.02$ & \mathDash{}           & \mathDash{}           & \mathDash{}\\ 
                MonoFlex \cite{zhang2021objects} & \mathDash{}& \second{23.64} & \first{17.51} & \first{14.83} & \mathDash{}           & \mathDash{}           & \mathDash{}& \mathDash{}           & \mathDash{}           & \mathDash{}           & \mathDash{}           & \mathDash{}           & \mathDash\\
                \gupNet{} (Reported)\!\cite{lu2021geometry}\!& \mathDash{}& $22.76$ & $16.46$ & $13.72$ & \second{31.07} & \second{22.94} & \second{19.75} & 57.62 & 42.33 & 37.59 & 61.78 & 47.06 & 40.88 \\
                \gupNet{} (Retrained)\!\cite{lu2021geometry}\!& \mathDash{}& $21.10$ & $15.48$ & $12.88$ & $28.58$ & $20.92$ & $17.83$ & $58.95$ & \second{43.99} & \second{38.07} & \second{64.60} & \second{47.76} &	\second{42.97} \\
                \hline
                \rowcolor{white}
                \textbf{\deviant (Ours)} & \mathDash{}& \first{24.63} & \second{16.54} & \second{14.52} & \first{32.60} &	\first{23.04} & \first{19.99} &	\first{61.00} & \first{46.00}	& \first{40.18} & \first{65.28} & \first{49.63} & \first{43.50} \\
                \myTopRule
            \end{tabular}
            }
        \end{table}

    \subsection{\kitti Test \monoThreeD}\label{sec:deviant_detection_results_kitti_test}

        \noIndentHeading{Cars.}
            \cref{tab:deviant_detection_results_kitti_test_cars} lists out the results of monocular \threeD detection and BEV evaluation on \kitti Test cars. 
            \cref{tab:deviant_detection_results_kitti_test_cars} results show that \deviant outperforms the \gupNet{} and several other \sota{} methods on both  tasks. 
            Except \ddThreeD \cite{park2021pseudo} and MonoDistill \cite{chong2022monodistill}, \deviant, an image-based method, also outperforms other methods that use extra information.

        \noIndentHeading{Cyclists and Pedestrians.} 
            \cref{tab:deviant_detection_results_kitti_test_ped_cyclist} lists out the results of monocular \threeD detection on \kitti Test Cyclist and Pedestrians. 
            The results show that \deviant achieves \sota{} results in the \imageOnly{} category on the challenging Cyclists, and is competitive on Pedestrians.

    \subsection{\kitti \val \monoThreeD}\label{sec:deviant_detection_results_kitti_val1}
    
        \noIndentHeading{Cars.}
            \cref{tab:deviant_detection_results_kitti_val1} summarizes the results of monocular \threeD detection and BEV evaluation on \kitti \val split at two \iouThreeD{} thresholds of $0.7$ and $0.5$ \cite{chen2020monopair,kumar2021groomed}. 
            We report the \textbf{median} model over 5 runs.
            The results show that \deviant outperforms the \gupNet{} \cite{lu2021geometry} baseline by a significant margin. 
            The biggest improvements shows up on the Easy set. 
            Significant improvements are also on the Moderate and Hard sets. 
            Interestingly, \deviant also outperforms \ddThreeD \cite{park2021pseudo} by a large margin when the large-dataset \pretrain{}ing is not done (denoted by \ddThreeD$^{-}$).

        \begin{figure}[!tb]
            \centering
            \begin{subfigure}{.45\linewidth}
              \centering
              \includegraphics[width=\linewidth]{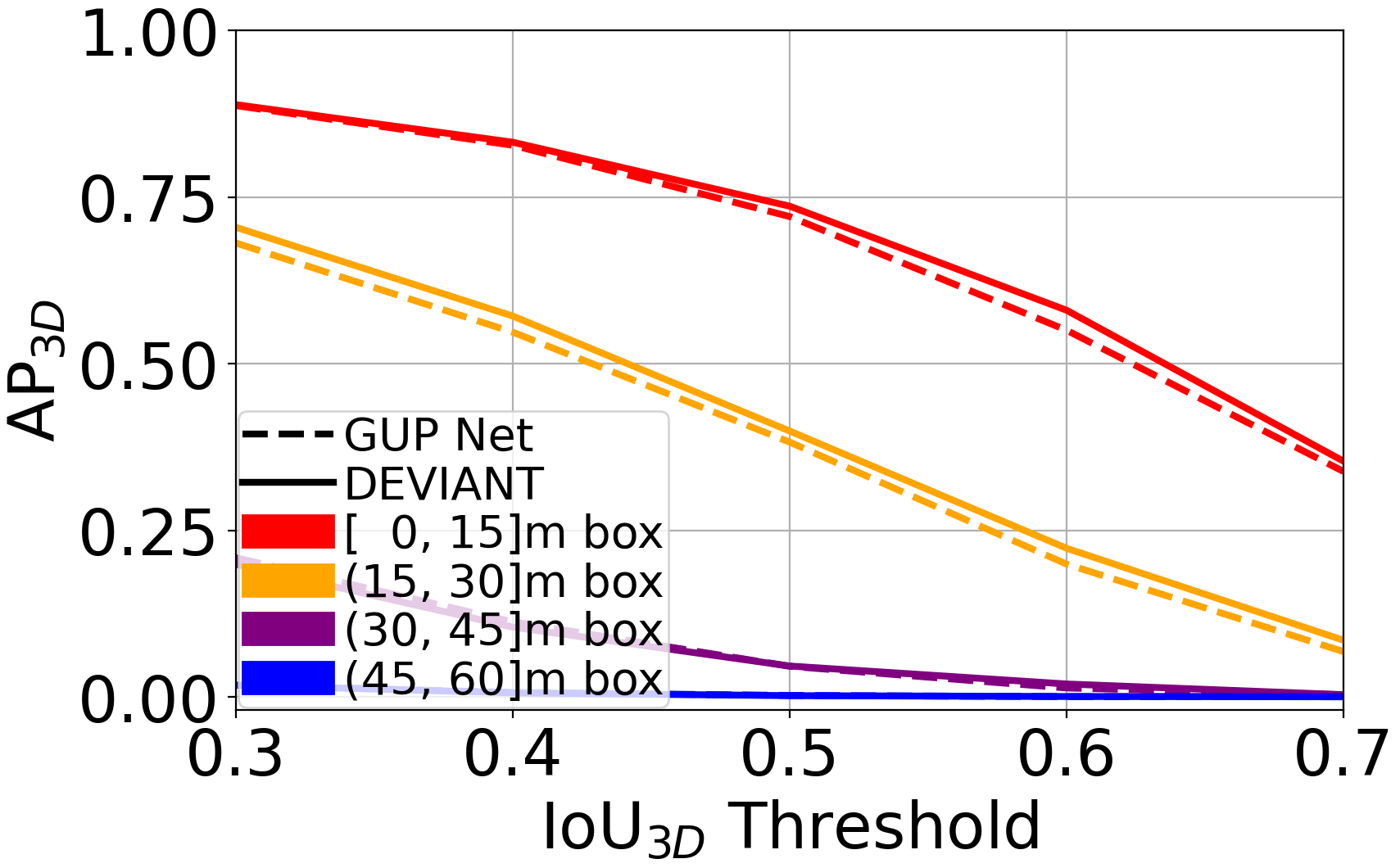}
              \caption{Linear Scale}
            \end{subfigure}%
            \begin{subfigure}{.45\linewidth}
              \centering
              \includegraphics[width=\linewidth]{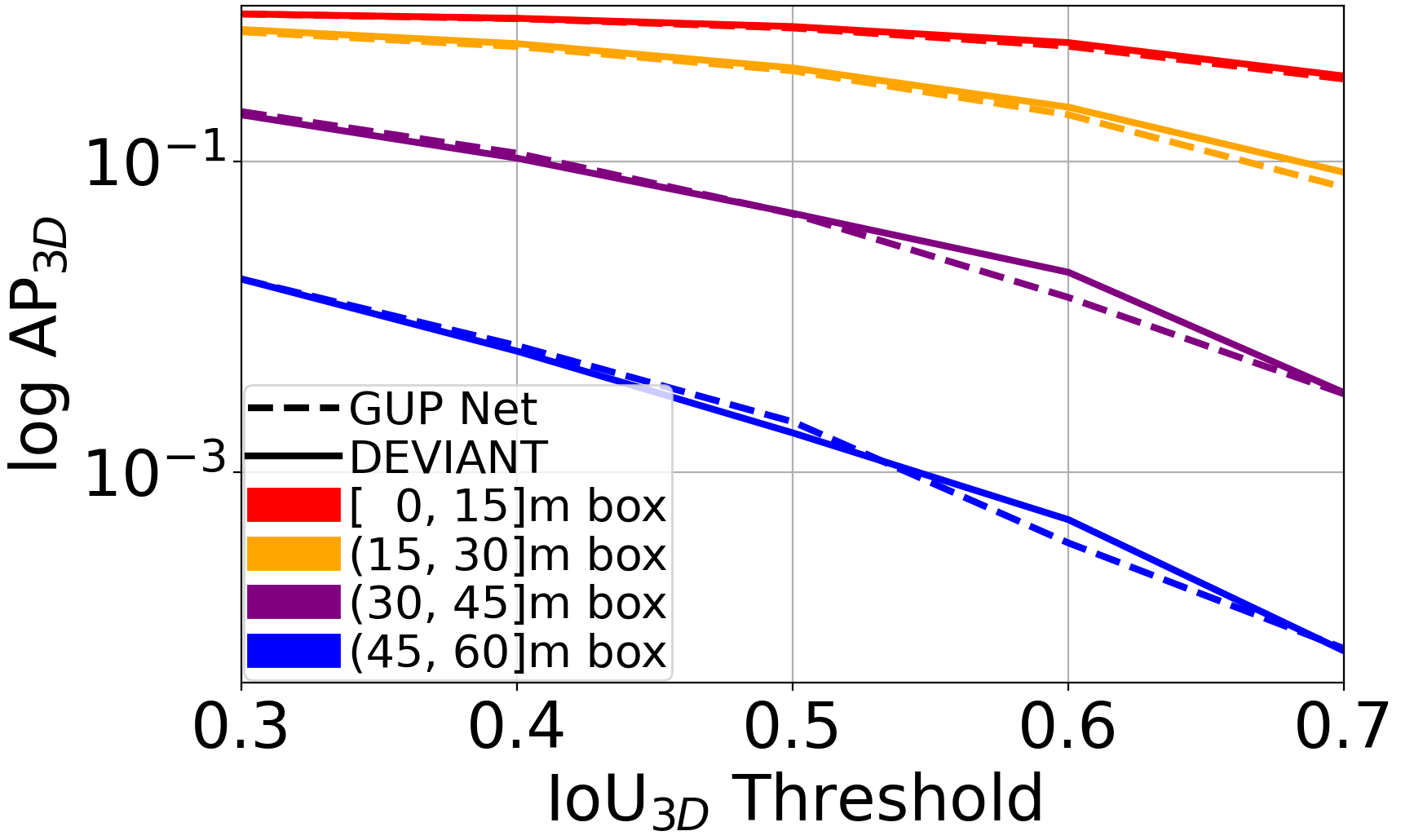}
              \caption{Log Scale}
            \end{subfigure}
            \caption[\apThreeD{} at different depths and \iouThreeD{} thresholds on \kitti \val Split.]{\textbf{\apThreeD{} at different depths and \iouThreeD{} thresholds} on \kitti \val Split. }
            \label{fig:deviant_ap_ground_truth_threshold}
        \end{figure}

        \noIndentHeading{\apThreeD{} at different depths and \iouThreeD{} thresholds.}
            We next compare the \apThreeD{} of \deviant and \gupNet{} in \cref{fig:deviant_ap_ground_truth_threshold} at different distances in meters and \iouThreeD{} matching criteria of $0.3\!\rightarrowRHD\!0.7$ as in \cite{kumar2021groomed}. 
            \cref{fig:deviant_ap_ground_truth_threshold} shows that \deviant is effective over \gupNet{} \cite{lu2021geometry} at all depths and higher \iouThreeD{} thresholds.

        \noIndentHeading{Cross-Dataset Evaluation.}
            \cref{tab:deviant_detection_cross_dataset} shows the result of our \kitti \val model on the \kitti \val and \nuscenes \cite{caesar2020nuscenes} \frontal{} \val images, using mean absolute error (MAE) of the depth of the boxes \cite{shi2021geometry}.
            More details are in \cref{sec:deviant_results_cross_dataset_additional}.
            \deviant outperforms \gupNet{} on most of the metrics on both the datasets, which confirms that \deviant generalizes better than CNNs. 
            \deviant performs exceedingly well in the cross-dataset evaluation than \cite{brazil2019m3d, shi2021geometry, lu2021geometry}.
            We believe this happens because \cite{brazil2019m3d, shi2021geometry, lu2021geometry} rely on data or geometry to get the depth, while \deviant is \equivariant{} to the depth translations, and therefore, outputs consistent depth. 
            So, \deviant is more robust to data distribution changes. 

        \begin{table}[!tb]
            \caption[Cross-dataset evaluation of the \kitti \val model on \kitti \val and \nuscenes \frontal~\val cars with depth MAE.]{\textbf{Cross-dataset evaluation}  of the \kitti \val model on \kitti \val and \nuscenes \frontal~\val cars with depth MAE (\downarrowRHDSmall). [Key: \firstKey{Best}, \secondKey{Second Best}]}
            \label{tab:deviant_detection_cross_dataset}
            \centering
            \scalebox{\scaleFraction}{
                \setlength{\tabcolsep}{0.1cm}
                \begin{tabular}{tlt ccc | c t ccc | ct}
                    \myTopRule
                    \addlinespace[0.01cm]
                    \multirow{2}{*}{Method} & \multicolumn{4}{ct}{\kitti \valOne} & \multicolumn{4}{ct}{\nuscenes \frontal~\val}\\\cline{2-9}
                    &$0\!-\!20$&$20\!-\!40$&$40\!-\!\infty$& All & $0\!-\!20$&$20\!-\!40$& $40\!-\!\infty$&All\\
                    \myTopRule
                    \mthreeDRPN \cite{brazil2019m3d} & $0.56$ & $1.33$ & $2.73$ & $1.26$ & $0.94$ & $3.06$ & $10.36$ & $2.67$\\
                    MonoRCNN \cite{shi2021geometry} &$0.46$& $1.27$ & $2.59$ & $1.14$ & $0.94$ & $2.84$ & $8.65$ & $2.39$\\
                    \gupNet{} \cite{lu2021geometry} & \second{0.45} & \second{1.10} & \second{1.85} & \second{0.89} & \second{0.82} & \second{1.70} & \second{6.20} & \second{1.45}\\
                    \hline
                    \bestKey{\deviant} & \first{0.40} & \first{1.09} & \first{1.80} & \first{0.87} & \first{0.76} & \first{1.60} & \first{4.50} & \first{1.26}\\        
                    \myTopRule
                \end{tabular}
            }
        \end{table}

        \noIndentHeading{Alternatives to \Equivariance.} We now compare with alternatives to \equivariance{} in the following paragraphs.
                
        \noIndentHeading{(a) Scale Augmentation.}
            A withstanding question in machine learning is the choice between \equivariance{} and data augmentation \cite{gandikota2021training}.
            \cref{tab:deviant_kitti_aug_vs_eq} compares \scaleEquivariance{} and scale augmentation.
            \gupNet{} \cite{lu2021geometry} uses scale-augmentation and therefore, \cref{tab:deviant_kitti_aug_vs_eq} shows that \equivariance{} also benefits models which use scale-augmentation. 
            This agrees with Tab. 2 of \cite{sosnovik2020sesn}, where they observe that both augmentation and \equivariance{} benefits classification on MNIST-scale dataset.

        \noIndentHeading{(b) Other \Equivariant{} Architectures.}
            We now benchmark adding depth (scale) \equivariance{} to a \twoD translation \equivariant{} CNN and a transformer which learns the \equivariance. 
            Therefore, we compare \deviant with \gupNet{} \cite{lu2021geometry} (a CNN), and  \detrThreeD \cite{wang2021detr3d} (a transformer) in \cref{tab:deviant_kitti_compare_eq}.
            As \detrThreeD~does not report \kitti results, we trained \detrThreeD~on \kitti using their public code.
            \deviant outperforms \gupNet{} and also surpasses \detrThreeD~by a large margin. This happens because learning \equivariance{} requires more data \cite{worrall2018cubenet} compared to architectures which hardcode \equivariance{} like CNN or \deviant.

        \begin{table}[!tb]
            \caption[Scale Augmentation vs Scale \Equivariance on \kitti \val cars.]{\textbf{Scale Augmentation vs Scale \Equivariance} on \kitti \val cars. [Key: \bestKey{Best}, Eqv= \Equivariance, Aug= Augmentation]}
            \label{tab:deviant_kitti_aug_vs_eq}
            \centering
            \scalebox{0.84}{
                \setlength{\tabcolsep}{0.08cm}
                \begin{tabular}{tl|c|cm ccc t ccc m ccc t ccct}
                    \myTopRule
                    \addlinespace[0.01cm]
                    \multirow{3}{*}{Method} & Scale & Scale & \multicolumn{6}{cm}{\iouThreeD{} $\geq 0.7$} & \multicolumn{6}{ct}{\iouThreeD{} $\geq 0.5$}\\
                    \cline{4-15}
                    & Eqv & Aug  & \multicolumn{3}{ct}{\apThreeDForty \bracketPercentage(\uparrowRHDSmall)} & \multicolumn{3}{cm}{\apBevForty \bracketPercentage(\uparrowRHDSmall)} & \multicolumn{3}{ct}{\apThreeDForty \bracketPercentage(\uparrowRHDSmall)} & \multicolumn{3}{ct}{\apBevForty \bracketPercentage(\uparrowRHDSmall)}\\
                    & & & Easy & Mod & Hard & Easy & Mod & Hard & Easy & Mod & Hard & Easy & Mod & Hard\\
                    \myTopRule
                    \gupNet\!\cite{lu2021geometry}&  & & $20.82$ & $14.15$ & $12.44$ & $29.93$ & $20.90$ & $17.87$ & \best{62.37} & $44.40$ & $39.61$ & \best{66.81} & $48.09$ &	$43.14$\\
                    & & \checkmark & $21.10$ & $15.48$ & $12.88$ & $28.58$ & $20.92$ & $17.83$ & $58.95$ & $43.99$ & $38.07$ & $64.60$ & $47.76$ &	$42.97$\\
                    \myTopRule
                    \textbf{\deviant} & \checkmark & & $21.33$ & $14.77$ & $12.57$ & $28.79$ & $20.28$ & $17.59$ & $59.31$ & $43.25$ & $37.64$ & $63.94$ & $47.02$ & $41.12$\\
                    & \checkmark & \checkmark & \best{24.63} & \best{16.54} & \best{14.52} & \best{32.60} &	\best{23.04} & \best{19.99} &	$61.00$ & \best{46.00}	& \best{40.18} & $65.28$ & \best{49.63} & \best{43.50} \\
                    \myTopRule
                \end{tabular}
            }
        \end{table}
        \begin{table}[!tb]
            \caption[Comparison of \Equivariant{} Architectures on~\kitti \val cars.]{\textbf{Comparison of \Equivariant{} Architectures} on~\kitti \val cars. [Key: \bestKey{Best}, Eqv= \Equivariance, $^\dagger$= Retrained]}
            \label{tab:deviant_kitti_compare_eq}
            \centering
            \scalebox{0.85}{
                \setlength\tabcolsep{0.1cm}
                \begin{tabular}{ml m c m ccc t ccc m ccc t cccm}
                    \myTopRule
                    \addlinespace[0.01cm]
                    \multirow{3}{*}{Method} & \multirow{3}{*}{Eqv} & \multicolumn{6}{cm}{\iouThreeD{} $\geq 0.7$} & \multicolumn{6}{cm}{\iouThreeD{} $\geq 0.5$}\\
                    \cline{3-14}
                    & & \multicolumn{3}{ct}{\apThreeDForty \bracketPercentage(\uparrowRHDSmall)} & \multicolumn{3}{cm}{\apBevForty \bracketPercentage(\uparrowRHDSmall)} & \multicolumn{3}{ct}{\apThreeDForty \bracketPercentage(\uparrowRHDSmall)} & \multicolumn{3}{cm}{\apBevForty \bracketPercentage(\uparrowRHDSmall)}\\
                    & & Easy & Mod & Hard & Easy & Mod & Hard & Easy & Mod & Hard & Easy & Mod & Hard\\ 
                    \myTopRule
                    DETR3D$^\dagger$\!\cite{wang2021detr3d} & Learned & $1.94$ & $1.26$ & $1.09$ & $4.41$ & $3.06$ & $2.79$ & $20.09$ & $13.80$ & $12.78$  & $26.51$ & $18.49$ & 	$17.36$\\
                    \gupNet\!\cite{lu2021geometry}& \twoD & $21.10$ & $15.48$ & $12.88$ & $28.58$ & $20.92$ & $17.83$ & $58.95$ & $43.99$ & $38.07$ & $64.60$ & $47.76$ &	$42.97$\\
                    \bestKey{\deviant} & \twoD+Depth & \best{24.63} & \best{16.54} & \best{14.52} & \best{32.60} &	\best{23.04} & \best{19.99} &	\best{61.00} & \best{46.00}	& \best{40.18} & \best{65.28} & \best{49.63} & \best{43.50} \\
                    \myTopRule
                \end{tabular}
            }
        \end{table}
        \begin{table}[!tb]
            \caption[Comparison with Dilated Convolution on~\kitti \val cars.]{\textbf{Comparison with Dilated Convolution} on~\kitti \val cars. [Key: \bestKey{Best}]}
            \label{tab:deviant_kitti_compare_dilation}
            \centering
            \scalebox{\scaleFraction}{
                \setlength\tabcolsep{0.1cm}
                \begin{tabular}{mc| cm ccc t ccc m ccc t cccm}
                    \myTopRule
                    \addlinespace[0.01cm]
                    \multirow{3}{*}{Method} & \multirow{3}{*}{Extra} & \multicolumn{6}{cm}{\iouThreeD{}$\geq 0.7$} & \multicolumn{6}{ct}{\iouThreeD{}$\geq 0.5$}\\\cline{3-14}
                    & & \multicolumn{3}{ct}{\apThreeDForty \bracketPercentage(\uparrowRHDSmall)} & \multicolumn{3}{cm}{\apBevForty \bracketPercentage(\uparrowRHDSmall)} & \multicolumn{3}{ct}{\apThreeDForty \bracketPercentage(\uparrowRHDSmall)} & \multicolumn{3}{cm}{\apBevForty \bracketPercentage(\uparrowRHDSmall)}\\
                    && Easy & Mod & Hard & Easy & Mod & Hard & Easy & Mod & Hard & Easy & Mod & Hard\\ 
                    \myTopRule
                    D4LCN \cite{ding2020learning} & Depth & $22.32$ & $16.20$ & $12.30$ & $31.53$ & $ 22.58$ & $17.87$ & \mathDash{} & \mathDash{} & \mathDash{} & \mathDash{} & \mathDash{} & \mathDash\\
                    DCNN \cite{yu2015multi} & \mathDash{}&$21.66$ & $15.49$ & $12.90$ & $30.22$ & $22.06$ & $19.01$ & $57.54$ & $43.12$ & $38.80$ & $63.29$ & $46.86$ & $42.42$\\
                    \bestKey{\deviant} & \mathDash{}& \best{24.63} & \best{16.54} & \best{14.52} & \best{32.60} &	\best{23.04} & \best{19.99} &	\best{61.00} & \best{46.00}	& \best{40.18} & \best{65.28} & \best{49.63} & \best{43.50}\\
                    \myTopRule
                \end{tabular}
            }
        \end{table}

        \noIndentHeading{(c) Dilated Convolution.}
            \deviant adjusts the receptive field based on the object scale, and so, we compare with the dilated CNN (DCNN) \cite{yu2015multi} and D4LCN \cite{ding2020learning} in \cref{tab:deviant_kitti_compare_dilation}.
            The results show that DCNN performs sub-par to \deviant.
            This is expected because dilation corresponds to integer scales \cite{worrall2019deep} while the scaling is generally a float in monocular detection.
            D4LCN \cite{ding2020learning} uses monocular depth as input to adjust the receptive field. 
            \deviant (without depth) also outperforms D4LCN on Hard cars, which are more distant. 

        \noIndentHeading{(d) Other Convolutions.}
            We now compare with other known convolutions in literature such as \LogPolar{} convolution \cite{zwicke1983new}, Dilated convolution \cite{yu2015multi} convolution and DISCO \cite{sosnovik2021disco} in \cref{tab:deviant_ablation}. 
            The results show that the \logPolar{} convolution does not work well, and \ses{} convolutions are better suited to embed depth (scale) equivariance.
            As described in~\cref{sec:deviant_log_polar}, we investigate the behavior of \logPolar{} convolution through a small experiment.
            We calculate the SSIM \cite{wang2004image} of the original image and the image obtained after the upscaling, \logPolar, inverse \logPolar, and downscaling blocks.
            We then average the SSIM over all \kitti \val images. 
            We repeat this experiment for multiple image heights and scaling factors.
            The ideal SSIM should have been one.
            However, \cref{fig:deviant_log_polar_ssim} 
            shows that SSIM 
            does not reach $1$ even after upscaling by $4$. 
            This result confirms that \logPolar{} convolution loses information at low resolutions resulting in inaccurate detection.
            
            Next, the results show that dilated convolution \cite{yu2015multi} performs sub-par to \deviant. 
            Moreover, DISCO \cite{sosnovik2021disco} also does not outperform \ses{} convolution which agrees with the \twoD tracking results of \cite{sosnovik2021disco}.

        \noIndentHeading{(e) Feature Pyramid Network (FPN).}
            Our baseline \gupNet{} \cite{lu2021geometry} uses FPN \cite{lin2017feature} and \cref{tab:deviant_detection_results_kitti_val1} shows that \deviant outperforms \gupNet. Hence, we conclude that \equivariance{} also benefits models which use FPN.

        \begin{figure}[!tb]
            \centering
            \begin{subfigure}{.45\linewidth}
              \centering
              \includegraphics[width=\linewidth]{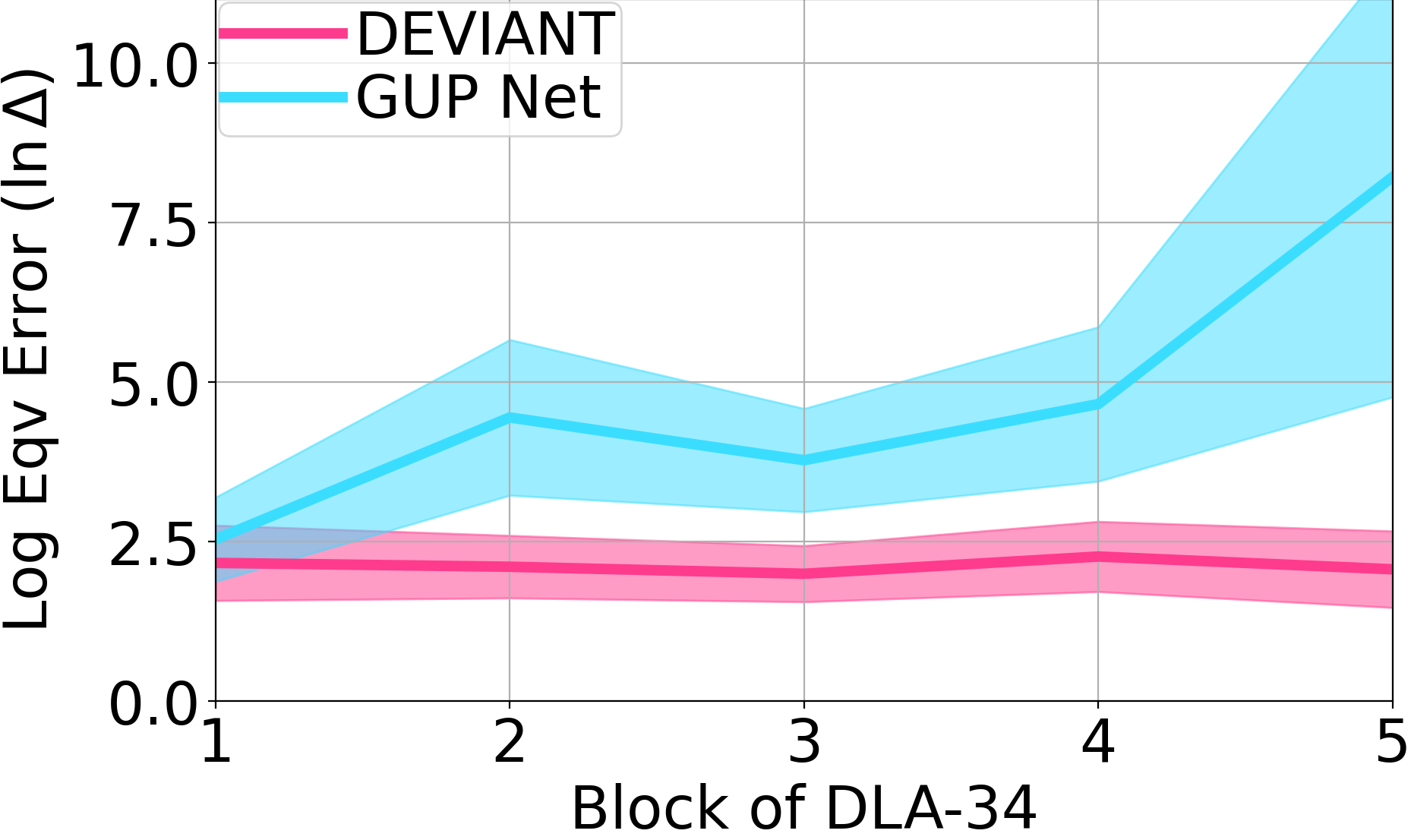}
              \caption{At blocks (depths) of backbone.}
            \end{subfigure}%
            \begin{subfigure}{.45\linewidth}
              \centering
              \includegraphics[width=0.96\linewidth]{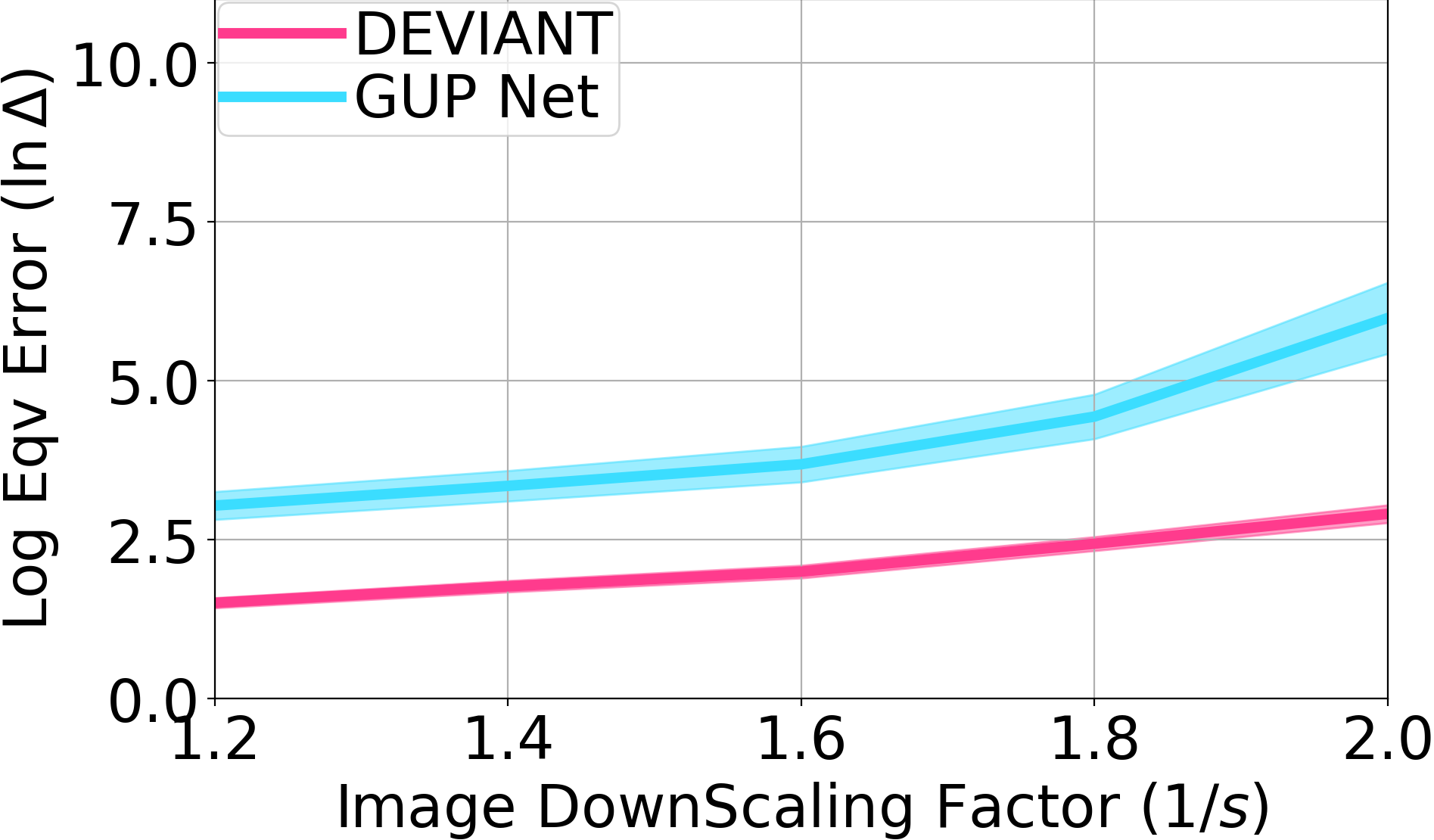}
              \caption{Varying scaling factors.}
              \label{fig:deviant_equiv_error_scaling}
            \end{subfigure}
            \caption[Log \Equivariance{} Error $(\Delta)$ comparison for \deviant and \gupNet{} at different blocks with random image scaling and block $3$ at different image scalings.]{\textbf{Log \Equivariance{} Error $(\Delta)$} comparison for \deviant and \gupNet{} at \textbf{(a)} different blocks with random image scaling factors \textbf{(b)} different image scaling factors at depth $3$. \deviant shows \textbf{lower} \scaleEquivariance{} error than vanilla \gupNet{} \cite{lu2021geometry}.}
            \label{fig:deviant_equiv_error}
        \end{figure}

        \noIndentHeading{Comparison of \Equivariance{} Error.}
            We next quantitatively evaluate the \scaleEquivariance{} of \deviant vs. \gupNet{} \cite{lu2021geometry}, using the \equivariance{} error~metric\cite{sosnovik2020sesn}.
            The \equivariance{} error $\Delta$ is the normalized difference between the scaled feature map and the feature map of the scaled image, and is given by
                $\Delta = \frac{1}{N} \sum_{i=1}^N \frac{||\transformationMath_{s_i} \mapping(\projectionOneIndexed) - \mapping(\transformationMath_{s_i} \projectionOneIndexed)||_2^2}{||\transformationMath_{s_i} \mapping(\projectionOneIndexed)||_2^2},$
            where $\mapping$ denotes the neural network, $\transformationMath_{s_i}$ is the scaling transformation for the image $i$, and $N$ is the total number of images.
            The \equivariance{} error is zero if the \scaleEquivariance{} is perfect.
            We plot the log of this error at different blocks of \deviant and \gupNet{} backbones and also plot at different downscaling of \kitti \val images in \cref{fig:deviant_equiv_error}.
            The plots show that \deviant has low \equivariance{} error than \gupNet.
            This is expected since the feature maps of the proposed \deviant are additionally \equivariant{} to scale \transformation s (depth translations).
            We also visualize the \equivariance{} error for a validation image and for the objects of this image in \cref{fig:deviant_equiv_error_qualitative,fig:deviant_equiv_error_qualitative_object} in the supplementary.
            The qualitative plots also show a lower error for the proposed \deviant, which agrees with \cref{fig:deviant_equiv_error}.
            \cref{fig:deviant_equiv_error_qualitative_object}\textcolor{link_color}{a} shows that \equivariance{} error is particularly low for nearby cars which also justifies the good performance of \deviant on Easy (nearby) cars in \cref{tab:deviant_detection_results_kitti_test_cars,tab:deviant_detection_results_kitti_val1}.
    
        \noIndentHeading{Does 2D Detection Suffer?}
            We now investigate whether \twoD detection suffers from using \deviant backbones in \cref{tab:deviant_results_kitti_compare_2d_3d}.
            The results show that \deviant introduces minimal decrease in the \twoD detection performance.
            This is consistent with \cite{sosnovik2021siamese}, who report that \twoD tracking improves with the \se{} networks.          
        \begin{table*}[!tb]
            \caption[3D~and 2D~detection on \kitti \val cars.]{
            \textbf{3D~and 2D~detection} on \kitti \val cars.}
            \label{tab:deviant_results_kitti_compare_2d_3d}
            \centering
            \scalebox{\scaleFraction}{
                \setlength\tabcolsep{0.1cm}
                \begin{tabular}{tl m ccc t ccc m ccc t ccct}
                    \myTopRule
                    \addlinespace[0.01cm]
                    \multirow{3}{*}{Method} & \multicolumn{6}{cm}{\iou~$\geq 0.7$} & \multicolumn{6}{ct}{\iou~$\geq 0.5$}\\\cline{2-13}
                    & \multicolumn{3}{ct}{\apThreeDForty \bracketPercentage(\uparrowRHDSmall)} & \multicolumn{3}{cm}{\apTwoDForty \bracketPercentage(\uparrowRHDSmall)} & \multicolumn{3}{ct}{\apThreeDForty \bracketPercentage(\uparrowRHDSmall)} & \multicolumn{3}{ct}{\apTwoDForty \bracketPercentage(\uparrowRHDSmall)}\\
                    & Easy & Mod & Hard & Easy & Mod & Hard & Easy & Mod & Hard & Easy & Mod & Hard\\ \myTopRule                
                    \gupNet{} \cite{lu2021geometry} & $21.10$ & $15.48$ & $12.88$ & $96.78$ & $88.87$ & $79.02$ & $58.95$ & $43.99$ & $38.07$ & $99.52$ & $91.89$ &	$81.99$ \\
                    \bestKey{\deviant (Ours)} & $24.63$ & $16.54$ & $14.52$ & $96.68$	& $88.66$	& $78.87$ & $61.00$ &	$46.00$ & $40.18$ & $97.12$ & $91.77$ &	$81.93$\\
                    \myTopRule
                \end{tabular}
            }
        \end{table*} 
        \begin{table*}[!tb]
            \caption[Ablation studies on \kitti \val cars.]{\textbf{Ablation studies} on \kitti \val cars.}
            \label{tab:deviant_ablation}
            \centering
            \scalebox{0.8}{
                \setlength{\tabcolsep}{0.05cm}
                \begin{tabular}{m c m l m ccc t ccc m ccc t ccc m}
                    \myTopRule
                    \addlinespace[0.01cm]
                    \multicolumn{2}{mcm}{\textbf{Change from \deviant :}} & \multicolumn{6}{cm}{\iouThreeD{} $\geq 0.7$} & \multicolumn{6}{cm}{\iouThreeD{} $\geq 0.5$}\\
                    \cline{1-14}
                    \multirow{2}{*}{Changed} & \multirow{2}{*}{From $\longrightarrowRHD$To} & \multicolumn{3}{ct}{\apThreeDForty \bracketPercentage(\uparrowRHDSmall)} & \multicolumn{3}{cm}{\apBevForty \bracketPercentage(\uparrowRHDSmall)} & \multicolumn{3}{ct}{\apThreeDForty \bracketPercentage(\uparrowRHDSmall)} & \multicolumn{3}{cm}{\apBevForty \bracketPercentage(\uparrowRHDSmall)}\\
                    && Easy & Mod & Hard & Easy & Mod & Hard & Easy & Mod & Hard & Easy & Mod & Hard\\
                    \myTopRule
                    & \ses$\rightarrowRHD$Vanilla            &$21.10$ & $15.48$ & $12.88$ & $28.58$ & $20.92$ & $17.83$ & $58.95$ & $43.99$ & $38.07$ & $64.60$ & $47.76$ &	$42.97$\\
                    Convolution &\ses$\rightarrowRHD$\LogPolar\!\cite{zwicke1983new}& $9.19$ & $6.77$ & $5.78$ & $16.39$ & $11.15$ & $9.80$ & $40.51$ & $27.62$ & $23.90$ & $45.66$ & $31.34$ & $25.80$\\
                    & \ses$\rightarrowRHD$Dilated\!\cite{yu2015multi} & $21.66$ & $15.49$ & $12.90$ & $30.22$ & $22.06$ & $19.01$ & $57.54$ & $43.12$ & $38.80$ & $63.29$ & $46.86$ & $42.42$\\ 
                    & \ses$\rightarrowRHD$DISCO\!\cite{sosnovik2021disco} & $20.21$ & $13.84$ & $11.46$ & $28.56$ & $19.38$ & $16.41$ & $55.22$ & $39.76$ & $35.37$ & $59.46$ & $43.16$ & $38.52$\\
                    \hline
                    Downscale
                    & $10\%\rightarrowRHD 5\%$& $24.24$	& $16.51$	& $14.43$	& $31.94$	& $22.86$	& $19.82$	& $60.64$	& $44.46$	& $40.02$	& $64.68$	& $49.30$	& $43.49$\\
                    $\alpha$ & $10\%\rightarrowRHD 20\%$& $22.19$	& $15.85$	& $13.48$	& $31.15$	& $23.01$	& $19.90$	& $61.24$	& $44.93$	& $40.22$	& $67.46$	& $50.10$	& $43.83$\\
                    \hline
                    BNP & \se$\rightarrowRHD$ Vanilla & $24.39$	& $16.20$	& $14.36$	& $32.43$	& $22.53$	& $19.70$	& $62.81$	& $46.14$	& $40.38$	& $67.87$	& $50.23$	& $44.08$\\  
                    \hline   
                    Scales
                    & $3\rightarrowRHD 1$& 
                    $23.20$	& $16.29$	& $13.63$	& $31.76$	& $23.23$	& $19.97$ & $61.90$	& $46.66$	& $40.61$	& $67.37$	& $50.31$	& $43.93$\\
                    & $3\rightarrowRHD 2$& 
                    $24.15$	& $16.48$	& $14.55$	& $32.42$	& $23.17$	& $20.07$	& $61.05$	& $46.34$	& $40.46$	& $67.36$	& $50.32$	& $44.07$\\
                    \hline
                    {---} & \textbf{\deviant (best)} & \best{24.63} & \best{16.54} & \best{14.52} & \best{32.60} &	\best{23.04} & \best{19.99} &	\best{61.00} & \best{46.00}	& \best{40.18} & \best{65.28} & \best{49.63} & \best{43.50}    \\
                    \myTopRule
                \end{tabular}
            }
        \end{table*}

    \subsection{Ablation Studies on \kitti \val}\label{sec:deviant_results_ablation}
        \cref{tab:deviant_ablation} compares the modifications of our approach on \kitti \val cars based on the experimental settings of \cref{sec:deviant_experiments}.
        
        \noIndentHeading{(a) Floating or Integer Downscaling?}
        We next investigate the question that whether one should use floating or integer downscaling factors for \deviant.
        We vary the downscaling factors as $(1\!+\!2\alpha, 1\!+\!\alpha, 1)$ and therefore, our scaling factor $s\!=\!\left(\frac{1}{1+2\alpha}, \frac{1}{1+\alpha}, 1\right)$.
        We find that $\alpha$ of $10\%$ works the best. 
        We again bring up the dilated convolution (Dilated) results at this point because dilation is a \scaleEquivariant{} operation for integer downscaling factors \cite{worrall2019deep} $(\alpha\!=\!100\%, s\!=\!0.5)$.
        \cref{tab:deviant_ablation} results suggest that the downscaling factors should be floating numbers.
        
        \noIndentHeading{(b) \se{} BNP.}
        As described in \cref{sec:deviant_steerable}, we ablate \deviant against the case when only convolutions are \se{} but BNP layers are not. 
        So, we place \MaxScale \cite{sosnovik2020sesn} immediately after every \ses{} convolution. 
        \cref{tab:deviant_ablation} shows that such a network performs slightly sub-optimal to our final model.

        \noIndentHeading{(c) Number of Scales.}
        We next ablate against the usage of Hermite scales. 
        Using three scales performs better than using only one scale especially on Mod and Hard objects, and slightly better than using two scales.

    \subsection{\waymo \val \monoThreeD}
        We also benchmark our method on the \waymo dataset \cite{sun2020scalability} which has more variability than \kitti.
        \cref{tab:deviant_waymo_val} shows the results on \waymo \val split. 
        The results show that \deviant outperforms the baseline \gupNet{} \cite{lu2021geometry} on multiple levels and multiple thresholds. 
        The biggest gains are on the nearby objects which is consistent with \cref{tab:deviant_detection_results_kitti_test_cars,tab:deviant_detection_results_kitti_val1}.
        Interestingly, \deviant also outperforms PatchNet \cite{ma2020rethinking} and PCT \cite{wang2021progressive} without using depth. 
        Although the performance of \deviant lags \caddn \cite{reading2021categorical}, it is important to stress that \caddn~uses \lidar{} data in training, while \deviant is an \imageOnly{} method. 

    \begin{table*}[!tb]
        \caption[\waymo \val vehicles detection results.]{\textbf{\waymo \val vehicles detection results}.
        [Key: \firstKey{Best}, \secondKey{Second Best}]}
        \label{tab:deviant_waymo_val}
        \centering
        \scalebox{0.85}{
            \setlength{\tabcolsep}{0.1cm}
            \begin{tabular}{m c m c m c m c m c | ccc m c | ccc m}
                \myTopRule
                & &  &  & \multicolumn{4}{cm}{\apThreeD{} \bracketPercentage(\uparrowRHDSmall)} & \multicolumn{4}{cm}{\apHThreeD~\bracketPercentage(\uparrowRHDSmall)} \\
                \cline{5-12}
                \multirow{-2}{*}{\iouThreeD} & \multirow{-2}{*}{Difficulty} & \multirow{-2}{*}{Method} & \multirow{-2}{*}{Extra} & All & 0-30 & 30-50 & 50-$\infty$ & All & 0-30 & 30-50 & 50-$\infty$ \\ 
                \myTopRule
                & & \caddn \cite{reading2021categorical} & \lidar & $5.03$ & $14.54$ & $1.47$ & $0.10$ & $4.99$ & $14.43$ & $1.45$ & $0.10$ \\
                & & \patchNet \cite{ma2020rethinking} in \cite{wang2021progressive} & Depth & $0.39$ & $1.67$ & $0.13$ & $0.03$ & $0.39$ & $1.63$ & $0.12$ & $0.03$ \\
                & & PCT \cite{wang2021progressive} & Depth & $0.89$ & $3.18$ & $0.27$ & $0.07$ & $0.88$ & $3.15$ & $0.27$ & $0.07$ \\
                $0.7$ & \levelOne & \mthreeDRPN \cite{brazil2019m3d} in \cite{reading2021categorical} & \mathDash{}& $0.35$ & $1.12$ & $0.18$ & $0.02$ & $0.34$ & $1.10$ & $0.18$ & $0.02$ \\
                & & \gupNet{} (Retrained)\cite{lu2021geometry} & \mathDash{}& \second{2.28} & \second{6.15} & \second{0.81} &	\first{0.03} & \second{2.27} & \second{6.11} & \second{0.80} &	\first{0.03}\\
                & & \textbf{\deviant (Ours)} & \mathDash{}& \first{2.69}	& \first{6.95} &	\first{0.99} &	\second{0.02} & \first{2.67}	& \first{6.90} &	\first{0.98} &	\second{0.02}\\
                \myTopRule
                & & \caddn \cite{reading2021categorical} & \lidar & $4.49$ & $14.50$ & $1.42$ & $0.09$ & $4.45$ & $14.38$ & $1.41$ & $0.09$ \\
                & & \patchNet \cite{ma2020rethinking} in \cite{wang2021progressive} & Depth & $0.38$ & $1.67$ & $0.13$ & $0.03$ & $0.36$ & $1.63$ & $0.11$ & $0.03$ \\
                &  & PCT \cite{wang2021progressive} & Depth & $0.66$ & $3.18$ & $0.27$ & $0.07$ & $0.66$ & $3.15$ & $0.26$ & $0.07$\\
                $0.7$ & \levelTwo & \mthreeDRPN \cite{brazil2019m3d} in \cite{reading2021categorical} & \mathDash{}& $0.33$ & {1.12} & {0.18} & \first{0.02} & $0.33$ & $1.10$ & $0.17$ & \first{0.02} \\
                & & \gupNet{} (Retrained)\cite{lu2021geometry} & \mathDash{}& \second{2.14} & \second{6.13} & \second{0.78} &	\first{0.02} & \second{2.12} & \second{6.08} & \second{0.77} &	\first{0.02} \\ 
                & & \textbf{\deviant (Ours)} & \mathDash{}& \first{2.52}	& \first{6.93} &	\first{0.95} &	\first{0.02} & \first{2.50}	& \first{6.87} &	\first{0.94} &	\first{0.02}\\
                \myTopRule
                & & \caddn \cite{reading2021categorical} & \lidar & $17.54$ & $45.00$ & $9.24$ & $0.64$ & $17.31$ & $44.46$ & $9.11$ & $0.62$ \\
                & & \patchNet \cite{ma2020rethinking} in \cite{wang2021progressive} & Depth & $2.92$ & $10.03$ & $1.09$ & $0.23$ & $2.74$ & $9.75$ & $0.96$ & $0.18$ \\
                & & PCT \cite{wang2021progressive} & Depth & $4.20$ & $14.70$ & $1.78$ & $0.39$ & $4.15$ & $14.54$ & $1.75$ & $0.39$\\
                $0.5$ & \levelOne & \mthreeDRPN \cite{brazil2019m3d} in \cite{reading2021categorical} & \mathDash{}& $3.79$ & $11.14$ & $2.16$ & \first{0.26} & $3.63$ & $10.70$ & $2.09$ & \second{0.21} \\
                & & \gupNet{} (Retrained)\cite{lu2021geometry} & \mathDash{}& \second{10.02} & \second{24.78} & \second{4.84} &	\second{0.22} & \second{9.94} & \second{24.59} & \second{4.78} &	\first{0.22} \\ 
                & & \textbf{\deviant (Ours)} & \mathDash{}& \first{10.98} & \first{26.85} & \first{5.13} &	$0.18$ & \first{10.89} & \first{26.64} & \first{5.08} & $0.18$  \\
                \myTopRule
                & & \caddn \cite{reading2021categorical} & \lidar & $16.51$ & $44.87$ & $8.99$ & $0.58$ & $16.28$ & $44.33$ & $8.86$ & $0.55$ \\
                & & \patchNet \cite{ma2020rethinking} in \cite{wang2021progressive} & Depth & $2.42$ & $10.01$ & $1.07$ & $0.22$ & $2.28$ & $9.73$ & $0.97$ & $0.16$ \\
                & & PCT \cite{wang2021progressive} & Depth & $4.03$ & $14.67$ & $1.74$ & $0.36$ & $4.15$ & $14.51$ & $1.71$ & $0.35$\\
                $0.5$ & \levelTwo  & \mthreeDRPN \cite{brazil2019m3d} in \cite{reading2021categorical} & \mathDash{}& $3.61$ & $11.12$ & $2.12$ & \first{0.24} & $3.46$ & $10.67$ & $2.04$ & \first{0.20} \\
                & & \gupNet{} (Retrained)\cite{lu2021geometry} & \mathDash{}& \second{9.39} & \second{24.69} & \second{4.67} &	\second{0.19} & \second{9.31} & \second{24.50} & \second{4.62} &	\second{0.19} \\ 
                & & \textbf{\deviant (Ours)} & \mathDash{}& \first{10.29} & \first{26.75} & \first{4.95} &	$0.16$ & \first{10.20} & \first{26.54} & \first{4.90} & $0.16$  \\
                \myTopRule
            \end{tabular}
        }
    \end{table*}

\section{Conclusions}
    This chapter studies the modeling error in monocular \threeD detection in detail and takes the first step towards convolutions equivariant to arbitrary \threeD translations in the projective manifold.
    Since the depth is the hardest to estimate for this task, this
    chapter proposes \deviantFull (\deviant) built with existing \scaleEquivariant{} steerable blocks. 
    As a result, \deviant is \equivariant{} to the depth translations in the projective \manifold{} whereas vanilla networks are not.
    The additional \depthEquivariance{} forces the \deviant to learn consistent depth estimates and therefore, \deviant achieves \sota{} detection results on \kitti and \waymo datasets in the \imageOnly{} category and performs competitively to methods using extra information.
    Moreover, \deviant works better than vanilla networks in cross-dataset evaluation.
    Future works include applying the idea to \pseudoLidar{} \cite{wang2019pseudo}, and monocular \threeD tracking.

    \noIndentHeading{Limitation.} \deviant does not model \threeD equivariance but only a special case of \threeD equivariance.
    Considerably less number of boxes are detected in the cross-dataset evaluation. 

%% file: images/deviant/teaser.tex
\begin{tikzpicture}[scale=0.21, every node/.style={scale=0.35}, every edge/.style={scale=0.35}]
\tikzset{vertex/.style = {shape=circle, draw=black!70, line width=0.06em, minimum size=1.4em}}
\tikzset{edge/.style = {-{Triangle[angle=60:.06cm 1]},> = latex'}}

    \draw [draw=axisShade, line width=0.05em, shorten <=0.5pt, shorten >=0.5pt, >=stealth]
           (0.5,2.0) node[]{}
        -- (18.2,2.0) node[]{};
    
    \draw [-{Triangle[angle=60:.1cm 1]}, line width=0.04em, shorten <=0.5pt, shorten >=0.5pt, >=stealth]
           (2.25, 2.0) node[]{}
        -- (14.75,2.0) node[pos=0.5, scale= 1.75, align= center]{{$\transX$~~~~}\\};
    
    \input{images/deviant/teaser_common}
    
    \draw[draw=rayShade, fill=rayShade, thick](7.5,10.0) circle (0.35) node[scale= 1.25]{};
    
    \draw [draw=rayShade, line width=0.1em, shorten <=0.5pt, shorten >=0.5pt, >=stealth]
       (7.5,10.0) node[]{}
    -- (4.65,6.2) node[]{};

    \node [inner sep=1pt, scale= 1.75] at (1.5, 4.7)  {$\projectionOne$};
    \node [inner sep=1pt, scale= 1.75] at (-2.65, -1.3)  {$\projectionTwo$};
    
    \draw [draw=projectionBorderShade, line width=0.02em, fill=projectionFillShade]    (9.2, 3.2) rectangle (13.8, 6.2) node[]{};
    \node[inner sep=0pt] (input) at (12.4,4.8) {\includegraphics[height=1.5cm]{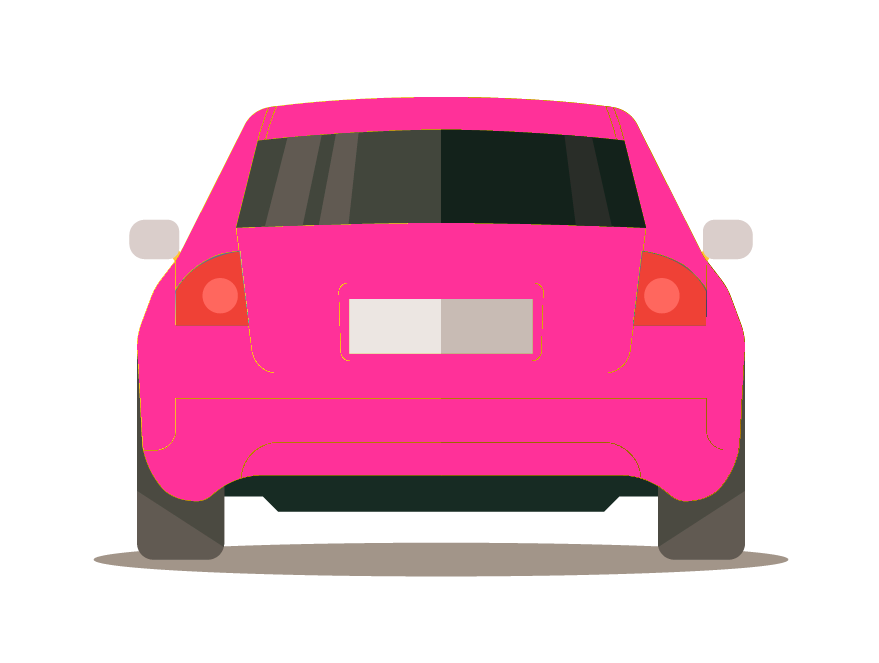}};
    \node [inner sep=1pt, scale= 1.75] at (14.5, 4.8)  {$\projectionTwo$};
        
    \draw [draw=rayShade, line width=0.1em, shorten <=0.5pt, shorten >=0.5pt, >=stealth]
           (7.5,10.0) node[]{}
        -- (11.08,6.2) node[]{};
    
    \draw [draw=rayShade, line width=0.1em, shorten <=0.5pt, shorten >=0.5pt, >=stealth]
           (12.4,4.8) node[]{}
        -- (15.4,2.0) node[]{}; 
    
    \draw[black,fill=black!50,rotate around={138:(15.4,2.0)}] (14.8,1.6) rectangle (16.0,2.4);
    \coordinate (c20) at (15.0, 2.42);
    \coordinate (c21) at (14.9, 2.9);
    \coordinate (c22) at (14.5, 2.38);
    \filldraw[draw=black, fill=gray!20] (c20) -- (c21) -- (c22) -- cycle;

    \draw [-{Triangle[angle=60:.1cm 1]}, draw=vanillaShade, line width=0.1em, shorten <=0.5pt, shorten >=0.5pt, >=stealth]
           (5.25, 4.5) node[]{}
        -- (10.75,4.5) node[pos=0.5, scale= 2.5, align= center]{\color{vanillaShade}{$\transU$}\\};

    \draw[draw=axisShadeDark, fill=axisShadeDark, thick](-1.65,10.0) circle (0.08) node[]{};
    
    \draw [-{Triangle[angle=60:.1cm 1]}, draw=axisShadeDark, line width=0.05em, shorten <=0.5pt, shorten >=0.5pt, >=stealth]
           (-1.65,10.0) node[]{}
        -- (0.3,10.0) node[scale= 1.75]{~~$x$};
        
    \draw [-{Triangle[angle=60:.1cm 1]}, draw=axisShadeDark, line width=0.05em, shorten <=0.5pt, shorten >=0.5pt, >=stealth]
           (-1.65,10.0) node[]{}
        -- (-1.65,8.35) node[text width=1cm,align=center,scale= 1.75]{~\\$y$};
    
    \draw [-{Triangle[angle=60:.1cm 1]}, draw=axisShadeDark, line width=0.05em, shorten <=0.5pt, shorten >=0.5pt, >=stealth]
           (-1.65,10.0) node[]{}
        -- (-0.75,11.0) node[scale= 1.75]{~~$z$};    
    
    \vspace{-0.48cm}
\end{tikzpicture}    

%% file: images/deviant/depth_equivariance.tex
\begin{tikzpicture}[scale=0.35, every node/.style={scale=0.58}, every edge/.style={scale=0.58}]
\tikzset{vertex/.style = {shape=circle, draw=black!70, line width=0.06em, minimum size=1.4em}}
\tikzset{edge/.style = {-{Triangle[angle=60:.06cm 1]},> = latex'}}


\node[inner sep=0pt, thick] (input) at (0,0) {\includegraphics[trim={5cm 0 5cm 0}, clip, height=2.0cm]{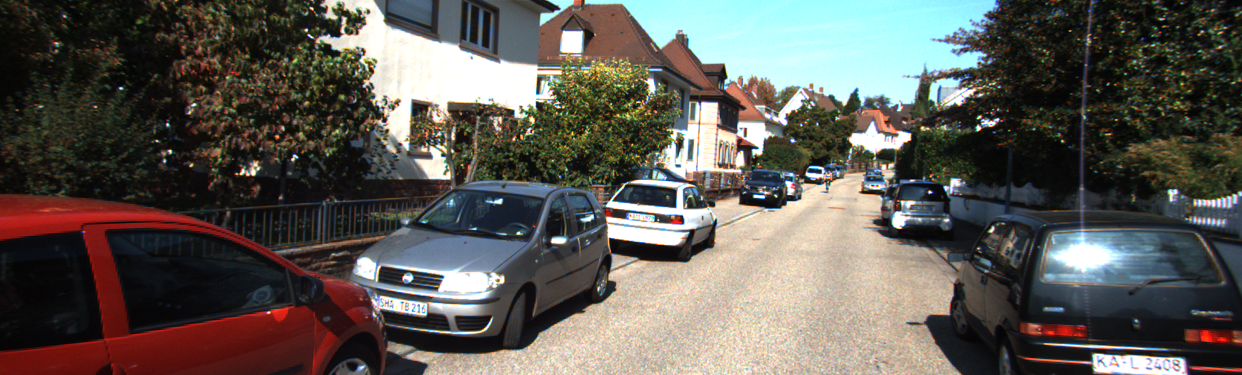}};
\draw [draw=projectionBorderShade, line width=0.05em]    (-4.25, -1.65) rectangle (4.25, 1.65) node[]{};

\draw [draw=projectionBorderShade, line width=0.05em, fill=black!80]    (-4.25, -7.65) rectangle (4.25, -4.35) node[]{};
\node[inner sep=0pt] (input) at (0,-6.0) {\includegraphics[trim={5cm 0 5cm 0}, clip, height=1.3cm]{images/deviant/000008.png}};

\draw [draw=projectionBorderShade, line width=0.05em]    (10.7, -1.7) rectangle (19.3, 1.7) node[]{};
\node[inner sep=0pt, thick] (input) at (15,0) {\includegraphics[trim={2.3cm 0 2.3cm 0}, clip, height=2.0cm]{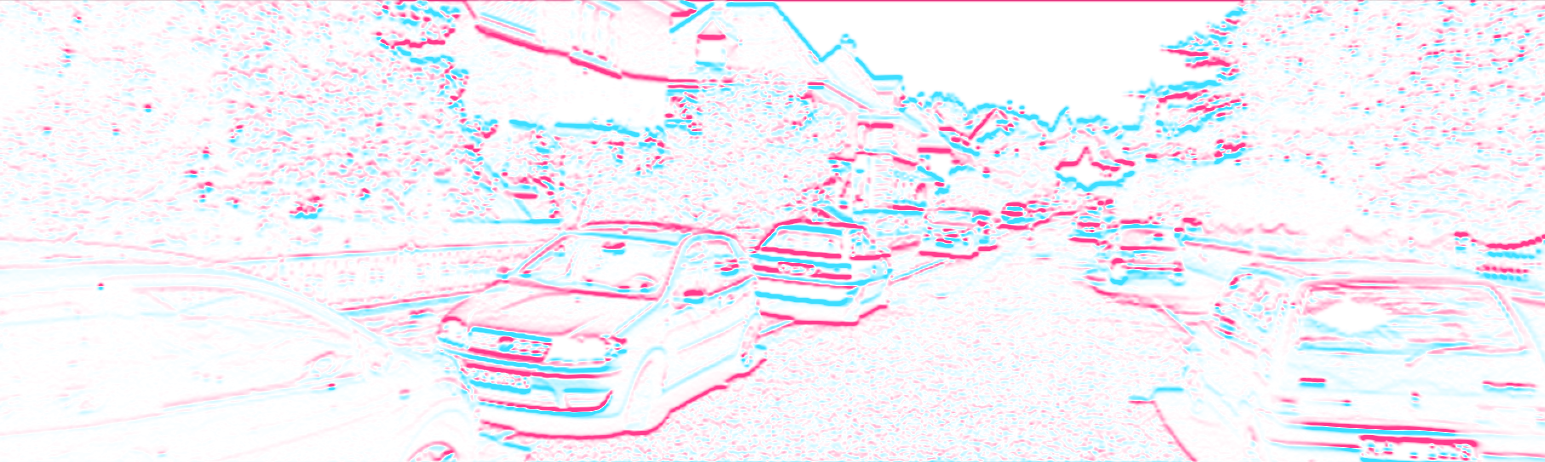}};

\draw [draw=projectionBorderShade, line width=0.05em, fill=white]    (10.75, -7.65) rectangle (19.25, -4.35) node[]{};
\node[inner sep=0pt, thick] (input) at (15,-6) {\includegraphics[trim={2.2cm 0 2.2cm 0}, clip, height=1.3cm]{images/deviant/filter_1.png}};

\draw [-{Triangle[angle=60:.1cm 1]}, draw=proposedShade, line width=0.1em, shorten <=0.5pt, shorten >=0.5pt, >=stealth]
       (0, -1.65) node[]{}
    -- (0,-4.35) node[pos=0.5, scale= 1.1, align= center]{~~~~~~~~~~~~~~~~~~~~~~~~~~~~~Depth Translation\\~~~~~~~~~~~~~~~~~~~~~~~~~~~~$=\transformationMath_\scaleNotation$(\cref{th:projective_scaled})};

\draw [-{Triangle[angle=60:.1cm 1]}, draw=proposedShade, line width=0.1em, shorten <=0.5pt, shorten >=0.5pt, >=stealth]
       (15, -1.65) node[]{}
    -- (15,-4.35) node[pos=0.5, scale= 1.1, align= center]{~~~~~~~~~~~~~~~~~~~~~~~~~~~~~~Depth Translation\\~~~~~~~~~~~~~$=\transformationMath_\scaleNotation$};

\draw [-{Triangle[angle=60:.1cm 1]}, draw=black, line width=0.05em, shorten <=0.5pt, shorten >=0.5pt, >=stealth]
       (4.25, -6) node[]{}
    -- (10.75, -6) node[pos=0.5, scale= 2.4, align= center]{};
\node [inner sep=1pt, scale= 2, align= center] at (7.5, -5.75)  {*~~~~~~~~~~~~~~};

\draw [-{Triangle[angle=60:.1cm 1]}, draw=black, line width=0.05em, shorten <=0.5pt, shorten >=0.5pt, >=stealth]
       (4.25, 0) node[]{}
    -- (10.75, 0) node[pos=0.5, scale= 2.4, align= center]{};
\node [inner sep=1pt, scale= 2, align= center] at (7.5, 0.25)  {*~~~~~~~~~~~~~~};

\draw [draw=projectionBorderShade, line width=0.05em, fill=white]    (5.8, 1.2) rectangle (8.2, -7.2) node[]{};
\node [inner sep=1pt, scale= 1.2, align= center] at (7.5, 1.8)  {\ses{} Convolution};

\node[inner sep=0pt, thick] (input) at (7,0) {\includegraphics[trim={0cm 0cm 0cm 22cm}, clip, height=1.25cm]{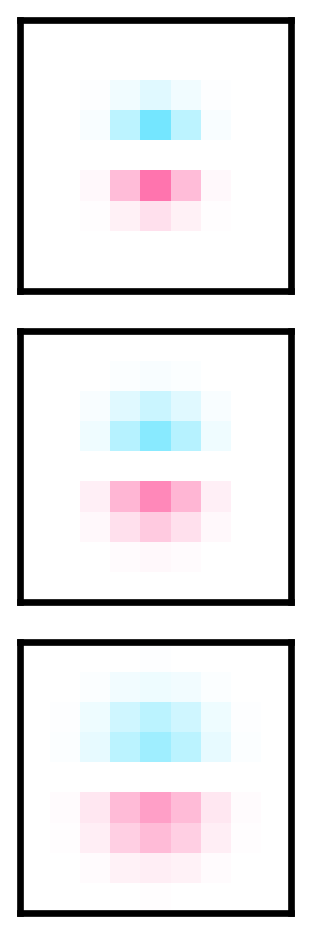}};
\node[inner sep=0pt, thick] (input) at (7,-6) {\includegraphics[trim={0cm 22cm 0cm 0cm}, clip, height=1.25cm]{images/deviant/sesn_basis_sample.png}};

\draw [-{Triangle[angle=60:.1cm 1]}, draw=black, densely dashed, line width=0.04em, shorten <=0.5pt, shorten >=0.5pt, >=stealth]
       (7, -5.1) node[]{}
    -- (7, -0.9) node[pos=0.5, scale= 0.9, align= center]{~~~~~~~~~$\transformationMath_{\scaleNotation^{-1}}$};

\end{tikzpicture}

%% file: chapters/seabird.tex
\chapter{
    \seabird: \seabirdFull with Dice Loss Improves Monocular 3D Detection of Large Objects
}
\label{chpt:seabird}


Monocular \threeD detectors achieve remarkable performance on cars and smaller objects. 
However, their performance drops on larger objects, leading to fatal accidents. 
Some attribute the failures to training data scarcity or their receptive field requirements of large objects.
In this chapter, we highlight this understudied problem of generalization to large objects.
We find that modern frontal detectors struggle to generalize to large objects even on nearly balanced datasets.
We argue that the cause of failure is the sensitivity of depth regression losses to noise of larger objects.
To bridge this gap, we comprehensively investigate regression and dice losses, examining their robustness under varying error levels and object sizes.
We mathematically prove that the dice loss leads to superior noise-robustness and model convergence for large objects compared to regression losses for a simplified case.
Leveraging our theoretical insights, we propose \seabird (\seabirdFull) as the first step towards generalizing to large objects.
\seabird effectively integrates BEV segmentation on foreground objects for 3D detection, with the segmentation head trained with the dice loss.
\seabird achieves \sota results on the \kittiThreeSixty leaderboard and improves existing detectors on the \nuscenes leaderboard, particularly for large objects. 


\section{Introduction}\label{sec:seabird_intro}

    Monocular \threeD object detection (\monoThreeD) task aims to estimate both the \threeD position and dimensions of objects in a scene from a single image. 
    Its applications span autonomous driving \cite{park2021pseudo,kumar2022deviant,li2022bevformer}, robotics \cite{saxena2008robotic}, and augmented reality \cite{alhaija2018augmented,Xiang2018RSS,park2019pix,merrill2022symmetry}, where accurate \threeD understanding of the environment is crucial. 
    Our study focuses explicitly on \threeD object detectors applied to autonomous vehicles (AVs), considering the challenges and motivations deviate drastically across different applications.

    \begin{figure}[!t]
        \centering
        \begin{minipage}[t]{.28\linewidth}
            \includegraphics[width=\linewidth]{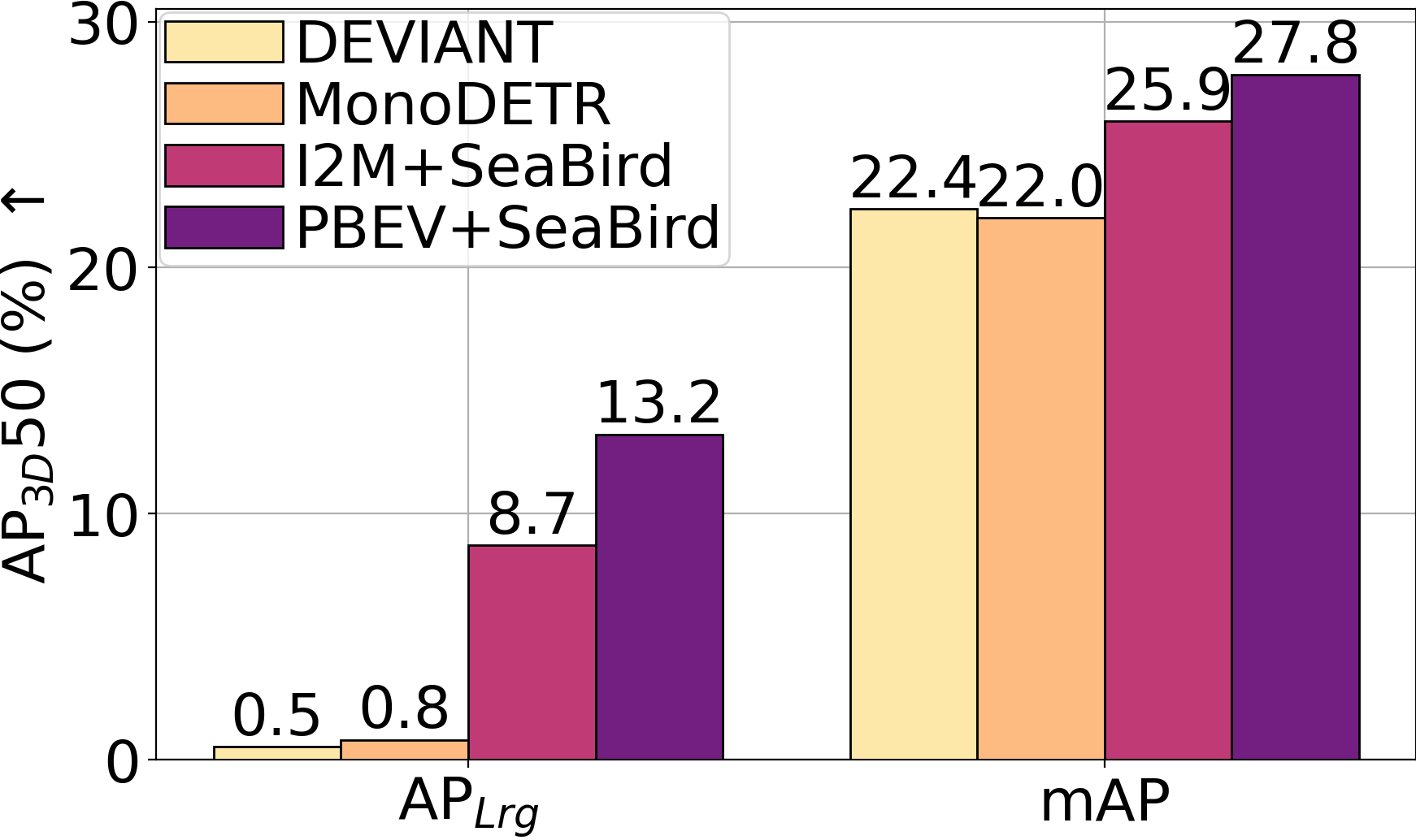}\\
            \vspace{-0.6cm}
            \captionof*{figure}{(a) Improve \kittiThreeSixty \sota.}
        \end{minipage}%
        \begin{minipage}[t]{.41\linewidth}
            \includegraphics[width=\linewidth]{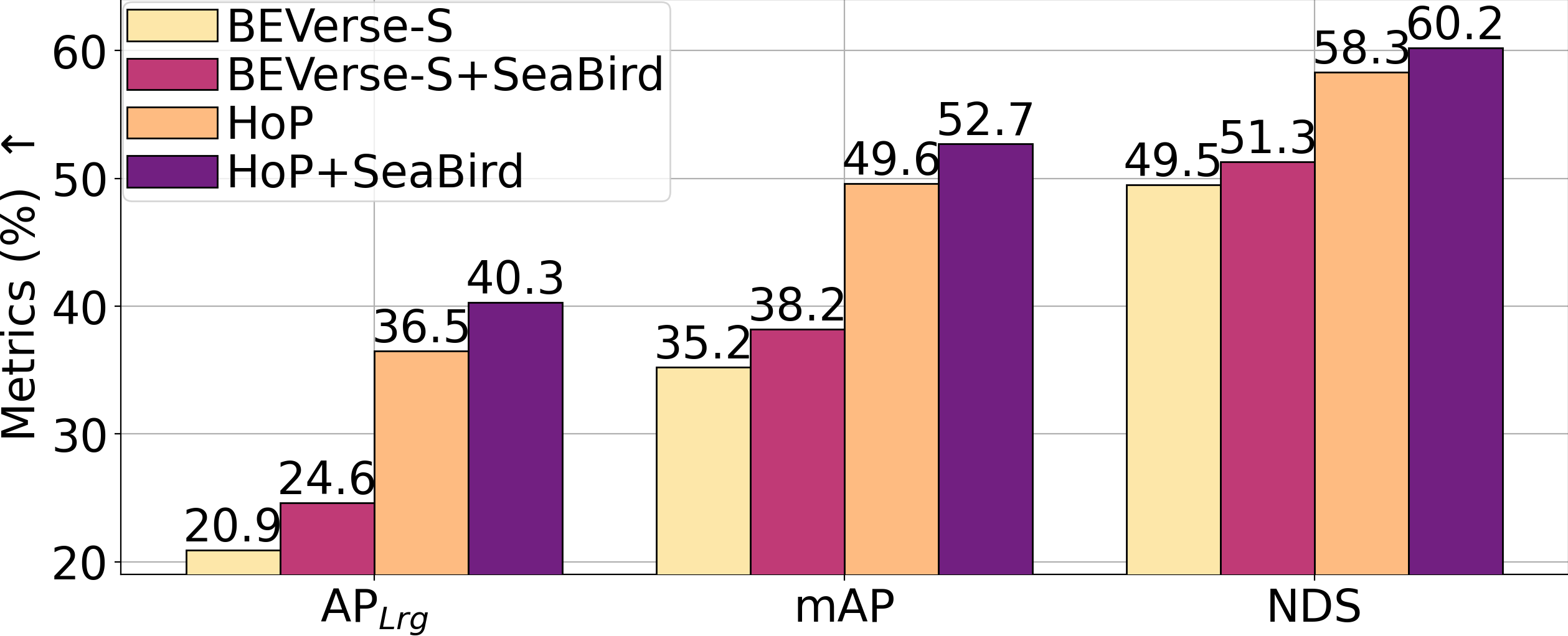}\\
            \vspace{-0.6cm}
            \captionof*{figure}{(b) Improve \nuscenes \val \sota.}
        \end{minipage}
        \begin{minipage}[t]{.28\linewidth}
            \includegraphics[width=\linewidth]{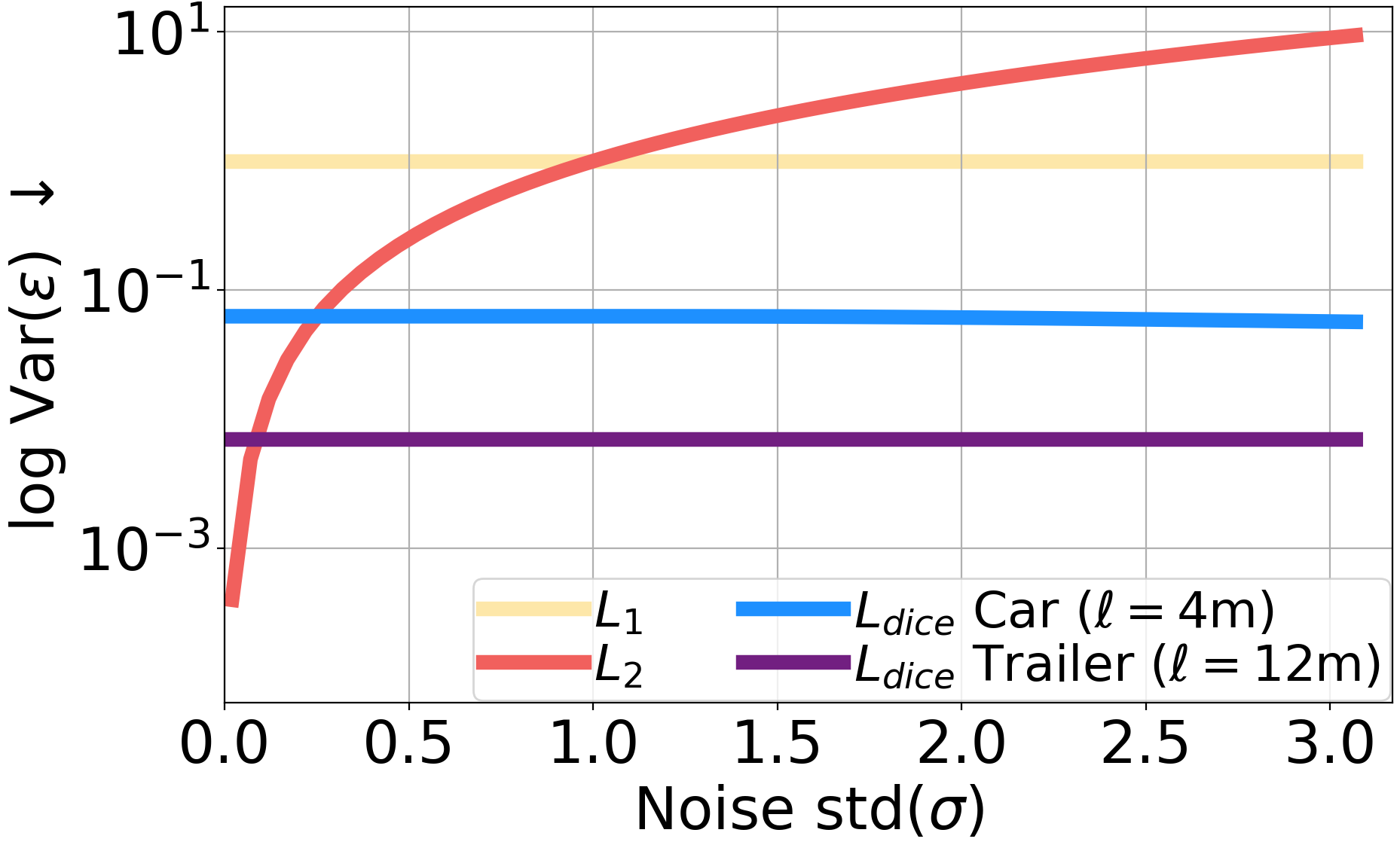}\\
            \vspace{-0.6cm}
            \captionof*{figure}{(c) Theory Advancement.}
        \end{minipage}
        \vspace{-0.2cm}
        \caption[\seabird Teaser]
        {\textbf{Teaser (a)} \sota frontal detectors struggle with large objects (low \MAPLarge) even on a nearly balanced \kittiThreeSixty dataset.
            Our proposed \seabird achieves significant \monoThreeD improvements, particularly for large objects. 
            \textbf{(b)} \seabird also improves two \sota \bev detectors, \beVerseSmall \cite{zhang2022beverse} and \hop \cite{zong2023hop} on the \nuscenes dataset, particularly for large objects.
            \textbf{(c)} Plot of convergence variance $\var(\funcNoise)$ of \dice and regression losses with the noise $\normalSig$ in depth prediction. 
            The $y$-axis denotes the deviation from the optimal weight, so the lower the better. \seabird leverages \textbf{\dice loss}, which we prove is more noise-robust than regression losses for large objects.
        }
        \label{fig:seabird_teaser}
    \end{figure}

    AVs demand object detectors that generalize to diverse intrinsics \cite{brazil2023omni3d}, camera-rigs \cite{jia2023monouni,tzofi2023towards}, rotations \cite{moon2023rotation}, weather and geographical conditions \cite{dong2023benchmarking} and also are robust to adversarial examples \cite{zhu2023understanding}.
    Since each of these poses a significant challenge, recent works focus exclusively on the generalization of object detectors to all these out-of-distribution shifts. 
    However, our focus is on the generalization of another type, which, thus far, has been understudied in the literature – {\it \monoThreeD generalization to large objects}. 
    
    Large objects like trailers, buses and trucks are harder to detect \cite{wu2023talk} in \monoThreeD, sometimes resulting in fatal accidents \cite{caldwell2022tesla,fernandez2023tesla}. 
    Some attribute these failures to training data scarcity \cite{zhu2019class} or the receptive field requirements \cite{wu2023talk} of large objects, but, to the best of our knowledge, no existing literature provides a comprehensive analytical explanation for this phenomenon.
    The goal of this chapter is, thus, to bring understanding and a first analytical approach to this real-world problem in the AV space – \monoThreeD generalization to large objects. 

    We conjecture that the generalization issue stems not only from limited training data or larger receptive field but also from the noise sensitivity of depth regression losses in \monoThreeD. 
    To substantiate our argument, we analyze the \monoThreeD performance of state-of-the-art (\sota) frontal detectors on the \kittiThreeSixty dataset \cite{liao2022kitti360}, which includes almost equal number ($1\!:\!2$) of large objects and cars.
    We observe that \sota detectors struggle with large objects on this  dataset (\cref{fig:seabird_teaser}\textcolor{link_color}{a}).
    Next, we carefully investigate the SGD convergence of losses used in \monoThreeD task and mathematically prove that the \dice loss, widely used in \bev segmentation, exhibits superior noise-robustness than the regression losses, particularly for large objects (\cref{fig:seabird_teaser}\textcolor{link_color}{c}).
    Thus, the \dice loss facilitates better model convergence than regression losses, improving \monoThreeD of large objects.

    Incorporating \dice loss in detection introduces unique challenges. 
    Firstly, the \dice loss does not apply to sparse detection centers and only incorporates depth information when used in the \bev space. 
    Secondly, naive joint training of \monoThreeD and \bev segmentation tasks with image inputs does not always benefit \monoThreeD task \cite{li2022bevformer,ma2022vision} due to negative transfer \cite{crawshaw2020multi}, and the underlying reasons remain unclear. 
    Fortunately, many \monoThreeD segmentors and detectors are in the \bev space, where the \bev segmentor can seamlessly apply \dice loss and the \bev detector can readily benefit from the segmentor in the same space. 
    To mitigate negative transfer, we find it effective to train the \bev segmentation head on the foreground detection categories. 
    
\begin{figure*}[!t]
        \centering
        \includegraphics[width=\linewidth]{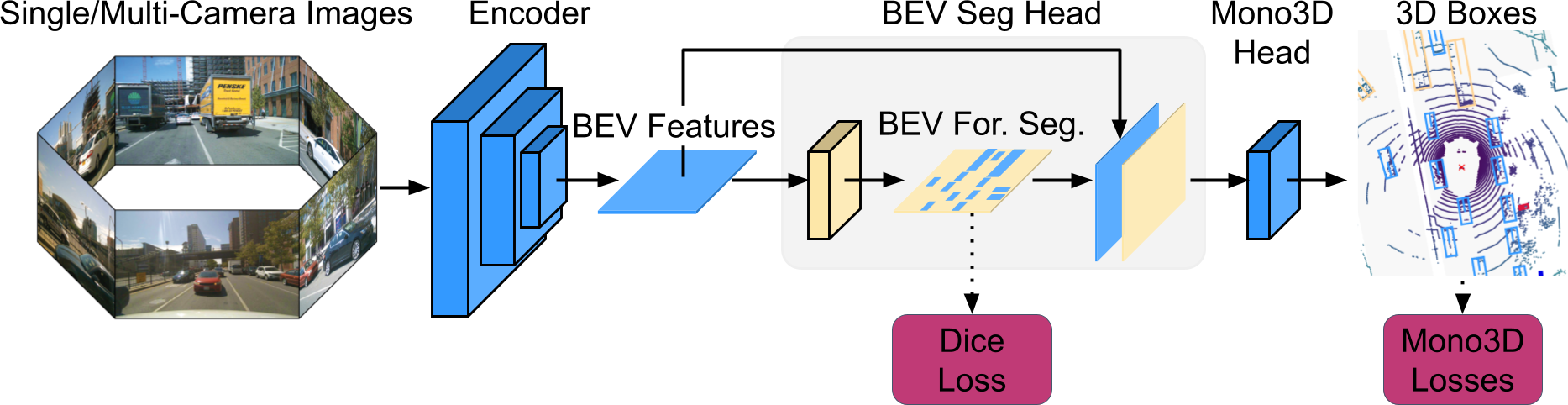}
        \caption[\seabird Pipeline.]
        {
            \textbf{\seabird Pipeline.} 
            \seabird uses the predicted \bev foreground segmentation (For. Seg.) map to predict accurate \threeD boxes for large objects.
            \seabird training protocol involves \bev segmentation pre-training with the noise-robust \dice loss and \monoThreeD fine-tuning.
        }
        \label{fig:seabird_pipeline}
    \end{figure*}

    Building upon our theoretical findings about the \dice loss, we propose a simple and effective pipeline called \seabirdFull (\seabird) for enhancing \monoThreeD of large objects. 
    \seabird employs a sequential approach for the \bev segmentation and \monoThreeD heads (\cref{fig:seabird_pipeline}).  
    \seabird first utilizes a \bev segmentation head to predict the segmentation of only foreground objects, supervised by the \dice loss. 
    The \dice loss offers superior noise-robustness for large objects, ensuring stable convergence, while focusing on foreground objects in segmentation mitigates negative transfer. 
    Subsequently, \seabird concatenates the resulting \bev segmentation map with the original \bev features as an additional feature channel and feeds this concatenated feature to a \monoThreeD head supervised by \monoThreeD losses\footnote{Only \monoThreeD head predicts additional \threeD attributes, namely object's height and elevation.}. 
    Building upon this, we adopt a two-stage training pipeline: the first stage exclusively focuses on training the \bev segmentation head with \dice loss, which fully exploits its noise-robustness and superior convergence in localizing large objects. The second stage involves both the detection loss and dice loss to finetune the \monoThreeD head.
    
    In our experiments, we first comprehensively evaluate \seabird and conduct ablations on the balanced single-camera \kittiThreeSixty dataset \cite{liao2022kitti360}. 
    \seabird outperforms the \sota baselines by a substantial margin. 
    Subsequently, we integrate \seabird as a plug-in-and-play module into two \sota detectors on the multi-camera \nuscenes dataset \cite{caesar2020nuscenes}.
    \seabird again significantly improves the original detectors, particularly on large objects. 
    Additionally, \seabird consistently enhances \monoThreeD performance across backbones with those two \sota detectors (\cref{fig:seabird_teaser}\textcolor{link_color}{b}), demonstrating its utility in both edge and cloud deployments.

    In summary, we make the following contributions:
    \begin{itemize}
        \item We highlight the understudied problem of generalization to large objects in \monoThreeD, showing that even on nearly balanced datasets, \sota frontal models struggle to generalize due to the noise sensitivity of regression losses.
        \item We mathematically prove that the \dice loss leads to superior noise-robustness and model convergence for large objects compared to regression losses for a simplified case and provide empirical support for more general settings.
        \item We propose \seabird, which 
        treats \bev segmentation head on foreground objects and \monoThreeD head sequentially and trains in a two-stage protocol to fully harness the noise-robustness of the \dice loss.
        \item We empirically validate our theoretical findings and show significant improvements, particularly for large objects, on both \kittiThreeSixty and \nuscenes leaderboards.
    \end{itemize}

\section{Related Works}

    \noIndentHeading{Mono3D.}
        \monoThreeD popularity stems from its high accessibility from consumer vehicles compared to \lidar/Radar-based detectors \cite{shi2019pointrcnn,yin2021center,long2023radiant} and computational efficiency compared to stereo-based detectors \cite{Chen2020DSGN}.
        Earlier approaches \cite{payet2011contours, chen2016monocular} leverage hand-crafted features, while the recent ones use deep networks. 
        Advancements include introducing new architectures \cite{shi2023multivariate,huang2022monodtr,xu2023mononerd}, equivariance \cite{kumar2022deviant, chen2023viewpoint}, losses \cite{brazil2019m3d,chen2020monopair}, uncertainty \cite{lu2021geometry,kumar2020luvli} and incorporating auxiliary tasks such as depth \cite{zhang2021objects,min2023neurocs}, 
        NMS \cite{shi2020distance,kumar2021groomed,liu2023monocular}, corrected extrinsics \cite{zhou2021monoef}, CAD models \cite{chabot2017deep, liu2021autoshape, lee2023baam} or \lidar \cite{reading2021categorical} in training.
        A particular line of work called \pseudoLidar \cite{wang2019pseudo, ma2019accurate} shows generalization by first estimating the depth, followed by a point cloud-based \threeD detector.
        
        Another line of work encodes image into latent \bev features \cite{ma2023towards} and attaches multiple heads for downstream tasks \cite{zhang2022beverse}. 
        Some focus on pre-training \cite{xie2022m2bev} and rotation-equivariant convolutions \cite{feng2022aedet}. 
        Others introduce new coordinate systems \cite{jiang2023polarformer}, queries \cite{luo2022detr4d,li2023fast}, or positional encoding \cite{shu2023dppe} in a transformer-based detection framework \cite{carion2020detr}.
        Some use pixel-wise depth \cite{huang2021bevdet}, object-wise depth \cite{chu2023oabev,choi2023depth,liu2021voxel}, or depth-aware queries \cite{zhang2023dabev}, while many utilize temporal fusion \cite{wang2022sts,wang2023stream,liu2023petrv2,brazil2020kinematic} to boost performance.
        A few use longer frame history \cite{park2022time,zong2023hop}, distillation \cite{klingner2023x3kd,wang2023distillbev} or stereo \cite{wang2022sts,li2023bevstereo}.
        We refer to \cite{ma20233d,ma2022vision} for the survey.
        \seabird also builds upon the \bev-based framework since it flexibly accepts single or multiple images as input and uses \dice loss. 
        Different from the majority of other detectors, 
        \seabird improves \monoThreeD of large objects using the power of \dice loss. 
        \seabird is also the first work to mathematically prove and justify this loss choice for large objects.

    \noIndentHeading{\bev Segmentation.}
        \bev segmentation typically utilizes \bev features transformed from \twoD image features. Various methods encode single or multiple images into \bev features using MLPs \cite{pan2020cross} or transformers \cite{roddick2020predicting,saha2022translating}. 
        Some employ learned depth distribution \cite{philion2020lift,hu2021fiery}, while others use attention \cite{saha2022translating,zhou2022cross} or attention fields \cite{chitta2021neat}. 
        \imageToMapsLong \cite{saha2022translating} utilizes polar ray, while \panopticBEVLong \cite{gosala2022bev} uses transformers. 
        FIERY \cite{hu2021fiery} introduces uncertainty modelling and temporal fusion, while Simple-\bev \cite{harley2022simple} uses \radar aggregation.
        Since \bev segmentation lacks object height and elevation, one also needs a \monoThreeD head to predict \threeD boxes.

    \noIndentHeading{Joint \monoThreeD and \bev Segmentation.}
        Joint \threeD detection and \bev segmentation using \lidar data \cite{shi2019pointrcnn,fan2022fully} as input benefits both tasks \cite{yang2023lidar,wang2023segmentation}. 
        However, joint learning on image data often hinders detection performance \cite{li2022bevformer,zhang2022beverse,xie2022m2bev,ma2022vision}, while the \bev segmentation improvement is inconsistent across categories \cite{ma2022vision}.
        Unlike these works which treat the two heads in parallel and decrease \monoThreeD performance \cite{ma2022vision}, \seabird treats the heads sequentially and increases \monoThreeD performance, particularly for large objects.

\section{\seabird}\label{sec:seabird_proposed}

    \seabird is driven by a deep understanding of the distinctions between monocular regression and \bev segmentation losses. 
    Thus, in this section, we delve into the problem and discuss existing results. 
    We then present our theoretical findings and, subsequently, introduce our pipeline.
    
\begin{figure}[!t]
        \centering
        \input{images/seabird/problem_setup}
        \caption[Problem setup, shifting of {predictions} in \bev and Cross Section (CS) view along the ray. ]
        {
            \textbf{(a) Problem setup}. The single-layer neural network takes an image $\image$ (or its features) and predicts depth $\depthPred$ and the object length $\length$.
            The noise $\noise$ is the additive error in depth prediction and is a normal random variable.
            The GT depth $\depthGT$ supervises the predicted depth $\depthPred$ with a loss $\loss$ in training.
            We assume the network predicts the GT length $\length$.
            Frontal detectors directly regress the depth with $\lOne$, $\lTwo$, or $\smoothLOne$ loss, while \seabird projects to \bev plane and supervises through \dice loss $\lDice$.
            \textbf{(b) Shifting of {predictions} (blue)} in \bev along the {ray} due to the noise $\noise$.
            \textbf{(c) Cross Section (CS) view} along the {ray} with classification scores $P(Z)$.
        }
        \label{fig:seabird_problem_setup}
    \end{figure}
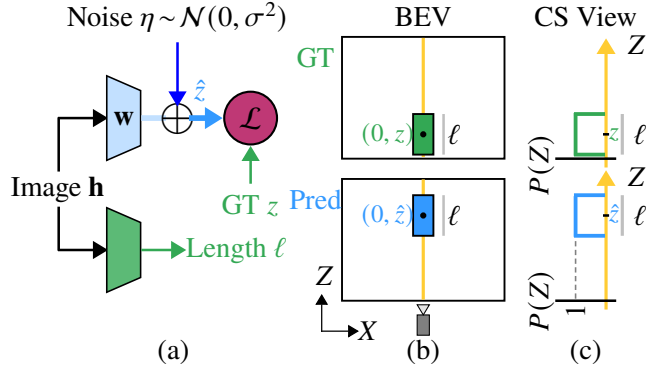

    We introduce the problem and refer to \cref{lemma:1} from the literature \cite{shalev2007pegasos, lacoste2012simpler}, which \emph{evaluates} loss quality by measuring the deviation of trained weight (after SGD updates) from the optimal weight. 
    \cref{fig:seabird_problem_setup}\textcolor{link_color}{a} illustrates the problem setup.
    Figs. \ref{fig:seabird_problem_setup}\textcolor{link_color}{b} and \ref{fig:seabird_problem_setup}\textcolor{link_color}{c} visualize the \bev and cross-section view, respectively. 
    Since this deviation depends on the gradient variance of losses, we next derive the gradient variance of the \dice loss in \cref{lemma:2}.
    By comparing the distance between trained weight and optimal weight, we assess the effectiveness of \dice loss versus MAE $(\lOne)$ and MSE $(\lTwo)$ losses in \cref{lemma:3}, and choose the representation and loss combination. 
    Combining these findings, we establish \cref{th:seabird_1} that the model trained with \dice loss achieves better \ap than the model trained with regression losses. 
    Finally, we present our pipeline, \seabird, which integrates \bev segmentation supervised by \dice loss for \monoThreeD.

    \subsection{Background and Problem Statement}
        \monoThreeD networks \cite{lu2021geometry, kumar2022deviant} commonly employ regression losses, such as $\lOne$ or $\lTwo$ loss, to compare the predicted depth with ground truth (GT) depth~\cite{kumar2022deviant, zhang2022beverse}. 
        In contrast, \bev segmentation utilizes \dice loss \cite{saha2022translating} or cross-entropy loss \cite{hu2021fiery} at each \bev location, comparing it with GT. 
        Despite these distinct loss functions, we evaluate their effectiveness under an idealized model, where we measure the model \emph{quality} by the expected deviation of trained weight (after SGD updates) from the optimal weight \cite{shalev2007pegasos}.

        \begin{lemma}\label{lemma:1}
            \textbf{Convergence analysis }\cite{shalev2007pegasos}.
            Consider a linear regression model with trainable weight $\layerWeight$ for depth prediction $\depthPred$ from an image $\image$.
            Assume the noise $\noise$ is an additive error in depth prediction and is a normal random variable $\normal(0, \normalVar)$.
            Also, assume SGD optimizes the model parameters with loss function $\loss$ during training with square summable steps $\step_j$, \thatIs $\stepSumTrue\!=\!\lim\limits_{t \rightarrow \infty} \sum\limits_{j=1}^\instant \step_j^2$ exists and $\noise$ is independent of the image. 
            Then, the expected deviation of the trained weight $\layerWeightConv$ from the optimal weight $\layerWeightOptimal$ obeys
            \begin{align}
                \label{eqn:conv:weight:dist}
                \expect\left(\norm{\layerWeightConv\!-\!\layerWeightOptimal}_2^2\right) &= \stepConstant \var(\funcNoise) + \uselessConstant,
            \end{align}
            where $\funcNoise\!=\!\frac{\partial \loss(\noise)}{\partial \noise}$ is the gradient of the loss $\loss$ wrt noise, $\stepConstant\!=\!\stepSumTrue\expect(\image^T\image)$ and $\uselessConstant$ are constants independent of the loss.
        \end{lemma}

        We refer to \cref{sec:seabird_proof_converged} for the proof.
        \cref{eqn:conv:weight:dist} demonstrates that training losses $\loss$ exhibit varying gradient variances $\var(\funcNoise)$. 
        Hence, comparing this term for different losses allows us to evaluate their quality.

    \subsection{Loss Analysis: \Dice vs. Regression}

        Given that \cite{shalev2007pegasos} provides the gradient variance $\var(\funcNoise)$, for $\lOne$ and $\lTwo$ losses, we derive the corresponding gradient variance for \dice and \iou losses in this chapter to facilitate comparison. 
        First, we express the \dice loss, $\lDice$, as a function of noise $\noise$ as per its definition from \cite{saha2022translating} for \cref{fig:seabird_problem_setup}\textcolor{link_color}{c} as:
        \begin{align}
            \lDice(\noise) = 1\!-\!2\frac{\text{Pred}~\text{GT}}{\text{Pred} + \text{GT}}
            &= \begin{cases}
                1\!-\!2\frac{\length-|\noise|}{2\length} \text{ , }|\noise|\le \length \\
                1 \qquad\qquad \text{, }|\noise|\ge \length
                \end{cases} \nonumber \\
            \implies \lDice(\noise) &= \begin{cases}
                \frac{|\noise|}{\length} \text{ , }|\noise|\le \length\\
                1\quad\text{, }|\noise|\ge \length 
               \end{cases} ,
            \label{eq:dice}
        \end{align}
        where $\length$ denotes the object length. 
        \cref{eq:dice} shows that the \dice loss $\lDice$ depends on the object size $\length$. 
        With the given \dice loss $\lDice$, we proceed to derive the following lemma:

    \begin{table}[!t]
        \caption[Convergence variance of training loss functions]{\textbf{Convergence variance} of training loss functions. 
        Gradient variance of $\lDice$ is more noise-robust for large objects, resulting in better detectors.
        We do not analyze cross-entropy loss theoretically since its Var$(\funcNoise)$ is infinite, but empirically in \cref{tab:seabird_ablation}.}
        \label{tab:seabird_optimality_bounds}
        \centering
        \scalebox{\scaleFraction}{
        \setlength\tabcolsep{0.1cm}
        \begin{tabular}{l|c|c}
            \myTopRule
            Loss $\loss$ & Gradient $\funcNoise$ & Var$(\funcNoise)$ (\downarrowRHDSmall)\\ 
            \hline
            $\lOne$ \cite{shalev2007pegasos} (App. \ref{sec:seabird_supp_var_lOne}) & $\sign(\noise)$ & $1$ \\
            $\lTwo$ \cite{shalev2007pegasos} (App. \ref{sec:seabird_supp_var_lTwo}) & $\noise$ & $\normalVar$ \\
            \Dice (\cref{lemma:2})& $\begin{cases}
                \frac{\sign(\noise)}{\length} \quad\text{  , }|\noise|\le \length \\
                0 \quad\quad~~~~\text{ , }|\noise|\ge \length 
            \end{cases}$& $\dfrac{1}{\length^2}\normalErf\left(\dfrac{\length}{\sqrt{2}\normalSig}\right)$\\
            \myTopRule
        \end{tabular}
        }
    \end{table}
    
    \begin{lemma}\label{lemma:2}
        \textbf{Gradient variance of \dice loss.}
        Let $\noise= \normal(0, \normalVar)$ be an additive normal random variable and $\length$ be the object length.
        Let $\normalErf$ be the error function.
        Then, the gradient variance of the \dice loss $\var_{dice}(\funcNoise)$ wrt noise $\noise$ is
        \begin{align}
            \var_{dice}(\funcNoise) &= \frac{1}{\length^2}\normalErf\left(\frac{\length}{\sqrt{2}\normalSig}\right).
            \label{eq:grad_var_dice}
        \end{align}
    \end{lemma}
        We refer to \cref{sec:seabird_supp_var_dice} for the proof.  
        \cref{eq:grad_var_dice} shows that gradient variance of the \dice loss $\var_{dice}(\funcNoise)$ also varies inversely to the object size $\length$ and the noise deviation $\normalSig$ (See \cref{sec:seabird_supp_dice_properties}). 
        These two properties of \dice loss are particularly beneficial for large objects. 

        \cref{tab:seabird_optimality_bounds} summarizes these losses, their gradients, and gradient  variances. 
        With $\var_{\dice}(\funcNoise)$ derived for the \dice loss, we now compare the deviation of trained weight with the deviations from $\lOne$ or $\lTwo$ losses, leading to our next lemma.

    \begin{figure}[!t]
        \centering
        \includegraphics[width=0.7\linewidth]{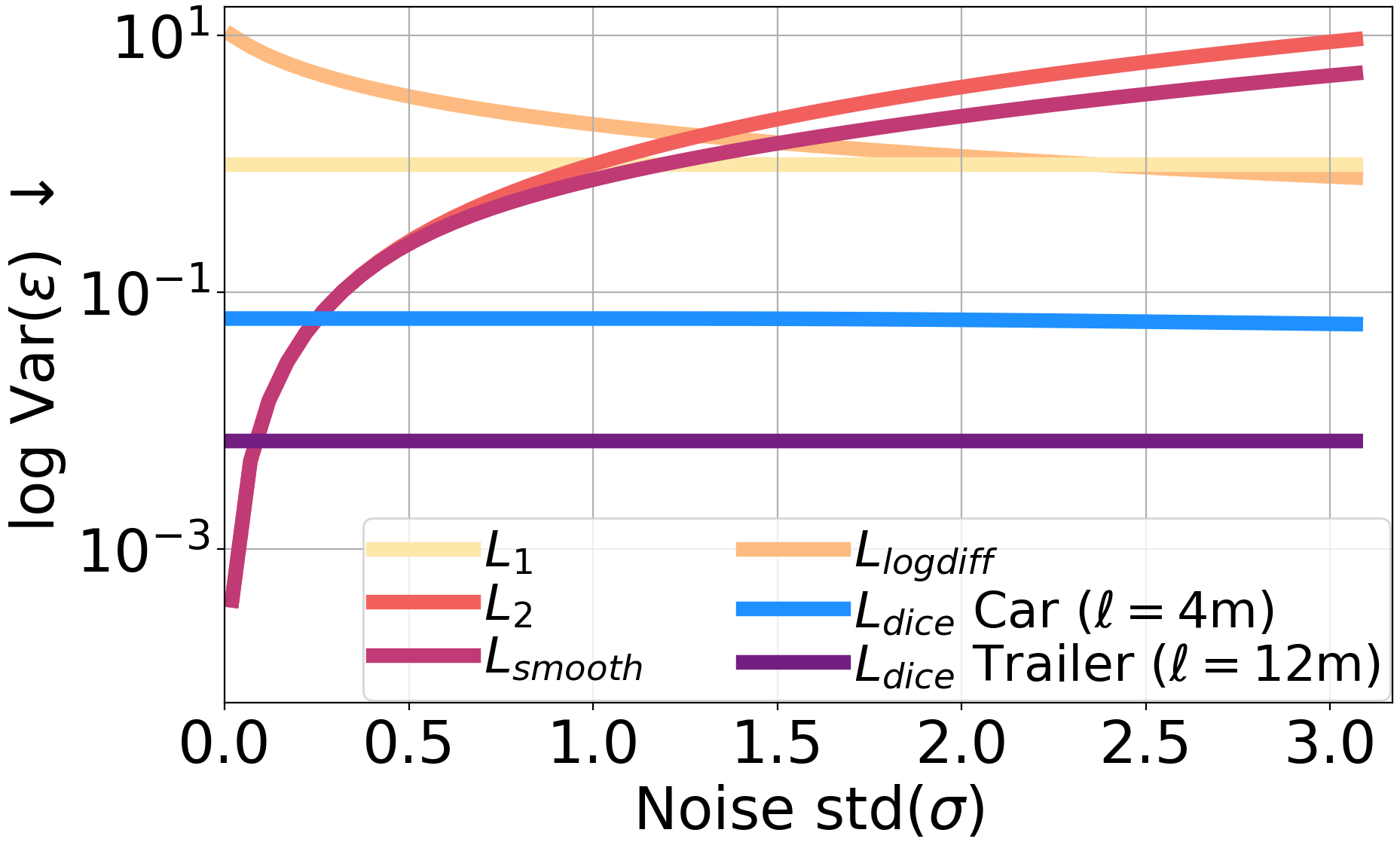}
        \caption[Plot of convergence variance Var$(\funcNoise)$ of loss functions with the noise $\normalSig$.]
        {\textbf{Plot of convergence variance} Var$(\funcNoise)$ of loss functions with the noise $\normalSig$. \Dice loss has minimum convergence variance with large noise, resulting in better detectors for large objects.}
        \label{fig:seabird_conv_analysis}
    \end{figure}
    \begin{lemma}\label{lemma:3}
        \textbf{\Dice model is closer to optimal weight than regression loss models.}
        Based on \cref{lemma:1} and assuming the object length $\length$ is a constant,  
        if $\normalSigTh$ is the solution of the equation $\normalVar\!=\!\frac{1}{\length^2}\normalErf\left(\frac{\length}{\sqrt{2}\normalSig}\right)$ and the noise deviation $\normalSig\!\ge\! \normalSigCr\!=\!\max \left(\normalSigTh, \frac{\sqrt{2}}{\length}\normalErfInv(\length^2)\right)$, then the converged weight $\layerWeightConvDice$ with the \dice loss $\lDice$ is better than the converged weight $\layerWeightConvReg$ with the $\lOne$ or $\lTwo$ loss, \thatIs
        \begin{align}
            \expect\left(\norm{\layerWeightConvDice-\layerWeightOptimal}_2\right) &\le \expect\left(\norm{\layerWeightConvReg-\layerWeightOptimal}_2\right).
        \end{align}
    \end{lemma}
        We refer to \cref{sec:seabird_supp_proof_lemma_3} for the proof.
        Beyond noise deviation threshold $\normalSigCr\!=\!\max \left(\normalSigTh, \frac{\sqrt{2}}{\length}\normalErfInv(\length^2)\right)$, the convergence gap between \dice and regression losses widens as the object size $\length$ increases. 
        \cref{fig:seabird_conv_analysis} depicts the superior convergence of \dice loss compared to regression losses under increasing noise deviation $\normalSig$ pictorially. 
        Taking the car category with $\length\!=\!4m$ and the trailer category with $\length\!=\!12m$ as examples, the noise threshold $\normalSigCr$, beyond which \dice loss exhibits better convergence, are $\normalSigCr\!=\!0.3m$ and $\normalSigCr\!=\!0.1m$ respectively.
        Combining these lemmas, we finally derive:\\
        \vspace{-0.6cm}
        \begin{theorem}\label{th:seabird_1}
            \textbf{\Dice model has better \apThreeD.}
            Assume the object length $\length$ is a constant and depth is the only source of error for detection.
            Based on \cref{lemma:1}, if $\normalSigTh$ is the solution of the equation $\normalVar\!=\!\frac{1}{\length^2}\normalErf\left(\frac{\length}{\sqrt{2}\normalSig}\right)$ and the noise deviation $\normalSig\!\ge\!\normalSigCr\!=\!\max \left(\normalSigTh, \frac{\sqrt{2}}{\length}\normalErfInv(\length^2)\right)$, then the Average Precision (\apThreeD) of the \dice model is better than \apThreeD from $\lOne$ or $\lTwo$ model.
        \end{theorem}
        We refer to \cref{sec:seabird_supp_proof_theorem_1} and \cref{tab:seabird_assumption_comp} for the proof and assumption comparisons respectively.

    \subsection{Discussions}

        \noIndentHeading{Comparing classification and regression losses.}
            We now explain how we compare classification (\dice) and regression losses.
            Our analysis assumes one-class classification in \bev segmentation with perfect predicted foreground scores $P(\varZ)=1$ (\cref{fig:seabird_problem_setup}\textcolor{link_color}{c}).
            Hence, \dice analysis focuses on object localization along the \bev ray  (\cref{fig:seabird_problem_setup}\textcolor{link_color}{b}) instead of classification probabilities thus allowing comparison of \dice and regression losses.
            \cref{lemma:1} links these losses by comparing the deviation of learned and optimal weights. 
    
        \noIndentHeading{Regression losses work better than \dice loss for regression tasks?}
            Our key message is NOT always! 
            We mathematically and empirically show that regression losses work better only when the \textbf{noise $\normalSig$ is less} in \cref{fig:seabird_conv_analysis}.

    \subsection{\seabird Pipeline}\label{sec:seabird_pipeline}
        
        \noIndentHeading{Architecture.} 
            Based on theoretical insights of \cref{th:seabird_1}, we propose \seabird, a novel pipeline, in \cref{fig:seabird_pipeline}. 
            To effectively involve the \dice loss which originally designed for segmentation task to assist \monoThreeD, \seabird treats \bev segmentation of foreground objects and \monoThreeD head sequentially. 
            Although \bev segmentation map provides depth information (hardest \cite{kumar2022deviant,ma2021delving} \monoThreeD parameter), it lacks elevation and height information for \monoThreeD task.
            To address this, \seabird concatenates \bev features with predicted \bev segmentation (\cref{fig:seabird_pipeline}), and feeds them into the detection head to predict \threeD boxes in a $7$-DoF representation: \bev \twoD position, elevation, \threeD dimension, and yaw. 
            Unlike most works \cite{zhang2022beverse, li2022bevformer} that treat segmentation and detection branches in parallel, the sequential design
            directly utilizes refined \bev localization information to enhance \monoThreeD. 
            Ablations in \cref{sec:seabird_ablation} validate this design choice.
            We defer the details of baselines to \cref{sec:seabird_experiments}.
            Notably, our foreground \bev segmentation supervision with \dice loss does not require dense \bev segmentation maps, as we efficiently prepare them from GT \threeD boxes.

        \noIndentHeading{Training Protocol.} 
            \seabird trains the \bev segmentation head first, employing the \dice loss between the predicted and the GT \bev semantic segmentation maps, which fully utilizes the \dice loss's noise-robustness and superior convergence in localizing large objects. 
            In the second stage, we jointly fine-tune the \bev segmentation head and the \monoThreeD head. 
            We validate the effectiveness of training protocol via the ablation in \cref{sec:seabird_ablation}.

\section{Experiments}\label{sec:seabird_experiments}

    \noIndentHeading{Datasets.}
        Our experiments utilize two datasets with large objects: \kittiThreeSixty \cite{liao2022kitti360} and \nuscenes \cite{caesar2020nuscenes} encompassing both single-camera and multi-camera configurations.
        We opt for \kittiThreeSixty instead of \kitti \cite{geiger2012we} for four reasons: 
        1) \kittiThreeSixty includes large objects, while \kitti does not; 
        2) \kittiThreeSixty exhibits a balanced distribution of large objects and cars; 
        3) an extended version, \kittiThreeSixtyPanoptic \cite{gosala2022bev}, includes \bev segmentation GT for ablation studies, while \kitti \threeD detection and the \semanticKITTI dataset \cite{behley2019semantickitti} do not overlap in sequences; 
        4) \kittiThreeSixty contains about $10\times$ more images than \kitti.
        We compare these datasets in \cref{tab:seabird_dataset_comparison} and show their skewness in \cref{fig:seabird_skew}.

        \begin{table}[!t]
            \caption[Datasets comparison.]
            {\textbf{Datasets comparison.} We use \kittiThreeSixty and \nuscenes datasets for our experiments. See \cref{fig:seabird_skew} for the skewness.}
            \label{tab:seabird_dataset_comparison}
            \centering
            \setlength\tabcolsep{0.1cm}
            \begin{tabular}{l m c c c c}
                & \kitti\!\cite{geiger2012we} & \waymo\!\cite{sun2020scalability} & \kittiThreeSixty\!\cite{liao2022kitti360} & \nuscenes\!\cite{caesar2020nuscenes}\\
                \myTopRule
                Large objects & \xmark & \xmark & \cmark & \cmark \\
                Balanced &	\xmark & \xmark & \cmark & \xmark\\
                \bev Seg. GT & \xmark & \cmark & \cmark & \cmark\\ 	
                \#images (k) & $4$ & $52$~\cite{kumar2022deviant} & $49$ & $168$ \\
            \end{tabular}
        \end{table}

    \noIndentHeading{Data Splits.}
        We use the following splits of the two datasets:
        \begin{itemize}
            \item \textit{\kittiThreeSixty Test split}: This benchmark \cite{liao2022kitti360} contains $300$ training and $42$ testing windows. 
            These windows contain $61{,}056$ training and $910$ testing images.

            \item \textit{\kittiThreeSixty \val split}: It partitions the official train into $239$ train and $61$ validation windows \cite{liao2022kitti360}. 
            This split contains $48{,}648$ training and $1{,}294$ validation images.

            \item \textit{\nuscenes Test split}: It has $34{,}149$ training and $6{,}006$ testing samples \cite{caesar2020nuscenes} from the six cameras. This split contains $204{,}894$ training and $36{,}036$ testing images.
            
            \item \textit{\nuscenes \val split}: It has $28{,}130$ training and $6{,}019$ validation samples \cite{caesar2020nuscenes} from the six cameras. This split contains $168{,}780$ training and $36{,}114$ validation images.
        \end{itemize}

    \noIndentHeading{Evaluation Metrics.}
        We use the following metrics:
        \begin{itemize}
            \item \textit{Detection}: \kittiThreeSixty uses the mean \apThreeDFifty percentage across categories to benchmark models \cite{liao2022kitti360}.
                \nuscenes \cite{caesar2020nuscenes} uses the \nuscenes Detection Score (\NDS) as the metric. \NDS is the weighted average of mean \ap (\MAP) and five TP metrics.
                We also report \MAP over large categories (truck, bus, trailers and construction vehicles), cars, and small categories (pedestrians, motorcyle, bicycle, cone and barrier) as \MAPLarge, \MAPCar and \MAPSmall respectively.
            \item \textit{Semantic Segmentation}: We report mean \iou over foreground and all categories at $200\!\times\!200$ resolution \cite{saha2022translating,zhang2022beverse}. 
        \end{itemize}

    \noIndentHeading{\kittiThreeSixty Baselines and \seabird Implementation.}
        Our evaluation on the \kittiThreeSixty focuses on the detectors taking single-camera image as input. 
        We evaluate \seabird pipelines against six \sota frontal detectors: \groomedNMS \cite{kumar2021groomed}, \monodle\cite{ma2021delving}, \gupNet \cite{lu2021geometry}, \deviant \cite{kumar2022deviant}, \cubeRCNN \cite{brazil2023omni3d} and \monodetr \cite{zhang2023monodetr}.
        The choice of these models encompasses anchor \cite{kumar2021groomed, brazil2023omni3d} and anchor-free methods \cite{ma2021delving,kumar2022deviant},
        CNN~\cite{ma2021delving,lu2021geometry}, group CNN~\cite{kumar2022deviant} and transformer-based \cite{zhang2023monodetr} architectures.
        Further, \monodle normalizes loss with GT box dimensions.
        
        Due to \seabird's \bev-based approach, we do not integrate it with these frontal view detectors. 
        Instead, we extend two \sota image-to-\bev segmentation methods, \imageToMapsLong (\imageToMaps) \cite{saha2022translating} and \panopticBEVLong (\panopticBEV) \cite{gosala2022bev} with \seabird. 
        Since both \bev segmentors already include their own implementations of the image encoder, the image-to-\bev transform, and the segmentation head, implementing the \seabird pipeline  only involves adding a detection head, which we chose to be \orBoxNet \cite{yi2021oriented}.
        \seabird extensions employ \dice loss for \bev segmentation, $\smoothLOne$ losses \cite{girshick2015fast} in the \bev space to supervise the \bev \twoD position, elevation, and \threeD dimension, and cross entropy loss to supervise orientation. 

    \noIndentHeading{\nuscenes Baselines and \seabird Implementation.}
        We integrate \seabird into two prototypical \bev-based detectors, \beVerse \cite{zhang2022beverse} and \hop \cite{zong2023hop} to prove the effectiveness of \seabird. 
        Our choice of these models encompasses both transformer and convolutional backbones, multi-head and single-head architectures, shorter and longer frame history, and non-query and query-based detectors. 
        This comprehensively allows us to assess \seabird's impact on large object detection.
        \beVerse employs a multi-head architecture with a transformer backbone and shorter frame history.
        \hop is single-head query-based \sota model utilizing \bevDetFourD \cite{huang2022bevdet4d} with CNN backbone, and longer frame history. 
    
        \beVerse \cite{zhang2022beverse} includes its own implementation of detection head and \bev segmentation head in parallel. 
        We reorganize the two heads to follow our sequential design and adhere to our training protocol for network training. 
        Since \hop \cite{zong2023hop} lacks a \bev segmentation head, we incorporate the one from \beVerse into this \hop extension with \seabird.

    \subsection{\kittiThreeSixty \monoThreeD}\label{sec:seabird_detection_results_kitti_360_val}

        \begin{table}[!t]
            \caption
            [\kittiThreeSixty Test detection results.]
            {\textbf{\kittiThreeSixty Test detection results.} 
            \seabird pipelines outperform all monocular baselines, and also outperform old \lidar baselines.
            Click for the \href{https://www.cvlibs.net/datasets/kitti-360/leaderboard_scene_understanding.php?task=box3d}{\kittiThreeSixty leaderboard} as well as our 
            \href{https://www.cvlibs.net/datasets/kitti-360/eval_bbox_detect_detail.php?benchmark=bbox3d&result=2c29dba83ec92b4efa4b9bf67d9dcae2bef57828}{\panopticBEVWithMethod}
            and 
            \href{https://www.cvlibs.net/datasets/kitti-360/eval_bbox_detect_detail.php?benchmark=bbox3d&result=7f8612f009cc35fbebe749a345b5e49158f1efa0}{\imageToMapsWithMethod} entries.
            [Key: \firstKey{Best}, \secondKey{Second Best}, L= \lidar, C= Camera, \retrained= Retrained].
            }
            \label{tab:seabird_det_results_kitti_360_test}
            \centering
            \scalebox{\scaleFraction}{
            \setlength\tabcolsep{0.04cm}
            \begin{tabular}{cc m l | c m c | c}
                \multicolumn{2}{cm}{Modality} & \multirow{2}{*}{Method} & \multirow{2}{*}{Venue} & \apThreeDFifty (\uparrowRHDSmall) & \apThreeDTwentyFive (\uparrowRHDSmall)\\ 
                L & C & & & \MAP~\bracketPercentage & \MAP~\bracketPercentage\\
                \myTopRule
                \checkmark & & \voteNet\cite{qi2019deep}~~~~~& ICCV19 & $3.40$ & $30.61$\\
                \checkmark & & \lidarBoxNet\cite{qi2019deep} & ICCV19 & $4.08$ & $23.59$\\
                \hline
                & \checkmark &GrooMeD~\retrained\cite{kumar2021groomed} & CVPR21 & $0.17$ & $16.12$\\
                & \checkmark &\monodle\retrained\cite{ma2021delving} & CVPR21 & $0.85$ & $28.99$\\
                & \checkmark &\gupNet\retrained\cite{lu2021geometry} & ICCV21 & $0.87$ & $27.25$\\
                & \checkmark &\deviant\retrained\cite{kumar2022deviant} & ECCV22 & $0.88$ & $26.96$\\
                & \checkmark &\cubeRCNN\retrained \cite{brazil2023omni3d} & CVPR23 & $0.80$ & $15.57$\\
                & \checkmark &\monodetr\retrained\cite{zhang2023monodetr} & ICCV23 & $0.79$ & $27.13$\\
                \rowcolor{my_gray}& \checkmark & \textbf{\imageToMapsWithMethod} & CVPR24 & \second{3.14} & \second{35.04}\\
                \rowcolor{my_gray}& \checkmark & \textbf{\panopticBEVWithMethod} & CVPR24 & \first{4.64} & \first{37.12} \\ 
            \end{tabular}
            }
        \end{table}

        \noIndentHeading{\kittiThreeSixty Test.}\label{sec:seabird_detection_results_kitti_360_test}
            \cref{tab:seabird_det_results_kitti_360_test} presents \kittiThreeSixty leaderboard results, demonstrating the superior performance of both 
            \seabird pipelines compared to all monocular baselines across all metrics. 
            Moreover, \panopticBEVWithMethod also outperforms both legacy \lidar baselines on all metrics, while \imageToMapsWithMethod surpasses them on the \apThreeDTwentyFive metric.
            
        \noIndentHeading{\kittiThreeSixty \val.}
            \cref{tab:seabird_det_seg_results_kitti_360_val} presents the results on \kittiThreeSixty \val split, reporting the \textbf{median} model over three different seeds with the model being the final checkpoint as \cite{kumar2022deviant}. 
            \seabird pipelines outperform all monocular baselines on all but one metric, similar to \cref{tab:seabird_det_results_kitti_360_test} results. 
            Due to the \dice loss in \seabird, the biggest improvement shows up on larger objects.
            \cref{tab:seabird_det_seg_results_kitti_360_val} also includes the upper-bound oracle, where we train the \orBoxNet with the GT \bev segmentation maps.

        \begin{figure}[!t]
            \centering
            \begin{subfigure}{.485\linewidth}
                \centering
                \includegraphics[width=\linewidth]{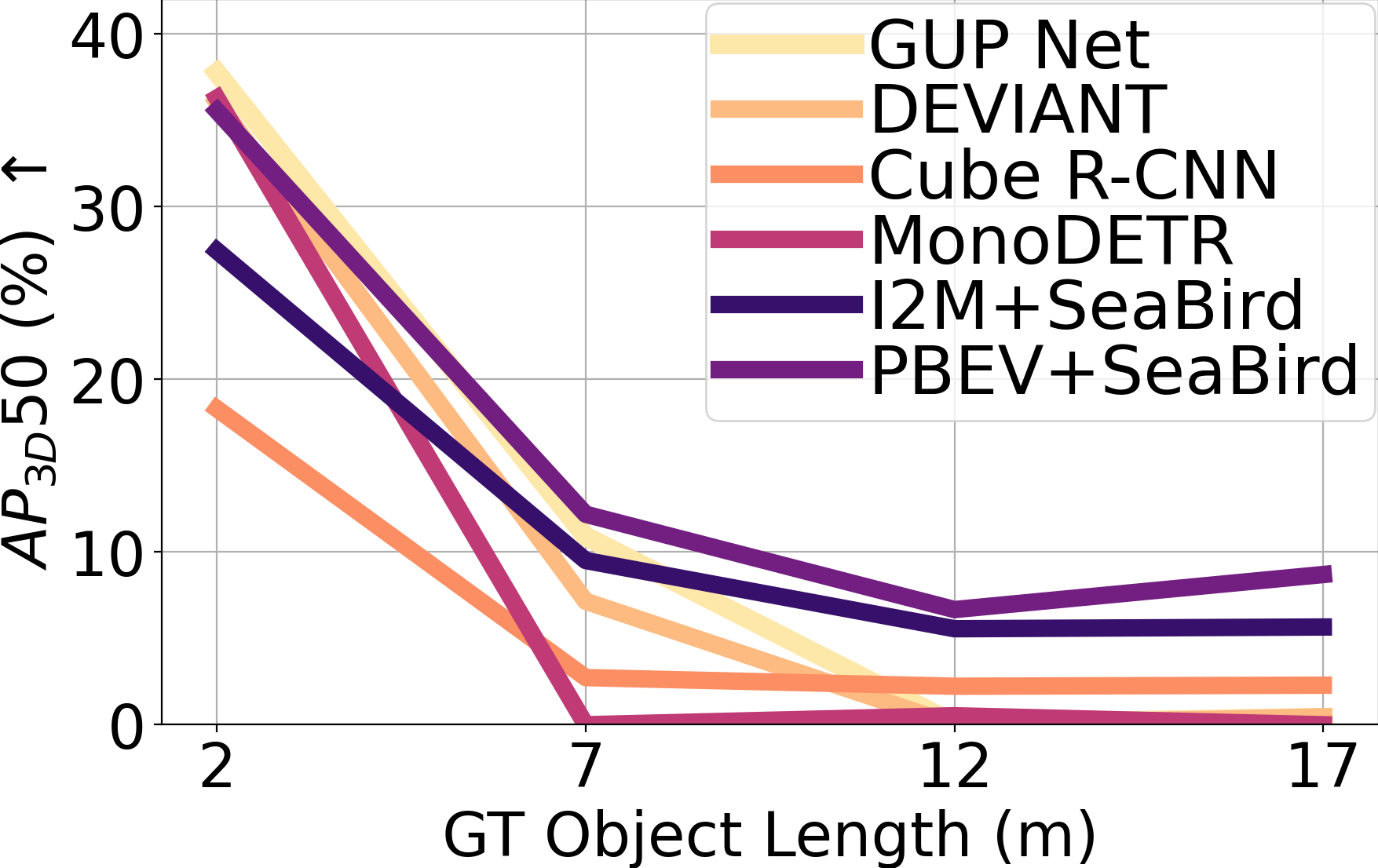}
                \caption{\apThreeDFifty comparison.}
            \end{subfigure}%
            \hfill
            \begin{subfigure}{.485\linewidth}
                \centering
                \includegraphics[width=\linewidth]{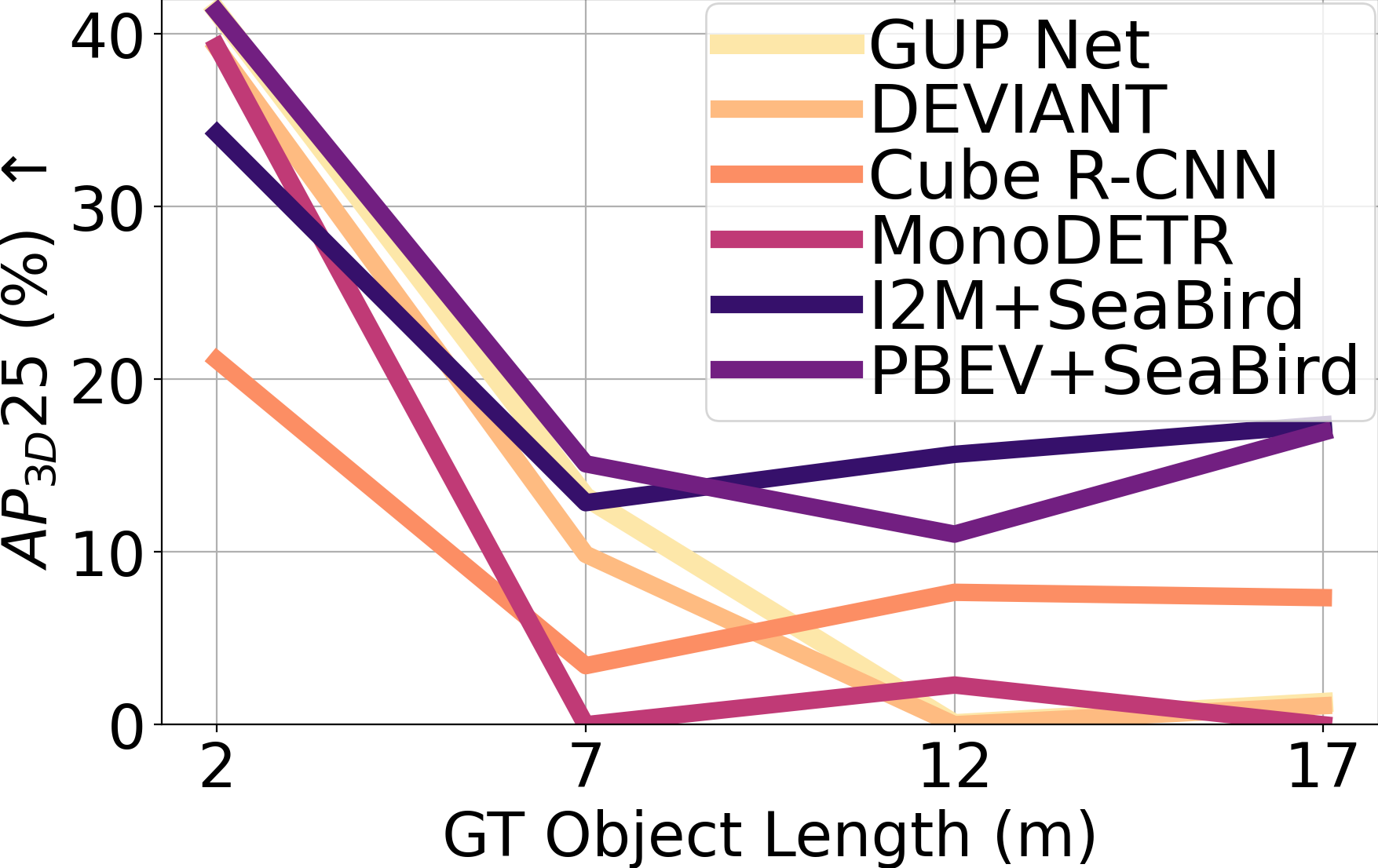}
                \caption{\apThreeDTwentyFive comparison.}
            \end{subfigure}
            \caption[Lengthwise \ap Analysis of four \sota detectors and two \seabird pipelines on \kittiThreeSixty \val split]
            {\textbf{Lengthwise \ap Analysis} of four \sota detectors and two \seabird pipelines on \kittiThreeSixty \val split. 
            \seabird pipelines outperform all baselines on large objects with over $10$m in length.}
            \label{fig:seabird_lengthwise_analysis}
        \end{figure}
        \begin{table*}[!t]
            \caption[\kittiThreeSixty \val detection and segmentation results.]
            {\textbf{\kittiThreeSixty \val detection and segmentation results.}
            \seabird pipelines outperform all frontal monocular baselines, particularly for large objects.
            \Dice loss in \seabird also improves the \bev only (w/o \dice) version of \seabird pipelines.
            \imageToMaps and \panopticBEV are \bev segmentors. 
            So, we do not report their \monoThreeD performance.
            [Key: \firstKey{Best}, \secondKey{Second Best}, \retrained= Retrained]
            }
            \label{tab:seabird_det_seg_results_kitti_360_val}
            \centering
            \scalebox{0.7}{
            \setlength\tabcolsep{0.15cm}
            \begin{tabular}{c m l | c c m bcc | bcc m ccc}
                \addlinespace[0.01cm]
                \multirow{2}{*}{View} & \multirow{2}{*}{Method} & \bev Seg & \multirow{2}{*}{Venue} &\multicolumn{3}{c|}{\apThreeDFifty~\bracketPercentage (\uparrowRHDSmall)} & \multicolumn{3}{cm}{\apThreeDTwentyFive~\bracketPercentage (\uparrowRHDSmall)} & \multicolumn{3}{c}{\bev Seg \iou~\bracketPercentage (\uparrowRHDSmall)}\\ 
                && Loss & & \MAPLarge & \MAPCar & \MAP & \MAPLarge & \MAPCar & \MAP & Large & Car & \meanFor \\
                \myTopRule
                \multirow{6}{*}{Frontal} & \groomedNMS\retrained\cite{kumar2021groomed} & \multirow{6}{*}{\mathDash} & CVPR21 & $0.00$	& $33.04$ & $16.52$ & $0.00$ & $38.21$ & $19.11$ & \mathDash	& \mathDash	& \mathDash\\
                & \monodle\retrained\cite{ma2021delving} &  & CVPR21 & $0.94$ & $44.81$ & $22.88$ & $4.64$ & $50.52$ & $27.58$ & \mathDash	& \mathDash	& \mathDash\\
                & \gupNet\retrained \cite{lu2021geometry} & & ICCV21 &
                ${0.54}$ & \second{45.11} & ${22.83}$ & $0.98$ & ${50.52}$ & ${25.75}$ & \mathDash	& \mathDash	& \mathDash	\\
                & \deviant\retrained \cite{kumar2022deviant} & & ECCV22 &
                $0.53$	& $44.25$ & $22.39$ & ${1.01}$ & $48.57$ & $24.79$ & \mathDash	& \mathDash	& \mathDash\\
                & \cubeRCNN\retrained \cite{brazil2023omni3d} & & CVPR23 & $0.75$ & $22.52$ & $11.63$ & $5.55$ & $27.12$ & $16.34$ & \mathDash	& \mathDash	& \mathDash\\
                & \monodetr\retrained\cite{zhang2023monodetr} & & ICCV23 & $0.81$ & $43.24$ & $22.02$ & $4.50$ & $48.69$ & $26.60$ & \mathDash	& \mathDash	& \mathDash\\
                \hline
                \multirow{7}{*}{\bev} & \imageToMaps\retrained \cite{saha2022translating} & \Dice & ICRA22 & \mathDash & \mathDash & \mathDash & \mathDash & \mathDash & \mathDash & $20.46$ & $38.04$	& $29.25$ \\
                & \imageToMapsWithMethod & \xmark & CVPR24 & $4.86$ & $45.09$ & $24.98$ & $26.33$ & $52.31$ & $39.32$ & $0.00$ & $7.07$ & $3.54$\\
                & \cellcolor{my_gray}\textbf{\imageToMapsWithMethod} & \cellcolor{my_gray}\Dice & \cellcolor{my_gray}CVPR24 &
                \cellcolor{my_gray}\second{8.71}	& \cellcolor{my_gray}$43.19$ & \cellcolor{my_gray}$25.95$	& \cellcolor{my_gray}\second{35.76}	& \cellcolor{my_gray}$52.22$ & \cellcolor{my_gray}\second{43.99} & \cellcolor{my_gray}${23.23}$ & \cellcolor{my_gray}${39.61}$	& \cellcolor{my_gray}${31.42}$	\\
                \hhline{|~|------------|}
                & \panopticBEV\retrained \cite{gosala2022bev} & CE & RAL22 & \mathDash & \mathDash & \mathDash & \mathDash & \mathDash & \mathDash & \second{23.83} & \first{48.54}	& \first{36.18} \\
                & \panopticBEVWithMethod & \xmark & CVPR24 & $7.64$ & \first{45.37} & \second{26.51} & $29.72$ & \first{53.86} & $41.79$ & $2.07$ & $1.47$	& $1.57$\\ 
                & \cellcolor{my_gray}\textbf{\panopticBEVWithMethod} & \cellcolor{my_gray}Dice & \cellcolor{my_gray}CVPR24 &
                \cellcolor{my_gray}\first{13.22}	& \cellcolor{my_gray}$42.46$ & \cellcolor{my_gray}\first{27.84}	& \cellcolor{my_gray}\first{37.15}	& \cellcolor{my_gray}\second{52.53}   & \cellcolor{my_gray}\first{44.84} & \cellcolor{my_gray}\first{24.30}	& \cellcolor{my_gray}\second{48.04}	& \cellcolor{my_gray}\second{36.17}	\\
                \hhline{|~|------------|}
                & Oracle (GT \bev) & & \mathDash &
                $26.77$ & $51.79$ & $39.28$ & $49.74$ & $56.62$ & $53.18$ & $100.00$ & $100.00$	& $100.00$	\\
            \end{tabular}
            }
        \end{table*}

        \noIndentHeading{Lengthwise \ap Analysis.}
            \cref{th:seabird_1} states that training a model with \dice loss should lead to lower errors and, consequently, a better detector for large objects.
            To validate this claim, we analyze the detection performance with \apThreeDFifty and \apThreeDTwentyFive metrics against the object's lengths.
            For this analysis, we divide objects into four bins based on their GT object length (max of sizes): $[0, 5), [5,10), [10, 15), [15+ m$.
            \cref{fig:seabird_lengthwise_analysis} shows that \seabird pipelines excel for large objects, where the baselines' performance drops significantly.

        \noIndentHeading{\bev Semantic Segmentation.} \label{sec:seabird_segmentation_results_kitti_360_val}
            \cref{tab:seabird_det_seg_results_kitti_360_val} also presents the \bev semantic segmentation results on the \kittiThreeSixty \val split.
            \seabird pipelines outperforms the baseline \imageToMaps \cite{saha2022translating}, and achieve similar performance to \panopticBEV \cite{gosala2022bev} in \bev segmentation. 
            We retrain all \bev segmentation models only on foreground detection categories for a fair comparison.

        \begin{table*}[!t]
            \caption[Ablation studies on \kittiThreeSixty \val]
            {\textbf{Ablation studies} on \kittiThreeSixty \val.
            [Key: \bestKey{Best}, \secondKey{Second Best}]
            }
            \label{tab:seabird_ablation}
            \centering
            \scalebox{0.77}{
            \setlength\tabcolsep{0.15cm}
            \begin{tabular}{l | l m bcc | bcc m cccc}
                \addlinespace[0.01cm]
                \multirow{2}{*}{Changed} & \multirow{2}{*}{From $\rightarrowRHD$ To} & \multicolumn{3}{c|}{\apThreeDFifty~\bracketPercentage (\uparrowRHDSmall)} & \multicolumn{3}{cm}{\apThreeDTwentyFive~\bracketPercentage (\uparrowRHDSmall)} & \multicolumn{4}{c}{\bev Seg \iou~\bracketPercentage (\uparrowRHDSmall)}\\ 
                && \MAPLarge & \MAPCar & \MAP & \MAPLarge & \MAPCar & \MAP & Large & Car & \meanFor & \meanEleven\\
                \myTopRule
                \multirow{4}{*}{Segmentation Loss} & \Dice$\rightarrowRHD$No Loss &$4.86$ & \best{45.09} & $24.98$ & $26.33$ & $52.31$ & $39.32$ & $0.00$ & $7.07$ & $3.54$ & \mathDash \\
                & \Dice$\rightarrowRHD$Smooth~$\lOne$& $7.63$ & $36.69$ & $22.16$ & $31.01$ & $47.51$ & $39.26$ & $17.16	$ & $34.67$ & $25.92$ & \mathDash\\
                & \Dice$\rightarrowRHD$MSE & $7.04$ & $35.59$ & $21.32$ & $30.90$ & $44.71$ & $37.81$ & $17.46$ & $34.85$ & $26.16$ & \mathDash\\
                & \Dice$\rightarrowRHD$CE & $7.06$ & $35.60$ & $21.33$ & \second{33.22} & $47.60$ & $40.41$ & $21.83$ & $38.11$ & $29.97$ & \mathDash\\
                \hline
                Segmentation Head & Yes$\rightarrowRHD$No & $7.52$ & $39.24$ & $23.38$ & $31.83$ & $47.88$ & $39.86$ & \mathDash & \mathDash & \mathDash & \mathDash \\
                Detection Head & Yes$\rightarrowRHD$No & \mathDash & \mathDash & \mathDash & \mathDash & \mathDash & \mathDash & $20.46$ & $38.04$	& $29.25$ & \mathDash \\
                \hline
                \multirow{2}{*}{Semantic Category} & For.$\rightarrowRHD$All & $1.61$ & \second{44.12} & $22.87$ & $15.36$ & $51.76$ & $33.56$ & $19.26$ & $34.46$ & $26.86$ & \first{24.34}\\
                & For.$\rightarrowRHD$Car & $4.17$ & $43.01$ & $23.59$ & $22.68$ & $51.58$ & $37.13$ & \mathDash & $40.28$ & $20.14$ & \mathDash	\\
                \hline
                Multi-head Arch.  & Sequential$\rightarrowRHD$Parallel & \best{9.12} & $40.27$ & $24.69$ & $32.45$ & $51.55$ & $42.00$ & $22.19$ & \second{40.37} & $31.28$ & \mathDash\\
                \hline 
                \bev Shortcut & Yes$\rightarrowRHD$No & $6.53$ & $38.12$ & $22.33$ & $32.05$ & \second{52.62} & \second{42.34} & \second{23.00} & \best{40.39} & \best{31.70}		& \mathDash\\
                \hline
                \multirow{2}{*}{Training Protocol} & S+J$\rightarrowRHD$J \cite{zhang2022beverse}& $7.42$ & $	42.73$ & \second{25.08} & $31.94$ & $49.88$ & $40.91$ & $22.91$ & $39.66$ & $31.29$	& \mathDash\\
                & S+J$\rightarrowRHD$D+J \cite{yang2023lidar}& $6.07$ & $43.43$ & $24.75$ & $29.24$ & \best{52.96} & $41.10$ & $20.71$ & $35.68$ & $28.20$ & \mathDash\\
                \hline
                \textbf{\imageToMapsWithMethod} & \mathDash & 
                \second{8.71}	& $43.19$ & \best{25.95}	& \best{35.76}	& $52.22$ & \best{43.99} & \best{23.23}	& $39.61$ & \second{31.42}	& \mathDash	\\
            \end{tabular}
            }
        \end{table*}

    \subsection{Ablation Studies on \kittiThreeSixty \val}\label{sec:seabird_ablation}

        \cref{tab:seabird_ablation} ablates \imageToMaps \cite{saha2022translating} +\seabird on the \kittiThreeSixty \val split, following the experimental settings of \cref{sec:seabird_detection_results_kitti_360_val}.

        \noIndentHeading{\Dice Loss.}
            \cref{tab:seabird_ablation} shows that both \dice loss and \bev representation are crucial to \monoThreeD of large objects.
            Replacing \dice loss with MSE or $\smoothLOne$ loss, or only \bev representation (w/o \dice) reduces \monoThreeD performance.

        \noIndentHeading{\monoThreeD and \bev Segmentation.}
            \cref{tab:seabird_ablation} shows that removing the segmentation head hinders \monoThreeD performance. 
            Conversely, removing detection head also diminishes the \bev segmentation performance for the segmentation model. 
            This confirms the mututal benefit of sequential \bev segmentation on foreground objects and \monoThreeD.

        \noIndentHeading{Semantic Category in \bev Segmentation.}
            We next analyze whether background categories play any role in \monoThreeD. 
            \cref{tab:seabird_ablation} shows that changing the foreground (For.) categories to foreground + background (All) does not help \monoThreeD. 
            This aligns with the observations of \cite{zhang2022beverse,xie2022m2bev,ma2022vision} that report lower performance on joint \monoThreeD and \bev segmentation with all categories.
            We believe this decrease happens because the network gets distracted while getting the background right.
            We also predict one foreground category (Car) instead of all in \bev segmentation.
            \cref{tab:seabird_ablation} shows that predicting all foreground categories in \bev segmentation is crucial for overall good \monoThreeD.

        \noIndentHeading{Multi-head Architecture.}
            \seabird employs a sequential architecture (Arch.) of segmentation and detection heads instead of parallel architecture.
            \cref{tab:seabird_ablation} shows that the sequential architecture outperforms the parallel one. 
            We attribute this \monoThreeD boost to the explicit object localization provided by segmentation in the \bev plane.

        \noIndentHeading{\bev Shortcut.}
            \cref{sec:seabird_pipeline} mentions that \seabird's \monoThreeD head utilizes both the \bev segmentation map and \bev features.
            \cref{tab:seabird_ablation} demonstrates that providing \bev features to the detection head is crucial for good \monoThreeD.
            This is because the \bev map lacks elevation information, and incorporating \bev features helps estimate elevation.

        \noIndentHeading{Training Protocol.}
            \seabird trains segmentor first and then jointly trains detector and segmentor (S+J).
            We compare with direct joint training (J) of \cite{zhang2022beverse} and training detection followed by joint training (D+J) of \cite{yang2023lidar}.
            \cref{tab:seabird_ablation} shows that \seabird training protocol works best.

        \begin{table*}[!t]
            \caption[\nuscenes Test detection results.]
            {\textbf{\nuscenes Test detection results.}
            \seabird pipelines achieve the best \MAPLarge among methods without Class Balanced Guided Sampling (CBGS) \cite{zhu2019class} and future frames.
            Results are from the nuScenes leaderboard or corresponding chapters on \vovNet or R101 backbones.
            [Key: 
            \firstKey{Best}, \secondKey{Second Best}, 
            S= Small, \reimplemented= Reimplementation, \cbgs= CBGS, \future= Future Frames.]
            }
            \label{tab:seabird_nuscenes_test}
            \centering
            \scalebox{0.8}{
            \setlength\tabcolsep{0.15cm}
            \begin{tabular}{l l | c c m b c c c c}
                Resolution & Method & BBone &  Venue & \MAPLarge\!(\uparrowRHDSmall) & \MAPCar\!(\uparrowRHDSmall) & \MAPSmall\!(\uparrowRHDSmall) & \MAP\!(\uparrowRHDSmall)  & \NDS\!(\uparrowRHDSmall) \\
                \myTopRule
                \multirow{6}{*}{$512\!\times\!1408$}
                & \bevDepth\cite{li2023bevdepth} in \cite{kim2023predict} & R101 & AAAI23 & \mathDash 
                & \mathDash & \mathDash & $39.6$ & $48.3$ \\
                & \bevStereo\cite{li2023bevstereo} in \cite{kim2023predict} & R101 & AAAI23 & \mathDash & \mathDash & \mathDash & $40.4$ & $50.2$ \\
                & P2D \cite{kim2023predict} & R101 & ICCV23 & \mathDash & \mathDash & \mathDash & $43.6$ & $53.0$ \\
                & \beVerseSmall\cite{zhang2022beverse}& Swin-S  & \arxiv & $24.4$ & $60.4$ & $47.0$ & $39.3$ & $53.1$ \\
                & \hop\reimplemented\cite{zong2023hop}  & R101 & ICCV23 & \second{36.0}    & \second{65.0}    & \second{53.9} & \second{47.9} & \first{57.5} \\
                & \cellcolor{my_gray}\textbf{\hopWithMethod} & \cellcolor{my_gray}R101 & \cellcolor{my_gray}CVPR24 & \cellcolor{my_gray}\first{36.6}    & \cellcolor{my_gray}\first{65.8}    & \cellcolor{my_gray}\first{54.7} & \cellcolor{my_gray}\first{48.6} & \cellcolor{my_gray}\second{57.0}\\
                \myTopRule
                \multirow{13}{*}{$640\!\times\!1600$}
                & SpatialDETR \cite{doll2022spatialdetr} & \vovNet & ECCV22 & $30.2$ & $61.0$ & $48.5$ & $42.5$ & $48.7$ \\
                & \threeDPPE\cite{shu2023dppe} & \vovNet & ICCV23 & \mathDash & \mathDash & \mathDash & $46.0$ & $51.4$ \\
                & \xThreeKDAll \cite{klingner2023x3kd} & R101 & CVPR23 & \mathDash & \mathDash & \mathDash & $45.6$    & $56.1$\\
                & \petrVTwo\cite{liu2023petrv2} & \vovNet & ICCV23 & $36.4$    & $66.7$    & $55.6$    & $49.0$    & $58.2$    \\
                & \veDet\cite{chen2023viewpoint} & \vovNet & CVPR23 & \second{37.1}    & $68.5$    & \first{57.7}    & $50.5$    & $58.5$\\ 
                & \frustumFormer\cite{wang2023frustumformer} & \vovNet & CVPR23 & \mathDash & \mathDash & \mathDash & \first{51.6}    & $58.9$\\    
                & MV2D \cite{wang2023object}  & \vovNet & ICCV23 & \mathDash & \mathDash & \mathDash & \second{51.1}    & \second{59.6}    \\
                & \hop\reimplemented\cite{zong2023hop} & \vovNet & ICCV23 & \second{37.1} & \second{68.7} & $55.6$ & $49.4$ & $58.9$ \\
                & \cellcolor{my_gray}\textbf{\hopWithMethod} & \cellcolor{my_gray}\vovNet & \cellcolor{my_gray}CVPR24 & \cellcolor{my_gray}\first{38.4}    & \cellcolor{my_gray}\first{70.2} & \cellcolor{my_gray}\second{57.4}    & \cellcolor{my_gray}\second{51.1}    & \cellcolor{my_gray}\first{59.7}    \\
                \hhline{|~|--------|}
                & \saBEV\cbgs\cite{zhang2023sabev} & \vovNet & ICCV23 & $40.5$ & $68.9$ & $60.5$ & $53.3$ & $62.4$ \\
                & \fbBEV\cbgs\cite{li2023fbbev} & \vovNet & ICCV23 & $39.3$    & $71.7$    & $61.6$   & $53.7$ & $62.4$\\
                & \cape\cbgs\cite{xiong2023cape} & \vovNet & CVPR23 & $41.3$    & $71.4$    & $63.3$    & $55.3$    & $62.8$\\
                & \sparseBEV\future\cite{liu2023sparsebev} & \vovNet & ICCV23 & $45.6$    & $76.3$    & $68.8$    & $60.3$    & $67.5$    \\
                \myTopRule
                \multirow{5}{*}{$900\!\times\!1600$} 
                & \parametricBEV\!\cite{yang2023parametric} & R101 & ICCV23 & \mathDash & \mathDash & \mathDash & $46.8$ & $49.5$ \\
                & \uvtr \cite{li2022unifying} & R101 & NeurIPS22 & $35.1$ & $67.3$ & $52.9$ & $47.2$ & $55.1$ \\
                & \bevFormer\cite{li2022bevformer} & \vovNet & ECCV22 & $34.4$ & $67.7$ & $55.2$ &
                $48.9$ & $56.9$ \\
                & \polarFormer\cite{jiang2023polarformer} & \vovNet & AAAI23 & $36.8$ & $68.4$ & $55.5$ & $49.3$ & $57.2$ \\
                & \stxd\cite{jang2023stxd} & \vovNet & NeurIPS23 & \mathDash & \mathDash & \mathDash & $49.7$ & $58.3$\\ 
                
            \end{tabular}
            }
        \end{table*}

        \begin{table*}[!t]
            \caption[\nuscenes \val detection results.]
            {\textbf{\nuscenes \val detection results.}
            \seabird pipelines outperform the two baselines \beVerse and \hop, particularly for large objects.
            We train all models without CBGS. 
            See \cref{tab:seabird_nuscenes_val_more} for a detailed comparison.
            [Key: 
            S= Small, T= Tiny, \released= Released, \reimplemented= Reimplementation]
            }
            \label{tab:seabird_nuscenes_val}
            \centering
            \scalebox{\scaleFractionSeaBird}{
            \setlength\tabcolsep{0.15cm}
            \begin{tabular}{l  l | c c m a l l l l}
                Resolution & Method & BBone & Venue & \multicolumn{1}{c}{\CYMyFix\MAPLarge (\uparrowRHDSmall)} & \multicolumn{1}{c}{\MAPCar (\uparrowRHDSmall)} & \multicolumn{1}{c}{\MAPSmall (\uparrowRHDSmall)} & \multicolumn{1}{c}{\MAP (\uparrowRHDSmall)}  & \multicolumn{1}{c}{\NDS (\uparrowRHDSmall)} \\
                \myTopRule
                \multirow{4}{*}{$256\!\times\!704$}  
                & \beVerseTiny\released\cite{zhang2022beverse}& \multirow{2}{*}{Swin-T} & \arxiv & $18.5$        & $53.4$ & $38.8$           & $32.1$        & $46.6$           \\
                & \textbf{+\seabird}         & & CVPR24 & \good{19.5}{1.0} & \good{54.2}{0.8} & \good{41.1}{2.3}    & \good{33.8}{1.5} & \good{48.1}{1.7} \\
                \hhline{|~|--------|}
                & \hop\released\cite{zong2023hop}  & \multirow{2}{*}{R50} & ICCV23 & $27.4$    & $57.2$    & $46.4$ & $39.9$ & $50.9$ \\
                & \textbf{+\seabird}      & & CVPR24 & \good{28.2}{0.8}    & \good{58.6}{1.4}    & \good{47.8}{1.4} & \good{41.1}{1.2} & \good{51.5}{0.6} \\
                \myTopRule
                \multirow{4}{*}{$512\!\times\!1408$} 
                &\beVerseSmall\released\cite{zhang2022beverse}& \multirow{2}{*}{Swin-S} &  \arxiv & $20.9$           & $56.2$           & $42.2$        & $35.2$        & $49.5$ \\
                & \textbf{+\seabird}                          & & CVPR24 & \good{24.6}{3.7}    & \good{58.7}{2.5}    & \good{45.0}{2.8}  & \good{38.2}{3.0} & \good{51.3}{1.8} \\
                \hhline{|~|--------|}
                & \hop\reimplemented\cite{zong2023hop}             & \multirow{2}{*}{R101} & ICCV23 & 
                $31.4$    & $63.7$    & $52.5$    & $45.2$    & $55.0$ \\ 
                & \textbf{+\seabird}                          & & CVPR24 & \good{32.9}{1.5}    & \good{65.0}{1.3}    & \good{53.1}{0.6} & \good{46.2}{1.0} & \bad{54.7}{0.3} \\
                \myTopRule
                \multirow{2}{*}{$640\!\times\!1600$}
                & \hop\reimplemented\cite{zong2023hop}             & \multirow{2}{*}{\vovNet} 
                & ICCV23 & $36.5$    & $69.1$    & $56.1$    & $49.6$    & $58.3$ \\
                & \textbf{+\seabird}                          & & CVPR24 & \good{40.3}{3.8}    & \good{71.7}{2.6}    & \good{58.8}{2.7} & \good{52.7}{3.1} & \good{60.2}{1.9} \\
            \end{tabular}
            }
        \end{table*}

    \subsection{\nuscenes \monoThreeD}
    
        We next benchmark \seabird on \nuscenes \cite{caesar2020nuscenes}, which encompasses more diverse object categories such as trailers, buses, cars and traffic cones, compared to \kittiThreeSixty \cite{liao2022kitti360}.

    \noIndentHeading{\nuscenes Test.}\label{sec:seabird_detection_nuscenes_test}
        \cref{tab:seabird_nuscenes_test} presents the results of incorportaing \seabird to the \hop models with the \vovNet and R101 backbones.
        \seabird with both \vovNet and R101 backbones outperform several \sota methods on the \nuscenes leaderboard, as well as the baseline \hop, on nearly every metric.
        Interestingly, \seabird pipelines also outperform several baselines which use higher resolution $(900\!\times\!1600)$ inputs. 
        Most importantly, \seabird pipelines achieve the highest \MAPLarge performance, providing empirical support for the claims of \cref{th:seabird_1}.

    \noIndentHeading{\nuscenes \val.}\label{sec:seabird_detection_nuscenes_val}
        \cref{tab:seabird_nuscenes_val} showcases the results of integrating \seabird with \beVerse \cite{zhang2022beverse} and \hop \cite{zong2023hop} at multiple resolutions, as described in \cite{zhang2022beverse,zong2023hop}.
        \cref{tab:seabird_nuscenes_val} demonstrates that integrating \seabird consistently improves these detectors on almost every metric at multiple resolutions.
        The improvements on \MAPLarge empirically support the claims of \cref{th:seabird_1} and validate the effectiveness of \dice loss and \bev segmentation in localizing large objects.

\section{Conclusions}\label{sec:seabird_conclusions}
    This chapter highlights the understudied problem of \monoThreeD generalization to large objects.
    Our findings reveal that modern frontal detectors struggle to generalize to large objects even when trained on balanced datasets.
    To bridge this gap, we investigate the regression and dice losses, 
    examining their robustness under varying error levels and object sizes.
    We mathematically prove that the dice loss 
    outperforms regression losses in noise-robustness and model convergence for large objects for a simplified case.
    Leveraging our theoretical insights, we propose \seabird (\seabirdFull) as the first step towards generalizing to large objects.
    \seabird effectively integrates \bev segmentation with the dice loss for \monoThreeD.
    \seabird achieves \sota results on the \kittiThreeSixty leaderboard and consistently improves existing detectors on the \nuscenes leaderboard, particularly for large objects.
    We hope that this initial step towards generalization will contribute to safer AVs.

    \noIndentHeading{Limitation.} \seabird does not fully solve the problem of generalization to large objects.

%% file: images/seabird/problem_setup.tex
\begin{tikzpicture}[scale=0.27, every node/.style={scale=0.45}, every edge/.style={scale=0.45}]
\tikzset{vertex/.style = {shape=circle, draw=black!70, line width=0.06em, minimum size=1.4em}}
\tikzset{edge/.style = {-{Triangle[angle=60:.06cm 1]},> = latex'}}

    \node [scale= 2, color=black] at (11.7, -7.5)  {(a)};

    \draw [draw=axisShadeDark, line width=0.08em, shorten <=0.5pt, shorten >=0.5pt, >=stealth]
           (6.05,1.3) node[]{}
        -- (6.05,4.1)  node[]{};
    \draw [-{Triangle[angle=60:.2cm 1]}, draw=axisShadeDark, line width=0.08em, shorten <=0.5pt, shorten >=0.5pt, >=stealth]
           (5.95,4.0) node[]{}
        -- (8.55,4.0)  node[]{};
        
    \draw [draw=axisShadeDark, line width=0.08em, shorten <=0.5pt, shorten >=0.5pt, >=stealth]
           (6.05,-0.3) node[]{}
        -- (6.05,-2.6)  node[]{};
    \draw [-{Triangle[angle=60:.2cm 1]}, draw=axisShadeDark, line width=0.08em, shorten <=0.5pt, shorten >=0.5pt, >=stealth]
           (5.95,-2.5) node[]{}
        -- (8.55,-2.5)  node[]{};
    
    \node[trapezium, draw=black!100, line width=0.05em, rotate=270, fill=my_blue!30, opacity=1.0, trapezium stretches=true, minimum width=2.5cm, minimum height=1cm, trapezium left angle=75, trapezium right angle=75] (t) at (9.25,4) {};

    \node[trapezium, draw=black!100, line width=0.05em, rotate=270, fill=darkGreen3!80, opacity=1.0, trapezium stretches=true, minimum width=2.5cm, minimum height=1cm, trapezium left angle=75, trapezium right angle=75] (t) at (9.25,-2.5) {};

    \node [scale= 2] at (6.0, 0.5)  {Image $\image$};
    \node [scale= 2] at (9.25, 4){$\layerWeight$};
    \node [scale= 2, color=darkGreen3] at (14.8, -2.5)  {Length $\length$};

    \draw [-{Triangle[angle=60:.2cm 1]}, draw=darkGreen3, line width=0.08em, shorten <=0.5pt, shorten >=0.5pt, >=stealth]
           (10.05,-2.5) node[]{}
        -- (12.4,-2.5)  node[]{};

    \draw [draw=my_blue!30, line width=0.18em, shorten <=0.5pt, shorten >=0.5pt, >=stealth]
           (10.05,4) node[]{}
        -- (11.4,4)  node[]{};
    \node [scale= 2] at (11.9, 4)  {$\bigoplus$}; 

    \node [scale= 2] at (11.9,9)  {Noise $\noise\!\sim\!\normal(0,\normalVar)$};

    \draw [-{Triangle[angle=60:.2cm 1]}, draw=blue, line width=0.08em, shorten <=0.5pt, shorten >=0.5pt, >=stealth]
           (11.9,7.8) node[]{}
        -- (11.9,4.5)  node[]{};
    \draw [-{Triangle[angle=60:.2cm 1]}, draw=my_blue, line width=0.18em, shorten <=0.5pt, shorten >=0.5pt, >=stealth]
           (12.4,4) node[]{}
        -- (14.2,4)  node[scale= 2]{};

    \node [scale= 2, color=my_blue] at (13.0,5.35)  {$\depthPred$};

    \draw[draw=black!100, fill=sns_orange, thick](15.5,4.0) circle (1.3) node[scale= 2]{$\loss$};

    \draw [-{Triangle[angle=60:.2cm 1]}, draw=darkGreen3, line width=0.08em, shorten <=0.5pt, shorten >=0.5pt, >=stealth]
           (15.5,0.7) node[]{}
        -- (15.5,2.7)  node[]{};

    \node [scale= 2, color=darkGreen3] at (15.5,-0.3)  {GT $\depthGT$};

    \node [scale= 2, color=black] at (24, -7.5)  {(b)};
    \draw [draw=black!100, line width=0.06em](20, 8) rectangle (28, 2) node[]{};
    \draw [draw=black!100, line width=0.06em](20, 1) rectangle (28,-5) node[]{};

    \draw[black,fill=black!50] (23.7,-5.7) rectangle (24.3,-6.7);
    \coordinate (c10) at (24,-5.7);
    \coordinate (c11) at (23.7,-5.2);
    \coordinate (c12) at (24.3,-5.2);
    \filldraw[draw=black, fill=gray!20] (c10) -- (c11) -- (c12) -- cycle;

    \draw [-{Triangle[angle=60:.15cm 1]}, draw=black, line width=0.04em, shorten <=0.5pt, shorten >=0.5pt, >=stealth]
           (19.0,-6.6) node[]{}
        -- (19.0,-4.6)  node[scale=2,text width=0.2cm]{$Z$\\~};

    \draw [-{Triangle[angle=60:.15cm 1]}, draw=black, line width=0.04em, shorten <=0.5pt, shorten >=0.5pt, >=stealth]
           (18.9,-6.5) node[]{}
        -- (20.9,-6.5)  node[scale=2]{~~$X$};

    \node [scale= 2] at (24,9)   {BEV};
    \node [scale= 2, color=darkGreen3] at (18.7,7) {GT};
    \node [scale= 2, color=my_blue  ] at (18.7,0) {Pred};
    
    \draw [draw=rayShade, line width=0.1em]
           (24,2.08) node[]{}
        -- (24,7.95)  node[]{};
    \draw [draw=black!100, line width=0.06em, fill=darkGreen3](23.5, 4.2) rectangle (24.5, 2.2) node[]{};
    \draw[draw=black!100, fill=black!100, thick](24,3.2) circle (0.1) node[]{};
    \node [scale= 1.75, color=darkGreen3] at (22.25,3.2)  {$(0, \depthGT)$};

    \draw [draw=gray!50, line width=0.08em]
           (25,4.2) node[]{}
        -- (25,2.2) node[pos=0.5,scale=2]{~~~$\length$};

    \draw [draw=rayShade, line width=0.1em]
           (24,0.98) node[]{}
        -- (24,-4.95)  node[]{};
    \draw [draw=black!100, line width=0.06em, fill=my_blue](23.5, 0.2) rectangle (24.5, -1.8) node[]{};
    \draw[draw=black!100, fill=black!100, thick](24,-0.8) circle (0.1) node[]{};
    \node [scale= 1.75, color=my_blue] at (22.25,-0.8)  {$(0, \depthPred)$};

    \draw [draw=gray!50, line width=0.08em]
           (25,0.2) node[]{}
        -- (25,-1.8) node[pos=0.5,scale=2]{~~~$\length$};

    
    \node [scale= 2, color=black] at (32, -7.5)  {(c)};
    \node [scale= 2] at (32,9)   {CS View};
    \draw [-{Triangle[angle=60:.2cm 1]}, draw=rayShade, line width=0.1em, shorten <=0.5pt, shorten >=0.5pt, >=stealth]
           (33, 1.5) node[]{}
        -- (33, 8) node[scale=2,text width=0.5cm]{~\\~~~~$\varZ$};
    \draw [draw=black!100, line width=0.08em]
           (30.5, 2) node[]{}
        -- (33.5, 2) node[]{};

    \draw [draw=darkGreen3, line width=0.12em]
           (31.42, 2.2) node[]{}
        -- (32.95, 2.2) node[]{};
    \draw [draw=darkGreen3, line width=0.12em]
           (31.5, 2.15) node[]{}
        -- (31.5, 4.25) node[]{};
    \draw [draw=darkGreen3, line width=0.12em]
           (31.42, 4.2) node[]{}
        -- (32.95, 4.2) node[]{};

    \draw [draw=black!100, line width=0.08em]
           (32.85, 3.2) node[]{}
        -- (33.15, 3.2) node[]{};
    \node [scale= 1.75, color=darkGreen3] at (33.4,3.2)  {$\depthGT$};

    \draw [draw=gray!50, line width=0.1em]
           (33.8, 2.2) node[]{}
        -- (33.8, 4.2) node[pos=0.5,scale=2,text width=0.5cm]{~~~$\length$};

    \node [scale=2, color=black,rotate=90] at (29.8,1.7)  {$P(\varZ)$};

    \draw [-{Triangle[angle=60:.2cm 1]}, draw=rayShade, line width=0.1em, shorten <=0.5pt, shorten >=0.5pt, >=stealth]
           (33, -5.5) node[]{}
        -- (33, 1.5) node[scale=2,text width=0.5cm]{~\\~~~~$\varZ$};
    \draw [draw=black!100, line width=0.08em]
           (30.5, -5) node[]{}
        -- (33.5, -5) node[]{};

    \draw [draw=my_blue  , line width=0.12em]
           (31.42, 0.2) node[]{}
        -- (32.95, 0.2) node[]{};
    \draw [draw=my_blue  , line width=0.12em]
           (31.5, 0.15) node[]{}
        -- (31.5, -1.85) node[]{};
    \draw [draw=my_blue  , line width=0.12em]
           (31.42, -1.8) node[]{}
        -- (32.95, -1.8) node[]{};

    \draw [draw=black!100, line width=0.08em]
           (32.85, -0.8) node[]{}
        -- (33.15, -0.8) node[]{};
    \node [scale= 1.75, color=my_blue] at (33.4,-0.7)  {$\depthPred$};

    \draw [draw=gray!50, line width=0.1em]
           (33.8, 0.2) node[]{}
        -- (33.8, -1.8) node[pos=0.5,scale=2,text width=0.5cm]{~~~$\length$};

    \draw [draw=black!50, dash pattern=on 2pt off 1.3pt, line width=0.05em]
           (31.5, -4.9) node[]{}
        -- (31.5, -1.9) node[]{};

    \node [scale=2, color=black,rotate=90] at (31.5,-5.5)  {$1$};
    \node [scale=2, color=black,rotate=90] at (29.8,-5.0)  {$P(\varZ)$};
    
\end{tikzpicture}

%% file: chapters/charmer.tex
\chapter{
    \charmer: Towards \charmerFull
}
\label{chpt:charm3r}

    To this end, we attempt generalizing \monoThreeD networks to occlusion, dataset and object sizes. 
    Monocular 3D object detectors, while effective on data from one ego camera height, struggle with unseen or out-of-distribution camera heights. 
    Existing methods often rely on Plucker embeddings, image transformations or data augmentation. 
    This chapter takes a step towards this understudied problem by investigating the impact of camera height variations on state-of-the-art (SoTA) Mono3D models.
    With a systematic analysis on the extended CARLA dataset with multiple camera heights, we observe that depth estimation is a primary factor influencing performance under height variations. 
    We mathematically prove and also empirically observe consistent negative and positive \trends in mean depth error of regressed and ground-based depth models, respectively, under camera height changes. 
    To mitigate this, we propose Camera Height Robust Monocular 3D Detector (CHARM3R), which averages both depth estimates within the model.
    \charmer significantly improves generalization to unseen camera heights, achieving \sota performance on the CARLA dataset. 

\section{Introduction}\label{sec:charmer_intro}

    Monocular \threeD object detection (\monoThreeD) task uses a single image to determine both the \threeD location and dimensions of objects. 
    This technology is essential for augmented reality \cite{alhaija2018augmented,Xiang2018RSS,park2019pix,merrill2022symmetry}, robotics \cite{saxena2008robotic}, and self-driving cars \cite{park2021pseudo,kumar2022deviant,li2022bevformer}, where accurate \threeD understanding of the environment is crucial. 
    Our research specifically focuses on using \threeD object detectors applied to autonomous vehicles (AVs), as they have unique challenges and requirements.

    AVs necessitate detectors that are robust to a wide range of intrinsic and extrinsic factors, including intrinsics \cite{brazil2023omni3d}, domains \cite{li2024unimode}, object size \cite{kumar2024seabird}, rotations \cite{zhou2021monoef,moon2023rotation}, weather conditions \cite{lin2024monotta,oh2024monowad}, and adversarial examples \cite{zhu2023understanding}. 
    Existing research primarily focusses on generalizing object detectors to these failure modes. 
    However, this work investigates the generalization of \monoThreeD to another type, which, thus far, has been relatively understudied in the literature – {\it \monoThreeD generalization to unseen ego camera \variations}.

        \begin{figure}[!t]
            \centering
            \includegraphics[width=0.7\linewidth]{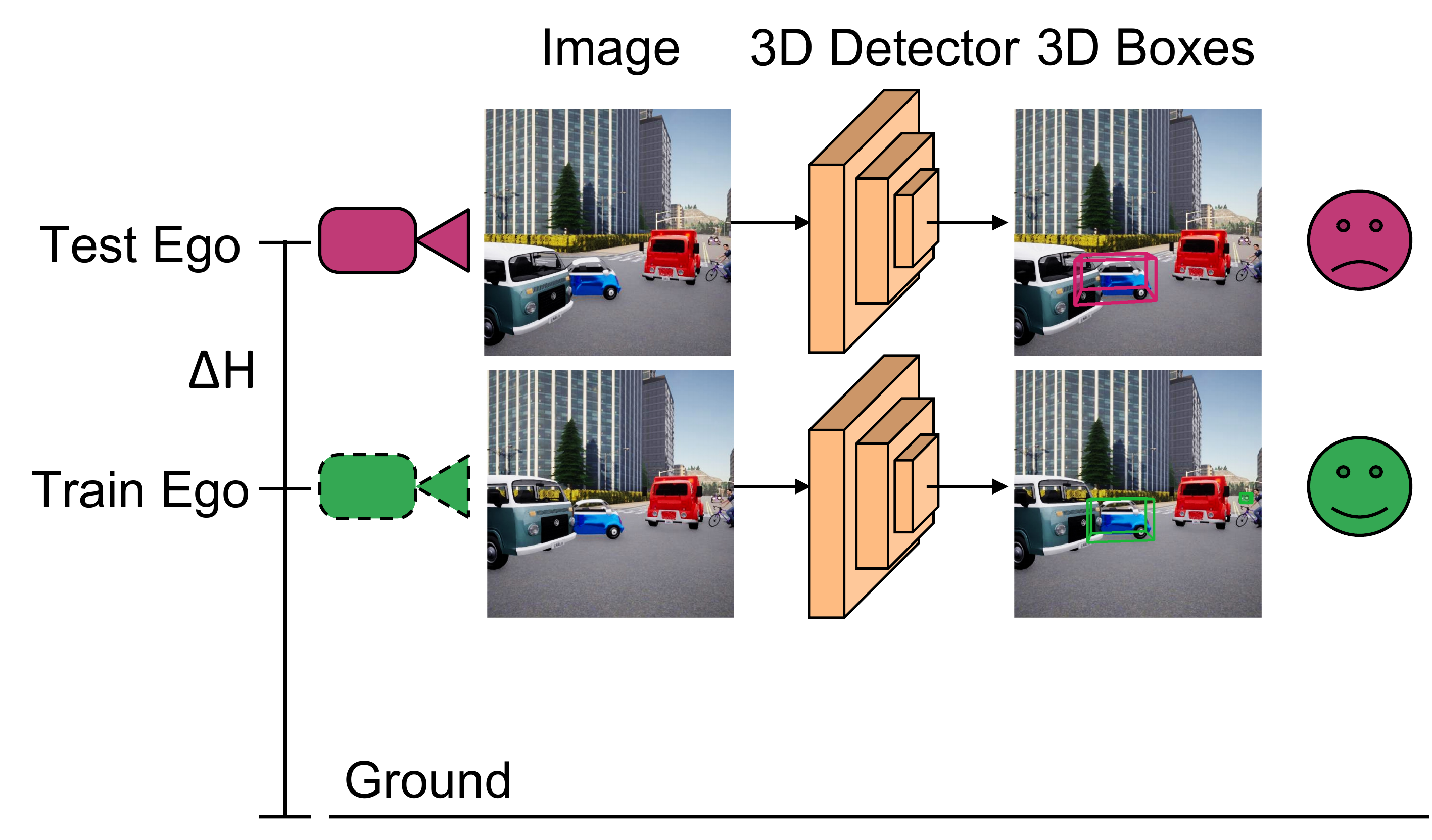}
            \caption[\charmer Teaser.]
            {\textbf{Teaser.} Changing ego \variation at inference quickly \textbf{drops} \monoThreeD performance of \sota detectors. A \variation change $\egoHeightChange$ of $0.76m$ in inference drops \apThreeD \bracketPercentage~by absolute $35$ points. }
            \label{fig:charmer_teaser}
        \end{figure}

        \begin{figure*}[!t]
            \centering
            \begin{subfigure}{.32\linewidth}
                \includegraphics[width=\linewidth]{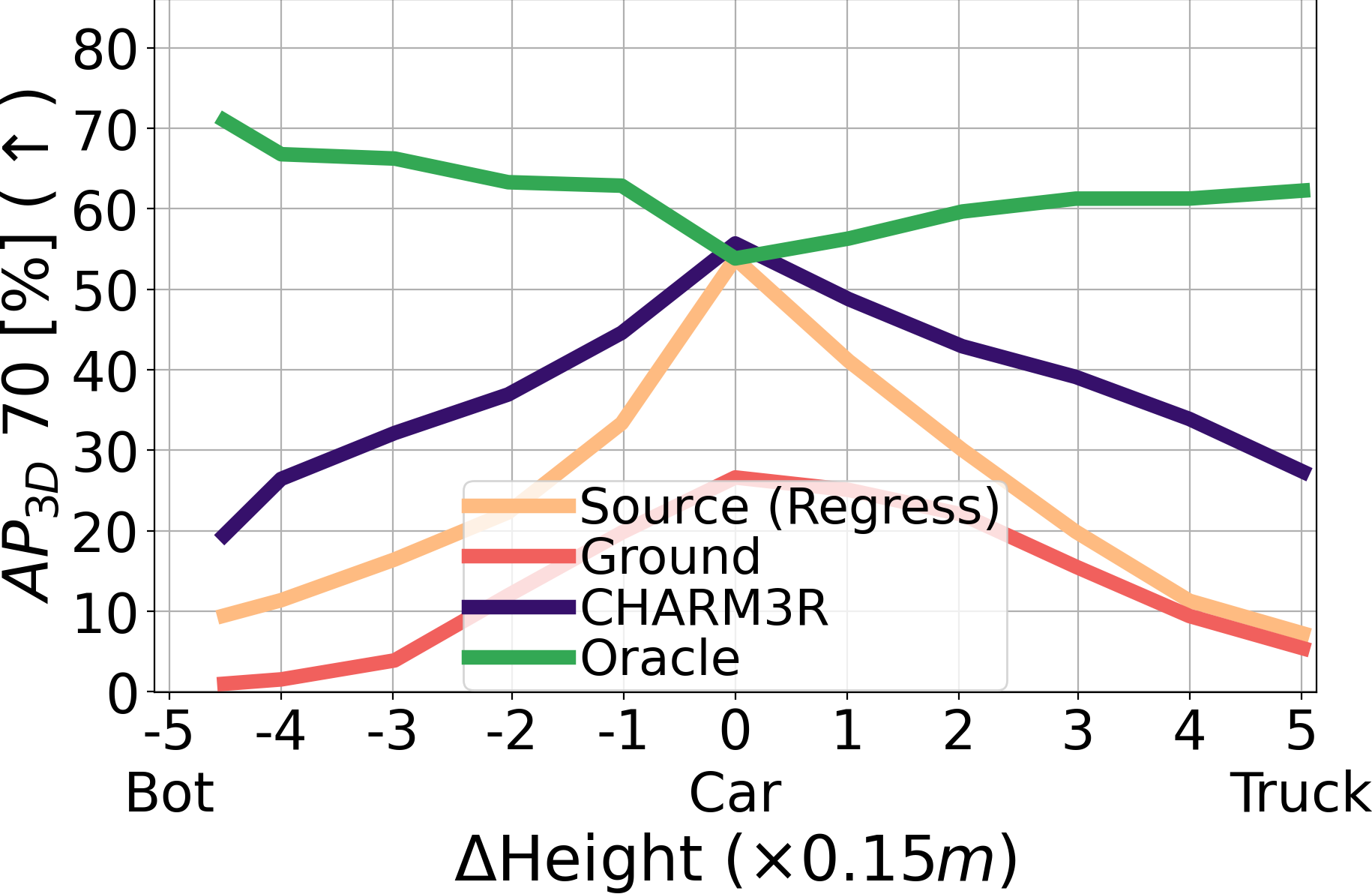}
                \caption{\apThreeDSeventy \bracketPercentage{} Results.}
                \label{fig:charmer_reason_seventy}
            \end{subfigure}%
            \hfill
            \begin{subfigure}{.32\linewidth}
                \includegraphics[width=\linewidth]{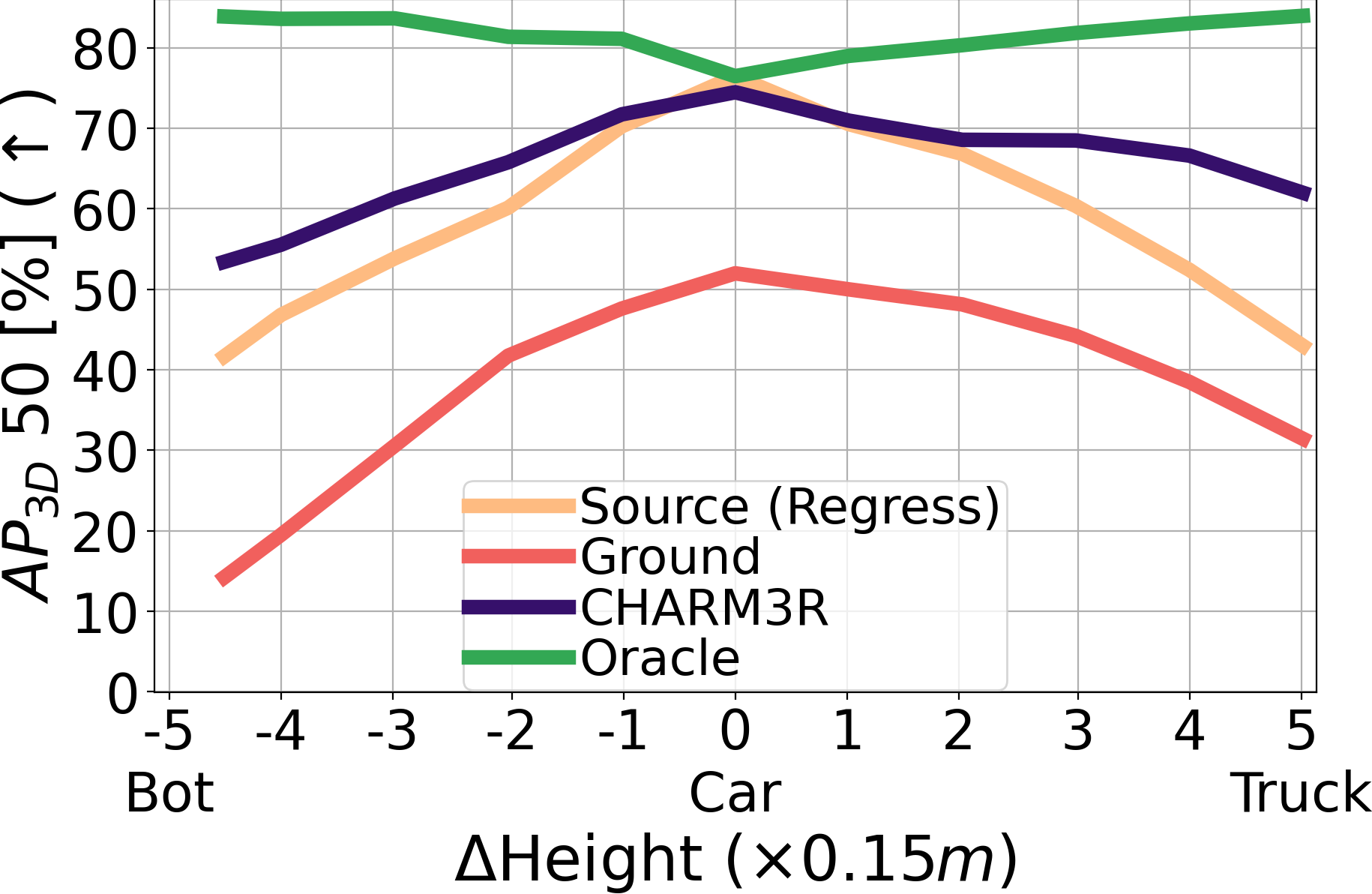}
                \caption{\apThreeDFifty \bracketPercentage{} Results.}
                \label{fig:charmer_reason_fifty}
            \end{subfigure}
            \hfill
            \begin{subfigure}{.32\linewidth}
                \includegraphics[width=\linewidth]{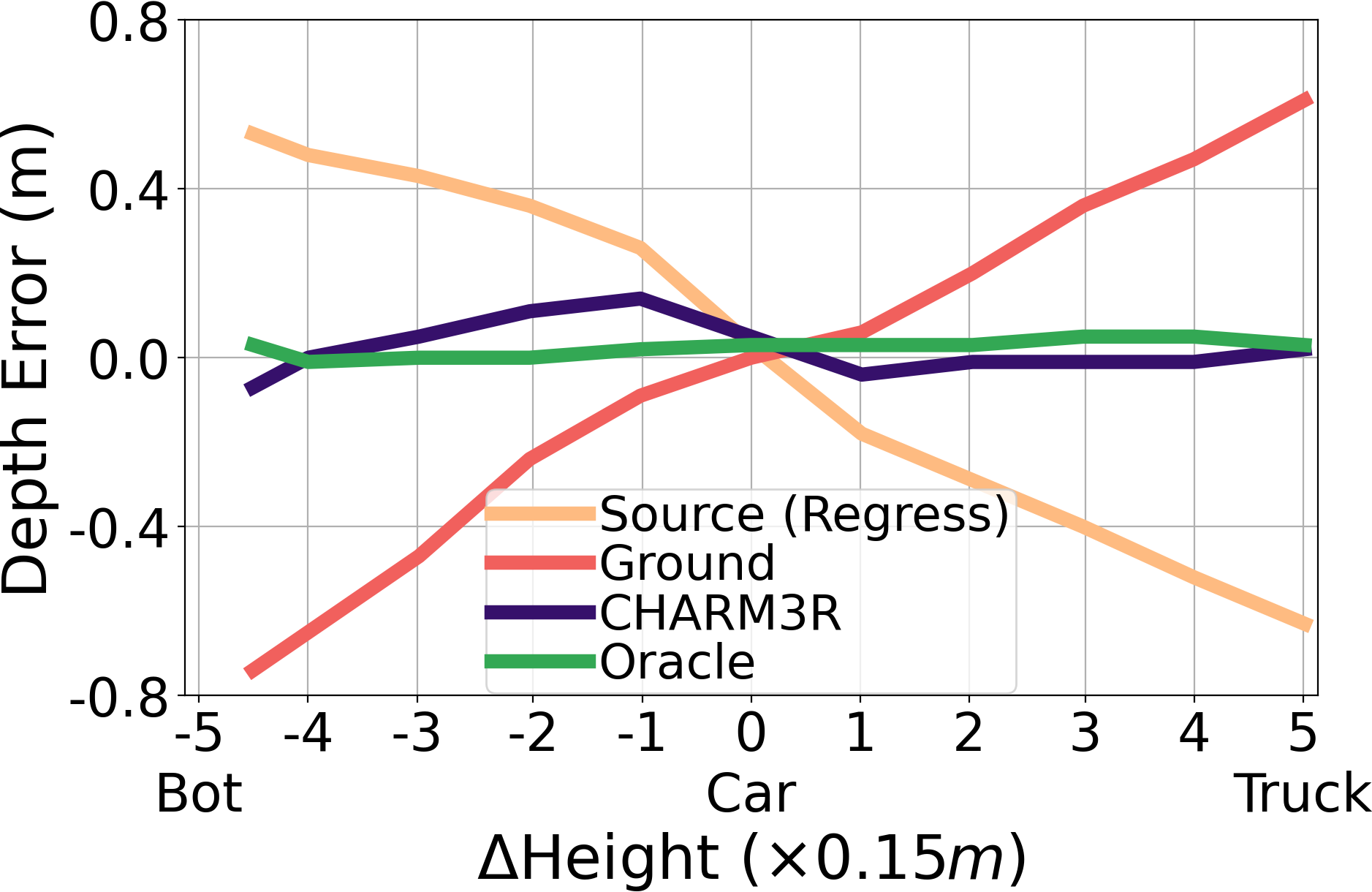}
                \caption{Depth error \trend on changing ego \variations.}
                \label{fig:charmer_reason_bias}
            \end{subfigure}
            \caption[The performance of \sota detector \gupNet drops significantly with changing ego \variations in inference.]
            {\textbf{Performance Comparison.} The performance of \sota detector \gupNet \cite{lu2021geometry} drops significantly with changing ego \variations in inference. 
            Ground-based model shows contrasting depth error (extrapolation) trend compared to regression-based depth models.
            Our proposed \textbf{\charmer exhibits greater robustness} to such variations by averaging regression and ground-based depth estimates. All methods, except the Oracle, are trained on car-height data $\egoHeightChange=0m$ and tested on data from bot to truck \variations.
            }
            \label{fig:charmer_reason}
        \end{figure*}


    The ego \variation of autonomous vehicles (AVs) varies significantly across different platforms and deployment scenarios. 
    While almost all training data is collected from a specific ego \variation, such as that of a passenger car, AVs are now deployed with substantially different ego \variation such as small bots or trucks. 
    Collecting, labeling datasets and retraining models for each possible \variation is not scalable \cite{tzofi2023towards}, computationally expensive and impractical. 
    Therefore, our work aims to address the challenge of \textit{generalizing} \monoThreeD models to \textit{unseen} ego \variations.

    Generalizing \monoThreeD to unseen ego \variations from \textbf{single ego \variation} data is challenging due to the following five reasons.
    First, neural models excel at In-Domain (\inDomain{}) generalization, but struggle with unseen Out-Of-Domain (\outDomain{}) generalization \cite{xu2021how,teney2023id}.
    Second, ego \variation changes induce projective transformations \cite{hartley2003multiple} that CNNs \cite{cohen2016group}, \deviant \cite{kumar2022deviant} or \vit \cite{dosovitskiy2021image} backbones do not effectively handle \cite{sarkar2024shadows}.
    Third, existing projective equivariant backbones \cite{macdonald2022enabling,mironenco2024lie} are limited to single-transform-per-image scenarios, while every pixel in a driving image undergoes a different depth-dependent transform. 
    Fourth, the non-linear nature \cite{burns1992non,hartley2003multiple} of projective transformations makes interpolation difficult.
    Finally, disentangled learning does not work for this problem since such approaches need at least two \variation data, while the training data here is from \textbf{single \variation}.
    Note that the generalization from single height to multi heights is more practical since multi-height data is unavailable in almost all real datasets.

    We first systematically analyze and quantify the impact of ego \variation on the performance of \monoThreeD models trained on a single ego \variation. 
    Leveraging the extended \carla dataset \cite{tzofi2023towards}, we evaluate the performance of state-of-the-art (\sota) \monoThreeD models under multiple ego \variations. 
    Our analysis reveals that \sota \monoThreeD models exhibit significant performance degradation when faced with large \variation changes in inference (\cref{fig:charmer_reason_seventy,fig:charmer_reason_fifty}). 
    Additionally, we empirically observe a consistent negative trend in the regressed object depth under \variation changes (\cref{fig:charmer_reason_bias}).
    Furthermore, we decompose the performance impact into individual sub-tasks and identify depth estimation as the primary contributor to this degradation. 

    Recent papers address ego \variation changes by using \plucker embeddings \cite{bahmani2024vd3d}, transforming target-height images to the original height, assuming constant depth \cite{li2024unidrive}, or by retraining with augmented data\cite{tzofi2023towards}.
    While these techniques do offer some effectiveness, image transformation fails (\cref{fig:charmer_det_results_carla_val}) under significant \variation changes due to real-world depth variations.
    The augmentation strategy requires complicated pipelines for data synthesis at target heights and also falls short when the target \variation is \outDomain or when the target \variation is unknown apriori during training.

    To effectively generalize \monoThreeD to unseen ego \variations, a detector should first disentangle the depth representation from ego parameters in training and produce a new representation with new ego parameters in inference, while also canceling the \trends. 
    We propose using the projected bottom \threeD center and ground depth in addition to the regressed depth. 
    While the ground depth is easily calculated from ego parameters and height, and can be changed based on the ego \variation, its direct application to \monoThreeD models is sub-optimal (a reason why ground plane is not used alone).
    However, we observe a consistent positive trend in ground depth, which contrasts with the negative trend in regressed depths.
    By averaging both depth estimates within the model, we effectively cancel these opposing \trends and improve \monoThreeD generalization to unseen ego \variations.

    In summary the main contributions of this work include:
    \begin{itemize}
        \item We attempt the understudied problem of \outDomain ego \variation robustness in \monoThreeD models from single \variation data.
        \item We mathematically prove systematic negative and positive \trends in the regressed and ground-based object depths, respectively, under ego \variation changes under simplified assumptions (\cref{theorem:1,theorem:2}).
        \item We propose simple averaging of these depth estimates within the model to effectively counteract these opposing \trends and generalize to unseen ego \variations (\cref{sec:depth_merge}).
        \item We empirically demonstrate \sota robustness to unseen ego \variation changes on the \carla dataset (\cref{tab:det_results_carla_val}).
    \end{itemize}

\section{Related Works}\label{sec:charmer_literature}

    \noIndentHeading{Extrapolation / \outDomain Generalization.}
        Neural models excel at \inDomain generalization, but struggle at \outDomain generalization \cite{xu2021how,teney2023id}. 
        There are two major classes of methods for good \outDomain classification. 
        The first does not use target data and relies on diversifying data \cite{teney2021unshuffling}, features \cite{yashima2022feature,tiwari2023overcoming}, predictions \cite{lee2022diversify}, gradients \cite{ross2018learning,teney2022evading} or losses \cite{sagawa2019distributionally,ruan2023towards,puli2023don}.
        Another class finetunes on small target data \cite{kirichenko2022last}.
        None of these papers attempt \outDomain generalization for regression tasks.
    
    \noIndentHeading{Mono3D.}
        \monoThreeD has gained significant popularity, offering a cost-effective and efficient solution for perceiving the \threeD world.
        Unlike its more expensive \lidar and \radar counterparts \cite{shi2019pointrcnn,yin2021center,long2023radiant}, or its computationally intensive stereo-based cousins \cite{Chen2020DSGN}, \monoThreeD relies solely on a single camera or multiple cameras with little overlaps.
        Earlier approaches to this task \cite{payet2011contours, chen2016monocular} relied on hand-crafted features, while the recent advancements use deep models. 
        Researchers explored a variety of approaches to improve performance, including architectural innovations \cite{huang2022monodtr,xu2023mononerd}, equivariance \cite{kumar2022deviant, chen2023viewpoint}, losses \cite{brazil2019m3d,chen2020monopair}, uncertainty \cite{lu2021geometry,kumar2020luvli} and depth estimation \cite{zhang2021objects,min2023neurocs,yan2024monocd}.
        A few use NMS \cite{kumar2021groomed,liu2023monocular}, corrected extrinsics \cite{zhou2021monoef}, CAD models \cite{chabot2017deep, liu2021autoshape, lee2023baam} or \lidar \cite{reading2021categorical} in training.
        Other innovations include \pseudoLidar \cite{wang2019pseudo, ma2019accurate}, diffusion \cite{ranasinghe2024monodiff,xu20243difftection}, \bev feature encoding \cite{jiang2024fsd,zhang2022beverse,li2024bevnext} or transformer-based \cite{carion2020detr} methods with modified positional encoding \cite{shu2023dppe,tang2024simpb,hou2024open},  queries\cite{li2023fast,zhang2023dabev,ji2024enhancing,chen2024learning} or query denoising \cite{liu2024ray}.
        Some use pixel-wise depth \cite{huang2021bevdet} or object-wise depth \cite{chu2023oabev,choi2023depth,liu2021voxel}.
        Many utilize temporal fusion with short \cite{wang2022sts,wang2023stream,liu2023petrv2,brazil2020kinematic} or long frame history \cite{park2022time,zong2023hop,changrecurrentbev} to boost performance.
        A few use distillation \cite{wang2023distillbev,kim2024labeldistill}, stereo \cite{wang2022sts,li2023bevstereo} or loss \cite{kumar2024seabird,liu2024multi} to improve these results further.
        For a comprehensive overview, we redirect readers to the surveys \cite{ma20233d,ma2022vision}.
        \charmer selects representative \monoThreeD models and improves their extrapolation to unseen camera \variations.

    \noIndentHeading{Camera Parameter Robustness.} 
        While several works aim for robust \lidar-based detectors \cite{hu2022investigating,wang2020train,yang2021st3d,xu2021spg,chang2024cmda},  planners \cite{yao2024improving} and map generators \cite{ranganatha2024semvecnet}, fewer studies focus on generalizing image-based detectors.
        Existing image-based techniques, such as self-training \cite{li2022unsupervised}, adversarial learning \cite{wang2023towards}, perspective debiasing \cite{lu2023towards}, and multi-view depth constraints \cite{chang2024unified}, primarily address datasets with variations in camera intrinsics and minor height differences of $0.2m$. 
        Some papers show robustness to other camera parameters such as intrinsics \cite{brazil2023omni3d}, and rotations \cite{zhou2021monoef,moon2023rotation}.
        \charmer specifically tackles the challenge of generalizing to scenarios with significant camera height changes, exceeding $0.7m$.
    
    \noIndentHeading{\Variation-Robustness.} 
        Image-based \threeD detectors such as \bevHeight \cite{jia2023monouni} and MonoUNI \cite{jia2023monouni} train multiple detectors at different heights, but always do \inDomain testing.
        Recent works address ego \variation changes by either using \plucker embeddings \cite{bahmani2024vd3d,zhang2024camerasrays} for video generation/pose estimation, by transforming target-height images to the original height, assuming constant depth \cite{li2024unidrive} for \monoThreeD, or by retraining with augmented data\cite{tzofi2023towards} for \bev segmentation.
        In contrast, we investigate the contrasting extrapolation behavior of regressed and ground-based depth estimators and average them for generalizing  \monoThreeD to unseen camera \variations.

    \noIndentHeading{Wide Baseline Setup.}
        Wide baseline setups are challenging due to issues like large occlusions, depth discontinuities \cite{strecha2003dense} and intensity variations \cite{strecha2004wide}.
        Unlike traditional wide-baseline setups with arbitrary baseline movements, generalization to unseen ego \variation requires handling baseline movements specifically along the vertical direction. 

\section{Notations and Preliminaries}

    We first list out the necessary notations and preliminaries which are used throughout this chapter.
    These are not our contributions and can be found in the literature \cite{hartley2003multiple,garnett20193d,guo2020gen}.

\begin{figure}[!t]
        \centering
        \includegraphics[width=\linewidth]{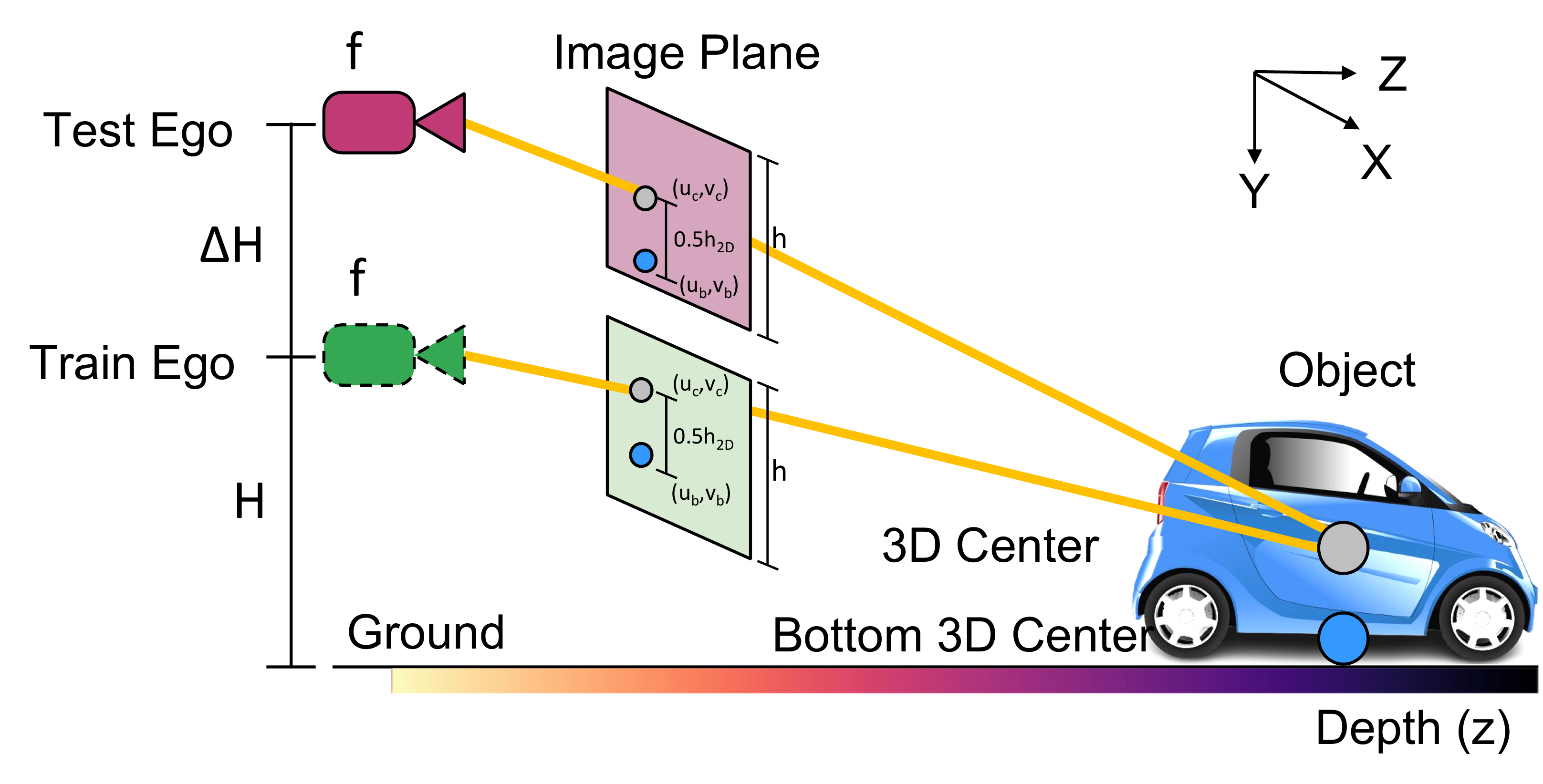}
        \caption[Problem Setup.]{\textbf{Problem Setup}. Note that changing ego height does not change the object depth $\depthGT$ but only its position $(\pixUCenter, \pixVCenter)$ in the image plane. A regressed-depth model uses this pixel position to estimate the depth and therefore, fails when the ego height is changed.}
        \label{fig:charmer_setup}
    \end{figure}

    \noIndentHeading{Notations.}
    Let $\intrinsic\!\in\!\realDomain^{3 \times 3}$ denote the camera intrinsic matrix, $\rotation\!\in\!\realDomain^{3 \times 3}$ the rotation matrix and $\extrinsicTrans\!\in\!\realDomain^{3 \times 1}$ the translation vector of the extrinsic parameters. 
    Also, $\zeroThreeD\!\in\!\realDomain^{3 \times 1}$ denotes the zero vector in \threeD.
    We denote the ego camera height on the car as $\egoHeight$, and the \variation change relative to this car as $\egoHeightChange$ meters.
    The camera intrinsics matrix $\intrinsic$ has focal length $\focal$ and principal point $(\ppointU, \ppointV)$.
    Let $(\pixU, \pixV)$ represent a pixel position in the camera coordinates, and  
    $(\pixUCenter, \pixVCenter)$ and $(\pixUBottom, \pixVBottom)$ denotes the projected \threeD center and bottom center respectively.
    $\heightImage$ denotes the height of the image plane.
    We show these notations pictorially in \cref{fig:charmer_setup}.

    \begin{figure*}[!t]
        \centering
        \includegraphics[width=1\linewidth]{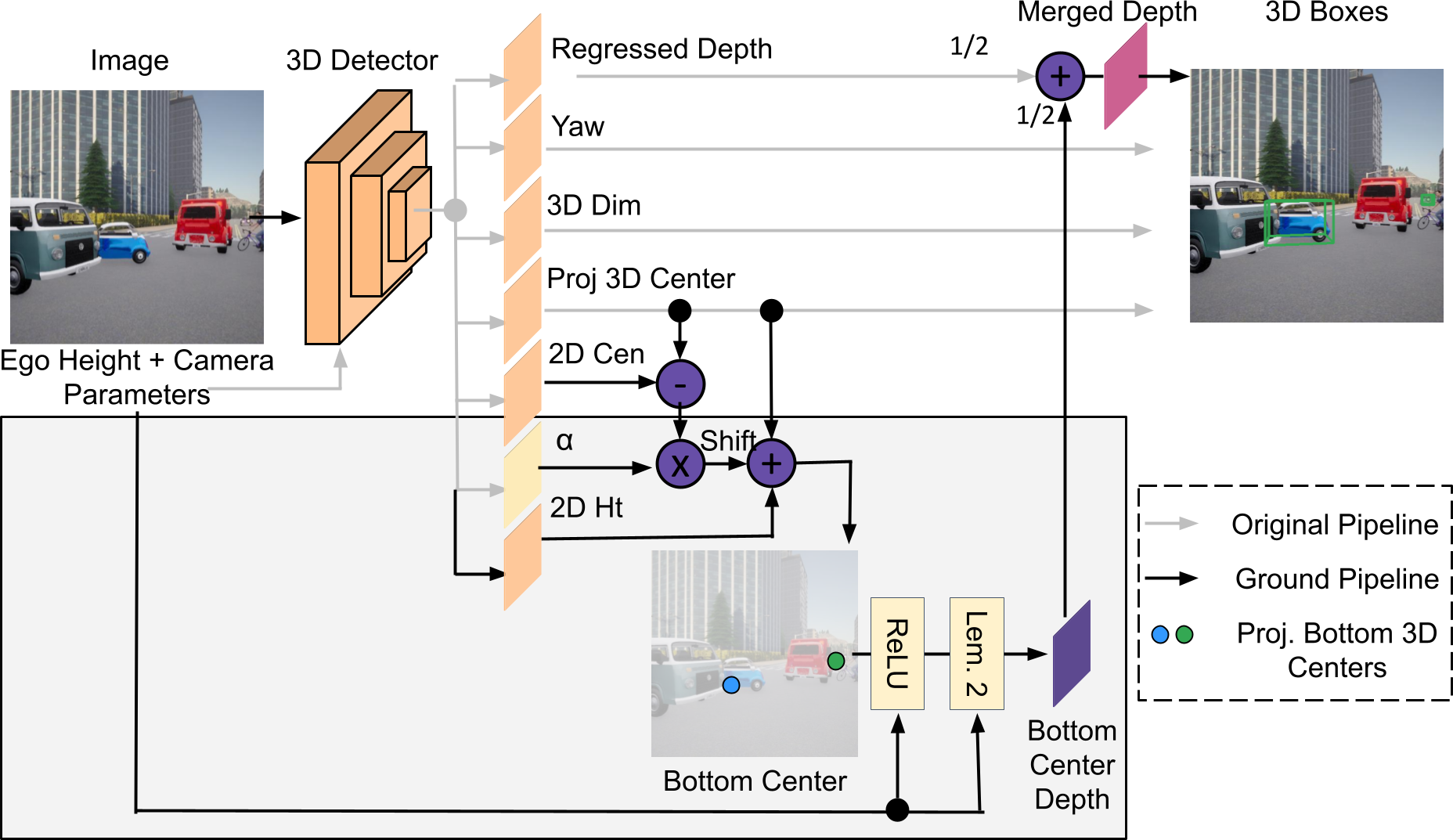}
        \caption[\charmer Overview.]
        {\textbf{\charmer Overview.} \charmer predicts the shift coefficient to obtain projected \threeD bottom centers to query the ground depth and then averages the ground-depth and the regressed depth estimates within the model itself to output final depth estimate of a bounding box.
        \charmer uses the results of \cref{theorem:1,theorem:2} that demonstrate that the ground and the regressed depth models show contrasting extrapolation behaviors.
        }
        \label{fig:charmer_overview}
    \end{figure*}

    \noIndentHeading{Pinhole Point Projection }\cite{hartley2003multiple}.
        The pinhole model relates a \threeD point $(\varX,\varY,\varZ)$ in the world coordinate system to its \twoD projected pixel $(\pixU, \pixV)$ in camera coordinates as:
        \begin{align}
        \begin{bmatrix}
        \pixU \\
        \pixV \\
        1 \\
        \end{bmatrix}\posZ
        =
        \begin{bmatrix}
        \intrinsic\;\;\;\zeroThreeD
        \end{bmatrix}
        \begin{bmatrix}
        \rotation\;\;\;\extrinsicTrans \\
        \zeroThreeD^T\;\;1
        \end{bmatrix}
        \begin{bmatrix}
        \varX \\
        \varY \\
        \varZ \\
        1
        \end{bmatrix},
        \label{eq.pinhole}
        \end{align}
        where $\posZ$ denotes the depth of pixel $(\pixU, \pixV)$.

    \noIndentHeading{Ground Depth Estimation }\cite{garnett20193d,guo2020gen}.
        While depth estimation in \monoThreeD is ill-posed, ground depth can be precisely determined given the camera parameters and height relative to the ground in the world coordinate system \cite{garnett20193d,guo2020gen,yang2023gedepth}.
        Since all datasets provide camera mounting height from the ground, we obtain the depth of ground plane pixels in closed form.
        
        \begin{lemma}\label{lemma:charmer_1}
            \textbf{Ground Depth of Pixel }\cite{garnett20193d,guo2020gen,yang2023gedepth}.
            Consider a pinhole camera model with intrinsics $\intrinsic$, rotation $\rotation$ and translation extrinsics $\extrinsicTrans$. Let matrix $\bm{A}\!=\!(a_{ij})\!=\!\rotation^{-1}\intrinsic^{-1}\in \realDomain^{3 \times 3}$, and $-\rotation^{-1}\extrinsicTrans$ as the vector $\bm{B}\!=\!(b_i) \in \realDomain^{3 \times 1}$. Then, the ground depth $\posZ$ for a pixel $(\pixU, \pixV)$ is 
            \begin{align}
                \posZ = \frac{\egoHeight - b_{2}}{a_{21}\pixU + a_{22}\pixV + a_{23}}.
                \label{eq:gd_depth}
            \end{align}
        \end{lemma}
        We refer to \cref{sec:charmer_proof_lemma} in the appendix for the derivation.

        \begin{lemma}\label{lemma:charmer_2}
        \textbf{Ground Depth of Pixel }For datasets with the rotation extrinsics $\rotation$ an identity, the depth estimate $\posZ$ from \cref{lemma:charmer_1} becomes
        \begin{align}
            \posZ &= \dfrac{\egoHeight - b_{2}}{~\dfrac{\pixV\!-\!\ppointV}{\focal}~}.
            \label{eq:gd_depth_simple}
        \end{align}
        \end{lemma}
        We refer to \cref{sec:charmer_proof_simple} for the proof.

\section{\charmer}\label{sec:charmer_method}

    In this section, we first mathematically prove the contrasting extrapolation behavior of regressed and ground-based object depths under varying camera \variations. 
    To mitigate the impact of these opposing \trends and improve generalization to unseen \variations, we propose \charmerFull or \charmer. 
    \charmer averages both these depth estimates within the model to mitigate these \trends and improves generalization to unseen \variations.
    \cref{fig:charmer_overview} shows the overview of \charmer.

    \subsection{Ground-based Depth Model}\label{sec:charmer_bottom_ground_depth}
        Outdoor driving scenes typically contain a ground region, unlike indoor scenes. 
        The ground depth varies with ego \variation, providing a valuable reference and prior for generalizing \monoThreeD to unseen ego \variations.

        \noIndentHeading{Bottom Center Estimation.}
            \cref{lemma:charmer_1} utilizes the ground plane depth from \cref{eq:gd_depth} to estimate object depths.
            The numerator in \cref{eq:gd_depth} can be negative, while depth is positive for forward facing cameras.
            To ensure positive depth values, we apply the Rectified Linear Unit (\relu) activation $(\max(\posZ, 0))$ to the numerator of \cref{eq:gd_depth}.
            This step promotes spatially continuous and meaningful ground depth representations, improving the training stability of \charmer. 
            Ablation in \cref{sec:charmer_ablation} confirm the effectiveness.
    
            In practice, \charmer leverages the projected \threeD center $(\pixUCenter,\pixVCenter)$, \twoD height information $\heightBoxTwoD$ and the \twoD center  $(\pixUCenterTwoD,\pixVCenterTwoD)$ to compute the projected bottom \threeD center $(\pixUBottom,\pixVBottom)$ as follows:
            \begin{align}
                \pixUBottom &= \pixUCenter~~; \quad 
                \pixVBottom = \pixVCenter + \frac{1}{2}\heightBoxTwoD + \alpha (\pixVCenter - \pixVCenterTwoD).
                \label{eq:bottom_center}
            \end{align}

            With the projected bottom center $(\pixUBottom,\pixVBottom)$ estimated, we query the ground plane depth at this point, as derived in \cref{lemma:charmer_1}. 
            Note that we do not use the \threeD height to calculate the bottom center since projecting this point requires the box depth, which is the quantity we aim to estimate. 
            We, now, analyze the extrapolation behavior of this ground-based depth model in the following theorem.
    
        \begin{theorem}\label{theorem:1}
            \textbf{Ground-based bottom center model has positive slope (trend) in extrapolation.}
            Consider a ground depth model that predicts $\depthPred$ from the projected bottom \threeD center $(\pixUBottom,\pixVBottom)$ image.
            Assuming the GT object depth $\depthGT$ is more than the ego height change $\egoHeightChange$, the mean depth error of the ground model exhibits a positive \trend \wrt the \variation change $\egoHeightChange$:
            \begin{align}
                \label{eqn:ground_depth_bias}
                \expect\Big(\newDepthGround - \depthGT\Big) &\approx
                \relu\left(\frac{1}{\pixVBottom\!-\!\ppointV}\right)\focal\egoHeightChange,
            \end{align}
            where $\focal$ is the focal length and $(\ppointU,\ppointV)$ is the optical center.
        \end{theorem}
        \cref{theorem:1} says that the ground model over-estimates and under-estimates depth as the ego \variation change $\egoHeightChange$ increases and decreases respectively.
        \begin{proof}            
            When the ego camera shifts by $\egoHeightChange~m$, the $y$-coordinate of the projected \threeD bottom center $\pixVBottom$ of a \threeD box becomes $\pixVBottom\!+\! \dfrac{\focal\egoHeightChange}{\depthGT}$. 
            Using \cref{eq:gd_depth_simple}, the new depth $\newDepthGround$ is
            \begin{align}
                \newDepthGround = \dfrac{\egoHeight + \egoHeightChange - b_{2}}{\dfrac{~\pixVBottom + \dfrac{\focal\egoHeightChange}{\depthGT} - \ppointV}{\focal}~}
                &= \dfrac{\egoHeight + \egoHeightChange - b_{2}}{~\dfrac{\pixVBottom\!-\!\ppointV}{\focal} + \dfrac{\egoHeightChange}{\depthGT}~}.
            \end{align}
            If the ego height change $\egoHeightChange$ is small compared to the object depth $\depthGT$, $\dfrac{\egoHeightChange}{\depthGT} \approx 0$.
            So, we write the above equation as
            \begin{align}
                \newDepthGround &\approx \dfrac{\egoHeight + \egoHeightChange - b_{2}}{\dfrac{\pixVBottom\!-\!\ppointV}{\focal}} = \oldDepthGround + \dfrac{\egoHeightChange}{~\dfrac{\pixVBottom\!-\!\ppointV}{\focal}~} \nonumber \\
                &\approx \depthGT + \noise + \dfrac{\focal\egoHeightChange}{\pixVBottom\!-\!\ppointV} \nonumber \\
                \implies \newDepthGround - \depthGT &\approx \noise + \dfrac{\focal\egoHeightChange}{\pixVBottom\!-\!\ppointV} \nonumber,
            \end{align}
            assuming the ground depth $\oldDepthGround$ at train height $\egoHeightChange=0$ is the GT depth $\depthGT$ added by a normal random variable $\noise$ with mean $0$ and variance $\normalVar$ as in \cite{kumar2024seabird}.
            Taking expectation on both sides, the mean depth error is     
            \begin{align}
                \expect\Big(\newDepthGround - \depthGT\Big) &\approx 
                \left(\dfrac{1}{\pixVBottom\!-\!\ppointV}\right)\focal\egoHeightChange, \nonumber
            \end{align}
            confirming the positive \trend of the mean depth error of the ground model \wrt the height change $\egoHeightChange$.

        The ground lies between the bottom part of the image plane/ image height ($h$) and the optical center $y$-coordinate $\ppointV$, and so $\pixVBottom-\ppointV > 0$.
        However, in practice, it could get negative in early stage of training. 
        To enforce non-negativity of this term, we pass $\pixVBottom\!-\!\ppointV$ through a \relu non-linearity to enforce $\pixVBottom\!-\!\ppointV$ is positive.
        \cref{sec:charmer_ablation} confirms that \relu remains important for good results.
        \end{proof}

    \subsection{Regression-based Depth Model}
        Most \monoThreeD models rely on regression losses, to compare the predicted depth with the GT depth~\cite{kumar2022deviant, zhang2022beverse}.
        We, next, derive the extrapolation behavior of such regressed depth model in the following theorem.

        \begin{theorem}\label{theorem:2}
            \textbf{Regressed model has negative slope (trend) in extrapolation}.
            Consider a regressed depth model trained on data from single ego \variation, predicting depth $\depthPred$ from the projected \threeD center $(\pixUCenter,\pixVCenter)$.
            Assuming a linear relationship between predicted depth and pixel position, the mean depth error of a regressed model exhibits a negative \trend \wrt the \variation change $\egoHeightChange$:
            \begin{align}
                \expect\Big(\newDepthRegress - \depthGT\Big) &= -\left(\dfrac{\depthSlope}{\depthGT}\right)\focal\egoHeightChange,
                \label{eqn:regress_depth_bias}
            \end{align}
            where $\depthSlope$ is a camera \variation independent positive constant.
        \end{theorem}

        \begin{figure}[!t]
            \centering
            \includegraphics[width=0.7\linewidth]{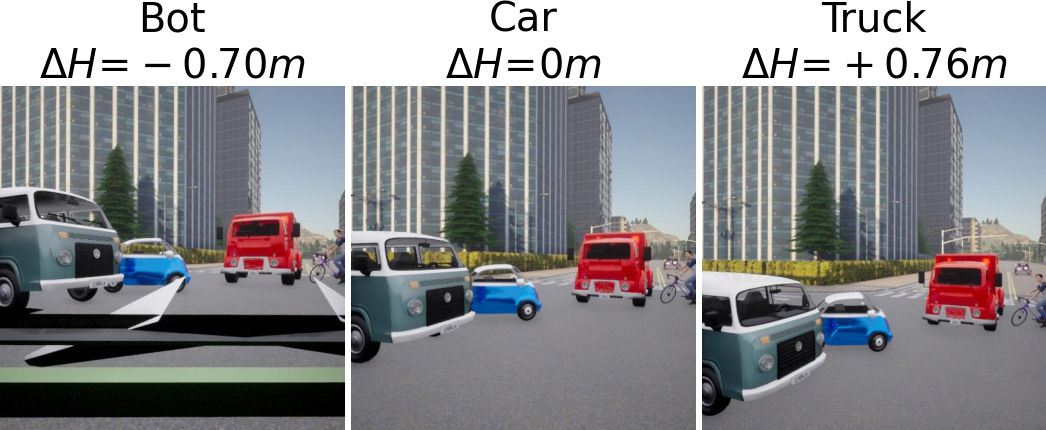}
            \includegraphics[width=0.7\linewidth]{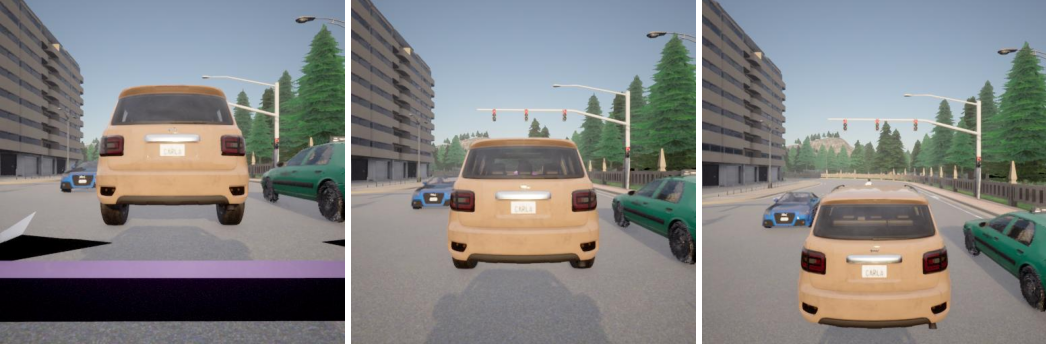}
            \caption[\carla \val samples with both negative and positive ego \variation changes ($\egoHeightChange$) covers AVs from bots to cars to trucks.]
            {\textbf{\carla \val samples} with both negative and positive ego \variation changes ($\egoHeightChange$) covers AVs from bots to cars to trucks.}
            \label{fig:charmer_carla_sample}
        \end{figure}

        \cref{theorem:2} says that regressed depth model under-estimates and over-estimates depth as the ego \variation change $\egoHeightChange$ increases and decreases respectively.

        \begin{table*}[!t]
            \caption[Error analysis of \gupNet trained on $\egoHeightChange\!=\!0m$]
            {\textbf{Error analysis} of \gupNet \cite{lu2021geometry} trained on $\egoHeightChange\!=\!0m$ on all height changes $\egoHeightChange$ of \carla \val split. 
            Depth remains the biggest source of error in inference on {unseen ego \variations}.
            }
            \label{tab:error_analysis}
            \centering
            \scalebox{0.8}{
            \setlength\tabcolsep{0.23cm}
            \begin{tabular}{ccccccc m dcd m dcd m dcdc }
                \addlinespace[0.01cm]
                \multicolumn{7}{cm}{Oracle Params. \downarrowRHDSmall~~/ $\egoHeightChange~(m)\rightarrowRHDSmall$ } &  \multicolumn{3}{cm}{\apThreeDSeventy~\bracketPercentage~(\uparrowRHDSmall)} & \multicolumn{3}{cm}{\apThreeDFifty~\bracketPercentage~(\uparrowRHDSmall)} & \multicolumn{3}{c}{\MDE $(m)~[\approx 0]$}\\
                $x$ & $y$ & $z$ & $l$ & $w$ & $h$ & $\theta$ & $-0.70$ & $0$ & $+0.76$ & $-0.70$ & $0$ & $+0.76$ & $-0.70$ & $0$ & $+0.76$ \\
                \myTopRule
                & & & & & & & $9.46$ & $53.82$ & $7.23$ & $41.66$ & $76.47$ & $40.97$ & $+0.53$ & $+0.03$ & $-0.63$\\
                \cmark & & & & & & & $15.95$ & $62.21$ & $12.74$ & $46.89$ & $76.78$ & $50.97$ & $+0.53$ & $+0.03$ & $-0.63$\\
                & \cmark  && & & & & $13.56$ & $59.55$ & $10.67$ & $44.93$ & $76.86$ & $49.84$ & $+0.53$ & $+0.03$ & $-0.63$\\
                &&  \cmark & & & & & $34.82$ & $69.99$ & $39.03$ & $68.10$ & $82.73$ & $76.24$ & $+0.00$ & $+0.00$ & $+0.00$\\
                \cmark & \cmark & \cmark & & & & & $65.44$ & $82.36$ & $80.70$ & $74.76$ & $84.93$ & $82.11$ & $+0.00$ & $+0.00$ & $+0.00$\\
                & & & \cmark & \cmark & \cmark & & $10.32$ & $56.24$ & $7.20$ & $42.04$ & $76.61$ & $42.03$ & $+0.53$ & $+0.03$ & $-0.63$\\
                \cmark & \cmark & \cmark & \cmark & \cmark & \cmark & & $75.86$ & $82.82$ & $82.08$ & $78.21$ & $85.17$ & $82.24$ & $+0.00$ & $+0.00$ & $+0.00$\\
                \cmark & \cmark & \cmark & \cmark & \cmark & \cmark & \cmark & $78.44$ & $85.20$ & $82.28$ & $78.44$ & $85.20$ & $82.28$ & $+0.00$ & $+0.00$ & $+0.00$\\
            \end{tabular}
            }
        \end{table*} 

        \begin{proof}
            Neural nets often use the $y$-coordinate of their projected \threeD center $\pixVCenter$ to predict depth \cite{dijk2019neural}.
            Consider a simple linear regression model for predicting depth. 
            Then, the regressed depth $\oldDepthRegress$ is
            \begin{align}
                \oldDepthRegress &= -\left(\dfrac{\depthGTMax\!-\!\depthGTMin}{\heightImage\!-\!\ppointV}\right)(\pixVCenter\!-\!\ppointV) + \depthGTMax \nonumber \\
                &= -\depthSlope(\pixVCenter\!-\!\ppointV) + \depthGTMax,
                \label{eq:regress_depth}
            \end{align}

            This linear regression model has a negative slope, with a positive slope parameter $\depthSlope$, and $\heightImage$ being the height of the image.
            This model predicts depth $\depthGTMin$ at pixel position $\pixVCenter\!=\!\heightImage$ and $\depthGTMax$ at principal point $\pixVCenter\!=\!\ppointV$. 
            When the ego camera shifts by $\egoHeightChange~m$, the projected center of the object becomes $\pixVCenter + \dfrac{\focal\egoHeightChange}{\depthGT}$. 
            Substituting this into the regression model of \cref{eq:regress_depth}, we obtain the new depth $\newDepthRegress$ as, 
            \begin{align}
                \newDepthRegress &= -\depthSlope\left(\pixVCenter\!+\! \dfrac{\focal\egoHeightChange}{\depthGT}\!-\!\ppointV\right) + \depthGTMax \nonumber \\
                &= -\depthSlope(\pixVCenter\!-\!\ppointV) + \depthGTMax -\left(\dfrac{\depthSlope}{\depthGT}\right)\focal\egoHeightChange \nonumber \\
                &= \oldDepthRegress -\left(\dfrac{\depthSlope}{\depthGT}\right)\focal\egoHeightChange \nonumber \\ 
                &= \depthGT +  \noise -\left(\dfrac{\depthSlope}{\depthGT}\right)\focal\egoHeightChange \nonumber \\
                \implies \newDepthRegress - \depthGT &= \noise -\left(\dfrac{\depthSlope}{\depthGT}\right)\focal\egoHeightChange \nonumber,
            \end{align}
            assuming the regressed depth $\oldDepthRegress$ at train height $\egoHeightChange=0$ is the GT depth $\depthGT$ added by a normal random variable $\noise$ with mean $0$ and variance $\normalVar$ as in \cite{kumar2024seabird}.
            Taking expectation on both sides, the mean depth error is 
            \begin{align}
                \expect\Big(\newDepthRegress - \depthGT\Big) &= -\left(\dfrac{\depthSlope}{\depthGT}\right)\focal\egoHeightChange, \nonumber
            \end{align}
            confirming the negative \trend of the mean depth error of the regressed depth model \wrt the height change $\egoHeightChange$.
        \end{proof}

    \subsection{Merging Depth Estimates.}\label{sec:depth_merge}

        \cref{theorem:1,theorem:2} prove that the ground and the regressed depth models show contrasting extrapolation behaviors.
        The former over-estimates the depth while the latter under-estimates depth as the ego \variation change $\egoHeightChange$ increases.
        \cref{fig:charmer_overview} shows how these two depth estimates are fused together. 
        Overall, \charmer leverages depth information from these two source sources (with different extrapolation behaviors) to improve the \monoThreeD generalization to unseen camera \variations. 
        \charmer starts with an input image, and estimates the depth of the object using two methods: ground and regressed depth. 
        \charmer outputs the projected bottom center of the object to query the ground depth (calculated from the ego camera parameters and its position and orientation relative to the ground plane as in \cref{lemma:charmer_1}).
        It also outputs another depth estimate based on regression. 
        The final step combines the two estimated depths with a simple average to cancel the opposing trends and obtain the refined depth estimates, resulting in a set of accurate and localized \threeD objects in the scene.

\section{Experiments}\label{sec:charmer_experiments}

    \noIndentHeading{Datasets.}
        Our experiments 
            utilize the simulated \carla dataset\footnote{The authors of \cite{tzofi2023towards} do not release their other Nvidia-Sim dataset.} from \cite{tzofi2023towards}, configured to mimic the \nuscenes~\cite{caesar2020nuscenes} dataset.
            We use this dataset for two reasons.
            First, this dataset reduces training and testing domain gaps, while existing public datasets lack data at multiple ego \variations.
            Second, recent paper \cite{tzofi2023towards} also use this dataset for their experiments. 
            The default \carla dataset sweeps camera height changes $\egoHeightChange$ from $0$ to $0.76m$, rendering a dataset every $0.076m$ (car to trucks). 
            To fully investigate the impact of camera height variations, we extend the original \carla dataset by introducing negative height changes. 
            The extended \carla dataset sweeps height changes $\egoHeightChange$ from $-0.70m$ to $0.76m$ with settings from bots to cars to trucks. 
            \cref{fig:charmer_carla_sample} illustrates sample images from this dataset.
            Note that we exclude $\egoHeightChange\!=\!-0.76m$ setting due to visibility obstructions caused by the ego vehicle's bonnet.
            

        \begin{table*}[!t]
            \caption[\carla \val Results.]
            {\textbf{\carla \val Results.}
            \charmer \textbf{outperforms} all other baselines, especially at bigger {unseen ego \variations}.
            All methods except Oracle are trained on car height $\egoHeightChange=0m$ and tested on bot to truck height data.
            [Key: \firstKey{Best}]
            }
            \centering
            \scalebox{0.8}{
            \setlength\tabcolsep{0.21cm}
            \begin{tabular}{l l m dcd m dcd m dcdc}
                \addlinespace[0.01cm]
                \multirow{2}{*}{\threeD Detector} & \multirow{2}{*}{Method $\downarrowRHDSmall$ / $\egoHeightChange~(m)\rightarrowRHDSmall$} & \multicolumn{3}{cm}{\apThreeDSeventy \bracketPercentage~(\uparrowRHDSmall)} & \multicolumn{3}{cm}{\apThreeDFifty \bracketPercentage~(\uparrowRHDSmall)} & \multicolumn{3}{c}{\MDE $(m)~[\approx 0]$}\\
                &  & $-0.70$ & $0$ & $+0.76$ & $-0.70$ & $0$ & $+0.76$ & $-0.70$ & $0$ & $+0.76$\\
                \myTopRule
                \multirow{6}{*}{\gupNet \cite{lu2021geometry}} 
                & Source  & $9.46$ & $53.82$ & $7.23$ & $41.66$ & $76.47$ & $40.97$ & $+0.53$ & $+0.03$ & $-0.63$\\
                & \plucker \cite{plucker1828analytisch} & $8.43$ & $55.56$ & $10.13$ & $37.10$ & \first{76.57} & $43.22$ & $+0.55$ & $+0.03$ & $-0.63$\\
                & \uniDrive \cite{li2024unidrive} & $10.73$ & $53.82$ & $5.54$ & $42.30$ & $76.46$ & $39.33$ & $+0.51$ & $+0.03$ & $-0.67$\\
                & \uniDrivePlus \cite{li2024unidrive} & $10.83$ & $53.82$ & $12.27$ & $47.81$ & $76.46$ & $53.08$ & $+0.39$ & $+0.03$ & $-0.48$\\
                & \cellcolor{methodColor}\textbf{\charmer} & \cellcolor{methodColor}\first{19.45} & \cellcolor{methodColor}\first{55.68} &  \cellcolor{methodColor}\first{27.33} & \cellcolor{methodColor}\first{53.40} & \cellcolor{methodColor}${74.47}$ & \cellcolor{methodColor}\first{61.98} & \cellcolor{methodColor}$+0.07$ & \cellcolor{methodColor}$+0.05$ & \cellcolor{methodColor}$-0.02$ \\
                \hhline{|~|-----------|}
                & Oracle & $70.96$ & $53.82$ & $62.25$ & $83.88$ & $76.47$ & $83.96$ & $+0.03$ & $+0.03$ & $+0.03$\\
                \myTopRule
                \multirow{6}{*}{\deviant \cite{kumar2022deviant}} 
                & Source  & $8.63$ & $50.18$ & $6.25$ & $40.24$ & $73.78$ & $41.74$ & $+0.46$ & $+0.01$ & $-0.65$ \\
                & \plucker \cite{plucker1828analytisch} & $8.43$ & \first{51.32} & $9.52$ & $38.24$ & \first{73.91} & $44.22$ & $+0.46$ & $+0.01$ & $-0.64$ \\
                & \uniDrive \cite{li2024unidrive} & $8.33$ & $50.18$ & $6.56$ & $41.40$ & $73.78$ & $41.27$ & $+0.46$ & $+0.01$ & $-0.64$ \\
                & \uniDrivePlus \cite{li2024unidrive} & $6.73$ & $50.18$ & $12.03$ & $42.91$ & $73.78$ & $52.36$ & $+0.37$ & $+0.01$ & $-0.47$ \\
                & \cellcolor{methodColor}\textbf{\charmer} & \cellcolor{methodColor}\first{17.11} & \cellcolor{methodColor}$48.74$ & \cellcolor{methodColor}\first{26.24} & \cellcolor{methodColor}\first{49.28} & \cellcolor{methodColor}${70.21}$ & \cellcolor{methodColor}\first{63.60} & \cellcolor{methodColor}$+0.01$ & \cellcolor{methodColor}$+0.03$ & \cellcolor{methodColor}$-0.02$  \\
                \hhline{|~|-----------|}
                & Oracle & $71.97$ & $50.18$ & $62.56$ & $84.56$ & $73.78$ & $83.94$ & $+0.03$ & $+0.01$ & $-0.02$ \\
            \end{tabular}
            }
            \label{tab:det_results_carla_val}
        \end{table*}
        \begin{figure*}[!t]
            \centering
            \begin{subfigure}{.3\linewidth}
                \includegraphics[width=\linewidth]{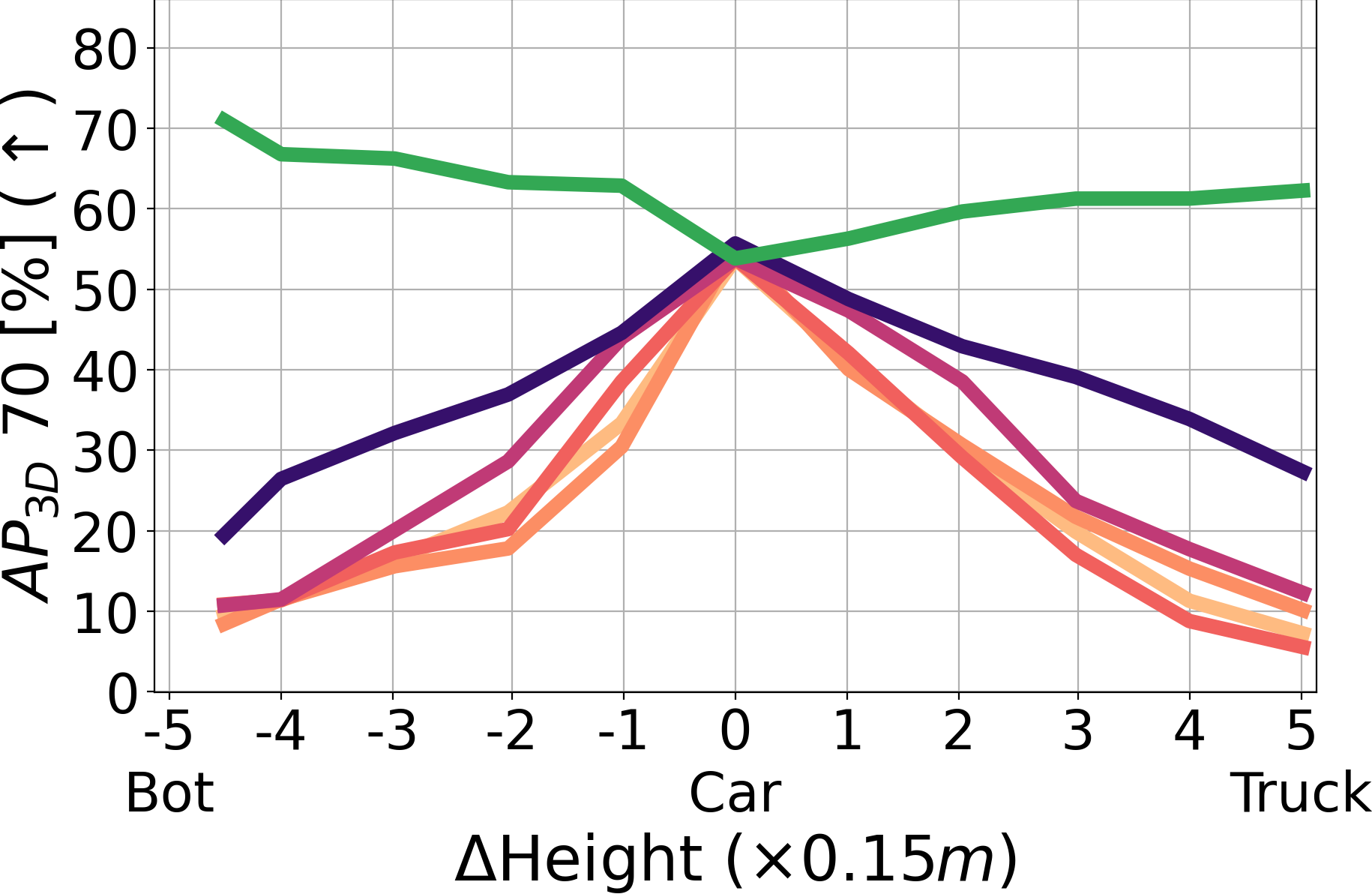}
                \caption{\apThreeDSeventy \bracketPercentage{} comparison.}
            \end{subfigure}%
            \hfill
            \begin{subfigure}{.3\linewidth}
                \includegraphics[width=\linewidth]{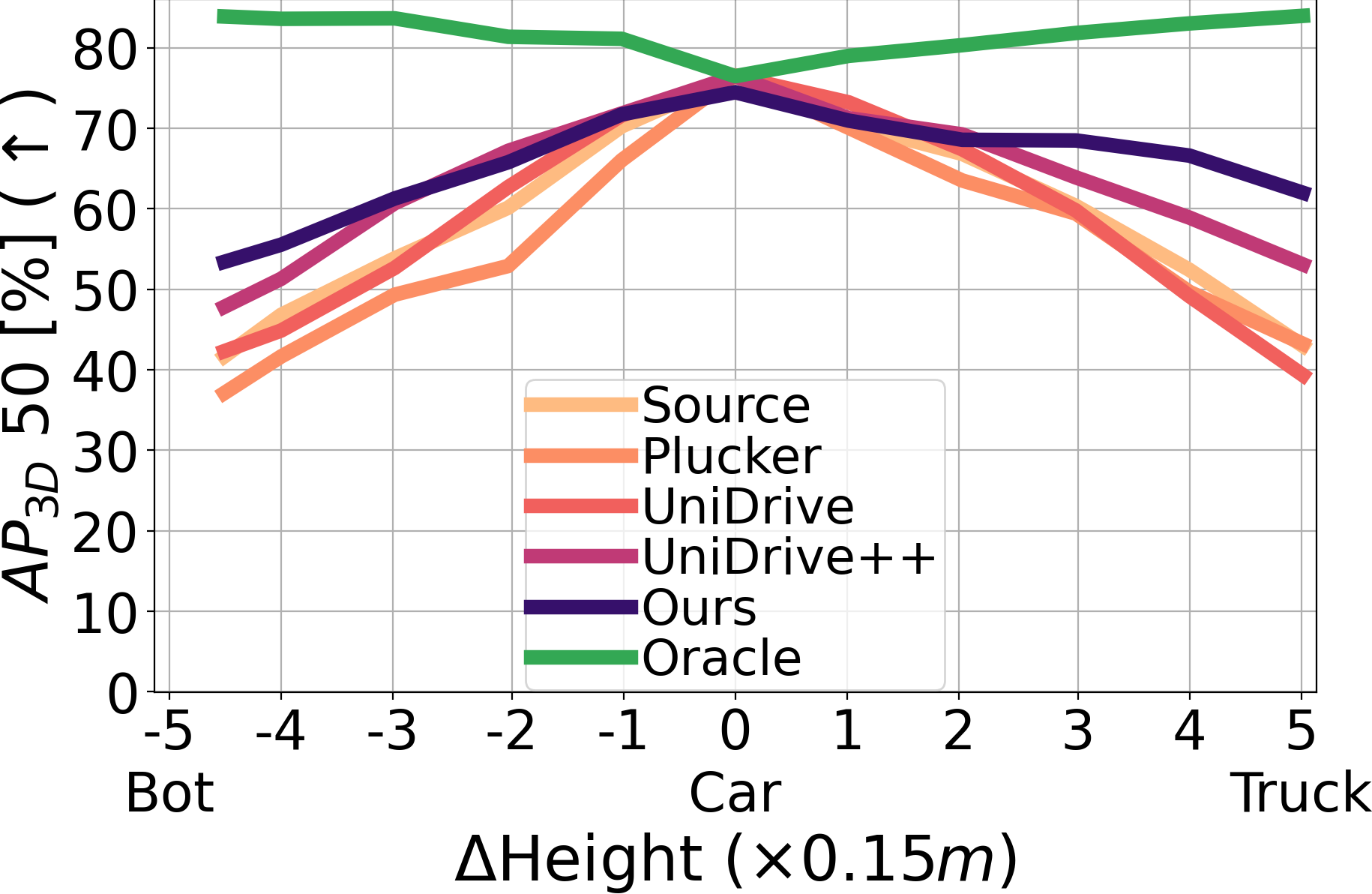}
                \caption{\apThreeDFifty \bracketPercentage{} comparison.}
            \end{subfigure}
            \hfill
            \begin{subfigure}{.3\linewidth}
                \includegraphics[width=\linewidth]{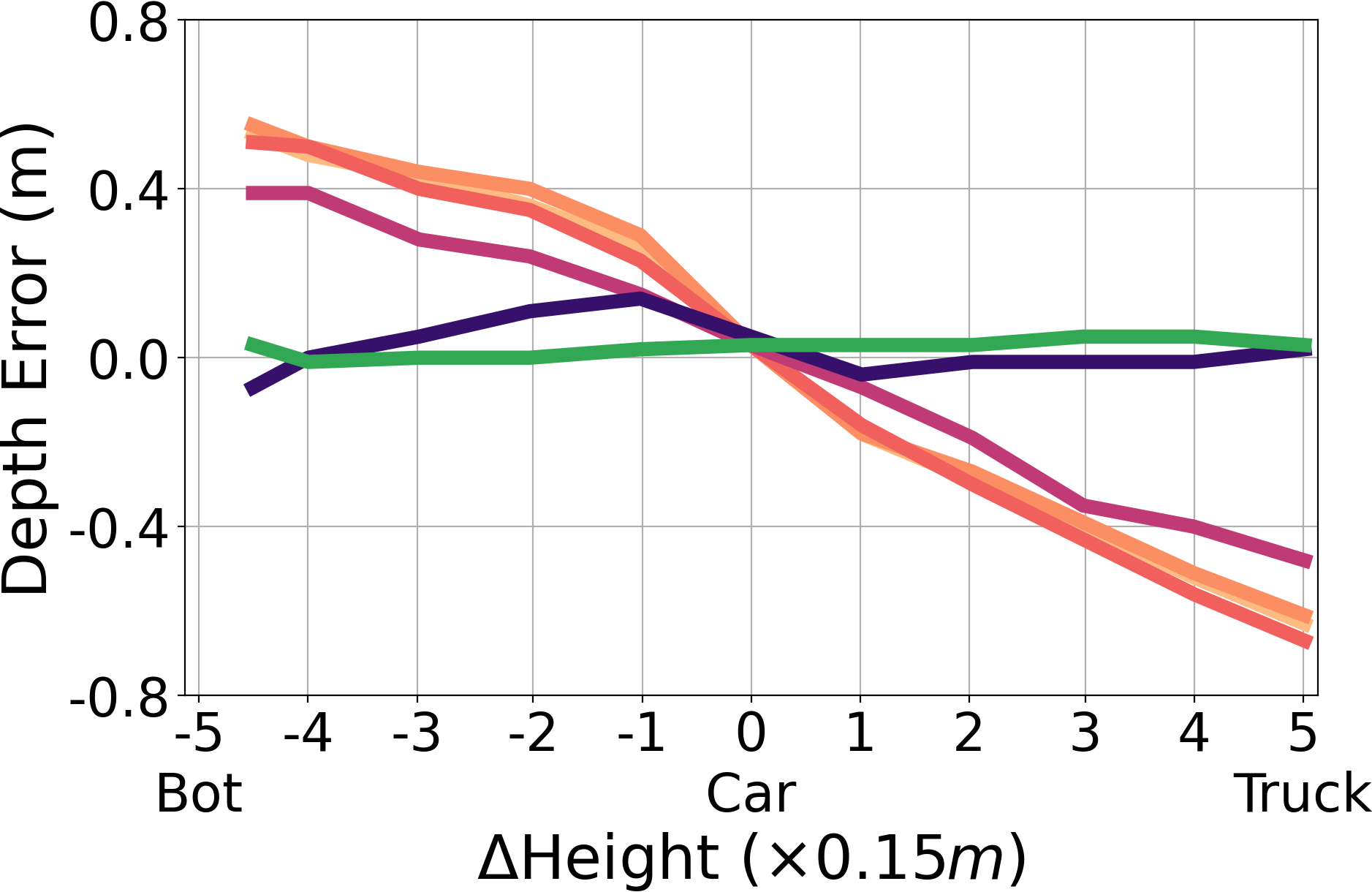}
                \caption{\MDE comparison.}
                \label{fig:charmer_mde_results_carla_val_gup}
            \end{subfigure}
            \caption[\carla \val Results on \gupNet.]
            {\textbf{\carla \val Results on \gupNet.}  
            \charmer \textbf{outperforms} all baselines, especially at bigger {unseen ego \variations}.
            All methods except Oracle are trained on car height and tested on all \variations.
            Results of inference on height changes of $-0.70,0$ and $0.76$ meters are in \cref{tab:det_results_carla_val}.
            See \cref{fig:charmer_det_results_carla_val} in the supplementary for another detector.
            }
            \label{fig:charmer_det_results_carla_val}
        \end{figure*}  

    \noIndentHeading{Data Splits.}
        Our experiments use the 
            \textit{\carla \val Split.} 
            This dataset split \cite{tzofi2023towards} contains $25{,}000$ images ($2{,}500$ scenes) from \textit{town03} map for training and $5{,}000$ images ($500$ scenes) from \textit{town05} map for inference on multiple ego \variation. Except for Oracle, we train all models on training images from the car height $(\egoHeightChange=0m)$.

    \noIndentHeading{Evaluation Metrics.} 
        We choose the \kitti \apThreeDSeventy percentage on the Moderate category \cite{geiger2012we} as our evaluation metric.
        We also report \apThreeDFifty percentage numbers following prior works \cite{brazil2020kinematic,kumar2021groomed}.
        Additionally, we report the mean depth error (\MDE) over predicted boxes with \iouTwoD overlap greater than $0.7$ with the GT boxes similar to \cite{kumar2022deviant}.
        Note that \MDE is different from MAE metric of \cite{kumar2022deviant} that it does not take absolute value.

    \noIndentHeading{Detectors.}
        We use the \gupNet \cite{lu2021geometry} and \deviant \cite{kumar2022deviant} as our base detectors. 
        The choice of these models encompasses CNN \cite{lu2021geometry} and group CNN-based \cite{kumar2022deviant} 
        architectures. 

    \noIndentHeading{Baselines.}
        We compare against the following baselines:
        \begin{itemize}
            \item \textit{Source}: This is the \monoThreeD model trained on the car height ($\egoHeightChange=0m$) data.
            \item \textit{\plucker Embeddings} \cite{plucker1828analytisch,teller1999determining}: Training a \monoThreeD model with \plucker embeddings to improve robustness as in \threeD pose estimation and reconstruction tasks. \plucker embeddings generalize the intrinsic-focused \camConvs \cite{facil2019camera} embeddings to camera extrinsics.
            \item \textit{\uniDrive} \cite{li2024unidrive}: Transforming unseen ego \variation (target) images to car height (source) assuming objects at fixed distance parameter ($50m$) and then passing to the \monoThreeD model.
            \item \textit{\uniDrivePlus} \cite{li2024unidrive}: \uniDrive with distance parameter optimized per dataset. 
        \item \textit{Oracle}: We also report the \textit{Oracle} \monoThreeD model, which is trained and tested on the \textbf{same} ego \variation. 
        The Oracle serves as the \textbf{upper bound} of all baselines.
        \end{itemize}

    \subsection{\carla Error Analysis}
        We first report the error analysis of the baseline \gupNet \cite{lu2021geometry} in \cref{tab:error_analysis} by replacing the predicted box data with the oracle parameters of the box as in \cite{ma2021delving,kumar2024seabird}.
        We consider the GT box to be an oracle box for predicted box if the euclidean distance is less than $4m$ \cite{kumar2024seabird}. 
        In case of multiple GT being matched to one box, we consider the oracle with the minimum distance.
        \cref{tab:error_analysis} shows that depth is the biggest source of error for \monoThreeD task under ego \variation changes as also observed for single height settings in \cite{ma2021delving,kumar2022deviant,kumar2024seabird}. 
        Note that the Oracle does not get $100\%$ results since we only replace box parameters in the baseline and consequently, the missed boxes in the baseline are not added.

        \begin{table*}[!t]
            \caption[\carla \val Results with \resNetEighteen backbone.]
            {\textbf{\carla \val Results with \resNetEighteen backbone}.
            \charmer \textbf{outperforms} all baselines, especially at bigger {unseen ego \variations}.
            All methods except Oracle are trained on car height $\egoHeightChange=0m$ and tested on bot to truck height data.
            [Key: \firstKey{Best}]
            }
            \label{tab:det_results_carla_val_resnet}
            \centering
            \scalebox{0.78}{
            \setlength\tabcolsep{0.23cm}
            \begin{tabular}{l l m dcd m dcd m dcd }
                \addlinespace[0.01cm]
                    \multirow{2}{*}{\threeD Detector} & \multirow{2}{*}{Method $\downarrowRHDSmall$ / $\egoHeightChange~(m)\rightarrowRHDSmall$} &  \multicolumn{3}{cm}{\apThreeDSeventy~\bracketPercentage~(\uparrowRHDSmall)} & \multicolumn{3}{cm}{\apThreeDFifty~\bracketPercentage~(\uparrowRHDSmall)} & \multicolumn{3}{c}{\MDE $(m)~[\approx 0]$}\\
                & & $-0.70$ & $0$ & $+0.76$ & $-0.70$ & $0$ & $+0.76$ & $-0.70$ & $0$ & $+0.76$\\
                \myTopRule
                \multirow{5}{*}{\gupNet \cite{lu2021geometry}} & Source & $10.13$ & \first{49.82} & $5.28$ & $47.15$  & \first{73.49} & $42.70$ & $+0.40$ & $+0.01$ & $-0.65$\\
                & \uniDrive \cite{li2024unidrive} & $10.05$ & \first{49.82} & $6.15$ & $47.15$  & \first{73.49} & $43.89$ & $+0.35$ & $+0.01$ & $-0.62$\\
                & \uniDrivePlus \cite{li2024unidrive} & $9.37$ & \first{49.82} & $13.00$ & $52.95$  & \first{73.49} & $55.57$ & $+0.31$ & $+0.01$ & $-0.46$\\
                & \cellcolor{methodColor}\textbf{\charmer} & \cellcolor{methodColor}\first{16.62} & \cellcolor{methodColor}$46.13$ & \cellcolor{methodColor}\first{24.50} & \cellcolor{methodColor}\first{57.00} & \cellcolor{methodColor}$67.83$ & \cellcolor{methodColor}\first{60.86} & \cellcolor{methodColor}$-0.15$ & \cellcolor{methodColor}$+0.00$ & \cellcolor{methodColor}$+0.07$ \\
                & Oracle & $70.25$ & $49.82$ & $62.93$  & $83.49$ & $73.49$ & $84.07$ & $-0.01$ & $+0.05$ & $+0.07$ \\
                \hline
                \multirow{5}{*}{\deviant \cite{kumar2022deviant}} & Source & $8.83$ & \first{49.88} & $4.43$ & $42.10$  & $72.79$ & $38.42$ & $+0.40$ & $+0.01$ & $-0.69$\\
                & \uniDrive \cite{li2024unidrive} & $8.21$ & $49.87$ & $3.75$ & $42.21$  & $72.79$ & $38.38$ & $+0.40$ & $+0.01$ & $-0.70$\\
                & \uniDrivePlus\cite{li2024unidrive} & $6.01$ & $49.87$ & $12.03$ & $43.99$  & $72.79$ & $50.67$ & $+0.38$ & $+0.01$ & $-0.50$\\
                & \cellcolor{methodColor}\textbf{\charmer} & \cellcolor{methodColor}\first{14.96} & \cellcolor{methodColor}$49.13$ & \cellcolor{methodColor}\first{23.66} & \cellcolor{methodColor}\first{52.68} & \cellcolor{methodColor}\first{72.95} & \cellcolor{methodColor}\first{60.98} & \cellcolor{methodColor}$-0.07$ & \cellcolor{methodColor}$+0.05$ & \cellcolor{methodColor}$+0.02$ \\
                & Oracle & $68.35$ & $49.88$ & $58.49$ & $84.03$  & $72.79$ & $83.42$ & $-0.04$ & $+0.01$ & $-0.1$\\
            \end{tabular}
            }
        \end{table*}

        \begin{table}[!t]
            \caption[Ablation Studies of \gupNet + \charmer on the \carla \val split on unseen ego \variations.]
            {\textbf{Ablation Studies} of \gupNet + \charmer on the \carla \val split on {unseen ego \variations}.
            [Key: \firstKey{Best}]
            }
            \label{tab:charmer_ablation}
            \centering
            \scalebox{0.78}{
            \setlength\tabcolsep{0.2cm}
            \begin{tabular}{l l m dcd m dcd m dcd }
                \addlinespace[0.01cm]
                    \multirow{2}{*}{Change} & \multirow{2}{*}{From \rightarrowRHDSmall To ~~~~/ $\egoHeightChange~(m)\rightarrowRHDSmall$} &  \multicolumn{3}{cm}{\apThreeDSeventy~\bracketPercentage~(\uparrowRHDSmall)} & \multicolumn{3}{cm}{\apThreeDFifty~\bracketPercentage~(\uparrowRHDSmall)} & \multicolumn{3}{c}{\MDE $(m)~[\approx 0]$}\\
                & & $-0.70$ & $0$ & $+0.76$ & $-0.70$ & $0$ & $+0.76$ & $-0.70$ & $0$ & $+0.76$\\
                \myTopRule
                \gupNet \cite{lu2021geometry} & \mathDash & $9.46$ & $53.82$ & $7.23$ & $41.66$  & $76.47$ & $40.97$ & $+0.53$ & $+0.05$ & $-0.63$\\
                \hline
                \multirow{4}{*}{Merge} & Regress+Ground \rightarrowRHDSmall~Regress & $9.46$ & $53.82$ & $7.23$ & $41.66$  & $76.47$ & $40.97$ & $+0.53$ & $+0.05$ & $-0.63$ \\
                & Regress+Ground \rightarrowRHDSmall~Ground & $0.98$ & $26.61$ & $5.39$ & $14.21$  & $51.97$ & $31.42$ & $-0.80$ & $-0.01$ & $+0.55$ \\
                & Within Model \rightarrowRHDSmall~Offline & $12.86$ & $47.66$ & $18.36$ & $49.86$  & $76.30$ & $54.38$ & $+0.24$ & $+0.02$ & $-0.28$ \\ 
                & Simple Avg \rightarrowRHDSmall~Learned Avg & $8.25$ & \first{56.49} & $9.53$ & $38.58$ & \first{76.82} & $43.13$ & $+0.56$ & $-0.03$ & $-0.62$\\
                \hline
                Ground & \relu \rightarrowRHDSmall~No \relu & $0.60$ & $52.94$ & $0.07$ & $15.66$  & $74.79$ & $4.50$ & $-1.09$ & $-0.01$ & $+1.34$\\
                \hline 
                Formulation & Product \rightarrowRHDSmall~Sum & $3.28$ & $37.22$ & $12.79$ & $17.28$  & $63.88$ & $47.09$ & $+0.56$ & $+0.09$ & $-0.22$\\
                \hline
                \cellcolor{methodColor}\textbf{\charmer} & \cellcolor{methodColor}\mathDash & \cellcolor{methodColor}\first{19.45} & \cellcolor{methodColor}$55.68$ & \cellcolor{methodColor}\first{27.33} & \cellcolor{methodColor}\first{53.40} & \cellcolor{methodColor}$74.47$ & \cellcolor{methodColor}\first{61.98} & \cellcolor{methodColor}$-0.07$ & \cellcolor{methodColor}$+0.05$ & \cellcolor{methodColor}$+0.02$ \\
                \hline
                Oracle & \mathDash & $70.96$ & $53.82$ & $62.25$ & $83.88$ & $76.47$ & $83.96$ & $+0.03$ & $+0.03$ & $+0.03$\\
            \end{tabular}
            }
        \end{table}

    \subsection{\carla \Variation Robustness Results}
        \cref{tab:det_results_carla_val} presents the \carla \val results, reporting the \textbf{median} model over three different seeds with the model being the final checkpoint as \cite{kumar2022deviant}.
        It compares baselines and our \charmer on all \monoThreeD models - \gupNet \cite{lu2021geometry}, and \deviant \cite{kumar2022deviant}. 
        Except for Oracle, all models are trained from car height data and tested on all ego heights.
        \cref{tab:det_results_carla_val} confirms that \charmer outperforms other baselines on all the  \monoThreeD models, and results in a better height robust detector. 
        We also plot these \apThreeD numbers and depth errors visually in \cref{fig:charmer_det_results_carla_val} for intermediate height changes to confirm our observations.
        The \MDE comparison in \cref{fig:charmer_mde_results_carla_val_gup} also shows the trend of baselines, while \charmer cancels the opposite trends in extrapolation.

        \noIndentHeading{Oracle Biases.}
        We further note biases in the Oracle models at big changes in ego \variation. 
        This agrees the observations of \cite{tzofi2023towards} in the \bev segmentation task. 
        While higher \apThreeD for a higher \variation could be explained by fewer occlusions due a higher \variation, higher \apThreeD at lower camera \variation is not explained by this hypothesis.
        We leave the analysis of higher Oracle numbers for a future work.

        \noIndentHeading{Results on Other Backbone.}
            We next investigate whether the extrapolation behavior holds for other backbones as well following \deviant \cite{kumar2022deviant}. 
            So, we benchmark on the \resNetEighteen backbone. 
            \cref{tab:det_results_carla_val_resnet} results show that extrapolation shows up in other backbones and \charmer again outperforms all baselines. 
            The biggest gains are in big camera height changes, which is consistent with \cref{tab:det_results_carla_val} results.

    \subsection{Ablation Studies}\label{sec:charmer_ablation}
        \cref{tab:charmer_ablation} ablates the design choices of \gupNet + \charmer on  \carla \val split, with the experimental setup of \cref{sec:charmer_experiments}.

        \noIndentHeading{Depth Merge.}
            We first analyze the impact of averaging the two depth estimates. 
            Merging both regressed and ground-based depth estimates is crucial for optimal performance. 
            Relying solely on the regressed depth gives good \inDomain performance but bad \ood performance. 
            Using only ground depth generalizes poorly in both \inDomain and \ood settings, which is why it is not used in modern \monoThreeD models.
            However, it has a contrasting extrapolation \MDE compared to regression models.
            While offline merging of depth estimates from regression-only and ground-only models also improves extrapolation, it is slower and lacks end-to-end training.
            We also experiment with changing the simple averaging of \charmer to learned averaging. 
            Simple average of \charmer outperforms learned one in \outDomain test because the learned average overfits to train distribution.

        \noIndentHeading{\relu{}ed Ground.}
            \cref{sec:charmer_bottom_ground_depth} says that \relu activation applied to the ground depth ensures spatial continuity and improves model training stability. 
            Removing the \relu leads to training instability and suboptimal extrapolation to camera \variation. (The training also collapses in some cases).

        \noIndentHeading{Formulation.}
            \charmer estimates the projected \threeD bottom center by using the projected \threeD center and the \twoD height prediction. 
            \cref{eq:bottom_center} predicts a coefficient $\alpha$ to determine the precise bottom center location. 
            Product means predicting $\alpha$ and then multiplying by $(\pixVCenter\!-\!\pixVCenterTwoD)$ to obtain the shift, while sum means directly predicting the shift $\alpha$.
            Replace this product formulation by the sum formulation of $\alpha$ confirms that 
            the product is more effective than the sum.

\section{Conclusions}

    This chapter highlights the understudied problem of \monoThreeD generalization to unseen ego \variations.
    We first systematically analyze the impact of camera height variations on state-of-the-art \monoThreeD models, identifying depth estimation as the primary factor affecting performance. 
    We mathematically prove and also empirically observe consistent negative and positive \trends in regressed and ground-based object depth estimates, respectively, under camera height changes. 
    This chapter then takes a step towards generalization to unseen camera heights and proposes \charmer. 
    \charmer averages both depth estimates within the model to mitigate these opposing \trends.
    \charmer significantly enhances the generalization of \monoThreeD models to unseen camera heights, achieving \sota performance on the \carla dataset.
    We hope that this initial step towards generalization will contribute to safer AVs.
    Future work involves extending this idea to more \monoThreeD models.

    \noIndentHeading{Limitation.} \charmer does not fully solve the generalization issue to unseen camera heights.

%% file: chapters/future.tex
\chapter{
    Conclusions and Future Research
}
\label{chpt:future}

In this thesis, we attempt generalizing \monoThreeD networks to occlusion, dataset, object sizes and camera heights. 
The backbones of our models is in all cases a convolutional neural network or a transformer backbone.
While the current \monoThreeD networks generalize fairly well across these shifts, they still suffer from the following issues:
\begin{itemize}
    \item They do not generalize to unseen datasets during training.
    \item They do not multiple handle tasks like depth prediction, semantic scene completion and \monoThreeD.
    \item They do not generalize to unknown or noisy camera extrinsics.
    \item They do not handle multiple camera models.
\end{itemize}

\noIndentHeading{Generalizing to Unseen Datasets.}
    Current multi-dataset trained baselines such as \cubeRCNN \cite{brazil2023omni3d} generalize poorly to datasets unseen in training. 
    In other words, these models do not generalize in cross-dataset settings. 
    Generalizing \monoThreeD to unseen datasets remains unsolved till date.
    We conjecture that the cause of limited generalization is the limited training data and specialized backbones which handle projective geometry.

\noIndentHeading{Generalizing to Multiple Tasks.}
    Tasks like metric depth prediction, semantic scene completion and \monoThreeD all represent \threeD scene understanding at varying levels of granularity from points to voxels to objects. 
    While there are networks that specialize in doing each task, a single model understands all these granularities as well as intermediate granularities remains an exciting direction for solidyfying \threeD understanding and task generalization.

\noIndentHeading{Generalizing to Unknown Extrinsics.}
    Current \monoThreeD methods work well when trained and tested on the same extrinsics.
    However, such methods do not work well when the camera extrinsics, is unknown during testing.
    Joint \monoThreeD and camera calibration remains an open problem.

\noIndentHeading{Generalizing to Camera Models.}
    Current methods handle only pinhole cameras, while the cameras available today also include fisheye and 360 camera models.
    Generalizing \monoThreeD networks to handle any camera model remains another open problem in this area.

Advances in \monoThreeD task enable diverse applications such as Autonomous Driving, Metaverse and robotics.
The goal of home robots is to assist humans in indoor activities, such as cooking or cleaning.
Future works which generalize \monoThreeD along these directions will make our limited \threeD scene understanding even more powerful.

%% file: appendices/publications.tex
\chapter{Publications}\label{chpt:appendix}

\noIndentHeading{First-Author Publications.}

\noindent A list of all first-authored peer-reviewed publications during the \phd program listed in reverse chronological order.

\begin{itemize} 
    \item \myself, Yuliang Guo, Zhihao Zhang, Xinyu Huang, Liu Ren and Xiaoming Liu. ``\charmer: Towards \charmerFull", ICCV, 2025 (under review).

    \item \myself, Yuliang Guo, Xinyu Huang, Liu Ren and Xiaoming Liu. ``\seabird: \seabirdFull with Dice Loss Improves 3D Detection of Large Objects", CVPR, 2024.  

    \item \myself, Garrick Brazil, Enrique Corona, Armin Parchami and Xiaoming Liu. ``\deviant: Depth Equivariant Network for Monocular 3D Object Detection", ECCV, 2022.

    \item \myself, Garrick Brazil and Xiaoming Liu. ``\groomedNMS: Grouped Mathematically Differentiable NMS for Monocular 3D Object Detection", CVPR, 2021.

    \item \myself{}$^*$, Tim Marks$^*$, Wenxuan Mou$^*$, Ye Wang, Michael Jones, Anoop Cherian, Toshi Koike-Akino, Xiaoming Liu and Chen Feng. ``LUVLi Face Alignment: Estimating Location, Uncertainty and Visibility Likelihood", CVPR, 2020.
\end{itemize}

\noIndentHeading{Other Publications.}

\begin{itemize}
    \item Yunfei Long, \myself, Xiaoming Liu and Daniel Morris. ``RICCARDO: Radar Hit Prediction and Convolution for Camera-Radar \threeD Object Detection", CVPR, 2025.

    \item Yuliang Guo, \myself,  Chen Zhao, Ruoyu Wang, Xinyu Huang, and Liu Ren. ``SUP-NeRF: A Streamlined Unification of Pose Estimation and NeRF for Monocular \threeD Object Reconstruction", ECCV, 2024.

    \item Shengjie Zhu, Girish Ganesan, \myself and Xiaoming Liu. ``RePLAy: Remove Projective LiDAR Depthmap Artifacts via Exploiting Epipolar Geometry", ECCV, 2024.

    \item Shengjie Zhu, \myself,  Masa Hu and Xiaoming Liu. ``Tame a Wild Camera: In-the-Wild Monocular Camera Calibration", NeurIPS, 2023.

    \item Vishal Asnani, \myself, Sua You and Xiaoming Liu. ``PrObeD: Proactive \twoD Object Detection Wrapper", NeurIPS, 2023.

    \item Garrick Brazil, \myself, Julian Straub, Nikhila Ravi, Justin Johnson and Georgia Gkioxari, ``Omni3D: A Large Benchmark and Model for \threeD Object Detection in the Wild", CVPR, 2023.

    \item Yunfei Long, \myself, Daniel Morris, Xiaoming Liu, Marcos Castro and Punarjay Chakravarty. ``RADIANT: \radar Image Association NeTwork for 3D Object Detection", AAAI, 2023.

    \item Thiago Serra, Xin Yu, \myself, and Srikumar Ramalingam. ``Scaling Up Exact Neural Network Compression by ReLU Stability", NeurIPS, 2021.

    \item \myself{}$^*$, Tim Marks$^*$, Wenxuan Mou$^*$, Chen Feng and Xiaoming Liu. ``UGLLI Face Alignment: Estimating Uncertainty with Gaussian Log-Likelihood Loss". ICCV Workshops, 2019.
\end{itemize}

%% file: appendices/groomed_appendix.tex
\chapter{\groomedNMS Appendix}\label{chpt:groomed_appendix}


\begin{table*}[t]
        \caption[Results on using Oracle NMS scores on \kitti \valOne cars detection]{\textbf{Results on using Oracle NMS scores on \kitti \valOne cars detection}. [Key: \bestKey{Best}]}
        \label{tab:oracle_nms}
        \centering
        \setlength\tabcolsep{0.1cm}
        \begin{tabular}{tl m ccc  m ccc m ccct}
            \myTopRule
            \multirow{2}{*}{NMS Scores} & \multicolumn{3}{cm}{\apThreeDForty ($\uparrowRHDSmall$)} & \multicolumn{3}{cm}{\apBevForty ($\uparrowRHDSmall$)} & \multicolumn{3}{ct}{\apTwoDForty ($\uparrowRHDSmall$)}\\ 
            & Easy & Mod & Hard & Easy & Mod & Hard & Easy & Mod & Hard\\ 
            \myTopRule
            \kinematicImage & $18.29$ & $13.55$ & $10.13$ & $25.72$ & $18.82$ & $14.48$ & $93.69$ & $84.07$ & $67.14$ \\
            Oracle \iouTwoD        & $9.36$ & $9.93$ & $6.40$  & $12.27$ & $10.43$ & $8.72$ & \best{99.18} & \best{95.66} & \best{85.77} \\
            Oracle \iouThreeD      & \best{87.93} & \best{73.10} & \best{60.91} & \best{93.47} & \best{83.61} & \best{71.31} & $80.99$ & $78.38$ & $67.66$\\
            \myTopRule
        \end{tabular}
    \end{table*}

\section{Detailed Explanation of NMS as a Matrix Operation}\label{sec:NMS_explanation}
    The rescoring process of the~\classicalNms~is greedy set-based~\cite{prokudin2017learning} and calculates the rescore for a box $i$ (Line $10$ of Alg.~\ref{alg:classical}) as 
    \begin{align}
        r_i &= s_i  \prod\limits_{j\in\vindex_{< i}}\left(1 -\pruneof{\overlap_{ij}}\right),
        \label{eq:classical_all}
    \end{align}
    where $\vindex_{< i}$ is defined as the box indices sampled from $\vindex$ having higher scores than box $i$. For example, let us consider that $\vindex=\{1,5, 7, 9\}$. Then, for $i=7,~ \vindex_{< i} = \{1,5\}$  while for $i=1, \vindex_{< i} = \phi$ with $\phi$ denoting the empty set.
    This is possible since we had sorted the scores $\score$ and $\overlapMat$ in decreasing order (Lines $2$-$3$ of Alg.~\ref{alg:diff_nms}) to remove the non-differentiable hard $\argmax$ operation of the \classicalNms~(Line $6$ of Alg.~\ref{alg:classical}).
    
    \classicalNmsCaps~only takes the overlap with unsuppressed boxes into account. Therefore, we generalize~\cref{eq:classical_all} by accounting for the 
    effect of all (suppressed and unsuppressed) boxes as
    \begin{align}
        r_i &= s_i  \prod\limits_{j=1}^{i-1}\left(1 -\pruneof{\overlap_{ij}}r_j\right).
        \label{eq:diff_nms_product}
    \end{align}
    The presence of $r_j$ on the RHS of \cref{eq:diff_nms_product} prevents suppressed boxes $r_j \approx 0$ from influencing other boxes hugely. 
    Let us say we have a box $b_2$ with a high overlap with an unsuppressed box $b_1$. The  \classicalNms~with a threshold pruning function assigns $r_2 = 0$ while \cref{eq:diff_nms_product} assigns $r_2$ a small non-zero value with a threshold pruning.
    
    Although \cref{eq:diff_nms_product} keeps $r_i \ge 0$, getting a closed-form recursion in $\rescore$ is not easy because of the product operation. To get a closed-form recursion with addition/subtraction in $\rescore$, we first carry out the polynomial multiplication and then ignore the higher-order terms as
    \begin{align}
        r_i &= s_i \left(1 - \sum\limits_{j=1}^{i-1}\pruneof{\overlap_{ij}}r_j + \mathcal{O}(n^2)\right) \nonumber\\
        &\approx s_i \left(1 - \sum\limits_{j=1}^{i-1}\pruneof{\overlap_{ij}}r_j \right) \nonumber\\
        &\approx s_i -  \sum\limits_{j=1}^{i-1}\pruneof{\overlap_{ij}}r_j.
        \label{eq:diff_nms_rescore_2}
    \end{align}
    Dropping the $s_i$ in the second term of \cref{eq:diff_nms_rescore_2} helps us get a cleaner form of \cref{eq:diff_nms_full_again}. Moreover, it does not change the nature of the NMS since the subtraction keeps the relation $r_i \le s_i$ intact as $\pruneof{\overlap_{ij}}$ and $r_j$ are both between $[0, 1]$. 
    
    We can also reach \cref{eq:diff_nms_rescore_2} directly as follows. \classicalNmsCaps~suppresses a box which has a high \iouTwoD~overlap with \emph{any} of the unsuppressed boxes ($r_j \approx 1$) to zero. 
    We consider \emph{any} as a logical non-differentiable OR operation and use logical OR $\bigvee$ operator's differentiable relaxation as $\sum$~\cite{van2020analyzing, li2019augmenting}. 
    We next use this relaxation with the other expression $\rescore  \le \score$.
    
    When a box shows overlap with more than two unsuppressed boxes, the term $\sum\limits_{j=1}^{i-1}\pruneof{\overlap_{ij}}r_j > 1$ in \cref{eq:diff_nms_rescore_2} or when a box shows high overlap with one unsuppressed box, the term $s_i < \pruneof{\overlap_{ij}}r_j$. In both of these cases, $r_i < 0$. So, we lower bound \cref{eq:diff_nms_rescore_2} with a $\max$ operation to ensure that $r_i \ge 0$. Thus,
    \begin{align}
        r_i &\approx \max\left(s_i - \sum\limits_{j=1}^{i-1}\pruneof{\overlap_{ij}}r_j,~0 \right).
        \label{eq:diff_nms_rescore_3}
    \end{align}

    We write the rescores $\rescore$ in a matrix formulation as
    \begin{align}
        \begin{bmatrix}
        r_1 \\
        r_2 \\
        r_3 \\
        \vdots \\
        r_n\\
        \end{bmatrix}
            &\!\approx\! 
        \max\left(
        \begin{bmatrix}
        s_1\\
        s_2\\
        s_3\\
        \vdots \\
        s_n\\
        \end{bmatrix}
        -
        \begin{bmatrix}
        0 & 0 & \dots & 0\\
        \pruneof{\overlap_{21}}\!&\!0\!&\!\dots\!&\!0\\
        \pruneof{\overlap_{31}}\!&\!\pruneof{\overlap_{32}}\!&\!\dots\!&\!0\\
        \vdots\!& \vdots & \vdots\!&\!\vdots \\
        \pruneof{\overlap_{n1}}\!&\!\pruneof{\overlap_{n2}} & \dots & 0 \\
        \end{bmatrix} 
        \begin{bmatrix}
        r_1 \\
        r_2 \\
        r_3 \\
        \vdots \\
        r_n\\
        \end{bmatrix}
        ,
        \begin{bmatrix}
        0 \\
        0 \\
        0 \\
        \vdots \\
        0\\
        \end{bmatrix}
        \right).
    \end{align}
    
    We next write the above equation compactly as
    \begin{align}
        \rescore &\approx \max(\score - \pruneMat\rescore,\zeroVector),
        \label{eq:diff_nms_recursive_2}
    \end{align}
    where $\pruneMat$, called the Prune Matrix, is obtained by element-wise operation of the pruning function $\prune$ on $\overlapMatLower$. Maximum operation makes \cref{eq:diff_nms_recursive_2} non-linear~\cite{kumar2013estimation} and, thus, difficult to solve. 
    
    However, for a differentiable NMS layer, we need to avoid the recursion. Therefore, we first solve \cref{eq:diff_nms_recursive_2} assuming the $\max$ operation is not present which gives us the solution $\rescore \approx \left( \identity + \pruneMat \right)^{-1}\score$. In general, this solution is not necessarily bounded between $0$ and $1$. Hence, we clip it explicitly to obtain the approximation
    \begin{align}
        \rescore \approx \clip{\left( \identity + \pruneMat \right)^{-1}\score},
        \label{eq:diff_nms_full_again}
    \end{align}
    which we use as the solution to \cref{eq:diff_nms_recursive_2}.

\section{Loss Functions}\label{sec:appendix_loss}
    We now detail out the loss functions used for training.
    The losses on the boxes before NMS, $\lossBefore$, is given by~\cite{brazil2020kinematic}
    \begin{align}
        \lossBefore &= \loss_\class + \loss_{\twoDMath} + \boxConfidence~\loss_{\threeDMath} \nonumber \\
        &\quad + \lossWeigh_{conf} (1- \boxConfidence),
    \end{align}
    where
    \begin{align}
        \loss_\class &= \text{CE}(b_\class, g_\class), \\
        \loss_{\twoDMath} &= -\log (\text{\iou}(b_{\twoDMath}, g_{\twoDMath})), \\
        \loss_{\threeDMath} &= \text{Smooth-L1} (b_{\threeDMath}, g_{\threeDMath}) \nonumber \\
        &\quad + \lossWeigh_{a} \text{CE}([b_{\theta_a}, b_{\theta_h}], [g_{\theta_a}, g_{\theta_h}]).
    \end{align}
    $\boxConfidence$ is the predicted self-balancing confidence of each box $b$, while $b_{\theta_a}$ and $b_{\theta_h}$ are its orientation bins~\cite{brazil2020kinematic}. 
    $g$ denotes the ground-truth.
    $\lossWeigh_{conf}$ is the rolling mean of most recent $\loss_{\threeDMath}$ losses per mini-batch~\cite{brazil2020kinematic}, while $\lossWeigh_{a}$ denotes the weight of the orientation bins loss.
    CE and Smoooth-L1 denote the Cross Entropy and Smooth L1 loss respectively.
    Note that we apply \twoD~and \threeD~regression losses as well as the confidence losses only on the foreground boxes.
    
    As explained in \cref{sec:target_loss}, the loss on the boxes after NMS, $\lossAfter$, is the \imageWise~\aploss, which is given by
    \begin{align}
        \lossAfter&=\!\loss_{\imageWise}\!=\!\frac{1}{N}\sum_{m=1}^N\!\apMath(\rescore^{(m)}, \text{target}(\boxes^{(m)})),
    \end{align}
    
    Let $\lossWeigh$ be the weight of the $\lossAfter$ term.
    Then, our overall loss function is given by 
    \begin{align}
        \loss &= \lossBefore + \lossWeigh \lossAfter \\
              &= \loss_\class + \loss_{\twoDMath} + \boxConfidence~\loss_{\threeDMath} + \lossWeigh_{conf} (1- \boxConfidence) \nonumber \\
              &\quad + \lossWeigh \loss_{\imageWise} \\
              &= \text{CE}(b_\class, g_\class) -  \log (\text{\iou}(b_{\twoDMath}, g_{\twoDMath})) \nonumber \\
              &\quad + \boxConfidence~\text{Smooth-L1} (b_{\threeDMath}, g_{\threeDMath}) \nonumber \\
              &\quad + \lossWeigh_{a}~\boxConfidence~\text{CE}([b_{\theta_a}, b_{\theta_h}], [g_{\theta_a}, g_{\theta_h}]) \nonumber \\
              &\quad + \lossWeigh_{conf} (1- \boxConfidence) + \lossWeigh \loss_{\imageWise}.
    \end{align}
    We keep $\lossWeigh_{a}= 0.35$ following~\cite{brazil2020kinematic}
    and $\lossWeigh = 0.05$. 
    Clearly, all our losses and their weights are identical to~\cite{brazil2020kinematic} except $\loss_{\imageWise}$.

    \begin{table*}[t]
        \caption[Detailed comparisons during inference on \kitti \valOne cars.]{\textbf{Detailed comparisons with other NMS} during inference on \kitti \valOne cars.}
        \label{tab:results_kitti_val1_other_nms_detailed}
        \centering
        \footnotesize
        \scalebox{\scaleFraction}{
        \setlength\tabcolsep{0.1cm}
        \begin{tabular}{tl t c m ccc t ccc m ccc t ccct}
            \myTopRule
            \addlinespace[0.01cm]
            \multirow{3}{*}{} &  \multirow{3}{*}{\shortstack{Inference\\NMS}} & \multicolumn{6}{cm}{\iouThreeD $\geq 0.7$} & \multicolumn{6}{ct}{\iouThreeD $\geq 0.5$}\\\cline{3-14}
            & & \multicolumn{3}{ct}{\apThreeDForty ($\uparrowRHDSmall$)} & \multicolumn{3}{cm}{\apBevForty ($\uparrowRHDSmall$)} & \multicolumn{3}{ct}{\apThreeDForty ($\uparrowRHDSmall$)} & \multicolumn{3}{ct}{\apBevForty ($\uparrowRHDSmall$)}\\
            & & Easy & Mod & Hard & Easy & Mod & Hard & Easy & Mod & Hard & Easy & Mod & Hard\\ 
            \myTopRule
            \kinematicImage~\cite{brazil2020kinematic}      &\classicalNmsShortCaps& $18.28$        & $13.55$        & $10.13$       & $25.72$        & $18.82$       & $14.48$       & $54.70$        & $39.33$        & $31.25$        & $60.87$        & $44.36$       & $34.48$       \\
            \kinematicImage~\cite{brazil2020kinematic}      &\softNmsShortCaps~\cite{bodla2017soft}                 & $18.29$        & $13.55$        & $10.13$       & $25.71$        & $18.81$        & $14.48$       & $54.70$       & $39.33$        & $31.26$        & $60.87$        & $44.36$       & $34.48$       \\
            \kinematicImage~\cite{brazil2020kinematic}      &\distanceNmsShortCaps~\cite{shi2020distance}          & $18.25$        & $13.53$        & $10.11$       & $25.71$        & $18.82$       & $14.48$        & $54.70$       & $39.33$        & $31.26$        & $60.87$        & $44.36$       & $34.48$       \\
            \kinematicImage~\cite{brazil2020kinematic}      &\groomedNMSShort~     & $18.26$        & $13.51$        & $10.10$       & $25.67$        & $18.77$       & $14.44$        & $54.59$       & $39.25$        & $31.18$        & $60.78$        & $44.28$       & $34.40$       \\
            \hline
            \groomedNMS                      &\classicalNmsShortCaps& $19.67$ & $14.31$  & $11.27$ & $27.38$ & $19.75$  & $15.93$ & $55.64$ & $41.08$ & $32.91$& $61.85$ & $44.98$ & $36.31$\\
            \groomedNMS                      &\softNmsShortCaps~\cite{bodla2017soft}& $19.67$ & $14.31$  & $11.27$ & $27.38$ & $19.75$  & $15.93$ & $55.64$ & $41.08$ & $32.91$& $61.85$ & $44.98$ & $36.31$\\
            \groomedNMS                      &\distanceNmsShortCaps~\cite{shi2020distance}          & $19.67$ & $14.31$  & $11.27$ & $27.38$ & $19.75$  & $15.93$ & $55.64$ & $41.08$ & $32.91$& $61.85$ & $44.98$ & $36.31$\\
            \groomedNMS                      &\groomedNMSShort~     & $19.67$ & $14.32$  & $11.27$ & $27.38$ & $19.75$  & $15.92$ & $55.62$ & $41.07$ & $32.89$& $61.83$ & $44.98$ & $36.29$\\
            \myTopRule
        \end{tabular}
        }
    \end{table*}

\section{Additional Experiments and Results}\label{sec:additional_exp}
    We now provide additional details and results evaluating our system's performance.

    \subsection{Training}\label{sec:training_additional}
        Training images are augmented using random flipping with probability $0.5$~\cite{brazil2020kinematic}.
        Adam optimizer~\cite{kingma2014adam} is used with batch size $2$, weight-decay $5\times10^{-4}$ and gradient clipping of $1$~\cite{brazil2019m3d, brazil2020kinematic}.
        Warmup starts with a learning rate $4 \times 10^{-3}$ following a poly learning policy with power $0.9$~\cite{brazil2020kinematic}.
        Warmup and full training phases take $80k$ and $50k$ mini-batches respectively for \valOne and~\valTwo~Splits~\cite{brazil2020kinematic} while take $160k$ and $100k$ mini-batches for Test Split. 

    \subsection{\kitti \valOne Oracle NMS Experiments}\label{sec:results_oracle_additional}
        As discussed in \cref{sec:Introduction}, to understand the effects of an inference-only NMS on \twoD~and \threeD~object detection, we conduct a series of oracle experiments.
        We create an oracle NMS by taking the Val Car boxes of \kitti \valOne Split from the baseline \kinematicImage~model \textit{before} NMS and replace their scores with their true \iouTwoD~or \iouThreeD with the ground-truth, respectively. 
        Note that this corresponds to the oracle because we do not know the ground-truth boxes during inference.
        We then pass the boxes with the oracle scores through the \classicalNms~and report the results in \cref{tab:oracle_nms}.
        
        The results show that the \apThreeD increases by a staggering $>60$ \ap~on Mod cars when we use oracle \iouThreeD as the NMS score.
        On the other hand, we only see an increase in \apTwoD~by $\approx 11$ \ap~on Mod cars when we use oracle \iouTwoD~as the NMS score.
        Thus, the relative effect of using oracle \iouThreeD NMS scores on \threeD~detection is more significant than using oracle \iouTwoD~NMS scores on \twoD~detection.
        In other words, the mismatch is greater between classification and \threeD~localization compared to the mismatch between classification and \twoD~localization.

    \subsection{\kitti \valOne 3D~Object Detection}\label{sec:results_kitti_val1_additional}
    
    \textbf{Comparisons with other NMS.}
        We compare \groomedNMS with the other NMS---\classicalNmsShort, \softNmsShortCaps~\cite{bodla2017soft} and \distanceNmsCaps~\cite{shi2020distance} and report the detailed results in \cref{tab:results_kitti_val1_other_nms_detailed}. 
        We use the publicly released \softNmsCaps~code and \distanceNmsCaps~code from the respective authors.
        The \distanceNmsCaps~model uses the class confidence scores divided by the uncertainty in $z$ (the most erroneous dimension in \threeD~localization~\cite{simonelli2021we}) of a box as the \distanceNmsCaps~\cite{shi2020distance} input. 
        Our model does not predict the uncertainty in $z$ of a box
        but predicts its self-balancing confidence (the \threeD~localization score). 
        Therefore, we use the class confidence scores multiplied by the self-balancing confidence as the \distanceNmsCaps~input.
        
        The results in \cref{tab:results_kitti_val1_other_nms_detailed} show that NMS inclusion in the training pipeline benefits the performance, unlike~\cite{bodla2017soft}, which suggests otherwise.
        Training with \groomedNMS helps because the network gets an additional signal through the \groomedNMS layer whenever the best-localized box corresponding to an object is not selected. 
        Moreover, \cref{tab:results_kitti_val1_other_nms_detailed} suggests that we can replace \groomedNMS with the~\classicalNms~in inference as the performance is almost the same even at \iouThreeD$=0.5$.

    \textbf{How good is the \classicalNms~approximation?}
        \groomedNMS uses several approximations to arrive at the matrix solution~\cref{eq:diff_nms_full_again}.
        We now compare how good these approximations are with the \classicalNms. 
        Interestingly, \cref{tab:results_kitti_val1_other_nms_detailed} shows that \groomedNMS is an excellent approximation to the \classicalNms~as the performance does not degrade after changing the NMS in inference.

    \subsection{\kitti \valOne Sensitivity Analysis}
        There are a few adjustable parameters for the \groomedNMS, such as the NMS threshold $\nmsThresh$, valid box threshold $\validBoxThresh$, the maximum group size $\alpha$, the weight $\lossWeigh$ for the $\lossAfter$, and $\beta$. We carry out a sensitivity analysis to understand how these parameters affect performance and speed, and how sensitive the algorithm is to these parameters.

        \begin{table}[!t]
            \caption[Sensitivity to NMS threshold $\nmsThresh$ on \kitti \valOne cars.]{\textbf{Sensitivity to NMS threshold $\nmsThresh$} on \kitti \valOne cars. [Key: \bestKey{Best}]}
            \label{tab:sensitivity_to_nms_thresh}
            \centering
            \setlength\tabcolsep{2.00pt}
            \begin{tabular}{tl m ccc  m ccct}
                \myTopRule
                \multirow{2}{*}{ } & \multicolumn{3}{cm}{\apThreeDForty ($\uparrowRHDSmall$)} & \multicolumn{3}{ct}{\apBevForty ($\uparrowRHDSmall$)}\\ 
                & Easy & Mod & Hard & Easy & Mod & Hard\\ 
                \myTopRule
                $\nmsThresh=0.3$ & $17.49$ & $13.32$ & $10.54$ & $26.07$ & $18.94$ & $14.61$\\
                $\nmsThresh=$\best{0.4} & \best{19.67} & \best{14.32} & \best{11.27} & \best{27.38} & \best{19.75} & \best{15.92}\\
                $\nmsThresh=0.5$ & $19.65$ & $13.93$ & $11.09$ & $26.15$ & $19.15$ & $14.71$\\
                \myTopRule
            \end{tabular}
        \end{table}
        
        \begin{table}[!t]
            \caption[Sensitivity to valid box threshold $\validBoxThresh$ on \kitti \valOne cars.]{\textbf{Sensitivity to valid box threshold }$\validBoxThresh$ on \kitti \valOne cars. [Key: \bestKey{Best}]}
            \label{tab:sensitivity_to_v}
            \centering
            \setlength\tabcolsep{2.00pt}
            \begin{tabular}{tl m ccc  m ccct}
                \myTopRule
                \multirow{2}{*}{ } & \multicolumn{3}{cm}{\apThreeDForty ($\uparrowRHDSmall$)} & \multicolumn{3}{ct}{\apBevForty ($\uparrowRHDSmall$)}\\ 
                & Easy & Mod & Hard & Easy & Mod & Hard\\ 
                \myTopRule
                $\validBoxThresh= 0.01$& $13.71$ & $9.65$ & $7.24$ & $17.73$ & $12.47$ & $9.36$\\
                $\validBoxThresh= 0.1$ & $19.37$ & $13.99$ & $10.92$ & $26.95$ & $19.84$ & $15.40$\\
                $\validBoxThresh= 0.2$ & $19.65$ & $14.31$ & $11.24$ & $27.35$ & $19.73$ & $15.89$\\
                $\validBoxThresh= 0.3$ & $19.67$ & $14.32$ & $11.27$ & $27.38$ & $19.75$ & $15.92$\\
                $\validBoxThresh= 0.4$ & ${19.67}$ & ${14.33}$ & ${11.28}$ & ${27.38}$ & ${19.76}$ & ${15.93}$\\
                $\validBoxThresh= 0.5$ & $19.67$ & $14.33$ & $11.28$ & $27.38$ & $19.76$ & $15.93$\\
                $\validBoxThresh=$ \best{0.6} & \best{19.67} & \best{14.33} & \best{11.29} & \best{27.39} & \best{19.77} & \best{15.95}\\
                \myTopRule
            \end{tabular}
        \end{table}

        \noIndentHeading{Sensitivity to NMS Threshold.} 
        We show the sensitivity to NMS threshold $\nmsThresh$ in  \cref{tab:sensitivity_to_nms_thresh}.
        The results in \cref{tab:sensitivity_to_nms_thresh} show that the optimal $\nmsThresh= 0.4$.
        This is also the $\nmsThresh$ in~\cite{brazil2019m3d, brazil2020kinematic}.
        
        \noIndentHeading{Sensitivity to Valid Box Threshold.} 
        We next show the sensitivity to valid box threshold $\validBoxThresh$ in \cref{tab:sensitivity_to_v}. 
        Our choice of $\validBoxThresh= 0.3$ performs close to the optimal choice.

\begin{figure}[!t]
            \centering
            \includegraphics[width=0.7\linewidth]{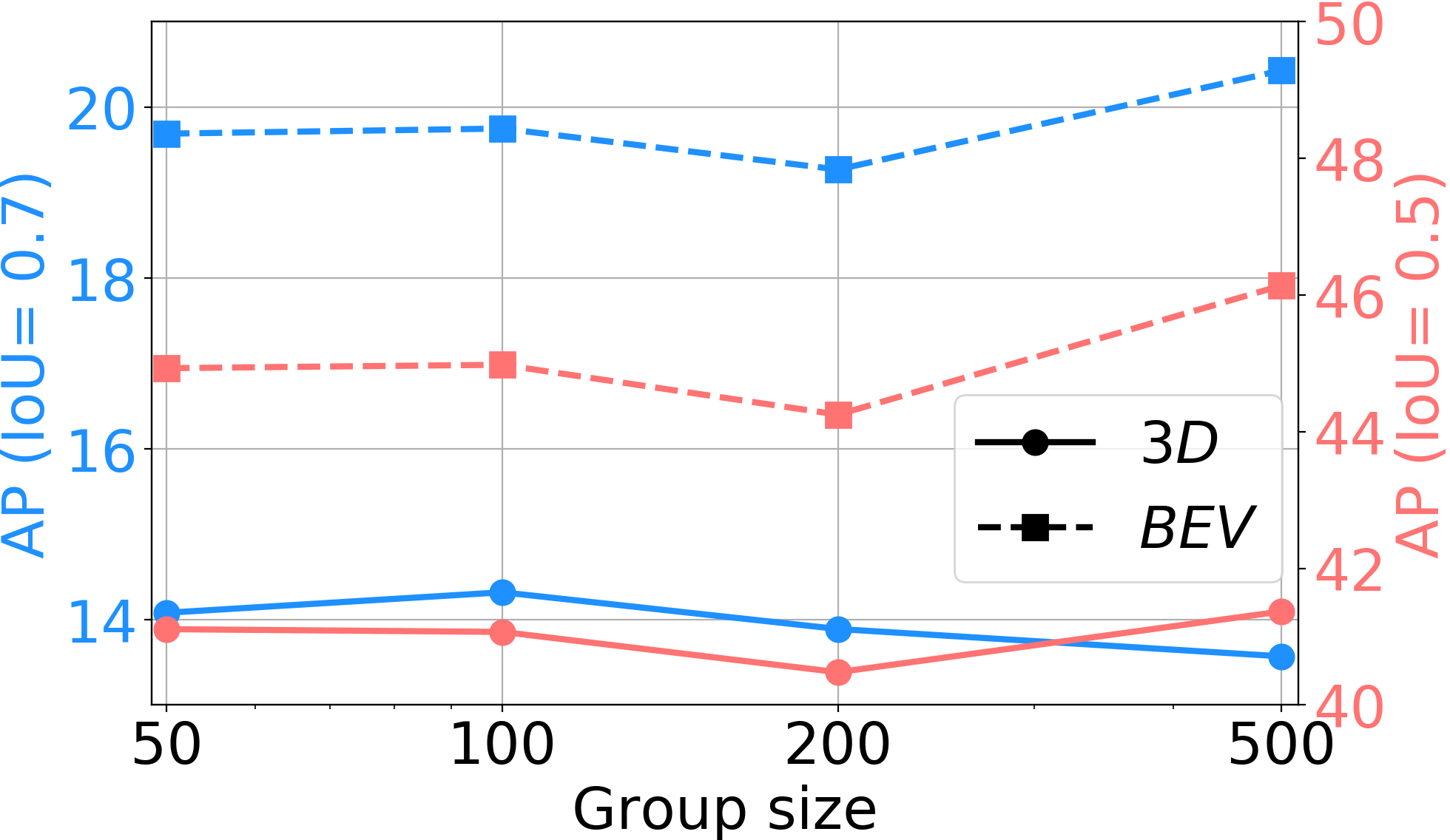}
            \caption[Sensitivity to group size $\groupSize$ on \kitti \valOne cars.]{\textbf{Sensitivity to group size $\groupSize$} on \kitti \valOne Moderate cars.}
            \label{fig:sensitivity_to_group_size}
        \end{figure}

        \noIndentHeading{Sensitivity to Maximum Group Size.}
            Grouping has a parameter group size $(\groupSize)$. 
            We vary this parameter and report \apThreeDForty and~\apBevForty at two different \iouThreeD thresholds on Moderate cars of \kitti \valOne Split in \cref{fig:sensitivity_to_group_size}. 
            We note that the best \apThreeDForty performance is obtained at $\groupSize= 100 $ and we, therefore, set $\groupSize= 100$ in our experiments.
        
        \noIndentHeading{Sensitivity to Loss Weight.}
            We now show the sensitivity to loss weight $\lossWeigh$ in \cref{tab:sensitivity_to_lambda}. 
            Our choice of $\lossWeigh= 0.05$ is the optimal value.
        
        \noIndentHeading{Sensitivity to Best Box Threshold.}
            We now show the sensitivity to the best box threshold $\beta$ in \cref{tab:sensitivity_to_beta}. 
            Our choice of $\beta= 0.3$ is the optimal value.
        
        \noIndentHeading{Conclusion.}
            \groomedNMS has minor sensitivity to $\nmsThresh, \groupSize, \lossWeigh$ and $\beta$, which is common in object detection. 
            \groomedNMS is not as sensitive to $\validBoxThresh$ since it only decides a box's validity. 
            Our parameter choice is either at or close to the optimal.
            The inference speed is only affected by $\groupSize$. Other parameters are used in training or do not affect inference speed.
    
        \begin{table}[!t]
            \caption[Sensitivity to loss weight $\lossWeigh$ on \kitti \valOne cars]{\textbf{Sensitivity to loss weight} $\lossWeigh$ on \kitti \valOne cars. [Key: \bestKey{Best}]}
            \label{tab:sensitivity_to_lambda}
            \centering
            \setlength\tabcolsep{2.00pt}
            \begin{tabular}{tl m ccc  m ccct}
                \myTopRule
                \multirow{2}{*}{ } & \multicolumn{3}{cm}{\apThreeDForty ($\uparrowRHDSmall$)} & \multicolumn{3}{ct}{\apBevForty ($\uparrowRHDSmall$)}\\ 
                & Easy & Mod & Hard & Easy & Mod & Hard\\ 
                \myTopRule
                $\lossWeigh= 0$ & $19.16$ & $13.89$ & $10.96$ & $27.01$ & $19.33$ & $14.84$\\
                \best{\lossWeigh= 0.05} & \best{19.67} & \best{14.32} & \best{11.27} & \best{27.38} & \best{19.75} & \best{15.92}\\
                $\lossWeigh= 0.1$ & $17.74$ & $13.61$ & $10.81$ & $25.86$	& $19.18$ & $15.57$\\
                $\lossWeigh= 1$   & $10.08$ & $ 7.26$ & $ 6.00$ & $14.44$ & $10.55$ & $8.41$\\
                \myTopRule
            \end{tabular}
        \end{table}

        \begin{table}[!t]
            \caption[Sensitivity to best box threshold $\beta$ on \kitti \valOne cars.]{\textbf{Sensitivity to best box threshold $\beta$} on \kitti \valOne cars. [Key: \bestKey{Best}]}
            \label{tab:sensitivity_to_beta}
            \centering
            \setlength\tabcolsep{2.00pt}
            \begin{tabular}{tl m ccc  m ccct}
                \myTopRule
                \multirow{2}{*}{ } & \multicolumn{3}{cm}{\apThreeDForty ($\uparrowRHDSmall$)} & \multicolumn{3}{ct}{\apBevForty ($\uparrowRHDSmall$)}\\ 
                & Easy & Mod & Hard & Easy & Mod & Hard\\ 
                \myTopRule
                $\beta= 0.1$ & $18.09$ & $13.64$ & $10.21$ & $26.52$ & $19.50$ & $15.74$\\ 
                \best{\beta= 0.3} & \best{19.67} & \best{14.32} & \best{11.27} & \best{27.38} & \best{19.75} & \best{15.92}\\ 
                $\beta= 0.4$ & $18.91$ & $14.02$ & $11.15$ & $27.11$ & $19.64$ & $15.90$\\
                $\beta= 0.5$ & $18.49$ & $13.66$ & $10.96$ & $27.01$ & $19.47$ & $15.79$\\
                \myTopRule
            \end{tabular}
        \end{table}

    \subsection{Qualitative Results}
\begin{figure*}[!tb]
            \centering
            \begin{subfigure}{\linewidth}
              \includegraphics[width=\linewidth]{images/groomed/qualitative/000514.png}
            \end{subfigure}
            \begin{subfigure}{\linewidth}
              \includegraphics[width=\linewidth]{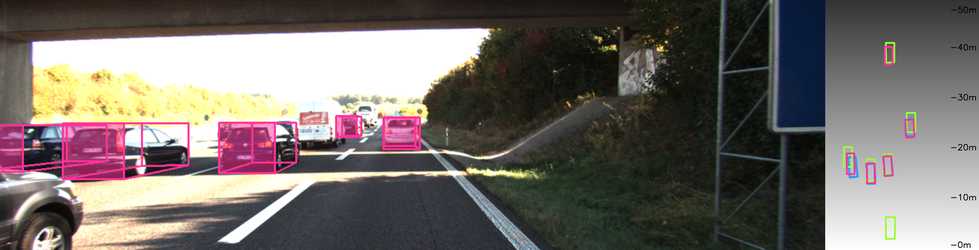}
            \end{subfigure}
            \begin{subfigure}{\linewidth}
              \includegraphics[width=\linewidth]{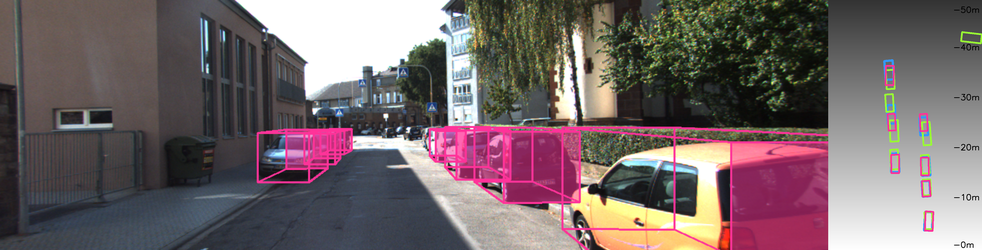}
            \end{subfigure}
            \begin{subfigure}{\linewidth}
              \includegraphics[width=\linewidth]{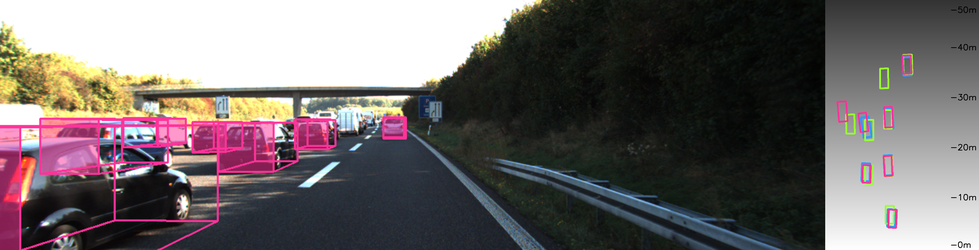}\\
            \end{subfigure}
            \caption[Qualitative Results.]{\textbf{Qualitative Results} (Best viewed in color). We depict the predictions of \groomedNMS (magenta) in image view on the left and the predictions of \groomedNMS, {\kinematicImage}~\cite{brazil2020kinematic} (blue), and {Ground Truth} (green) in BEV on the right. In general, {\groomedNMS} predictions are more closer to the {ground truth} than {\kinematicImage}~\cite{brazil2020kinematic}.}
            \label{fig:qualitative}
        \end{figure*}

        We next show some qualitative results of models trained on \kitti \valOne Split in \cref{fig:qualitative}. We depict the predictions of \groomedNMS in image view on the left and the predictions of \groomedNMS, \kinematicImage~\cite{brazil2020kinematic}, and ground truth in BEV on the right. In general, \groomedNMS predictions are more closer to the ground truth than \kinematicImage~\cite{brazil2020kinematic}.

    \subsection{Demo Video of \groomedNMS}   
        We next include a short demo video of our \groomedNMS model trained on \kitti \valOne Split. 
        We run our trained model independently on each frame of the three \kitti raw \cite{geiger2013vision} sequences - \textsc{2011\_10\_03\_drive\_0047}, \textsc{2011\_09\_29\_drive\_0026} and \textsc{2011\_09\_26\_drive\_0009}.
        None of the frames from these three raw sequences appear in the training set of \kitti \valOne Split.
        We use the camera matrices available with the raw sequences but do not use any temporal information. 
        Overlaid on each frame of the raw input videos, we plot the projected \threeD~boxes of the predictions and also plot these \threeD~boxes in the BEV. 
        We set the frame rate of this demo at $10$ fps.
        The demo is also available in HD at \url{https://www.youtube.com/watch?v=PWctKkyWrno}.
        In the demo video, notice that the orientation of the boxes are stable despite not using any temporal information.

%% file: appendices/deviant_appendix.tex
\chapter{\deviant Appendix}\label{chpt:deviant_appendix}

\section{Supportive Explanations}\label{sec:deviant_supplementary_explanation}
    
    We now add some explanations which we could not put in the main chapter because of the space constraints.
    
    \subsection{\Equivariance{} vs Augmentation} 
        \Equivariance{} adds suitable inductive bias to the backbone \cite{cohen2016group, dieleman2016exploiting} and is not learnt.
        Augmentation adds transformations to the input data during training or inference.
        
        \Equivariance{} and data augmentation have their own pros and cons.
        \Equivariance{} models the physics better, is mathematically principled and is so more agnostic to data distribution shift compared to the data augmentation.
        A downside of \equivariance{} compared to the augmentation is \equivariance{} requires mathematical modelling, may not always exist \cite{burns1992non}, is not so intuitive and generally requires more flops for inference.
        On the other hand, data augmentation is simple, intuitive and fast, but is not mathematically principled. 
        The choice between \equivariance{} and data augmentation is a withstanding question in machine learning \cite{gandikota2021training}.
    
    \subsection{Why do \twoD CNN detectors generalize?}
        We now try to understand why \twoD CNN detectors generalize well.
        Consider an image $\projectionOne(\pixU, \pixV)$ and $\mapping$ be the CNN.
        Let $\transformationMath_\translation$ denote the translation in the $(\pixU,\pixV)$ space.
        The \twoD translation \equivariance{} \cite{bronstein2021convolution, bronstein2021geometric, rath2020boosting} of the CNN means that 
        \begin{align}
            \mapping(\transformationMath_\translation\projectionOne(\pixU, \pixV)) &=
            \transformationMath_\translation\mapping(\projectionOne(\pixU, \pixV)) \nonumber \\
            \implies \mapping(\projectionOne(\pixU + \transU, \pixV + \transV)) &=
            \mapping(\projectionOne(\pixU, \pixV)) + (\transU, \transV)
            \label{eq:translation_equivariance}
        \end{align}
        where $(\transU, \transV)$ is the translation in the $(\pixU,\pixV)$ space.
        
        Assume the CNN predicts the object position in the image as $(\pixUTwo, \pixVTwo)$. 
        Then, we write
            \begin{align}
                \mapping(\projectionOne(\pixU, \pixV)) &= (\predictionU, \predictionV)
            \end{align}
        
        Now, we want the CNN to predict the output the position of the same object translated by $(\transU, \transV)$.
        The new image is thus $\projectionOne(\pixU + \transU, \pixV + \transV)$.
        The CNN easily predicts the translated position of the object because all CNN is to do is to invoke its \twoD translation \equivariance{} of \cref{eq:translation_equivariance}, and translate the previous prediction by the same amount.
        In other words,
        \begin{align*}
            \mapping(\projectionOne(\pixU + \transU, \pixV + \transV)) &=    
            \mapping(\projectionOne(\pixU, \pixV)) + (\transU, \transV)\\
            &=  (\predictionU, \predictionV) + (\transU, \transV) \\
            &=  (\predictionU + \transU, \predictionV + \transV)
        \end{align*}
        Intuitively, \equivariance{} is a disentaglement method. 
        The \twoD translation \equivariance{} disentangles the \twoD translations $(\transU, \transV)$ from the original image $\projectionOne$ and therefore, the network generalizes to unseen \twoD translations.

    \subsection{Existence and Non-existence of \Equivariance{}}
        The result from \cite{burns1992non} says that generic projective \equivariance{} does not exist in particular with rotation transformations.
        We now show an example of when the \equivariance{} exists and does not exist in the projective manifold in \cref{fig:deviant_eqv_exists,fig:deviant_non_existence} respectively.
    
    \begin{figure}[!tb]
        \centering
        \input{images/deviant/translational_equiv_exist}
        \caption[\Equivariance{} exists for the \plane{} when there is depth translation of the ego camera.]
        {
            \textbf{\Equivariance{} exists} for the \plane{} when there is depth translation of the ego camera. 
            Downscaling converts image $\projectionOne$ to image $\projectionTwo$.
        }
        \label{fig:deviant_eqv_exists}
    \end{figure}
    \begin{figure}[!tb]
        \centering
        \input{images/deviant/translational_equiv_not_exist}
        \caption[Example of non-existence of \equivariance{} when there is $180^\circ$ rotation of the ego camera.]{
            \textbf{Example of non-existence of \equivariance{} }\cite{burns1992non} when there is $180^\circ$ rotation of the ego camera. 
            No transformation can convert image $\projectionOne$ to image $\projectionTwo$.
        }
        \label{fig:deviant_non_existence}
    \end{figure}

    \subsection{Why do not Monocular 3D CNN detectors generalize?}
        Monocular \threeD CNN detectors do not generalize well because they are not \equivariant{} to arbitrary \threeD translations in the projective manifold.
        To show this, let $\pointCloudOne(\varX, \varY, \varZ)$ denote a \threeD point cloud.
        The monocular detection network $\mapping$ operates on the projection $\projectionOne(\pixU, \pixV)$ of this point cloud $\pointCloudOne$ to output the position $(\predictionX, \predictionY, \predictionZ)$ as
        \begin{align*}
            \mapping( \projectionOperator \pointCloudOne(\varX, \varY, \varZ)) &= (\predictionX, \predictionY, \predictionZ) \\
            \implies \mapping(   \projectionOne(\pixU, \pixV) ) &= (\predictionX, \predictionY, \predictionZ),   
        \end{align*}
        where $\projectionOperator$ denotes the projection operator.
        We translate this point cloud by an arbitrary \threeD translation of $(\transX,\transY,\transZ)$ to obtain the new point cloud  $\pointCloudOne(\varX + \transX, \varY + \transY, \varZ + \transZ)$. 
        Then, we again ask the monocular detector $\mapping$ to do prediction over the translated point cloud.
        However, we find that
        \begin{align*}
            \mapping( \projectionOperator \pointCloudOne(\varX + \transX, \varY + \transY, \varZ + \transZ)) &\ne \mapping(\projectionOne(\pixU + \projectionOperator (\transX, \transY,\transZ), \pixV + \projectionOperator (\transX, \transY, \transZ))) \\
            &= \mapping(\projectionOne(\pixU, \pixV)) + \projectionOperator (\transX, \transY, \transZ)  \\
            \implies \mapping( \projectionOperator \pointCloudOne(\varX + \transX, \varY + \transY, \varZ + \transZ)) &\ne \mapping(\projectionOperator \pointCloudOne(\varX, \varY, \varZ)) + \projectionOperator (\transX, \transY, \transZ)
        \end{align*}
        In other words, the projection operator $\projectionOperator$ does not distribute over the point cloud $\pointCloudOne$ and arbitrary \threeD translation of $(\transX,\transY,\transZ)$.
        Hence, if the network $\mapping$ is a vanilla CNN (existing monocular backbone), it can no longer invoke its \twoD translation \equivariance{} of  \cref{eq:translation_equivariance} to get the new \threeD coordinates $(\predictionX + \transX, \predictionY + \transY, \predictionZ + \transZ)$.
        
        Note that the \lidar{} based \threeD detectors with \threeD convolutions do not suffer from this problem because they do not involve any projection operator $\projectionOperator$.
        Thus, this problem exists only in monocular \threeD detection. 
        This makes monocular \threeD detection different from \twoD and \lidar{} based \threeD object detection.

    \subsection[Overview of Planar Transformations.]{Overview of Planar Transformations: \cref{th:projective_bigboss}}
    
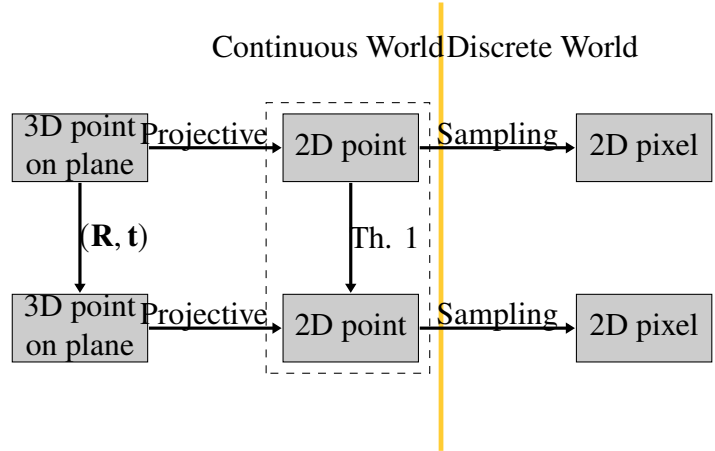
\begin{figure}[!htb]
            \centering
            \input{images/deviant/theorem_1_overview}
            \caption[Overview of Planar Transformations, which links the planarity and projective transformations in the continuous world.]
            {
                \textbf{Overview} of \cref{th:projective_bigboss} (Example 13.2 from \cite{hartley2003multiple}), which links the planarity and projective transformations in the continuous world.
            }
            \label{fig:deviant_high_level}
        \end{figure}

        We now pictorially provide the overview of \cref{th:projective_bigboss} (Example 13.2 from \cite{hartley2003multiple}), which links the planarity and projective transformations in the continuous world in \cref{fig:deviant_high_level}.
    
    \subsection[Approximation of Scale Transformations]{Approximation of Scale Transformations: \cref{th:projective_scaled}}\label{sec:deviant_approximation_proof}
        We now give the approximation under which \cref{th:projective_scaled} is valid.
        We assume that the ego camera does not undergo any rotation. Hence, we substitute $\rotation=\identity$ in \cref{eq:bigboss} to get
            \begin{align}
                \projectionOne(\pixU-\ppointU, \pixV-\ppointV)
                &= \projectionTwo\left(\focal\dfrac{\left(1\mySign\transX\frac{m}{p}\right)\pixUMinusPointU\mySign\transX\frac{n}{p}\pixVMinusPointV 
                \mySign\transX\frac{o}{p}\focal}{\transZ\frac{m}{p}\pixUMinusPointU \mySign\transZ\frac{n}{p}\pixVMinusPointV+\left(1\mySign\transZ\frac{o}{p}
                \right)\focal} ,
                \right. \nonumber \\
                &\quad\quad\left.\focal\dfrac
                {\transY\frac{m}{p}\pixUMinusPointU +  \left(1\mySign\transY\frac{n}{p}\right)\pixVMinusPointV \mySign \transY\frac{o}{p}\focal
                }
                {\transZ\frac{m}{p}\pixUMinusPointU 
                + \transZ\frac{n}{p} \pixVMinusPointV+\left(1\mySign\transZ\frac{o}{p}\right)\focal} \right).
                \label{eq:projective_no_rotation}
            \end{align}
        Next, we use the assumption that the ego vehicle moves in the $z$-direction as in \cite{brazil2020kinematic},  \thatIs, substitute $\transX\!=\!\transY\!=\!0$ to get
        \begin{align}
            \projectionOne(\pixU\!-\!\ppointU, \pixV\!-\!\ppointV) 
            &= \projectionTwo\left(\dfrac{\pixU-\ppointU}{\frac{\transZ}{\focal}\frac{m}{p}\pixUMinusPointU \mySign\frac{\transZ}{\focal}\frac{n}{p}\pixVMinusPointV +\left(1\mySign\transZ\frac{o}{p}\right)},  \right. \nonumber \\
            &\quad\quad\quad~\left.\dfrac{\pixV-\ppointV}{\frac{\transZ}{\focal}\frac{m}{p}\pixUMinusPointU \mySign\frac{\transZ}{f}\frac{n}{p}\pixVMinusPointV +\left(1\mySign\transZ\frac{o}{p}\right)} \right).
            \label{eq:projective_only_z}
        \end{align}

        The \plane{} is $mx + ny + oz + p= 0$. 
        We consider the planes in the front of camera. 
        Without loss of generality, consider $p < 0$ and $o > 0$.
    
        We first write the denominator $D$ of RHS term in \cref{eq:projective_only_z} as
        \begin{align}
            D &= \frac{\transZ}{\focal}\frac{m}{p}\pixUMinusPointU \mySign\frac{\transZ}{f}\frac{n}{p}\pixVMinusPointV +\left(1\mySign\transZ\frac{o}{p}\right) \nonumber \\
            &= 1 + \frac{\transZ}{p} \left(\frac{m}{f}\pixUMinusPointU + \frac{n}{f}\pixVMinusPointV + o \right) \nonumber
        \end{align}
        Because we considered \plane s in front of the camera, $p < 0$. 
        Also consider $\transZ < 0$, which implies $\transZ / p > 0$.
        Now, we bound the term in the parantheses of the above equation as
        \begin{align*}
            D &\le 1 +  \frac{\transZ}{p} \norm{\frac{m}{f}\pixUMinusPointU + \frac{n}{f}\pixVMinusPointV + o } \\
              &\le 1 + \frac{\transZ}{p} \left( \norm{\frac{m}{f}\pixUMinusPointU} + \norm{\frac{n}{f}\pixVMinusPointV} + \norm{o} \right) \quad \text{by Triangle inequality}\\
              &\le 1 + \frac{\transZ}{p} \left(\frac{\norm{m}}{f}\frac{W}{2} + \frac{\norm{n}}{f}\frac{H}{2} + o \right), \pixUMinusPointU \le \frac{W}{2}, \pixVMinusPointV \le \frac{H}{2}, \norm{o}= o\\
              &\le 1 + \frac{\transZ}{p} \left(\frac{\norm{m}}{f}\frac{W}{2} + \frac{\norm{n}}{f}\frac{W}{2} + o \right), H \le W\\
              &\le 1 + \frac{\transZ}{p} \left( \frac{(\norm{m} + \norm{n}) W}{2f} + o \right) , \quad 
        \end{align*}
        
\begin{figure}[!htb]
            \centering
            \input{images/deviant/the_assumption}
            \caption[Approximation of Scale Transformations with parallel and approximated planes.]
            {
                \textbf{Approximation of \cref{th:projective_scaled}}. Bold shows the \plane{}  parallel to the image plane. The dotted line shows the approximated \plane{}.
            }
            \label{fig:deviant_assumption}
        \end{figure}
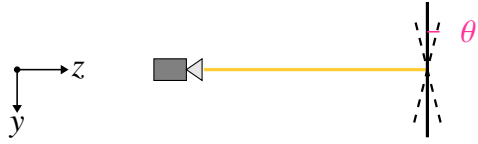

        If the coefficients of the patch plane $m,n,o$, its width $W$ and focal length $f$ follow the relationship $\frac{(\norm{m} + \norm{n}) W}{2f} << o$, the \plane{} is ``approximately''  parallel to the image plane.
        Then, a few quantities can be ignored in the denominator $D$ to get
        \begin{align}
            \text{D} &\approx 1 + \transZ\frac{o}{p} 
        \end{align}
        Therefore, the RHS of \cref{eq:projective_only_z} gets simplified and we obtain
        \begin{align}
            &\transformationMath_\scaleNotation :\projectionOne(\pixU-\ppointU, \pixV-\ppointV)
            \approx \projectionTwo\left(\dfrac{\pixU-\ppointU}{1\mySign\transZ\frac{o}{p}}, \dfrac{\pixV-\ppointV}{1\mySign\transZ\frac{o}{p}}\right)
        \end{align}
        An immediate benefit of using the approximation is \cref{eq:projective_scaled} does not depend on the distance of the \plane{} from the camera. 
        This is different from wide-angle camera assumption, where the ego camera is assumed to be far from the \plane{}.
        Moreover, \plane s need not be perfectly aligned with the image plane for \cref{eq:projective_scaled}. Even small enough perturbed \plane s work.
        We next show the approximation in the \cref{fig:deviant_assumption} with $\theta$ denoting the deviation from the perfect parallel plane.
        The deviation $\theta$ is about $3$ degrees for the \kitti{} dataset while it is $6$ degrees for the \waymo{} dataset.
        
        \forExample{} The following are valid \plane s for \kitti{} images whose focal length $\focal = 707$ and width $W = 1242$.
        \begin{align}
            -0.05x + 0.05y + z &= 30 \nonumber \\
            0.05x - 0.05y + z &= 30 
        \end{align}
        The following are valid \plane s for \waymo{} images whose focal length $\focal = 2059$ and width $W = 1920$.
        \begin{align}
            -0.1x + 0.1y + z &= 30 \nonumber \\
            0.1x - 0.1y + z &= 30 
        \end{align}
        
        Although the assumption is slightly restrictive, we believe our method shows improvements on both \kitti{} and \waymo{} datasets because the car patches are approximately parallel to image planes and also because the depth remains the hardest parameter to estimate \cite{ma2021delving}.

    \subsection{\ScaleEquivariance{} of \ses{} Convolution for Images}\label{sec:deviant_scale_eqv_proof}
        \cite{sosnovik2020sesn} derive the \scaleEquivariance{} of \ses{} convolution for a \oneD{} signal.
        We simply follow on their footsteps to get the \scaleEquivariance{} of \ses{} convolution for a \twoD image $\projectionOne(\pixU,\pixV)$ for the sake of completeness. 
        Let the scaling of the image $\projectionOne$ be $\scaleNotation$. 
        Let $\conv$ denote the standard vanilla convolution and $\filter$ denote the convolution filter.
        Then, the convolution of the downscaled image $\transformationMath_{\scaleNotation}(\projectionOne)$ with the filter $\filter$ is given by
        \begin{align}
            &\left[\transformationMath_{\scaleNotation}(\projectionOne)\conv\filter \right](\pixU,\pixV) \nonumber\\
            &= \int\int\projectionOne\left(\frac{\pixUTwo}{\scaleNotation}, \frac{\pixVTwo}{\scaleNotation}\right)\filter(\pixUTwo-\pixU, \pixVTwo-\pixV) d\pixUTwo d\pixVTwo \nonumber\\
            &= \scaleNotation^2 \int\int\projectionOne\left(\frac{\pixUTwo}{\scaleNotation}, \frac{\pixVTwo}{\scaleNotation}\right)\filter\left( \scaleNotation\frac{\pixUTwo-\pixU}{\scaleNotation}, \scaleNotation\frac{\pixVTwo-\pixV}{\scaleNotation}\right) d\left(\frac{\pixUTwo}{\scaleNotation}\right) d\left(\frac{\pixVTwo}{\scaleNotation} \right)\nonumber\\
            &= \scaleNotation^2 \int\int\projectionOne\left(\frac{\pixUTwo}{\scaleNotation}, \frac{\pixVTwo}{\scaleNotation}\right)\transformationMath_{\scaleNotation^{-1}}\left[\filter\left( \frac{\pixUTwo-\pixU}{\scaleNotation}, \frac{\pixVTwo-\pixV}{\scaleNotation}\right)\right] d\left(\frac{\pixUTwo}{\scaleNotation}\right) d\left(\frac{\pixVTwo}{\scaleNotation} \right)\nonumber\\
            &= \scaleNotation^2 \int\int\projectionOne\left(\frac{\pixUTwo}{\scaleNotation}, \frac{\pixVTwo}{\scaleNotation}\right)\transformationMath_{\scaleNotation^{-1}}\left[\filter\left( \frac{\pixUTwo}{\scaleNotation}-\frac{\pixU}{\scaleNotation}, \frac{\pixVTwo}{\scaleNotation}-\frac{\pixV}{\scaleNotation}\right)\right] d\left(\frac{\pixUTwo}{\scaleNotation}\right) d\left(\frac{\pixVTwo}{\scaleNotation} \right)\nonumber\\
            &= \scaleNotation^2 \left[\projectionOne \conv \transformationMath_{\scaleNotation^{-1}}(\filter)\right]\left(\frac{\pixU}{\scaleNotation},\frac{\pixV}{\scaleNotation}\right) \nonumber\\
            &= \scaleNotation^2 \transformationMath_{\scaleNotation}\left[\projectionOne \conv\transformationMath_{\scaleNotation^{-1}}(\filter)\right](\pixU,\pixV).
            \label{eq:scale_eq_orig}
        \end{align}

        Next, \cite{sosnovik2020sesn} re-parametrize the SES filters by writing $\filter_\sigma(\pixU,\pixV) = \frac{1}{\sigma^2}\filter\left(\frac{\pixU}{\sigma},\frac{\pixV}{\sigma}\right)$. 
        Substituting in \cref{eq:scale_eq_orig}, we get
        \begin{align}
            \left[\transformationMath_{\scaleNotation}(\projectionOne)\conv\filter_\sigma \right](\pixU,\pixV)
            &= \scaleNotation^2 \transformationMath_{\scaleNotation}\left[\projectionOne \conv\transformationMath_{\scaleNotation^{-1}}(\filter_\sigma)\right](\pixU,\pixV)
            \label{eq:scale_eq_interm}
        \end{align}
        
        Moreover, the re-parametrized filters are separable \cite{sosnovik2020sesn} by construction and so, one can write
        \begin{align}
            \filter_\sigma(\pixU,\pixV) &= \filter_\sigma(\pixU)\filter_\sigma(\pixV).
        \end{align}
        
        The re-parametrization and separability leads to the important property that
        \begin{align}
            \transformationMath_{\scaleNotation^{-1}}\left( \filter_\sigma(\pixU,\pixV) \right)
            &= \transformationMath_{\scaleNotation^{-1}}\left( \filter_\sigma(\pixU)\filter_\sigma(\pixV) \right) \nonumber\\
            &= \transformationMath_{\scaleNotation^{-1}}\left( \filter_\sigma(\pixU)\right)\transformationMath_{\scaleNotation^{-1}}\left( \filter_\sigma(\pixV) \right) \nonumber\\
            &= \scaleNotation^{-2} \filter_{\scaleNotation^{-1} \sigma}(\pixU)\filter_{\scaleNotation^{-1} \sigma}(\pixV) \nonumber \\
            &= \scaleNotation^{-2} \filter_{\scaleNotation^{-1} \sigma}(\pixU, \pixV).
        \end{align}
        
        Substituting above in the RHS of \cref{eq:scale_eq_interm}, we get
        \begin{align}
            \left[\transformationMath_{\scaleNotation}(\projectionOne)\conv\filter_\sigma \right](\pixU,\pixV)
            &= \scaleNotation^2 \transformationMath_{\scaleNotation}\left[\projectionOne \conv \scaleNotation^{-2} \filter_{\scaleNotation^{-1} \sigma}\right](\pixU,\pixV) \nonumber \\
            \implies \left[\transformationMath_{\scaleNotation}(\projectionOne)\conv\filter_\sigma \right](\pixU,\pixV) &= \transformationMath_{\scaleNotation}\left[\projectionOne \conv \filter_{\scaleNotation^{-1} \sigma}\right](\pixU,\pixV),
            \label{eq:scale_eq_final}
        \end{align}
        which is a cleaner form of \cref{eq:scale_eq_orig}.
        \cref{eq:scale_eq_final} says that convolving the downscaled image with a filter is same as the
        downscaling the result of convolving the image with the upscaled filter \cite{sosnovik2020sesn}. 
        This additional constraint regularizes the scale (depth) predictions for the image, leading to better generalization.
        

    \subsection{Why does \deviant generalize better compared to CNN backbone?}\label{sec:deviant_why_better_generalize}
        \deviant models the physics better compared to the CNN backbone. 
        CNN generalizes better for \twoD detection because of the \twoD translation \equivariance{} in the Euclidean manifold.
        However, monocular \threeD detection does not belong to the Euclidean manifold but is a task of the projective manifold.
        Modeling translation \equivariance{} in the correct manifold improves generalization.
        For monocular \threeD detection, we take the first step towards the general \threeD translation \equivariance{} by embedding \equivariance{} to depth translations. 
        The \threeD \depthEquivariance{} in \deviant uses \cref{eq:scale_eq_interm} and thus imposes an additional constraint on the feature maps.
        This additional constraint results in consistent depth estimates from the current image and a virtual image (obtained by translating the ego camera), and therefore, better generalization than CNNs. 
        On the other hand, CNNs, by design, do not constrain the depth estimates from the current image and a virtual image (obtained by translating the ego camera), and thus, their depth estimates are entirely data-driven. 

    \subsection{Why not Fixed Scale Assumption?}
        We now answer the question of keeping the fixed scale assumption. If we assume fixed scale assumption, then vanilla convolutional layers have the right equivariance. 
        However, we do not keep this assumption because the ego camera translates along the depth in driving scenes and also, because the depth is the hardest parameter to estimate \cite{ma2021delving} for monocular detection. 
        So, zero depth translation or fixed scale assumption is always violated.

    \subsection{Comparisons with Other Methods}
        We now list out the differences between different convolutions and monocular detection methods in \cref{tab:deviant_compare_methods}.
        Kinematic3D \cite{brazil2020kinematic} does not constrain the output at feature map level, but at system level using Kalman Filters.
        The closest to our method is the Dilated CNN (DCNN) \cite{yu2015multi}. 
        We show in \cref{tab:deviant_kitti_compare_dilation} that \deviant outperforms Dilated CNN.

        \begin{table}[!tb]
            \caption[Comparison of Methods on the basis of inputs, convolution kernels, outputs and whether output are scale-constrained.]{\textbf{Comparison of Methods} on the basis of inputs, convolution kernels, outputs and whether output are scale-constrained.
            }
            \label{tab:deviant_compare_methods}
            \centering
            \scalebox{\scaleFraction}{
                \setlength\tabcolsep{0.1cm}
                \begin{tabular}{ml m c m ccc  m cccm}
                    \myTopRule
                    \addlinespace[0.01cm]
                    \multirow{2}{*}{Method} & Input & \#Conv & \multirow{2}{*}{Output} & Output Constrained\\
                     & Frame & Kernel & & for Scales?\\
                    \myTopRule
                    Vanilla CNN & 1 & 1 & 4D & \xmark \\
                    Depth-Aware \cite{brazil2019m3d} & 1 & $>1$ & \fourD & \xmark \\
                    Dilated CNN \cite{yu2015multi} & 1 & $>1$ & \fiveD & Integer \cite{worrall2019deep}\\
                    \textbf{\deviant} & 1 & $>1$ & \fiveD & Float\\
                    \hline
                    Depth-guided\cite{ding2020learning} & 1 + Depth & 1 & \fourD & Integer \cite{worrall2019deep}\\
                    Kinematic3D \cite{brazil2020kinematic} & $>1$ & 1 & \fiveD & \xmark \\
                    \myTopRule
                \end{tabular}
            }
        \end{table}

    \subsection{Why is Depth the hardest among all parameters?}
        Images are the \twoD projections of the \threeD scene, and therefore, the depth is lost during projection.
        Recovering this depth is the most difficult to estimate, as shown in Tab.~1 of \cite{ma2021delving}.
        Monocular detection task involves estimating \threeD center, \threeD dimensions and the yaw angle. 
        The right half of Tab.~1 in \cite{ma2021delving} shows that if the ground truth \threeD center is replaced with the predicted center, the detection reaches a minimum. 
        Hence, \threeD center is the most difficult to estimate among center, dimensions and pose.
        Most monocular \threeD detectors further decompose the \threeD center into projected (\twoD) center and depth. 
        Out of projected center and depth, Tab.~1 of \cite{ma2021delving} shows that replacing ground truth depth with the predicted depth leads to inferior detection compared to replacing ground truth projected center with the predicted projected center.
        Hence, we conclude that depth is the hardest parameter to estimate.

\section{Implementation Details}\label{sec:deviant_implement_details}
    We now provide some additional implementation details for facilitating reproduction of this work.

    \begin{figure}[!t]
        \centering
        \input{images/deviant/steerable_idea}
        \caption[\ses{} convolution and \MaxScale]
        {\textbf{(a) \ses{} convolution} \cite{ghosh2019scale,sosnovik2020sesn} The non-trainable basis functions multiply with learnable weights $\weight$ to get kernels. The input then convolves with these kernels to get multi-scale \fiveD{} output. \textbf{(b) \MaxScale} \cite{sosnovik2020sesn} takes $\max$ over the scale dimension of the \fiveD{} output and converts it to \fourD. [Key: $*$ = Vanilla convolution.]}
        \label{fig:deviant_steerable_idea}
    \end{figure}
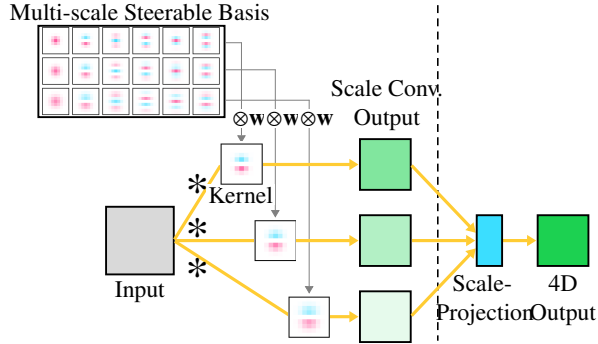

    \begin{figure}[!t]
        \centering
        \includegraphics[width=0.7\linewidth]{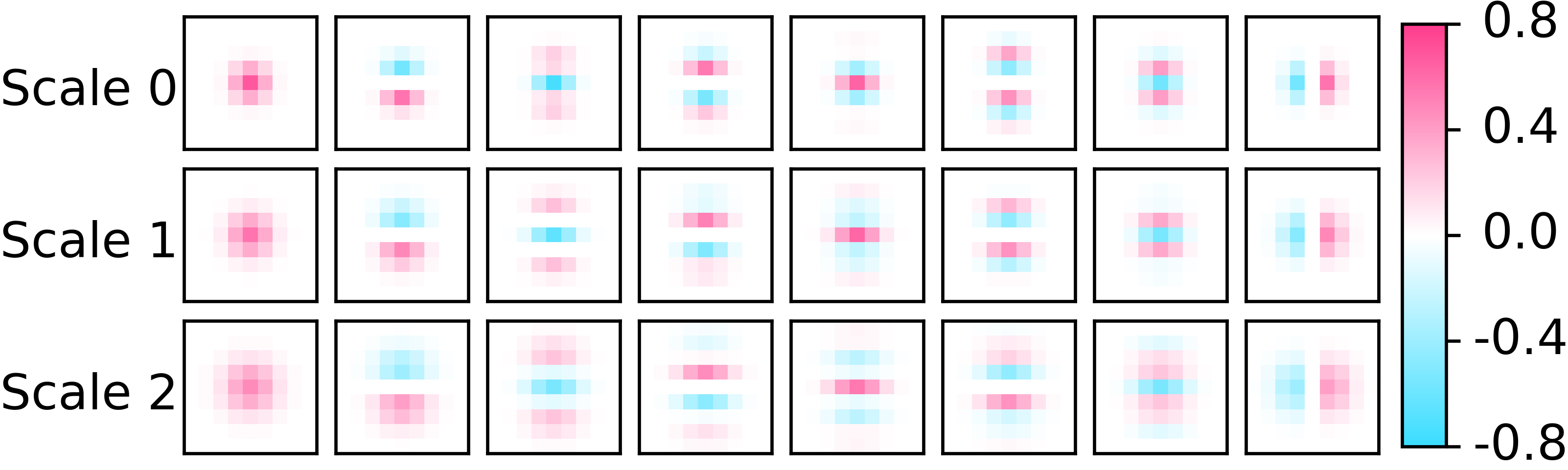}
        \caption[Steerable Basis for $7\!\times\!7$ \ses ~convolution filters.]
        {\textbf{Steerable Basis} \cite{sosnovik2020sesn} for $7\!\times\!7$ \ses ~convolution filters. (Showing only $8$ of the $49$ members for each scale).}
        \label{fig:deviant_steerable_basis}
    \end{figure}

    \subsection{Steerable Filters of \ses{} Convolution}\label{sec:deviant_steerable_additional}

        We use the \scaleEquivariant{} steerable blocks proposed by \cite{sosnovik2021siamese} for our \deviant backbone. 
        We now share the implementation details of these steerable filters.
        
        \noIndentHeading{Basis.}
        Although steerable filters can use any linearly independent functions as their basis, we stick with the Hermite polynomials as the basis \cite{sosnovik2021siamese}.
        Let $(0,0)$ denote the center of the function and $(\pixU, \pixV)$ denote the pixel coordinates. 
        Then, the filter coefficients~$\psi_{\sigma n m}$ \cite{sosnovik2021siamese} are         
        \begin{align}
            \psi_{\sigma n m} &= \frac{A}{\sigma^2} H_n \left(\frac{u}{\sigma}\right) H_m \left(\frac{v}{\sigma}\right) e^{-\frac{u^2 + v^2}{\sigma^2}} 
        \end{align}
        $H_n$ denotes the Probabilist's Hermite polynomial of the $n$th order, and $A$ is the normalization constant. 
        The first six Probabilist's Hermite polynomials are
        \begin{align}
            H_0(x) &= 1\\
            H_1(x) &= x\\
            H_2(x) &= x^2 - 1\\
            H_3(x) &= x^3 - 3x\\
            H_4(x) &= x^4 - 6x^2+3
        \end{align}
        \cref{fig:deviant_steerable_basis} visualizes some of the \ses{} filters and shows that the basis is indeed at different scales.

    \subsection{Monocular 3D Detection}\label{sec:deviant_detection_training_additional}

        \noIndentHeading{Architecture.}
            We use the \dla{} \cite{yu2018deep} configuration, with the standard Feature Pyramid Network (FPN) \cite{lin2017feature}, binning and ensemble of uncertainties.
            FPN is a bottom-up feed-forward CNN that computes feature maps with a downscaling factor of $2$, and a top-down network that brings them back to the high-resolution ones.
            There are total six feature maps levels in this FPN.
            
            We use \dla{} as the backbone for our baseline \gupNet{} \cite{lu2021geometry}, while we use \ses-\dla{} as the backbone for \deviant.
            We also replace the \twoD pools by \threeD pools with pool along the scale dimensions as $1$ for \deviant.
            
            We initialize the vanilla CNN from \imageNet{} weights. 
            For \deviant, we use the regularized least squares \cite{sosnovik2021siamese} to initialize the trainable weights in all the Hermite scales from the \imageNet{} \cite{deng2009imagenet} weights.
            Compared to initializing one of the scales as proposed in \cite{sosnovik2021siamese}, we observed more stable convergence in initializing all the Hermite scales.
            
            We output three foreground classes for \kitti{} dataset. 
            We also output three foreground classes for \waymo{} dataset ignoring the Sign class \cite{reading2021categorical}.

        \noIndentHeading{Datasets.}
            We use the publicly available \kitti{},\waymo{} and \nuscenes{} datasets for our experiments.
            \kitti{} is available at \url{http://www.cvlibs.net/datasets/kitti/eval_object.php?obj_benchmark=3d} under 
            CC BY-NC-SA 3.0 License.
            \waymo{} is available at \url{https://waymo.com/intl/en_us/dataset-download-terms/} under the Apache License, Version 2.0.
            \nuscenes{} is available at \url{https://www.nuscenes.org/nuscenes} under 
            CC BY-NC-SA 
            4.0 International Public License.

        \noIndentHeading{Augmentation.}
            Unless otherwise stated, we horizontal flip the training images with probability $0.5$, and use scale augmentation as $0.4$ as well for all the models \cite{lu2021geometry} in training.

        \noIndentHeading{Pre-processing.}
            The only pre-processing step we use is image resizing.
            \begin{itemize}
                \item \textit{\kitti{}.}
            We resize the $[370, 1242]$ sized~\kitti{} images, and bring them to the $[384, 1280]$ resolution \cite{lu2021geometry}.
            
                \item \textit{\waymo{}.}
            We resize the $[1280, 1920]$ sized~\waymo{} images, and bring them to the $[512, 768]$ resolution. This resolution preserves their aspect ratio. 
            \end{itemize}
        
        \noIndentHeading{Box Filtering.}
            We apply simple hand-crafted rules for filtering out the boxes. 
            We ignore the box if it belongs to a class different from the detection class.
            \begin{itemize}
                \item \textit{\kitti{}.} 
                We train with boxes which are atleast $2m$ distant from the ego camera, and with visibility $> 0.5$ \cite{lu2021geometry}. 
            
                \item \textit{\waymo{}.} 
                We train with boxes which are atleast $2m$ distant from the ego camera.
                The \waymo{} dataset does not have any occlusion based labels. 
                However, \waymo{} provides the number of \lidar{}  points inside each \threeD box which serves as a proxy for the occlusion.
                We train the boxes which have more than $100$ \lidar{} points for the vehicle class and have more than $50$ \lidar{} points for the cyclist and pedestrian class. 
            \end{itemize}

        \noIndentHeading{Training.}
            We use the training protocol of \gupNet{} \cite{lu2021geometry} for all our experiments.
            Training uses the Adam optimizer \cite{kingma2014adam} and weight-decay $1\times10^{-5}$ .
            Training dynamically weighs the losses using Hierarchical Task Learning (HTL)  \cite{lu2021geometry} strategy keeping $K$ as $5$ \cite{lu2021geometry}. 
            Training also uses a linear warmup strategy in the first $5$ epochs to stabilize the training.
            We choose the model saved in the last epoch as our final model for all our experiments.
            \begin{itemize}
                \item \textit{\kitti .} 
                We train with a batch size of $12$ on single Nvidia A100 (40GB) GPU for $140$ epochs. 
                Training starts with a learning rate $1.25 \times 10^{-3}$ with a step decay of $0.1$ at the $90$th and the $120$th epoch.

                \item \textit{\waymo .}
                We train with a batch size of $40$ on single Nvidia A100 (40GB) GPU for $30$ epochs because of the large size of the \waymo{} dataset.
                Training starts with a learning rate $1.25 \times 10^{-3}$ with a step decay of $0.1$ at the $18$th and the $26$th epoch.
            \end{itemize}

        \noIndentHeading{Losses.}
            We use the \gupNet{} \cite{lu2021geometry} multi-task losses before the NMS for training. The total loss $\loss$ is given by
            \begin{align}
                \loss &= \loss_\heatmap + \loss_{\twoDMath ,\offset} + \loss_{\twoDMath,\size} + \loss_{\threeDMath\twoDMath,\offset}+ \loss_{\threeDMath,angle}\nonumber\\
                &\quad\quad + \loss_{\threeDMath,l} + \loss_{\threeDMath,w} + \loss_{\threeDMath,h} + \loss_{\threeDMath,depth}.
            \end{align}
            The individual terms are given by
            \begin{align}
                \loss_\heatmap &= \text{Focal}(\class^b, \class^g), \\
                \loss_{\twoDMath ,\offset} &= \lOne (\delta_{\twoDMath}^b, \delta_{\twoDMath}^g), \\   
                \loss_{\twoDMath,\size} &= \lOne (w_{\twoDMath}^b, w_{\twoDMath}^g) + \lOne (h_{\twoDMath}^b, h_{\twoDMath}^g), \\
                \loss_{\threeDMath\twoDMath,\offset} &= \lOne(\delta_{\threeDMath\twoDMath}^b, \delta_{\threeDMath\twoDMath}^g)\\
                \loss_{\threeDMath,angle} &= \text{CE}(\alpha^b, \alpha^g) \\
                \loss_{\threeDMath,l} &= \lOne(\mu_{l{\threeDMath}}^b, \delta_{l\threeDMath}^g) \\
                \loss_{\threeDMath,w} &= \lOne(\mu_{w{\threeDMath}}^b, \delta_{w{\threeDMath}}^g) \\
                \loss_{\threeDMath,h} &= \frac{\sqrt{2}}{\sigma_{h\threeDMath}}\lOne(\mu^b_{h\threeDMath}, \delta_{h\threeDMath}^g) + \ln(\sigma_{h\threeDMath}) \\
                \loss_{\threeDMath,depth} &= \frac{\sqrt{2}}{\sigma_d}\lOne(\mu^b_{d}, \mu_{d}^g) + \ln(\sigma_d),
            \end{align}
            where, \begin{align}
                \mu^b_d &= \focal \frac{\mu_{h\threeDMath}^b}{h_{\twoDMath}^b} + \mu_{d,pred} \\
                \sigma_d &= \sqrt{ \left( \focal \frac{\sigma_{h\threeDMath}}{h_{\twoDMath}^b} \right)^2 + \sigma_{d,pred}^2 }.
            \end{align}
            
            The superscripts $b$ and $g$ denote the predicted box and ground truth box respectively. 
            CE and Focal denote the Cross Entropy and Focal loss respectively.
            
            The number of heatmaps depends on the number of output classes.
            $\delta_{\twoDMath}$ denotes the deviation of the \twoD center from the center of the heatmap.
            $\delta_{\threeDMath\twoDMath, \offset}$ denotes the deviation of the projected \threeD center from the center of the heatmap.
            The orientation loss is the cross entropy loss between the binned observation angle of the prediction and the ground truth.
            The observation angle $\alpha$ is split into $12$ bins covering $30^\circ$ range.
            $\delta_{l\threeDMath}, \delta_{w\threeDMath}$ and $\delta_{h\threeDMath}$ denote the deviation of the \threeD length, width and height of the box from the class dependent mean size respectively.
            
            The depth is the hardest parameter to estimate\cite{ma2021delving}. 
            So, \gupNet{} uses in-network ensembles to predict the depth. 
            It obtains a Laplacian estimate of depth from the \twoD height, while it obtains another estimate of depth from the prediction of depth. 
            It then adds these two depth estimates.
    
        \noIndentHeading{Inference.}\label{sec:deviant_det_testing_additional}
            Our testing resolution is same as the training resolution. 
            We do not use any augmentation for test/validation.
            We keep the maximum number of objects to $50$ in an image, and we multiply the class and predicted confidence to get the box's overall score in inference as in \cite{kumar2021groomed}.
            We consider output boxes~with scores greater than a threshold of $0.2$ for \kitti{} \cite{lu2021geometry} and $0.1$ for \waymo{} \cite{reading2021categorical}.

\section{Additional Experiments and Results}\label{sec:deviant_additional_exp}
    We now provide additional details and results of the experiments evaluating \deviant's performance.

    \subsection{\kitti{} \val Split}

        \begin{table*}[!tb]
            \caption[Generalization gap on~\kitti{} \val cars.]{\textbf{Generalization gap} (\downarrowRHDSmall) on~\kitti{} \val cars. Monocular detection has huge generalization gap between training and inference sets. [Key: \bestKey{Best}]}
            \label{tab:deviant_kitti_compare_generalization_big}
            \centering
            \scalebox{0.78}{
                \setlength{\tabcolsep}{0.1cm}
                \begin{tabular}{tl|c m c m ccc t ccc m ccc t ccct}
                    \myTopRule
                    \addlinespace[0.01cm]
                    \multirow{3}{*}{Method} & Scale  & \multirow{3}{*}{Set} & \multicolumn{6}{cm}{\iouThreeD{} $\geq 0.7$} & \multicolumn{6}{ct}{\iouThreeD{} $\geq 0.5$}\\
                    \cline{4-15}
                    & Eqv & & \multicolumn{3}{ct}{\apThreeDForty \bracketPercentage(\uparrowRHDSmall)} & \multicolumn{3}{cm}{\apBevForty \bracketPercentage(\uparrowRHDSmall)} & \multicolumn{3}{ct}{\apThreeDForty \bracketPercentage(\uparrowRHDSmall)} & \multicolumn{3}{ct}{\apBevForty \bracketPercentage(\uparrowRHDSmall)}\\
                    & & & Easy & Mod & Hard & Easy & Mod & Hard & Easy & Mod & Hard & Easy & Mod & Hard\\
                    \myTopRule
                    \multirow{3}{*}{\gupNet{} \cite{lu2021geometry}} & & Train & $91.83$ & $74.87$ & $67.43$ & $95.19$ & $80.95$ & $73.55$ & $99.50$ & $93.62$ & $86.22$ & $99.56$ & $93.88$ & $86.46$\\
                    & & Val & $21.10$ & $15.48$ & $12.88$ & $28.58$ & $20.92$ & $17.83$ & $58.95$ & $43.99$ & $38.07$ & $64.60$ & $47.76$ &	$42.97$\\
                    \cline{4-15}
                    & & \COFix Gap & \COFix$70.73$ & \COFix\best{59.39} & \COFix$54.55$ & \COFix$66.61$ & \COFix$60.03$ & \COFix$55.72$ & \COFix$40.55$ & \COFix$49.63$ & \COFix\best{48.15} & \COFix$34.96$ & \COFix$46.12$ & \COFix\best{43.49}\\
                    \myTopRule
                    \multirow{3}{*}{\textbf{\deviant}}  & \multirow{3}{*}{\checkmark} & Train & $91.09$ & $76.19$ & $67.16$ & $94.76$ & $82.61$ & $75.51$ & $99.37$ & $93.56$ & $88.57$ & $99.50$ & $93.87$ & $88.90$\\
                    & & Val & $24.63$ & $16.54$ & $14.52$ & $32.60$ & $23.04$ & $19.99$ & $61.00$ & $46.00$ & $40.18$ & $65.28$ & $49.63$ & $43.50$\\
                    \cline{4-15}
                    & & \COFix Gap & \COFix\best{66.46} & \COFix$59.65$ & \COFix\best{52.64} & \COFix\best{62.16} & \COFix\best{59.57} & \COFix\best{55.52} & \COFix\best{38.37} & \COFix\best{47.56} & \COFix$48.39$ & \COFix\best{34.22} & \COFix\best{44.24} & \COFix$45.40$\\
                    \myTopRule
                \end{tabular}
            }
        \end{table*}
    
        \noIndentHeading{Monocular Detection has Huge Generalization Gap.}
            As mentioned in \cref{sec:deviant_intro}, we now show that the monocular detection has huge generalization gap between training and inference.
            We report the object detection performance on the train and validation (val) set for the two models on \kitti{} \val split in \cref{tab:deviant_kitti_compare_generalization_big}. 
            \cref{tab:deviant_kitti_compare_generalization_big} shows that the performance of our baseline \gupNet{} \cite{lu2021geometry} and our \deviant is huge on the training set, while it is less than one-fourth of the train performance on the val set.
            
            We also report the {generalization~gap} (in pink) metric \cite{wu2021rethinking} in \cref{tab:deviant_kitti_compare_generalization_big}, which is the difference between training and validation performance.
            The generalization gap at both the thresholds of $0.7$ and $0.5$ is huge.
    
        \begin{table}[!tb]
            \caption[Comparison on multiple backbones on \kitti{} \val cars.]{\textbf{Comparison on multiple backbones} on \kitti{} \val cars. [Key: \bestKey{Best}]}
            \label{tab:deviant_kitti_compare_backbones}
            \centering
            \scalebox{0.78}{
                \setlength\tabcolsep{0.1cm}
                \begin{tabular}{ml m c m ccc t ccc m ccc t cccm}
                    \myTopRule
                    \addlinespace[0.01cm]
                    \multirow{3}{*}{BackBone} & \multirow{3}{*}{Method} & \multicolumn{6}{cm}{\iouThreeD{} $\geq 0.7$} & \multicolumn{6}{cm}{\iouThreeD{} $\geq 0.5$}\\
                    \cline{3-14}
                    & & \multicolumn{3}{ct}{\apThreeDForty \bracketPercentage(\uparrowRHDSmall)} & \multicolumn{3}{cm}{\apBevForty \bracketPercentage(\uparrowRHDSmall)} & \multicolumn{3}{ct}{\apThreeDForty \bracketPercentage(\uparrowRHDSmall)} & \multicolumn{3}{cm}{\apBevForty \bracketPercentage(\uparrowRHDSmall)}\\
                    & & Easy & Mod & Hard & Easy & Mod & Hard & Easy & Mod & Hard & Easy & Mod & Hard\\ 
                    \myTopRule
                    \resNetEighteen & \gupNet\!\cite{lu2021geometry}& $18.86$ & $13.20$ & $11.01$ & $26.05$ & $19.37$ & $16.57$ & $54.90$ & $40.65$ & $34.98$ & $60.54$ & $46.13$ & $40.12$\\
                    & \bestKey{\deviant} & $20.27$ & $14.21$ & $12.56$ & $28.09$ & $20.32$ & $17.49$ & $55.75$ & $42.41$ & $36.97$ & $60.82$ & $46.43$ & $40.59$\\
                    \myTopRule
                    \dla &\gupNet\!\cite{lu2021geometry} & $21.10$ & $15.48$ & $12.88$ & $28.58$ & $20.92$ & $17.83$ & $58.95$ & $43.99$ & $38.07$ & $64.60$ & $47.76$ &	$42.97$\\
                    & \bestKey{\deviant} & \best{24.63} & \best{16.54} & \best{14.52} & \best{32.60} &	\best{23.04} & \best{19.99} &	\best{61.00} & \best{46.00}	& \best{40.18} & \best{65.28} & \best{49.63} & \best{43.50} \\
                    \myTopRule
                \end{tabular}
            }
        \end{table}

        \noIndentHeading{Comparison on Multiple Backbones.}
            A common trend in \twoD object detection community is to show improvements on multiple backbones \cite{wang2020scale}. 
            \ddThreeD \cite{park2021pseudo} follows this trend and also reports their numbers on multiple backbones.
            Therefore, we follow the same and compare with our baseline on multiple backbones on~\kitti{} \val cars in \cref{tab:deviant_kitti_compare_backbones}.
            \cref{tab:deviant_kitti_compare_backbones} shows that \deviant shows consistent improvements over \gupNet{} \cite{lu2021geometry} in \threeD object detection on multiple backbones, proving the effectiveness of our proposal.
               
        \noIndentHeading{Comparison with Bigger CNN Backbones.}
            Since the SES blocks increase the Flop counts significantly compared to the vanilla convolution block, we next compare \deviant with bigger CNN backbones with comparable GFLOPs and FPS/ wall-clock time (instead of same configuration) in \cref{tab:deviant_detection_with_bigger_cnn}.
            We compare \deviant with \dlaOneZeroTwo{} and \dlaOneSixNine{} - two biggest DLA networks with \imageNet{} weights\footnote{Available at \url{http://dl.yf.io/dla/models/imagenet/}} on \kitti{} \val split. 
            We use the fvcore library\footnote{\url{https://github.com/facebookresearch/fvcore}} to get the parameters and flops.
            \cref{tab:deviant_detection_with_bigger_cnn} shows that \deviant again outperforms the bigger CNN backbones, especially on nearby objects. 
            We believe this happens because the bigger CNN backbones have more trainable parameters than \deviant, which leads to overfitting.
            Although \deviant takes more time compared to the CNN backbones, \deviant still keeps the inference almost real-time.
            
        \begin{table}[!tb]
            \caption[Results with bigger CNNs having similar flops on \kitti{} \val cars.]{\textbf{Results with bigger CNNs having similar flops} on \kitti{} \val cars. 
            [Key: \firstKey{Best}]
            }
            \label{tab:deviant_detection_with_bigger_cnn}
            \centering
            \scalebox{0.75}{
                \setlength\tabcolsep{1.00pt}
                \begin{tabular}{tl m c m c m c m c m c m ccc m ccct}
                    \myTopRule
                    \multirow{2}{*}{Method} & \multirow{2}{*}{BackBone} & Param (\downarrowRHDSmall) & Disk Size (\downarrowRHDSmall) & Flops (\downarrowRHDSmall) & Infer (\downarrowRHDSmall)& \multicolumn{3}{cm}{\apThreeD{} \iouThreeD{}$\ge 0.7$ (\uparrowRHDSmall)} & \multicolumn{3}{ct}{\apThreeD{} \iouThreeD{}$\ge 0.5$ (\uparrowRHDSmall)}\\ 
                    & & (M) & (MB) & (G) & (ms) & Easy & Mod & Hard & Easy & Mod & Hard\\
                    \myTopRule
                    \gupNet{} \cite{lu2021geometry} & \dlaThirtyFour & \first{16} & \first{235} & \first{30} & \first{20} & $21.10$ & $15.48$ & $12.88$ & $58.95$ & $43.99$ & $38.07$\\
                    \gupNet{} \cite{lu2021geometry} & \dlaOneZeroTwo & $34$ & $583$ & $70$ & $25$ & $20.96$ & $14.64$ & $12.80$ & $57.06$	& $41.78$	& $37.26$\\
                    \gupNet{} \cite{lu2021geometry} & \dlaOneSixNine & $54$ & $814$ & $114$ & $30$ & $21.76$ & $15.35$ & $12.72$ & $57.60$	& $43.27$ & $37.32$\\
                    \hline
                    \deviant & SES-\dlaThirtyFour & \first{16} & $236$ & $235$ & $40$ & \first{24.63} & \first{16.54} & \first{14.52} & \first{61.00} & \first{46.00}	& \first{40.18}\\
                    \myTopRule
                \end{tabular}
            }
        \end{table}

        \noIndentHeading{Performance on Cyclists and Pedestrians.}
            \cref{tab:deviant_detection_results_kitti_valone_ped_cyclist} lists out the results of \threeD object detection on \kitti{} \val Cyclist and Pedestrians. 
            The results show that \deviant is competitive on challenging Cyclist and achieves \sota{} results on Pedestrians on the \kitti{} \val split.

        \begin{table}[!tb]
            \caption[\kitti{} \val cyclists and pedestrians results.]{\textbf{Results on \kitti{} \val cyclists and pedestrians} (Cyc/Ped) (\iouThreeD$\geq\!0.5$). [Key: \firstKey{Best}, \secondKey{Second Best}]
            }
            \label{tab:deviant_detection_results_kitti_valone_ped_cyclist}
            \centering
            \scalebox{\scaleFraction}{
                \setlength\tabcolsep{0.1cm}
                \begin{tabular}{ml m c m ccc  m cccm}
                    \myTopRule
                    \addlinespace[0.01cm]
                    \multirow{2}{*}{Method} & \multirow{2}{*}{Extra} &\multicolumn{3}{cm}{Cyc \apThreeDForty \bracketPercentage(\uparrowRHDSmall)} & \multicolumn{3}{cm}{Ped \apThreeDForty \bracketPercentage(\uparrowRHDSmall)}\\ 
                    & & Easy & Mod & Hard & Easy & Mod & Hard\\ 
                    \myTopRule
                    \groomedNMS \cite{kumar2021groomed} & \mathDash{}& $0.00$ & $0.00$ & $0.00$ & $3.79$ & $2.71$ & $2.61$\\
                    MonoDIS \cite{simonelli2019disentangling}& \mathDash{}& $1.52$ & $0.73$ & $0.71$ & $3.20$ & $2.28$ & $1.71$\\ 
                    \monoDISMulti \cite{simonelli2020disentangling}& \mathDash{}   & $2.70$        & $1.50$        & $1.30$    & $9.50$        & $7.10$        & $5.70$          \\
                    \gupNet{} (Retrained) \cite{lu2021geometry} & \mathDash{}  & \first{4.41} & \second{2.17} & \second{2.03} & \second{9.37} & \second{6.84} & \first{5.73} \\
                    \hline	 
                    \rowcolor{white}
                    \bestKey{\deviant (Ours)}                        & \mathDash{}  & \second{4.05} & \first{2.20} & \first{2.14}  & \first{9.85} & \first{7.18} & \second{5.42} \\
                    \myTopRule
                \end{tabular}
            }
        \end{table}

        \noIndentHeading{Cross-Dataset Evaluation Details.}\label{sec:deviant_results_cross_dataset_additional}
            For cross-dataset evaluation, we test on all $3{,}769$ images of the \kitti{} \val split, as well as all frontal $6{,}019$ images of the \nuscenes{} \val{} split \cite{caesar2020nuscenes}, as in \cite{shi2021geometry}.
            We first convert the \nuscenes{} \val{} images to the KITTI format using the \texttt{export\_kitti}\footnote{\url{https://github.com/nutonomy/nuscenes-devkit/blob/master/python-sdk/nuscenes/scripts/export\_kitti.py}} function in the nuscenes devkit.
            We keep \kitti{} \val images in the $[384, 1280]$ resolution, while we keep the \nuscenes{} \val{} images in the $[384, 672]$ resolution to preserve the aspect ratio. 
            For \mthreeDRPN \cite{brazil2019m3d}, we bring the \nuscenes{} \val{} images in the $[512, 910]$ resolution.
            
            Monocular \threeD object detection relies on the camera focal length to back-project the projected centers into the \threeD space. 
            Therefore, the \threeD centers depends on the focal length of the camera used in the dataset. 
            Hence, one should take the camera focal length into account while doing cross-dataset evaluation. 
            We now calculate the camera focal length of a dataset as follows. 
            We take the camera matrix $\projectionOperator$ and calculate the normalized focal length $\focalNormalized = \frac{2 f_y}{H}$, where $H$ denotes the height of the image. 
            The normalized focal length $\focalNormalized$ for the \kitti{} dataset is $3.82$, while the normalized focal length $\focalNormalized$ for the \nuscenes{} dataset is $2.82$.
            Thus, the \kitti{} and the \nuscenes{} images have a different focal length \cite{wang2020train}.
            
            \mthreeDRPN \cite{brazil2019m3d} does not normalize w.r.t. the focal length. 
            So, we explicitly correct and divide the depth predictions of \nuscenes{} images from the \kitti{} model by $3.82/2.82= 1.361$ in the  \mthreeDRPN \cite{brazil2019m3d} codebase.
            The \gupNet{} \cite{lu2021geometry} and \deviant codebases use normalized coordinates \thatIs{} they normalize w.r.t. the focal length. 
            So, we do not explicitly correct the focal length for \gupNet{} and \deviant predictions.
            
            We match predictions to the ground truths using the \iouTwoD{} overlap threshold of $0.7$ \cite{shi2021geometry}. 
            After this matching, we calculate the Mean Average Error (MAE) of the depths of the predicted and the ground truth boxes \cite{shi2021geometry}.

        \noIndentHeading{Stress Test with Rotational and/or xy-translation Ego Movement.} 
            \cref{th:projective_scaled} uses translation along the depth as the sole ego movement. 
            This assumption might be valid for the current outdoor datasets and benchmarks, but is not the case in the real world. 
            Therefore, we conduct stress tests on how tolerable \deviant and \gupNet{} \cite{lu2021geometry} are when there is rotational and/or $xy$-translation movement on the vehicle.
    
            First, note that \kitti{} and \waymo{} are already large-scale real-world datasets, and our own dataset might not be a good choice.
            So, we stick with \kitti{} and \waymo{} datasets.
            We manually choose $306$ \kitti{} \val images with such ego movements and again compare performance of \deviant and \gupNet{} on this subset in \cref{tab:deviant_detection_with_rotational_movement}.
            The average distance of the car in this subset is $27.69$ m $(\pm 16.59$ m$)$, which suggests a good variance and unbiasedness in the subset.
            \cref{tab:deviant_detection_with_rotational_movement} shows that both the \deviant backbone and the CNN backbone show a drop in the detection performance by about $4$ AP points on the Mod cars of ego-rotated subset compared to the all set. 
            This drop experimentally confirms the theory that both the \deviant backbone and the CNN backbone do not handle arbitrary \threeD rotations. 
            More importantly, the table shows that \deviant maintains the performance improvement over \gupNet{} \cite{lu2021geometry} under such movements.

            Also, \waymo{} has many images in which the ego camera shakes.
            Improvements on \waymo{} (\cref{tab:deviant_waymo_val}) also confirms that \deviant{} outperforms \gupNet{} \cite{lu2021geometry} even when there is rotational or $xy$-translation ego movement.

        \begin{table}[!tb]
            \caption[Stress Test with rotational and $xy$-translation ego movement on \kitti{} \val cars.]
            {\textbf{Stress Test} with rotational and $xy$-translation ego movement on \kitti{} \val cars.
            [Key: \firstKey{Best}]
            }
            \label{tab:deviant_detection_with_rotational_movement}
            \centering
            \centering
            \scalebox{\scaleFraction}{
                \setlength\tabcolsep{0.1cm}
                \begin{tabular}{tl m l m ccc  m ccct}
                    \myTopRule
                    \multirow{2}{*}{Set }  & \multirow{2}{*}{Method }  & \multicolumn{3}{cm}{\apThreeD{} \iouThreeD{}$\ge 0.7$ (\uparrowRHDSmall)} & \multicolumn{3}{ct}{\apThreeD{} \iouThreeD{}$\ge 0.5$ (\uparrowRHDSmall)}\\ 
                    & & Easy & Mod & Hard & Easy & Mod & Hard\\ 
                    \myTopRule
                    Subset & \gupNet{} \cite{lu2021geometry} & $17.22$ & $11.43$ & $9.91$ & $47.47$ & $35.02$ & $32.63$\\
                    $(306)$ & \deviant{} & \first{20.17} & \first{12.49} & \first{10.93} & \first{49.81} & \first{36.93} & \first{34.32}\\
                    \myTopRule
                    \kitti{} \val & \gupNet{} \cite{lu2021geometry} & $21.10$ & $15.48$ & $12.88$ & $58.95$ & $43.99$ & $38.07$\\
                    $(3769)$ & \deviant{} & \first{24.63} & \first{16.54} & \first{14.52} & \first{61.00} & \first{46.00}	& \first{40.18}\\
                    \myTopRule
                \end{tabular}
            }
        \end{table}

        \noIndentHeading{Comparison of Depth Estimates from Monocular Depth Estimators and 3D Object Detectors.}
            We next compare the depth estimates from monocular depth estimators and depth estimates from monocular \threeD object detectors on the foreground objects. 
            We take a monocular depth estimator BTS \cite{lee2019big} model trained on \kitti{} Eigen split. 
            We next compare the depth error for all and foreground objects (cars) on \kitti{} \val split using MAE (\downarrowRHDSmall) metric in \cref{tab:deviant_detection_depth_estimator_depth} as in \cref{tab:deviant_detection_cross_dataset}.
            We use the MSeg \cite{lambert2020mseg} to segment out cars in the driving scenes for BTS.
            \cref{tab:deviant_detection_depth_estimator_depth} shows that the depth from BTS is not good for foreground objects (cars) beyond $20+$ m range.
            Note that there is a data leakage issue between the \kitti{} Eigen train split and the \kitti{} \val split \cite{simonelli2021we} and therefore, we expect more degradation in performance of monocular depth estimators after fixing the data leakage issue.

        \begin{table}[!tb]
            \caption[Comparison of Depth Estimates of monocular depth estimators and 3D object detectors on \kitti{} \val cars.]{\textbf{Comparison of Depth Estimates} of monocular depth estimators and 3D object detectors on \kitti{} \val cars. Depth from a depth estimator BTS is not good for foreground objects (cars) beyond $20+$ m range.
            [Key: \firstKey{Best}, \secondKey{Second Best}]
            }
            \label{tab:deviant_detection_depth_estimator_depth}
            \centering
            \scalebox{\scaleFraction}{
                \setlength\tabcolsep{0.1cm}
                \begin{tabular}{tl m cc m ccc  m ccct}
                    \myTopRule
                    \multirow{2}{*}{Method} & Depth & Ground & \multicolumn{3}{cm}{Back+ Foreground} & \multicolumn{3}{ct}{Foreground (Cars)}\\
                    & at & Truth & $0\!-\!20$ & $20\!-\!40$ & $40\!-\!\infty$ &$0\!-\!20$ & $20\!-\!40$ & $40\!-\!\infty$\\ 
                    \myTopRule
                    \gupNet{} \cite{lu2021geometry} & \threeD Center & \threeD Box & \mathDash & \mathDash & \mathDash & $0.45$ & \second{1.10} & \second{1.85}\\
                    \deviant & \threeD Center & \threeD Box & \mathDash & \mathDash & \mathDash & \second{0.40} & \first{1.09} & \first{1.80}\\
                    \hline
                    BTS \cite{lee2019big} & Pixel & \lidar{} &  $0.48$ & $1.30$ & $1.83$ & \first{0.30} & $1.22$ & $2.16$\\
                    \myTopRule
                \end{tabular}
            }
        \end{table}

        \begin{figure}[!tb]
            \centering
            \includegraphics[width=0.45\linewidth]{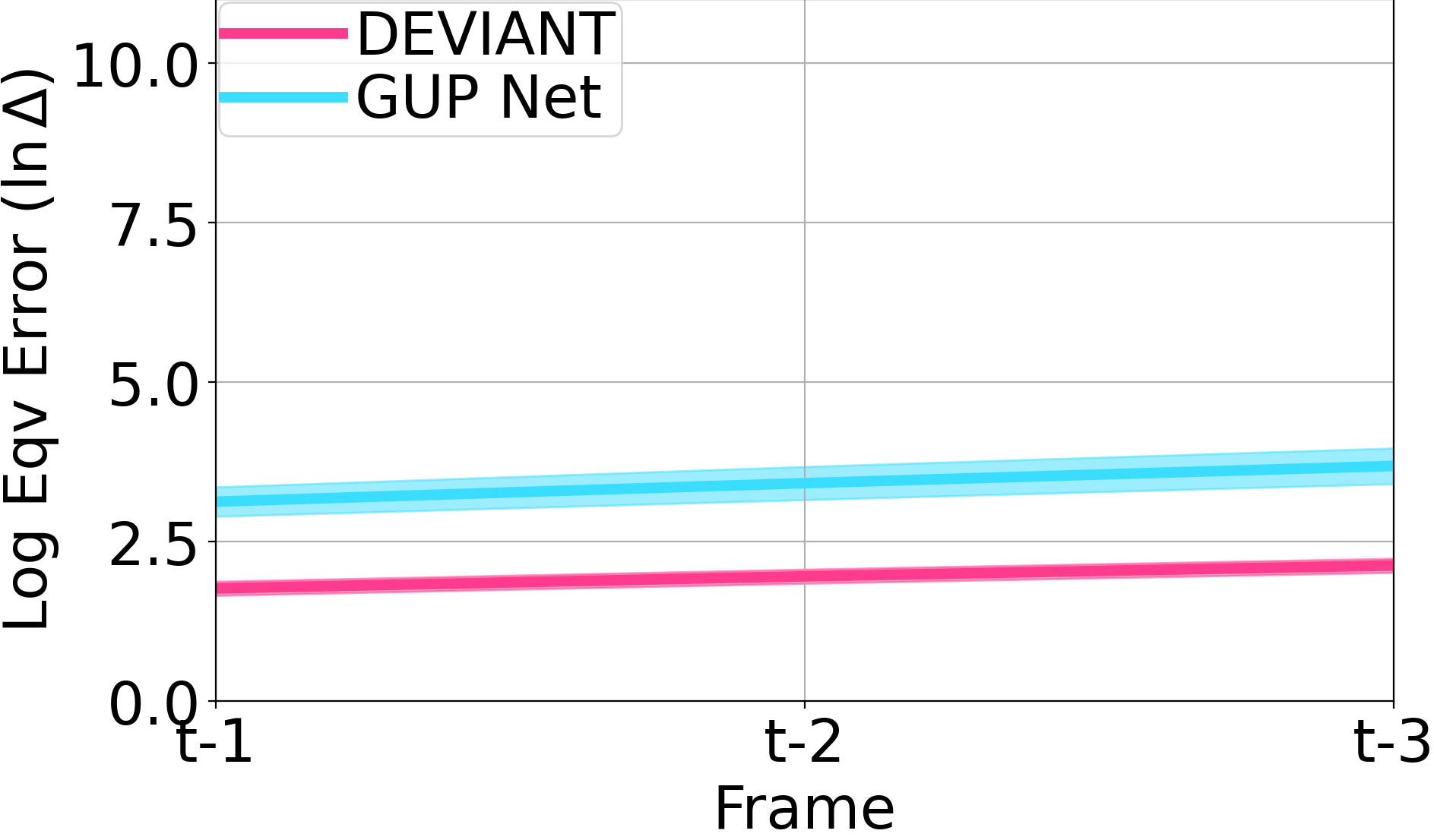}
            \caption[\Equivariance{} error $(\Delta)$ comparison for \deviant and \gupNet{} on previous three frames of the \kitti{} monocular videos at block $3$ in the backbone.]
            {\textbf{\Equivariance{} error $(\Delta)$} comparison for \deviant and \gupNet{} on previous three frames of the \kitti{} monocular videos at block $3$ in the backbone.}
            \label{fig:deviant_equiv_error_mono_videos}
        \end{figure}

        \noIndentHeading{\Equivariance{} Error for \kitti{} Monocular Videos.}
            A better way to compare the scale \equivariance{} of the \deviant and \gupNet{} \cite{lu2021geometry} compared to \cref{fig:deviant_equiv_error}, is to compare \equivariance{} error on real images with depth translations of the ego camera.
            The \equivariance{} error $\Delta$ is the normalized difference between the scaled feature map and the feature map of the scaled image, and is given by
            \begin{align}
                \Delta &= \frac{1}{N} \sum_{i=1}^N \frac{||\transformationMath_{s_i} \mapping(\projectionOneIndexed) - \mapping(\transformationMath_{s_i} \projectionOneIndexed)||_2^2}{||\transformationMath_{s_i} \mapping(\projectionOneIndexed)||_2^2},
                \label{eq:equiv_error}
            \end{align}
            where $\mapping$ denotes the neural network, $\transformationMath_{s_i}$ is the scaling transformation for the image $i$, and $N$ is the total number of images.
            Although we do evaluate this error in \cref{fig:deviant_equiv_error}, the image scaling in \cref{fig:deviant_equiv_error} does not involve scene change because of the absence of the moving objects.
            Therefore, evaluating on actual depth translations of the ego camera makes the \equivariance{} error evaluation more realistic.
            We next carry out this experiment and report the \equivariance{} error on three previous frames of the val images of the \kitti{} \val split as in \cite{brazil2020kinematic}.
            We plot this \equivariance{} error in \cref{fig:deviant_equiv_error_mono_videos} at block $3$ of the backbones because the resolution at this block corresponds to the output feature map of size $[96, 320]$.
            \cref{fig:deviant_equiv_error_mono_videos} is similar to \cref{fig:deviant_equiv_error_scaling}, and shows that \deviant achieves lower \equivariance{} error.
            Therefore, \deviant has better \equivariance{} to depth translations (scale \transformation{} s) than \gupNet{}\cite{lu2021geometry} in real scenarios.

        \noIndentHeading{Model Size, Training, and Inference Times.}
            Both \deviant and the baseline \gupNet{} have the same number of trainable parameters, and therefore, the same model size. 
            \gupNet{} takes $4$ hours to train on \kitti{} \val and $0.02$ ms per image for inference on a single Ampere A100 (40 GB) GPU.
            \deviant takes $8.5$ hours for training and $0.04$ ms per image for inference on the same GPU.
            This is expected because \se{} models use more flops \cite{zhu2019scale, sosnovik2020sesn} and, therefore, \deviant takes roughly twice the training and inference time as \gupNet.

        \noIndentHeading{Reproducibility.}
            As described in \cref{sec:deviant_detection_results_kitti_val1}, we now list out the five runs of our baseline \gupNet{} \cite{lu2021geometry} and \deviant in \cref{tab:deviant_runs_results_kitti_val1}.
            \cref{tab:deviant_runs_results_kitti_val1} shows that \deviant outperforms \gupNet{} in all runs and in the average run.

        \begin{table}[!tb]
            \caption[Five Different Runs on \kitti{} \val cars.]{\textbf{Five Different Runs} on \kitti{} \val cars. 
            [Key: \bestKey{Average}]
            }
            \label{tab:deviant_runs_results_kitti_val1}
            \centering
            \scalebox{\scaleFraction}{
            \setlength\tabcolsep{0.1cm}
            \begin{tabular}{ml m c m ccc t ccc m ccc t cccm}
                \myTopRule
                \addlinespace[0.01cm]
                \multirow{3}{*}{Method} & \multirow{3}{*}{Run} & \multicolumn{6}{cm}{\iouThreeD{} $\geq 0.7$} & \multicolumn{6}{cm}{\iouThreeD{} $\geq 0.5$}\\\cline{3-14}
                & & \multicolumn{3}{ct}{\apThreeDForty \bracketPercentage(\uparrowRHDSmall)} & \multicolumn{3}{cm}{\apBevForty \bracketPercentage(\uparrowRHDSmall)} & \multicolumn{3}{ct}{\apThreeDForty \bracketPercentage(\uparrowRHDSmall)} & \multicolumn{3}{cm}{\apBevForty \bracketPercentage(\uparrowRHDSmall)}\\
                & & Easy & Mod & Hard & Easy & Mod & Hard & Easy & Mod & Hard & Easy & Mod & Hard\\
                \myTopRule
                & 1 & $21.67$ & $14.75$ & $12.68$ & $28.72$ & $20.88$ & $17.79$ & $58.27$ & $43.53$ & $37.62$ & $63.67$ & $47.37$ & $42.55$\\
                & 2& $21.26$ & $14.94$ & $12.49$ & $28.39$ & $20.40$ & $17.43$ & $59.20$ & $43.55$ & $37.63$ & $64.06$ & $47.46$ & $42.67$\\
                \gupNet{} \cite{lu2021geometry} 
                & 3 & $20.87$ & $15.03$ & $12.61$ & $28.66$ & $20.56$ & $17.48$ & $60.19$ & $44.08$ & $39.36$ & $65.26$ & $49.44$ & $43.17$\\
                & 4 & $21.10$ & $15.48$ & $12.88$ & $28.58$ & $20.92$ & $17.83$ & $58.95$ & $43.99$ & $38.07$ & $64.60$ & $47.76$ &	$42.97$ \\
                & 5 & $22.52$ & $15.92$ & $13.31$ & $30.77$ & $22.40$ & $19.36$ & $59.91$ & $44.00$ & $39.30$ & $64.94$ & $48.01$ & $43.08$\\
                \cline{2-14}
                & Avg & \best{21.48} & \best{15.22} & \best{12.79} & \best{29.02} & \best{21.03} & \best{17.98} & \best{59.30} & \best{43.83} & \best{38.40} & \best{64.51} & \best{48.01} & \best{42.89}\\
                \myTopRule
                & 1 & $23.19$ & $15.84$ & $14.11$ & $29.82$ & $21.93$ & $19.16$ & $60.19$ & $45.52$ & $39.86$ & $66.32$ & $49.39$ & $43.38$\\
                & 2 & $23.33$ & $16.12$ & $13.54$ & $31.22$ & $22.64$ & $19.64$ & $61.59$ & $46.33$ & $40.35$ & $67.49$ & $50.26$ & $43.98$\\
                \deviant & 3 & $24.12$ & $16.37$ & $14.48$ & $31.58$ & $22.52$ & $19.65$ & $62.51$ & $46.47$ & $40.65$ & $67.33$ & $50.24$ & $44.16$\\ 
                & 4 & $24.63$ & $16.54$ & $14.52$ & $32.60$ &	$23.04$ & $19.99$ &	$61.00$ & $46.00$	& $40.18$ & $65.28$ & $49.63$ & $43.50$ \\
                & 5 & $25.82$ & $17.69$ & $15.07$ & $33.63$ & $23.84$ & $20.60$ & $62.39$ & $46.46$ & $40.61$ & $67.55$ & $50.51$ & $45.80$\\
                \cline{2-14}
                & Avg & \best{24.22} & \best{16.51} & \best{14.34} & \best{31.77} & \best{22.79} & \best{19.81}  & \best{61.54} & \best{46.16} & \best{40.33} & \best{66.79} & \best{50.01} & \best{44.16}\\
                \myTopRule
            \end{tabular}
            }
        \end{table}

        \noIndentHeading{Experiment Comparison.}
            We now compare the experiments of different chapters in \cref{tab:deviant_expt_comparison}.
            To the best of our knowledge, 
            the experimentation in \deviant is more than the experimentation of most monocular \threeD object detection chapters.

        \begin{table}[!tb]
            \caption[Experiments Comparison.]{\textbf{Experiments Comparison}.
            }
            \label{tab:deviant_expt_comparison}
            \centering
            \scalebox{\scaleFraction}{
                \setlength\tabcolsep{0.1cm}
                \begin{tabular}{m l m c m ccc m}
                    \myTopRule
                    Method & Venue & Multi-Dataset & Cross-Dataset & Multi-Backbone\\
                    \myTopRule
                    \groomedNMS \cite{kumar2021groomed} & CVPR21 & \mathDash{} & \mathDash{} & \mathDash{}\\
                    MonoFlex \cite{zhang2021objects} & CVPR21 & \mathDash{} & \mathDash{} & \mathDash{}\\
                    \caddn \cite{reading2021categorical} & CVPR21 & \cmark & \mathDash{} & \mathDash{}\\
                    \hline
                    MonoRCNN \cite{shi2021geometry} & ICCV21 & \mathDash{} & \cmark & \mathDash{}\\
                    \gupNet{} \cite{lu2021geometry} & ICCV21 & \mathDash{} & \mathDash{} & \mathDash{}\\
                    \ddThreeD \cite{park2021pseudo} & ICCV21 & \cmark & \mathDash{} & \cmark\\
                    \hline
                    PCT \cite{wang2021progressive} & NeurIPS21 & \cmark & \mathDash{} & \cmark\\
                    MonoDistill \cite{chong2022monodistill} & ICLR22 & \mathDash{} & \mathDash{} & \mathDash{}\\
                    \hline
                    \monoDISMulti \cite{simonelli2020disentangling} & TPAMI20 & \cmark & \mathDash{} & \mathDash{}\\
                    MonoEF \cite{zhou2021monoef} & TPAMI21 & \cmark & \mathDash{} & \mathDash{}\\
                    \hline
                    \best{\deviant} & - & \cmark & \cmark & \cmark\\
                    \myTopRule
                \end{tabular}
            }
        \end{table}

    \subsection{Qualitative Results}

\begin{figure}[!tb]
            \centering
            \begin{subfigure}{\figureScaleFraction\linewidth}
                \includegraphics[width=0.9\linewidth]{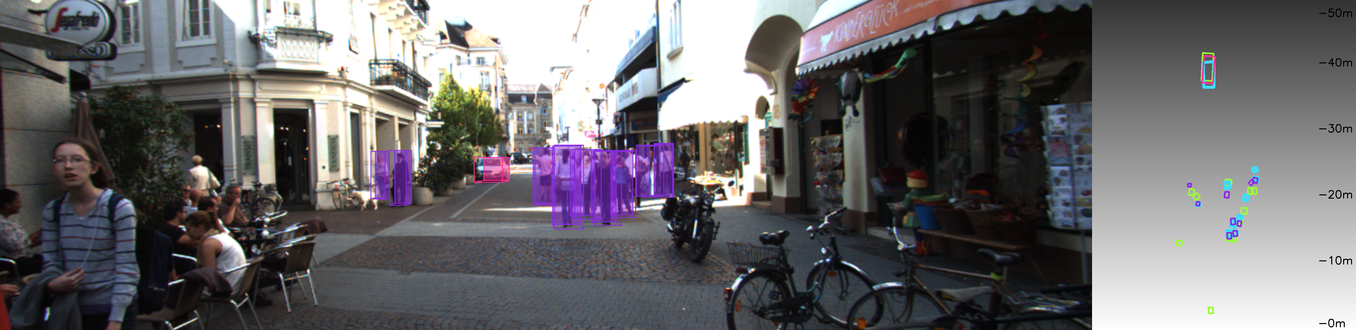}
            \end{subfigure}
            \begin{subfigure}{\figureScaleFraction\linewidth}
                \includegraphics[width=0.9\linewidth]{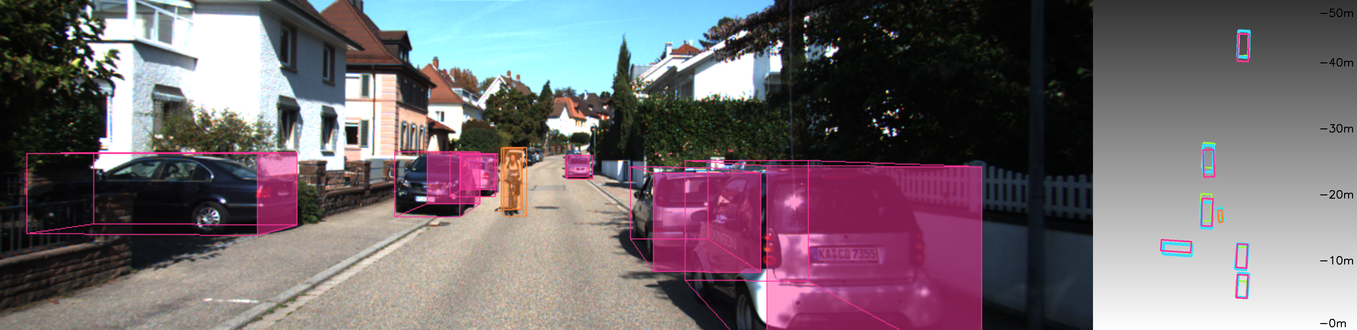}
            \end{subfigure}
            \begin{subfigure}{\figureScaleFraction\linewidth}
                \includegraphics[width=0.9\linewidth]{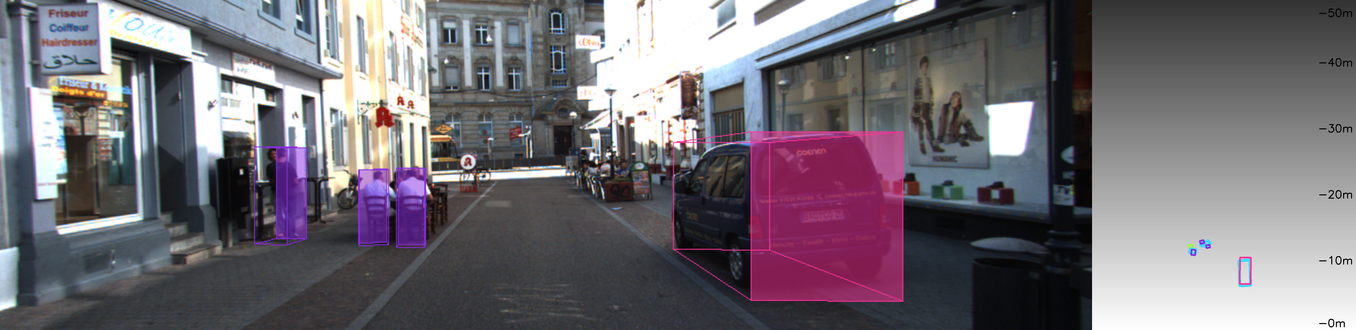}
            \end{subfigure}
            \begin{subfigure}{\figureScaleFraction\linewidth}
                \includegraphics[width=0.9\linewidth]{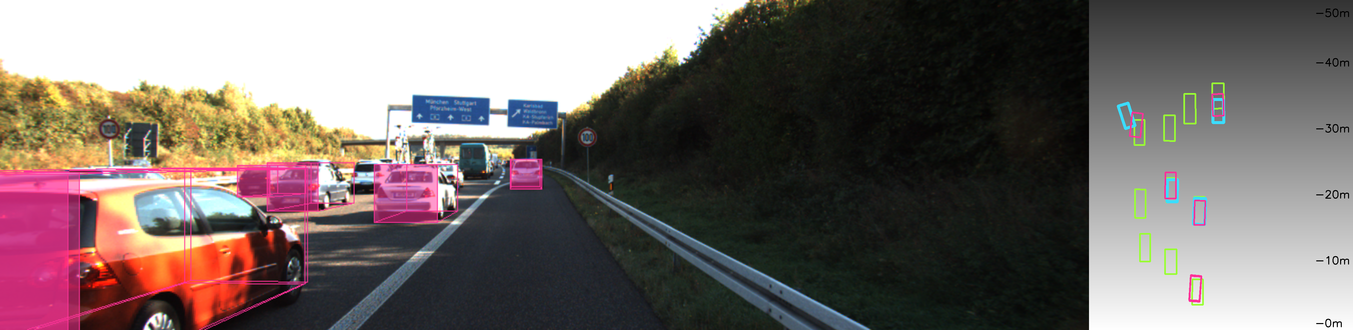}
            \end{subfigure}
            \begin{subfigure}{\figureScaleFraction\linewidth}
                \includegraphics[width=0.9\linewidth]{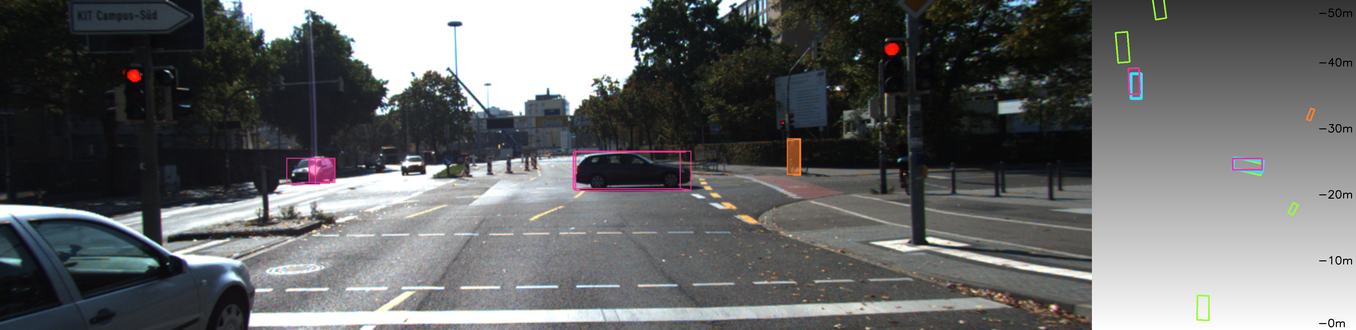}
            \end{subfigure}
            \begin{subfigure}{\figureScaleFraction\linewidth}
                \includegraphics[width=0.9\linewidth]{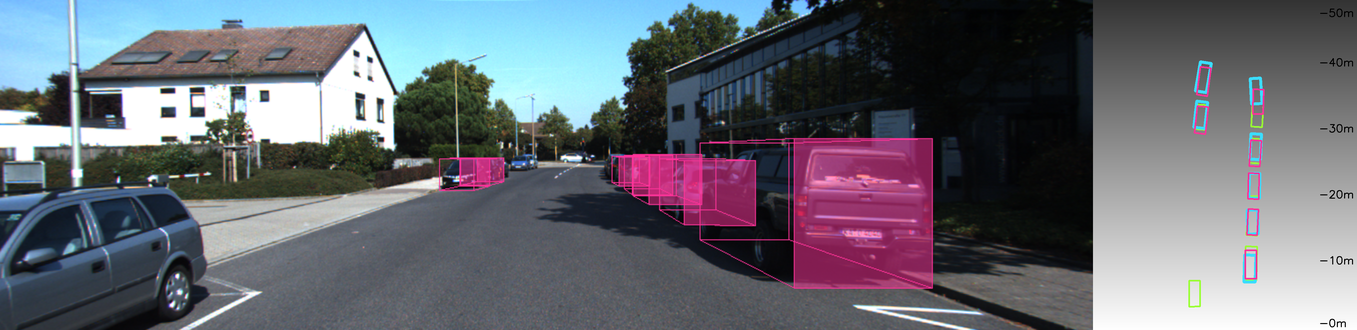}
            \end{subfigure}
            \caption[\kitti{} Qualitative Results.]{\textbf{\kitti{} Qualitative Results}. 
            \deviant{} predictions in general are more accurate than {\gupNet} \cite{lu2021geometry}.
            [Key: {Cars} (pink), {Cyclists} (orange) and {Pedestrians} (violet) of \deviant; {all classes of \gupNet (cyan)}, and {Ground Truth} (green) in BEV]. }
            \label{fig:deviant_qualitative_kitti}
        \end{figure}

        \noIndentHeading{\kitti{}.}
            We next show some more qualitative results of models trained on \kitti{} \val split in \cref{fig:deviant_qualitative_kitti}. 
            We depict the predictions of \deviant in image view on the left and the predictions of \deviant and \gupNet{} \cite{lu2021geometry}, and ground truth in BEV on the right. 
            In general, \deviant predictions are more closer to the ground truth than \gupNet{} \cite{lu2021geometry}.
            
\begin{figure}[!tb]
            \centering
            \begin{subfigure}{\figureScaleFraction\linewidth}
                \includegraphics[trim={0 2.5cm 0 5.0cm},clip,width=0.9\linewidth]{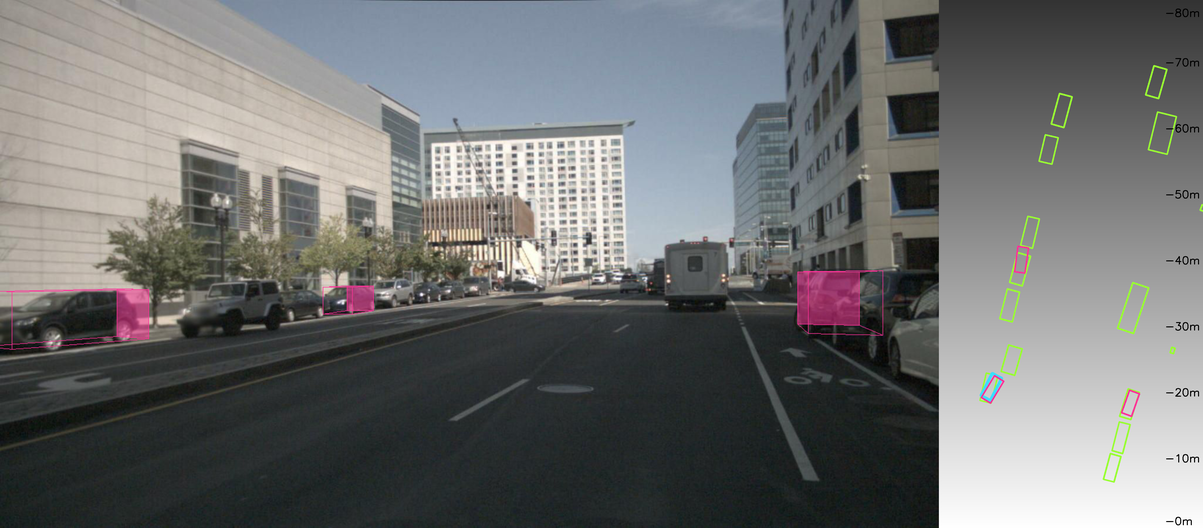}
            \end{subfigure}
            \begin{subfigure}{\figureScaleFraction\linewidth}
                \includegraphics[trim={0 2.5cm 0 5.0cm},clip,width=0.9\linewidth]{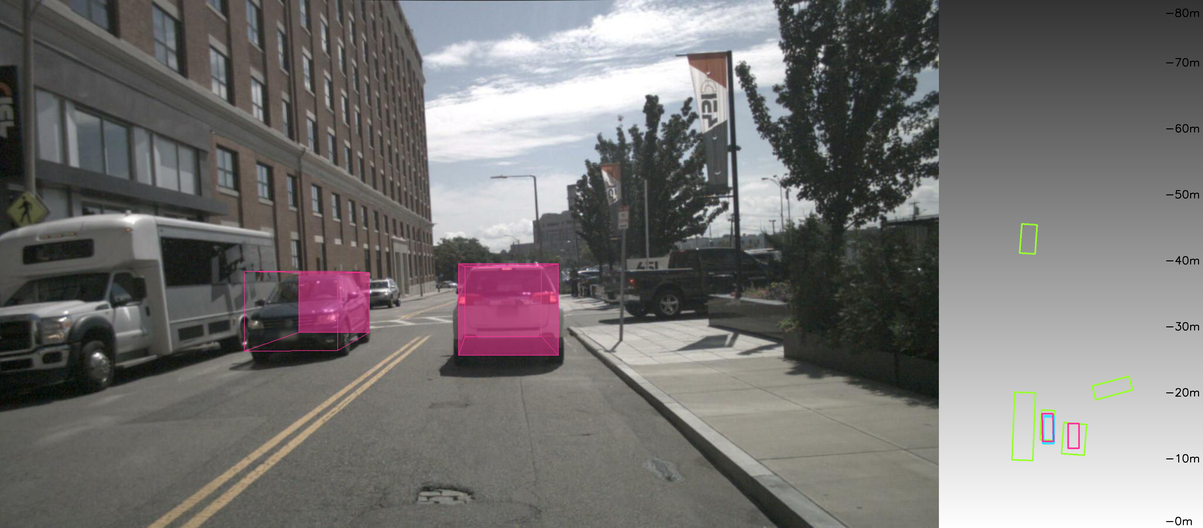}
            \end{subfigure}
            \begin{subfigure}{\figureScaleFraction\linewidth}
                \includegraphics[trim={0 2.5cm 0 5.0cm},clip,width=0.9\linewidth]{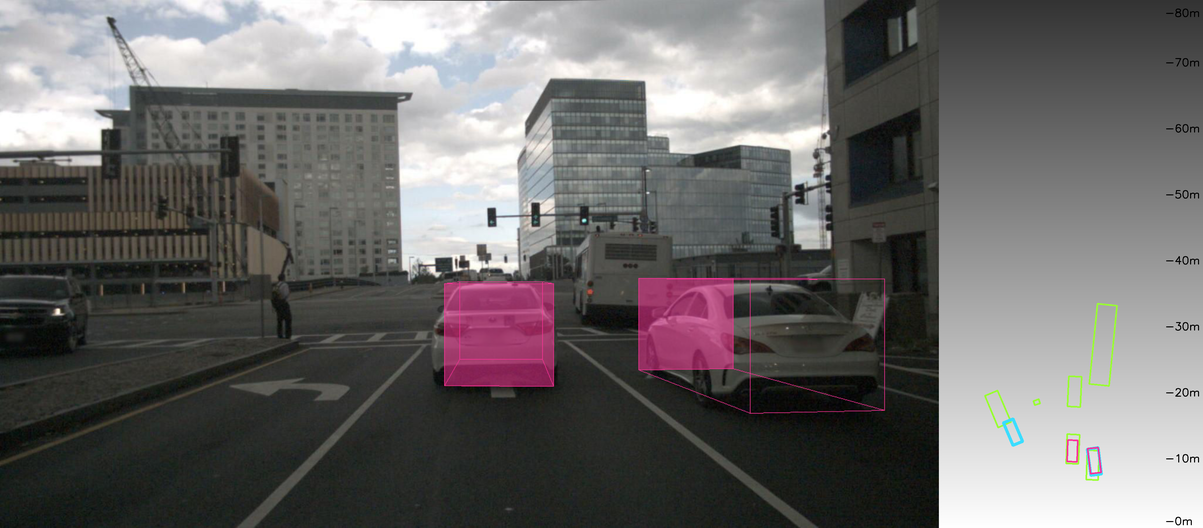}
            \end{subfigure}
            \begin{subfigure}{\figureScaleFraction\linewidth}
                \includegraphics[trim={0 2.5cm 0 5.0cm},clip,width=0.9\linewidth]{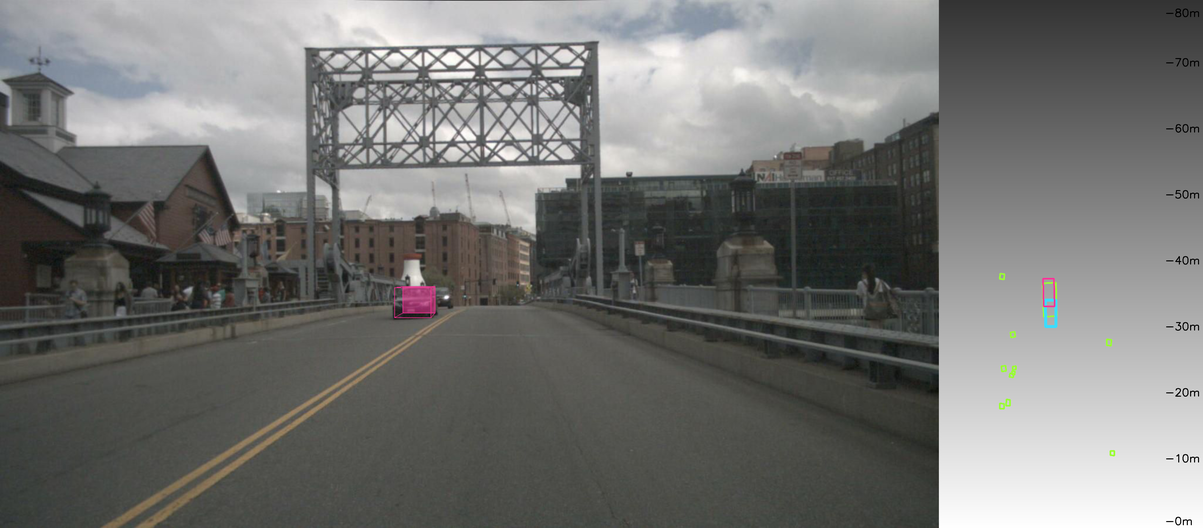}
            \end{subfigure}
            \begin{subfigure}{\figureScaleFraction\linewidth}
                \includegraphics[trim={0 2.5cm 0 5.0cm},clip,width=0.9\linewidth]{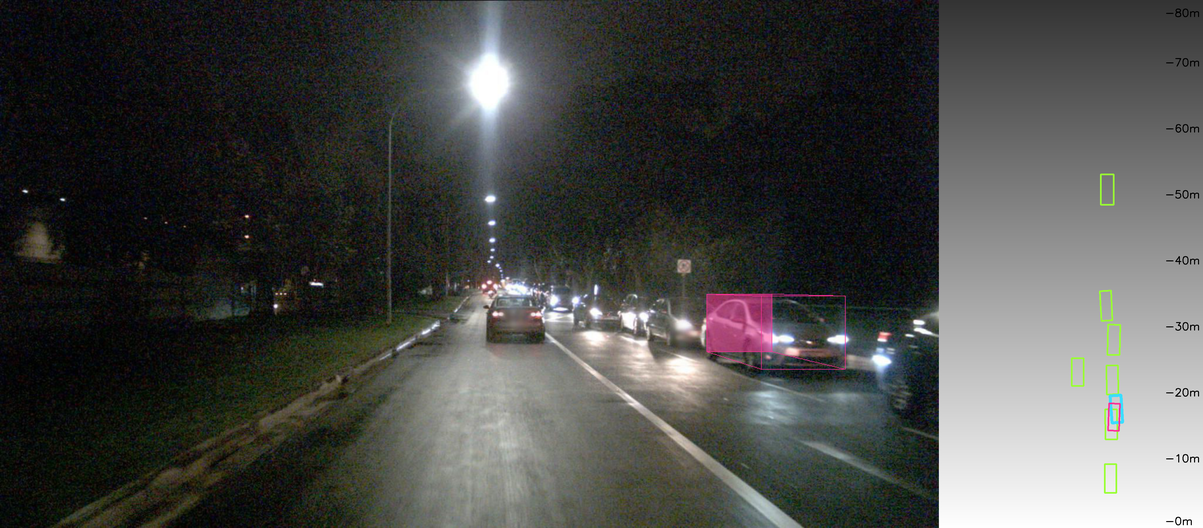}
            \end{subfigure}
            \begin{subfigure}{\figureScaleFraction\linewidth}
                \includegraphics[trim={0 2.5cm 0 5.0cm},clip,width=0.9\linewidth]{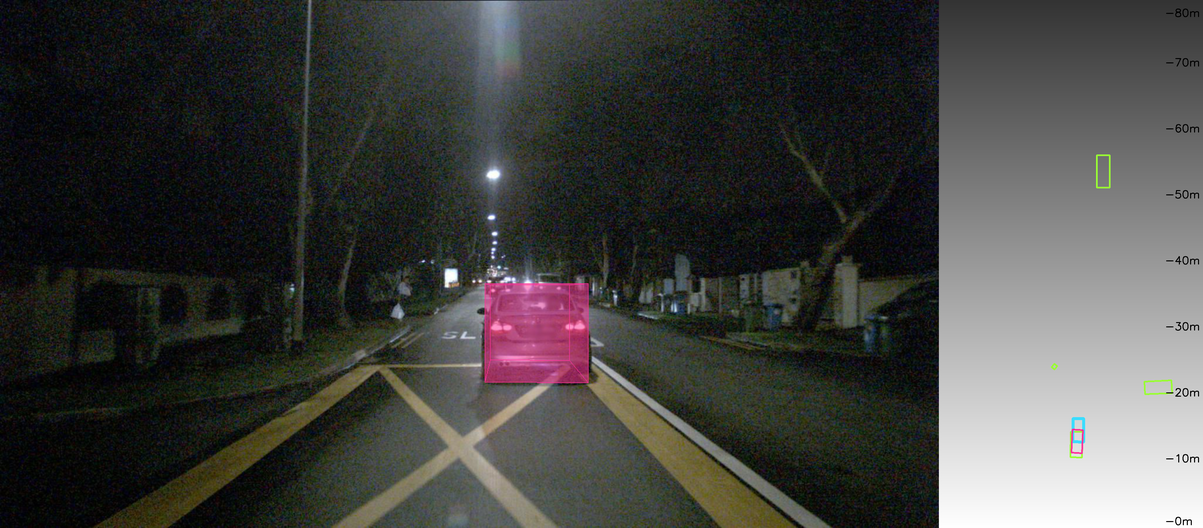}
            \end{subfigure}
            \caption[\nuscenes{} Cross-Dataset Qualitative Results.]{\textbf{\nuscenes{} Cross-Dataset Qualitative Results.}
            \deviant{} predictions in general are more accurate than {\gupNet} \cite{lu2021geometry}.
            [Key: {Cars}  of \deviant (pink); \textcolor{set1_cyan}{Cars} of \gupNet (cyan), and {Ground Truth} (green) in BEV]. }
            \label{fig:deviant_qualitative_nusc_kitti}
        \end{figure}

        \noIndentHeading{\nuscenes{} Cross-Dataset Evaluation.}
            We then show some qualitative results of \kitti{} \val model evaluated on \nuscenes{} frontal in \cref{fig:deviant_qualitative_nusc_kitti}. 
            We again observe that \deviant predictions are more closer to the ground truth than \gupNet{} \cite{lu2021geometry}.
            Also, considerably less number of boxes are detected in the cross-dataset evaluation \thatIs{} on \nuscenes{}. 
            We believe this happens because of the domain shift.
            
\begin{figure}[!tb]
            \centering
            \begin{subfigure}{\waymoFigureScaleFraction\linewidth}
                \includegraphics[trim={0 2cm 0 2cm},clip,width=0.9\linewidth]{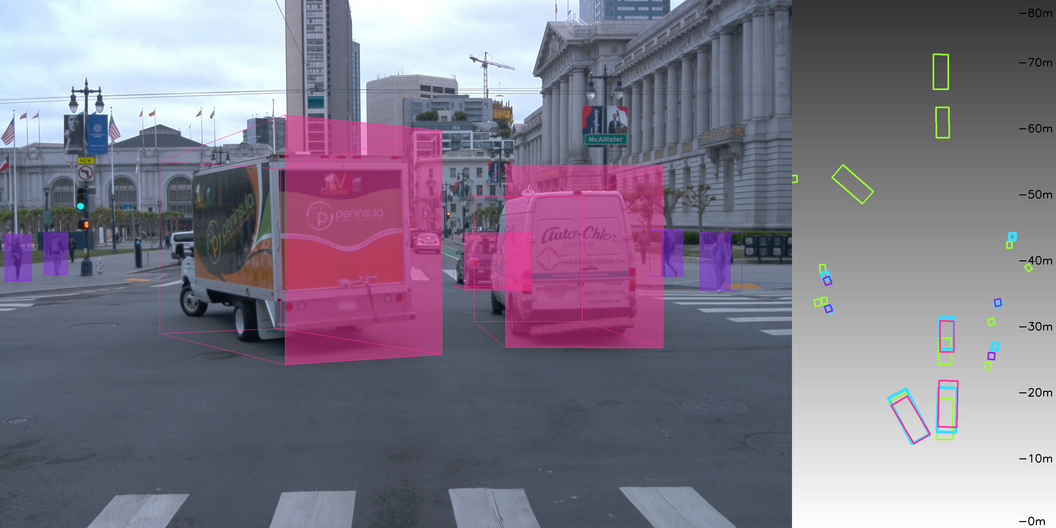}
            \end{subfigure}
            \begin{subfigure}{\waymoFigureScaleFraction\linewidth}
                \includegraphics[trim={0 2.0cm 0 2.0cm},clip,width=0.9\linewidth]{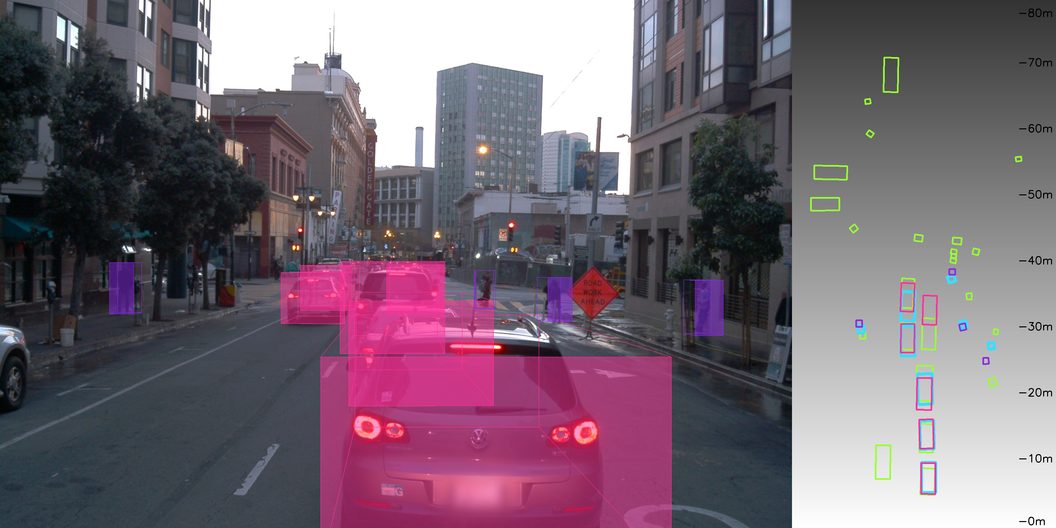}
            \end{subfigure}
            \begin{subfigure}{\waymoFigureScaleFraction\linewidth}
                \includegraphics[trim={0 2.0cm 0 2.0cm},clip,width=0.9\linewidth]{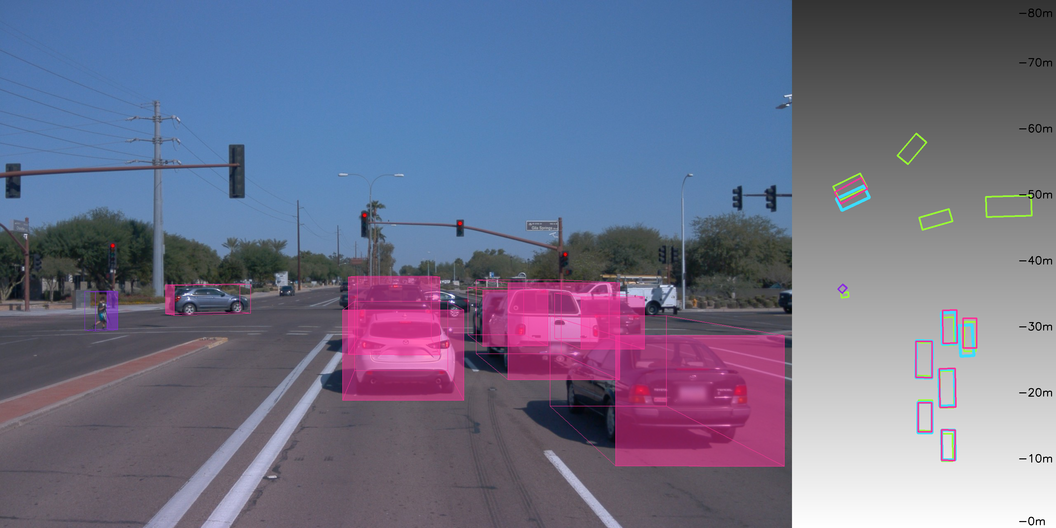}
            \end{subfigure}
            \begin{subfigure}{\waymoFigureScaleFraction\linewidth}
                \includegraphics[trim={0 2.0cm 0 2.0cm},clip,width=0.9\linewidth]{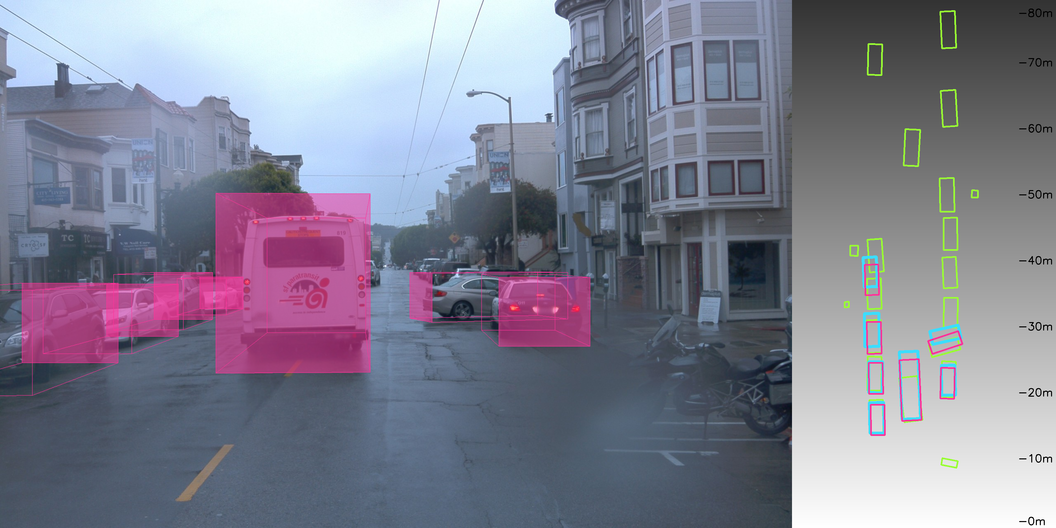}
            \end{subfigure}
            \begin{subfigure}{\waymoFigureScaleFraction\linewidth}
                \includegraphics[trim={0 2cm 0 2cm},clip,width=0.9\linewidth]{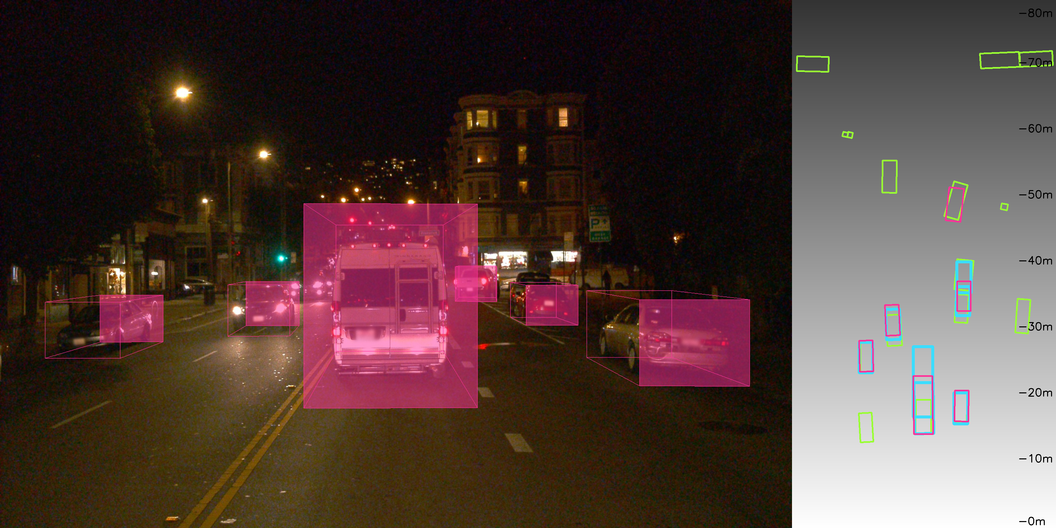}
            \end{subfigure}
            \begin{subfigure}{\waymoFigureScaleFraction\linewidth}
                \includegraphics[trim={0 2.0cm 0 2.0cm},clip,width=0.9\linewidth]{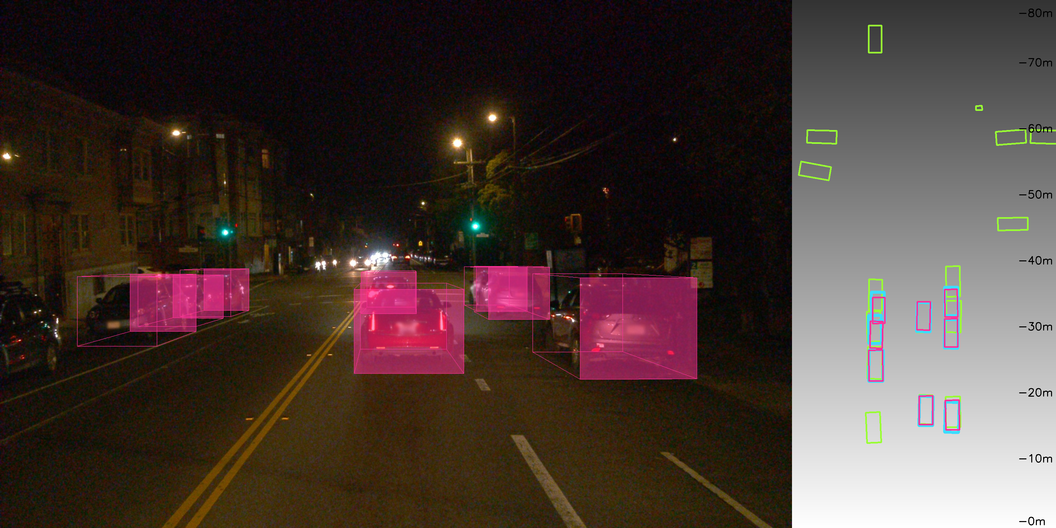}
            \end{subfigure}
            \caption[\waymo{} Qualitative Results.]{
            \textbf{\waymo{} Qualitative Results}. 
            \deviant predictions in general are more accurate than {\gupNet} \cite{lu2021geometry}.
            [Key: {Cars} (pink), {Cyclists} (orange) and {Pedestrians} (violet) of \deviant; {all classes of \gupNet (cyan)}, and {Ground Truth} (green) in BEV].}
            \label{fig:deviant_qualitative_waymo}
        \end{figure}

        \noIndentHeading{\waymo{}.}
            We now show some qualitative results of models trained on \waymo{} \val{} split in \cref{fig:deviant_qualitative_waymo}. 
            We again observe that \deviant predictions are more closer to the ground truth than \gupNet{} \cite{lu2021geometry}.

    \begin{figure}[!tb]
        \centering
        \begin{subfigure}[align=bottom]{.548\linewidth}
              \centering
              \includegraphics[width=\linewidth]{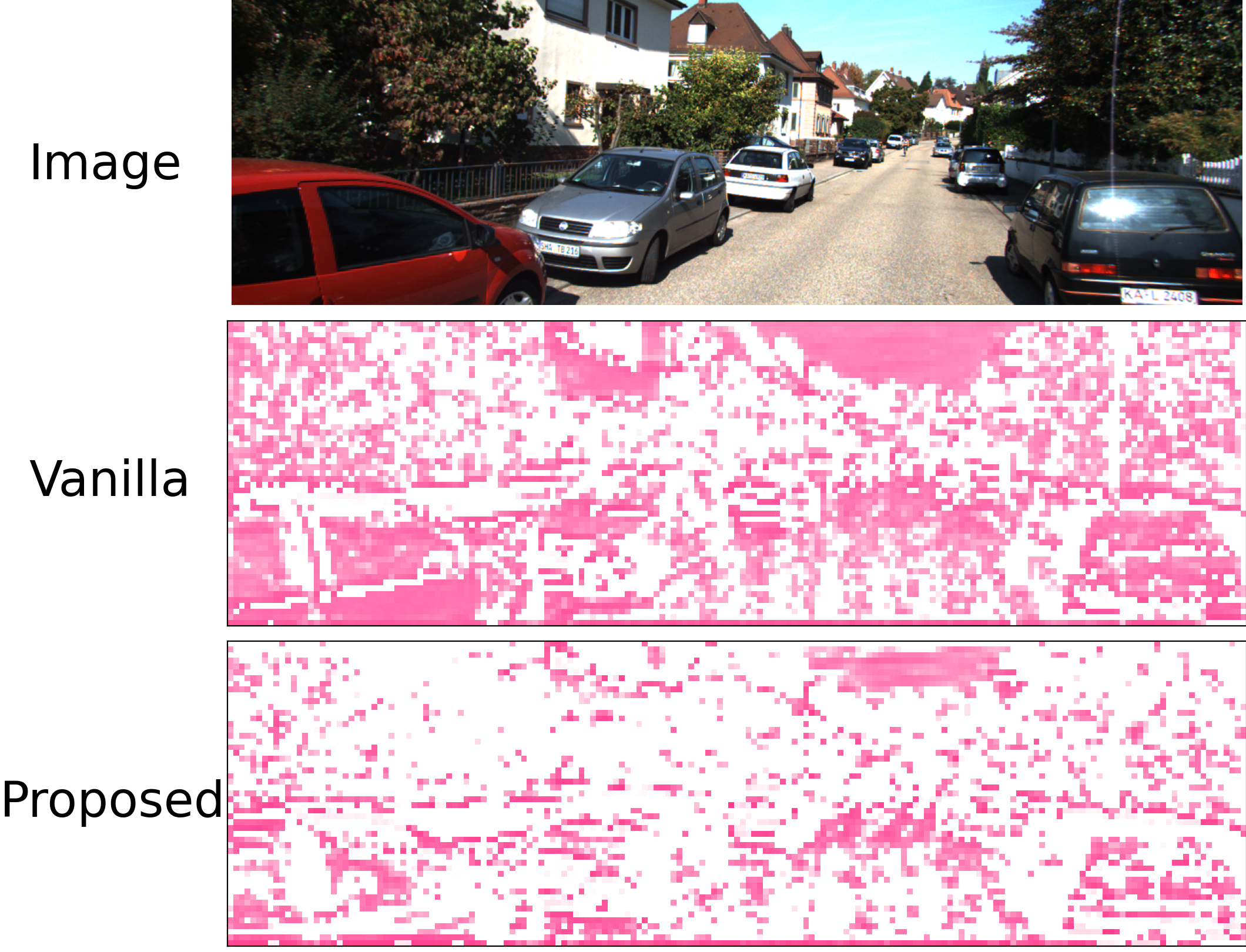}
              \caption{Depth \equivariance{} error (\downarrowRHDSmall).}
              \label{fig:deviant_equiv_error_qualitative}
        \end{subfigure}
        \begin{subfigure}[align=bottom]{.38\linewidth}
            \centering
            \includegraphics[width=\linewidth]{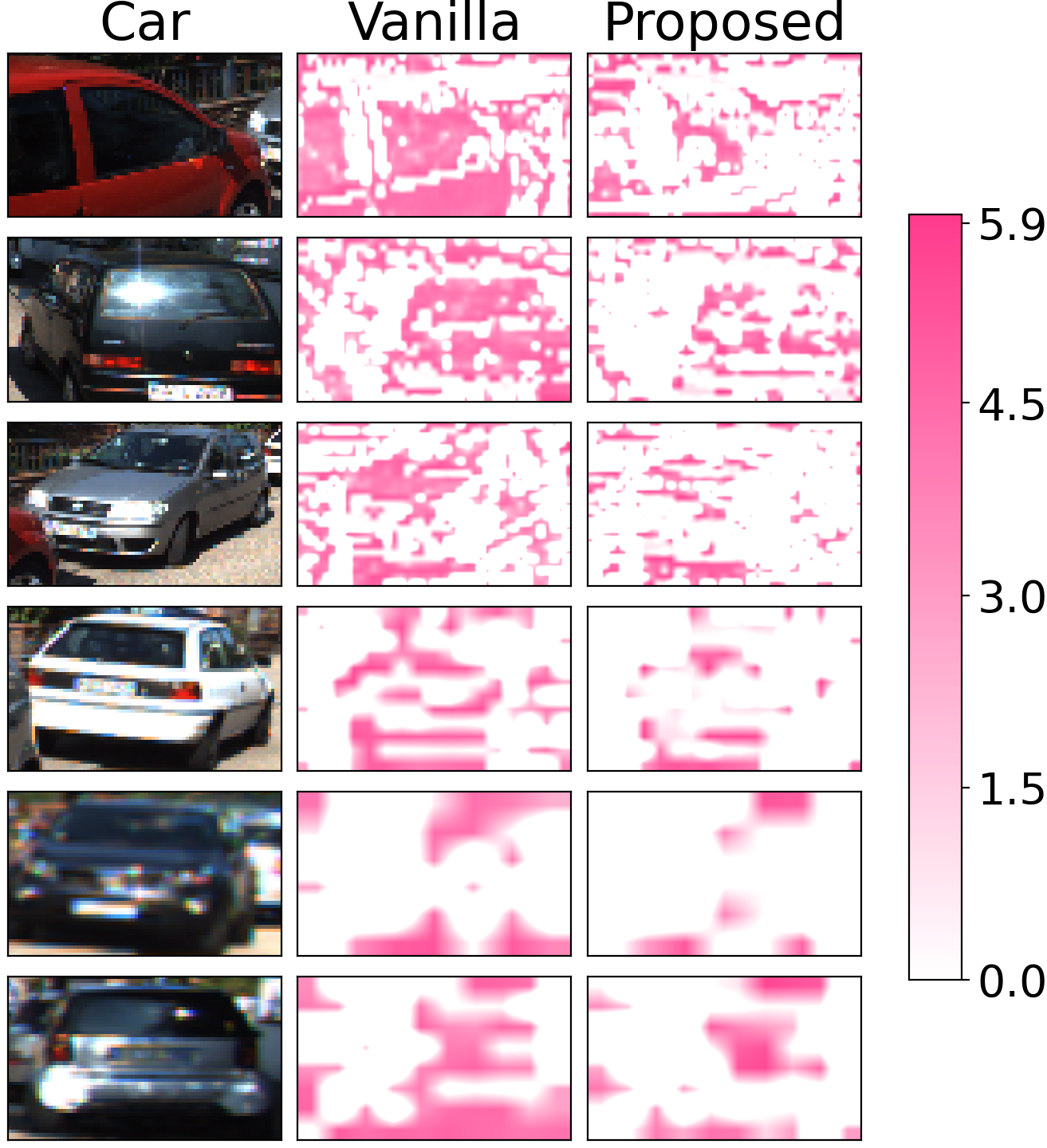}
            \caption{Error~(\downarrowRHDSmall) on objects.}
            \label{fig:deviant_equiv_error_qualitative_object}
        \end{subfigure}
        \caption[Depth (scale) \equivariance{} error of vanilla \gupNet{} and proposed \deviant on image and foreground objects.]
        {
        \textbf{(a) Depth (scale) \equivariance{} error } of vanilla \gupNet{} \cite{lu2021geometry} and proposed \deviant. (See \cref{sec:deviant_detection_results_kitti_val1} for details)
        \textbf{(b) Error on objects.} The proposed backbone has less depth \equivariance{} error than vanilla CNN backbone.
        }
        \label{fig:deviant_equiv_error_qualitative_both}
    \end{figure}
    
    \subsection{Demo Videos of \deviant}   
        
        \noIndentHeading{Detection Demo. }
            We next put a short demo video of our \deviant model trained on \kitti{} \val split at \url{https://www.youtube.com/watch?v=2D73ZBrU-PA}. 
            We run our trained model independently on each frame of \textsc{2011\_09\_26\_drive\_0009} \kitti{} raw \cite{geiger2013vision}.
            The video belongs to the City category of the \kitti{} raw video.
            None of the frames from the raw video appear in the training set of \kitti{} \val split \cite{kumar2021groomed}.
            We use the camera matrices available with the video but do not use any temporal information. 
            Overlaid on each frame of the raw input videos, we plot the projected \threeD boxes of the predictions and also plot these \threeD boxes in the BEV.
            We set the frame rate of this demo at $10$
            fps as in \kitti{}.
            The attached demo video demonstrates very stable and impressive results because of the additional equivariance to depth translations in \deviant which is absent in vanilla CNNs.
            Also, notice that the orientation of the boxes are stable despite not using any temporal information. 
            
        \noIndentHeading{\Equivariance{} Error Demo.}
            We next show the \depthEquivariance{} (\scaleEquivariance{}) error demo of one of the channels from the vanilla \gupNet{} and our proposed method at \url{https://www.youtube.com/watch?v=70DIjQkuZvw}.
            As before, we report at block $3$ of the backbones which corresponds to output feature map of the size $[96, 320]$.
            The \equivariance{} error demo indicates more white spaces which confirms that \deviant achieves lower equivariance~error compared to the baseline \gupNet{} \cite{lu2021geometry}. 
            Thus, this demo agrees with \cref{fig:deviant_equiv_error_qualitative}. 
            This happens because depth (scale) \equivariance{} is additionally hard-baked into \deviant, while the vanilla \gupNet{} is not \equivariant{} to depth translations (scale \transformation s).


%% file: images/deviant/translational_equiv_exist.tex
\begin{tikzpicture}[scale=0.28, every node/.style={scale=0.50}, every edge/.style={scale=0.50}]
\tikzset{vertex/.style = {shape=circle, draw=black!70, line width=0.06em, minimum size=1.4em}}
\tikzset{edge/.style = {-{Triangle[angle=60:.06cm 1]},> = latex'}}

    \input{images/deviant/teaser_common}

    \draw [draw=black!100, line width=0.04em, fill=gray!50, opacity=0.7]    (6.75, 9.25) rectangle (8.25, 10.75) node[]{};
    \node [inner sep=1pt, scale= 1.5,text width=3cm] at (14.5, 8.0)  {Patch Plane\\ $m\varX+n\varY+o\varZ+p= 0$};

    \draw[draw=rayShade, fill=rayShade, thick](7.5,10.0) circle (0.35) node[scale= 1.25]{};

    \draw [draw=rayShade, line width=0.1em, shorten <=0.5pt, shorten >=0.5pt, >=stealth]
       (7.5,10.0) node[]{}
    -- (4.65,6.2) node[]{};
    
    \draw [draw=black!100, line width=0.05em, shorten <=0.5pt, shorten >=0.5pt, >=stealth]
       (7.8,9.5) node[]{}
    -- (10.3,8.0) node[]{};

    \node [inner sep=1pt, scale= 1.75] at (8.5, 4.8)   {$\projectionOne(\pixU, \pixV)$};
    \node [inner sep=1pt, scale= 1.75] at (5.0, -1.1) {$\projectionTwo(\pixUTwo, \pixVTwo)$};
    \node [inner sep=1pt, scale= 1.25] at (10.1,10.0)   {$(\posX,\posY,\posZ)$};
    
    \draw[draw=black, fill=black, thick](4.7, 5.5) circle (0.12) node[scale= 1.1]{~\quad\quad\quad$(\!\ppointU,\!\ppointV\!)$};
    \draw[draw=black, fill=black, thick](0.55, -0.5) circle (0.12) node[scale= 1.1]{~\quad\quad\quad$(\!\ppointU,\!\ppointV\!)$};

    \draw [{|}-{|}, draw=black!60, line width=0.05em, shorten <=0.5pt, shorten >=0.5pt, >=stealth]
       (2.1,4.8) node[]{}
    -- (0.45,2.8) node[pos=0.5, scale= 1.75, align= center]{\color{black}{$\focal$}~~~~\\};

    \draw [{|}-{|}, draw=black!60, line width=0.05em, shorten <=0.5pt, shorten >=0.5pt, >=stealth]
       (-2.5,-1.2) node[]{}
    -- (-4.05,-3.2) node[pos=0.5, scale= 1.75, align= center]{\color{black}{$\focal$}~~~~\\};


    
    \draw[draw=axisShadeDark, fill=axisShadeDark, thick](-1.65,10.0) circle (0.08) node[]{};
    
    \draw [-{Triangle[angle=60:.1cm 1]}, draw=axisShadeDark, line width=0.05em, shorten <=0.5pt, shorten >=0.5pt, >=stealth]
           (-1.65,10.0) node[]{}
        -- (0.3,10.0) node[scale= 1.75]{~~$x$};
        
    \draw [-{Triangle[angle=60:.1cm 1]}, draw=axisShadeDark, line width=0.05em, shorten <=0.5pt, shorten >=0.5pt, >=stealth]
           (-1.65,10.0) node[]{}
        -- (-1.65,8.35) node[text width=1cm,align=center,scale= 1.75]{~\\$y$};
    
    \draw [-{Triangle[angle=60:.1cm 1]}, draw=axisShadeDark, line width=0.05em, shorten <=0.5pt, shorten >=0.5pt, >=stealth]
           (-1.65,10.0) node[]{}
        -- (-0.75,11.0) node[scale= 1.75]{~~$z$};
    
\end{tikzpicture}

%% file: images/deviant/translational_equiv_not_exist.tex
\begin{tikzpicture}[scale=0.28, every node/.style={scale=0.50}, every edge/.style={scale=0.50}]
\tikzset{vertex/.style = {shape=circle, draw=black!70, line width=0.06em, minimum size=1.4em}}
\tikzset{edge/.style = {-{Triangle[angle=60:.06cm 1]},> = latex'}}

\node[inner sep=0pt] (input) at (7.5,10.0) {\includegraphics[height=3.0cm]{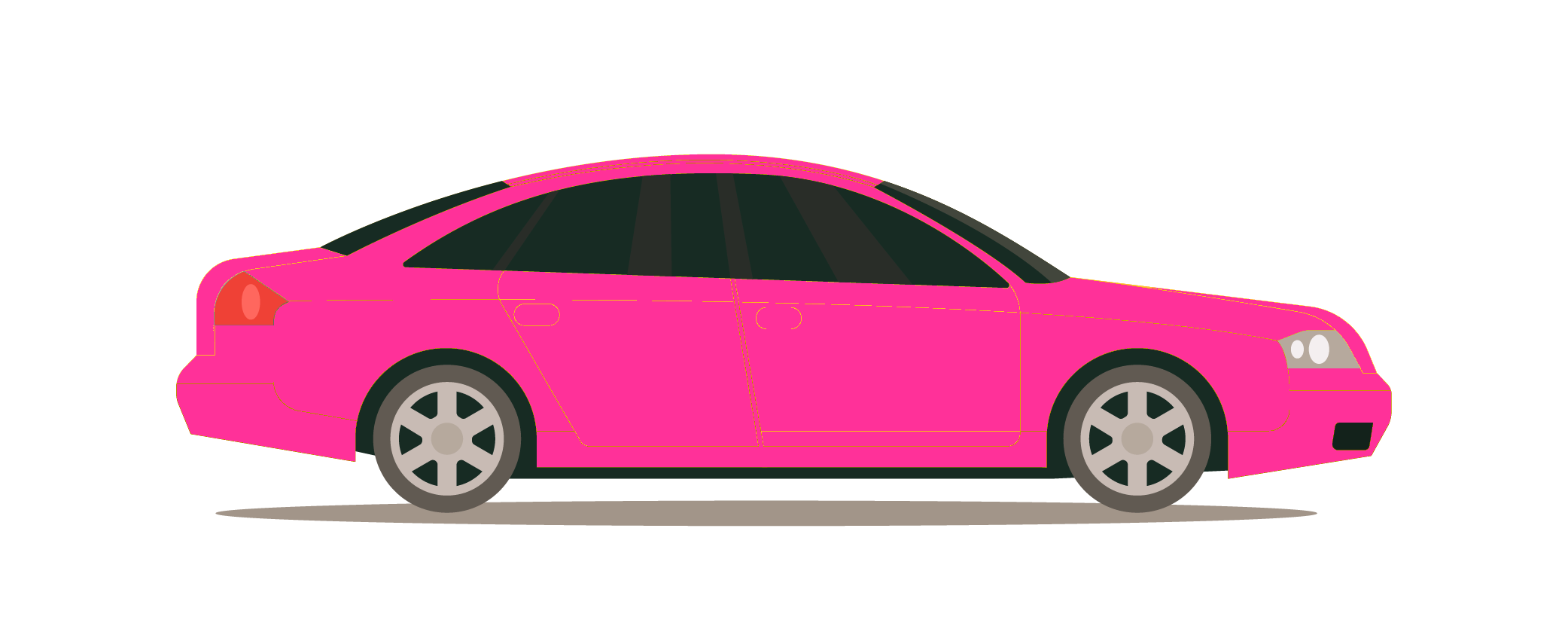}};
    
\draw [draw=projectionBorderShade, line width=0.02em, fill=projectionFillShade]    (-3.2, 4.2) rectangle (2.8, 7.2) node[]{};
\node[inner sep=0pt] (input) at (0.0,5.8) {\includegraphics[height=1.5cm]{images/deviant/pink_car_back.png}};

\draw [draw=projectionBorderShade, line width=0.02em, fill=projectionFillShade]    (13.2, 4.2) rectangle (18.8, 7.2) node[]{};
\node[inner sep=0pt] (input) at (16,5.8) {\includegraphics[height=1.5cm]{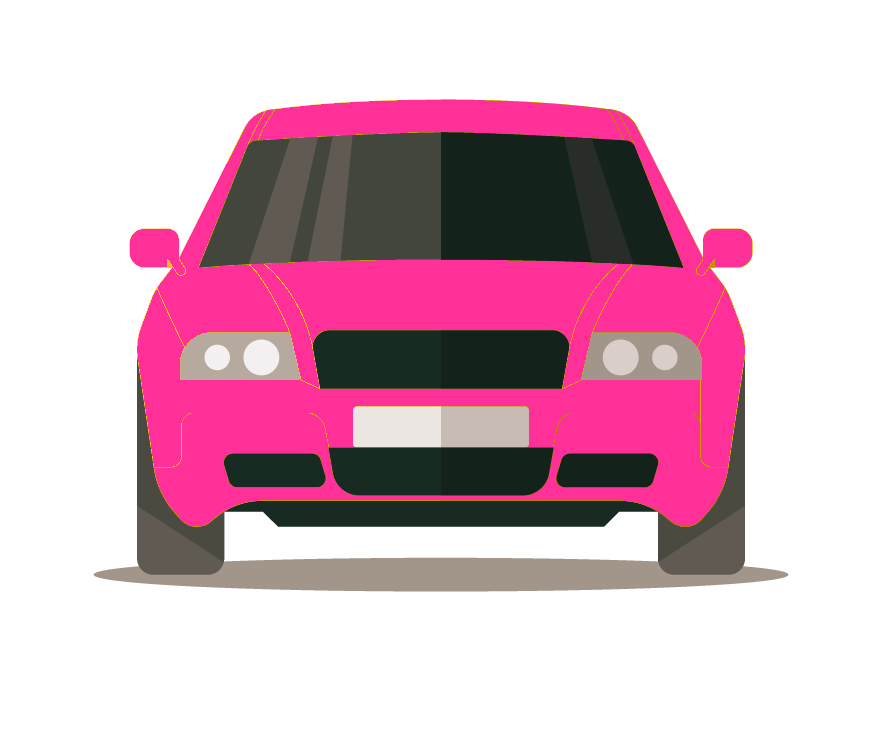}};

\node [inner sep=1pt, scale= 1.75] at (4.8, 5.8)  {$\projectionOne(\pixU, \pixV)$};
\node [inner sep=1pt, scale= 1.75] at (21.0, 5.8) {$\projectionTwo(\pixUTwo, \pixVTwo)$};

\draw [draw=rayShade, line width=0.1em, shorten <=0.5pt, shorten >=0.5pt, >=stealth]
       (2.7,10.0) node[]{}
    -- (-1.5,10.0) node[]{};

\draw [draw=rayShade, line width=0.1em, shorten <=0.5pt, shorten >=0.5pt, >=stealth]
       (12.06,10.0) node[]{}
    -- (15.76,10.0) node[]{};
    
\draw[black,fill=black!50] (-1.6,10.4) rectangle (-0.4,9.6);
\coordinate (c10) at (-0.4,10.0);
\coordinate (c11) at (0.18,9.6);
\coordinate (c12) at (0.18,10.4);
\filldraw[draw=black, fill=gray!20] (c10) -- (c11) -- (c12) -- cycle;

\draw[black, fill=black!50, line width=0.05em] (15.76,9.6) rectangle (16.96,10.4);
\coordinate (c00) at (15.76,10.0);
\coordinate (c01) at (15.18,9.6);
\coordinate (c02) at (15.18,10.4);
\filldraw[draw=black, fill=gray!20, line width=0.05em] (c00) -- (c01) -- (c02) -- cycle;

\end{tikzpicture}

%% file: images/deviant/theorem_1_overview.tex
\begin{tikzpicture}[scale=0.30, every node/.style={scale=0.5}, every edge/.style={scale=0.50}]
\tikzset{vertex/.style = {shape=circle, draw=black!70, line width=0.06em, minimum size=1.4em}}
\tikzset{edge/.style = {-{Triangle[angle=60:.06cm 1]},> = latex'}}

\draw[black,fill=black!20] (-2, 0) rectangle (4, 3);
\node [inner sep=1pt, scale= 2, align= center] at (1, 1.5)  {\threeD{} point\\on plane};

\draw[black,fill=black!20] (-2, 8) rectangle (4, 11);
\node [inner sep=1pt, scale= 2, align= center] at (1, 9.5)  {\threeD{} point\\on plane};

\draw [-{Triangle[angle=60:.1cm 1]},draw=black, line width=0.1em, shorten <=0.5pt, shorten >=0.5pt, >=stealth]
      (1,8) node[]{}
    -- (1,3) node[]{};
\node [inner sep=1pt, scale= 2] at (2.5, 5.5)  {$(\rotation,\translation)$};

\draw [-{Triangle[angle=60:.1cm 1]},draw=black, line width=0.1em, shorten <=0.5pt, shorten >=0.5pt, >=stealth]
      (4,1.5) node[]{}
    -- (10,1.5) node[]{};
\node [inner sep=1pt, scale= 2] at (6.5, 2)  {Projective};

\draw [-{Triangle[angle=60:.1cm 1]},draw=black, line width=0.1em, shorten <=0.5pt, shorten >=0.5pt, >=stealth]
      (4,9.5) node[]{}
    -- (10,9.5) node[]{};
\node [inner sep=1pt, scale= 2] at (6.5, 10)  {Projective};

\draw[black,fill=black!20] (10, 0) rectangle (16, 3);
\node [inner sep=1pt, scale= 2] at (13, 1.5)  {\twoD{} point};

\draw[black,fill=black!20] (10, 8) rectangle (16, 11);
\node [inner sep=1pt, scale= 2] at (13, 9.5)  {\twoD{} point};

\draw [-{Triangle[angle=60:.1cm 1]},draw=black, line width=0.1em, shorten <=0.5pt, shorten >=0.5pt, >=stealth]
      (13,8) node[]{}
    -- (13,3) node[]{};
\node [inner sep=1pt, scale= 2] at (14.5, 5.5)  {Th. \ref{th:projective_bigboss}};

\draw[black,dashed] (9.25, -0.5) rectangle (16.5, 11.5);

\draw [draw=rayShade, line width=0.15em, shorten <=0.5pt, shorten >=0.5pt, >=stealth]
      (17,16) node[]{}
    -- (17,-4) node[]{};
\node [inner sep=1pt, scale= 2] at (12, 14)  {Continuous World};
\node [inner sep=1pt, scale= 2] at (21.5, 14)  {Discrete World};

\draw [-{Triangle[angle=60:.1cm 1]},draw=black, line width=0.1em, shorten <=0.5pt, shorten >=0.5pt, >=stealth]
      (16,1.5) node[]{}
    -- (23,1.5) node[]{};
\node [inner sep=1pt, scale= 2] at (19.5, 2)  {Sampling};

\draw [-{Triangle[angle=60:.1cm 1]},draw=black, line width=0.1em, shorten <=0.5pt, shorten >=0.5pt, >=stealth]
      (16,9.5) node[]{}
    -- (23,9.5) node[]{};
\node [inner sep=1pt, scale= 2] at (19.5, 10)  {Sampling};

\draw[black,fill=black!20] (23, 0) rectangle (29, 3);
\node [inner sep=1pt, scale= 2] at (26, 1.5)  {\twoD{} pixel};

\draw[black,fill=black!20] (23, 8) rectangle (29, 11);
\node [inner sep=1pt, scale= 2] at (26, 9.5)  {\twoD{} pixel};

\end{tikzpicture}

%% file: images/deviant/the_assumption.tex
\begin{tikzpicture}[scale=0.36, every node/.style={scale=0.6}, every edge/.style={scale=0.60}]
\tikzset{vertex/.style = {shape=circle, draw=black!70, line width=0.06em, minimum size=1.4em}}
\tikzset{edge/.style = {-{Triangle[angle=60:.06cm 1]},> = latex'}}

\draw[black,fill=black!50] (-1.6,10.4) rectangle (-0.4,9.6);
\coordinate (c10) at (-0.4,10.0);
\coordinate (c11) at (0.18,9.6);
\coordinate (c12) at (0.18,10.4);
\filldraw[draw=black, fill=gray!20] (c10) -- (c11) -- (c12) -- cycle;

\draw [draw=rayShade, line width=0.1em, shorten <=0.5pt, shorten >=0.5pt, >=stealth]
       (0.2,10.0) node[]{}
    -- (8.6,10.0) node[]{};


\draw [, draw=axisShadeDark, line width=0.1em]
       (8.5,7.5) node[]{}
    -- (8.5,12.5) node[text width=1cm,align=center,scale= 1.75]{};

\draw [dashed, draw=axisShadeDark, line width=0.07em, shorten <=0.5pt, shorten >=0.5pt, >=stealth]
       (8.0,8.0) node[]{}
    -- (9.0,12.0) node[text width=1cm,align=center,scale= 1.75]{};

\draw [draw=my_magenta, line width=0.05em, shorten <=0.5pt, shorten >=0.5pt, >=stealth]
       (8.45,11.4) node[]{}
    -- (9.0,11.4) node[text width=1cm,align=center,scale= 1.75]{~~~~~\textcolor{my_magenta}{$\theta$}};

\draw [dashed, draw=axisShadeDark, line width=0.07em, shorten <=0.5pt, shorten >=0.5pt, >=stealth]
       (9.0,8.0) node[]{}
    -- (8.0,12.0) node[text width=1cm,align=center,scale= 1.75]{};

    \draw[draw=axisShadeDark, fill=axisShadeDark, thick](-6.65,10.0) circle (0.08) node[]{};
    
    \draw [-{Triangle[angle=60:.1cm 1]}, draw=axisShadeDark, line width=0.05em, shorten <=0.5pt, shorten >=0.5pt, >=stealth]
           (-6.65,10.0) node[]{}
        -- (-4.7,10.0) node[scale= 1.75]{~~$z$};
        
    \draw [-{Triangle[angle=60:.1cm 1]}, draw=axisShadeDark, line width=0.05em, shorten <=0.5pt, shorten >=0.5pt, >=stealth]
           (-6.65,10.0) node[]{}
        -- (-6.65,8.35) node[text width=1cm,align=center,scale= 1.75]{~\\$y$};
    

\end{tikzpicture}

%% file: images/deviant/steerable_idea.tex
\begin{tikzpicture}[scale=0.45, every node/.style={scale=0.75}, every edge/.style={scale=0.75}]
\tikzset{vertex/.style = {shape=circle, draw=black!70, line width=0.06em, minimum size=1.4em}}
\tikzset{edge/.style = {-{Triangle[angle=60:.06cm 1]},> = latex'}}

    \node[inner sep=0pt] (input) at (-3.24,2) {\includegraphics[trim={1.58cm 0 4.2cm 0},clip,height=1.65cm]{images/deviant/sesn_basis_eff_size_7.png}};
    \draw [draw=black!100, line width=0.06em]    (-0.5, 3.35) rectangle (-6.0, 0.65) node[]{};

    \draw [draw=black!100, line width=0.06em, fill= gray!30]    (-4, -4) rectangle (-2, -2) node[]{};

    \draw [draw=rayShade, line width=0.1em, shorten <=0.5pt, shorten >=0.5pt, >=stealth]
           (-2,-3) node[]{}
        -- (-0.6,-0.75) node[]{};
    
    \draw [draw=rayShade, line width=0.1em, shorten <=0.5pt, shorten >=0.5pt, >=stealth]
           (-2,-3) node[]{}
        -- (0.5,-3) node[]{};
        
    \draw [draw=rayShade, line width=0.1em, shorten <=0.5pt, shorten >=0.5pt, >=stealth]
           (-2,-3) node[]{}
        -- (1.48,-5.25) node[]{};
        
    \draw [draw=gray, line width=0.02em, shorten <=0.5pt, shorten >=0.5pt, >=stealth]
           (-0.5, 2.85) node[]{}
        -- (0.05, 2.85) node[]{};
    \draw [-{Triangle[angle=60:.1cm 1]}, draw=gray, line width=0.02em, shorten <=0.5pt, shorten >=0.5pt, >=stealth]
           (0.0, 2.9) node[]{}
        -- (0.0,-0.14) node[]{};

    \draw [draw=gray, line width=0.02em, shorten <=0.5pt, shorten >=0.5pt, >=stealth]
           (-0.5, 2.05) node[]{}
        -- (1.05, 2.05) node[]{};
    \draw [-{Triangle[angle=60:.1cm 1]},draw=gray, line width=0.02em, shorten <=0.5pt, shorten >=0.5pt, >=stealth]
           (1., 2.1) node[]{}
        -- (1.,-2.35) node[]{};

    \draw [draw=gray, line width=0.02em, shorten <=0.5pt, shorten >=0.5pt, >=stealth]
           (-0.5, 1.15) node[]{}
        -- (2.05, 1.15) node[]{};
    \draw [-{Triangle[angle=60:.1cm 1]},draw=gray, line width=0.02em, shorten <=0.5pt, shorten >=0.5pt, >=stealth]
           (2, 1.2) node[]{}
        -- (2,-4.6) node[]{};
        
    \node [inner sep=1pt, scale= 1.00, align= center, fill=white] at (0.23,  0.6)  {$\otimes\weight$};
    \node [inner sep=1pt, scale= 1.00, align= center, fill=white] at (1.23,  0.6)  {$\otimes\weight$};
    \node [inner sep=1pt, scale= 1.00, align= center, fill=white] at (2.23,  0.6)  {$\otimes\weight$};
    
    \node [inner sep=1pt, scale= 1.00, align= center, fill=white] at (-.02,  -1.65)  {Kernel};    

    \draw [-{Triangle[angle=60:.1cm 1]}, draw=rayShade, line width=0.1em, shorten <=0.5pt, shorten >=0.5pt, >=stealth]
           (0.5,-.75) node[]{}
        -- (3.5,-.75) node[]{};

    \draw [-{Triangle[angle=60:.1cm 1]}, draw=rayShade, line width=0.1em, shorten <=0.5pt, shorten >=0.5pt, >=stealth]
           (1.5,-3) node[]{}
        -- (3.5,-3) node[]{};

    \draw [-{Triangle[angle=60:.1cm 1]}, draw=rayShade, line width=0.1em, shorten <=0.5pt, shorten >=0.5pt, >=stealth]
           (2.5,-5.25) node[]{}
        -- (3.5,-5.25) node[]{};

    \node[inner sep=0pt] (input) at (0.,-0.75) {\includegraphics[trim={0.5cm 22cm 0.5cm 0cm},clip,height=0.8cm]{images/deviant/sesn_basis_sample.png}};
    \node[inner sep=0pt] (input) at (1.,-3) {\includegraphics[trim={0.5cm 11cm 0.5cm 11cm},clip,height=0.8cm]{images/deviant/sesn_basis_sample.png}};
    \node[inner sep=0pt] (input) at (2,-5.25) {\includegraphics[trim={0.5cm 0cm 0.5cm 22cm},clip,height=0.8cm]{images/deviant/sesn_basis_sample.png}};

    \draw [draw=black!100, line width=0.06em, fill= darkGreen!50]    (3.5, -1.50) rectangle (5.0,  0) node[]{};
    \draw [draw=black!100, line width=0.06em, fill= darkGreen!30]    (3.5, -3.75) rectangle (5.0, -2.25) node[]{};
    \draw [draw=black!100, line width=0.06em, fill= darkGreen!10]    (3.5, -4.50) rectangle (5.0, -6.00) node[]{};

    \draw [draw=black, line width=0.05em, shorten <=0.5pt, shorten >=0.5pt, >=stealth, densely dashed]
           (5.8, 4.0) node[]{}
        -- (5.8,-6.0) node[]{};

    \draw [-{Triangle[angle=60:.1cm 1]}, draw=rayShade, line width=0.1em, shorten <=0.5pt, shorten >=0.5pt, >=stealth]
           (5.0,-.75) node[]{}
        -- (6.95,-2.8) node[]{};

    \draw [-{Triangle[angle=60:.1cm 1]}, draw=rayShade, line width=0.1em, shorten <=0.5pt, shorten >=0.5pt, >=stealth]
           (5.0,-3.0) node[]{}
        -- (6.95,-3.0) node[]{};

    \draw [-{Triangle[angle=60:.1cm 1]}, draw=rayShade, line width=0.1em, shorten <=0.5pt, shorten >=0.5pt, >=stealth]
           (5.0,-5.25) node[]{}
        -- (6.95,-3.2) node[]{};

    \draw [-{Triangle[angle=60:.1cm 1]}, draw=rayShade, line width=0.1em, shorten <=0.5pt, shorten >=0.5pt, >=stealth]
           (7.25,-3.0) node[]{}
        -- (8.75 ,-3.0) node[]{};

    \draw [draw=black!100, line width=0.06em, fill=set1_cyan]    (6.95, -3.75) rectangle (7.7, -2.25) node[]{};
    
    \draw [draw=black!100, line width=0.06em, fill= darkGreen!90]    (8.75, -3.75) rectangle (10.25, -2.25) node[]{};

    \node [inner sep=1pt, scale= 1.00, align= center] at (-3.00,  3.75)  {Multi-scale Steerable Basis};
    
    \node [inner sep=1pt, scale= 1.00, align= center] at (-3.0, -4.5)  {Input};
    
    \node [inner sep=1pt, scale= 2.00, align= center] at (-1.3, -1.5)  {*};
    \node [inner sep=1pt, scale= 2.00, align= center] at (-1.3, -2.8)  {*};
    \node [inner sep=1pt, scale= 2.00, align= center] at (-1.3, -4.0)  {*};

    \node [inner sep=1pt, scale= 1.00, align= center] at (4.3,  1.00)  {Scale Conv.\\Output};
    
    \node [inner sep=1pt, scale= 1.00, align= center] at (7.2, -4.75)  {Scale-~~\\Projection};
    \node [inner sep=1pt, scale= 1.00, align= center] at (9.5, -4.75)  {\fourD\\Output};

\end{tikzpicture}

%% file: appendices/seabird_appendix.tex
\chapter{\seabird Appendix}\label{chpt:seabird_appendix}

\section{Additional Explanations and Proofs}\label{sec:seabird_additional}

    We now add some explanations and proofs which we could not put in the main chapter because of the space constraints.

    \subsection{Proof of Converged Value}\label{sec:seabird_proof_converged}
        We first bound the converged value from the optimal value. 
        These results are well-known in the literature \cite{shalev2007pegasos, lacoste2012simpler}. 
        We reproduce the result from using our notations for completeness.
        \begin{align}
            &\expect\left(\norm{\layerWeightConv\!-\!\layerWeightOptimal}_2^2\right) \nonumber\\
            &= \expect\left(\norm{\layerWeightConv\!-\!\layerWeightMean + \layerWeightMean\!-\!\layerWeightOptimal}^2_2\right) \nonumber \\
            &= \expect\left(\left(\layerWeightConv\!-\!\layerWeightMean + \layerWeightMean\!-\!\layerWeightOptimal\right)^T\left(\layerWeightConv\!-\!\layerWeightMean + \layerWeightMean\!-\!\layerWeightOptimal\right)\right) \nonumber \\
            &= \expect((\layerWeightConv\!-\!\layerWeightMean)^T(\layerWeightConv\!-\!\layerWeightMean)) + \expect((\layerWeightMean\!-\!\layerWeightOptimal)^T(\layerWeightMean\!-\!\layerWeightOptimal)) \nonumber \\
            &~~~~+ 2\expect((\layerWeightConv\!-\!\layerWeightMean)^T(\layerWeightMean\!-\!\layerWeightOptimal)) \nonumber \\
            &= \var(\layerWeightConv) + \expect((\layerWeightMean\!-\!\layerWeightOptimal)^T(\layerWeightMean\!-\!\layerWeightOptimal)) 
            \label{eq:weight_bound_with_mean}
        \end{align}
        where $\layerWeightMean=\expect(\layerWeightConv)$ is the mean of the layer weight and 
         $\var(\mathbf{\weight})$ denotes the variance of $\sum_\instantTwo w_\instantTwo^2$.
    
        \noIndentHeading{SGD.}
        We begin the proof by writing the value of $\layerWeightTime$ at every step.
        The model uses SGD, and so, the weight $\layerWeightTime$ after $\instant$ gradient updates is
        \begin{align}
            \layerWeightTime &= \layerWeightZero - \step_1 {}^\loss\gradient_1 - \step_2 {}^\loss\gradient_2 - \cdots - s_\instant \gradTime,
            \label{eq:sgd_step}
        \end{align}
        where $\gradTime$ denotes the gradient of $\layerWeight$ at every step $\instant$.
        Assume the loss function under consideration $\loss$ is $\loss = f(\layerWeightTimeVanilla\image - \depthGT) = f(\noise)$. Then, we have,
        \begin{align}
            \gradTime &= \dfrac{\partial \loss}{\partial \layerWeightTimeVanilla} \nonumber \\
            &= \dfrac{\partial \loss(\layerWeightTimeVanilla\image - \depthGT)}{\partial \layerWeightTimeVanilla} \nonumber \\
            &= \dfrac{\partial \loss(\layerWeightTimeVanilla\image - \depthGT)}{\partial (\layerWeightTimeVanilla\image - \depthGT)} \dfrac{\partial (\layerWeightTimeVanilla\image - \depthGT)}{\partial \layerWeightTimeVanilla} \nonumber \\
            &= \dfrac{\partial \loss(\noise)}{\partial \noise} \image \nonumber \\
            &= \image \dfrac{\partial \loss(\noise)}{\partial \noise} \nonumber \\
            \implies \gradTime &= \image \funcNoise,
        \end{align}
        with $\funcNoise = \dfrac{\partial \loss(\noise)}{\partial \noise}$ is the gradient of the loss function wrt noise. 
        
        \noIndentHeading{Expectation and Variance of Gradient $\gradTime$}
        Since the image $\image$ and noise $\noise$ are statistically independent, the image and the noise gradient $\noise$ are also statistically independent. 
        So, the expected gradients
        \begin{align}
            \expect(\gradTime) &= \expect(\image) \expect(\funcNoise) = 0.
            \label{eq:mean_gradTime}
        \end{align}
    
        Note that if the loss function is an even function (symmetric about zero), its gradient $\funcNoise$ is an odd function (anti-symmetric about $0$), and so its mean $\expect(\funcNoise) = 0$. 
        
        Next, we write the gradient variance $\var(\gradTime)$ as
        \begin{align}
            \var(\gradTime) = \var(\image \funcNoise) 
            &= \expect(\image^T\image) \expect(\funcNoise^2) - \expect^2(\image)\expect^2(\funcNoise) \nonumber \\
            &= \expect(\image^T\image) \left[ \var(\funcNoise) + \expect^2(\funcNoise) \right] \nonumber\\
            &~~~~ - \expect^2(\image)\expect^2(\funcNoise) \nonumber \\
            \implies \var(\gradTime) &= \expect(\image^T\image)\var(\funcNoise)\quad\text{as } \expect(\funcNoise)=0
            \label{eq:var_gradTime}
        \end{align}
    
        \noIndentHeading{Expectation and Variance of Converged Weight $\layerWeightTime$}
        We first calculate the expected  converged weight as
        \begin{align}
            \expect(\layerWeightTime) &= 
            \expect(\layerWeightZero) + \left(\sum_{\instantTwo=1}^\instant \step_\instantTwo  \expect\left(\gradTimeTwo\right)  \right) ,\text{using \cref{eq:sgd_step}} \nonumber \\
            &= \mathbf{0} \quad\text{using \cref{eq:mean_gradTime}} \nonumber \\
            \implies \expect(\layerWeightConv) &= \lim_{t\rightarrow\infty} \expect(\layerWeightTime) \nonumber \\ 
            \implies \expect(\layerWeightConv) &= \layerWeightMean = \mathbf{0}
            \label{eq:mean_conv_weight}
        \end{align}
    
        We finally calculate the variance of the converged weight. Because the SGD step size is independent of the gradient, we write using \cref{eq:sgd_step}, 
        \begin{align}
            \var(\layerWeightTime) &= \var(\layerWeightZero) + \step_1^2\var\left( \gradient_1\right) + \step_2^2\var\left(\gradient_2\right) \nonumber\\
            &~~~~+ \cdots + \step_\instant^2\var\left(\gradTime\right)
        \end{align}
    
        Assuming the gradients $\gradTime$ are drawn from an identical distribution, we have
        \begin{align}
            \var(\layerWeightTime) &= \var(\layerWeightZero) + \left(\sum_{\instantTwo=1}^\instant \step_\instantTwo^2\right)  \var\left(\gradTime\right) \nonumber \\
            \implies \var(\layerWeightConv) &= \lim_{t\rightarrow\infty} \var(\layerWeightTime) \nonumber \\
           &= \var(\layerWeightZero) + \left(\lim_{t\rightarrow\infty} \sum_{\instantTwo=1}^\instant \step_\instantTwo^2\right)  \var\left(\gradTime\right) \nonumber \\ 
            \implies \var(\layerWeightConv) &= \var(\layerWeightZero) + \stepSumTrue \var\left(\gradTime\right)
            \label{eq:weight_bound_2}
        \end{align}
        An example of square summable step-sizes of SGD is $\step_\instantTwo= \frac{1}{\instantTwo}$, and then the constant $\stepSumTrue = \sum\limits_{\instantTwo=1}  \step_\instantTwo^2 = \frac{\pi^2}{6}$.
        This assumption is also satisfied by modern neural networks since their training steps are always finite. 
    
        Substituting \cref{eq:var_gradTime} in \cref{eq:weight_bound_2}, we have
        \begin{align}
            \var(\layerWeightConv) &= \var(\layerWeightZero) + \stepSumTrue \expect(\image^T\image)\var(\funcNoise)
            \label{eq:var_conv_weight}
        \end{align}
        Substituting mean and variances from \cref{eq:mean_conv_weight,eq:var_conv_weight} in \cref{eq:weight_bound_with_mean}, we have
        \begin{align}
            \expect\left(\norm{\layerWeightConv\!-\!\layerWeightOptimal}_2^2\right) 
            &= \var(\layerWeightZero) + \stepSumTrue \expect(\image^T\image)  \var(\funcNoise) \nonumber \\
            &\qquad + \expect(||\layerWeightOptimal||^2) \nonumber \\
            &= \stepSumTrue \expect(\image^T\image) \var(\funcNoise) + \var(\layerWeightZero) \nonumber \\
            &\qquad + \expect(||\layerWeightOptimal||^2)  \nonumber \\
            \implies \expect\left(\norm{\layerWeightConv\!-\!\layerWeightOptimal}_2^2\right) &= \stepConstant \var(\funcNoise) + \uselessConstant,
        \end{align}
        where $\funcNoise = \dfrac{\partial \loss(\noise)}{\partial \noise}$ is the gradient of the loss function wrt noise,  and $\stepConstant = \stepSumTrue \expect(\image^T\image)$ and $\uselessConstant$ are terms independent of the loss function $\loss$.

    \subsection{Comparison of Loss Functions}

        \cref{eqn:conv:weight:dist} shows that different losses $\loss$ lead to different $\var(\funcNoise)$. 
        Hence, comparing this term for different losses asseses the quality of losses.

        \subsubsection{Gradient Variance of MAE Loss}\label{sec:seabird_supp_var_lOne}
            The result on MAE $(\lOne)$ is well-known in the literature \cite{shalev2007pegasos, lacoste2012simpler}. 
            We reproduce the result from \cite{shalev2007pegasos, lacoste2012simpler} using our notations for completeness.
            
            The $\lOne$ loss is 
            \begin{align}
                \lOne(\noise) &= |\depthPred - \depthGT|_1 = |\layerWeightTime\image - \depthGT|_1 = |\noise|_1\nonumber \\
                \implies \funcNoise &= \dfrac{\partial \lOne(\noise)}{\partial \noise} = \sign(\noise) 
            \end{align}
            Thus, $\funcNoise = \sign(\noise)$ is a Bernoulli random variable with $p(\funcNoise) = 1/2 \text{ for } \funcNoise= \pm 1$. 
            So, mean $\expect(\funcNoise) = 0$ and variance $\var(\funcNoise) = 1$. 

        \subsubsection{Gradient Variance of MSE Loss}\label{sec:seabird_supp_var_lTwo} 
            The result on MSE $(\lTwo)$ is well-known in the literature \cite{shalev2007pegasos, lacoste2012simpler}. 
            We reproduce the result from \cite{shalev2007pegasos, lacoste2012simpler} using our notations for completeness.          
            The $\lTwo$ loss is 
            \begin{align}
                \lTwo(\noise) &= 0.5|\depthPred - \depthGT|^2 = 0.5|\noise|^2 = 0.5\noise^2\nonumber \\
                \implies \funcNoise &= \dfrac{\partial \lTwo(\noise)}{\partial \noise} = \noise
            \end{align}
            Thus, $\funcNoise = \noise$ is a normal random variable \cite{shalev2007pegasos}.
            So, mean $\expect(\funcNoise) = 0$ and variance $\var(\funcNoise) = \var(\noise) = \normalVar$.

        \subsubsection{Gradient Variance of \Dice Loss. (Proof of \cref{lemma:2})}\label{sec:seabird_supp_var_dice}

            \begin{proof}\let\qed\relax
                We first write the gradient of \dice loss as a function of noise $(\noise)$ as follows:
                \begin{align}
                    \funcNoise &= \dfrac{\partial \lDice(\noise)}{\partial \noise} = \begin{cases}
                        \dfrac{\sign(\noise)}{\length} \text{ , }|\noise|\le \length \\
                        0 \quad~~~~~~\text{ , }|\noise|\ge \length 
                    \end{cases} 
                \end{align}
                The gradient of the loss $\funcNoise$ is an odd function and so, its mean $\expect(\funcNoise) = 0$. Next, we write its variance $\var(\funcNoise)$ as
                \begin{align}
                    \var(\funcNoise) = \var(\noise) &= \frac{1}{\length^2}\int\limits_{-\length}^\length \dfrac{1}{\sqrt{2\pi}\normalSig} e^{-\frac{\noise^2}{2\normalVar}} d\noise \nonumber \\
                    &= \frac{2}{\length^2}\int\limits_{0}^\length \dfrac{1}{\sqrt{2\pi}\normalSig} e^{-\frac{\noise^2}{2\normalVar}} d\noise \nonumber \\
                    &= \frac{2}{\length^2}\int\limits_{0}^{\length/\normalSig} \dfrac{1}{\sqrt{2\pi}} e^{-\frac{\noise^2}{2}} d\noise \nonumber \\
                    &= \frac{2}{\length^2}\left[ \int\limits_{-\infty}^{\length/\normalSig} \dfrac{1}{\sqrt{2\pi}} e^{-\frac{\noise^2}{2}} d\noise  - \frac{1}{2}\right] \nonumber \\
                    &= \frac{2}{\length^2}\left[ \normalCDF\left(\frac{\length}{\normalSig}\right) - \frac{1}{2}\right] \label{eq:dice_cdf}\\
                    &\quad\text{~~~~~~~~~~~where, } \normalCDF \text{ is the normal CDF}\nonumber 
                \end{align}
                We write the CDF $\normalCDF(x)$ in terms of error function $\normalErf$ as:
                \begin{align}
                    \normalCDF(x) &= \dfrac{1}{2} + \dfrac{1}{2}\normalErf\left(\dfrac{x}{\sqrt{2}}\right)
                \end{align}
                $\text{for } x \ge 0$.
                Next, we put $x = \dfrac{\length}{\normalSig}$ to get
                \begin{align}
                     \normalCDF\left(\frac{\length}{\normalSig}\right) &= \frac{1}{2} + \frac{1}{2}\normalErf\left(\frac{\length}{\sqrt{2}\normalSig}\right)
                \end{align}
                Substituting above in \cref{eq:dice_cdf}, we obtain
                \begin{align}
                \var(\funcNoise) &= \frac{2}{\length^2}\left[ \frac{1}{2} + \frac{1}{2}\normalErf\left(\frac{\length}{\sqrt{2}\normalSig}\right) - \frac{1}{2}\right]\nonumber \\
                \implies \var(\funcNoise) &= \frac{1}{\length^2}\normalErf\left(\frac{\length}{\sqrt{2}\normalSig}\right)
                \end{align}
            \end{proof}

    \subsection[Proof of Dice Model Being Better]{Proof of Dice Model Being Better\cref{lemma:3}}\label{sec:seabird_supp_proof_lemma_3}
    \begin{proof}\let\qed\relax
        It remains sufficient to show that 
        \begin{align}
            \expect\left(\norm{\layerWeightConvDice-\layerWeightOptimal}_2\right) &\le \expect\left(\norm{\layerWeightConvReg-\layerWeightOptimal}_2\right) \nonumber \\
            \implies \expect\left(\norm{\layerWeightConvDice-\layerWeightOptimal}_2^2\right) &\le \expect\left(\norm{\layerWeightConvReg-\layerWeightOptimal}_2^2\right) \label{eq:show}
        \end{align}
        Using \cref{lemma:1}, the above comparison is a comparison between the gradient variance of the loss wrt noise $\var(\funcNoise)$. 
        Hence, we compute the gradient variance of the loss $\loss$, \thatIs, $\var(\funcNoise)$ of regression and \dice losses to derive this lemma.

        \noIndentHeading{Case 1 $\normalSig \le 1$:} Given \cref{tab:seabird_optimality_bounds}, if $\normalSig \le 1$, the minimum deviation in converged regression model comes from the $\lTwo$ loss. 
        The difference in the estimates of regression loss and the \dice loss 
        \begin{align}
            \expect\left(\norm{\layerWeightConvReg-\layerWeightOptimal}_2^2\right) &- \expect\left(\norm{\layerWeightConvDice-\layerWeightOptimal}_2^2\right)\nonumber\\
            &\propto \normalVar - \frac{1}{\length^2}\normalErf\left(\frac{\length}{\sqrt{2}\normalSig}\right) 
        \end{align}

        Let $\normalSigTh$ be the solution of the equation $\normalVar = \dfrac{1}{\length^2}\normalErf\left(\dfrac{\length}{\sqrt{2}\normalSig}\right)$.
        Note that the above equation has unique solution $\normalSigTh$ since $\normalVar$ is a strictly increasing function wrt $\normalSig$ for $\normalSig > 0$, while $\dfrac{1}{\length^2}\normalErf\left(\dfrac{\length}{\sqrt{2}\normalSig}\right)$ is a strictly decreasing function wrt $\normalSig$ for $\normalSig > 0$.
        If the noise has $\normalSig \ge \normalSigTh$, the RHS of the above equation $\ge 0$, which means \dice loss converges better than the regression loss. 

        \noIndentHeading{Case 2 $\normalSig \ge 1$:} Given \cref{tab:seabird_optimality_bounds}, if $\normalSig \ge 1$, the minimum deviation in converged regression model comes from the $\lOne$ loss. 
        The difference in the regression and \dice loss estimates: 
        \begin{align}
            \expect\left(\norm{\layerWeightConvReg-\layerWeightOptimal}_2^2\right) &- \expect\left(\norm{\layerWeightConvDice-\layerWeightOptimal}_2^2\right)\nonumber\\
            &\propto 1- \frac{1}{\length^2}\normalErf\left(\frac{\length}{\sqrt{2}\normalSig}\right) 
        \end{align}
        If the noise has $\normalSig \ge \dfrac{\sqrt{2}}{\length}\normalErfInv(\length^2)$, the RHS of the above equation $\ge 0$, which means \dice loss is better than the regression loss.
        For objects such as cars and trailers which have length $\length > 4m$, this is trivially satisfied.

        Combining both cases, \dice loss outperforms the $\lOne$ and $\lTwo$ losses if the noise deviation $\normalSig$ exceeds the critical threshold $\normalSigCr$, \thatIs
        \begin{align}
            \normalSig > \normalSigCr = \max \left( \normalSigTh, \dfrac{\sqrt{2}}{\length}\normalErfInv(\length^2)\right).
            \label{eq:final_bound} 
        \end{align}
    \end{proof}

    \subsection[Proof of Convergence Analysis]{Proof of Convergence Analysis \cref{th:seabird_1}}\label{sec:seabird_supp_proof_theorem_1}
    \begin{proof}
        Continuing from \cref{lemma:3}, the advantage of the trained weight obtained from \dice loss over the trained weight obtained from regression losses further results in
        \begin{align}
            \var(\layerWeightConvDice) &\le \var(\layerWeightConvReg) \nonumber \\
            \implies \expect(|\layerWeightConvDice\image-\depthGT|) &\le \expect(|\layerWeightConvReg\image-\depthGT|) \nonumber \\
            \implies \expect(|\depthPredDice-\depthGT|) &\le \expect(|\depthPredReg-\depthGT|) \nonumber \\
            \implies \expect(\!{}^d\iouThreeDMath) &\ge \expect(\!{}^r\iouThreeDMath),
        \end{align}
        assuming depth is the only source of error. 
        Because \apThreeD is an non-decreasing function of \iouThreeD, the inequality remains preserved. 
        Hence, we have ${}^d$\apThreeD $\ge {}^r$\apThreeD. 
    \end{proof}
    Thus, the average precision from the \dice model is better than the regression model, which means a better detector.

    \subsection{Properties of \Dice Loss.}\label{sec:seabird_supp_dice_properties}  

        We next explore the properties of model in \cref{lemma:3} trained with \dice loss.
        From \cref{lemma:1}, we write
        \begin{align}
            \expect\left(\norm{\layerWeightConvDice-\layerWeightOptimal}^2_2\right) 
            &= \stepConstant \var(\funcNoise) + \uselessConstant \nonumber
        \end{align}
        Substituting the result of \cref{lemma:2}, we have
        \begin{align}
            \expect\left(\norm{\layerWeightConvDice-\layerWeightOptimal}^2_2\right) &= \frac{\stepConstant}{\length^2}\normalErf\left(\frac{\length}{\sqrt{2}\normalSig} \right)+\uselessConstant
            \label{eq:dice_convergence}
        \end{align}

        chapter \cite{birnbaum1942inequality} says that for a normal random variable $X$ with mean $0$ and variance $1$ and for any $x > 0$, we have
        \begin{align}
            \frac{\sqrt{4+x^2}-x}{2} \sqrt{\frac{1}{2\pi}} e^{-\frac{x^2}{2}} &\le P\left(X > x\right) \nonumber \\
            \implies \frac{1}{x+\sqrt{4+x^2}} \sqrt{\frac{2}{\pi}} e^{-\frac{x^2}{2}} &\le P\left(X > x\right) \nonumber \\
            \implies \frac{1}{x+\sqrt{4+x^2}} \sqrt{\frac{2}{\pi}} e^{-\frac{x^2}{2}}  &\le 1 - P\left(X \le x\right) \nonumber \\
            \implies \frac{1}{x+\sqrt{4+x^2}} \sqrt{\frac{2}{\pi}} e^{-\frac{x^2}{2}}  &\le 1\!-\!\frac{1}{2}\!-\!\int_0^{x} \frac{1}{\sqrt{2\pi}} e^{-\frac{X^2}{2}} dX \nonumber \\
            \implies \frac{1}{x+\sqrt{4+x^2}} \sqrt{\frac{2}{\pi}} e^{-\frac{x^2}{2}}  &\le \frac{1}{2}\!-\!\int_0^{x} \frac{1}{\sqrt{2\pi}} e^{-\frac{X^2}{2}} dX \nonumber \\
            \implies \frac{1}{x+\sqrt{4+x^2}} \sqrt{\frac{2}{\pi}} e^{-\frac{x^2}{2}}  &\le \frac{1}{2} - \int_0^{\frac{x}{\sqrt{2}}} \frac{1}{\sqrt{\pi}} e^{-X^2} dX \nonumber \\
            \implies \frac{1}{x+\sqrt{4+x^2}} \sqrt{\frac{2}{\pi}} e^{-\frac{x^2}{2}}  &\le \frac{1}{2} - \frac{1}{2}\normalErf\left(\frac{x}{\sqrt{2}}\right) \nonumber \\
            \implies \normalErf\left(\frac{x}{\sqrt{2}}\right) &\le 1 - \frac{2}{x+\sqrt{4+x^2}} \sqrt{\frac{2}{\pi}} e^{-\frac{x^2}{2}} \nonumber
        \end{align}
        Substituting $x=\dfrac{\length}{\normalSig}$ above, we have,
        \begin{align}
            \normalErf\left(\frac{\length}{\sqrt{2}\normalSig} \right) &\le 1 - \frac{2\normalSig}{\length+\sqrt{4\normalVar+\length^2}} \sqrt{\frac{2}{\pi}} e^{-\frac{\length^2}{2\normalVar}}
            \label{eq:big}
        \end{align}

        \noIndentHeading{Case 1: Upper bound.}  
            The RHS of \cref{eq:big} is clearly less than $1$ since the term in the RHS after subtraction is positive.
            Hence, 
            \begin{align}
                \normalErf\left(\frac{\length}{\sqrt{2}\normalSig} \right) &\le 1 \nonumber
            \end{align}
            Substituting above in \cref{eq:dice_convergence}, we have 
            \begin{align}
                \expect\left(\norm{\layerWeightConvDice-\layerWeightOptimal}^2_2\right) &\le  \frac{\stepConstant }{\length^2} + \uselessConstant
            \end{align}
            Clearly, the deviation of the trained model with the \dice loss is inversely proportional to the object length $\length$. 
            The deviation from the optimal is less for large objects.

        \begin{table}[!t]
            \caption[Assumption comparison of Convergence Analysis vs \monoThreeD models]
            {\textbf{Assumption comparison} of Convergence Analysis of \cref{th:seabird_1} vs \monoThreeD models.}
            \label{tab:seabird_assumption_comp}
            \centering
            \scalebox{\scaleFraction}{
            \setlength\tabcolsep{0.15cm}
            \begin{tabular}{l m c m c }
                \addlinespace[0.01cm]
                & \cref{th:seabird_1} & \monoThreeD Models\\ 
                \myTopRule
                Regression & Linear & Non-linear \\
                Noise $\noise$ PDF & Normal & Arbitrary\\
                Noise \& Image & Independent & Dependent\\
                Object Categories & $1$ & Multiple \\
                Object Size $\length$ & Ideal & Non-ideal\\
                Error & Depth & All $7$ parameters\\
                Loss $\loss$ & $\lOne, \lTwo$, \dice & $\smoothLOne, \lTwo$, \dice, CE \\ 
                Optimizers & SGD & SGD, Adam, AdamW\\
                Global Optima & Unique & Multiple \\
            \end{tabular}
            }
        \end{table}

        \noIndentHeading{Case 2: Infinite Noise variance} $\normalVar \rightarrow \infty$. 
            Then, one of the terms in the RHS of \cref{eq:big} $\dfrac{2\normalSig}{\length+\sqrt{4\normalVar+\length^2}} \rightarrow 1$.
            Moreover, $\dfrac{\length}{\normalSig} \rightarrow 0 \implies e^{-\frac{\length^2}{2\normalVar}} \approx \left(1 - \dfrac{\length^2}{2\normalVar}\right)$. 
            So, RHS of \cref{eq:big} becomes
            \begin{align}
                \normalErf\left(\frac{\length}{\sqrt{2}\normalSig} \right) &\approx 1 - \sqrt{\frac{2}{\pi}} \left(1 - \frac{\length^2}{2\normalVar}\right) \nonumber \\
                \implies \normalErf\left(\frac{\length}{\sqrt{2}\normalSig} \right) &\approx \left( 1 + \sqrt{\frac{2}{\pi}} + \sqrt{\frac{2}{\pi}}\frac{\length^2}{2\normalVar}\right)
            \end{align}
            Substituting above in \cref{eq:dice_convergence}, we have 
            \begin{align}
            \expect\left(\norm{\layerWeightConvDice-\layerWeightOptimal}^2_2\right) &\approx \frac{\stepConstant }{\length^2} \left( 1 + \sqrt{\frac{2}{\pi}} + \sqrt{\frac{2}{\pi}}\frac{\length^2}{2\normalVar}\right) \nonumber \\ 
                &~~~~ + \uselessConstant 
            \end{align} 
            Thus, the deviation from the optimal weight is inversely proportional to the noise deviation $\normalVar$.
            Hence, the deviation from the optimal weight decreases as $\normalVar$ increases for the \dice loss.
            This property provides noise-robustness to the model trained with the \dice loss.

    \subsection{Notes on Theoretical Result}\label{sec:seabird_supp_theory_notes}

        \noIndentHeading{Assumption Comparisons.}
            The theoretical result of \cref{th:seabird_1} relies upon several assumptions. 
            We present a comparison between the assumptions made by \cref{th:seabird_1} and those underlying \monoThreeD models, in \cref{tab:seabird_assumption_comp}. 
            While our analysis depends on these assumptions, it is noteworthy that the results are apparent even in scenarios where the assumptions do not hold true.
            Another advantage of having a linear regression setup is that this setup has a unique global minima (because of its convexity).

\begin{figure}[!t]
            \centering
            \includegraphics[width=0.6\linewidth]{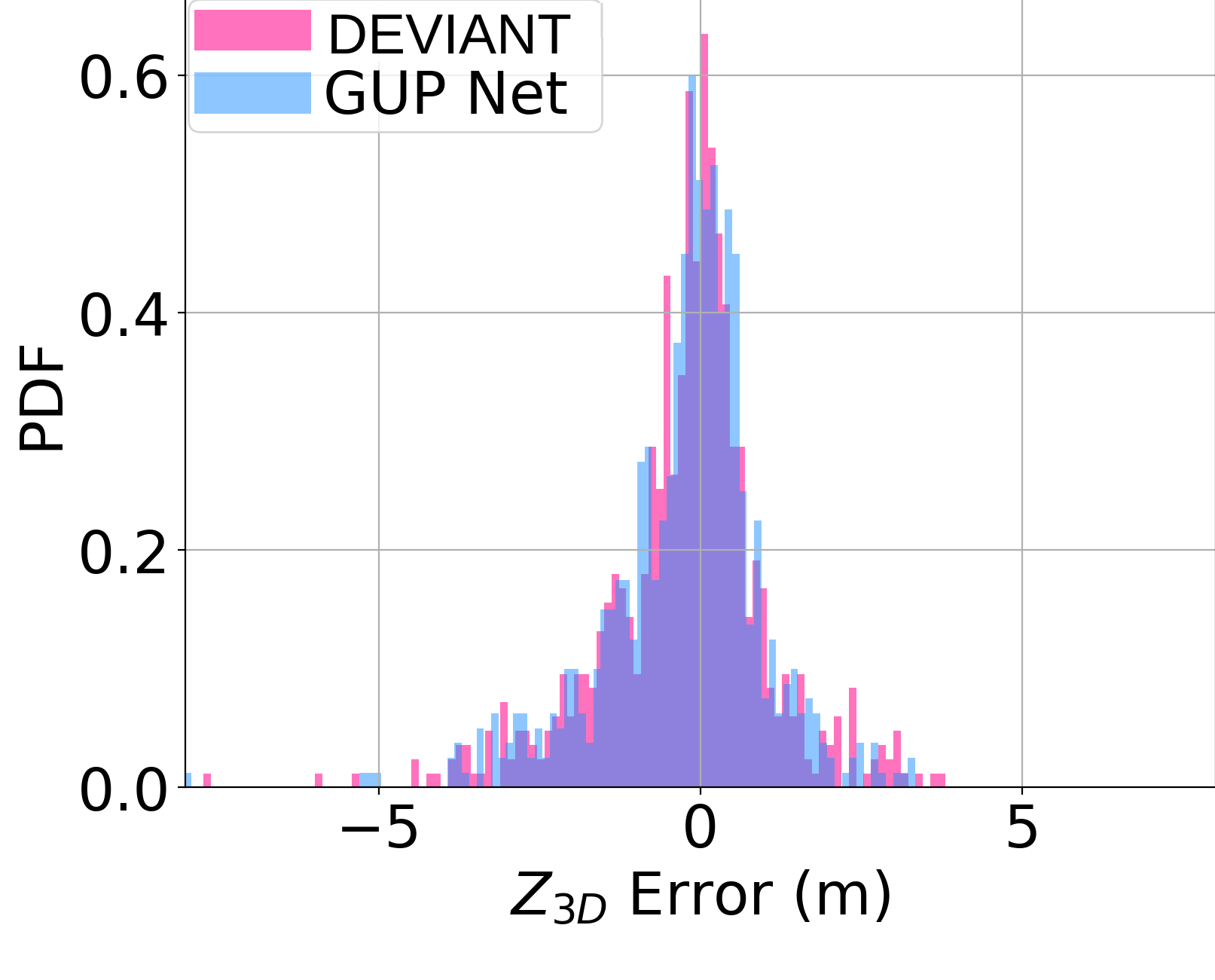}
            \caption[Depth error histogram of released \gupNet and \deviant on the \kitti \val cars.]
            {\textbf{Depth error histogram} of released \gupNet and \deviant \cite{kumar2022deviant} on the \kitti \val cars.
            The histogram shows that depth error is close to the Gaussian random variable.
            }
            \label{fig:seabird_depth_error_histogram}
        \end{figure}

        \noIndentHeading{Nature of Noise $\noise$.} 
            \cref{th:seabird_1} assumes that the noise $\noise$ is a normal random variable $\normal(0,\normalVar)$. 
            To verify this assumption, we take the two \sota released models \gupNet \cite{lu2021geometry} and \deviant \cite{kumar2022deviant} on the \kitti \cite{geiger2012we} \val cars. 
            We next plot the depth error histogram of both these models in \cref{fig:seabird_depth_error_histogram}.
            This figure confirms that the depth error is close to the Gaussian random variable.
            Thus, this assumption is quite realistic.

        \noIndentHeading{\cref{th:seabird_1} Requires Assumptions?}
            We agree that \cref{th:seabird_1} requires assumptions for the proof.
            However, our theory does have empirical support; most \monoThreeD works have no theory. 
            So, our theoretical attempt for \monoThreeD is a step forward! 
            We leave the analysis after relaxing some or all of these assumptions for future avenues.

        \noIndentHeading{Does \cref{th:seabird_1} Hold in Inference?}
            Yes, \cref{th:seabird_1} holds even in inference.
            \cref{th:seabird_1} relies on the converged weight $\layerWeightConv$, which in turn depends on the training data distribution. 
            Now, as long as the training and testing data distribution remains the same (a fundamental assumption in ML), \cref{th:seabird_1} holds also during inference.

    \subsection{More Discussions}

        \noIndentHeading{\seabird improves because it removes depth estimation and integrates \bev segmentation}.
            We clarify to remove this confusion. 
            First, \seabird also estimates depth. 
            \seabird depth estimates are better because of good segmentation, a \textit{form} of depth (thanks to \dice loss). 
            Second, predicted \bev segmentation needs processing with the \threeD head to output depth; so it can not replace depth estimation. 
            Third, integrating segmentation over all categories degrades \monoThreeD performance (\cite{li2022bevformer} and our \cref{tab:seabird_ablation} Sem. Category).

        \noIndentHeading{Why evaluation on outdoor datasets?}
            We experiment with outdoor datasets in this chapter because indoor datasets rarely have large objects (mean length $>6m$).

\section{Implementation Details}\label{sec:seabird_supp_implement_details}

        \noIndentHeading{Datasets.}
            Our experiments use the publicly available \kittiThreeSixty, \kittiThreeSixtyPanoptic and \nuscenes datasets.
            \kittiThreeSixty is available at \url{https://www.cvlibs.net/datasets/kitti-360/download.php} under CCA-NonCommercial-ShareAlike 
            (CC BY-NC-SA) 3.0 License.
            \kittiThreeSixtyPanoptic is available at \url{http://panoptic-bev.cs.uni-freiburg.de/} under Robot Learning License Agreement.
            \nuscenes is available at \url{https://www.nuscenes.org/nuscenes} under CC BY-NC-SA 4.0 International Public License.

        \noIndentHeading{Data Splits.}
            We detail out the detection data split construction of the \kittiThreeSixty dataset.
            \begin{itemize}
                \item \textit{\kittiThreeSixty Test split}: This detection benchmark \cite{liao2022kitti360} contains $300$ training and $42$ testing windows. 
                These windows contain $61{,}056$ training and $9{,}935$ testing images.
                The calibration exists for each frame in training, while it exists for every $10^\text{th}$ frame in testing.
                Therefore, our split consists of $61{,}056$ training images, while we run monocular detectors on $910$ test images (ignoring uncalibrated images).
    
                \item \textit{\kittiThreeSixty \val split}: The \kittiThreeSixty detection \val split partitions the official train into $239$ train and $61$ validation windows \cite{liao2022kitti360}. 
                The original \val split \cite{liao2022kitti360} contains $49{,}003$ training and $14{,}600$ validation images.
                However, this original \val split has the following three issues:
                \begin{itemize}
                    \item Data leakage (common images) exists in the training and validation windows.
                    \item Every \kittiThreeSixty image does not have the corresponding \bev semantic segmentation GT in the \kittiThreeSixtyPanoptic \cite{gosala2022bev} dataset, making it harder to compare \monoThreeD and \bev segmentation performance.
                    \item The \kittiThreeSixty validation set has higher sampling rate compared to the testing set.
                \end{itemize}
                To fix the data leakage issue, we remove the common images from training set and keep them only in the validation set. 
                Then, we take the intersection of \kittiThreeSixty and \kittiThreeSixtyPanoptic datasets to ensure that every image has corresponding \bev segmentation segmentation GT.
                After these two steps, the training and validation set contain $48{,}648$ and $12{,}408$ images with calibration and semantic maps. 
                Next, we subsample the validation images by a factor of $10$ as in the testing set. 
                Hence, our \kittiThreeSixty \val split contains $48{,}648$ training images and $1{,}294$ validation images.
                
            \end{itemize}

        \begin{figure}[!t]
            \centering
            \includegraphics[width=\linewidth]{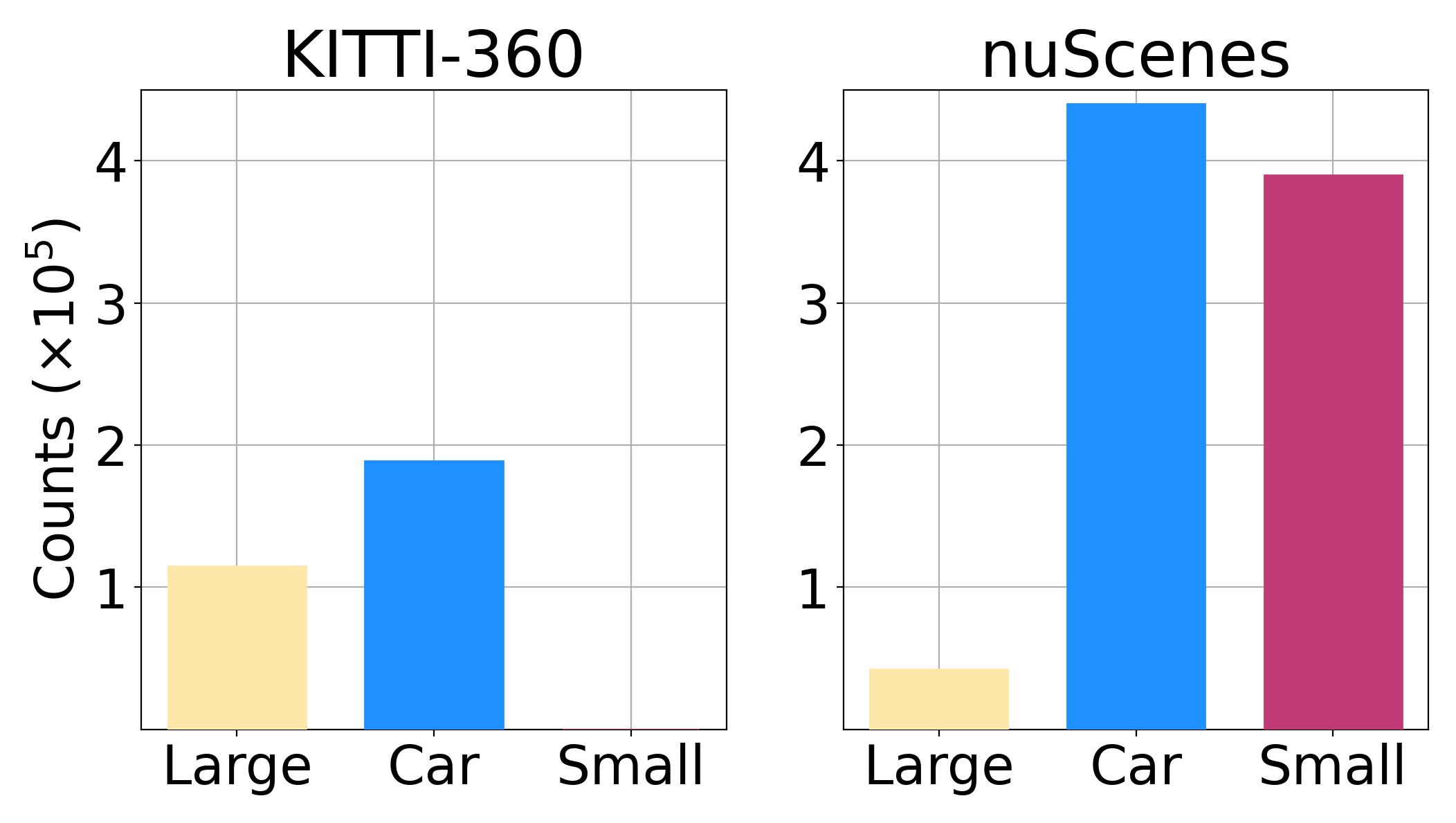}
            \caption[Skewness in datasets.]{
                \textbf{Skewness in datasets.} 
                The ratio of {large} (yellow) objects to other objects is approximately $1\!:\!2$ in \kittiThreeSixty \cite{liao2022kitti360}, while the skewness is about $1\!:\!21$ in \nuscenes \cite{caesar2020nuscenes}.
            }
            \label{fig:seabird_skew}
        \end{figure}

        \noIndentHeading{Augmentation.}
            We keep the same augmentation strategy as our baselines for the respective models.

        \noIndentHeading{Pre-processing.} We resize images to preserve their aspect ratio.
            \begin{itemize}
                \item \textit{\kittiThreeSixty.}
                    We resize the $[376, 1408]$ sized \kittiThreeSixty images, and bring them to the $[384, 1438]$ resolution.
            
                \item \textit{\nuscenes.}
                    We resize the $[900, 1600]$ sized \nuscenes images, and bring them to the $[256, 704]$, $[512, 1408]$ and $[640, 1600]$ resolutions as our baselines \cite{zhang2022beverse,zong2023hop}. 
            \end{itemize}

        \noIndentHeading{Libraries.}
            \imageToMaps and \panopticBEV experiments use PyTorch \cite{paszke2019pytorch}, while \beVerse and \hop use MMDetection3D \cite{mmdet3d2020}.

        \noIndentHeading{Architecture.}
            \begin{itemize}
                \item \textit{\imageToMapsWithMethod.}\imageToMaps \cite{saha2022translating} uses \resNetEighteen as the backbone with the standard Feature Pyramid Network (FPN) \cite{lin2017feature} and a transformer to predict depth distribution. 
                FPN is a bottom-up feed-forward CNN that computes feature maps with a downscaling factor of $2$, and a top-down network that brings them back to the high-resolution ones.
                There are total four feature maps levels in this FPN.
                We use the \orBoxNet with \resNetEighteen \cite{he2016deep} as the detection head.
                \item \textit{\panopticBEVWithMethod.}\panopticBEV \cite{gosala2022bev} uses \efficientDet \cite{tan2020efficientdet} as the backbone.
                We use \orBoxNet with \resNetEighteen \cite{he2016deep} as the detection head.
                \item \textit{\beVerseWithMethod.} \beVerse \cite{zhang2022beverse} uses Swin transformers \cite{liu2021swin} as the backbones.
                We use the original heads without any configuration change.
                \item \textit{\hopWithMethod.} \hop \cite{zong2023hop} uses \resNetFifty,  \resNetOneHundredOne \cite{he2016deep} and \vovNet \cite{park2021pseudo} as the backbones.
                Since \hop does not have the segmentation head, we use the one in \beVerse as the segmentation head.
            \end{itemize}
            We initialize the CNNs and transformers from \imageNet weights except for \vovNet, which is pre-trained on $15$ million \lidar data..
            We output two and ten foreground categories for \kittiThreeSixty and \nuscenes datasets respectively.

        \noIndentHeading{Training.}
            We use the training protocol as our baselines for all our experiments.
            We choose the model saved in the last epoch as our final model for all our experiments.
            \begin{itemize}
                \item \textit{\imageToMapsWithMethod.} 
                Training uses the Adam optimizer \cite{kingma2014adam}, a batch size of $30$, an exponential decay of $0.98$ \cite{saha2022translating} and gradient clipping of $10$ on single Nvidia A100 ($80$GB) GPU.
                We train the \bev Net in the first stage with a learning rate $1.0\!\times\!10^{-4}$ for $50$ epochs \cite{saha2022translating} .
                We then add the detector in the second stage and finetune with the first stage weight with a learning rate $0.5\!\times\!10^{-4}$ for $40$ epochs. 
                Training on \kittiThreeSixty \val takes a total of $100$ hours.
                For Test models, we finetune \imageToMaps \val stage 1 model with train+val data for $40$ epochs.
                \item \textit{\panopticBEVWithMethod.} 
                Training uses the Adam optimizer \cite{kingma2014adam} with Nesterov, a batch size of $2$ per GPU on eight Nvidia RTX A6000 ($48$GB) GPU.
                We train the \panopticBEV with the \dice loss in the first stage with a learning rate $2.5\!\times\!10^{-3}$ for $20$ epochs.
                We then add the \orBoxNet in the second stage and finetune with the first stage weight with a learning rate $2.5\!\times\!10^{-3}$ for $20$ epochs. 
                \panopticBEV decays the learning rate by $0.5$ and $0.2$ at $10$ and $15$ epoch respectively. 
                Training on \kittiThreeSixty \val takes a total of $80$ hours.
                For Test models, we finetune \panopticBEV \val stage 1 model with train+val data for $10$ epochs on four GPUs.
                \item \textit{\beVerseWithMethod.}
                Training uses the AdamW optimizer \cite{loshchilov2019decoupled}, a sample size of $4$ per GPU, the one-cycle policy \cite{zhang2022beverse} and gradient clipping of $35$ on eight Nvidia RTX A6000 ($48$GB) GPU \cite{zhang2022beverse}.
                We train the segmentation head in the first stage with a learning rate $2.0\!\times\!10^{-3}$ for $4$ epochs. 
                We then add the detector in the second stage and finetune with the first stage weight with a learning rate $2.0\!\times\!10^{-3}$ for $20$ epochs \cite{zhang2022beverse}. 
                Training on \nuscenes takes a total of $400$ hours.
                \item \textit{\hopWithMethod.}
                Training uses the AdamW optimizer \cite{loshchilov2019decoupled}, a sample size of $2$ per GPU, and gradient clipping of $35$ on eight Nvidia A100 ($80$GB) GPUs \cite{zong2023hop}.
                We train the segmentation head in the first stage with a learning rate $1.0\!\times\!10^{-4}$ for $4$ epochs. 
                We then add the detector in the second stage and finetune with the first stage weight with a learning rate $1.0\!\times\!10^{-4}$ for $24$ epochs \cite{zhang2022beverse}. 
                \nuscenes training takes a total of $180$ hours.
                For Test models, we finetune val model with train+val data for $4$ more epochs.
            \end{itemize}

        \noIndentHeading{Losses.}
            We train the \bev Net of \seabird in Stage 1 with the dice loss.
            We train the final \seabird pipeline in Stage 2 with the following loss:
            \begin{align}
                \loss &= \loss_{det} + \weightSeg \loss_{seg},
            \end{align}
            with $\loss_{seg}$ being the dice loss and $\weightSeg$ being the weight of the dice loss in the baseline. 
            We keep the $\weightSeg = 5$.
            If the segmentation loss is itself scaled such as \panopticBEV uses the $\loss_{seg}$ as $7$, we use $\weightSeg = 35$ with detection.

        \noIndentHeading{Inference.} 
            We report the performance of all \kittiThreeSixty and \nuscenes models by inferring on single GPU card.
            Our testing resolution is same as the training resolution. 
            We do not use any augmentation for test/validation. 
            
            We keep the maximum number of objects is $50$ per image for \kittiThreeSixty models. 
            We use score threshold of $0.1$ for \kittiThreeSixty models and class dependent threshold for \nuscenes models as in \cite{zhang2022beverse}.
            \kittiThreeSixty evaluates on windows and not on images. 
            So, we use a \threeD center-based NMS \cite{kumar2021groomed} to convert image-based predictions to window-based predictions for \seabird and all our \kittiThreeSixty baselines.
            This NMS uses a threshold of $4$m for all categories, and keeps the highest score \threeD box if multiple \threeD boxes exist inside a window.

\section{Additional Experiments and Results}\label{sec:seabird_additional_exp}

        \begin{table}[!t]
            \caption[Error analysis on \kittiThreeSixty \val]
            {\textbf{Error analysis} on \kittiThreeSixty \val.
            }
            \label{tab:seabird_error_analysis}
            \centering
            \scalebox{\scaleFraction}{
            \setlength\tabcolsep{0.05cm}
            \begin{tabular}{ccccccc m bcc m bcc}
                \addlinespace[0.01cm]
                \multicolumn{7}{cm}{Oracle} & \multicolumn{3}{cm}{\apThreeDFifty~\bracketPercentage (\uparrowRHDSmall)} & \multicolumn{3}{c}{\apThreeDTwentyFive~\bracketPercentage (\uparrowRHDSmall)}\\ 
                $x$ & $y$ & $z$ & $l$ & $w$ & $h$ & $\theta$ & \MAPLarge & \MAPCar & \MAP & \MAPLarge & \MAPCar & \MAP\\
                \myTopRule
                & & & & & & & $8.71$	& $43.19$ & $25.95$	& $35.76$	& $52.22$ & $43.99$\\
                \cmark & & & & & & & $9.78$	& $41.63$ & $25.70$	& $36.07$	& $50.63$ & $43.35$\\
                & \cmark  && & & & & $9.57$	& $46.08$ & $27.82$	& $34.65$	& $53.03$ & $43.84$\\
                &&  \cmark & & & & & $9.90$	& $42.32$ & $27.11$	& $39.66$	& $53.08$ & $46.37$\\
                \cmark & \cmark & \cmark & & & & & $19.90$	& $47.37$ & $33.63$	& $41.84$	& $52.53$ & $47.19$\\
                & & & \cmark & \cmark & \cmark & & $9.49$	& $45.67$ & $27.58$	& $33.43$	& $51.53$ & $42.48$\\
                \cmark & \cmark & \cmark & \cmark & \cmark & \cmark & & $37.09$	& $46.27$ & $41.68$ & $44.58$	& $51.15$	& $47.87$\\
                \cmark & \cmark & \cmark & \cmark & \cmark & \cmark & \cmark & $37.02$	& $47.03$ & $42.02$ & $44.46$	& $51.50$	& $47.98$\\
                \hline
            \end{tabular}
            }
        \end{table}

    We now provide additional details and results of the experiments evaluating \seabird{}’s performance.           

    \subsection{\kittiThreeSixty \val Results}

        \noIndentHeading{Error Analysis.}
            We next report the error analysis of the \seabird in \cref{tab:seabird_error_analysis} by replacing the predicted box data with the oracle box data as in \cite{ma2021delving}.
            We consider the GT box to be an oracle box for predicted box if the euclidean distance is less than $4m$. 
            In case of multiple GT being matched to one box, we consider the oracle with the minimum distance.
            \cref{tab:seabird_error_analysis} shows that depth is the biggest source of error for \monoThreeD task as also observed in \cite{ma2021delving}. 
            Moreover, the oracle does not lead to perfect results since the \kittiThreeSixtyPanoptic GT \bev semantic is only upto $50m$, while the \kittiThreeSixty evaluates all objects (including objects beyond $50m$).

        \begin{table}[!t]
            \caption[Complexity analysis on \kittiThreeSixty \val.]
            {\textbf{Complexity analysis} on \kittiThreeSixty \val.}
            \label{tab:seabird_complexity_analysis}
            \centering
            \scalebox{\scaleFraction}{
            \setlength\tabcolsep{0.08cm}
            \begin{tabular}{cm c m c c c}
                Method & \monoThreeD & Inf. Time (s) & Param (M) & Flops (G)\\
                \myTopRule
                \gupNet \cite{lu2021geometry} & \cmark & $0.02$ & $16$ & $30$\\
                \deviant \cite{kumar2022deviant} & \cmark & $0.04$ & $16$ & $235$\\
                \imageToMaps \cite{saha2022translating} & \xmark & $0.01$ & $40$ & $80$\\
                \imageToMapsWithMethod & \cmark & $0.02$ & $53$ & $130$\\
                \panopticBEV \cite{gosala2022bev} & \xmark & $0.14$ & $24$ & $229$\\
                \panopticBEVWithMethod & \cmark & $0.15$ & $37$ & $279$\\ 
            \end{tabular}
            }
        \end{table} 

        \begin{table}[!t]
            \caption[\kittiThreeSixty \val results with naive baseline finetuned for large objects.]
            {\textbf{\kittiThreeSixty \val results with naive baseline finetuned for large objects.}
            \seabird pipelines comfortably outperform this naive baseline on large objects.
            [Key: \firstKey{Best}, \secondKey{Second Best}, \retrained= Retrained]
            }
            \label{tab:seabird_det_seg_results_kitti_360_val_naive}
            \centering
            \scalebox{0.8}{
            \setlength\tabcolsep{0.15cm}
            \begin{tabular}{l | c m bcc | bcc m ccc}
                \addlinespace[0.01cm]
                \multirow{2}{*}{Method} & \multirow{2}{*}{Venue} &\multicolumn{3}{c|}{\apThreeDFifty~\bracketPercentage (\uparrowRHDSmall)} & \multicolumn{3}{cm}{\apThreeDTwentyFive~\bracketPercentage (\uparrowRHDSmall)} & \multicolumn{3}{c}{\bev Seg \iou~\bracketPercentage (\uparrowRHDSmall)}\\ 
                & & \MAPLarge & \MAPCar & \MAP & \MAPLarge & \MAPCar & \MAP & Large & Car & \meanFor \\
                \myTopRule
                \gupNet\retrained \cite{lu2021geometry} & ICCV21 &
                ${0.54}$ & \first{45.11} & ${22.83}$ & $0.98$ & ${50.52}$ & ${25.75}$ & \mathDash	& \mathDash	& \mathDash	\\
                \gupNet (Large FT)~\retrained \cite{lu2021geometry} & ICCV21 &
                ${0.56}$ & \mathDash & ${0.28}$ & $2.56$ & \mathDash & ${1.28}$ & \mathDash	& \mathDash	& \mathDash	\\
                \rowcolor{my_gray}\textbf{\imageToMapsWithMethod} & CVPR24 &
                \second{8.71}	& \second{43.19} & \second{25.95}	& \second{35.76}	& \second{52.22} & \second{43.99} & \second{23.23}	& \second{39.61}	& \second{31.42}	\\
                \rowcolor{my_gray}\textbf{\panopticBEVWithMethod} & CVPR24 &
                \first{13.22}	& $42.46$ & \first{27.84}	& \first{37.15}	& \first{52.53} & \first{44.84} & \first{24.30}	& \first{48.04}	& \first{36.17}	\\
            \end{tabular}
            }
        \end{table}

        \begin{table}[!t]
            \caption[Impact of denoising \bev segmentation maps with \mirNet on \kittiThreeSixty \val.]
            {\textbf{Impact of denoising} \bev segmentation maps with \mirNet \cite{zamir2022learning} on \kittiThreeSixty \val with \imageToMapsWithMethod. Denoising does not help.
            [Key: \bestKey{Best}]
            }
            \label{tab:seabird_ablation_more}
            \centering
            \scalebox{\scaleFraction}{
            \setlength\tabcolsep{0.15cm}
            \begin{tabular}{c m bcc | bcc m cccc}
                \addlinespace[0.01cm]
                \multirow{2}{*}{Denoiser} & \multicolumn{3}{c|}{\apThreeDFifty~\bracketPercentage (\uparrowRHDSmall)} & \multicolumn{3}{cm}{\apThreeDTwentyFive~\bracketPercentage (\uparrowRHDSmall)} & \multicolumn{3}{c}{\bev Seg \iou~\bracketPercentage (\uparrowRHDSmall)}\\ 
                & \MAPLarge & \MAPCar & \MAP & \MAPLarge & \MAPCar & \MAP & Large & Car & \meanFor\\
                \myTopRule
                \cmark & $2.73$ & \best{43.77} & $23.25$ & $14.34	$ & $51.23$ & $32.79$ & $21.42$ & \best{39.72} & $30.57$ \\
                \xmark & 
                \best{8.71}	& $43.19$ & \best{25.95}	& \best{35.76}	& \best{52.22} & \best{43.99} & \best{23.23}	& $39.61$ & \best{31.42}	\\
            \end{tabular}
            }
        \end{table}

        \noIndentHeading{Computational Complexity Analysis.}
            We next compare the complexity analysis of \seabird pipeline in \cref{tab:seabird_complexity_analysis}. 
            For the flops analysis, we use the fvcore library as in \cite{kumar2022deviant}.

        \noIndentHeading{Naive baseline for Large Objects.}
            We next compare \seabird against a naive baseline for large objects detection, such as by fine-tuning \gupNet only on larger objects.
            \cref{tab:seabird_det_seg_results_kitti_360_val_naive} shows that \seabird pipelines comfortably outperform this baseline as well.

        \noIndentHeading{Does denoising \bev images help?}
            Another potential addition to the \seabird framework is using a denoiser between segmentation and detection heads. 
            We use the \mirNet \cite{zamir2022learning} as our denoiser and train the \bev segmentation head, denoiser and detection head in an end-to-end manner.
            \cref{tab:seabird_ablation_more} shows that denoising does not increase performance but the inference time.
            Hence, we do not use any denoiser for \seabird.
    
        \noIndentHeading{Sensitivity to Segmentation Weight.}
            We next study the impact of segmentation weight on \imageToMapsWithMethod in \cref{tab:seabird_det_seg_results_sensitivity} as in \cref{sec:seabird_ablation}.
            \cref{tab:seabird_det_seg_results_sensitivity} shows that $\weightSeg=5$ works the best for the \monoThreeD of large objects.

        \begin{table}[!t]
            \caption[Segmentation loss weight $\weightSeg$ sensitivity on \kittiThreeSixty \val with \imageToMapsWithMethod.]
            {\textbf{Segmentation loss weight $\weightSeg$ sensitivity} on \kittiThreeSixty \val with \imageToMapsWithMethod.
            $\weightSeg=5$ works the best.
            [Key: \bestKey{Best}]
            }
            \label{tab:seabird_det_seg_results_sensitivity}
            \centering
            \scalebox{\scaleFraction}{
            \setlength\tabcolsep{0.15cm}
            \begin{tabular}{c m bcc m bcc m ccc}
                \addlinespace[0.01cm]
                \multirow{2}{*}{$\weightSeg$} & \multicolumn{3}{cm}{\apThreeDFifty~\bracketPercentage (\uparrowRHDSmall)} & \multicolumn{3}{cm}{\apThreeDTwentyFive~\bracketPercentage (\uparrowRHDSmall)} & \multicolumn{3}{c}{\bev Seg \iou~\bracketPercentage (\uparrowRHDSmall)}\\ 
                & \MAPLarge & \MAPCar & \MAP & \MAPLarge & \MAPCar & \MAP & Large & Car & \meanFor\\
                \myTopRule
                $0$ &$4.86$ & \best{45.09} & $24.98$ & $26.33$ & $52.31$ & $39.32$ & $0$ & $7.07$ & $3.54$ \\
                $1$ & $7.07$ & $41.71$ & $24.39$ & $32.92$ & $52.9$ & $42.91$ & $23.78$ & $40.58$ & $32.18$\\
                $3$ & $7.26$ & $43.45$ & $25.36$ & $34.47$ & \best{52.54} & $43.51$ & \best{23.40} & \best{40.15} & \best{31.78}\\
                $5$ & \best{8.71}	& $43.19$ & \best{25.95}	& \best{35.76} & $52.22$ & \best{43.99} & $23.23$	& $39.61$	& $31.42$ \\
                $10$ & $7.69$ & $43.41$ & $25.55$ & $34.22$ & $50.97$ & $42.60$ & $22.15$ & $39.83$ & $30.99$\\
            \end{tabular}
            }
        \end{table}
        
        \begin{table}[!t]
            \caption[Reproducibility results on \kittiThreeSixty \val with \imageToMapsWithMethod.]
            {\textbf{Reproducibility results} on \kittiThreeSixty \val with \imageToMapsWithMethod.
            \seabird outperforms \seabird without dice loss in the median and average cases.
            [Key: \firstKey{Best}, \secondKey{Second Best}]
            }
            \label{tab:seabird_det_seg_results_kitti_val_repeat}
            \centering
            \scalebox{\scaleFraction}{
            \setlength\tabcolsep{0.15cm}
            \begin{tabular}{c m c m bcc m bcc m cccc}
                \addlinespace[0.01cm]
                \multirow{2}{*}{Dice} &\multirow{2}{*}{Seed} & \multicolumn{3}{cm}{\apThreeDFifty~\bracketPercentage (\uparrowRHDSmall)} & \multicolumn{3}{cm}{\apThreeDTwentyFive~\bracketPercentage (\uparrowRHDSmall)} & \multicolumn{3}{c}{\bev Seg \iou~\bracketPercentage (\uparrowRHDSmall)}\\ 
                & & \MAPLarge & \MAPCar & \MAP & \MAPLarge & \MAPCar & \MAP & Large & Car & \meanFor \\
                \myTopRule
                \multirow{4}{*}{\xmark}
                & $111$ &$3.81$ & $44.63$ & $24.22$ & $24.96$ & $53.15$ & $39.06$ & $0$ & $5.99$ & $3.00$ \\
                & $444$ &$4.86$ & $45.09$ & $24.98$ & $26.33$ & $52.31$ & $39.32$ & $0$ & $7.07$ & $3.54$ \\
                & $222$ &$5.79$ & $46.71$ & $26.25$ & $24.32$ & $54.06$ & $39.19$ & $0$ & $5.32$ & $2.66$ \\
                \rowcolor{my_gray}& Avg & $4.82$ & $45.58$ & \second{25.15} & $25.20$ & $53.17$ & \second{39.19} & $0$ & $6.13$ & $3.06$\\
                \hline
                \multirow{4}{*}{\cmark}
                &  $111$ & $7.87$ & $44.03$ & $25.95$ & $33.55$ & $53.93$ & $43.74$ & $22.64$ & $40.64$ & $31.64$\\
                &  $444$ & $8.71$	& $43.19$ & $25.95$	& $35.76$ & $52.22$ & $43.99$ & $23.23$	& $39.61$	& $31.42$\\
                &  $222$ & $8.71$ & $42.87$ & $25.79$ & $34.71$ & $51.72$ & $43.22$ & $22.74$ & $40.01$ & $31.38$ \\
                \rowcolor{my_gray}& Avg & $8.43$ & $43.36$ & \first{25.90} & $34.67$ & $52.62$ & \first{43.65} & $22.87$ & $40.09$ & $31.48$ \\
            \end{tabular}
            }
        \end{table}

        \begin{table}[!t]
            \caption[\Dice vs regression on methods with depth estimation.]
            {\textbf{\Dice vs regression on methods with depth estimation}. 
            \Dice model again outperforms regression loss models, particularly for large objects.
            [Key: 
            \firstKey{Best}, \secondKey{Second Best}]
            }
            \label{tab:seabird_dice_vs_regression_on_depth}
            \centering
            \scalebox{\scaleFraction}{
            \setlength\tabcolsep{0.08cm}
            \begin{tabular}{l  l | c c c m b c c c c}
            Resolution & Method & BBone & Venue & Loss &\MAPLarge (\uparrowRHDSmall)& \MAPCar (\uparrowRHDSmall) & \MAPSmall (\uparrowRHDSmall) & \MAP (\uparrowRHDSmall) & \NDS (\uparrowRHDSmall) \\
            \myTopRule
            \multirow{4}{*}{$256\!\times\!704$} & \multirow{4}{*}{\hop\!+\seabird} & \multirow{4}{*}{R50} & ICCV23 & \mathDash & \second{27.4}    & \second{57.2}    & \second{46.4} & \second{39.9} & \second{50.9}\\
            & & & \mathDash & $\lOne$ & $27.0$    & $57.1$    & $46.5$ & $39.7$ & $50.7$\\
            & & & \mathDash & $\lTwo$ & \CYMyFix & \multicolumn{4}{c}{Did Not Converge} \\
            & & & CVPR24 & Dice & \first{28.2}    & \first{58.6}    & \first{47.8}    & \first{41.1}    & \first{51.5} \\
            \end{tabular}
        }
        \end{table}

        \noIndentHeading{Reproducibility.}
            We ensure reproducibility of our results by repeating our experiments for $3$ random seeds.
            We choose the final epoch as our checkpoint in all our experiments as \cite{kumar2022deviant}.
            \cref{tab:seabird_det_seg_results_kitti_val_repeat} shows the results with these seeds. 
            \seabird outperforms \seabird without \dice loss in the median and average cases. 
            The biggest improvement shows up on larger objects.

    \subsection{\nuscenes Results}
        
        \noIndentHeading{Extended \val Results.}
            Besides showing improvements upon existing detectors in \cref{tab:seabird_nuscenes_val} on the \nuscenes \val split, we compare with more recent \sota detectors with large backbones in \cref{tab:seabird_nuscenes_val_more}. 

        \noIndentHeading{Dice vs regression on depth estimation methods}.
            We report \hop+R50 config, which uses depth estimation and compare losses in \cref{tab:seabird_dice_vs_regression_on_depth}.
            \cref{tab:seabird_dice_vs_regression_on_depth} shows that \Dice model again outperforms regression loss models.

        \noIndentHeading{\seabird Compatible Approaches.}
             \seabird conditions the detection outputs on segmented \bev features and so, requires foreground \bev segmentation.
             So, all approaches which produce latent \bev map in \cref{tab:seabird_nuscenes_test,tab:seabird_nuscenes_val} are compatible with \seabird.
             However, approaches which do not produce \bev features such as \sparseBEV \cite{liu2023sparsebev} are incompatible with \seabird.

        \begin{table}[!t]
            \caption[\nuscenes \val Detection results.]
            {\textbf{\nuscenes \val Detection results}. 
            \seabird pipelines outperform the baselines, particularly for large objects.
            [Key: 
            \firstKey{Best}, \secondKey{Second Best}, 
            B= Base, S= Small, T= Tiny, \released= Released, \reimplemented= Reimplementation, \cbgs= CBGS]
            }
            \label{tab:seabird_nuscenes_val_more}
            \centering
            \scalebox{0.8}{
            \setlength\tabcolsep{0.15cm}
            \begin{tabular}{l  l | c c m b c c c c}
                Resolution & Method & BBone & Venue &\MAPLarge (\uparrowRHDSmall)& \MAPCar (\uparrowRHDSmall) & \MAPSmall (\uparrowRHDSmall) & \MAP (\uparrowRHDSmall) & \NDS (\uparrowRHDSmall) \\
                \myTopRule
                \multirow{7}{*}{$256\!\times\!704$}
                & CAPE\released\cite{xiong2023cape} & R50 & CVPR23 & $18.5$    & $53.2$    & $38.1$    & $31.8$    & $44.2$\\
                & \petrVTwo\cite{liu2023petrv2} & R50 & ICCV23 & \mathDash    & \mathDash    & \mathDash    & $34.9$    & $45.6$    \\
                & SOLOFusion\released\cbgs\cite{park2022time} & R50 & ICLR23 & $26.5$	& \second{57.3} & \first{48.5} & \second{40.6} & $49.7$\\
                & \beVerseTiny\released\cite{zhang2022beverse}& Swin-T & \arxiv & $18.5$        & $53.4$ & $38.8$           & $32.1$        & $46.6$           \\
                & \cellcolor{my_gray}\textbf{\beVerseTinyWithMethod}         & \cellcolor{my_gray}Swin-T & \cellcolor{my_gray}CVPR24 & \cellcolor{my_gray}$19.5$ & \cellcolor{my_gray}$54.2$ & \cellcolor{my_gray}$41.1$    & \cellcolor{my_gray}$33.8$ & \cellcolor{my_gray}$48.1$\\
                &\hop\released\cite{zong2023hop}  & R50 & ICCV23 & \second{27.4}    & $57.2$    & $46.4$ & $39.9$ & \second{50.9} \\
                & \cellcolor{my_gray}\textbf{\hopWithMethod}      & \cellcolor{my_gray}R50 & \cellcolor{my_gray}CVPR24 & \cellcolor{my_gray}\first{28.2}    & \cellcolor{my_gray}\first{58.6}    & \cellcolor{my_gray}\second{47.8} & \cellcolor{my_gray}\first{41.1} & \cellcolor{my_gray}\first{51.5} \\
                \myTopRule
                \multirow{8}{*}{$512\!\times\!1408$}
                & \threeDPPE\cite{shu2023dppe} & R101 & ICCV23 & \mathDash & \mathDash & \mathDash & $39.1$ & $45.8$ \\
                & STS \cite{wang2022sts} & R101 & AAAI23 & \mathDash & \mathDash & \mathDash & $43.1$ & $52.5$ \\
                & P2D \cite{kim2023predict} & R101 & ICCV23 & \mathDash & \mathDash & \mathDash & $43.3$ & $52.8$ \\
                &\bevDepth\cite{li2023bevdepth}            & R101 & AAAI23 & \mathDash & \mathDash & \mathDash   & $41.8$  & $53.8$   \\
                &\bevDetFourD\cite{huang2022bevdet4d}                & R101 & \arxiv & \mathDash & \mathDash & \mathDash & $42.1$   & $54.5$     \\
                &\beVerseSmall\released\cite{zhang2022beverse}& Swin-S & \arxiv & $20.9$           & $56.2$           & $42.2$        & $35.2$        & $49.5$ \\ 
                & \cellcolor{my_gray}\textbf{\beVerseSmallWithMethod}                          & \cellcolor{my_gray}Swin-S & \cellcolor{my_gray}CVPR24 & \cellcolor{my_gray}$24.6$    & \cellcolor{my_gray}$58.7$    & \cellcolor{my_gray}$45.0$  & \cellcolor{my_gray}$38.2$ & \cellcolor{my_gray}$51.3$ \\
                & \hop\reimplemented\cite{zong2023hop}             & R101 & ICCV23 & 
                \second{31.4}    & \second{63.7}    & \second{52.5}    & \second{45.2}    & \first{55.0} \\
                & \cellcolor{my_gray}\textbf{\hopWithMethod}                          & \cellcolor{my_gray}R101 & \cellcolor{my_gray}CVPR24 & \cellcolor{my_gray}\first{32.9}    & \cellcolor{my_gray}\first{65.0}    & \cellcolor{my_gray}\first{53.1} & \cellcolor{my_gray}\first{46.2} & \cellcolor{my_gray}\second{54.7} \\
                \myTopRule
                \multirow{6}{*}{$640\!\times\!1600$}
                & BEVDet \cite{huang2021bevdet} & \vovNet & \arxiv & $29.6$ & $61.7$ & $48.2$ & $42.1$ & $48.2$ \\
                & \petrVTwo\cite{liu2023petrv2} & R101 & ICCV23 & \mathDash    & \mathDash    & \mathDash    & $42.1$    & $52.4$    \\
                & CAPE\released\cite{xiong2023cape} & \vovNet & CVPR23 & $31.2$    & $63.2$    & $51.9$    & $44.7$    & $54.4$\\
                & \bevDetFourD\cbgs\cite{huang2022bevdet4d} & Swin-B & \arxiv & \mathDash    & \mathDash    & \mathDash & $42.6$ & $55.2$\\ 
                & \hop\reimplemented\cite{zong2023hop}  & \vovNet & ICCV23 & \second{36.5}    & \second{69.1}    & \second{56.1}    & \second{49.6}    & \second{58.3} \\
                & \cellcolor{my_gray}\textbf{\hopWithMethod}               & \cellcolor{my_gray}\vovNet & \cellcolor{my_gray}CVPR24 & \cellcolor{my_gray}\first{40.3}    & \cellcolor{my_gray}\first{71.7}    & \cellcolor{my_gray}\first{58.8} & \cellcolor{my_gray}\first{52.7} & \cellcolor{my_gray}\first{60.2} \\
                \myTopRule
                \multirow{6}{*}{$900\!\times\!1600$} 
                & \fcosThreeD\!\cite{wang2021fcos3d} & R101 & ICCVW21 & \mathDash & \mathDash & \mathDash & $34.4$ & $41.5$ \\
                & \pgd\cite{wang2021probabilistic} & R101 & CoRL21 & \mathDash & \mathDash & \mathDash & $36.9$ & $42.8$ \\
                & \detrThreeD\cite{wang2021detr3d}& R101 & CoRL21 & $22.4$ & $60.3$ & $41.1$ & $34.9$ & $43.4$ \\
                & \petr \cite{liu2022petr} & R101 & ECCV22 & \mathDash & \mathDash & \mathDash & $37.0$ & $44.2$ \\
                & \bevFormer\cite{li2022bevformer} & R101 & ECCV22 & $27.7$ & $48.5$ & $34.5$ & $41.5$ & $51.7$ \\
                & \polarFormer\cite{jiang2023polarformer} & \vovNet & AAAI23 & \mathDash & \mathDash & \mathDash & $50.0$ & $56.2$ \\
            \end{tabular}
            }
        \end{table}

    \subsection{Qualitative Results}

\begin{figure}[!t]
            \centering
            \begin{subfigure}{\figureScaleFraction\linewidth}
                \includegraphics[width=\linewidth]{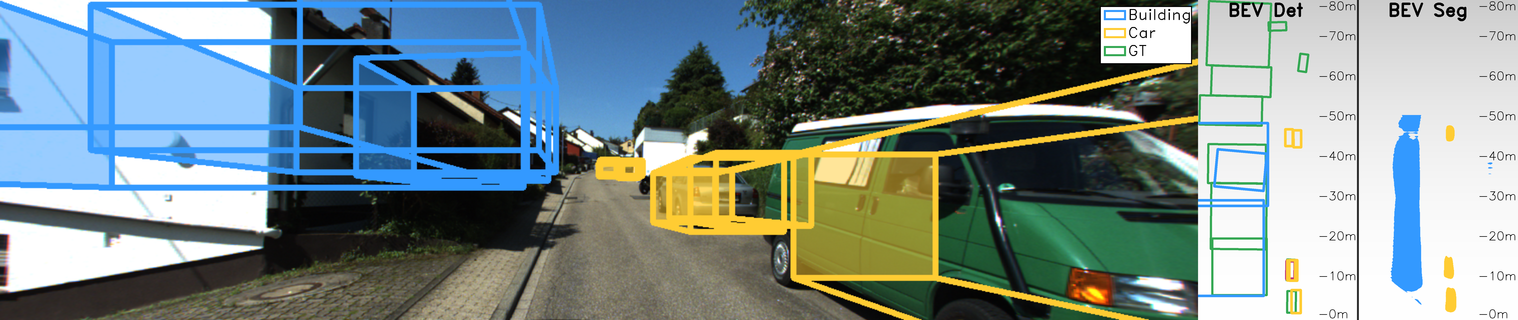}
            \end{subfigure}
            \begin{subfigure}{\figureScaleFraction\linewidth}
                \includegraphics[width=\linewidth]{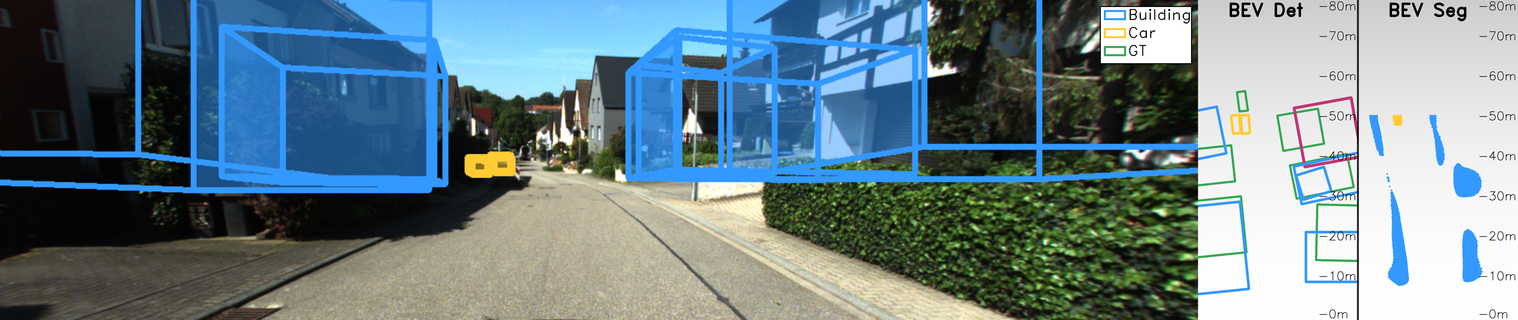}
            \end{subfigure}
            \begin{subfigure}{\figureScaleFraction\linewidth}
                \includegraphics[width=\linewidth]{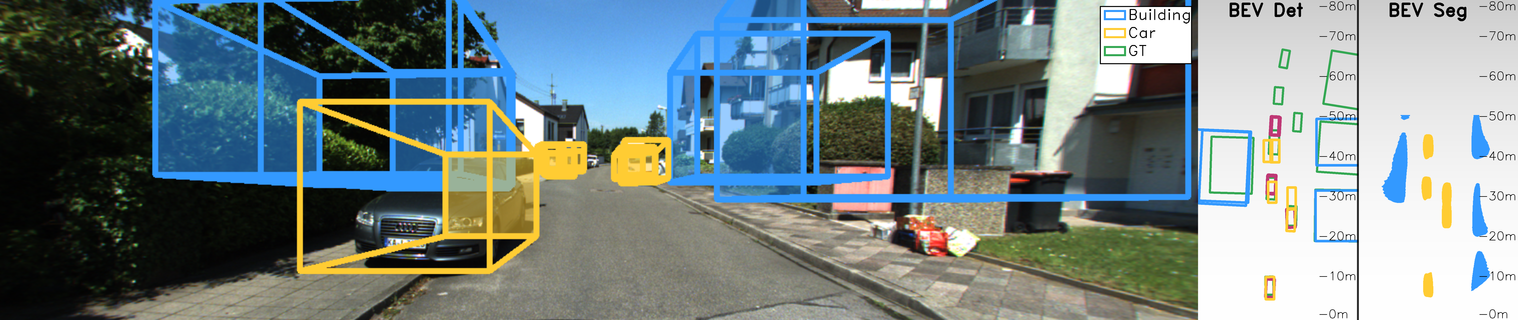}
            \end{subfigure}
            \begin{subfigure}{\figureScaleFraction\linewidth}
                \includegraphics[width=\linewidth]{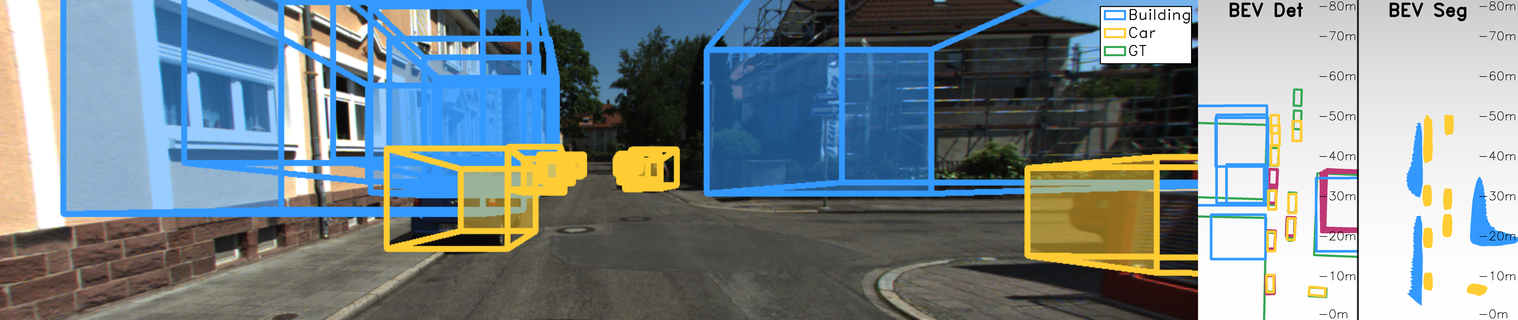}
            \end{subfigure}
            \begin{subfigure}{\figureScaleFraction\linewidth}
                \includegraphics[width=\linewidth]{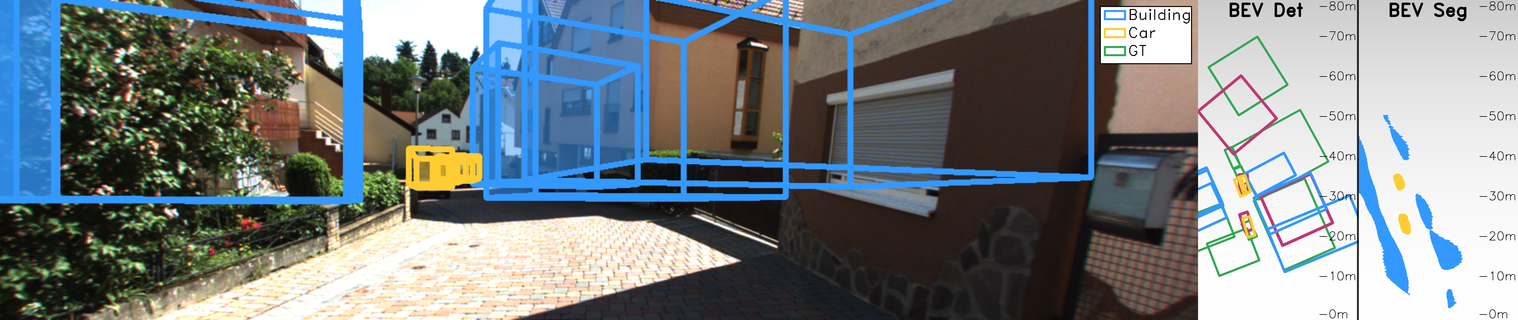}
            \end{subfigure}
            \begin{subfigure}{\figureScaleFraction\linewidth}
                \includegraphics[width=\linewidth]{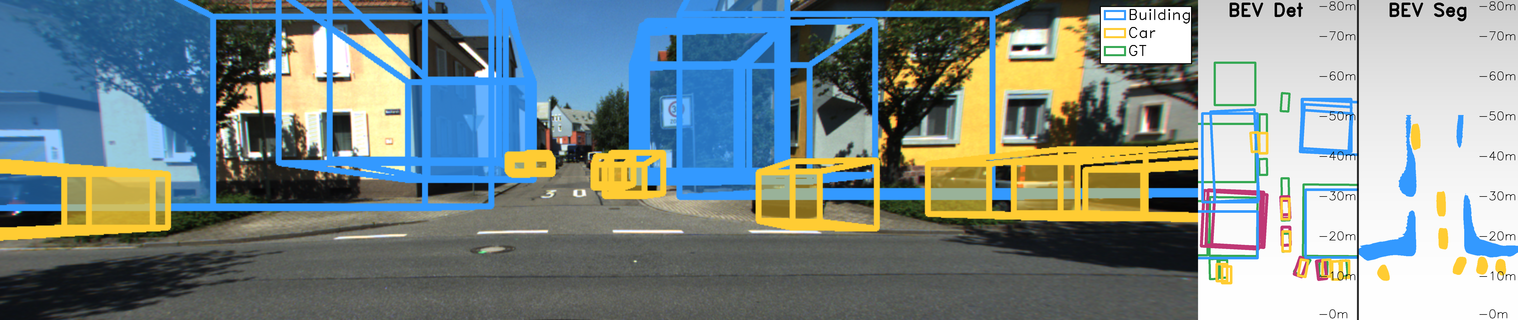}
            \end{subfigure}
            \caption[\kittiThreeSixty Qualitative Results.]{\textbf{\kittiThreeSixty Qualitative Results}. 
            \panopticBEVWithMethod detects more {large objects} (buildings, in blue) than {\monodetr} \cite{zhang2023monodetr} in orange.
            We depict the predictions of \panopticBEVWithMethod in the image view on the left, the predictions of \panopticBEVWithMethod, the baseline \monodetr \cite{zhang2023monodetr}, and ground truth in \bev in the middle, and \bev semantic segmentation predictions from \panopticBEVWithMethod on the right. 
            [Key: {Buildings} (in blue) and {Cars} (in yellow) of \panopticBEVWithMethod; {all classes} (pink) of \monodetr \cite{zhang2023monodetr}, and {Ground Truth} (in green) in BEV]. 
            }
            \label{fig:qualitative_kitti_360}
        \end{figure}

        \noIndentHeading{\kittiThreeSixty.}
            We now show some qualitative results of models trained on \kittiThreeSixty \val split in \cref{fig:qualitative_kitti_360}. 
            We depict the predictions of \panopticBEVWithMethod in image view on the left, the predictions of \panopticBEVWithMethod, the baseline \monodetr \cite{zhang2023monodetr}, predicted and GT boxes in \bev in the middle and \bev semantic segmentation predictions from \panopticBEVWithMethod on the right. 
            In general, \panopticBEVWithMethod detects more larger objects (buildings) than \gupNet\cite{lu2021geometry}.

\begin{figure}[!t]
            \centering
            \begin{subfigure}{\figureScaleFraction\linewidth}
                \includegraphics[width=\linewidth]{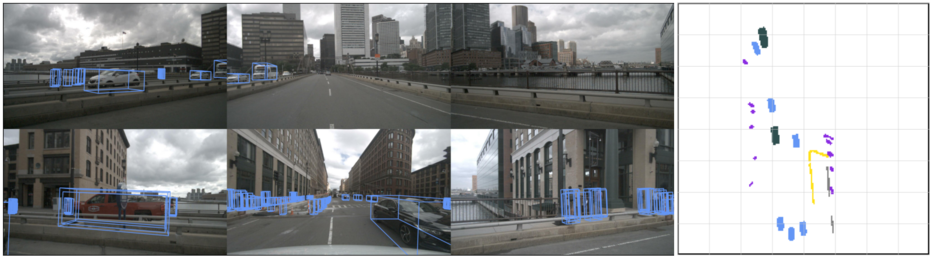}
            \end{subfigure}
            \begin{subfigure}{\figureScaleFraction\linewidth}
                \includegraphics[width=\linewidth]{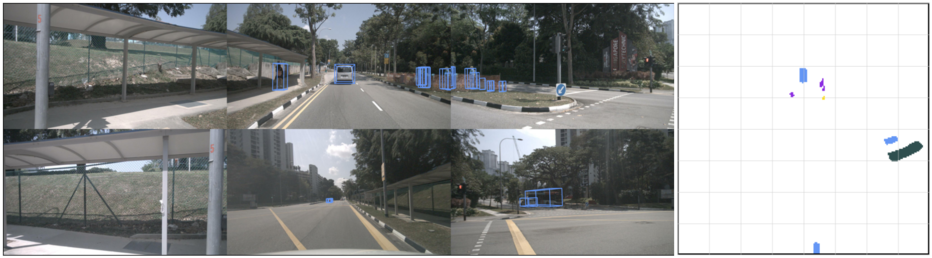}
            \end{subfigure}
            \begin{subfigure}{\figureScaleFraction\linewidth}
                \includegraphics[width=\linewidth]{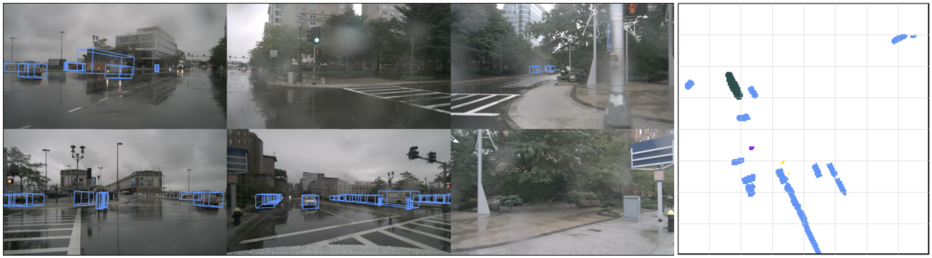}
            \end{subfigure}
            \begin{subfigure}{\figureScaleFraction\linewidth}
                \includegraphics[width=\linewidth]{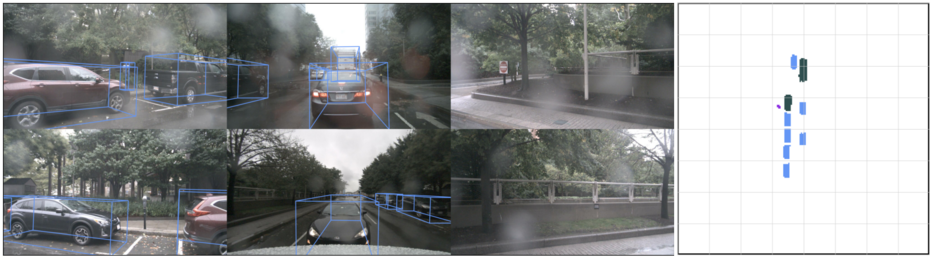}
            \end{subfigure}
            \begin{subfigure}{\figureScaleFraction\linewidth}
                \includegraphics[width=\linewidth]{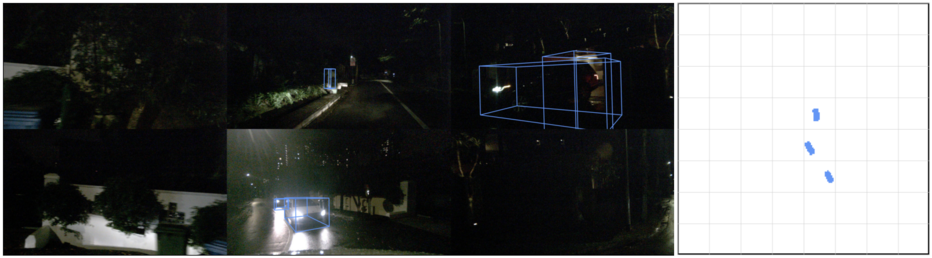}
            \end{subfigure}
            \caption[\nuscenes Qualitative Results.]{\textbf{\nuscenes Qualitative Results}. 
            The first row shows the front\_left, front, and front\_right cameras, while the second row shows the back\_left, back, and back\_right cameras. 
            [Key: {Cars} (blue), {Vehicles} (green), {Pedestrian} (violet), {Cones} (yellow) and {Barrier} (gray) of \beVerseSmallWithMethod at $200\!\times\!200$ resolution in \bev].}
            \label{fig:qualitative_nuscenes}
        \end{figure}

        \noIndentHeading{\nuscenes.}
            We now show some qualitative results of models trained on \nuscenes \val split in \cref{fig:qualitative_nuscenes}.
            As before, we depict the predictions of \beVerseSmallWithMethod in image view from six cameras on the left and \bev semantic segmentation predictions from \seabird on the right.

        \noIndentHeading{\kittiThreeSixty Demo Video.} 
            We next put a short demo video of \panopticBEVWithMethod model
            trained on \kittiThreeSixty \val split compared with \monodetr at\\ \url{https://www.youtube.com/watch?v=SmuRbMbsnZA}.
            We run our trained model independently on each frame of \kittiThreeSixty. 
            None of the frames from the raw video appear in the training set of \kittiThreeSixty \val split. 
            We use the camera matrices available with the video but do not use
            any temporal information. 
            Overlaid on each frame of the raw input videos, we
            plot the projected \threeD boxes of the predictions, predicted and GT boxes in \bev in the middle and \bev semantic segmentation predictions from \panopticBEVWithMethod. 
            We set the frame rate of this demo at $5$ fps similar to \cite{kumar2022deviant}. 
            The demo video demonstrates impressive results on larger objects.

%% file: appendices/charmer_appendix.tex
\chapter{\charmer Appendix}\label{chpt:charmer_appendix}

\section{Additional Details and Proof}\label{sec:additional_proof}
     We now add more details and proofs which we could not put in the main paper because of the space constraints.

    \subsection{Proof of Ground Depth \cref{lemma:1}}\label{sec:charmer_proof_lemma}
        We reproduce the proof from \cite{yang2023gedepth} with our notations for the sake of completeness of this work.
        \begin{proof}
            We first rewrite the pinhole projection \cref{eq.pinhole} as: 
            \begin{align}
            \begin{bmatrix}
            \varX \\
            \varY \\
            \varZ \\
            \end{bmatrix}
            =\rotation^{-1}
            (\intrinsic^{-1}
            \begin{bmatrix}
            \pixU\\
            \pixV\\
            1 \\
            \end{bmatrix}
            \posZ - \extrinsicTrans). 
            \label{eq.rewrite}
            \end{align}
            We now represent the ray shooting from the camera optical center through each pixel as $\ray(\pixU, \pixV, \posZ)$.
            Using the matrix $\bm{A}\!=\!(a_{ij})\!=\!\rotation^{-1}\intrinsic^{-1}$, and the vector $\bm{B}\!=\!(b_i)\!=\!-\rotation^{-1}\extrinsicTrans$, we define the parametric ray as: 
            \begin{align}
            \ray(\pixU, \pixV, \posZ) :
            \begin{cases}
            \varX = (a_{11}u +a_{12}v + a_{13}) \posZ +  b_1  \\
            \varY = (a_{21}u +a_{22}v + a_{23}) \posZ +  b_2  \\
            \varZ = (a_{31}u +a_{32}v + a_{33}) \posZ +  b_3  \\
            \end{cases}
            \label{eq.ray}
            \end{align}
            Moreover, the ground at a distance $h$ can be described by a plane, which is determined by the point $(0, \egoHeight, 0)$ in the plane and the normal vector $\overrightarrow{n}=(0,1,0)$:
            \begin{align}
            \ray \boldsymbol{\cdot} \overrightarrow{n} = \egoHeight.
            \label{eq.plane}
            \end{align}
            Then, the ground depth is the intersection point between this ray and the ground plane. 
            Combining \cref{eq.ray,eq.plane}, the ground depth $\posZ$ of the pixel $(\pixU, \pixV)$ is:
            \begin{align}
                (a_{21}u +a_{22}v + a_{23}) \posZ +  b_2  &= \egoHeight \nonumber \\
                \implies \posZ &= \frac{\egoHeight - b_{2}}{a_{21}u + a_{22}v + a_{23}}.
            \end{align}
        \end{proof}

    \subsection{Proof of \cref{lemma:charmer_2}}\label{sec:charmer_proof_simple}
        We next derive \cref{lemma:charmer_2} from \cref{lemma:charmer_1} as follows. 
        \begin{proof}
        {
            \begin{align}
            \bm{A}\!=\!(a_{ij})&=\rotation^{-1}\intrinsic^{-1}\!=\!\bm{I}^{-1} 
            \begin{bmatrix}
            \focal & 0 & \ppointU\\
            0 & \focal & \ppointV\\
            0 & 0 & 1\\
            \end{bmatrix}^{-1} \nonumber\\
            &=\bm{I} \begin{bmatrix}
            \frac{1}{\focal} & 0 & \frac{-\ppointU}{\focal}\\
            0 & \frac{1}{\focal} & \frac{-\ppointV}{\focal}\\
            0 & 0 & 1\\
            \end{bmatrix}, \nonumber
            \end{align}
            }
            \noindent with rotation matrix $\rotation$ is identity $\bm{I}$ for forward cameras. So, $a_{21}\!=\!0,a_{22}\!=\!\frac{1}{\focal},a_{23}\!=\!\frac{-\ppointV}{\focal}$. Substituting $a_{21},a_{22},a_{23}$ in \cref{eq:gd_depth}, we get \cref{eq:gd_depth_simple}.
        \end{proof}

    \subsection{Extension to Camera not parallel to Ground}\label{sec:not_parallel}
        Following Sec. 3.3 of GEDepth \cite{yang2023gedepth}, we use the camera pitch $\delta$, and generalize \cref{eq:gd_depth} to obtain ground depth as 
        \begin{align}
        \posZ &=\!\frac{\egoHeight\!-\!b_{2}\!\cos\delta\!-\!b_{3}\!\sin\delta}{[a_{21}\pixU\!+\!a_{22}\pixV\!+\!a_{23}]\cos\delta\!+\![a_{31}\pixU\!+\!a_{32}\pixV\!+\!a_{33}]\sin\delta} \nonumber \\
        &=\frac{H-b_{2}\cos\!\delta-b_{3}\!\sin\delta}{\frac{\pixV-\ppointV}{\focal}\cos\delta + \sin\delta}
        \label{eq:not_parallel}
        \end{align}.
        Note that if camera pitch $\delta=0$, this reduces to the usual form of \cref{eq:gd_depth} and \cref{eq:gd_depth_simple} respectively.
        Also, Th. 1 has a more general form with the pitch value, and remains valid for majority of the pitch angle ranges.

    \subsection{Extension to Not-flat Roads}\label{sec:not_flat}
        For non-flat roads, we assume that the road is made of multiple flat `pieces` of roads each with its own slope and we predict the slope of each pixel as in GEDepth \cite{yang2023gedepth}.
        To predict slope $\Hat{\delta}$ of each pixel, we first define a set of $N$ discrete slopes: $\{\tau_i,i\!=\!1,...,N\}$. 
        We compute each pixel slope by linearly combining the discrete slopes with the predicted probability distribution $\{\Hat{p_i}\!\in\![0,1],\sum_{i}\!\Hat{p_i}\!=\!1\}$ over $N$ slopes $\Hat{\delta}\!=\!\sum_{i}\Hat{p_i}\tau_i$.
        We train the network to minimize the total loss:
        $L_{\text{total}}\!=\!L_{\text{det}}\!+\!\lambda_{\text{slope}} L_{\text{slope}}(\delta,\hat{\delta})$, 
        where $L_{\text{det}}$ are the detection losses, and $L_{\text{slope}}$ is the slope classification loss.
        We next substitute the predicted slope in \cite{eq:not_parallel}.
        We do not run this experiment since planar ground is reasonable assumption for most driving scenarios within some distance.
    
    \subsection{Unrealistic assumptions of \cref{theorem:2}} 
        We partially agree. 
        These assumptions reflect the observations of \cite{dijk2019neural}. 
        Also, our theory has empirical support; most \monoThreeD works have no theory. 
        So, our theoretical attempt is a step forward!
        

            \begin{figure*}[!t]
                \centering
                \begin{subfigure}{.38\linewidth}
                  \centering
                  \includegraphics[width=\linewidth]{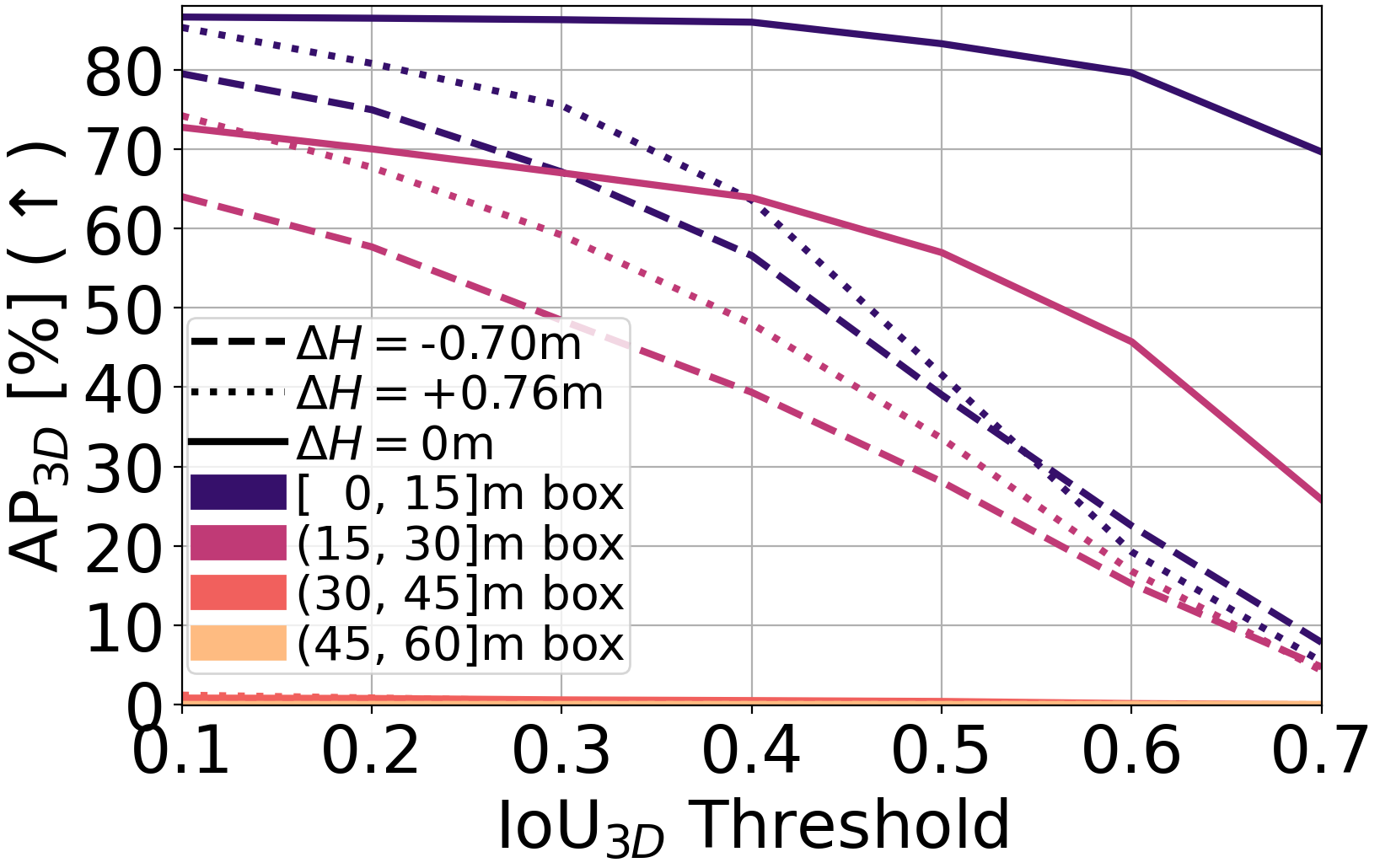}
                  \caption{\gupNet}
                \end{subfigure}~~~~
                \begin{subfigure}{.38\linewidth}
                  \centering
                  \includegraphics[width=\linewidth]{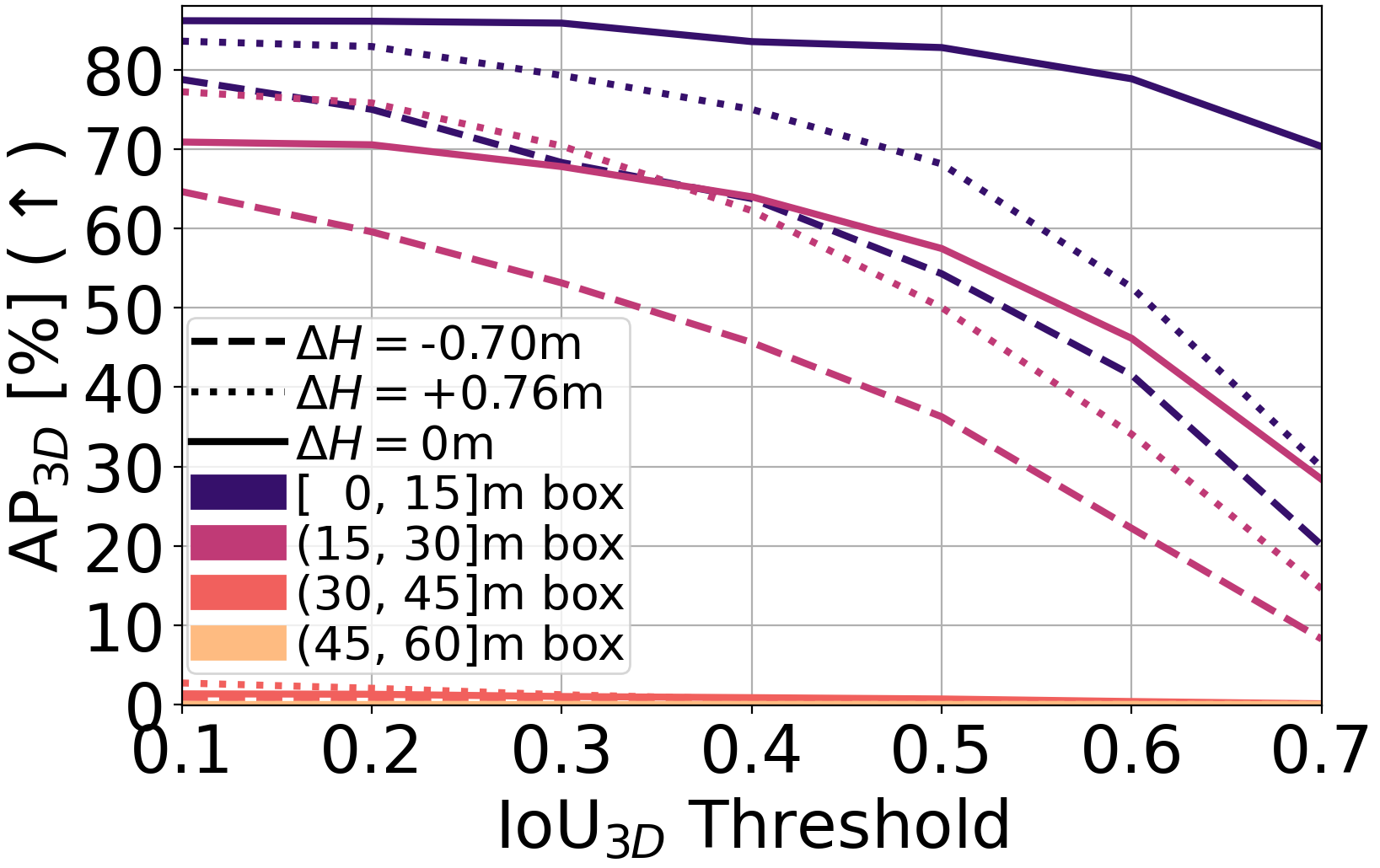}
                  \caption{\charmer}
                \end{subfigure}
                \caption{\textbf{\carla \val \apThreeD{} at different depths and \iouThreeD{} thresholds} with \gupNet. \charmer shows biggest gains on \iouThreeD $> 0.3$ for $[0, 30]m$ boxes. }
                \label{fig:carla_ap_ground_truth_threshold}
            \end{figure*}

            \begin{figure*}[!t]
                \centering
                \begin{subfigure}{.32\linewidth}
                    \includegraphics[width=\linewidth]{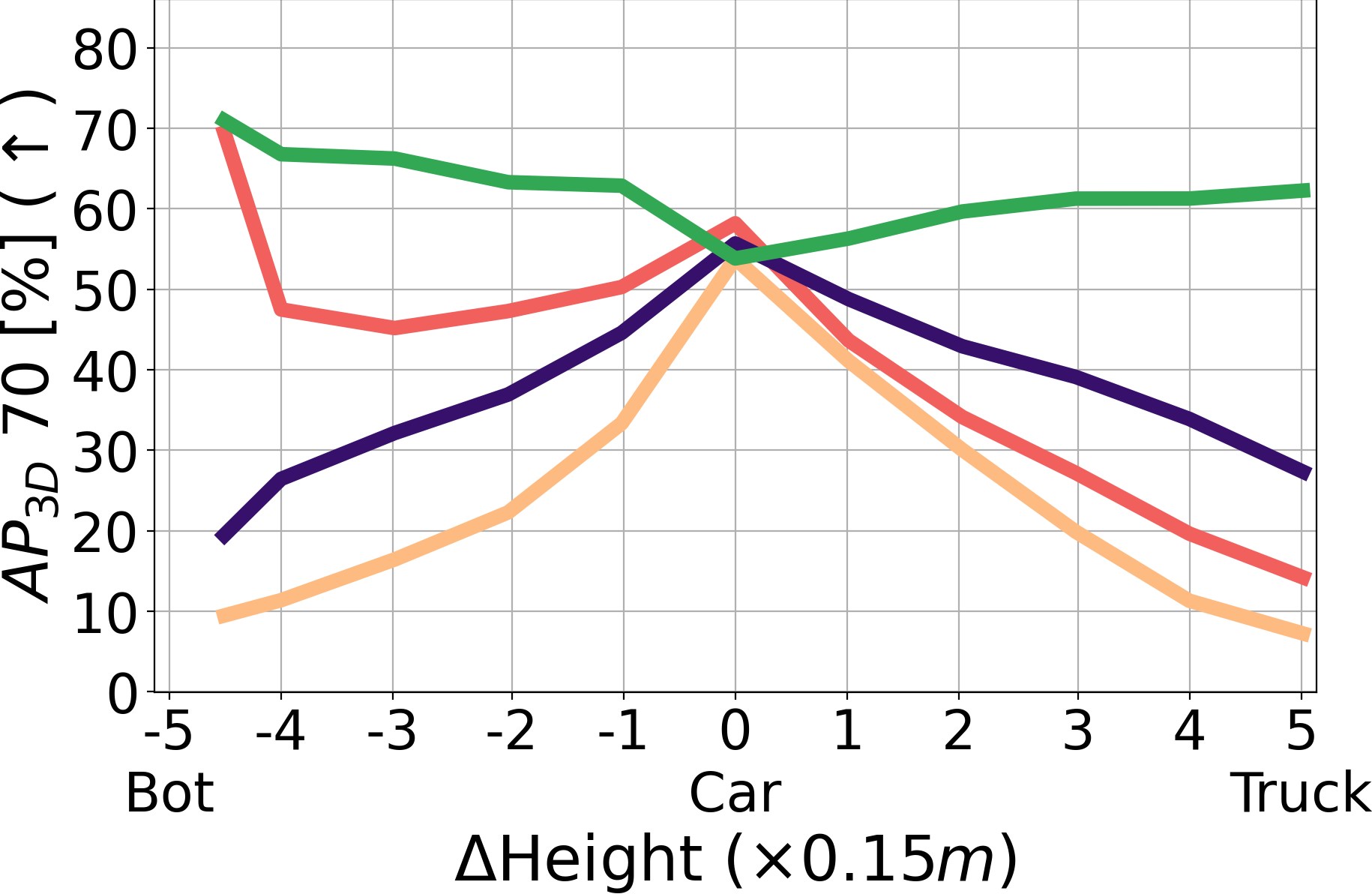}
                    \caption{\apThreeDSeventy \bracketPercentage{} comparison.}
                \end{subfigure}%
                \hfill
                \begin{subfigure}{.32\linewidth}
                    \includegraphics[width=\linewidth]{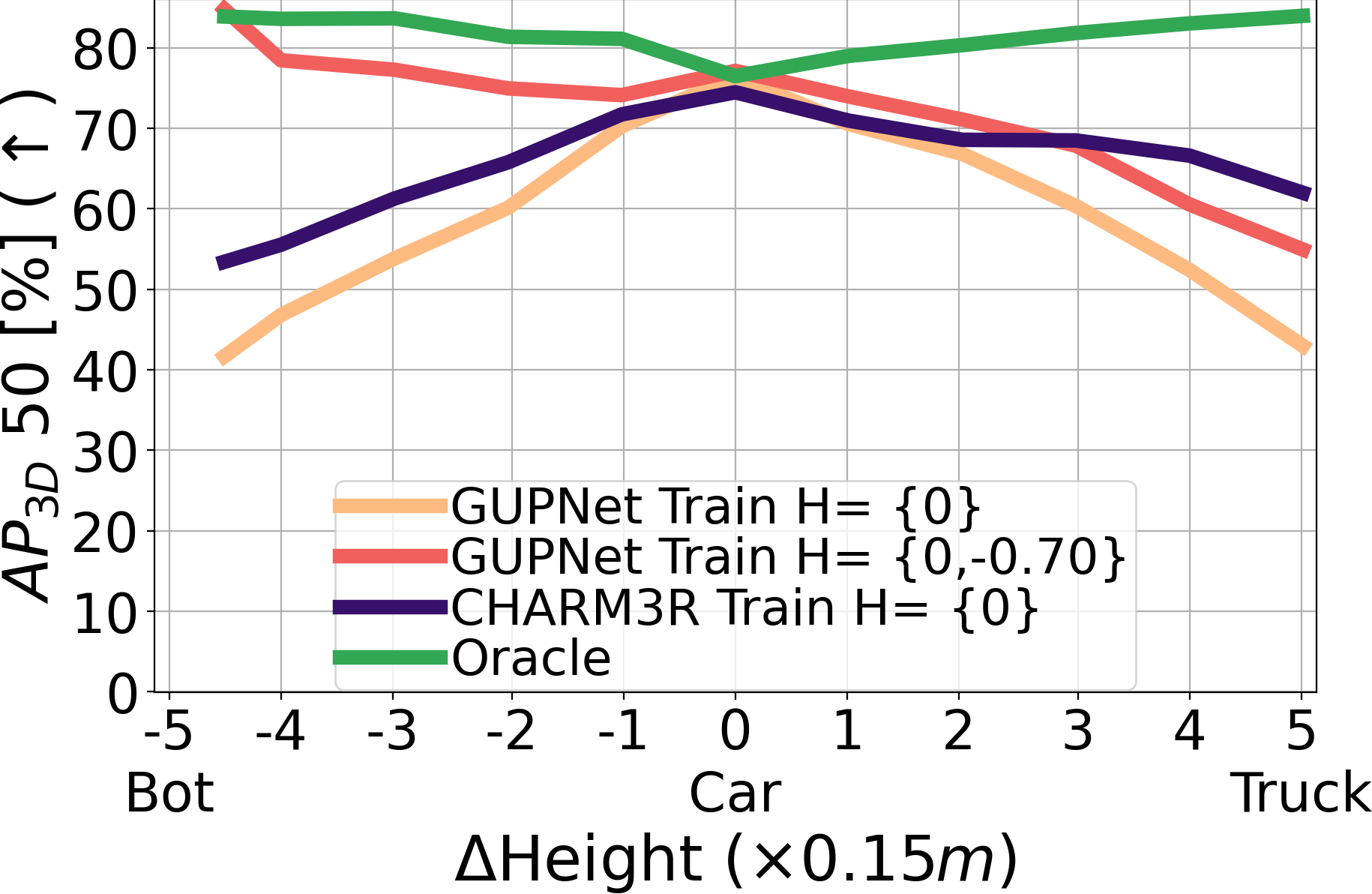}
                    \caption{\apThreeDFifty \bracketPercentage{} comparison.}
                \end{subfigure}
                \hfill
                \begin{subfigure}{.32\linewidth}
                    \includegraphics[width=\linewidth]{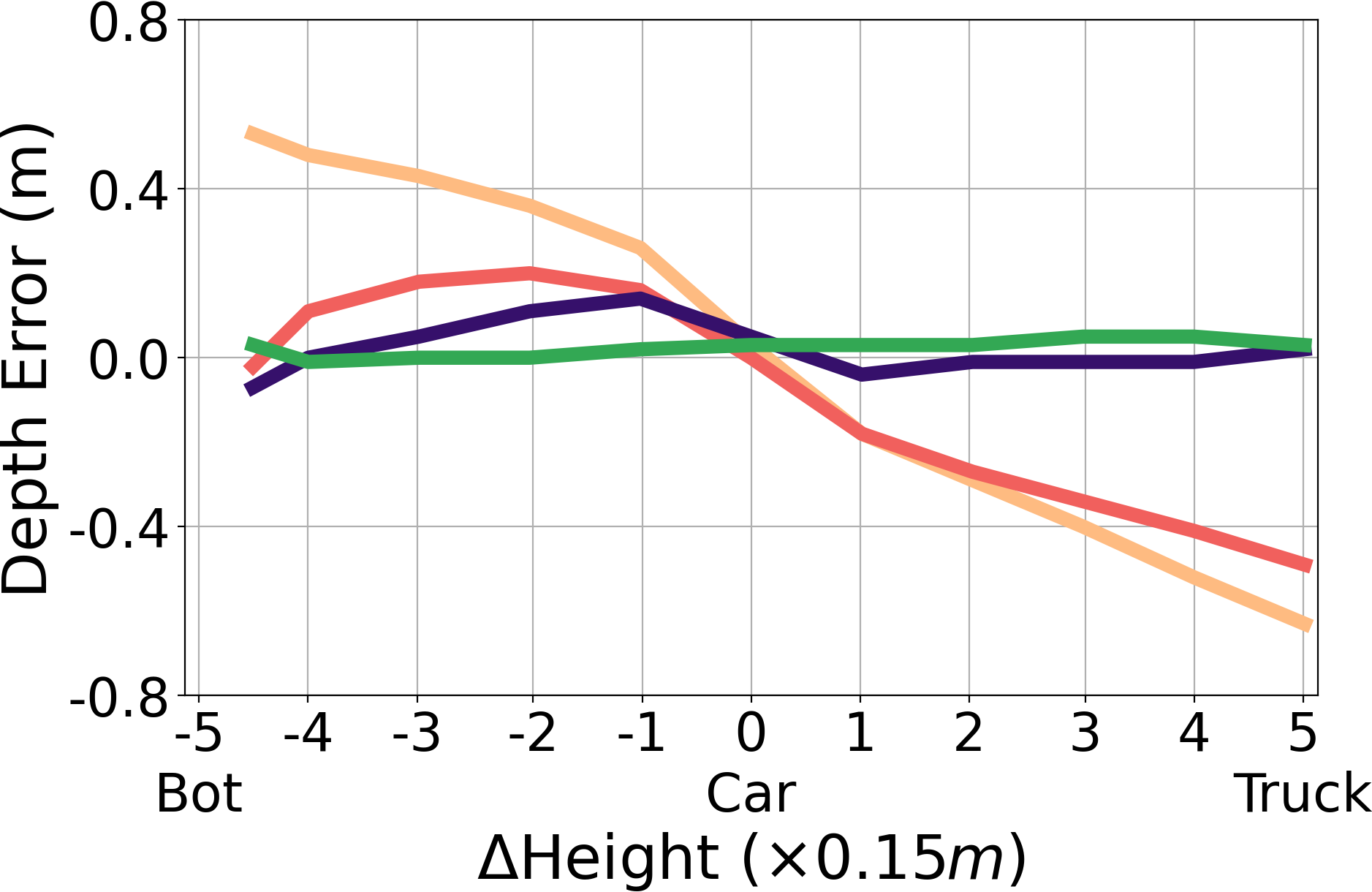}
                    \caption{\MDE comparison.}
                \end{subfigure}
                \caption{\textbf{\carla \val Results with \gupNet} detector after augmentation of \cite{tzofi2023towards}.  
                Training a detector with both $\egoHeightChange=-0.70m$ and $\egoHeightChange=0m$ images produces better results at $\egoHeightChange=-0.70m$ and $\egoHeightChange=0m$, but \textbf{fails at unseen height images $\egoHeightChange=+0.76m$}.
                \charmer \textbf{outperforms} all baselines, especially at unseen bigger height changes.
                All methods except Oracle are trained on car height and tested on all heights.
                }
                \label{fig:det_results_carla_val_aug_more}
            \end{figure*}

\newpage
\section{Additional Experiments }\label{sec:additional_expt}

            \begin{figure*}[!t]
                \centering
                \begin{subfigure}{.32\linewidth}
                    \includegraphics[width=\linewidth]{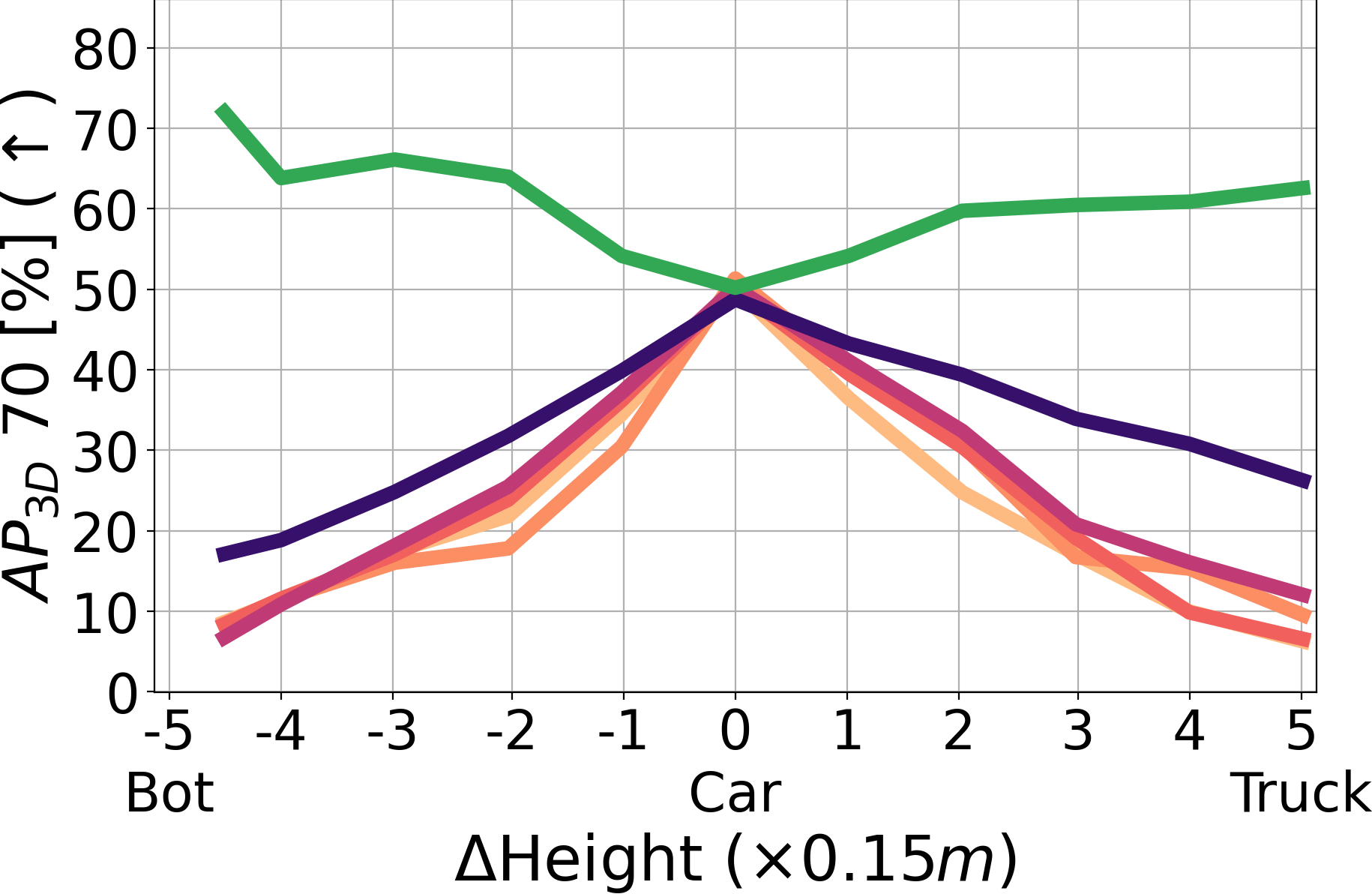}
                    \caption{\apThreeDSeventy \bracketPercentage{} comparison.}
                \end{subfigure}%
                \hfill
                \begin{subfigure}{.32\linewidth}
                    \includegraphics[width=\linewidth]{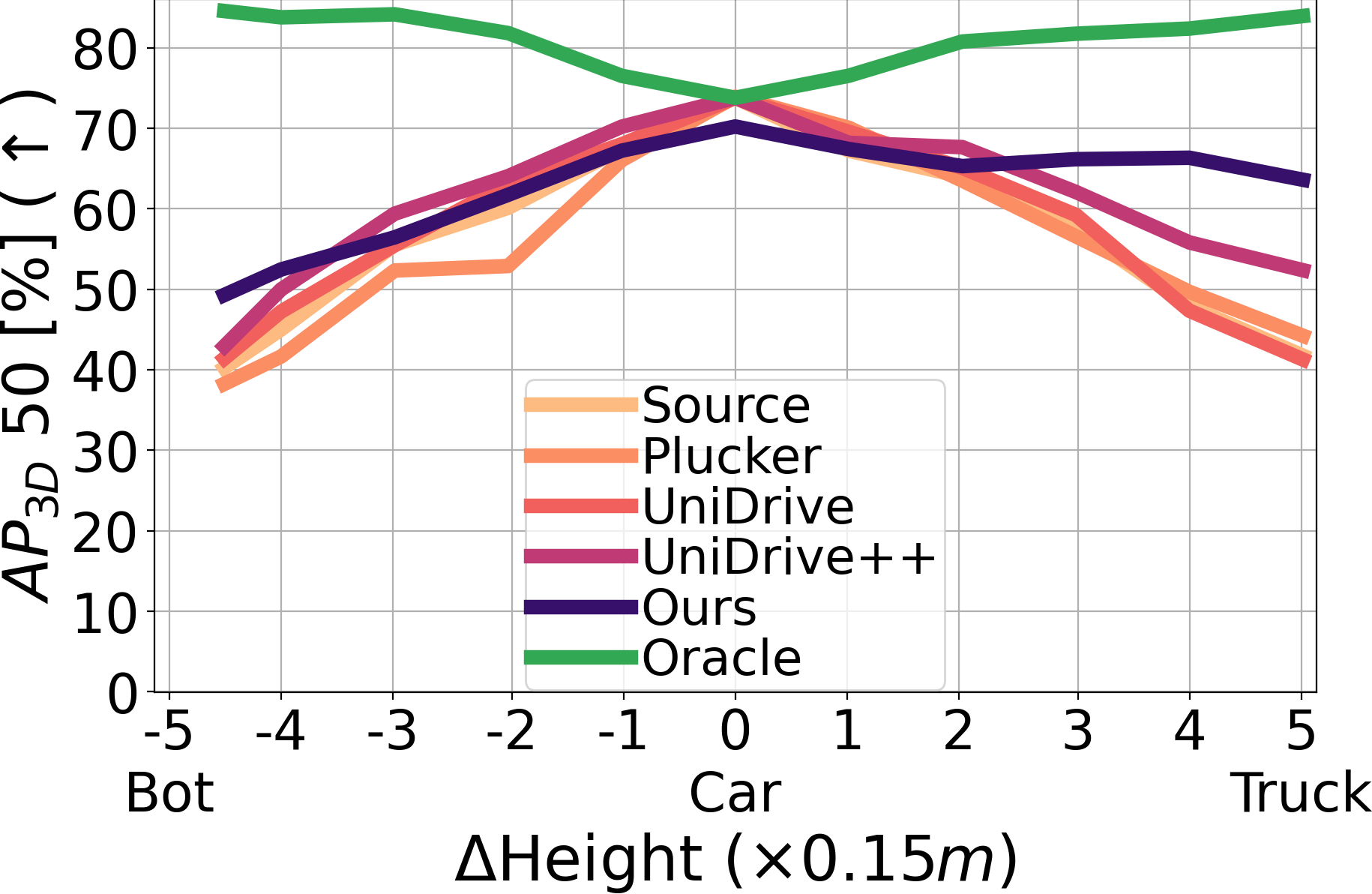}
                    \caption{\apThreeDFifty \bracketPercentage{} comparison.}
                \end{subfigure}
                \hfill
                \begin{subfigure}{.32\linewidth}
                    \includegraphics[width=\linewidth]{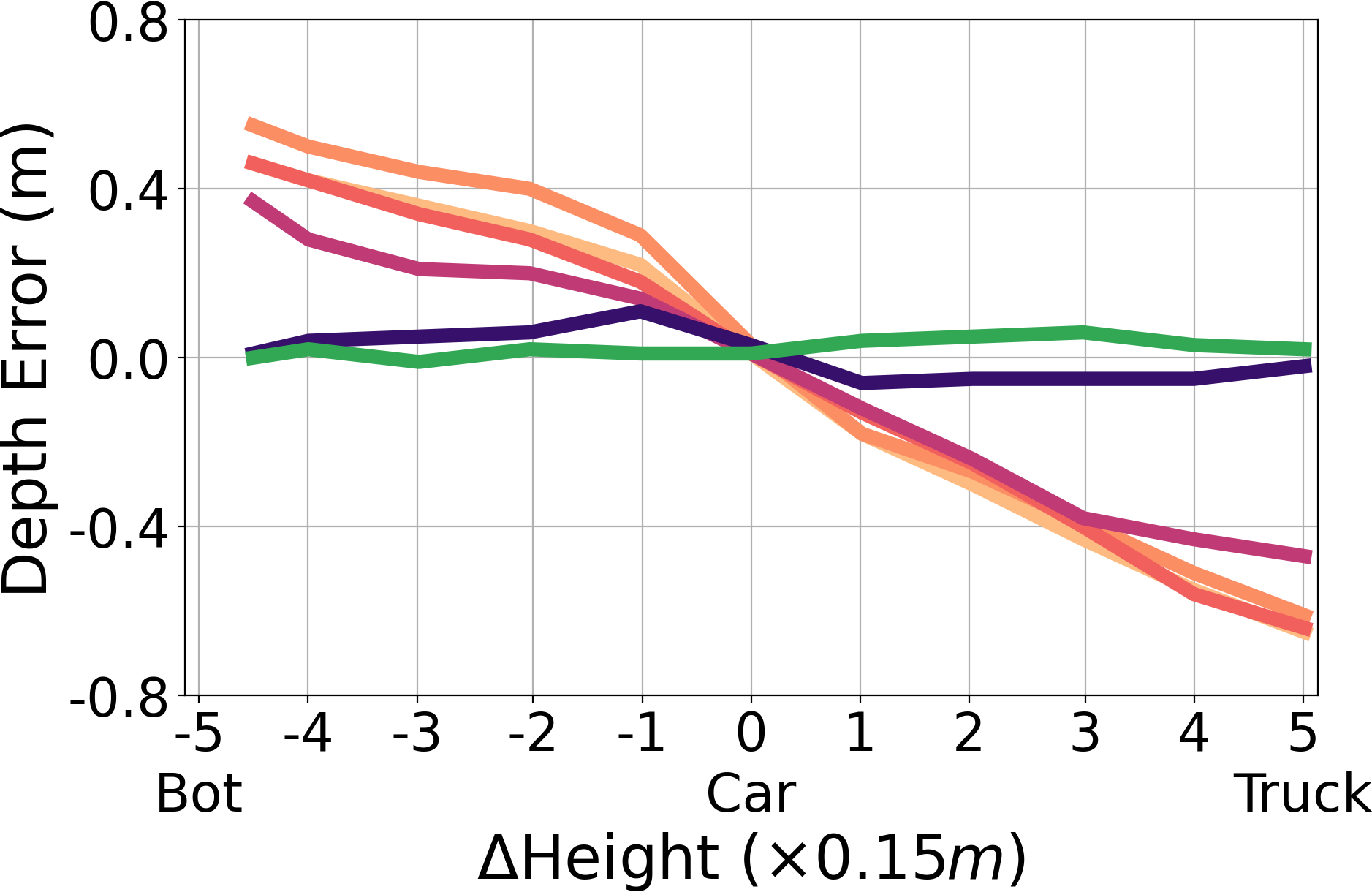}
                    \caption{\MDE comparison.}
                \end{subfigure}
                \caption{\textbf{\carla \val Results with \deviant} detector.  
                \charmer \textbf{outperforms} all baselines, especially at bigger height changes.
                All methods except Oracle are trained on car height and tested on all heights.
                Results of inference on height changes of $-0.70,0$ and $0.76$ meters are in \cref{tab:det_results_carla_val}.
                }
                \label{fig:det_results_carla_val_more}
            \end{figure*}

    We now provide additional details and results of the experiments evaluating \charmer{}’s performance.  

    \subsection{\carla \val Results}
        We first analyze the results on the synthetic \carla dataset further.

        \noIndentHeading{AP at different distances and thresholds.}
            We next compare the \apThreeD of the baseline \gupNet and \charmer in \cref{fig:carla_ap_ground_truth_threshold} at different distances in meters and \iouThreeD matching criteria of $0.1-0.7$ as in \cite{kumar2022deviant}. 
            \cref{fig:carla_ap_ground_truth_threshold} shows that \charmer is effective over \gupNet at all depths and higher \iouThreeD thresholds.
            \charmer shows biggest gains on \iouThreeD $> 0.3$ for $[0, 30]m$ boxes.

        \noIndentHeading{Comparison with Augmentation-Methods.}
            \cref{sec:charmer_intro} of the paper says that the augmentation strategy falls short when the target height is \ood. 
            We show this in \cref{fig:det_results_carla_val_aug_more}. 
            Since authors of \cite{tzofi2023towards} do not release the NVS code, we use the ground truth images from height change $\egoHeightChange=-0.70m$ in training.
            \cref{fig:det_results_carla_val_aug_more} confirms that augmentation also improves the performance on $\egoHeightChange=-0.70m$ and $\egoHeightChange=0m$, but again falls short on unseen ego \variations $\egoHeightChange=+0.76m$.
            On the other hand, \charmer (even though trained on $\egoHeightChange=-0.70m$) outperforms such augmentation strategy at unseen ego heights $\egoHeightChange=+0.76m$. 
            This shows the complementary nature of \charmer over augmentation strategies.
    
        \noIndentHeading{Reproducibility.}
            We ensure reproducibility of our results by repeating our experiments for 3 random seeds. 
            We choose the final epoch as our checkpoint in all our experiments as \cite{kumar2022deviant,kumar2024seabird}. 
            \cref{tab:det_results_carla_val_reproduce} shows the results with these seeds.
            \charmer outperforms the baseline \gupNet in both median and average cases.

        \begin{table*}[!t]
            \caption{\textbf{Reproducibility Results.}
            \charmer \textbf{outperforms} all other baselines on \carla \val split, especially at bigger {unseen ego \variations} in both median (Seed=$444$) and average cases.
            All except Oracle are trained on car height $\egoHeightChange=0m$ and tested on bot to truck height data.
            [Key: \firstKey{Best}]
            }
            \label{tab:det_results_carla_val_reproduce}
            \centering
            \scalebox{0.78}{
            \setlength\tabcolsep{0.23cm}
            \begin{tabular}{l l m dcd m dcd m dcdc}
                \addlinespace[0.01cm]
                \multirow{2}{*}{\threeD Detector} & \multirow{2}{*}{Seed $\downarrowRHDSmall$ / $\egoHeightChange~(m)\rightarrowRHDSmall$} & \multicolumn{3}{cm}{\apThreeDSeventy \bracketPercentage~(\uparrowRHDSmall)} & \multicolumn{3}{cm}{\apThreeDFifty \bracketPercentage~(\uparrowRHDSmall)} & \multicolumn{3}{c}{\MDE $(m)~[\approx 0]$}\\
                &  & $-0.70$ & $0$ & $+0.76$ & $-0.70$ & $0$ & $+0.76$ & $-0.70$ & $0$ & $+0.76$\\
                \myTopRule
                \multirow{4}{*}{\gupNet \cite{lu2021geometry}} 
                & $111$  & $12.24$ & $55.98$ & $7.53$ & $44.14$ & $76.37$ & $41.32$ & $+0.48$ & $+0.00$ & $-0.64$\\
                & $444$  & $9.46$ & $53.82$ & $7.23$ & $41.66$ & $76.47$ & $40.97$ & $+0.53$ & $+0.03$ & $-0.63$\\
                & $222$  & $10.35$ & $52.94$ & $10.79$ & $41.67$ & $75.80$ & $46.45$ & $+0.53$ & $+0.01$ & $-0.57$\\
                \hhline{|~|-----------|}
                & Average  & $10.68$ & $54.25$ & $8.52$ & $42.49$ & \first{76.21} & $43.58$ & $+0.51$ & $+0.01$ & $-0.61$\\
                \myTopRule
                \multirow{4}{*}{+~\charmer} 
                & $111$  & $19.99$ & $58.16$ & $29.96$ & $54.15$ & $74.10$ & $64.27$ & $+0.09$ & $+0.00$ & $-0.03$\\
                & $444$  & $19.45$ & $55.68$ & $27.33$ & $53.40$ & $74.47$ & $61.98$ & $+0.07$ & $+0.05$ & $-0.02$\\
                & $222$  & $17.41$ & $53.57$ & $27.77$ & $54.30$ & $74.83$ & $64.42$ & $+0.12$ & $+0.01$ & $-0.09$\\
                \hhline{|~|-----------|}
                & Average  & \first{18.95} & \first{55.80} & \first{28.35} & \first{53.95} & $74.47$ & \first{63.56} & $+0.09$ & $+0.02$ & $-0.05$\\
                \myTopRule
                Oracle & \mathDash & $70.96$ & $53.82$ & $62.25$ & $83.88$ & $76.47$ & $83.96$ & $+0.03$ & $+0.03$ & $+0.03$\\
            \end{tabular}
            }
        \end{table*}

        \begin{table*}[!t]
            \caption{\textbf{\nuscenes to \coda \val Results.}
            \charmer \textbf{outperforms} all baselines, especially at {unseen height changes}.
            [Key: \firstKey{Best}, \secondKey{Second Best}, Ped= Pedestrians]
            }
            \label{tab:det_results_nusc_to_coda}
            \centering
            \scalebox{\scaleFraction}{
            \setlength\tabcolsep{0.15cm}
            \begin{tabular}{l l m dc m dc }
                \addlinespace[0.00cm]
                \multirow{2}{*}{\threeD Detector} & \multirow{2}{*}{Method} &\multicolumn{2}{cm}{Car \apThreeDFifty~\bracketPercentage~(\uparrowRHDSmall)} & \multicolumn{2}{c}{Ped \apThreeDThirty~\bracketPercentage~(\uparrowRHDSmall)} \\
                & & \coda & \nuscenes & \coda & \nuscenes \\
                \myTopRule
                \multirow{5}{*}{\!\gupNet{} \cite{lu2021geometry}} 
                & Source  & $0.02$ & \first{18.42} & $0.01$ &\first{2.93} \\
                & \uniDrive \cite{li2024unidrive} & $0.02$ & \first{18.42} & $0.01$ &\first{2.93}\\
                & \uniDrivePlus{}\!\cite{li2024unidrive} & \second{0.03} & \first{18.42} & \second{0.02} &\first{2.93}\\
                & \cellcolor{methodColor}\textbf{\charmer} & \cellcolor{methodColor}\first{0.30} & \cellcolor{methodColor}\second{14.80} & \cellcolor{methodColor}\first{0.05} & \cellcolor{methodColor}\second{1.26}\\
                \hhline{|~|-----|}
                & Oracle & $28.56$ & $18.42$ & $30.31$ & $2.93$ \\
            \end{tabular}
            }
        \end{table*}

        \begin{figure*}[!t]
            \centering
            \begin{subfigure}{.38\linewidth}
              \centering
              \includegraphics[width=\linewidth]{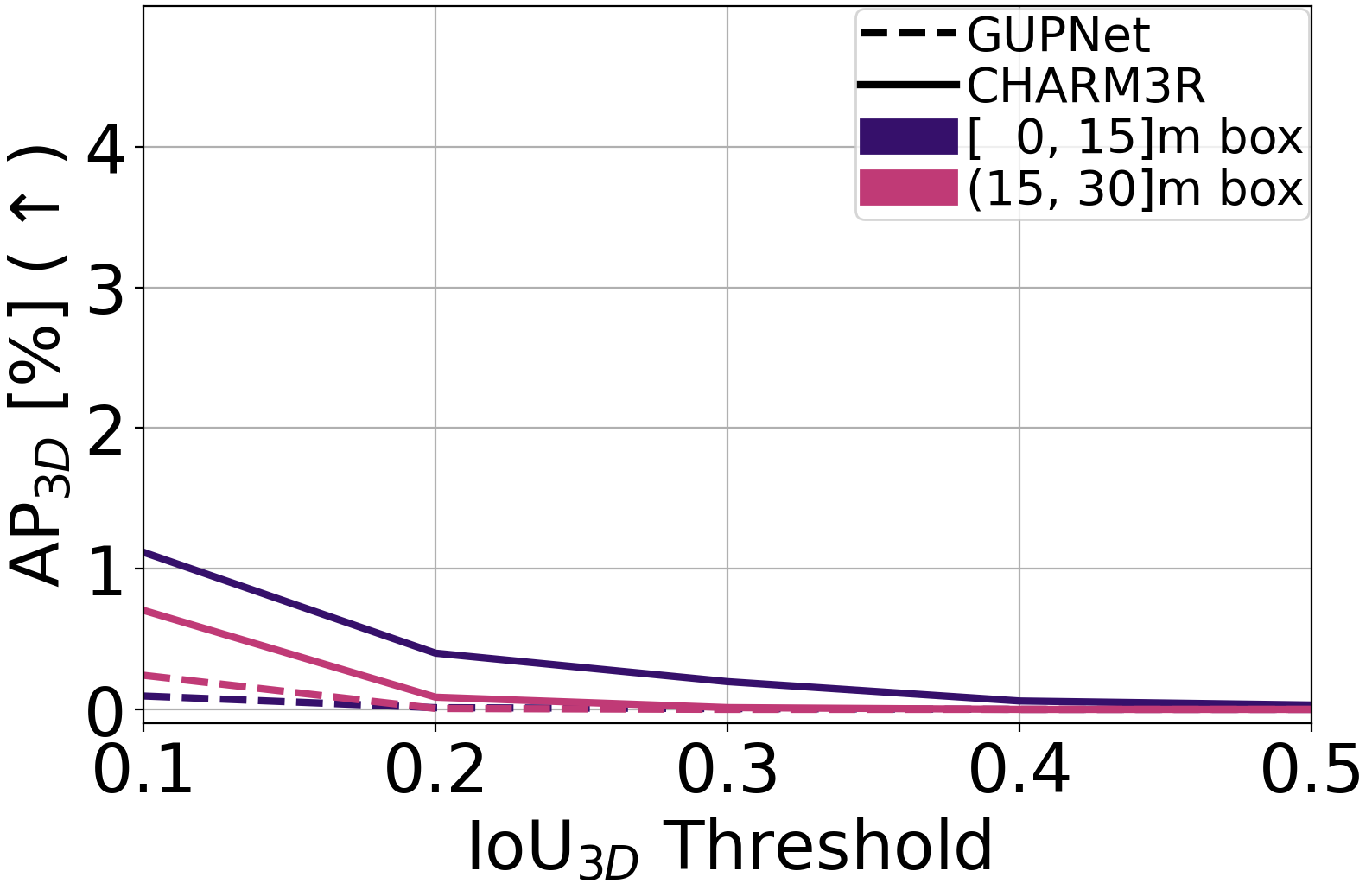}
              \caption{\coda Car}
            \end{subfigure}~~~~
            \begin{subfigure}{.38\linewidth}
              \centering
              \includegraphics[width=\linewidth]{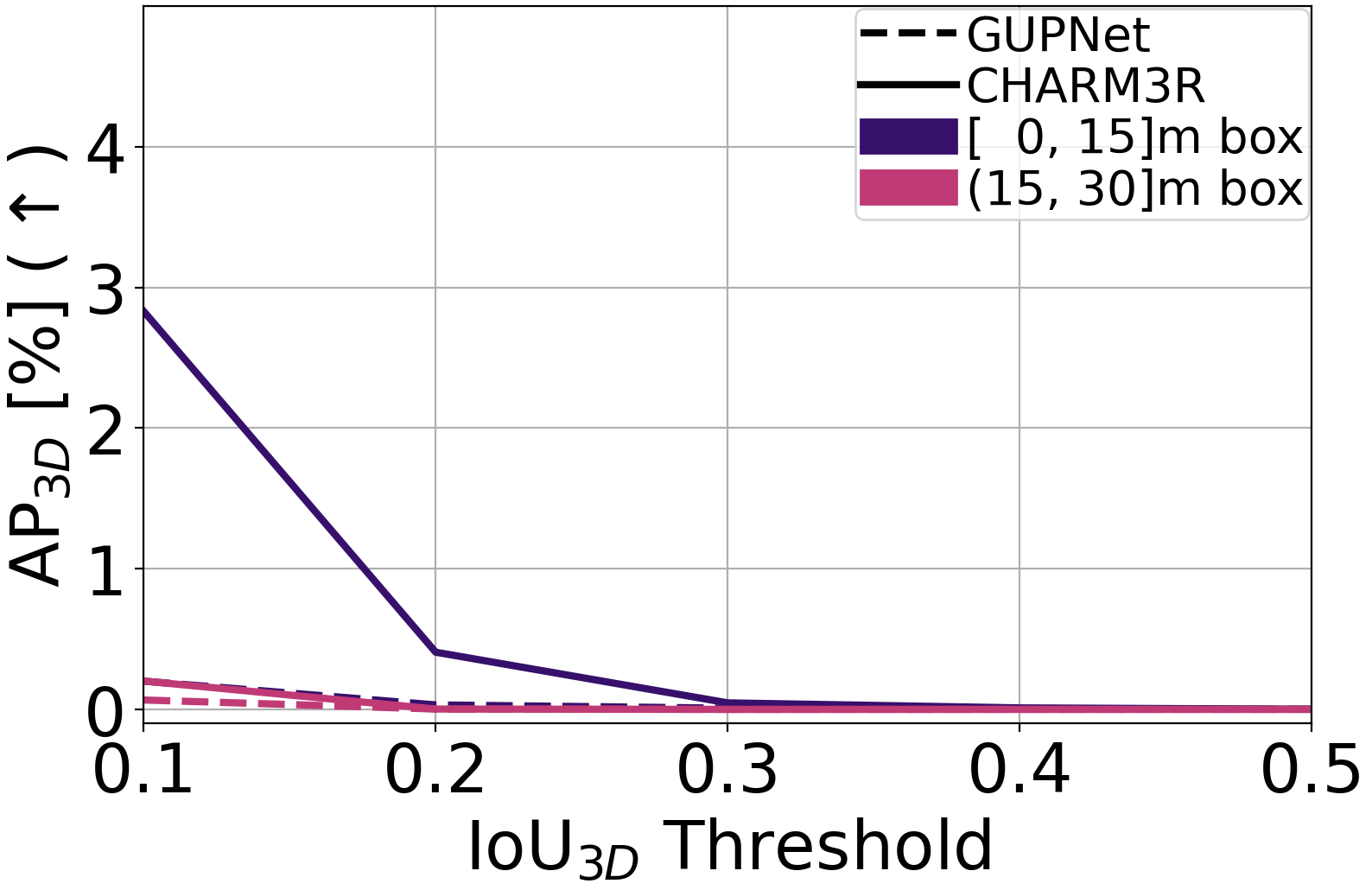}
              \caption{\coda Pedestrian}
            \end{subfigure}
            \caption{\textbf{\coda \val \apThreeD{} at different depths and \iouThreeD{} thresholds} with \gupNet trained on \nuscenes. \charmer shows biggest gains on \iouThreeD $< 0.3$ for $[0, 30]m$ boxes. }
            \label{fig:coda_ap_ground_truth_threshold}
        \end{figure*}

        \noIndentHeading{Results with \deviant.}
            We next additionally plot the robustness of \charmer with other methods on the \deviant detector \cite{kumar2022deviant} in \cref{fig:det_results_carla_val_more}
            The figure confirms that \charmer works even with \deviant and produces \sota robustness to unseen ego \variations.





    \newpage
    \subsection{\nuscenes $\rightarrowRHD$ \coda \val Results}
        To test our claims further in real-life, we use two real datasets: the \nuscenes dataset \cite{caesar2020nuscenes} and the recently released \coda \cite{zhang2023utcoda} datasets. 
        \nuscenes has ego camera at height $1.51m$ above the ground, while the \coda is a robotics dataset with ego camera at a height of $0.75m$ above the ground.
        This experiment uses the following data split:
        \begin{itemize}
            \item \textit{\nuscenes \val Split.}
            This split \cite{caesar2020nuscenes} contains $28{,}130$ training and $6{,}019$ validation images from the front camera as \cite{kumar2022deviant}. 
            \item \textit{\coda \val Split.} 
            This split \cite{zhang2023utcoda} contains $19{,}511$ training and $4{,}176$ validation images. We only use this split for testing.
        \end{itemize}

        We train the \gupNet detector with $10$ \nuscenes classes and report the results with the \kitti metrics on both \nuscenes val and \coda \val splits.
        
        \noIndentHeading{Main Results.}
            We report the main results in \cref{tab:det_results_nusc_to_coda} paper. 
            The results of \cref{tab:det_results_nusc_to_coda} shows gains on both Cars and Pedestrians classes of \coda val dataset. 
            The performance is very low, which we believe is because of the domain gap between \nuscenes and \coda datasets.
            These results further confirm our observations that unlike \twoD detection, generalization across unseen datasets remains a big problem in the \monoThreeD task.

        \noIndentHeading{AP at different distances and thresholds.}
            To further analyze the performance, we next plot the \apThreeD of the baseline \gupNet and \charmer in \cref{fig:coda_ap_ground_truth_threshold} at different distances in meters and \iouThreeD matching criteria of $0.1-0.5$ as in \cite{kumar2022deviant}. 
            \cref{fig:coda_ap_ground_truth_threshold} shows that \charmer is effective over \gupNet at all depths and lower \iouThreeD thresholds.
            \charmer shows biggest gains on \iouThreeD $< 0.3$ for $[0, 30]m$ boxes.
            The gains are more on the Pedestrian class on \coda since \coda captures UT Austin campus scenes, and therefore, has more pedestrians compared to cars. 
            \nuscenes captures outdoor driving scenes in Boston and Singapore, and therefore, has more cars compared to pedestrians.
            We describe the statistics of these two datasets in \cref{tab:dataset_statistics}.

             \begin{table}[!t]
                \caption{\textbf{Dataset statistics.}
                \nuscenes \val has more Cars compared to Pedestrians, while \coda \val has more Pedestrians than Cars.
                }
                \label{tab:dataset_statistics}
                \centering
                \scalebox{\scaleFraction}{
                \setlength\tabcolsep{0.1cm}
                \begin{tabular}{l c m c m ccc }
                    \addlinespace[0.00cm]
                    \textbf{\val} & \textbf{Ego Ht} $(m)$ & \textbf{\#Images} & \textbf{Car} $(k)$ & \textbf{Ped} $(k)$\\
                    \myTopRule
                    \nuscenes & $1.51$ & $6{,}019$ & $18$  & $7$ \\
                    \coda & $0.75$ & $4{,}176$ & $4$ & $86$ \\
                \end{tabular}
                }
            \end{table}

    \subsection{Qualitative Results.}

\begin{figure*}[!t]
            \centering
            \begin{subfigure}{0.4\linewidth}
                \includegraphics[width=\linewidth]{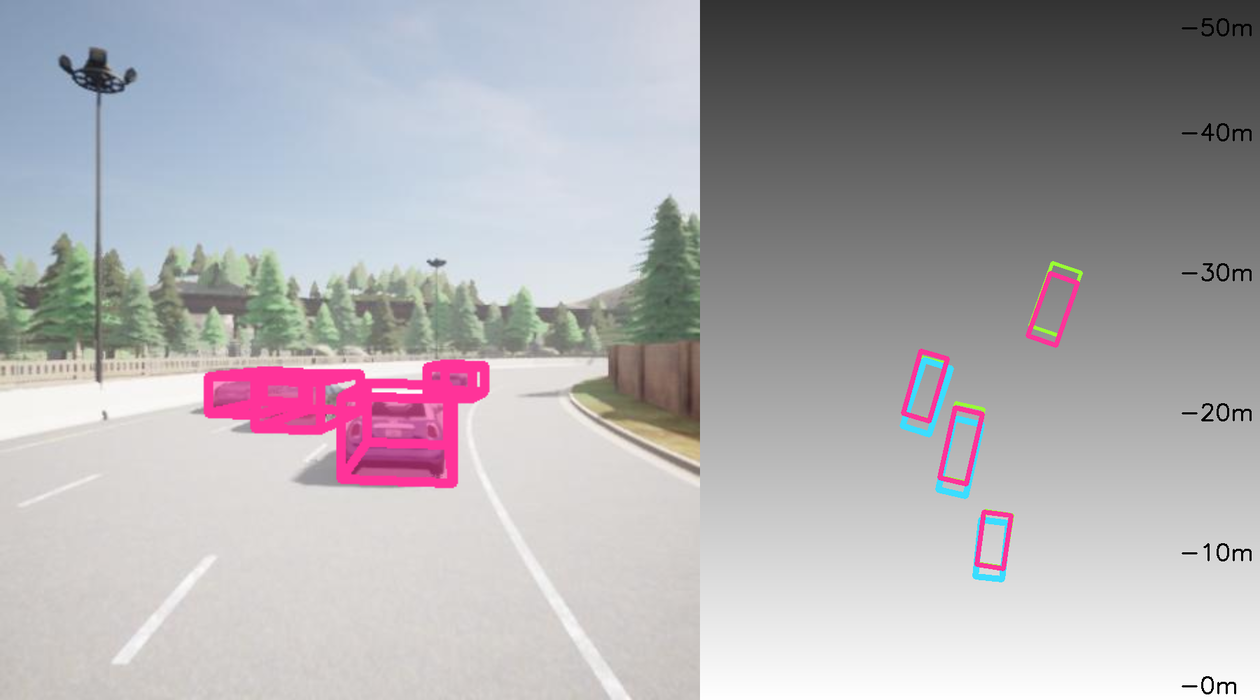}
            \end{subfigure}
            \begin{subfigure}{0.4\linewidth}
                \includegraphics[width=\linewidth]{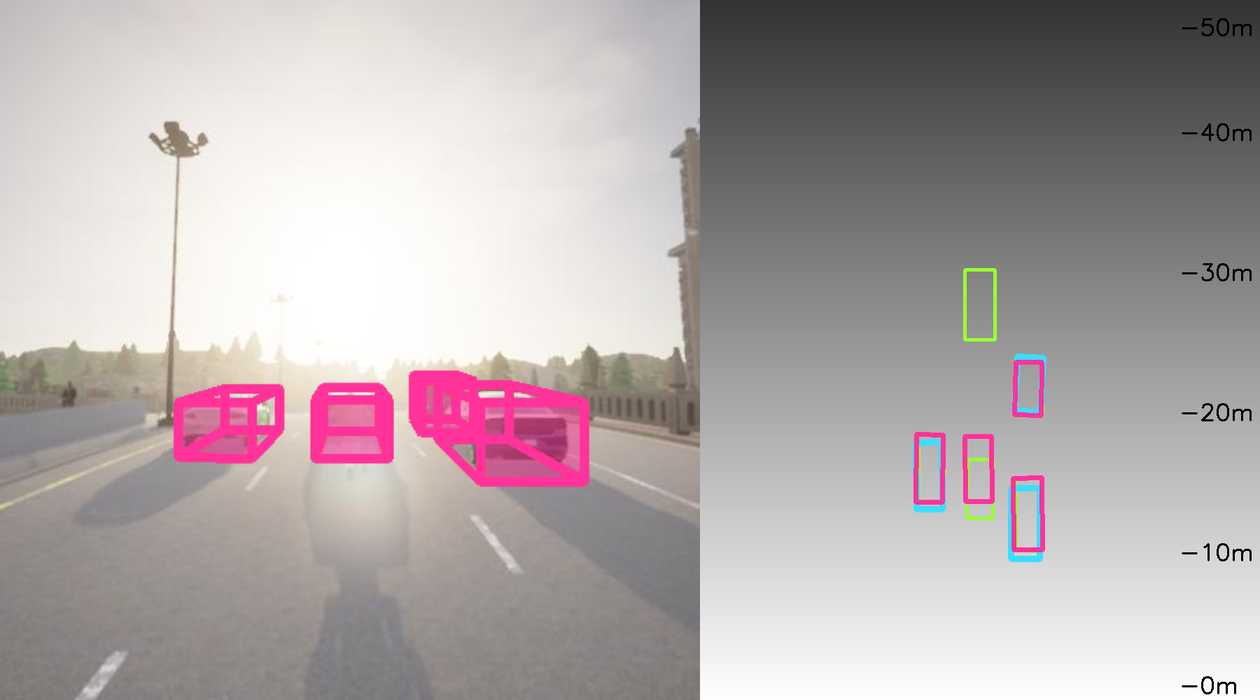}
            \end{subfigure}
            \begin{subfigure}{0.4\linewidth}
                \includegraphics[width=\linewidth]{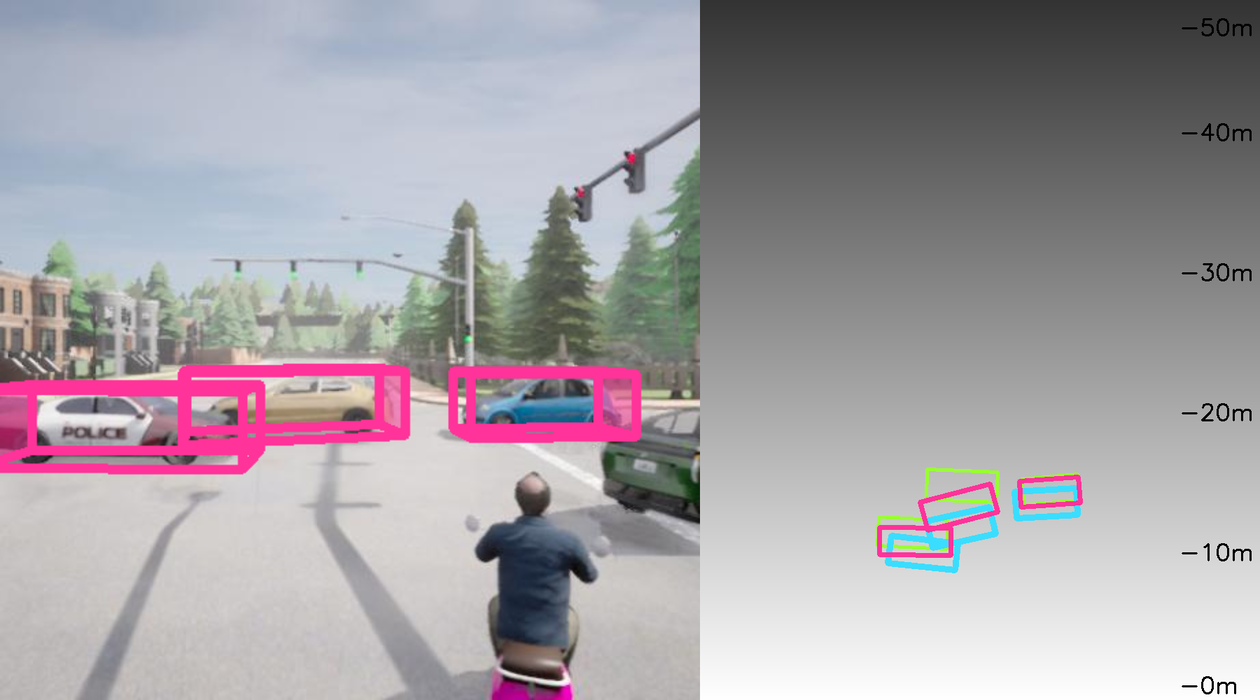}
            \end{subfigure}
            \begin{subfigure}{0.4\linewidth}
                \includegraphics[width=\linewidth]{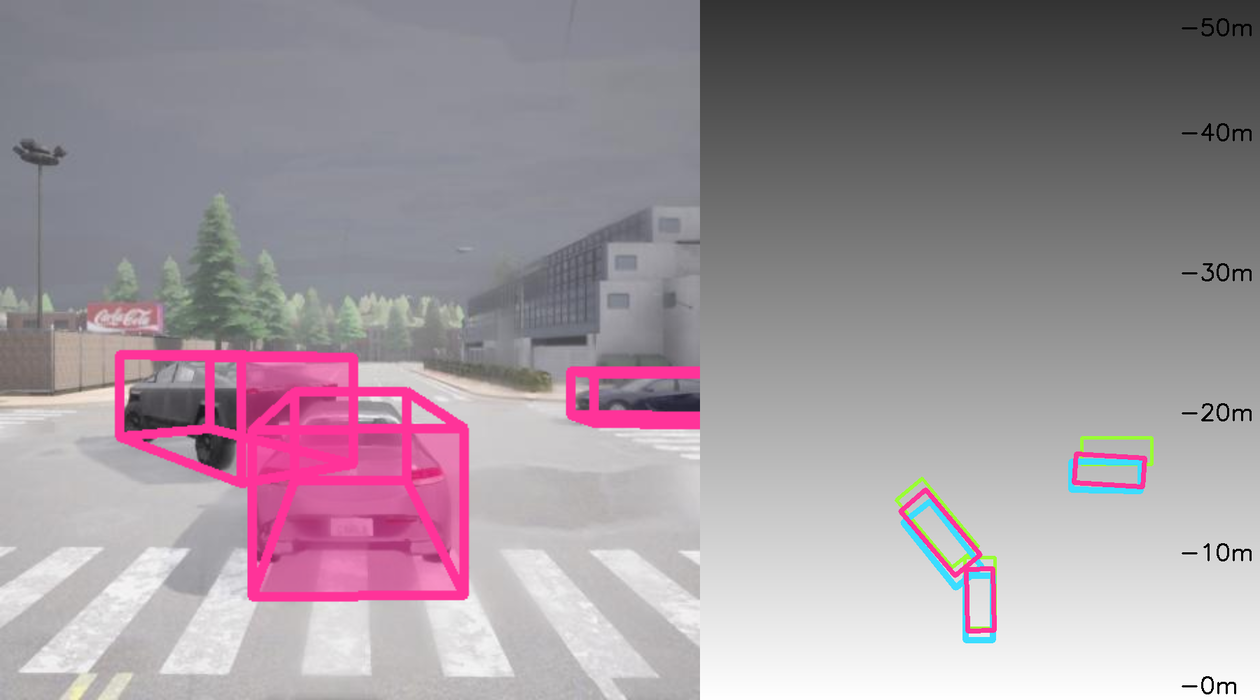}
            \end{subfigure}
            \begin{subfigure}{0.4\linewidth}
                \includegraphics[width=\linewidth]{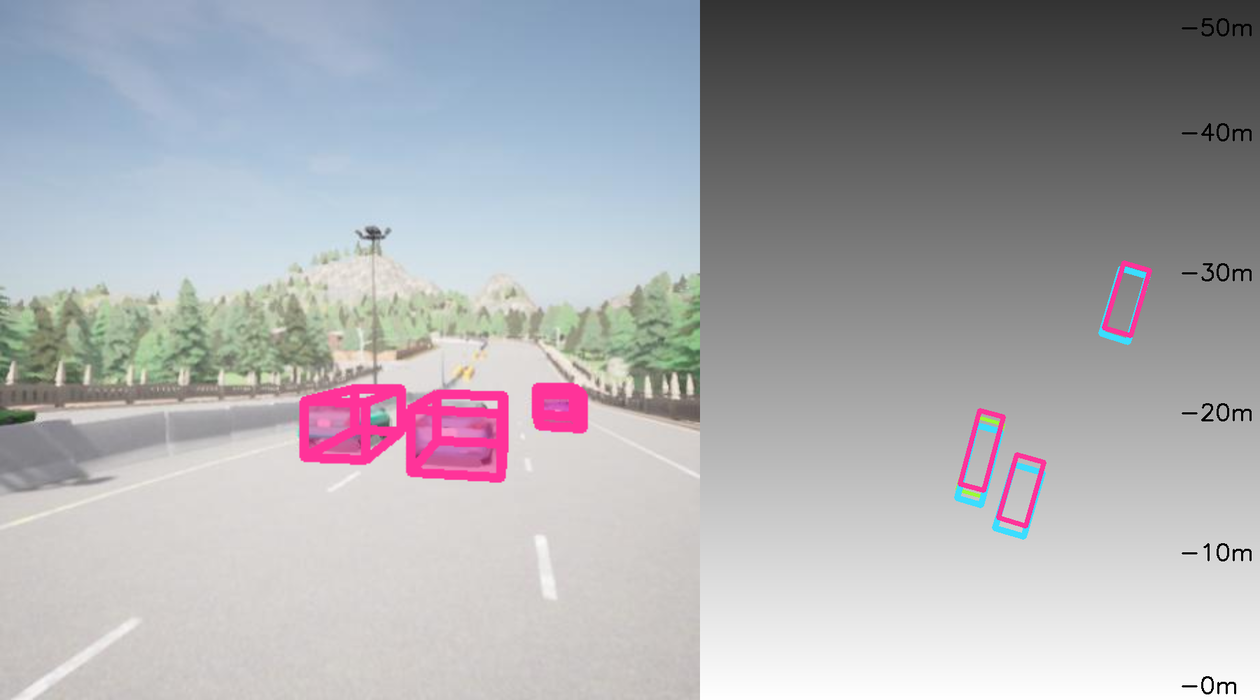}
            \end{subfigure}
            \begin{subfigure}{0.4\linewidth}
                \includegraphics[width=\linewidth]{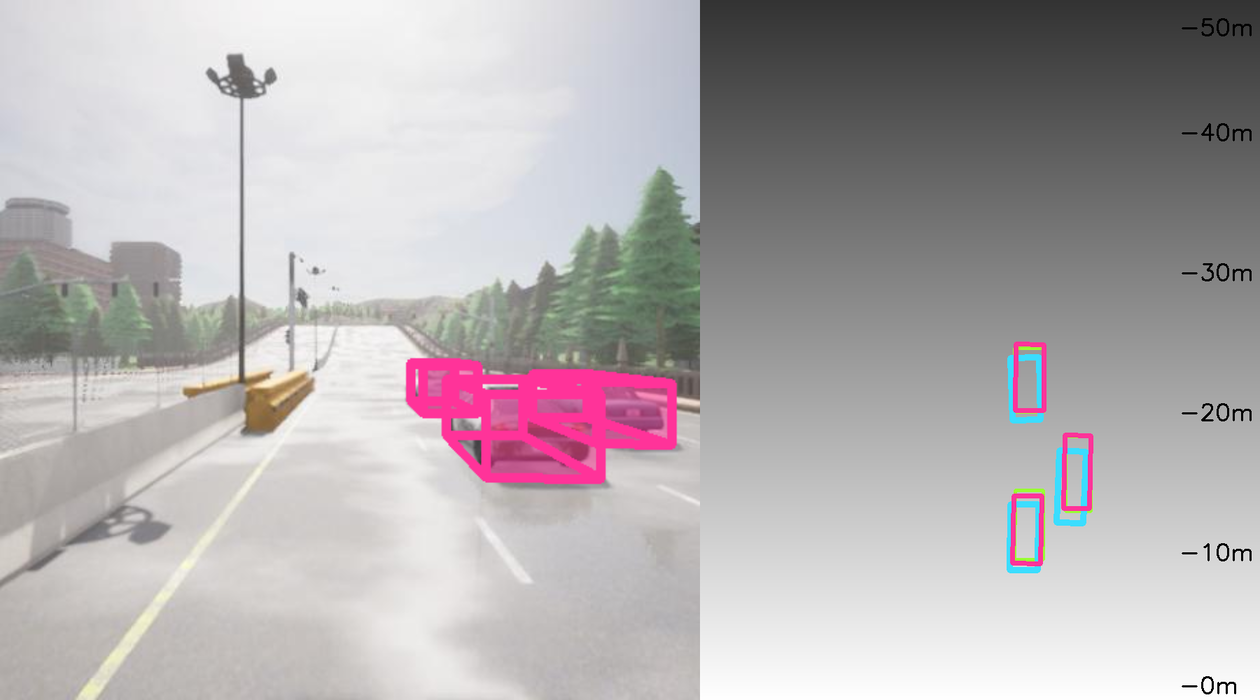}
            \end{subfigure}
            \begin{subfigure}{0.4\linewidth}
                \includegraphics[width=\linewidth]{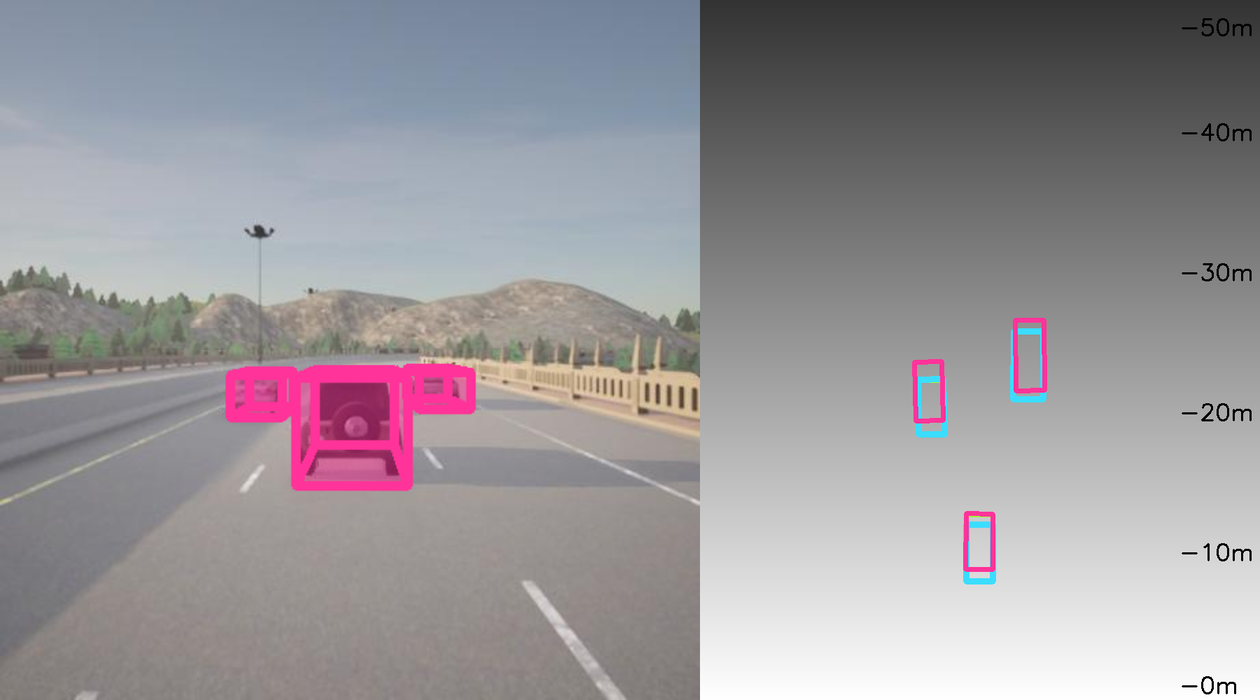}
            \end{subfigure}
            \begin{subfigure}{0.4\linewidth}
                \includegraphics[width=\linewidth]{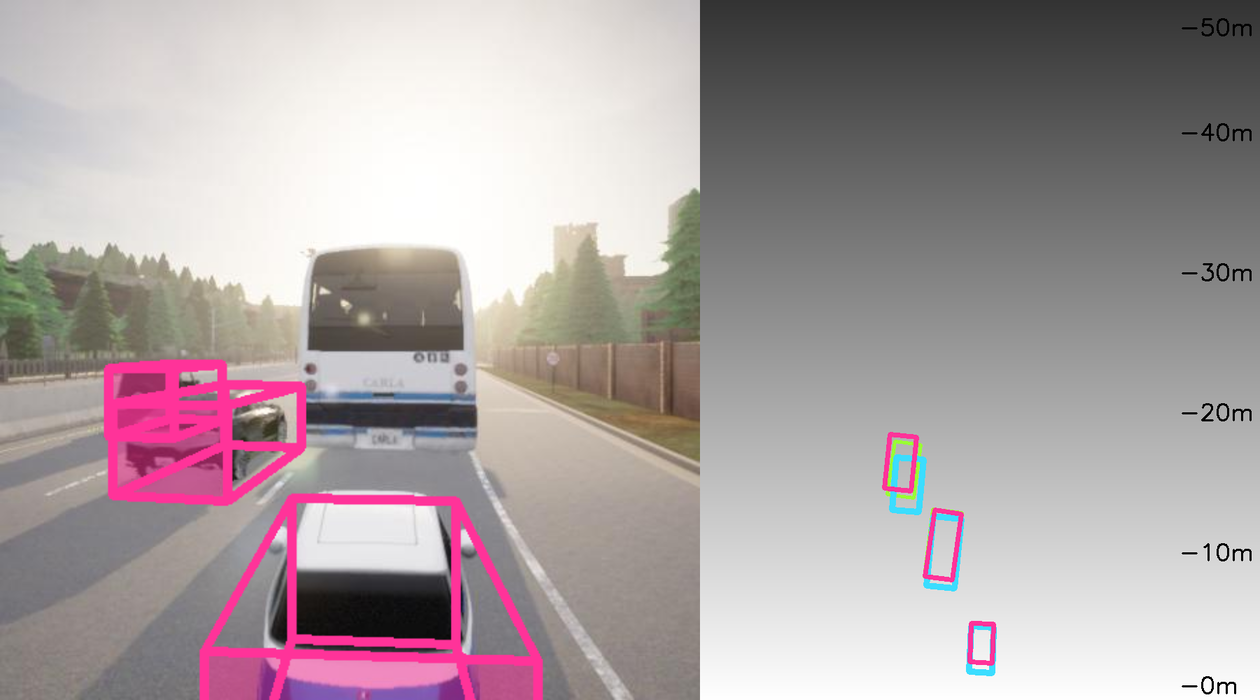}
            \end{subfigure}
            \begin{subfigure}{0.4\linewidth}
                \includegraphics[width=\linewidth]{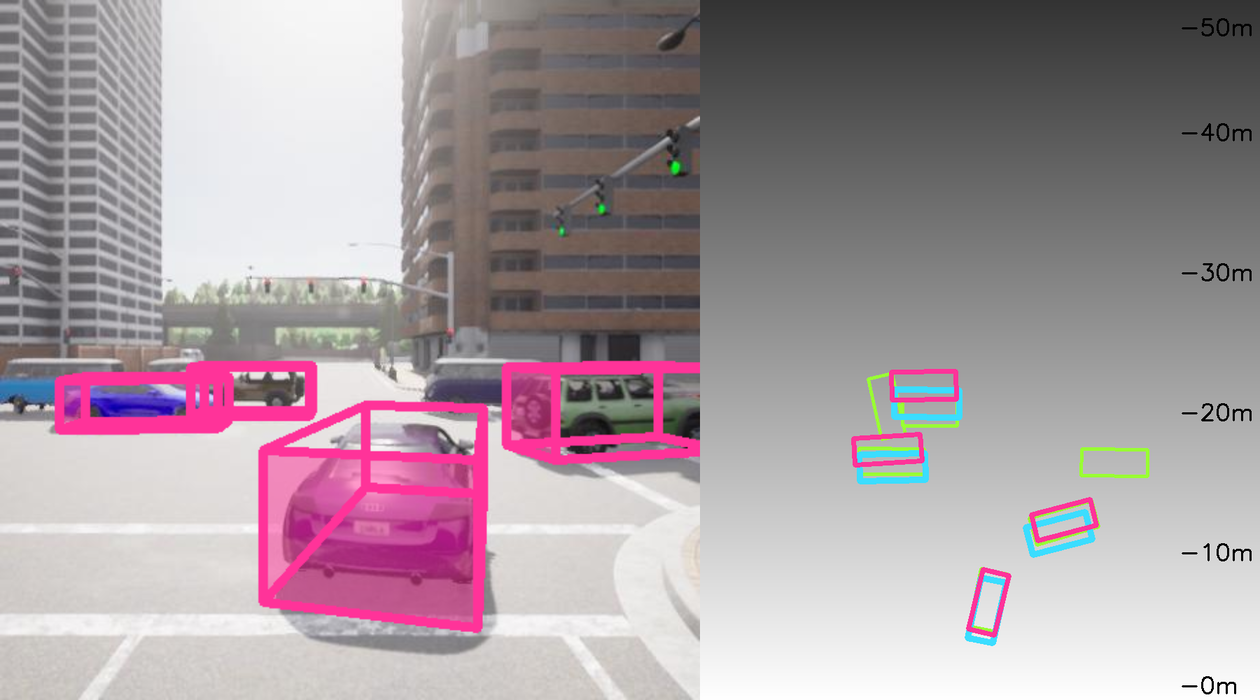}
            \end{subfigure}
            \begin{subfigure}{0.4\linewidth}
                \includegraphics[width=\linewidth]{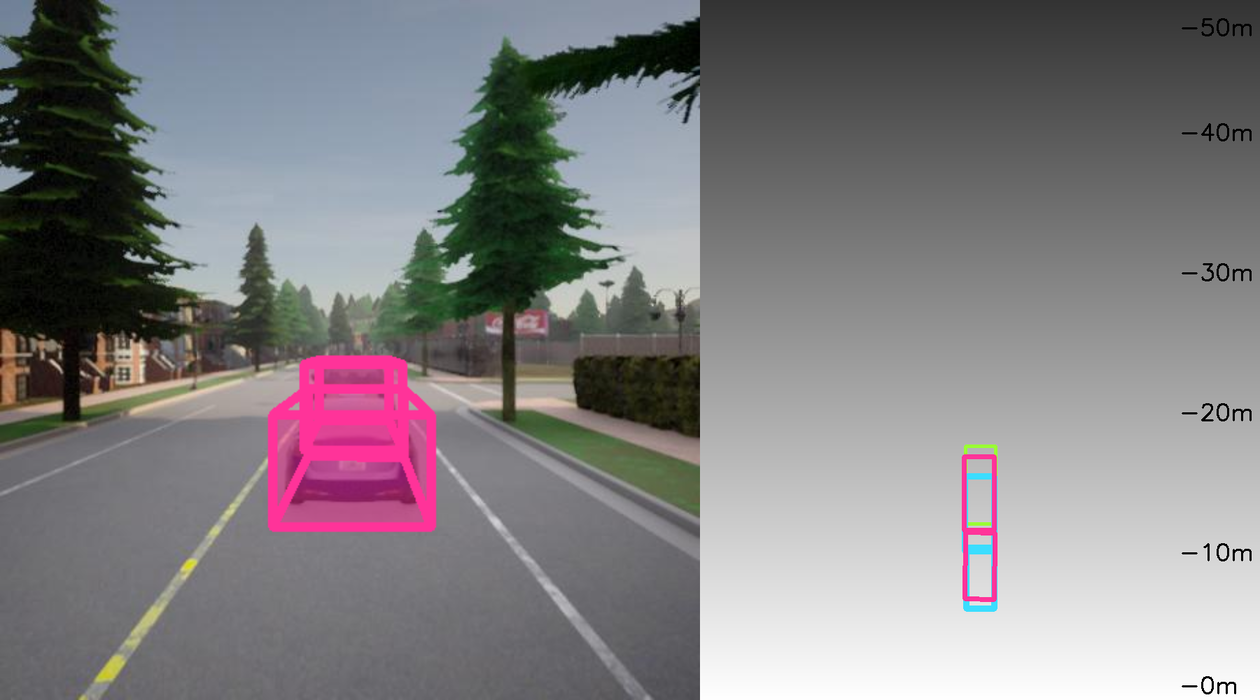}
            \end{subfigure}
            \caption{\textbf{\carla \val Qualitative Results}. 
            \charmer detects objects more accurately than \gupNet\cite{lu2021geometry}, making \charmer more robust to camera height changes.
            The regression-based baseline \gupNet mostly underestimates the depth which qualitatively justifies the claims of \cref{theorem:2}. 
            All methods are trained on \carla images at car height $\egoHeightChange=0m$ and evaluated on $\egoHeightChange=+0.76m$.
            [Key: {Cars} (pink) of \charmer.
            ; {Cars} (cyan) of \gupNet, and {Ground Truth} (green) in BEV.
            }
            \label{fig:qualitative_carla}
        \end{figure*}

            \noIndentHeading{\carla.}
                We now show some qualitative results of models trained on \carla \val split from car height $(\egoHeightChange=0m)$ and tested on truck height $(\egoHeightChange=+0.76m)$ in \cref{fig:qualitative_carla}. 
                We depict the predictions of \charmer in image view on the left, the predictions of \charmer, the baseline \gupNet \cite{lu2021geometry}, and GT boxes in \bev on the right. 
                In general, \charmer detects objects more accurately than \gupNet\cite{lu2021geometry}, making \charmer more robust to camera height changes.
                The regression-based baseline \gupNet mostly underestimates the depth of \threeD boxes with positive ego height changes, which qualitatively justifies the claims of \cref{theorem:2}. 
    
\begin{figure*}[!t]
            \centering
            \begin{subfigure}{0.48\linewidth}
                \includegraphics[width=\linewidth]{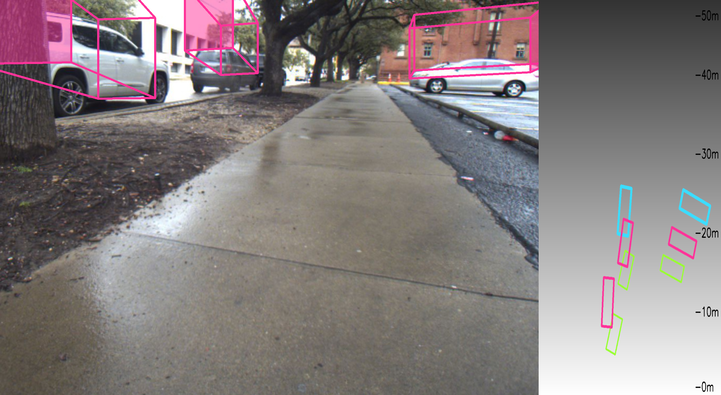}
            \end{subfigure}
            \begin{subfigure}{0.48\linewidth}
                \includegraphics[width=\linewidth]{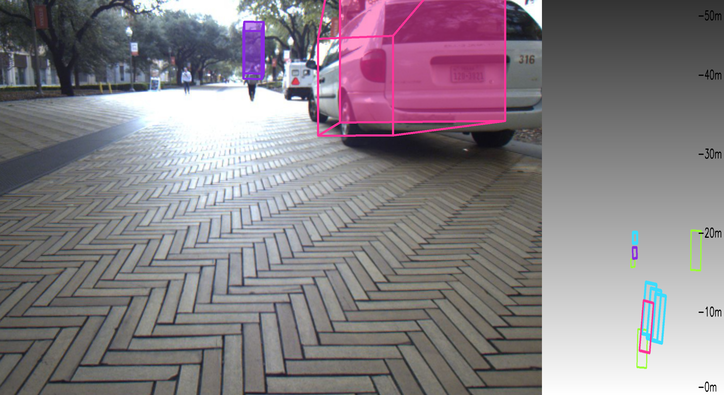}
            \end{subfigure}
            \begin{subfigure}{0.48\linewidth}
                \includegraphics[width=\linewidth]{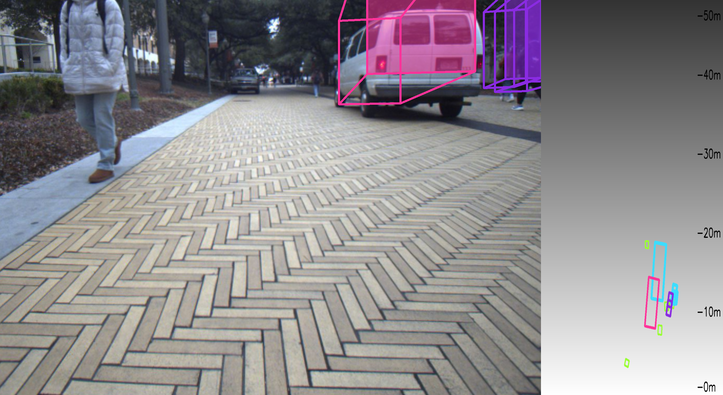}
            \end{subfigure}
            \begin{subfigure}{0.48\linewidth}
                \includegraphics[width=\linewidth]{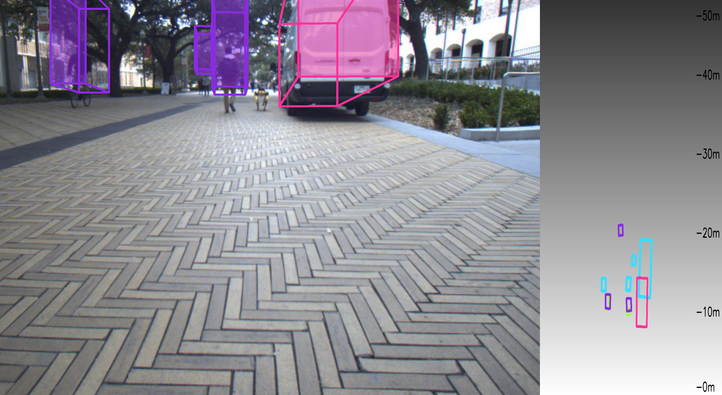}
            \end{subfigure}
            \begin{subfigure}{0.48\linewidth}
                \includegraphics[width=\linewidth]{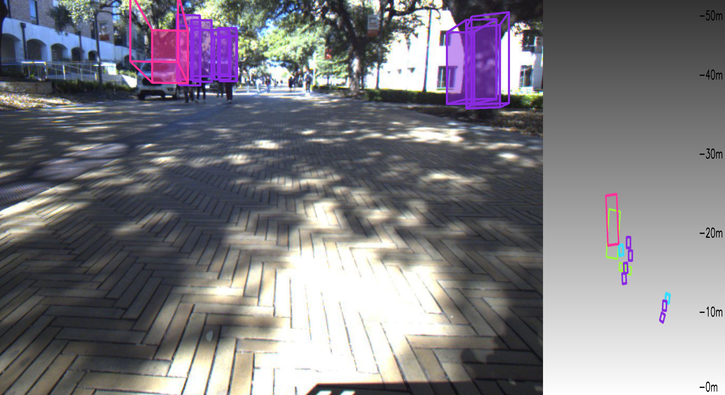}
            \end{subfigure}
            \begin{subfigure}{0.48\linewidth}
                \includegraphics[width=\linewidth]{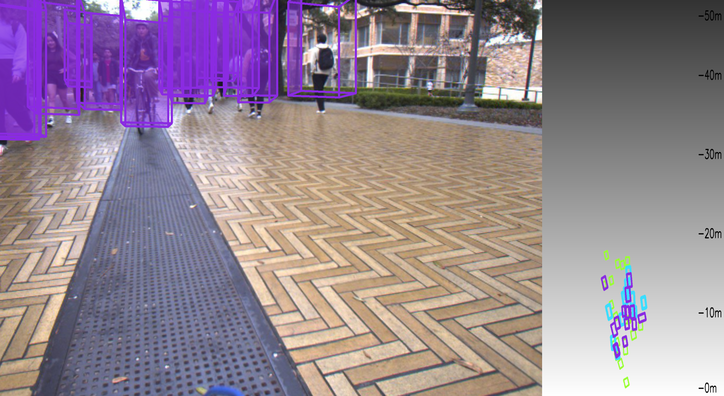}
            \end{subfigure}
            \begin{subfigure}{0.48\linewidth}
                \includegraphics[width=\linewidth]{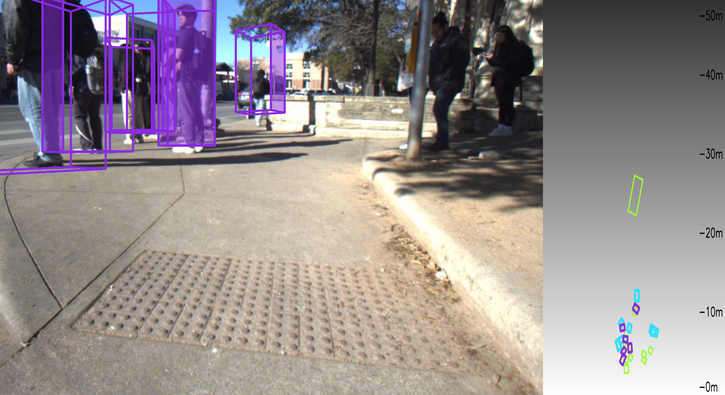}
            \end{subfigure}
            \begin{subfigure}{0.48\linewidth}
                \includegraphics[width=\linewidth]{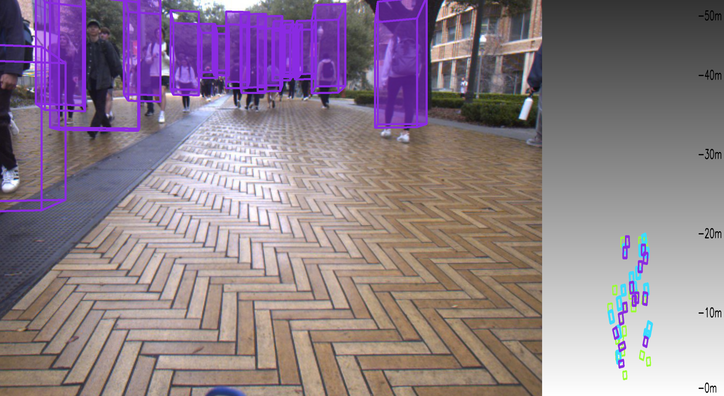}
            \end{subfigure}
            \caption[\coda \val Qualitative Results]
            {
            \textbf{\coda \val Qualitative Results}. 
            \charmer detects objects more accurately than \gupNet\cite{lu2021geometry}, making \charmer more robust to camera height changes.
            All methods are trained on \nuscenes dataset and evaluated on \coda dataset.
            [Key: {Cars} (pink) and {Pedestrian} (violet) of \charmer.
            ; {all classes} (cyan) of \gupNet, and {Ground Truth} (green) in BEV.
            }
            \label{fig:qualitative_coda}
        \end{figure*}

            \noIndentHeading{\coda.}
                We now show some qualitative results of models trained on \coda \val split in \cref{fig:qualitative_coda}.
                As before, we depict the predictions of \charmer in image view image view on the left, the predictions of \charmer, the baseline \gupNet \cite{lu2021geometry}, and GT boxes in \bev on the right. 
                In general, \charmer detects objects more accurately than the baseline \gupNet\cite{lu2021geometry}, making \charmer more robust to camera height changes.
                Also, considerably less number of boxes are detected in the cross-dataset evaluation \thatIs{} on \coda \val. 
                We believe this happens because of the domain shift.

